\RequirePackage[l2tabu, orthodox]{nag}

\documentclass[12pt,phd,a4paper,oneside]{ucl_thesis}

\usepackage{blindtext}

\usepackage{emptypage}

\usepackage{graphicx}

\usepackage{float}

\usepackage{amsmath}

\usepackage{gensymb}
\usepackage{textcomp}

\usepackage{setspace}
\usepackage{multirow}

\usepackage{bibentry}

\usepackage[format=hang,font=small,labelfont=bf]{caption}

\usepackage{etoolbox}

\usepackage{microtype}

\usepackage[numbers,sort&compress]{natbib}
\usepackage[dvipsnames]{xcolor}

\usepackage{pdfpages}
\usepackage{pax}

\usepackage[]{amsfonts}
\usepackage{amsmath}
\usepackage{nicefrac}       %

\usepackage{subfig}
\usepackage[justification=justified]{caption}

\usepackage{todonotes}	
\usepackage{graphicx}
\usepackage{coseoul}
\usepackage{titletoc}
\usepackage{tikz}
\usetikzlibrary{calc,patterns,decorations.pathreplacing,shapes.multipart}
\usepackage[page,header]{appendix}

\graphicspath{{figures/}}

\usepackage{algorithm}
\usepackage{algpseudocode}
\usepackage{makecell}
\usepackage{multirow}

\date{June 14, 2023}

\newcommand{\smoothnesstitle}{The importance of model smoothness in deep learning}

\newcommand{\specialcell}[2][c]{%
  \begin{tabular}[#1]{@{}c@{}}#2\end{tabular}}

\usepackage{amsmath, latexsym}
\usepackage{amsfonts}
\usepackage{mathtools}
\usepackage{ntheorem}
\usepackage{dsfont}
\usepackage{booktabs}
\usepackage{xfrac}
\usepackage{bbm}

\usepackage[export]{adjustbox}

\usepackage{bm}
\usepackage{listings}

\usepackage[unicode=true,pdfusetitle]{hyperref}

\usepackage[most]{tcolorbox}
\usepackage{xparse}
\usepackage{lipsum}
\usepackage{changepage}
\usepackage{enumitem}
\usepackage{pifont}

\usepackage{bm}
\usepackage{amssymb}
\usepackage{xifthen}

\newtheorem{theorem}{Theorem}[section]
\newtheorem{lemma}[theorem]{Lemma}
\newtheorem{remark}{Remark}[section]
\newenvironment{proof}{\paragraph{Proof:}}{\hfill$\square$}
\newtheorem{theo}{Theorem}[section]
\newtheorem{definition}{Definition}[section]

\newtheorem{proposition}{Proposition}[section]

\newcommand{\xmark}{\ding{55}}

\newtheorem{innercustomthm}{Theorem}
\newenvironment{customthm}[1]
  {\renewcommand\theinnercustomthm{#1}\innercustomthm}
  {\endinnercustomthm}

\newtheorem{innercustomcor}{Corollary}
\newenvironment{customcor}[1]
  {\renewcommand\theinnercustomcor{#1}\innercustomcor}
  {\endinnercustomcor}

\newcommand{\tph}{\tau}

\usepackage{etoolbox} %

\definecolor{538blue}{RGB}{48,162,218}
\definecolor{538red}{RGB}{252,79,48}
\definecolor{538yellow}{RGB}{229,174,56}
\definecolor{538green}{RGB}{109,144,79}
\definecolor{538purple}{RGB}{129, 15, 124}
\definecolor{538grey}{RGB}{139,139,139}
\definecolor{538magenta}{RGB}{200,100,200}

\newcommand{\catgames}{54}
\newcommand{\dqngames}{54}

\DeclareRobustCommand\legend[1]{%
  \tikz\draw[#1] (0,0) (0,\the\dimexpr\fontdimen22\textfont2\relax)
  -- (8pt,\the\dimexpr\fontdimen22\textfont2\relax);%
}

\newcommand{\qedwhite}{\hfill \ensuremath{\Box}}

\newcommand\cut[1]{}

\usepackage[percent]{overpic}
\newcommand{\norm}[1]{\left\lVert#1\right\rVert}
\newcommand{\rainbow}{Rainbow}

\newcommand{\decorate}[2][]{%
    \ifthenelse{\isempty{#1}}{#2}{%
          \ifthenelse{\equal{\detokenize{#1}}{\detokenize{sn}}}{\hat{#2}}{\bar{#2}}%
          }}

\newcommand{\weight}[2][]{\mathbf{\decorate[#1]{W}}_{#2}}

\newcommand{\lrp}{{\upsilon_{\phi}}}
\newcommand{\lrt}{{\upsilon_{\theta}}}

\newcommand{\owntag}[1]{\stepcounter{equation}   \tag{#1, \theequation} }

\newcommand{\squishlist}{
   \begin{list}{$\bullet$}
    { \setlength{\itemsep}{0pt}      \setlength{\parsep}{3pt}
      \setlength{\topsep}{3pt}       \setlength{\partopsep}{0pt}
      \setlength{\leftmargin}{1.5em} \setlength{\labelwidth}{1em}
      \setlength{\labelsep}{0.5em} } }

\newcommand{\squishlisttwo}{
   \begin{list}{$\bullet$}
    { \setlength{\itemsep}{0pt}    \setlength{\parsep}{0pt}
      \setlength{\topsep}{0pt}     \setlength{\partopsep}{0pt}
      \setlength{\leftmargin}{2em} \setlength{\labelwidth}{1.5em}
      \setlength{\labelsep}{0.5em} } }

\newcommand{\squishend}{
    \end{list}  }

\newtheorem{cor}{Corollary}[section]

\DeclareMathOperator{\Tr}{Tr}

\newcommand{\sign}{\mbox{sign}}

\newcommand{\myvec}[1]{\mathbf{#1}}

\newcommand{\myvecsym}[1]{\bm{#1}}

\newcommand{\vphi}{\myvecsym{\phi}}

\newcommand{\vpsi}{\myvecsym{\psi}}

\newcommand{\vtheta}{\myvecsym{\theta}}

\newcommand{\vxi}{\myvecsym{\xi}}

\newcommand{\vb}{\myvec{b}}

\newcommand{\vg}{\myvec{g}}
\newcommand{\vh}{\myvec{h}}

\newcommand{\vm}{\myvec{m}}

\newcommand{\vo}{\myvec{o}}

\newcommand{\vs}{\myvec{s}}
\newcommand{\vt}{\myvec{t}}
\newcommand{\vu}{\myvec{u}}
\newcommand{\vv}{\myvec{v}}

\newcommand{\vx}{\myvec{x}}

\newcommand{\vy}{\myvec{y}}

\newcommand{\vz}{\myvec{z}}

\newcommand{\vA}{\myvec{A}}

\newcommand{\vC}{\myvec{C}}

\newcommand{\vH}{\myvec{H}}
\newcommand{\vI}{\myvec{I}}
\newcommand{\vJ}{\myvec{J}}

\newcommand{\vM}{\myvec{M}}

\newcommand{\vP}{\myvec{P}}

\newcommand{\vR}{\myvec{R}}

\newcommand{\vW}{\myvec{W}}
\newcommand{\vX}{\myvec{X}}

\newcommand{\vZ}{\myvec{Z}}

\newcommand{\hatdotvtheta}{\bm{{\dot{\tilde{\theta}}}}}

\newcommand{\gamesvf}{u}

\newcommand{\jacparam}[2]{\frac{d #2}{d #1}}
\newcommand{\jactwoparam}[3]{\frac{d #3}{d #1 d #2}}
\newcommand{\evaljac}[3]{\jacparam{#1}{#2}\bigg\rvert_{#3}}
\newcommand{\evaljacgames}[3]{\evaljac{#1}{#2}{#3}}
\newcommand{\secondorderjac}[2]{\frac{d #2}{d^2 #1}}
\newcommand{\csecondorderfam}{\mathcal{C}(\secondorderjac{\vtheta}{f})} 
\newcommand{\csecondorderfamv}{\mathcal{C}(\secondorderjac{\vpsi}{\gamesvf})} 
\newcommand{\raw}[2]{\frac{d}{d #1}\left(#2\right)} 
\newcommand{\evaljacdata}[4]{\evaljac{#1}{#2(\cdot; #4)}{#3}}

\newcommand{\jacthetaf}{\jacparam{\vtheta}{f}}
\newcommand{\jacphif}{\jacparam{\vphi}{f}}

\newcommand{\jacthetag}{\jacparam{\vtheta}{g}}
\newcommand{\jacphig}{\jacparam{\vphi}{g}}
\newcommand{\nablavpsiu}{\jacparam{\vpsi}{u}}

\newcommand{\miditerationindex}{t'}

\newcommand{\diag}{\mbox{diag}}

\newcommand{\union}{\cup}

\newcommand{\be}{\begin{equation}}
\newcommand{\ee}{\end{equation}}
\newcommand{\bea}{\begin{eqnarray}}
\newcommand{\eea}{\end{eqnarray}}
\newcommand{\beaa}{\begin{eqnarray*}}
\newcommand{\eeaa}{\end{eqnarray*}}

\DeclareMathAlphabet{\mathpzc}{OT1}{pzc}{m}{n}

\newtheorem{corollary}{Corollary}[chapter]

\begin{document}

\pagenumbering{roman}

\author{Mihaela Claudia Rosca}

\title{On discretisation drift and smoothness regularisation in neural network training}
\department{Department of Computer Science}

\maketitle
\makedeclaration

\begin{abstract} %
\vspace{-1em}
The deep learning recipe of casting real-world problems as mathematical optimisation and tackling the optimisation by training deep neural networks using gradient-based optimisation has undoubtedly proven to be a fruitful one. The understanding behind \textit{why} deep learning works, however, has lagged behind its practical significance. We aim to make steps towards an improved understanding of deep learning with a focus on optimisation and model regularisation.
We start by investigating gradient descent (GD), a discrete-time algorithm at the basis of most popular deep learning optimisation algorithms. Understanding the dynamics of GD has been hindered by the presence of discretisation drift, the numerical integration error between GD and its often studied continuous-time counterpart, the negative gradient flow (NGF). 
To add to the toolkit available to study GD, we derive novel continuous-time flows that account for discretisation drift. Unlike the NGF, these new flows can be used to describe learning rate specific behaviours of GD, such as training instabilities observed in supervised learning and two-player games.
We then translate insights from continuous time into mitigation strategies for unstable GD dynamics, by constructing novel learning rate schedules and regularisers that do not require additional hyperparameters.
Like optimisation, smoothness regularisation is another pillar of deep learning's success with wide use in supervised learning and generative modelling. 
Despite their individual significance, the interactions between smoothness regularisation and optimisation have yet to be explored.
We find that smoothness regularisation affects optimisation across multiple deep learning domains, and that incorporating smoothness regularisation in reinforcement learning leads to a performance boost that can be recovered using adaptions to  optimisation methods.
We end by showing
that optimisation can also affect smoothness, as discretisation drift can act as an implicit smoothness regulariser in neural network training.
\end{abstract}

\begin{impactstatement}
\vspace{-1em}
Our work focuses on understanding and improving components of deep learning systems. As such, it provides two avenues for impact: first, by providing insights that can be useful to other researchers in the field and second, by helping downstream applications that use such systems.

Inside deep learning academic research, this work provides a set of novel tools and insights concerning two main building blocks of deep learning systems, namely optimisation and models. Our insights include novel continuous-time flows to analyse optimisation dynamics, as well as a rethinking of the effect of smoothness constraints imposed on deep learning models. We have accompanied theoretical results with experimental frameworks for analysis and validation, including approaches that can stabilise training and reduce hyperparameter sensitivity and thus computational costs.
Future theoretical research directions based on our line of work include specialising our theoretical results to specific classes of neural networks, further expanding the family of available continuous-time flows capturing optimisation dynamics, and finding new beneficial or detrimental implicit regularisation forces in deep learning optimisation.  Promising avenues for practical impact include new methods for training  deep learning models, and finding new approaches for model selection. Throughout this work, we provide insights across multiple deep learning domains, including supervised learning, two-player games, and reinforcement learning.  The work included here has been presented at multiple conferences and published in journals.

While we do not tackle specific applications directly, this work can have impact outside the research domain by improving the stability of deep learning models used in applications across multiple domains; examples include image classification and generation, as well as game play using reinforcement learning.
\end{impactstatement}

\begin{acknowledgements}
\vspace{-1em}
One of the biggest opportunities I have been provided with throughout my PhD studies was to have not one, but two incredible supervisors. My primary supervisor, Marc Deisenroth, has been an wonderful source of guidance and strength. Marc, you have raised my bar for rigour in thinking and writing, and have shown me that technical excellence does not come in spite of kindness, but due to kindness; for that, I will be eternally grateful. My second supervisor, Shakir Mohamed, has been my champion and supporter for many years. Shakir, working with you has been the privilege of a life-time; your breadth of technical knowledge is outstanding, and your careful and deliberate application of it through a meticulous choice of impactful research directions should serve as a guide to us all.

I have also been lucky to have an amazing set of collaborators throughout my PhD: Andriy Mnih, Arthur Gretton, Benoit Dherin, Chongli Qin, Claudia Clopath, David G.T. Barrett, Florin Gogianu, Lucian Busoniu, Michael Figurnov, Michael Munn, Razvan Pascanu, Theophane Weber, Tudor Berariu, and Yan Wu. Special thanks to Benoit Dherin, for many evening chats on everything to do with deep learning optimisation and regularisation; to Yan Wu, for a long term fruitful collaboration; to Theophane Weber for many chats on reinforcement learning; to Tudor Berariu and Florin Gogianu for being so much fun while doing research; and to Arthur Gretton for guidance early in my PhD. 

This work was done as part of the DeepMind-UCL PhD program, and I am grateful to those organising it and championing this program. Thanks also Frederic Besse, and to Sarah Hodkinson and Claudia Pope for their support. 

During my PhD I was lucky to write a book chapter on Implicit Generative models with  Balaji Lakshminarayanan and Shakir Mohamed, for  Kevin P. Murphy's second edition of `Machine Learning: a Probabilistic Perspective'. Beyond allowing me to share my enthusiasm for this area of machine learning, a great clarity of thinking followed the writing process, which helped fledge some of the new ideas presented in this work. I thus must thank Kevin, Balaji, and Shakir for this opportunity.

\vspace{2em}
Thanks also go to my examination committee, Ferenc Huszár and Patrick Rebeschini, for their time and an incredible and thoughtful discussion on my work.
\vspace{2em}

I also have many reasons to be grateful to those who have supported me personally during this time and beyond. I am here because my parents have prioritised my education and encouraged me throughout the way. I also had the most loving of grandmothers, and I am sure she would have loved to see me pursue this degree.

I have been fortunate to have a great set of friends, especially in the though times of coronavirus lockdowns. Niklas, you have been there for me since the times of the Imperial labs, and have always had faith in me; there are no words that can express how grateful I am. Fabio, you are a consistent, reliable, and wise friend; you are also raising my favourite humans, who are a great source of happiness for me. Many thanks also go to Ada, Dhruva, David,  George, James, Ira, Sascha, Siqi, Slemi, Victor, and many other friends, for the time spent together and the great memories.

\end{acknowledgements}

\setcounter{tocdepth}{2} 

\tableofcontents

\listoffigures

\listoftables

\pagenumbering{arabic}

\chapter{Introduction}

Why has deep learning been so successful in recent decades at solving a myriad of problems, such as scientific discovery and applications \citep{jumper2021highly,ravuri2021skilful,fawzi2022discovering}, language and image generation~\citep{ramesh2022hierarchical,brown2020language,hoffmann2022training,sauer2023stylegan}, and super level human game play in intricate games~\citep{silver2017mastering,alphazero,berner2019dota}?
A common answer to the question of why deep learning has been so successful is that it scales well with increases in amount of data and computational resources, both of which have drastically increased in recent years. 
But this answer raises further questions, since many of the reasons why deep learning scales well have yet to be fully uncovered. Particularly, we still do not know why the recipe behind deep learning---using simple gradient-based algorithms to optimise deep neural networks---is so effective on large datasets.
One valid response to this observation is to state that it might not matter \textit{why} deep learning works, but \textit{that} it does and to continue exploring its benefits on a wide range of domains, as well as continue the trend of scaling up data, architectures, and compute.
This approach works, can lead to progress, and has been widely used in other sciences. This is not the approach we take here, however. 
Throughout this thesis, we take the point of view that understanding why a system works is not only desirable, but necessary in order to ensure targeted research progress, by exploring the areas most likely to lead to an acceleration of new methods, increased resource efficiency, and safe deployment.
In taking this point of view, we follow the footsteps of a growing body of work aiming to uncover, examine, and understand deep learning systems~\citep{arora2018optimization,igr,igr_sgd,arora2018understanding,mescheder2017numerics,nagarajan2019generalization,nagarajan2017gradient,keskar2016large,dinh2017sharp,jiang2019fantastic,smith2018dont}.

There are many avenues worthy of study inside the deep learning domain, ranging from investigating why certain model architectures work well~\citep{li2018visualizing,understanding_batch_norm,brock2021characterizing,ding2022overparameterization}, to studying why certain types of generative models or learning principles outperform others~\citep{sriperumbudur2009integral,arjovsky2017wasserstein,fedus2017many}. 
Here, we choose to focus on the study of optimisation and model regularisation in the form of smoothness regularisation. We are motivated by their generality and applicability across deep learning domains from supervised learning, probabilistic modelling, and reinforcement learning; their significant impact on training performance and test set and out of distribution generalisation~\citep{hoffman2020robust,miyato2018spectral,dcgan}; and the opportunity for analysis due to the vast gap between their aforementioned practical performance and the understanding of their mechanisms. Unanswered questions include: What causes instability in deep learning? How can instability be mitigated? Why is there not more instability? What are the implicit regularisation effects of popular optimisers? What are the effects of smoothness regularisation? What are the interactions between model regularisation and optimisation in deep learning? Which optimisation and regularisation effects are domain specific and which transfer across deep learning domains? These are some of the questions we will tackle in this thesis.

\textbf{A motivating example}. Generative Adversarial Networks (GANs)~\citep{goodfellow2014generative} are a type of generative model that has led to many image generation breakthroughs \cite{biggan,sauer2023stylegan,karras2019style}. Many GAN variants have been proposed, some based on minimising different distributional divergences or distances \citep{nowozin2016f,arjovsky2017wasserstein,gulrajani2017improved,binkowski2018demystifying,ls_gan} while others use convergence analysis around a Nash equilibrium~\citep{mescheder2017numerics,nagarajan2017gradient}; each of these variants are  theoretically appealing and have different properties worth studying. Yet, the biggest successes in GANs have consistently come from changes to optimisation and regularisation: the choice of optimiser~\citep{dcgan}, large batch sizes~\citep{biggan}, and incorporating smoothness regularisation~\citep{miyato2018spectral} are the key ingredients of all successful GANs, while the loss functions used are often those closest to the original formulation~\citep{biggan,miyato2018spectral,zhang2019self,karras2021alias}. Similar optimisation and regularisation techniques have also allowed the expansion of GANs from continuous input domains, such as images, to discrete input domains, such as text~\citep{scratch_gan}. The GAN example showcases what opportunities are available by understanding and improving optimisation and smoothness regularisation, opportunities we aim to explore here.

\section{Optimisation in deep learning}

Machine learning is the science of casting real world problems such as prediction, generation, and action into optimisation problems.
Thus, many machine learning problems can be formulated as
\begin{align}
 \min_{\vtheta} E(\vtheta),
\end{align}
where $E$ is a loss function suitably chosen for the underlying task and $\vtheta \in \mathbb{R}^D$. For many machine learning tasks, $E$ is an expected value of an unknown data distribution $p^*(\vx)$; $E$ gets estimated  given a training dataset $\mathcal{D}$ of unbiased samples from $p^*(\vx)$: ${E(\vtheta) = \mathbb{E}_{p^*(\vx)} E(\vtheta; \vx) \approx \frac{1}{|\mathcal{D}|} \sum_{\vx_i \in \mathcal{D}} E(\vtheta; \vx_i)}$. 
A distinction is made between the model and the objective function components of $E(\vtheta; \vx)$, with the model $f(\vtheta; \vx)$ depending on the parameters $\vtheta$, and the objective function  depending only on the model's output.
In deep learning, $f$ is a neural network, $\vtheta$ are its parameters, and $E$ is non-convex. Due to the high dimensionality of $\vtheta$ or large dataset size, many optimisation algorithms are too costly or inefficient to be used in deep learning, as they scale unfavourably with parameter or dataset size. Luckily, however, we can use the compositional structure of neural networks to compute the gradient $\nabla_{\vtheta} E \in \mathbb{R}^{D}$ using the backpropagation algorithm~\citep{werbos1982applications,rumelhart1986learning,lecun1989backpropagation}. From there, we can use first-order gradient-based iterative algorithms, such as gradient descent and its variants~\citep{cauchy1847methode,robbins1951stochastic,kingma2014adam,tieleman2012rmsprop}.

Studying the behaviour of optimisation algorithms can be challenging, even for algorithms as seemingly simple as gradient descent. Typical questions of study include: Will the algorithm converge to a local minima? Does the algorithm get stuck in saddle points?  When does the algorithm fail to converge and when it does converge, how quickly does it converge? The complexity of these issues gets compounded when studying the optimisation of deep neural networks with millions to billions of parameters that do not satisfy commonly used assumptions such as convexity; this often creates a discrepancy between what theory analyses and what is used in practice.
Fundamentally, the deep learning optimisation community is interested in answering the questions: \textit{Why do first-order optimisers work so well in training neural networks}
and \textit{How can we improve optimisation in deep learning?}
Progress in this area has been made recently both through theoretical and empirical means of studying neural network optimisation dynamics~\citep{du2019gradient,zou2020gradient,du2017gradient,cohen2021gradient,keskar2016large,lewkowycz2020large}.
Novel insights challenge commonly held assumptions, such as the belief that local minima and saddle points are a major challenge for neural network optimisation due to high parameter dimensionality~\citep{du2019gradient,zou2020gradient,du2017gradient,elkabetz2021continuous}.
There has also been a growing body of work showing the importance of the Hessian $\nabla_{\vtheta}^2 E$ in supervised learning optimisation, and specifically on the connection between the learning rate and the largest Hessian eigenvalue and observed instabilities in training~\citep{cohen2021gradient,keskar2016large,lewkowycz2020large}; we will examine the importance of the Hessian in a new light in Chapter~\ref{ch:pf}.

Many analyses of discrete-time methods, such as gradient descent, take a continuous-time approach~\citep{glendinning1994stability,saxe2013exact,nagarajan2017gradient,lampinen2018analytic,arora2018optimization,advani2020high,elkabetz2021continuous,vardi2021implicit,franca2020,balduzzi2018mechanics} as continuous-time proofs tend to be easier to construct~\citep{may1976simple,elkabetz2021continuous}. 
The downside of taking a continuous-time and not discrete-time approach to optimisation analysis is that one does not directly analyse the update of interest; the discrepancy between the discrete-time trajectory and its continuous counterpart, which we will call discretisation drift, can lead to conclusions that do not transfer from continuous-time analysis to discrete-time algorithms~\citep{yaida2018fluctuation,liu2021noise}. 
 Discretisation drift is still not greatly understood and has only recently come into attention in deep learning~\citep{symmetry,igr,igr_sgd}.
The incorporation of techniques from the numerical integration community that construct continuous-time flows accounting for discretisation drift
 has shed light on the implicit regularisation effects of gradient descent and highlighted the importance of the learning rate in generalisation~\citep{igr,igr_sgd}. We will use the same numerical integration techniques in Chapter~\ref{ch:pf}, where we find a new continuous-time flow that, unlike existing flows, can capture instabilities induced by gradient descent. 
 We then use insights from our continuous-time analysis to devise an automatic learning rate schedule that can trade-off training stability and generalisation performance.

\begin{figure}[t]
\begin{subfloat}[Single objective.]{
\includegraphics[width=0.45\columnwidth]{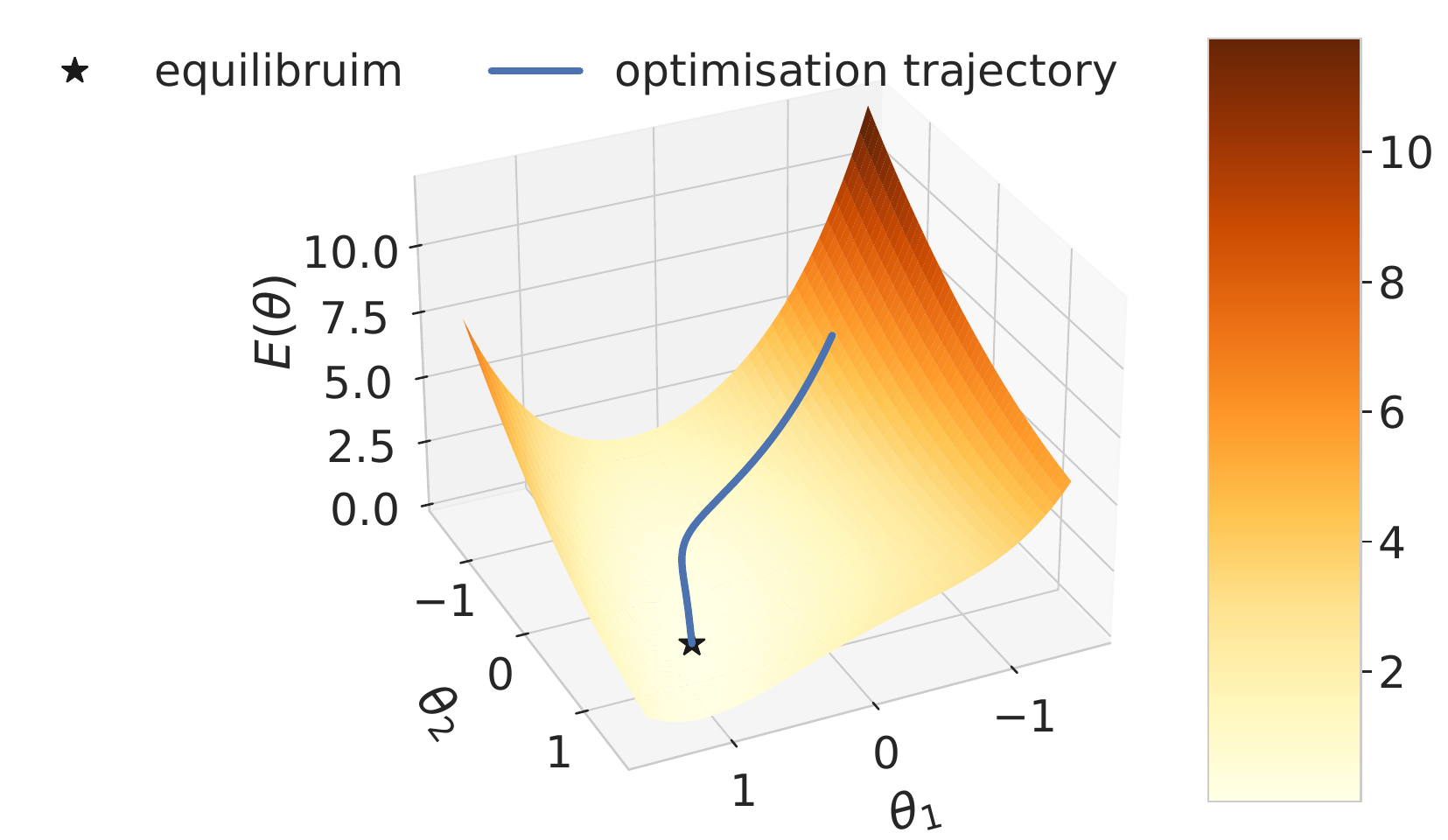}
\label{fig:single_objective_intro}
}
\end{subfloat}
\begin{subfloat}[Zero-sum games.]{
\includegraphics[width=0.45\columnwidth]{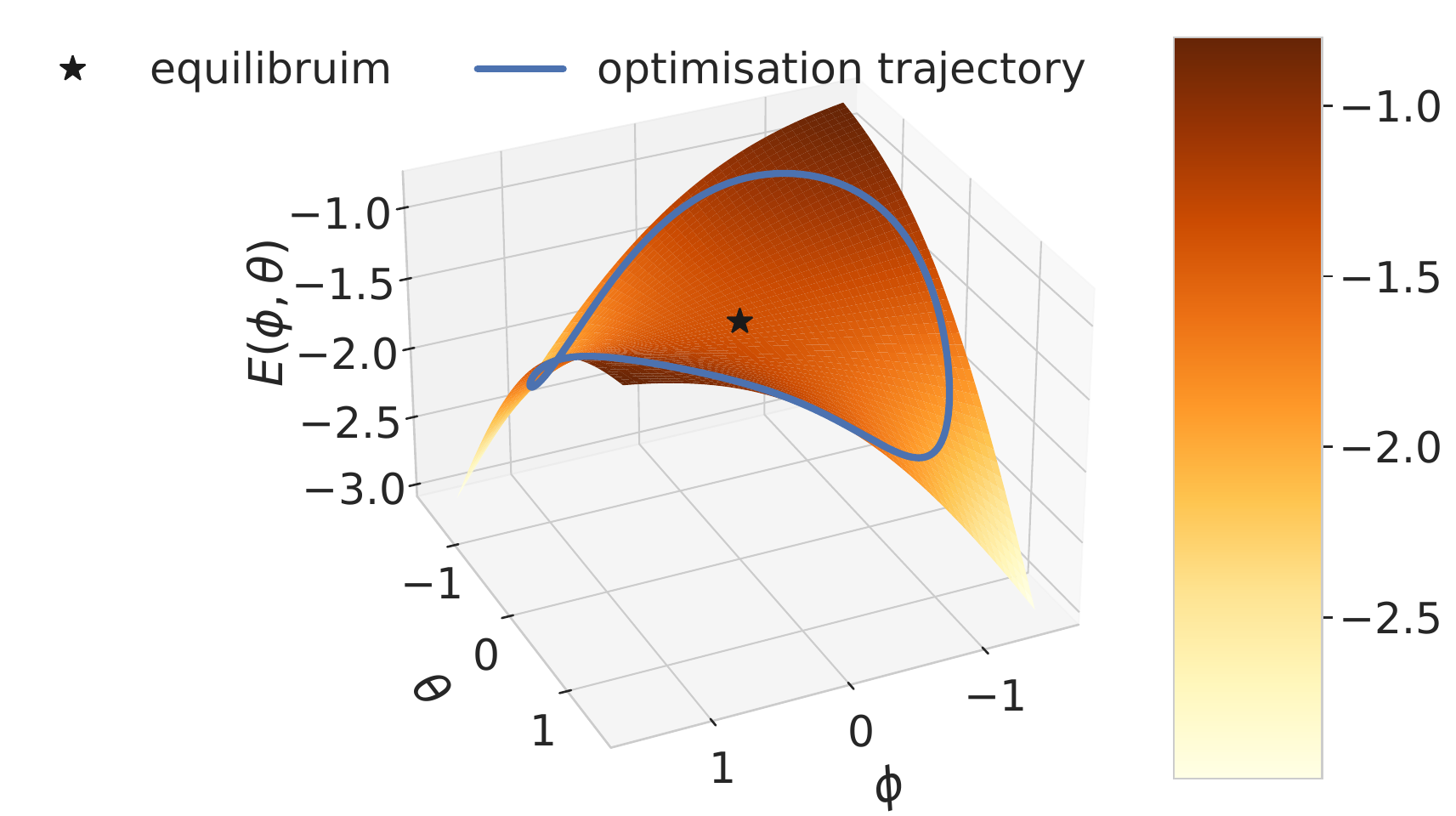}
\label{fig:two_player_games_intro}
}
\end{subfloat}
\caption[Overview of optimisation in deep learning.]{\textbf{Overview of optimisation in deep learning.} Many optimisation problems in deep learning can be seen as either single-objective problems, where all parameters are updated to minimise the same loss, or multi objective problems where where different sets of parameters are updated to minimise different objectives; when the number of objectives is two, these are called two-player games. A special case of two-player games are adversarial games such as zero-sum games, where the objective of one player is to minimise a function $E$, while the objective of the other player is to maximise the same function. This adversarial structure can lead to challenges in optimisation, with approaches that lead to converge to local minima in single objective optimisation \subref{fig:single_objective_intro} resulting in cyclic trajectories in zero sum games \subref{fig:two_player_games_intro}.}
\label{fig:optimisation_introduction}
\end{figure}

While many studies focus on optimisation dynamics in the single-objective case, such as supervised learning, strides have also been made towards understanding optimisation in two-player games~\citep{mescheder2017numerics,balduzzi2018mechanics,competitive_sgd,nagarajan2017gradient,odegan}. Unlike in the single-objective setting, in two-player games not all parameters minimise the same objective, as each player has its own set of parameters, here denoted by $\vphi$ and $\vtheta$, and its own objective:
\begin{align}
 &\min_{\vphi} E_{\vphi}(\vphi, \vtheta) \\
 &\min_{\vtheta} E_{\vtheta}(\vphi, \vtheta).
\end{align}  
Two-player games can be adversarial, where the two players have opposing objectives, and even zero sum, where one player's gain is another's loss: ${E_{\vphi}(\vtheta, \vphi) = - E_{\vtheta}(\vtheta, \vphi)}$. 
We visualise a simple two-player zero-sum game in Figure~\ref{fig:two_player_games_intro}, showing that adversarial dynamics can make it challenging to reach a local equilibrium compared to a single-objective counterpart, shown in Figure~\ref{fig:single_objective_intro}.

Interest in the analysis of two-player game optimisation has been fuelled by interest in GANs \citep{goodfellow2014generative}, a generative model trained via an adversarial two player game.
While GANs have been a successful generative model, they have a reputation of being notoriously hard to train, and  this has served as a motivation for finding approaches to stabilise and improve the dynamics of adversarial two-player games. 
Analysing the behaviour and convergence of two-player games has led to progress in understanding the sources of instability and divergence in games and has led to many a regulariser that can stabilise training~\citep{odegan,mescheder2018training,nagarajan2017gradient,balduzzi2018mechanics,wang2019solving,mazumdar2019finding}. We add to this body of work by further investigating sources of instability in two-player games, coming from discretisation drift rather than the original continuous-time system, in Chapter~\ref{ch:dd_gans}. By modelling discretisation drift, we devise explicit regularisers that stabilise training without requiring any additional hyperparameter sweep.

\section{Smoothness w.r.t. inputs}

Optimisation provides an approach to approximate an unknown decision surface required for prediction, generation, or action from an available dataset. That is, given an optimal predictor $f^*$, optimisation provides a recipe to find a parametrised function $f(\cdot;\vtheta)$ close to $f^*$ on the training dataset. But there are often many such optimisation solutions, each with different properties outside the training data.
How useful this approximation is in the underlying application depends on how well it performs on \textit{new data}. 
This is an essential desideratum of machine learning algorithms, and it is relevant for generalisation, continual learning \citep{parisi2019continual}, and transfer learning~\citep{zhuang2020comprehensive}.   Since we know from the no-free-lunch theorem~\citep{shalev2014understanding} that no machine learning algorithm will perform better than all other algorithms for any data distribution and prediction problem, to provide guarantees beyond the training data requires making assumptions about the data distribution, the structure of $f^*$, or both.
One approach to encode beliefs about $f^*$ into $f(\cdot;{\vtheta})$ is through what is often referred to in machine learning as an \textit{inductive bias}~\citep{mitchell1980need}\footnote{When we talk about smoothness with respect to inputs, we often write the predictor as a function of data, given parameters. When discussing optimisation and the focus is on parameters, we write the reverse for clarity.}.

\begin{figure}[t]
\centering
\begin{subfloat}[Very smooth fit, \\ hurts training data fit.]{
\includegraphics[width=0.3\columnwidth]{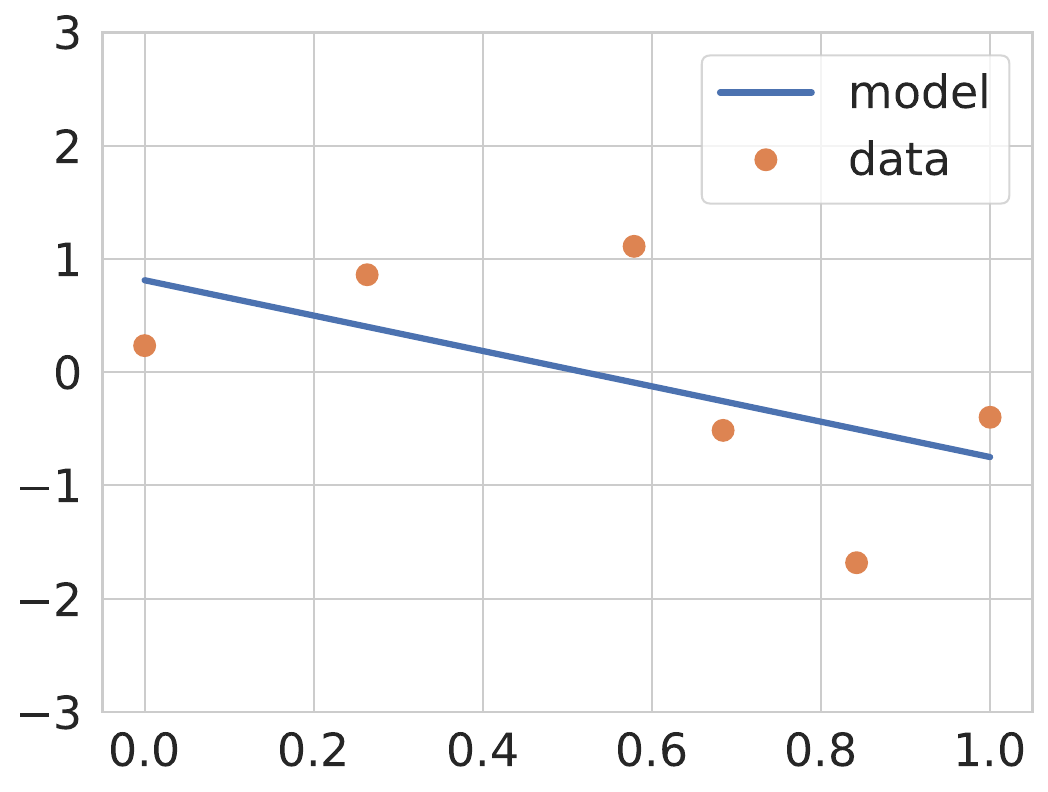}
\label{fig:intro_too_smooth}
}\end{subfloat}
\begin{subfloat}[Desired level of smoothness.]{
\includegraphics[width=0.3\columnwidth]{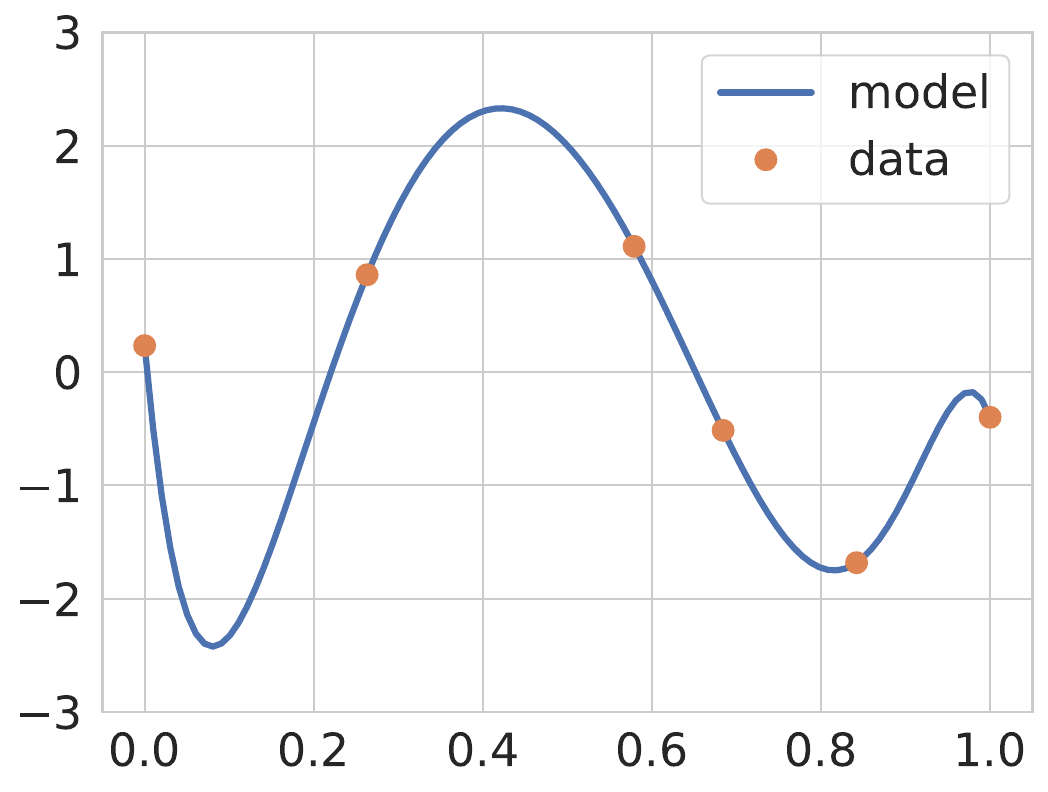}
}
\end{subfloat}
\begin{subfloat}[Too smooth fit, \\ hurts generalisation.]{
\includegraphics[width=0.3\columnwidth]{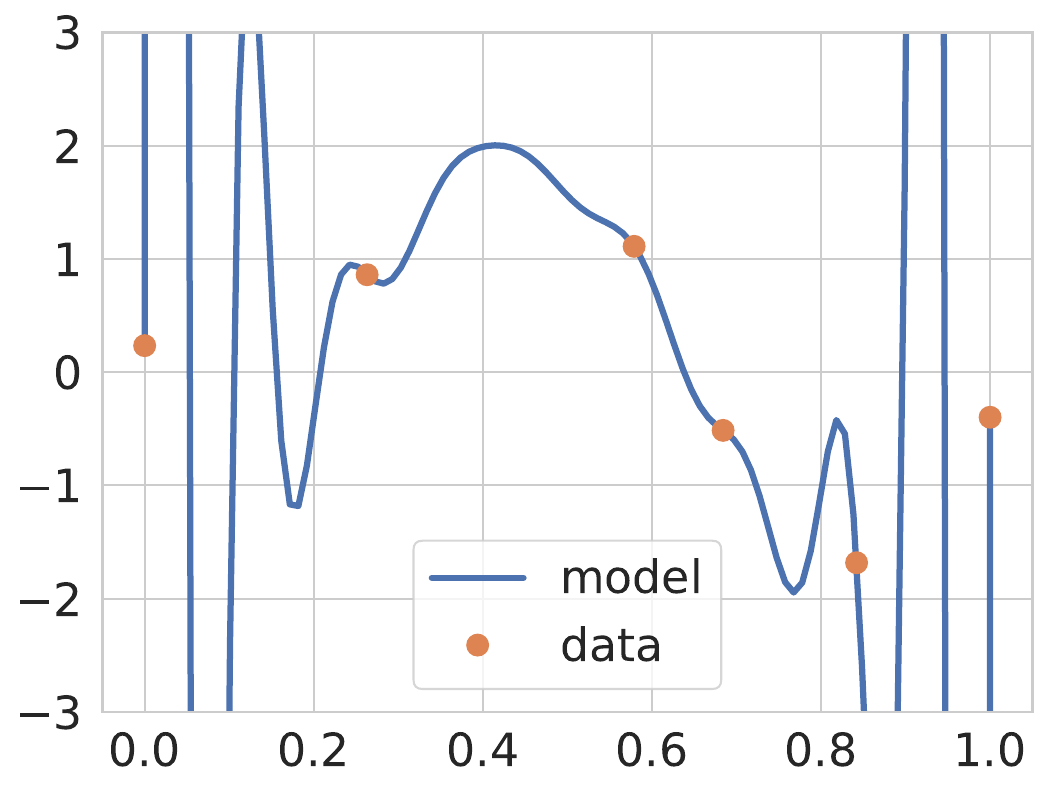}
\label{fig:intro_too_unsmooth}
}
\end{subfloat}
\caption[The importance of smoothness on the model fit.]{The importance of smoothness with respect to inputs on the model fit: too much smoothness can hurt capacity \subref{fig:intro_too_smooth}, while not enough smoothness can hurt generalisation \subref{fig:intro_too_unsmooth}.}
\label{fig:smooth_ex}
\end{figure}

In accordance with Occam's razor~\citep{mackay1991bayesian}, we often want to encode a simplicity inductive bias: we are searching for the least complex functions that can explain the data. One approach to formalising simplicity is by measuring the effect changes in function input have on the function's output. If small changes in function input do not result in large changes in function output, we call that function \textit{smooth}. A smoothness inductive bias encodes a preference for learning smooth functions: between two functions that fit the data equally well, the smoothest one should be preferred. We show an intuitive example highlighting the importance of smoothness in Figure~\ref{fig:smooth_ex}. Figure~\ref{fig:intro_too_smooth} exemplifies how too much smoothness can hurt training performance by reducing model capacity, while Figure~\ref{fig:intro_too_unsmooth} shows how the lack of smoothness can lead to overfitting.

 The formal definition of smoothness is often taken to be Lipschitz smoothness or related measures. 
 A function $f(\cdot; \vtheta): \mathcal{X} \rightarrow \mathcal{Y}$ is $K$-Lipschitz  if
 \begin{align}
  \norm{f(\vx_1; \vtheta) - f(\vx_2; \vtheta)}_{\mathcal{Y}} \le K \norm{\vx_1-\vx_2}_{\mathcal{X}}  \hspace{3em} \forall \vx_1, \vx_2 \in \mathcal{X},
\end{align}
where $K$ is known as the Lipschitz constant of function $f(\cdot; \vtheta)$. Throughout this manuscript, when we refer to the smoothness of a neural network, we refer to the smoothness w.r.t. data $\vx$, and not w.r.t. parameters $\vtheta$.

For many a machine learning task smoothness with respect to inputs constitutes a reasonable prior;  this is can be easily illustrated in the image domain, where the prediction or action that ought to be taken does not change when a few pixels in the inputs change: an image of a dog still encodes a dog if a few pixels are modified. Thus, it does not come as a surprise that despite having vastly different motivations and implementations,  many existing deep learning regularisation methods target a smoothness inductive bias, including $L_2$ regularisation, dropout~\citep{dropout}, early stopping. 
We will discuss in detail how these regularisation approaches are connected to smoothness in Chapter~\ref{ch:smoothness}, formalise these connections in Chapter~\ref{ch:gc}, as well as link smoothness to important phenomena in deep learning, such as double descent \cite{belkin2021fear,Belkin15849,deep_double_descent}.

As the importance of smoothness is beginning to be recognised by the deep learning community, methods that directly target smoothness regularisation have been developed~\citep{miyato2018spectral,yoshida2017spectral,hoffman2020robust}. These regularisation methods have been incorporated in supervised learning to help generalisation~\citep{hoffman2020robust,bartlett2017spectrally} and adversarial robustness~\citep{sokolic2017robust,novak2018sensitivity,cisse2017parseval}, as well as into generative models leading to improvements in performance and stability~\citep{miyato2018spectral,vahdat2020nvae,zhang2019self}. We investigate the importance of smoothness regularisation and its effects in Chapter~\ref{ch:smoothness} as well as incorporate it into a new problem domain, reinforcement learning, in Chapter~\ref{ch:rl}.

\section{Interactions between optimisation and smoothness regularisation}

We have thus far discussed optimisation and smoothness regularisation as separate pillars of a deep learning system. Optimisation---concerned with changes in \textit{parameters}---affects training performance, speed, and stability. 
Smoothness---concerned with changes in \textit{inputs}---can be a powerful inductive bias avoiding learning overly complex models.
What is often overlooked is the \textit{implicit bias optimisation has on the model class being learned}.
While neural architectures and model regularisers define the class of functions we can represent with our models, optimisation methods determine \textit{which} functions in the function class we can learn~\citep{igr}. Likewise, different model architectures are easier to learn than others, with ResNets~\citep{he2016deep} providing a prime example of an architecture that aids optimisation~\citep{li2018visualizing}, followed by model normalisation techniques such as batch normalisation~\citep{batch_norm}, which have been shown to primarily benefit optimisation~\citep{understanding_batch_norm}.

While not previously studied, we show that smoothness regularisation and  optimisation heavily interact in deep learning.
We start by showing examples of this interaction in Chapter~\ref{ch:smoothness}, where we observe that smoothness regularisation methods can interact with hyperparameter choices such as learning rates. 
 We then highlight how in some problem domains smoothness regularisation leads to increased performance by improving optimisation dynamics in Chapter~\ref{ch:rl}. In Chapter~\ref{ch:gc}, we show how the implicit regularisation induced by the discretisation drift of an optimisation method can lead to a smoothness inductive bias, and that the strengh of this inductive bias is dependent on optimsation hyperparameters such as learning rate and batch size.

Throughout this thesis, we will see examples of how optimisation can affect the learned model class and how the model smoothness can affect optimisation. While it is tempting to think of each component of a deep learning system individually, it can hinder progress and misguide our intuitions. We will use these observations to argue for a cohesive view between optimisation and regularisation in deep learning.

\newpage
\section*{Relationship to published papers}

The presented thesis is based on the following published works:
\begin{itemize}
    \setlength\itemsep{-0.1em}
    \item \textit{On a continuous-time model of gradient descent dynamics and instability in deep learning}. \textbf{Mihaela Rosca}, Yan Wu, Chongli Qin, Benoit Dherin. Transactions of Machine Learning Research, 2023.
    \item \textit{Discretization Drift in Two-Player Games}. \textbf{Mihaela Rosca}, Yan Wu, Benoit Dherin and David G. T. Barrett. International Conference for Machine Learning, 2021.
    \item \textit{A case for new neural network smoothness constraints}. \textbf{Mihaela Rosca}, Theophane Weber, Arthur Gretton, Shakir Mohamed. 1st I Can’t Believe It’s Not Better Workshop (ICBINB@NeurIPS 2020), Vancouver, Canada.
    \item \textit{Spectral Normalisation for Deep Reinforcement Learning: an Optimisation Perspective}. Florin Gogianu, Tudor Berariu, \textbf{Mihaela Rosca}, Claudia Clopath, Lucian Busoniu, Razvan Pascanu. International Conference for Machine Learning, 2021.
    \item \textit{Why neural networks find simple solutions: The many regularizers of geometric complexity}. Benoit Dherin, Michael Munn, \textbf{Mihaela Rosca}, David G.T. Barrett. Neural Information Processing Systems, 2022.
\end{itemize}

\noindent Other work performed as part of the PhD program that is not discussed here:
\begin{itemize}
    \setlength\itemsep{-0.1em}
    \item Chapter on implicit generative models in the second edition of `Machine Learning: a Probabilistic Perspective' by Kevin P. Murphy. \textbf{Mihaela Rosca}, Balaji Lakshminarayanan and Shakir Mohamed.
    \item \textit{Monte Carlo gradient estimation in machine learning}. Shakir Mohamed, \textbf{Mihaela Rosca}, Michael Figurnov, Andriy Mnih. Equal contributions. Journal of Machine Learning Research, 21(132), pp.1--62. 2020.
    \item Proposed and advised a project for a master student at UCL. The topic of study was the optimisation landscape in offline reinforcement learning.
    \item Co-organised the ICML 2021 workshop on `Continuous Time Perspectives in Machine Learning' \url{https://icml.cc/virtual/2022/workshop/13452}.
\end{itemize}

\subsection*{Overview of thesis}

\begin{itemize}
    \setlength\itemsep{0em}
    \item Chapter~\ref{ch:opt_intro} is an overview of the optimisation concepts required for the rest of this thesis.
    \item Chapter~\ref{ch:pf} consists of the work published in \textit{On a continuous-time model of gradient descent dynamics and instability in deep learning}~\citep{rosca2022on}. My contributions to this paper were: I started the project and investigated the idea of finding a new continuous-time flow for gradient descent; did all the proofs (except the justification of Thm~\ref{thm:principal_ode}); I wrote all the code, ran all the experiments, and did all the visualisations; I wrote the paper.
    \item Chapter~\ref{ch:dd_gans} consists of the work published in \textit{Discretization Drift in Two-Player Games}~\citep{rosca2021discretization}, together with novel insights presented in this thesis regarding the use of different learning rates for the two-players (Section~\ref{sec:bea_diff_lr_games}) as well as expanding the flow derived in Chapter~\ref{ch:pf} to games (Section~\ref{sec:bea_pf_games}). My contributions to the paper were: I initiated this project by suggesting to expand the work of~\citet{igr} to two-player games; derived the main results; did the relevant literature survey; I wrote all the code, ran all the experiments and did all the visualisations; and I wrote most of the paper.
    \item Chapter~\ref{ch:bea_stochasticity} consists of unpublished work aiming to develop a new framework for constructing modified flows and implicit regularisers in deep learning.
    \item Chapter~\ref{ch:smoothness} consists of the work published in the workshop paper \textit{A case for new neural network smoothness constraints}~\citep{rosca2020case}. This work is the result of the early reading I have done as part of the PhD program. I wrote the code and performed all the experiments for this work, and wrote the paper.
    \item Chapter~\ref{ch:rl} is based on the paper \textit{Spectral Normalisation for Deep Reinforcement Learning: an Optimisation Perspective}~\citep{gogianu2021spectral} together with the application of the insights from the paper applied to GANs (Section~\ref{sec:sn_gans_after_rl}), which is unpublished novel work I did while writing this thesis. The paper contents were rewritten and the text is mine.  My contributions to the paper were: I added to the view that Spectral Normalisation helps RL through optimisation, helped derive the optimisers that allow us to recover the performance of Spectral Normalisation, and suggested experiments and analysis such as plotting the change in Jacobian of the critic against performance as well as provide intuition and advice on the application on Spectral Normalisation and other smoothness regularisation methods; I did not contribute with any coding or experiments to the paper.
    \item Chapter~\ref{ch:gc} is based on \textit{Why neural networks find simple solutions: The many regularizers of Geometric Complexity}~\citep{dherinneural}. The paper contents were rewritten and the text is mine.  My contributions to the paper were: I provided a broader context and related work on smoothness regularisation in deep learning; I suggested, analysed, and visualised the double descent experiments that led to showing that Geometric Complexity (GC) follows the traditional U-shape curve associated with complexity measures; provided coding and debugging help throughout the project, which enabled the use of GC explicit regularisation, improved evaluation and training. I also contributed to writing and the rebuttal. I did not run experiments for this project, but provided guidance for what experiments to run. I did not derive the main results in the paper, but was involved in related discussions throughout the project.
\end{itemize}
\clearpage

\chapter{Optimisation in deep learning}
\label{ch:opt_intro}

In this chapter we describe the optimisation tools and methods we will require in the rest of this thesis.
When describing the commonly used algorithms in deep learning we often take a continuous-time approach. 
Compared to a discrete-time approach, a continuous-time approach tends to be more amenable to proof construction~\citep{elkabetz2021continuous}, having many tools available, including stability analysis, and can be used to construct conserved quantities~\citep{symmetry}. Indeed, the ease of analysis of continuous-time approaches has lead to a general uptake into continuous-time methods in deep learning recently, with use in generative models, architectures, and optimisation~\citep{chen2018continuous,chen2018neural,kidger2022neural,grathwohl2018ffjord,song2021maximum,odegan}.

\section{The direction of steepest descent}

Once we have formulated our machine learning problem into an optimisation problem
\begin{align}
 \min_{\vtheta \in \mathbb{R}^D} E(\vtheta), 
 \label{eq:opt_problem_2}
\end{align}
we have to choose an optimisation algorithm with which to solve the above problem.
 Solving the problem in Eq~\eqref{eq:opt_problem_2} in closed form is intractable for the problems we are interested in. Thus, the optimisation algorithms used are iterative, and aim to answer the question: given a current set of parameters $\vtheta$, what is a direction of descent of function $E$? 
The simplest choice of the direction is given by the negative gradient $- \nabla_{\vtheta} E(\vtheta)$, known as \textit{the direction of steepest descent}.
To see why, consider the continuous-time flow, often referred to as the \textit{gradient flow} or \textit{negative gradient flow} (NGF)
\begin{equation}
    \dot{\vtheta} = - \nabla_{\vtheta} E.
\label{eq:ngf_first}
\end{equation}
 If parameters $\vtheta$ follow the NGF in continuous-time, $E(\vtheta)$ will decrease until a stationary point $\nabla_{\vtheta} E = \mathbf{0}$ is reached, since
\begin{equation}
    \frac{d E}{d t} = \frac{d \vtheta}{d t}^T \nabla_{\vtheta} E  =  - \left(\nabla_{\vtheta} E\right)^T  \nabla_{\vtheta} E = -  || \nabla_{\vtheta} E||^2 \le 0.
    \label{eq:e_min_ngf}
\end{equation}
If $E$ is convex, following the NGF will find the global minimum.
If $E$ is not convex, following the NGF in continuous-time will reach a local minimum (a proof will be provided in Corollary~\ref{cor:ngf_local_minimum}). A point $\vtheta^* \in \mathbb{R}^D$ is a local minimum if there is a neighbourhood $\mathcal{V}$
around $\vtheta^*$ such that $\forall \vtheta' \in \mathcal{V} \hspace{1em} E(\vtheta') \le E(\vtheta^*)$; a more amenable characterisation is that local minima are stationary points $\nabla_{\vtheta} E(\vtheta^*) = \mathbf{0}$ with a positive semi-definite Hessian, i.e. $\nabla_{\vtheta}^2 E(\vtheta^*)$ has only non-negative eigenvalues. It is often convenient to also define strict local minima, i.e. local minima where eigenvalues are strictly positive. 

We have already found two insights that would follow us throughout the rest of this thesis: the ease of analysis in continuous-time, and the importance of the negative gradient. Before exploring the optimisation algorithms using these insights, we briefly turn to the question of \textit{computing} the gradient.

\vspace{-0.2em}
\subsection{Computing the gradient}
\label{sec:stochastic_opt}
To compute the gradient $\nabla_{\vtheta} E$ we have to consider the specification of the loss function $E$.
In machine learning, $E$ often takes the form of an expectation or sum of expectations; if the expectation is under the data distribution $p^*(\vx)$ and we can write $E(\vtheta) = \mathbb{E}_{p^*(\vx)} l(\vtheta; \vx)$, we have
\begin{align}
\nabla_{\vtheta} E(\vtheta) = \nabla_{\vtheta} \mathbb{E}_{p^*(\vx)} l(\vtheta; \vx) =  \mathbb{E}_{p^*(\vx)} \nabla_{\vtheta} l(\vtheta; \vx).
\label{eq:swap}
\end{align}
We can now compute an unbiased estimate of the gradient via Monte Carlo estimation:
\begin{align}
\nabla_{\vtheta} E(\vtheta) = \mathbb{E}_{p^*(\vx)}\nabla_{\vtheta} l(\vtheta; \vx) \approx \frac{1}{N}\sum_{i=1}^{N}\nabla_{\vtheta} l( \vtheta; \vx_i), \hspace{1em} \vx_i \sim p^*(\vx).
\label{eq:mc}
\end{align}
In deep learning, the gradient $\nabla_{\vtheta} l( \vtheta; \vx_i)$ is often computed via backpropagation, i.e. using the compositional structure of neural networks to compute gradients in a recursive fashion via the chain rule.
If the entire dataset is used to approximate the gradient in Eq~\eqref{eq:mc}, this corresponds to \textit{full-batch training}; if an unbiased sample from the dataset is used, the training procedure is referred to as \textit{mini-batch training}. 

Additional challenges arise for objectives of the form $E(\vtheta) = \mathbb{E}_{p(\vx; \vtheta)} l(\vx)$, as is the case for some reinforcement learning and generative modelling objectives. In such situations, the change of expectation and gradient in Eq~\eqref{eq:swap} is no longer possible.
For certain family of distributions $p(\vx; \vtheta)$, however, gradient estimators are available.
In later chapters we will use the pathwise estimator, where the random variable $X$ with distribution $p(\vx; \vtheta)$ can be written as $X = g(Z; \vtheta)$, in which case we can use the change of variable formula for probability distributions: 
\begin{align}
\mathbb{E}_{p(\vx; \vtheta)} l(\vx) = \mathbb{E}_{p(\vz)} l(g(Z; \vtheta)).
\end{align}
From here we can write
\begin{align}
\nabla_{\vtheta}\mathbb{E}_{p(\vx; \vtheta)} l(\vx) = \nabla_{\vtheta}\mathbb{E}_{p(\vz)} l(g(Z; \vtheta)) = \mathbb{E}_{p(\vz)} \nabla_{\vtheta} l(g(Z; \vtheta)),
\end{align}
which is again amenable to Monte Carlo estimation.
We refer to our general overview for more details on the pathwise estimator as well as other estimators \citep{mohamed2019monte}.

\section{Algorithms: from gradient descent to Adam}
\label{sec:opt_algo}

We have seen how following the negative gradient in continuous-time leads to convergence to a local minimum.
Implementing infinitely small updates in continuous time on our discrete computers is not feasible, however, and thus we need a discrete-time algorithm for optimisation. Discretising continuous-time flows is a heavily studied topic with its own area of research in applied mathematics, numerical integration~\citep{hairer2006geometric}.
Numerical integrators provide approaches to approximate the solution of the flow at time $h$ and initial conditions $\vtheta(0)$, which we denote as $\vtheta(h; \vtheta(0))$.

A simple discretisation method is Euler discretisation, which when applied to the NGF leads to
\begin{align}
    \vtheta(h; \vtheta(0)) &= \vtheta(0) + \int_{0}^h \dot{\vtheta}(t) d t = \vtheta(0) - \int_{0}^h \nabla_{\vtheta} E(\vtheta(t)) d t \\
     &\approx \vtheta(0) - \int_{0}^h \nabla_{\vtheta} E(\vtheta(0)) d t = \vtheta(0) - h \nabla_{\vtheta} E(\vtheta(0)). \label{eq:euler_int}
\end{align} 
Setting $\vtheta(0) = \vtheta_{t-1}$ in the above equation leads to the familiar gradient descent update:
\begin{equation}
    \vtheta_t = \vtheta_{t-1} - h \nabla_{\vtheta} E(\vtheta_{t-1}).
\end{equation}
By assuming that the vector field of the flow---the gradient $\nabla_{\vtheta} E$---does not change over the time interval $h$ we obtained a fast discrete-time algorithm. 
However, due to the approximation in Eq~\eqref{eq:euler_int}, following gradient descent is not the same as following the NGF: the numerical integration error leads to a difference between the gradient descent and NGF trajectories. Throughout this thesis we will refer to this difference in trajectories as \textit{discretisation drift} (often also known as \textit{discretisation error}).
Due to discretisation drift, there are no guarantees that gradient descent will reach a local minimum. Unlike the NGF, 
  gradient descent can diverge or converge to saddle points (stationary points that have at least one negative eigenvalue of the Hessian).

\begin{figure}[t]
\begin{subfloat}[Stability.]{
\includegraphics[width=0.35\columnwidth]{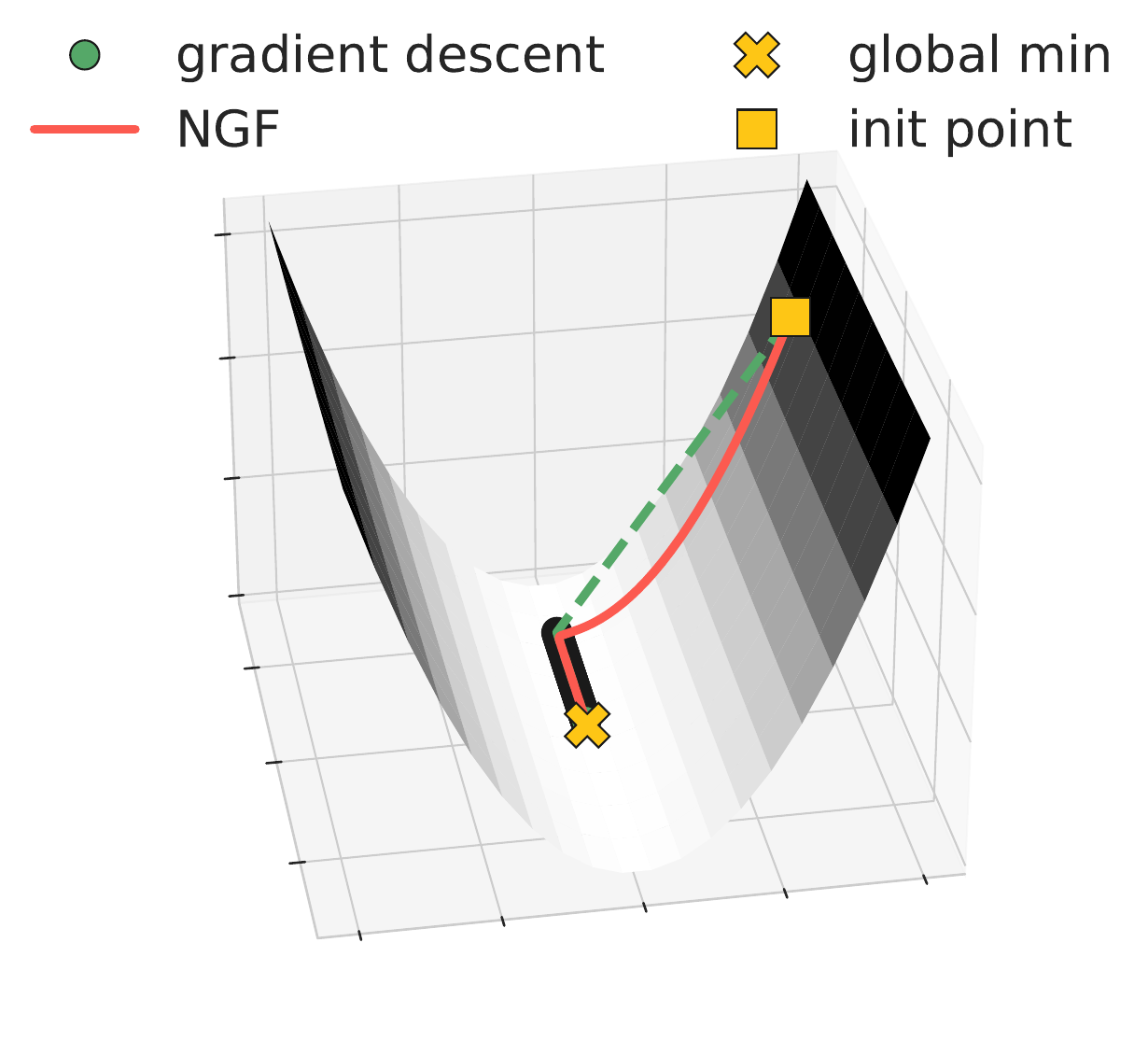}
\label{fig:dd_intro_1}
}
\end{subfloat}%
\begin{subfloat}[Instability.]{
\includegraphics[width=0.32\columnwidth]{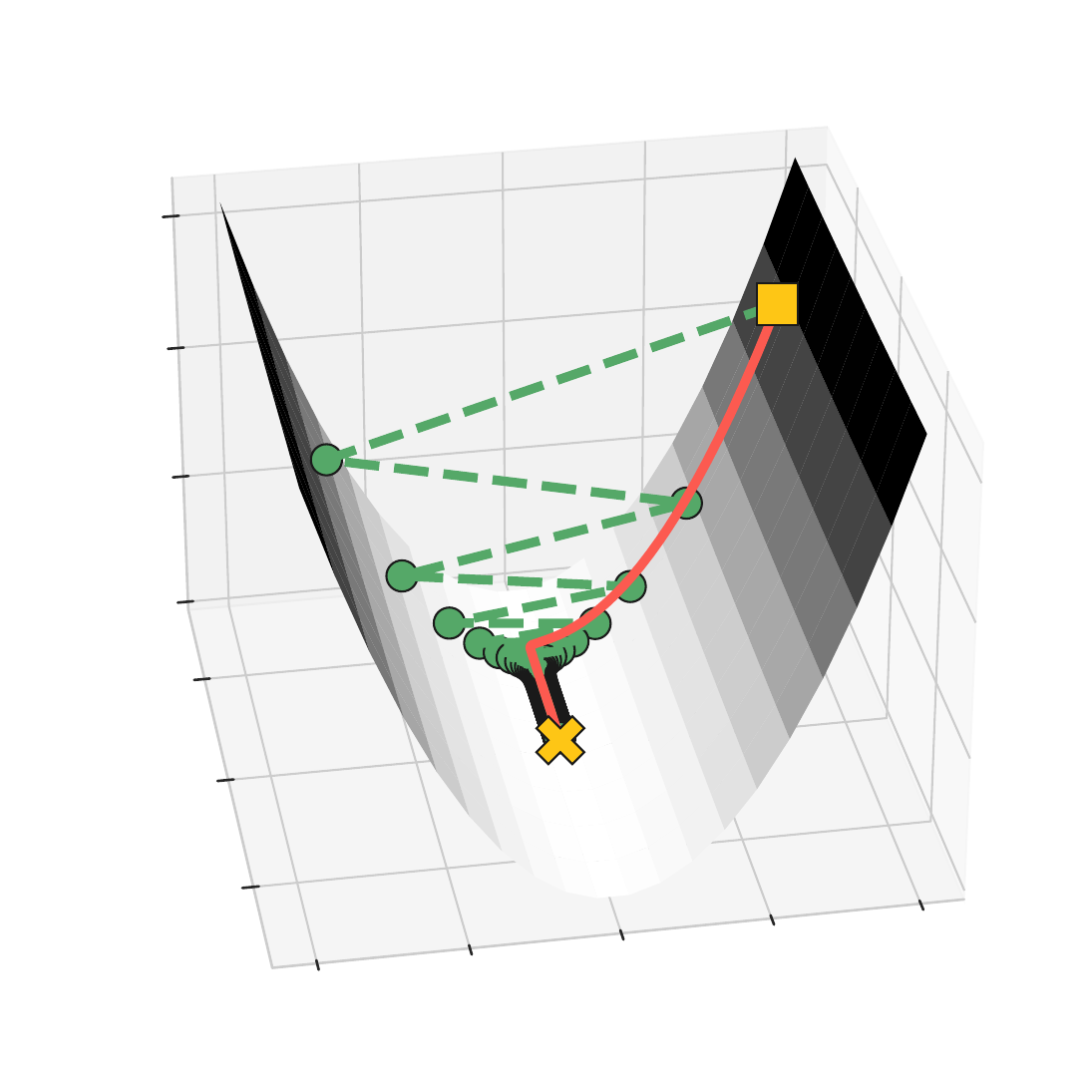}
\label{fig:dd_intro_2}
}
\end{subfloat}%
\begin{subfloat}[Divergence.]{
\includegraphics[width=0.32\columnwidth]{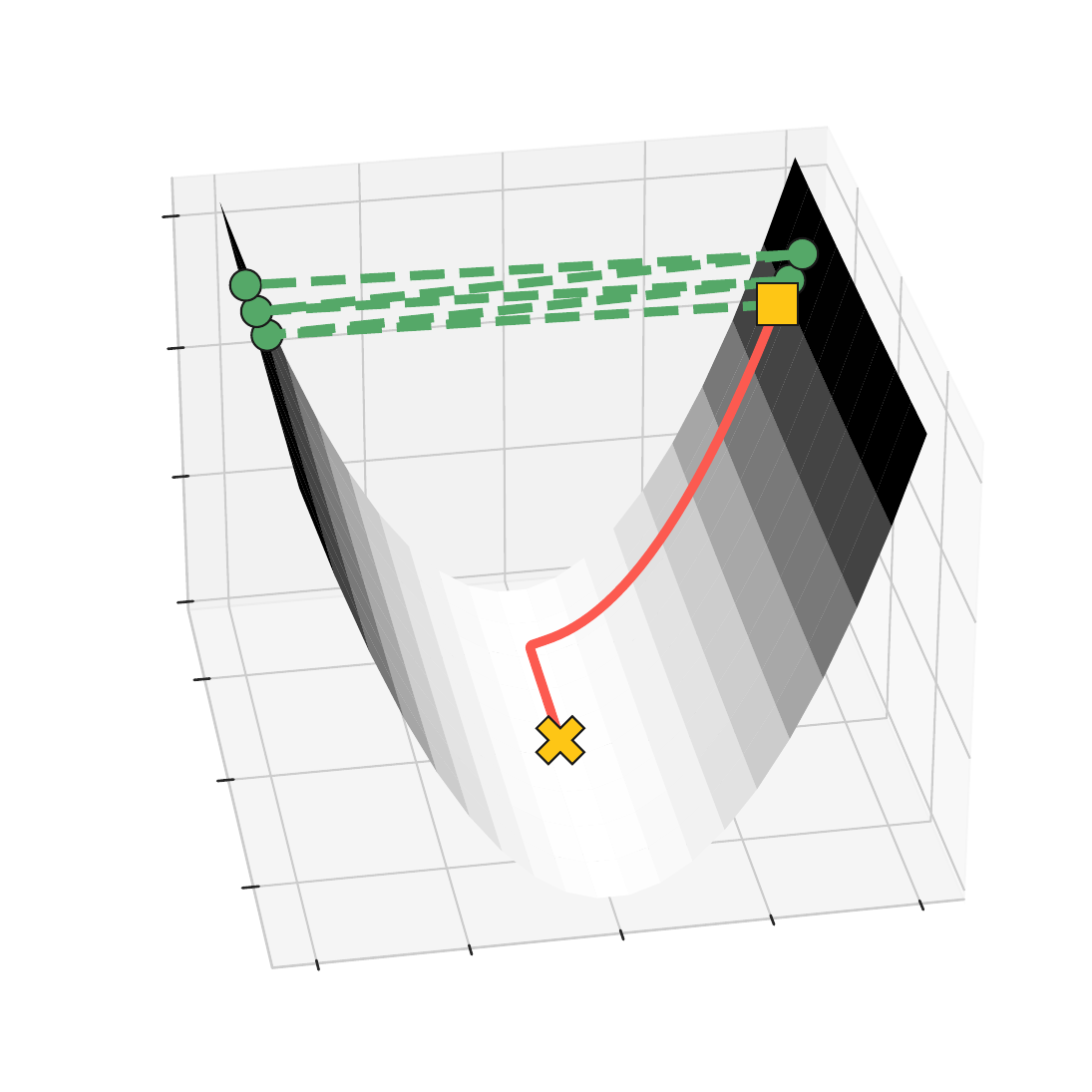}
\label{fig:dd_intro_3}
}
\end{subfloat}
\caption[The importance of the learning rate in gradient descent.]{The importance of the learning rate in gradient descent: small learning rates lead to convergence to local minima but slow to converge \subref{fig:dd_intro_1}, while larger learning rates can lead to instability \subref{fig:dd_intro_2} and divergence \subref{fig:dd_intro_3}.}
\label{fig:dd_intro}
\end{figure}

We visualise the effects of discretisation drift in Figure~\ref{fig:dd_intro}. Discretisation drift can speed up training (Figure~\ref{fig:dd_intro_1}), destabilise it (Figure~\ref{fig:dd_intro_2}) or lead to divergence  (Figure~\ref{fig:dd_intro_3}).
Importantly, when using gradient descent, the discrete counterpart of the NGF, for large $h$ the function $E$ can \textit{increase}, leading to training instabilities.
Understanding \textit{when} gradient descent leads to instabilities can be challenging, due to intricate dependencies between the learning rate and the shape of $E$; this gets compounded in the case of deep learning due to the very high dimensional nature of the parameter space. We will tackle this question later in this thesis.

Recent work has made great strides in understanding the behaviour of gradient descent in deep learning. Questions of study include whether gradient descent converges to local or global minima~\citep{du2019gradient,jacot2018neural,du2018gradient,bartlett2018gradient}, the prevalence and effect of saddle points~\citep{dauphin2014identifying,du2017gradient},
the interaction between neural architectures and optimisation~\citep{li2018visualizing,understanding_batch_norm,li2020convolutional}, the connection between learning rates and Hessian eigenvalues and instabilities~\citep{cohen2021gradient},
 the effect of the learning rate on generalisation~\citep{igr,li2019towards}.

\textbf{Rprop and RMSProp}. We have derived gradient descent as a discretisation of the NGF; we used the NGF since it decreases the value of the objective function $E$. We note, however, that only the sign of the gradient for each parameter is relevant in order to achieve a descent direction locally. To see why, consider constants $c_i >0$ and the system:
\begin{equation}
   \dot{\vtheta_i} = -  c_i \nabla_{\vtheta_i} E.
\end{equation}
This flow, like the NGF, also decreases $E$, since as $c_i >0$
\begin{equation}
    \frac{d E}{d t} = \sum_i \frac{d E}{d\vtheta_i} \frac{d \vtheta_i}{d t} =  \sum_i  \frac{d E}{d\vtheta_i} (-  c_i \nabla_{\vtheta_i} E) = -  \sum_i  c_i (\nabla_{\vtheta_i} E)^2\le 0.
    \label{eq:c_i_ode}
\end{equation}
This observation is the motivation for Rprop~\citep{rprop}, which instead of using  the gradient as the parameter update, uses only its sign: $sign(\nabla_{\vtheta_i} E) = \frac{\nabla_{\theta_i} E}{\sqrt{\left(\nabla_{\vtheta_i} E\right)^2}}$. We can think of Rprop as the discretisation of the following flow at each iteration:
\begin{equation}
   \dot{\vtheta_i} = -  \frac{1}{|\nabla_{\vtheta_i} E(\vtheta_0)|} \nabla_{\vtheta_i} E .
\end{equation}
While this can solve issues with exploding gradients and an imbalance between the magnitude of the updates of different parameters, it can be crude: Rprop cannot distinguish between areas of space where gradient are positive, but very small or areas of space with very large gradients; the update in both of these cases would be the learning rate $h$. The solution proposed to this issue was RMSprop~\citep{tieleman2012rmsprop}, which instead of normalising the gradient of each parameter by $\left(\nabla_{\vtheta_i} E\right)^2$, it normalises the gradient by a moving average of $\left(\nabla_{\vtheta_i} E\right)^2$ obtained from previous iterations. By using element-wise operations, we can write the RMSprop update as
\begin{align}
\vv_t &= \beta \vv_{t-1} + (1-\beta) \nabla_{\vtheta} E(\vtheta_{t-1})^2; \hspace{5em}
\vtheta_t = \vtheta_{t-1} - h \frac{\nabla_{\vtheta} E(\vtheta_{t-1})}{\sqrt{\vv_t}},
\end{align}
where $\beta$ is a hyperparameter often set to values in $[0.9, 0.999]$.
Since the denominator is positive, the RMSprop update has the same sign as the gradient, but a different magnitude. The use of moving averages can dampen the effect of a large gradient, which would lead to instabilities in gradient descent, but still accounts for the the magnitude of the gradient, not only its sign, unlike Rprop.
RMSprop and its variants are often used in reinforcement learning~\citep{mnih2015human,kearney2018tidbd,hessel2018rainbow}.

\textbf{Momentum}. Moving averages are an important staple of optimisation algorithms, beyond adjusting the magnitude of the local gradient, as we have seen with RMSprop. The idea behind momentum algorithms is to speed up or stabilise training by using previous iteration gradients. 

To provide intuition regarding momentum, we note that along a trajectory towards a local minimum the sign of the gradient will be the same throughout that trajectory. 
Let $\vtheta^*$  be a local minimum of $E(\vtheta)$ and $\vtheta_{t+1}$ and $\vtheta_t$ iterates along a trajectory towards $\vtheta^*$ and consider parameter index $i$ such that $sign(\vtheta_{t+1,i} - \vtheta^*_{i}) = sign(\vtheta_{t, i} - \vtheta^*_{i})$; this encodes the assumption that the trajectory goes towards a $\vtheta^*$ without escaping it in dimension $i$. But $sign(\vtheta_{t+1, i} - \vtheta^*_i) = sign(\nabla_{\vtheta} E (\vtheta_t)_i)$ as they are both the direction which minimises $E$. Thus, $sign(\nabla_{\vtheta} E (\vtheta_t)_i)=sign( \nabla_{\vtheta} E(\vtheta_{t+1})_i)$. This tells us that as long as a trajectory is not jumping over a local minimum in a direction, the sign of the gradient in that direction does not change. 
Hence, one can speed up training by taking bigger steps in that direction by accounting for the previous updates; alternatively this is often framed as an approach to speed up training in areas of low curvature (where by definition the gradient does not change substantially)~\citep{sutskever2013importance}.
This intuition suggests the following algorithm~\citep{polyak1964some}:
\begin{align}
\vtheta_t = \vtheta_{t-1} - h \sum_{i=1}^t \beta^{i-1} \nabla_{\vtheta}  E(\vtheta_{t-i}),
\end{align}
where the update is given by a geometrical average of the gradients at previous iterations, and $\beta$ is the decay rate, a hyperparameter with $\beta < 1$.
We can rewrite the momentum algorithm in its commonly known form:
\begin{align}
\vv_t &= \beta \vv_{t-1} - h \nabla_{\vtheta} E(\vtheta_{t-1}); \hspace{10em}
\vtheta_t = \vtheta_{t-1} + \vv_t .
\end{align}
Traditionally momentum has been seen as a discretisation of the second order flow
\begin{align}
{\ddot{\vtheta}} + c_v \dot{\vtheta} = - \nabla_{\vtheta} E (\vtheta),
\end{align}
from where the vector $\vv$ can be introduced:
\begin{align}
 {\dot{\vv}} = - c_v \vv - \nabla_{\vtheta} E (\vtheta); \hspace{10em}
 {\dot{\vtheta}} = \vv
\end{align}
which, when setting $c_v = \frac{1-\beta}{h}$ and Euler discretised with learning rates $h$ and $1$, leads to the discrete updates above. Momentum can also be seen as a discretisation of the first-order continuous-time flow, by combining implicit and explicit Euler discretisation \citep{symmetry}:
\begin{equation}
\dot{\vtheta} = -\frac{1}{1 - \beta} \nabla_{\vtheta} E .
\label{eq:momentum_ode}
\end{equation}
Since $\beta < 1 \implies \frac{1}{1 - \beta} > 1$, this is in line with our previous intuition of speeding up training. As with gradient descent, however, this intuition does not account for the effect of large learning rates leading to discretisation errors; while following the flow in Eq~\eqref{eq:momentum_ode} always decreases $E$, following the trajectory of the momentum optimiser instead loses that guarantee.
\begin{figure}[t]
\begin{subfloat}[$\beta=0.9$ and small $h$.]{
\includegraphics[width=0.33\columnwidth]{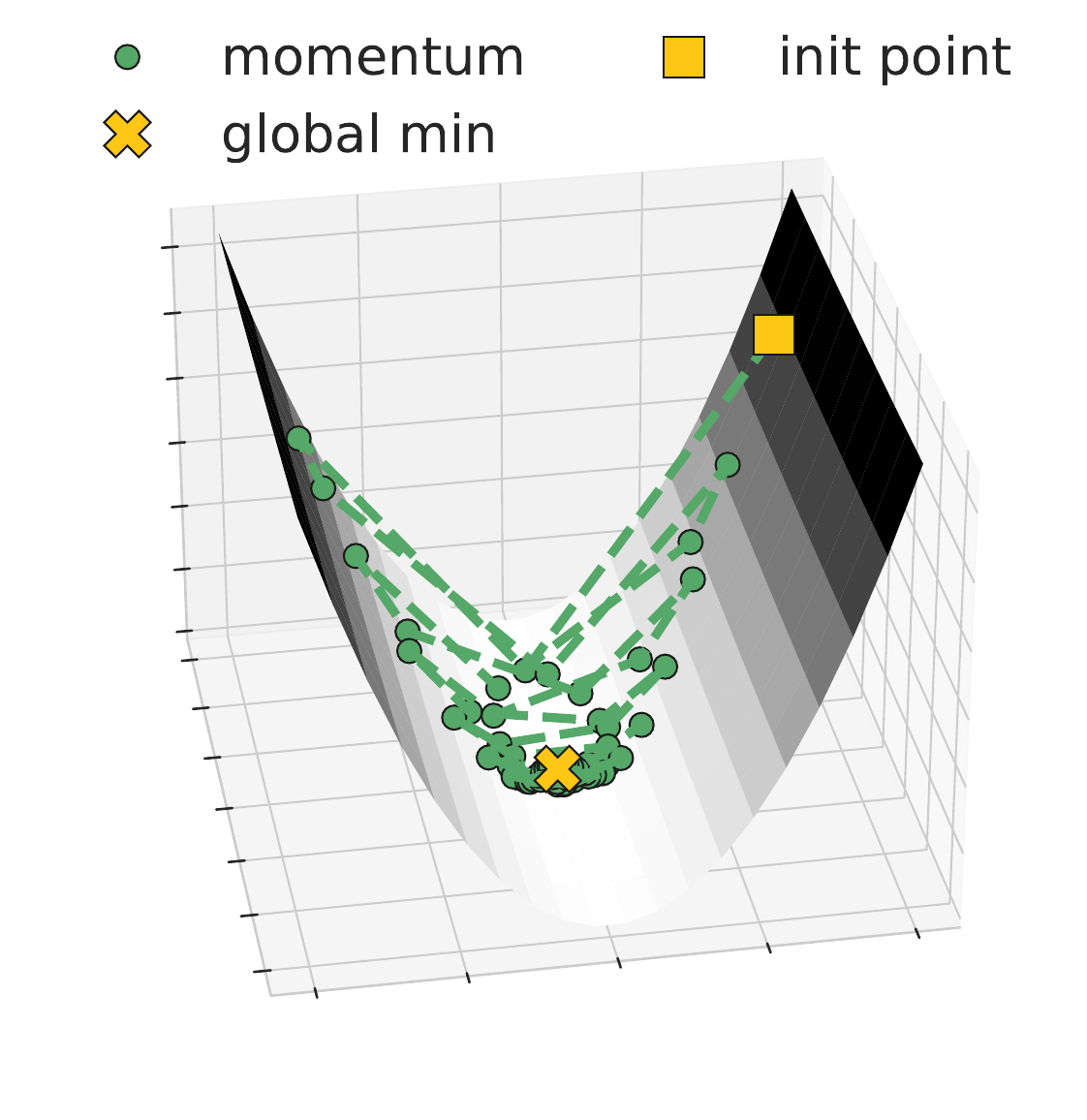}
\label{fig:mom1}
}
\end{subfloat}%
\begin{subfloat}[$\beta=0.9$ and large $h$.]{
\includegraphics[width=0.33\columnwidth]{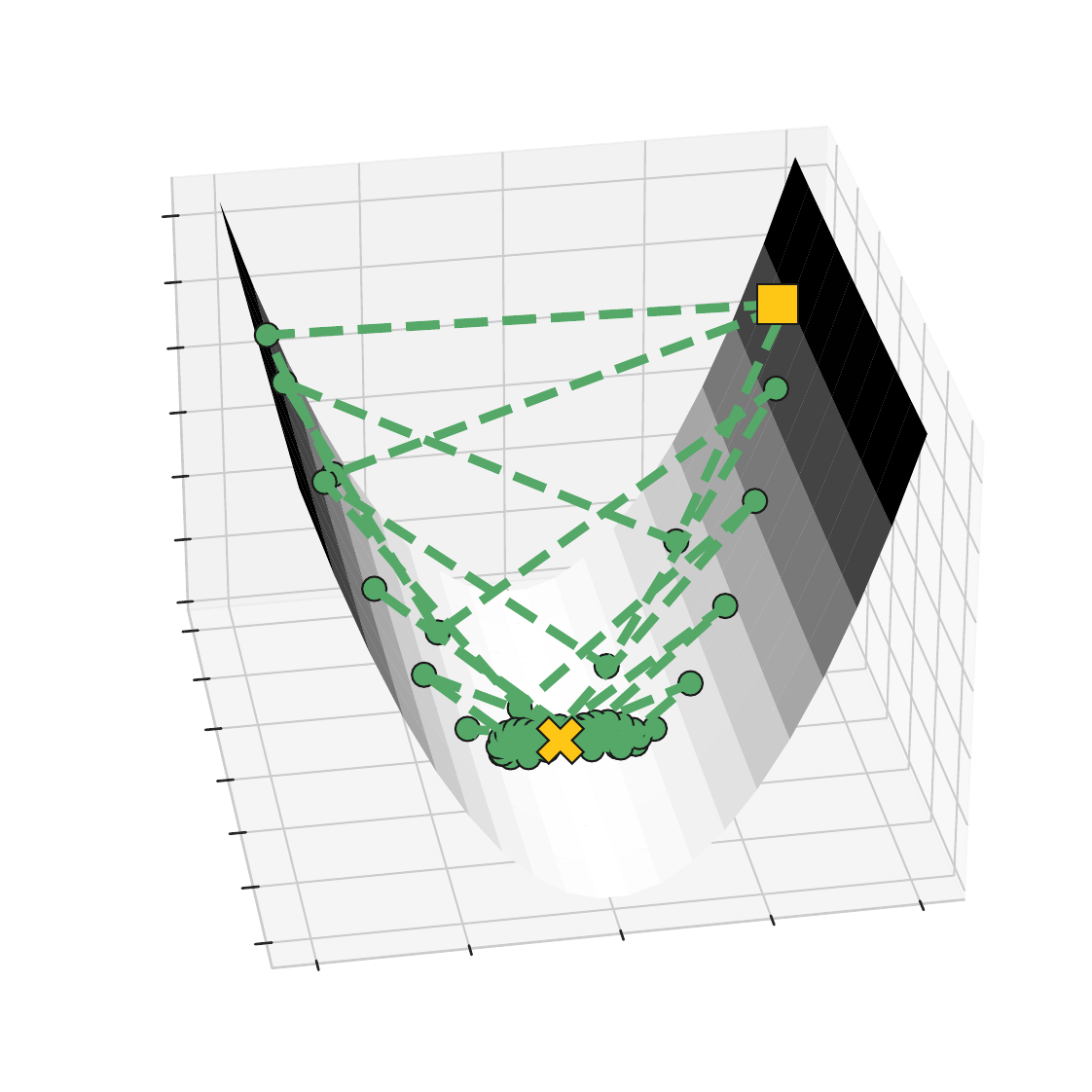}
\label{fig:mom2}
}
\end{subfloat}%
\begin{subfloat}[$\beta=0.5$ and large $h$.]{
\includegraphics[width=0.33\columnwidth]{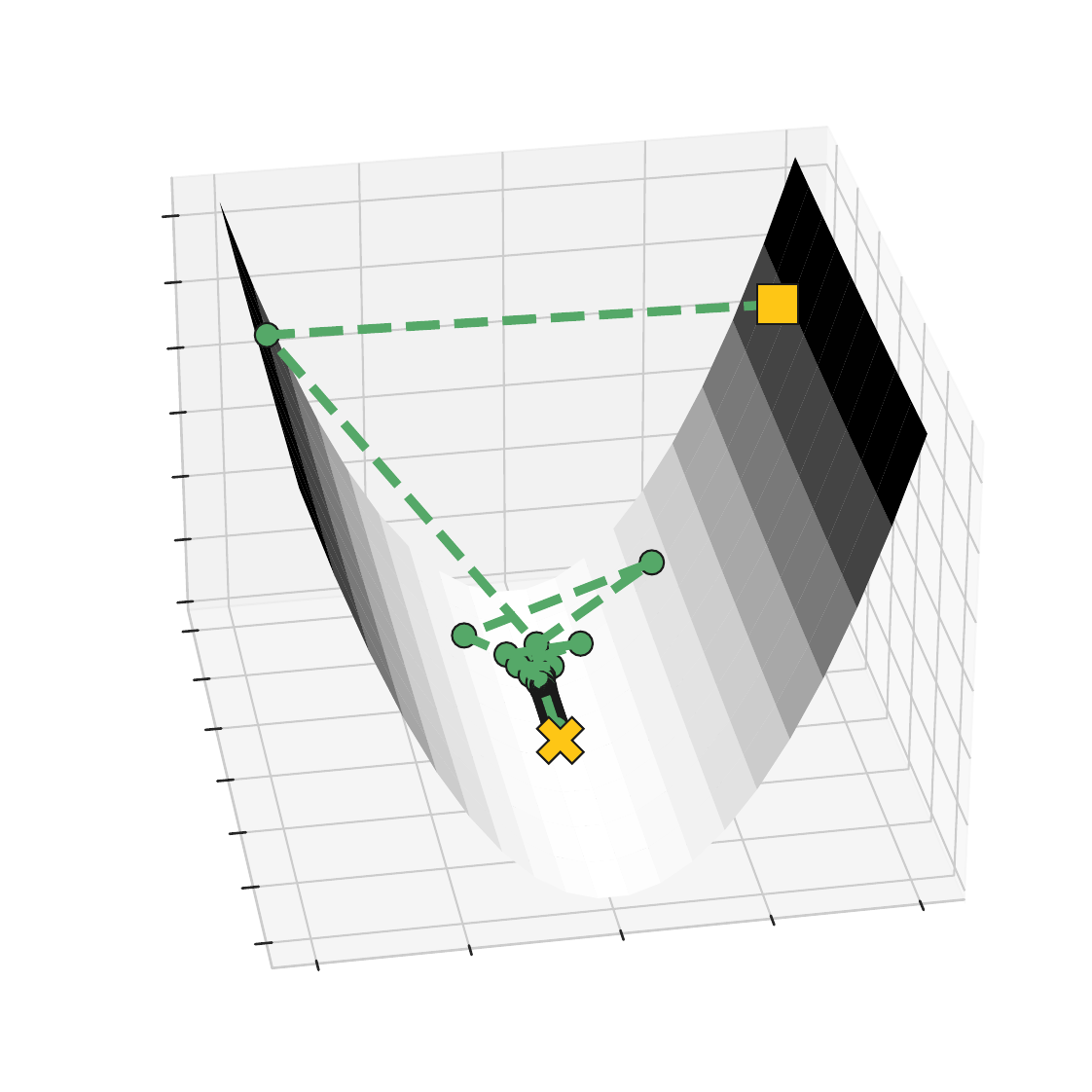}
\label{fig:mom3}
}
\end{subfloat}%
\caption[The importance of the learning rate and decay rate in gradient descent with momentum.]{The importance of the learning rate $h$ and decay rate $\beta$ in gradient descent with momentum. \subref{fig:mom1}: momentum can speed up training but can also introduce instabilities when the gradient changes direction; the learning rate used is the same as in Figure~\ref{fig:dd_intro_1}, where gradient descent converged to the minimum without instabilities. \subref{fig:mom2} and \subref{fig:mom3}: for learning rates where gradient descent diverges, as we observed in Figure~\ref{fig:dd_intro_3}, momentum can stabilise training by dampening the effect of local large gradient magnitudes and large learning rates; from this simple example one can already observe the intricacies of the interactions of decay rates and learning rates when momentum is used.}
\label{fig:mom}
\end{figure}
We show the effect of $\beta$ and $h$ on the optimisation trajectory in Figure~\ref{fig:mom}.
While momentum can speed up training and reach a neighbourhood of a local minimum quicker than vanilla gradient descent, for large decay rates $\beta$ it can lead to large updates that move further away from the equilibrium. We show this effect
  in Figure~\ref{fig:mom1}, where we observe that despite using a small learning rate, with a high decay rate $\beta$ momentum can oscillate around the minimum significantly before reaching it, while gradient descent with the same learning rate does not oscillate around the minimum (Figure~\ref{fig:dd_intro_1}).  This effect can be mitigated by using a smaller decay rate, as exemplified in Figure~\ref{fig:mom3}.
  Dampening the strength of the current gradient update through momentum can also stabilise training, as we show in Figure~\ref{fig:mom2}. While for the same learning rate gradient descent diverges (Figure~\ref{fig:dd_intro_3}), momentum does not, since the large local gradients moving away from the equilibrium are dampened by the moving averages obtained from past gradients. For quadratic functions, this difference between gradient descent and momentum can be seen in the smallest learning rate which leads to divergence: while gradient descent diverges if $h > 2/\lambda_0$ where $\lambda_0$ is the largest eigenvalue of the Hessian $\nabla_{\vtheta}^2 E$, for momentum divergence occurs when $h > 2/\lambda_0 + 2\beta/\lambda_0\ge 2/\lambda_0$ since $\beta \ge 0$~\citep{cohen2021gradient}. We will come back to the importance of the learning rate relative to the Hessian eigenvalues in later chapters.

Gradient descent with momentum is still very much used for deep learning optimisation and has been credited with increased generalisation performance compared to other optimisers~\citep{gupta2021adam,zhou2020towards}, though recent work has shown that other well tuned methods can outperform momentum if well tuned~\citep{kingma2014adam}.

\textbf{The Adam optimiser} \citep{kingma2014adam}, perhaps the most popular optimisation algorithm in deep learning, combines some of the previously discussed insights: it uses momentum and an RMSprop-like approach of dividing by a moving average of the square gradient (all operations below are element-wise):
\begin{align}
\vm_t &= \beta_1 \vm_{t-1} + (1 - \beta_1) \nabla_{\vtheta}E(\vtheta_{t-1})\\
\vv_t &= \beta_2 \vv_{t-1} + (1 - \beta_2) \nabla_{\vtheta}E(\vtheta_{t-1})^2\\
\vtheta_t &= \vtheta_{t-1} - h \frac{\frac{1}{1 - \beta_1^t} \vm_t}{\sqrt{\frac{1}{1 - \beta_2^t}\vv_t} + \epsilon},
\label{eq:last_adam}
\end{align}
where $\beta_1, \beta_2$ and $\epsilon$ are hyperparameters usually set to $\beta_2 = 0.999$ and $\beta_1 \in [0, 0.9]$ and $\epsilon \in [10^{-1}, 10^{-8}]$ with the default at $10^{-8}$.
The division in Eq \eqref{eq:last_adam} is obtained using a bias correction argument for the moving averages. To see this, we write:
\begin{align}
\vm_t &= (1 - \beta_1) \sum_{i=1}^t \beta_1^{i-1} \nabla_{\vtheta}E(\vtheta_{t-i}).
\label{eq:adam_m_expanded}
\end{align}
The authors make the argument that if the gradients at iteration $i$ $\nabla_{\vtheta}E(\vtheta_{i})$ are drawn from the distribution $p_{g,i}$
\begin{align}
\mathbb{E}_{p_m} [\vm_t] &=   (1 - \beta_1) \sum_{i=1}^t \beta_1^{i-1} \mathbb{E}_{p_{g, i}} \left[\nabla_{\vtheta}E(\vtheta_{t-i})\right]\\
                     & = (1 - \beta_1^{t}) \mathbb{E}_{p_g} \left[\nabla_{\vtheta}E(\vtheta_{t-1})\right] + \vxi,
\end{align}
where $\vxi$ is a bias term accounting for the changes in the expected value of the gradients at different iterations; it will be $\mathbf{0}$ if the gradient mean does not change between iterations otherwise it is expected to be small due to the decaying contribution of older gradients.
Thus, to ensure the expected value of the moving average stays the same as the gradient and does not decrease steadily as training progresses, the division in Eq~\eqref{eq:last_adam} is used; a similar argument can be made for the second moment. This ensures that in expectation the update (if a small $\epsilon$ is used), is approximately of magnitude $h$ for each parameter. We note that while the assumption that the bias $\vxi$ is small does not hold for certain areas of training, such as when the gradient changes direction due to instability around an equilibrium, the above argument holds in areas of low curvature (i.e. areas of space where the gradient does not change substantially); this is something we will use later.

\begin{figure}[t!]
\begin{subfloat}[$\beta_1=0.9$, small $h$, $\epsilon=10^{-8}$.]{
\includegraphics[width=0.32\columnwidth]{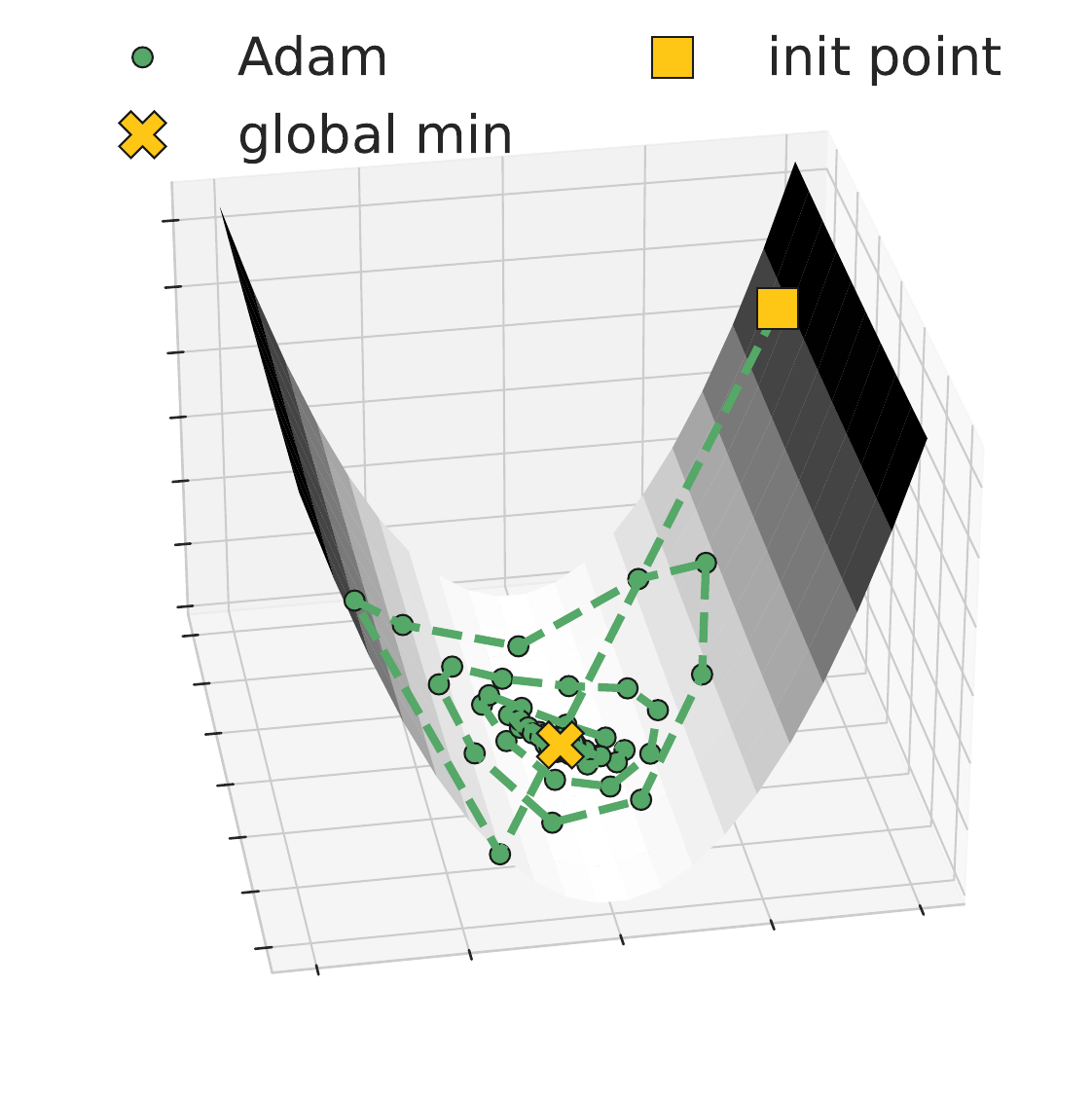}
\label{fig:adam1}
}
\end{subfloat}%
\begin{subfloat}[$\beta_1=0.9$, large $h$, $\epsilon=10^{-8}$.]{
\includegraphics[width=0.33\columnwidth]{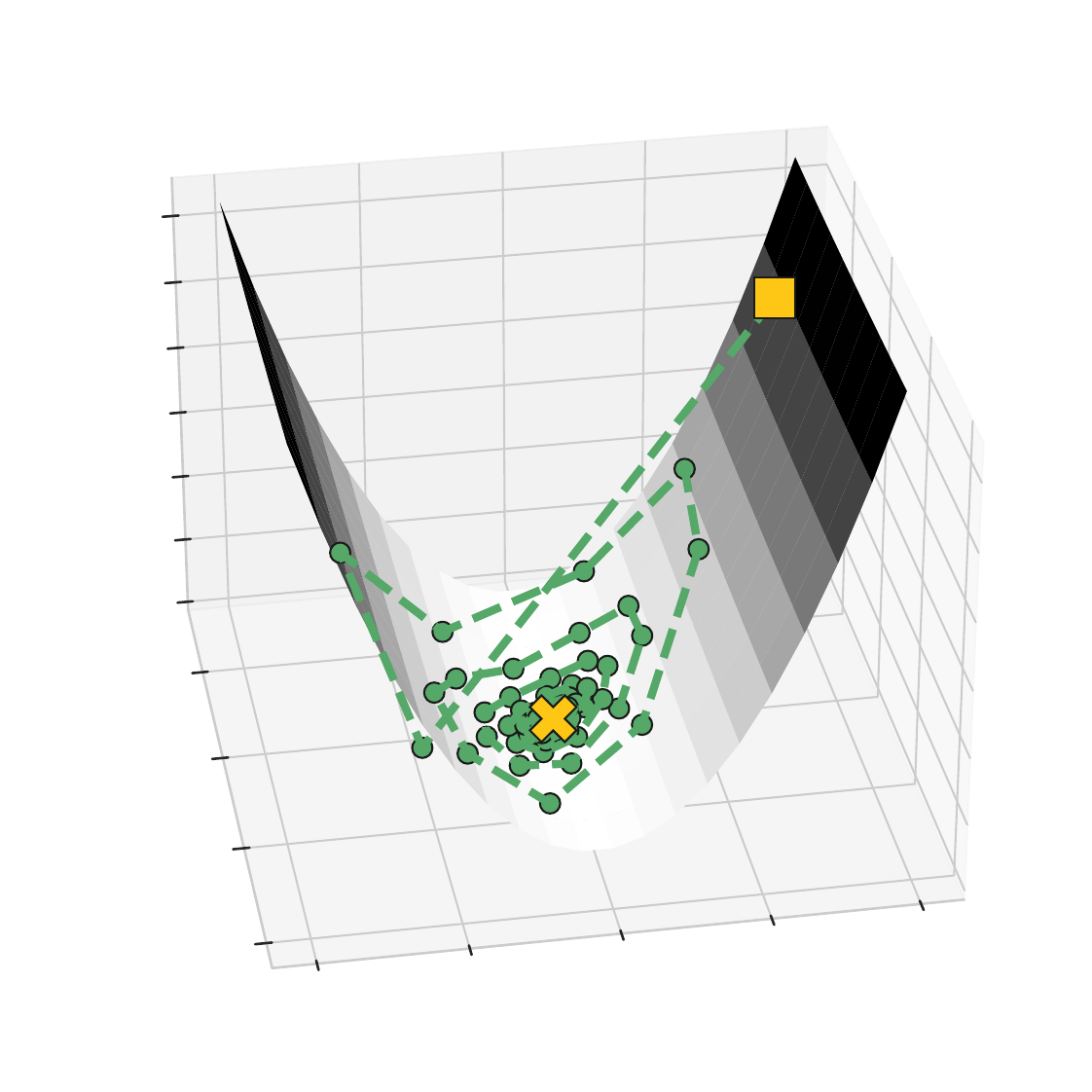}
\label{fig:adam2}
}
\end{subfloat}%
\begin{subfloat}[$\beta_1=0.5$, large $h$, $\epsilon=10^{-8}$.]{
\includegraphics[width=0.33\columnwidth]{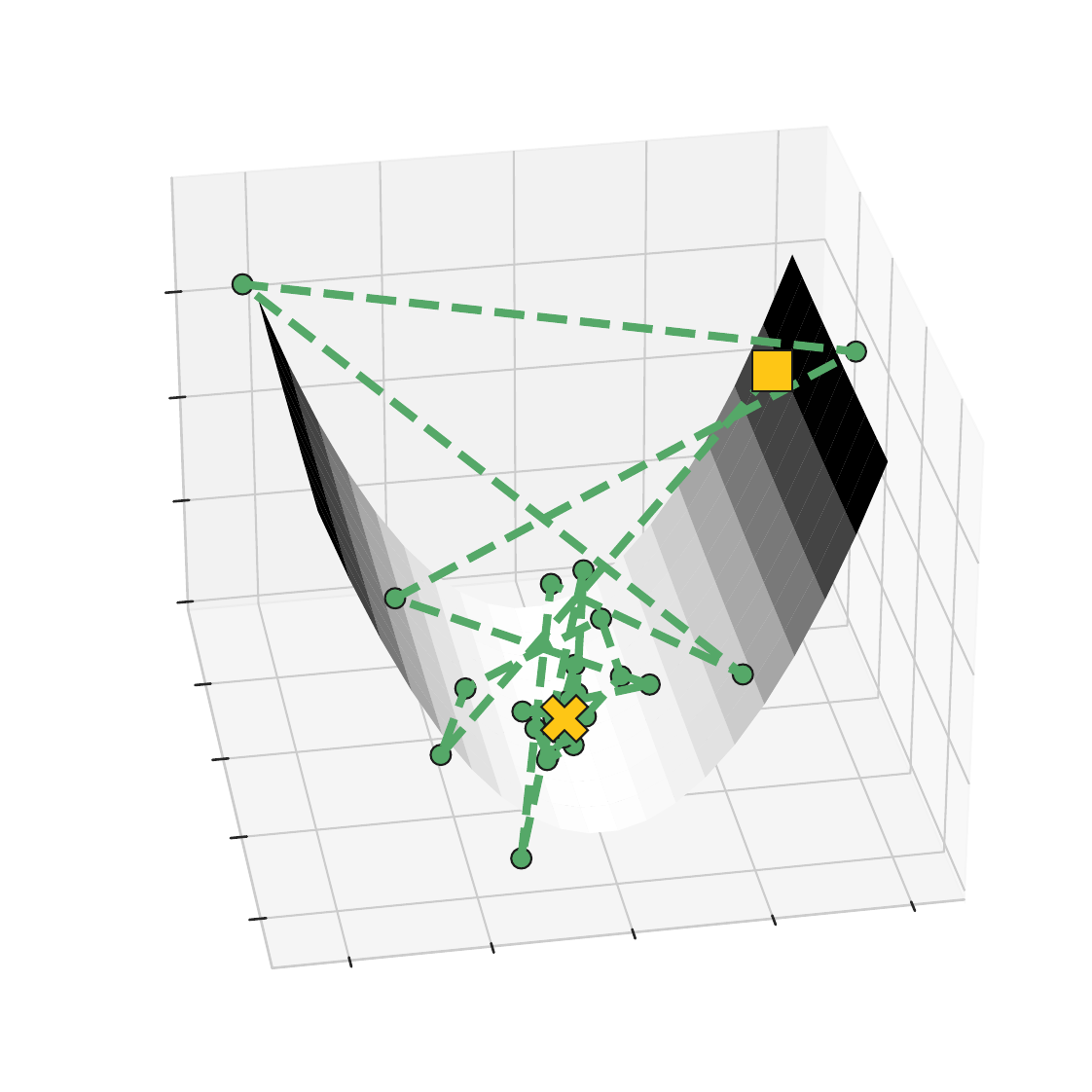}
\label{fig:adam3}
}
\end{subfloat}
\begin{subfloat}[$\beta_1=0.9$, small $h$, $\epsilon=10^{-2}$.]{
\includegraphics[width=0.32\columnwidth]{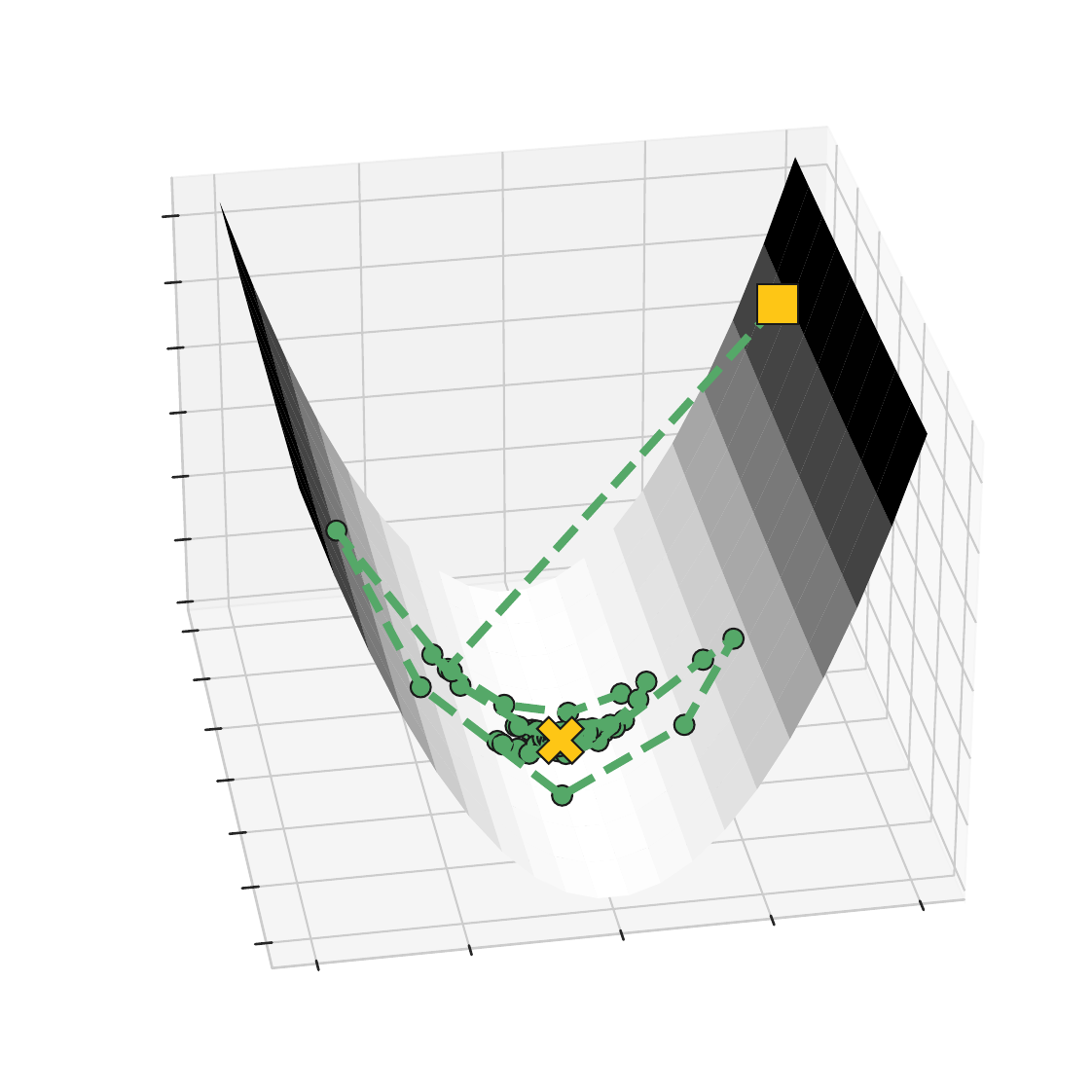}
\label{fig:adam1_large_eps}
}
\end{subfloat}%
\begin{subfloat}[$\beta_1=0.9$, small $h$, $\epsilon=10^{-2}$.]{
\includegraphics[width=0.33\columnwidth]{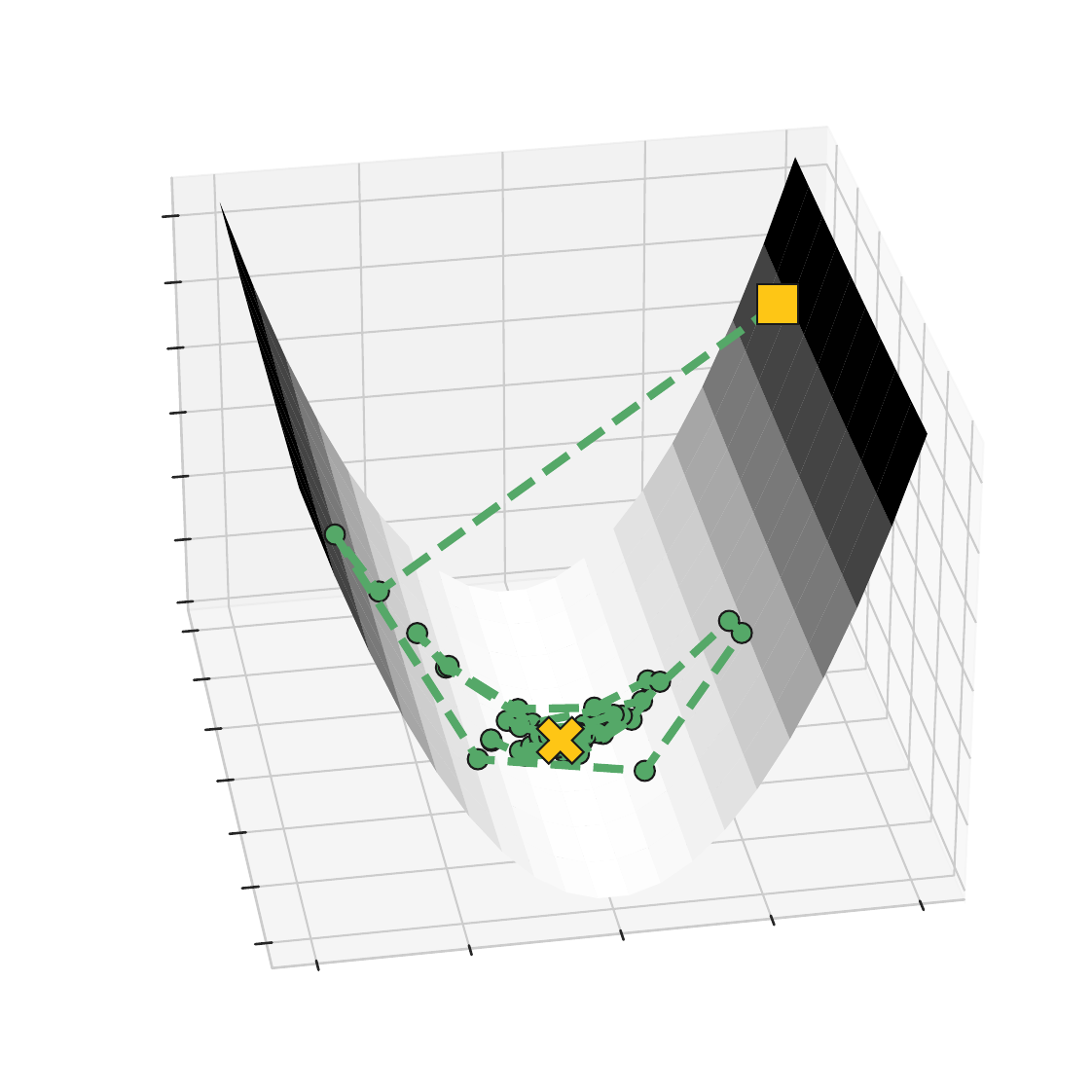}
\label{fig:adam2_large_eps}
}
\end{subfloat}%
\begin{subfloat}[$\beta_1=0.5$, large $h$, $\epsilon=10^{-2}$.]{
\includegraphics[width=0.33\columnwidth]{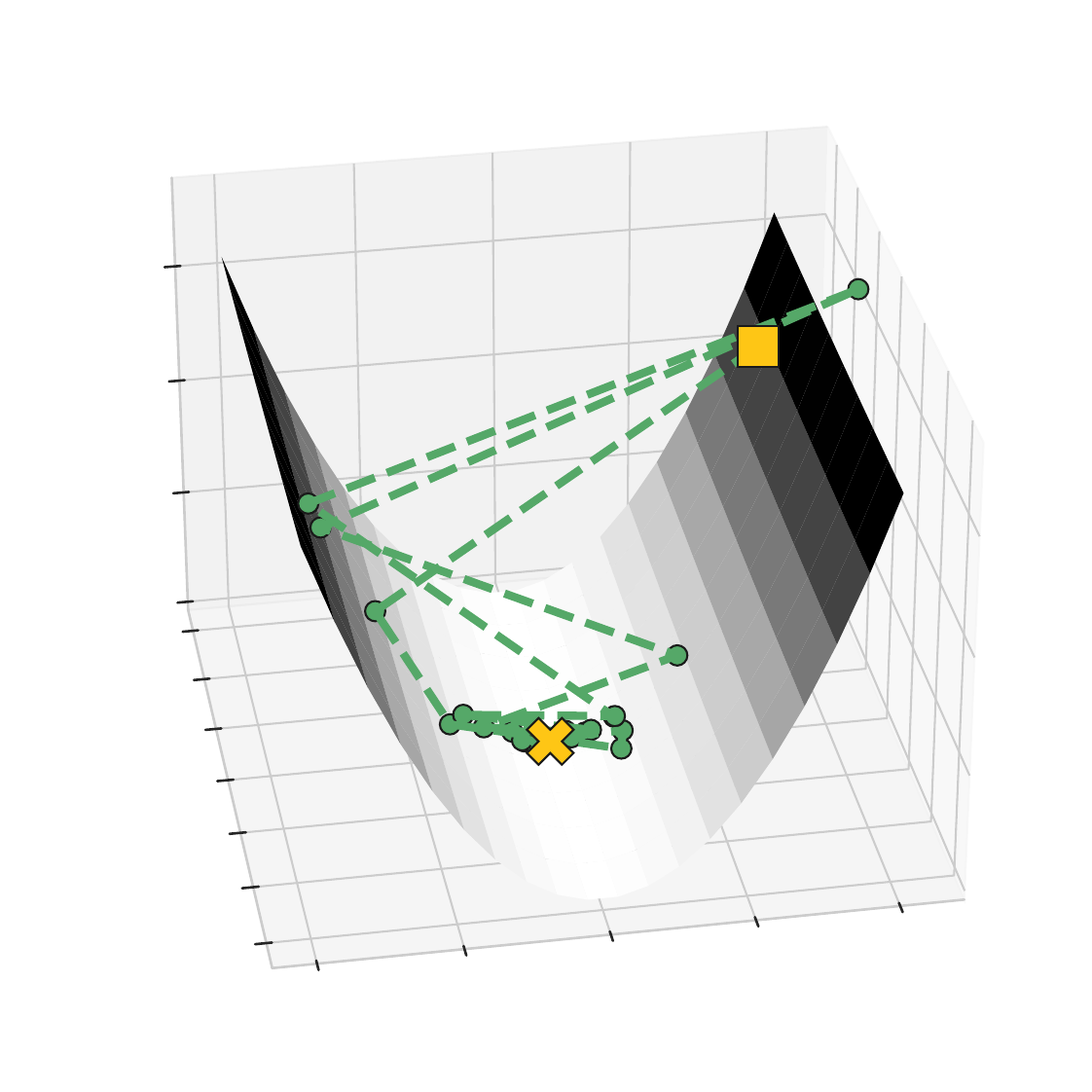}
\label{fig:adam3_large_eps}
}
\end{subfloat}%
\caption[The effect of hyperparameters in Adam.]{The importance of the Adam hyperparameters: learning rate $h$ and decay rate $\beta_1$ and $\epsilon$. We use the default $\beta_2 = 0.999$. Top row: the update normalisation in Adam can stabilise training but can also lead to instability compared to momentum; this row uses $\epsilon=10^{-8}$, the default value and the other hyperparameters are the same as in Figure~\ref{fig:mom} which shows the corresponding momentum trajectories. Bottom row: while keeping other hyperparameters fixed, changing $\epsilon$ to $10^{-2}$ leads to vastly different behaviours and can stabilise training by avoiding dividing by small values.}
\label{fig:adam}
\end{figure}

By investigating the commonly used Adam hyperparameters, $\beta_1 \in [0, 0.9]$ and $\beta_2 = 0.999$, we get intuition about its behaviour. Since $\beta_1$ is smaller than $\beta_2$, Adam is more sensitive to the sign of the gradient (controlled by $\beta_1$) than its magnitude (controlled by $\beta_2$). Figures~\ref{fig:adam1} and~\ref{fig:adam2} show that the exponential moving averages in the denominator of the Adam update can dampen large updates compared to momentum (the corresponding momentum figures are shown in Figure~\ref{fig:mom1} and~\ref{fig:mom2}), but oscillations around the minimum still occur. 
When a low decay rate for the nominator is used, such as $\beta_1 = 0.5$, in conjunction with a large $h$, instabilities can occur if small gradients are followed by a large gradient, since $\vm_t$ will be large but $\vv_t$ will remain small due to $\beta_2 = 0.999$. We see this in Figure~\ref{fig:adam3}, which we can contrast with momentum for $\beta=\beta_1$ and the same learning rate $h$ in Figure~\ref{fig:mom3}: momentum exhibits stable training, and does not exit the area around the equilibrium. Relatedly, the use of moving averages in the denominator $\vv_t$ has been linked to the lack of convergence in Adam for simple online convex optimisation problems~\citep{reddi2019convergence}.  

The hyperparameter $\epsilon$ does not change the sign of the update, but it can change its magnitude. Specifically, a high $\epsilon$ decreases the magnitude of the parameter update. We show the effect of $\epsilon$ in Figure~\ref{fig:adam}, where we show that using $\epsilon =10^{-2}$ (bottom row) can lead to more stable trajectories compared to $\epsilon =10^{-8}$ (top row).

Adam is a very popular the optimiser in deep learning, and has been attributed with progress in various fields including GANs~\citep{dcgan}, and training modern models such as transformers~\citep{liu2020understanding,choi2019empirical}. The $\epsilon$ hyperparameter in Adam has been noted to be important for certain machine learning applications \cite{choi2019empirical}, including reinforcement learning, where the $\epsilon$ used is large compared to supervised learning \citep{bellemare2017distributional,hessel2018rainbow}.

\section{Continuous-time methods}

We now describe the continuous-time approaches we will use in the rest of this thesis.

\subsection{A brief overview of stability analysis}
\label{sec:stability_analysis_overview}
Stability analysis is a tool for continuous-time systems that helps decide whether $\vtheta^* \in \mathbb{R}^D$ is a stable fixed point for a continuous-time flow 
\begin{align}
 \dot{\vtheta} = f(\vtheta),
\end{align}
where $f: \mathbb{R}^D \rightarrow \mathbb{R}^D$ is the vector field of the flow. 
The most used definitions of stability are Lyapunov stability and asymptotic stability. Lyapunov stability requires
\begin{align}
\forall \epsilon > 0, \exists \delta > 0 \hspace{3em} \norm{\vtheta(0) - \vtheta^*} < \delta \implies \ \forall t \ge 0 \ \norm{\vtheta(t) - \vtheta^*} < \epsilon.
\end{align}
Asymptotic stability requires Lyapunov stability and additionally
\begin{align}
\exists \delta > 0 \hspace{3em} \norm{\vtheta(0) - \vtheta^*} < \delta \implies \lim_{t \rightarrow \infty} \norm{\vtheta(t) - \vtheta^*} = \mathbf{0}.
\end{align}
The rate of convergence of asymptotic stability is exponential if
\begin{align}
\exists \delta, C, \alpha > 0 \hspace{1em}   \norm{\vtheta(0) - \vtheta^*} < \delta \implies \forall t \ge 0 \ \norm{\vtheta(t) - \vtheta^*} \le C \norm{\vtheta(0) - \vtheta^*} e^{- \alpha t}.
\end{align}
Exponential asymptotic stability can be established using the following criterion:
\begin{remark} (From~\citep{hartman1960lemma}) A fixed point $\vtheta^*$ with $ f(\vtheta^*) = \mathbf{0}$ where the  Jacobian of the vector field $\jacthetaf(\vtheta^*)$ has eigenvalues with strictly negative real part is a stable fixed point of the flow $\dot{\vtheta} = f(\vtheta)$, with asymptotic exponential convergence. In contrast, if a fixed point has a Jacobian with an eigenvalue with strictly positive real part, that is an unstable equilibrium and the flow will not converge to it.
\label{rem:stability_analysis}
\end{remark}
Remark~\ref{rem:stability_analysis} provides us with a useful test to establish convergence based on a fixed point's Jacobian eigenvalues. 
The test is inconclusive for equilibria with non-strictly negative real part Jacobian eigenvalues, i.e. Jacobians where a subset of eigenvalues are have strictly negative real part while some have $0$ real part.
Throughout this thesis, when performing stability analysis of a continuous-time system we will be testing exponential asymptotic stability using the above criterion.
\begin{corollary}[The NGF and local minima.] The NGF $ \dot{\vtheta} = - \nabla_{\vtheta} E$ is only locally attracted to local minima of $E$, and all strict local minima are exponentially attractive under the NGF.
\label{cor:ngf_local_minimum}
\end{corollary}
This follows from the application of Remark~\ref{rem:stability_analysis}. If a stationary point ${\nabla_{\vtheta} E(\vtheta^*) = \mathbf{0}}$ is attractive under the NGF, the eigenvalues of the Jacobian $\jacparam{\vtheta}{(-\nabla_{\vtheta}E)}\left(\vtheta^*\right)$ have strictly negative or 0 real part. Since $\jacparam{\vtheta}{(-\nabla_{\vtheta}E)}\left(\vtheta^*\right) = - \nabla_{\vtheta}^2 E$ is a symmetric matrix, this condition is equivalent to the eigenvalues of $\nabla_{\vtheta}^2 E$ being non-negative, the local minimum condition. Conversely, since strict local minima have strictly positive eigenvalues, the Jacobian has strictly negative eigenvalues at strict local minima, leading to exponential local stability under the NGF.

\subsection{Backward error analysis}
\label{sec:bea}

Backward error analysis (BEA) is a tool in numerical analysis developed to understand the discretisation error of numerical integrators. We now present an overview of how to use BEA in the context of the gradient descent update $\vtheta_t = \vtheta_{t-1} - h \nabla_{\vtheta} E(\vtheta_{t-1})$ with $\vtheta \in \mathbb{R}^D$; for a general overview see~\citet{hairer2006geometric}.
BEA provides a modified vector field
\begin{equation}
\tilde f_n(\vtheta) =  - \nabla_{\vtheta} E + h f_1(\vtheta) + \cdots + h^n f_n(\vtheta)
\end{equation}
 by finding functions $f_1:\mathbb{R}^D \rightarrow \mathbb{R}^D$, ..., $f_n: \mathbb{R}^D \rightarrow \mathbb{R}^D$, such that the solution of the modified flow at order $n$, that is,
\begin{align}
    \bm{{\dot{\vtheta}}} =  - \nabla_{\vtheta} E + h f_1(\vtheta) + \cdots + h^n f_n(\vtheta)
\label{eq:general_modified_vector_field}
\end{align} follows the discrete dynamics of the gradient descent update with an error $\| \vtheta_t - \vtheta(h)\|$ of order $\mathcal O(h^{n+2})$, where $\vtheta(h)$ is the solution of the modified equation truncated at order $n$ at time $h$ and $\vtheta(0) = \vtheta_{t-1}$. 
The full modified vector field with all orders ($n \rightarrow \infty$)
\begin{equation}
\tilde f(\vtheta) = - \nabla_{\vtheta} E + h f_1(\vtheta) + \cdots + h^n f_n(\vtheta) + \cdots
\label{eq:bea_series}
\end{equation}
is usually divergent and only forms an asymptotic expansion. What BEA provides is the Taylor expansion in $h$ of an unknown $h$-dependent vector field $f_h(\vtheta)$ developed at $h=0$:
\begin{equation}
    \tilde f(\vtheta) = \textrm{Taylor}_{h=0} f_h(\vtheta).
\end{equation}
Thus, a strategy for finding $f_h$ is to find a series of the form in Eq~\eqref{eq:bea_series} via BEA and then find the function  $f_h$ such that its Taylor expansion in $h$ at 0 results in the found series.
Using this approach we can find the flow $ \dot{\vtheta} = f_h(\vtheta)$ which  describes the gradient descent step $\vtheta_t = \vtheta_{t-1} - h \nabla_{\vtheta} E(\vtheta_{t-1})$ exactly. 

While flows obtained using BEA are constructed to approximate one gradient descent step, the same flows can be used over multiple gradient descent steps as shown in Section~\ref{sec:multiple_steps_proof} in the Appendix.

\citet{igr} used this technique to find the $\mathcal{O}(h^2)$ correction term of gradient descent in the single-objective setting. They showed that for a model with parameters  $\vtheta$ and loss $E(\vtheta)$, optimised with gradient descent $\vtheta_t = \vtheta_{t-1} - h\nabla_{\vtheta} E(\vtheta)$, the first-order modified equation is 
\begin{align} 
\dot{\vtheta} = -\nabla_{\vtheta}E   -\frac{h}{2} \nabla_{\vtheta}^2 E \nabla_{\vtheta} E,
\label{eq:first_igr}
\end{align}
which can be written as
\begin{align}
\dot{\vtheta} =-\nabla_{\vtheta} \tilde E(\vtheta) = -\nabla_{\vtheta}\left(E(\vtheta) + \frac h4 \norm{\nabla_{\vtheta} E(\vtheta)}^2\right).
\end{align}
Thus, gradient descent can be seen as implicitly minimising the modified loss
\begin{align}
\tilde E(\vtheta) = E(\vtheta) + \frac h4 \norm{\nabla_{\vtheta} E(\vtheta)}^2.
\label{eq:sup_learning_igr}
\end{align}
This shows that when training models with gradient descent, there is an implicit regularisation effect, dependent on learning rate $h$, which biases learning towards paths with low gradient norms. The authors refer to this phenomenon as `implicit gradient regularisation'; we will thus refer to the flow in Eq~\eqref{eq:first_igr} as the IGR flow.

\subsection{BEA proofs}
\label{sec:bea_proofs}
The general structure of BEA proofs is as follows: start with a Taylor expansion in $h$ of the modified flow in Eq~\eqref{eq:general_modified_vector_field}; write each term in the Taylor expansion as a function of $\nabla_{\vtheta} E$ and the desired $f_i$ using the chain rule; group together terms of the same order in $h$ in the expansion; and identify $f_i$ such that all terms of $\mathcal{O}(h^p)$ are 0 for $p \ge 2$, as is the case in the gradient descent update. A formal overview of BEA proofs can be found in Section \ref{sec:bea_proof_structure} in the Appendix.

We now exemplify how to use BEA to find the IGR flow we showed in Eq~\eqref{eq:first_igr}~\citep{igr}.  Since we are only looking for the first correction term, we only need to find $f_1$.
We perform a Taylor expansion to find the value of $\vtheta(h)$ up to order $\mathcal O(h^{3})$ and then identify $f_1$ from that expression such that the error $\| \vtheta_t - \vtheta(h)\|$ is of order $\mathcal O(h^{3})$.
We have
\begin{align}
\vtheta(h) = \vtheta_{t-1} + h \vtheta^{(1)}(\vtheta_{t-1}) + \frac{h^2}{2} \vtheta^{(2)}(\vtheta_{t-1}) +  \mathcal{O}(h^3).
\end{align}
We know by the definition of the modified vector field in Eq~\eqref{eq:general_modified_vector_field} that
\begin{align}
\vtheta^{(1)} = - \nabla_{\vtheta} E + h f_1({\vtheta}).
\end{align}
We can then use the chain rule to obtain
\begin{align}
\vtheta^{(2)} = \frac{d \left(- \nabla_{\vtheta} E + h f_1({\vtheta})\right)}{dt}  = - \frac{d \nabla_{\vtheta} E}{dt} + \mathcal{O}(h) = \nabla_{\vtheta}^2 E \nabla_{\vtheta} E + \mathcal{O}(h).
\end{align}
Thus,
\begin{align}
\vtheta(h) = \vtheta_{t-1} - h \nabla_{\vtheta} E(\vtheta_{t-1}) + h^2 f_1(\vtheta_{t-1}) + \frac{h^2}{2}  \nabla_{\vtheta}^2 E (\vtheta_{t-1})\nabla_{\vtheta} E (\vtheta_{t-1})+  \mathcal{O}(h^3).
\end{align}
We can then write
\begin{align}
\vtheta_t - \vtheta(h) &= \vtheta_{t-1} - h \nabla_{\vtheta} E(\vtheta_{t-1}) - \vtheta(h) \\
&= h^2 f_1(\vtheta_{t-1}) + \frac{h^2}{2}  \nabla_{\vtheta}^2 E (\vtheta_{t-1})\nabla_{\vtheta} E (\vtheta_{t-1})+  \mathcal{O}(h^3).
\end{align}
For the error to be of order $\mathcal{O}(h^3)$ the terms of order $\mathcal{O}(h^2)$ have to be $\mathbf{0}$.
This entails 
\begin{align}
f_1(\vtheta_{t-1}) =  -\frac{1}{2} \nabla_{\vtheta}^2 E(\vtheta_{t-1}) \nabla_{\vtheta} E(\vtheta_{t-1}),
\label{eq:igr_proof_value_iterate}
\end{align}
from which we conclude to $f_1 =  -\frac{1}{2} \nabla_{\vtheta}^2 E \nabla_{\vtheta} E $ leading to Eq~\eqref{eq:first_igr}.

\section{Multiple objective optimisation}
\label{sec:intro_multi_objective_optimisation}

We have thus far discussed optimisers for single-objective problems, which encompasses many machine learning use cases, including supervised learning and certain formulations of reinforcement learning and generative modelling.
There are, however, use cases in machine learning beyond single-objective problems, including GANs and types of reinforcement learning, such as actor-critic frameworks. 
These settings are often formulated via nested minimisation problems:
\begin{align}
 \min_{\vtheta}   E_{\vtheta}(\arg \min_{\vphi} E_{\vphi}(\vphi, \vtheta), \vtheta).
 \label{eq:nested_op}
\end{align}
Solving nested optimisation problems is computationally expensive, as it would require solving the inner optimisation problem for each parameter update of the outer optimisation problem. Thus, in deep learning practice nested optimisation problems are often translated into problems of the form
\begin{align}
 &\min_{\vphi} E_{\vphi}(\vphi, \vtheta) \label{eq:prob_games_1} \\
 &\min_{\vtheta} E_{\vtheta}(\vphi, \vtheta) \label{eq:prob_games_2}. \
\end{align}
Multi-objective optimisation problems of this kind are often cast as a game. Each player has a set of actions to take, in the form of an update for their parameters, $\vphi$ for the first player, and $\vtheta$ for the second. The loss functions are the negative of the player's respective payoff. As it is common in the literature, we will often adapt the game theory nomenclature in this setting;
 this connection allows for a bridge between the machine learning and the game theory communities.

When translating problems of the form in Eq~\eqref{eq:nested_op} into optimisation procedures that preserve the nested structure of a game it is common to use one of the single-objective algorithms discussed Section~\ref{sec:opt_algo} to update the first player's parameters, followed by the use of the same algorithm to update the second player's parameters. This approach is often referred to as \textit{alternating updates}; we summarise it in Algorithm~\ref{alg:alt_updates}. Optimisation approaches that further account for the nested game structure do exist~\citep{metz2016unrolled}, but they tend to be more computationally prohibitive and thus less used in practice. 
In some settings, there is no nested structure or the nested structure is further discarded and both players are updated simultaneously, as we show in Algorithm~\ref{alg:sim_updates}.

\begin{algorithm}[t]
\caption{Alternating updates}\label{alg:alt_updates}
\begin{algorithmic}
\State $k \ge 0$; the number of first player updates for each second player update
\State Choice of optimiser\_routine($E$, current\_value); e.g: $gd(E, \vtheta_t) = \vtheta_t - h \nabla_{\vtheta} E(\vtheta_t)$
\State Initialise $\vphi_0, \vtheta_0$
\State $t\gets 0$
\While {\text{training}}
\State $\vphi \gets \vphi_t$
 \For {$i$ in \{1, ..., $k$\} }
    \State $\vphi \gets$ optimiser\_routine($E_{\vphi}(\cdot, \vtheta_t), \vphi)$
 \EndFor
\State $\vphi_{t+1} \gets \vphi$
\State $\vtheta_{t +1} \gets$ optimiser\_routine($E_{\vtheta}(\vphi_{t+1}, \cdot), \vtheta_t)$
\State $t\gets t+1$
\EndWhile
\end{algorithmic}
\end{algorithm}

\begin{algorithm}[t]
\caption{Simultaneous updates}\label{alg:sim_updates}
\begin{algorithmic}
\State Initialise $\vphi_0, \vtheta_0$
\State Choice of optimiser\_routine($E$, current\_value); e.g: $gd(E, \vtheta_t) = \vtheta_t - h \nabla_{\vtheta} E(\vtheta_t)$
\While {\text{training}}
\State $\vphi_{t +1} \gets$ optimiser\_routine($E_{\vphi}(\cdot, \vtheta_t), \vphi_t)$
\State $\vtheta_{t +1} \gets$ optimiser\_routine($E_{\vtheta}(\vphi_t, \cdot), \vtheta_t)$
\State $t\gets t+1$
\EndWhile
\end{algorithmic}
\end{algorithm}

Simultaneous gradient descent updates derived for solving the problem in Eqs~\eqref{eq:prob_games_1} and~\eqref{eq:prob_games_2}---using gradient descent as the optimiser in Algorithm~\ref{alg:sim_updates}---can be seen as the Euler discretisation of the continuous-time flow
\begin{align}
&\dot{\vphi} = -\nabla_{\vphi} E_{\vphi} \label{eq:gen_games_ode1} \\
&\dot{\vtheta} = -\nabla_{\vtheta} E_{\vtheta} \label{eq:gen_games_ode2}.
\end{align}
Analysis of these game dynamics is more challenging than in the single objective case (shown in Eq~\eqref{eq:e_min_ngf}), as we can no longer ascertain that following such flows minimises the corresponding loss functions since
\begin{align}
\frac{dE_{\vphi}}{d t} = \frac{d \vphi}{d t}^T \nabla_{\vphi} E_{\vphi} + \frac{d \vtheta}{d t}^T \nabla_{\vtheta} E_{\vphi}
 = - \norm{\nabla_{\vphi} E_{\vphi}}^2 -  \nabla_{\vtheta} E_{\vtheta}^T \nabla_{\vtheta} E_{\vphi},
\end{align}
which need not be negative. Furthermore, it is unclear how to transfer the alternating structure that preserves player order from Algorithm~\ref{alg:alt_updates} into a continuous-time flow; we will address this in a later chapter. Nonetheless, the flows in Eqs~\eqref{eq:gen_games_ode1} and~\eqref{eq:gen_games_ode2} are often studied to understand the behaviour of gradient descent in nested optimisation and games~\citep{nagarajan2017gradient,balduzzi2018mechanics,jin2020local,odegan}.

The challenges pertaining to analysing optimisation in games get further compounded when considering the notions of equilibria. The natural translation of local minimum from single-objective optimisation to games is a local Nash equilibrium. A local Nash equilibrium is a point in parameter space $(\vphi^*, \vtheta^*)$ such that no player has an incentive to deviate from $(\vphi^*, \vtheta^*)$ to another point in its local neighbourhood; that is $\vphi^*$ is a local minimum for $E_{\vphi}(\cdot, \vtheta^*)$ and $\vtheta^*$ is a local minimum for $E_{\vtheta}(\vphi^*, \cdot)$. Equivalently, $\nabla_{\vphi} E_{\vphi}(\vphi^*, \vtheta^*) = \mathbf{0}$ and $\nabla_{\vtheta} E_{\vtheta}(\vphi^*, \vtheta^*) = \mathbf{0}$ and $\nabla_{\vphi}^2
 E_{\vphi}(\vphi^*, \vtheta^*)$ and $\nabla_{\vtheta}^2 E_{\vtheta}(\vphi^*, \vtheta^*)$ are positive semi-definite, i.e. do not have negative eigenvalues.

While in the single-objective case, local minima are the only exponentially attractive equilibria of the NGF (see Corollary~\ref{cor:ngf_local_minimum}),
this \textit{correspondence between desired equilibrium and attractive dynamics does not generally occur in games}. By applying stability analysis (Remark~\ref{rem:stability_analysis}), we know that the system in Eqs~\eqref{eq:gen_games_ode1}
and~\eqref{eq:gen_games_ode2} is attracted to equilibria for which the Jacobian 
\begin{align}
  \vJ =
\begin{bmatrix}
  -\jactwoparam{\vphi}{\vphi}{E_{\vphi}} & - \jactwoparam{\vtheta}{\vphi}{E_{\vphi}} \\
  - \jactwoparam{\vphi}{\vtheta}{E_{\vtheta}} & -\jactwoparam{\vtheta}{\vtheta}{E_{\vtheta}} 
\end{bmatrix} 
\end{align}
evaluated at $(\vphi^*, \vtheta^*)$ has only eigenvalues with strictly negative real part.
In contrast, a Nash equilibrium requires that the block diagonals of $\vJ$,  namely $-\nabla_{\vphi}^2 E_{\vphi}(\vphi^*, \vtheta^*)$ and $-\nabla_{\vtheta}^2 E_{\vtheta}(\vphi^*, \vtheta^*)$, have negative eigenvalues.
Since for an unconstrained matrix there is no relationship between its eigenvalues and the eigenvalues of its block diagonals, there is no inclusion relationship between Nash equilibria and locally attractive equilibria of the underlying continuous game dynamics that applies to all two-player games.

For zero-sum games, i.e. games where  $E_{\vphi} = - E_{\vtheta} = E$, a strict Nash equilibrium will be locally attractive when following the flow in Eqs~\eqref{eq:gen_games_ode1}
and~\eqref{eq:gen_games_ode2}. To see why, we have to examine the eigenvalues of the flow's Jacobian
\begin{align}
 \vJ = 
\begin{bmatrix}
  - \jactwoparam{\vphi}{\vphi}{E} & - \jactwoparam{\vtheta}{\vphi}{E} \\
 \jactwoparam{\vphi}{\vtheta}{E} &  \jactwoparam{\vtheta}{\vtheta}{E} 
\end{bmatrix}
\end{align}
evaluated at a strict Nash equilibrium of such game.
 Due to the zero-sum formulation, the off-diagonal blocks of $\vJ$ satisfy $- \jactwoparam{\vtheta}{\vphi}{E} = - (\jactwoparam{\vphi}{\vtheta}{E})^T$, and we have 
\begin{align}
 \vJ + \vJ^T = \begin{bmatrix}
  - \nabla_{\vphi}^2 E  & \mathbf{0} \\
  \mathbf{0} &  \nabla_{\vtheta}^2 E
\end{bmatrix}.
\end{align}
Thus,  $\vJ + \vJ^T$ at the Nash equilibrium $(\vphi^*, \vtheta^*)$ is a symmetric block diagonal matrix with negative definite blocks, $-\nabla_{\vphi}^2 E(\vphi^*, \vtheta^*) $ and $\nabla_{\vtheta}^2 E(\vphi^*, \vtheta^*)$, and thus has negative eigenvalues. From here it follows that $\vJ$ has eigenvalues with strictly negative real part, and thus satisfies the condition of Remark~\ref{rem:stability_analysis}.
We note that the reverse does not hold, as being locally attractive in this case is a weaker condition than that of a Nash equilibrium.
This observation together with the realisation that in deep learning certain games might not have Nash equilibria~\citep{farnia2020gans}, has lead the search for other equilibrium measures including local stability and Stackelberg equilibria~\citep{fiez2019convergence,berard2019closer,wang2019solving}.

\section{Conclusion}

In this section we have introduced the main optimisation algorithms used in deep learning, together with their main challenges and the analytical tools we will employ  in the rest of this thesis.

\chapter{A new continuous-time model of gradient descent}
\label{ch:pf}

\newcommand{\rebuttalrthree}[1]{\textcolor{black}{#1}}
\newcommand{\rebuttalrtwo}[1]{\textcolor{black}{#1}}
\newcommand{\rebuttalrone}[1]{\textcolor{black}{#1}}

    The recipe behind the success of deep learning has been training neural networks with gradient-based optimisation on large datasets.
    Understanding the behaviour of gradient descent, however, and particularly its instability, has lagged behind its empirical success. 
    To add to the theoretical tools available to study gradient descent we propose \textit{the principal flow} (PF), a continuous-time flow that approximates gradient descent dynamics. To the best of our knowledge, the PF is the only continuous flow that captures the divergent and oscillatory behaviours of gradient descent, including escaping local minima and saddle points. Through its dependence on the eigen-decomposition of the Hessian the PF sheds light on the recently observed edge of stability phenomena in deep learning~\citep{cohen2021gradient}. Using our new understanding of instability we propose a learning rate adaptation method that enables us to control the trade-off between training stability and test set evaluation performance.

\section{Introduction}

Our goal is to use continuous-time models to understand the behaviour of gradient descent.  Using continuous-time dynamics to understand discrete-time systems opens up tools from dynamical systems, such as stability analysis, and has a long history in optimisation and machine learning~\citep{glendinning1994stability,saxe2013exact,nagarajan2017gradient,lampinen2018analytic,arora2018optimization,advani2020high,elkabetz2021continuous,vardi2021implicit,franca2020,igr,igr_sgd}. Most theoretical analysis of gradient descent using continuous-time systems uses the negative gradient flow, but this has well-known limitations, such as not being able to explain any behaviour contingent on the learning rate.
To mitigate these limitations 
we find a new continuous-time flow that
reveals important new roles of the Hessian in gradient descent training. 
To do so, we use backward error analysis (BEA), a method with a long history in the numerical integration community~\citep{hairer2006geometric} that has only recently been used in the deep learning context~\citep{igr,igr_sgd}.

We find that the proposed flow sheds new light on gradient descent stability, including, but not limited to, divergent and oscillatory behaviour around a fixed point.
Instability---areas  of training where the loss consistently increases---and edge of stability behaviours~\citep{cohen2021gradient}---areas of training where the loss does not behave monotonically but decreases over long time periods---are pervasive in deep learning and occur for all learning rates and architectures \citep{cohen2021gradient,gur2018gradient,gilmer2021loss,lewkowycz2020large}. We use our novel insights to understand and mitigate these instabilities.

\rebuttalrthree{The structure of this chapter is as follows:}
\begin{itemize}
    \item \rebuttalrthree{We discuss the advantages of a continuous-time approach in Section \ref{sec:motivation}, where we also highlight the limitations of existing continuous-time flows.}
    \item We introduce \textbf{the principal flow} (PF), a flow \rebuttalrthree{in complex space} defined by the eigen-decomposition of the Hessian (Section~\ref{sec:principal_flow}). To the best of our knowledge the PF is the first continuous-time flow that captures that gradient descent can diverge around local minima and saddle points. \rebuttalrthree{We show that using a complex flow is crucial in understanding instabilities in gradient descent.}
    \item We show the PF \rebuttalrthree{is better than existing flows at modelling neural network training dynamics} in Section~\ref{sec:the_pf_and_nns}. In Section~\ref{sec:instability_deep_learning}, we use the PF to shed new light on edge of stability behaviours in deep learning. \rebuttalrthree{We do so by connecting changes in the loss and Hessian eigenvalues with core quantities exposed by the PF and neural network landscapes explored through the behaviour of gradient flows.}
    \item \rebuttalrthree{Through a continuous-time perspective we demonstrate empirically how to control the trade-off between stability and performance in deep learning in Section \ref{sec:stabilising_training}.  We do so using DAL (Drift Adjusted Learning), an approach to setting the learning rate dynamically based on insights on instability derived from the PF.}
    \item \rebuttalrthree{We end by showcasing the potential of integrating our continuous-time approach with other optimisation schemes and highlighting how the PF can be used as a tool for existing continuous-time analyses in Section \ref{sec:future_work}}.
\end{itemize}

\textbf{Notation}: We denote as $E$ the loss function, $\vtheta$ the parameter vector of dimension $D$, $\nabla_{\vtheta}^2 E \in \mathbb{R}^{D \times D}$ the loss Hessian and $\lambda_i$ the Hessian's $i$th largest eigenvalue with $\vu_i$ the corresponding eigenvector. Since if $\vu_i$ is an eigenvector of $\nabla_{\vtheta}^2 E$ so is $-\vu_i$, we always use $\vu_i$ such that $\Re[\nabla_{\vtheta} E^T \vu_i] \ge 0$; this has no effect on our results and is only used for convenience and we used the notation $\Re(z)$ to be the real part of $z$. For a continuous-time flow $\vtheta(h)$ refers to the solution of the flow at time $h$.

\section{Continuous time models of gradient descent}
\label{sec:motivation}

The aim of this work is to understand the dynamics of gradient descent updates with learning rate $h$
\begin{align}
    \vtheta_t = \vtheta_{t-1} - h \nabla_{\vtheta} E(\vtheta_{t-1})
    \label{eq:gd-basic}
\end{align}
from the perspective of continuous dynamics. 
When using continuous-time dynamics to understand gradient descent it is most common to use \textit{the negative gradient flow} (NGF) 
\begin{align}
\dot{\vtheta} = - \nabla_{\vtheta} E.
\label{eq:ngf}
\end{align}
Gradient descent can be obtained from the NGF through Euler numerical integration, with an error of $\mathcal{O}(h^2)$ after one gradient descent step, see Section~\ref{sec:opt_algo}. 
\rebuttalrthree{
Studying gradient descent and its behaviour around equilibria and beyond has thus taken two main approaches: directly studying the discrete updates of Eq~\eqref{eq:gd-basic} \citep{bartlett2018gradient,bartlett2018representing,mescheder2017numerics,gunasekar2018implicit,du2019gradient,allen2019convergence,du2019width,ziyin2021sgd,liu2021noise}, or the continuous-time NGF of Eq~\eqref{eq:ngf} \citep{glendinning1994stability,saxe2013exact,nagarajan2017gradient,lampinen2018analytic,arora2018optimization,advani2020high,elkabetz2021continuous,vardi2021implicit,franca2020,balduzzi2018mechanics}.
The appeal of continuous-time systems lies in their connection with dynamical systems and the plethora of tools that thus become available, such as stability analysis; the simplicity by which conserved quantities can be obtained \citep{du2018algorithmic,franca2020}; and analogies that can be constructed through similarities with physical systems \citep{franca2020}. Because of the availability of tools for the analysis of continuous-time systems, it has been previously noted that discrete-time approaches are often more challenging and discrete-time proofs are often inspired from continuous-time ones \citep{may1976simple,elkabetz2021continuous}. We use an example to showcase the ease of continuous-time analyses: when following the NGF the loss $E$ decreases since $\frac{dE}{dt} = - ||\nabla_{\vtheta}{E}||^2$, as we derived in Eq~\ref{eq:e_min_ngf}.  Showing that and \textit{when} following the discrete-time gradient descent update in Eq~\eqref{eq:gd-basic} is more challenging and requires adapting the analysis on the form of the loss function $E$. Classical convergence guarantees associated with other optimisation approaches, such as natural gradient, are also derived in continuous-time \citep{amari1998natural,ollivier2015riemannian,ollivier2015riemannian2}.
By analysing the properties of continuous-time systems one can also determine whether optimisers should more closely follow the underlying continuous-time flow \citep{song2018accelerating,odegan}, what regularisers should be constructed to ensure convergence or stability \citep{nagarajan2017gradient,balduzzi2018mechanics,rosca2021discretisation}, construct converge guarantees in functional space for infinitely wide networks \citep{jacot2018neural,lee2019wide}.
}

\subsection{\rebuttalrthree{Limitations of existing continuous-time flows}}

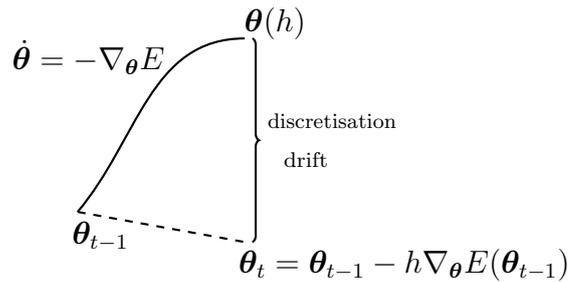
\begin{figure}[t]
\centering
\begin{tikzpicture}[every text node part/.style={align=center,inner sep=0,outer sep=0}][overlay]
\coordinate (theta_t_minus_1) at (0,1);
\coordinate (theta_t) at (2.2,0.6);

\coordinate (cont_theta_t) at (2.2,3.3);

\node(draw) at ($(theta_t_minus_1) + (+0.3,-0.32)$) {$\vtheta_{t-1}$};
\node(draw) at ($(theta_t) + (2.1,-0.3)$) {$\vtheta_{t} = \vtheta_{t-1} - h \nabla_{\vtheta} E(\vtheta_{t-1})$};

\coordinate (first_time_transition) at ($(theta_t_minus_1) + (+0.55,1.6)$);

\draw [thick,dashed] (theta_t_minus_1) -- (theta_t);

\draw [thick]  (theta_t_minus_1) to[out=50,in=180]  node[near end,above,yshift=-0.4cm, xshift=-1.3cm] {$\dot{\vtheta} = -\nabla_{\vtheta} E$} (cont_theta_t);

\node(draw) at ($(cont_theta_t) + (+0.4,0.2)$) {$\vtheta(h)$};

\draw [
    thick,
    decoration={
        brace,
        mirror,
        raise=0.1cm
    },
    decorate
] (theta_t) -- (cont_theta_t)
node [pos=0.5,anchor=west,xshift=0.15cm,yshift=-0.cm,text width=1cm,align=left] { \scriptsize {\color{black}discretisation \protect\newline drift}};
\end{tikzpicture}
\caption[Defining discretisation drift.]{\textbf{Discretisation drift}. Using continuous-time flows to understand gradient descent is limited by the gap between the discrete and continuous dynamics. In the case of the negative gradient flow $\dot{\vtheta} = -\nabla_{\vtheta} E$, we call this gap \textit{discretisation drift}. Since the negative gradient flow always minimises $E$, instabilities in gradient descent---areas where the loss $E$ increases---are due to discretisation drift.}
\label{fig:dd_def}
\end{figure}

The well-known discrepancy between Euler integration and the NGF, often called \textit{discretisation error} or \textit{discretisation drift} (Figure~\ref{fig:dd_def}) leads to certain limitations when using the NGF to describe gradient descent: the NGF cannot explain divergence around a local minimum for high learning rates or convergence to flat minima as often seen in the training of neural networks.  
Critically, since the NGF does not depend on the learning rate, it cannot explain any learning rate dependent behaviour.

The appeal of continuous-time methods together with the limitations of the NGF have inspired the machine learning community to look for other continuous-time systems that may better approximate the gradient descent trajectory.  One approach to constructing continuous-time flows approximating gradient descent that takes into account the learning rate is backward error analysis (BEA); for an overview of BEA, see Section~\ref{sec:bea}. Using this approach, \citet{igr} introduce the Implicit Gradient Regularisation flow (IGR flow):
\begin{align} 
\dot{\vtheta} = -\nabla_{\vtheta}E   -\frac{h}{2} \nabla_{\vtheta}^2 E \nabla_{\vtheta} E,
\label{eq:modified_flow_igr}
\end{align}
which tracks the dynamics of the gradient descent step $\vtheta_t = \vtheta_{t-1} - h \nabla_{\vtheta} E(\vtheta_{t-1})$ with an error of $\mathcal{O}(h^3)$, thus reducing the order of the error compared to the NGF.
Unlike the NGF flow, the IGR flow depends on the learning rate $h$. This dependence explains certain properties of gradient descent, such as avoiding trajectories with high gradient norm; the authors connect this behaviour to convergence to flat local minima.

\begin{figure}[t]
\centering
\hspace{1em}
\begin{subfloat}[2D convex case.]{
 \includegraphics[width=0.45\columnwidth]{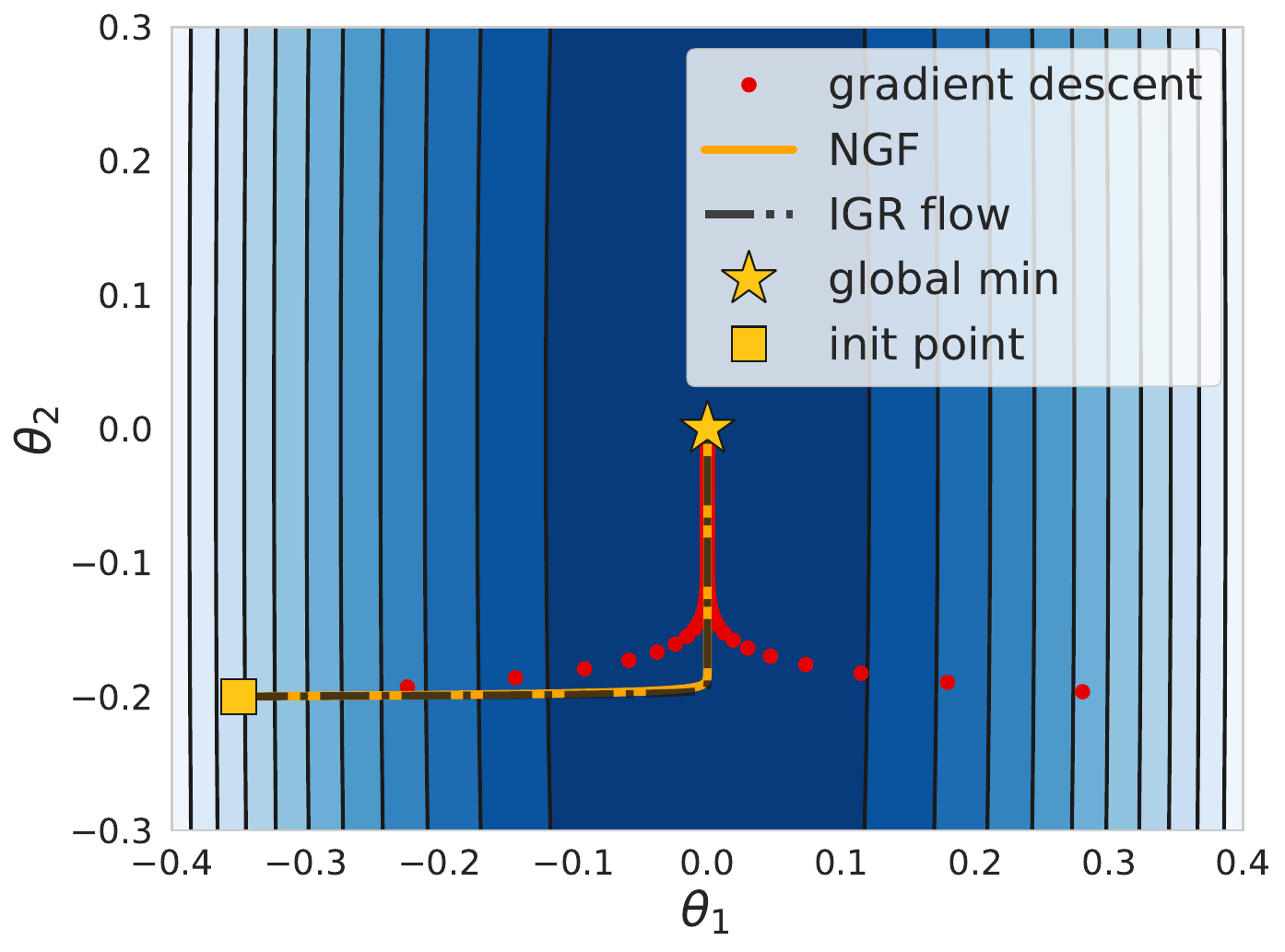}
 }\end{subfloat}%
 \begin{subfloat}[Banana function.]{
  \includegraphics[width=0.45\columnwidth]{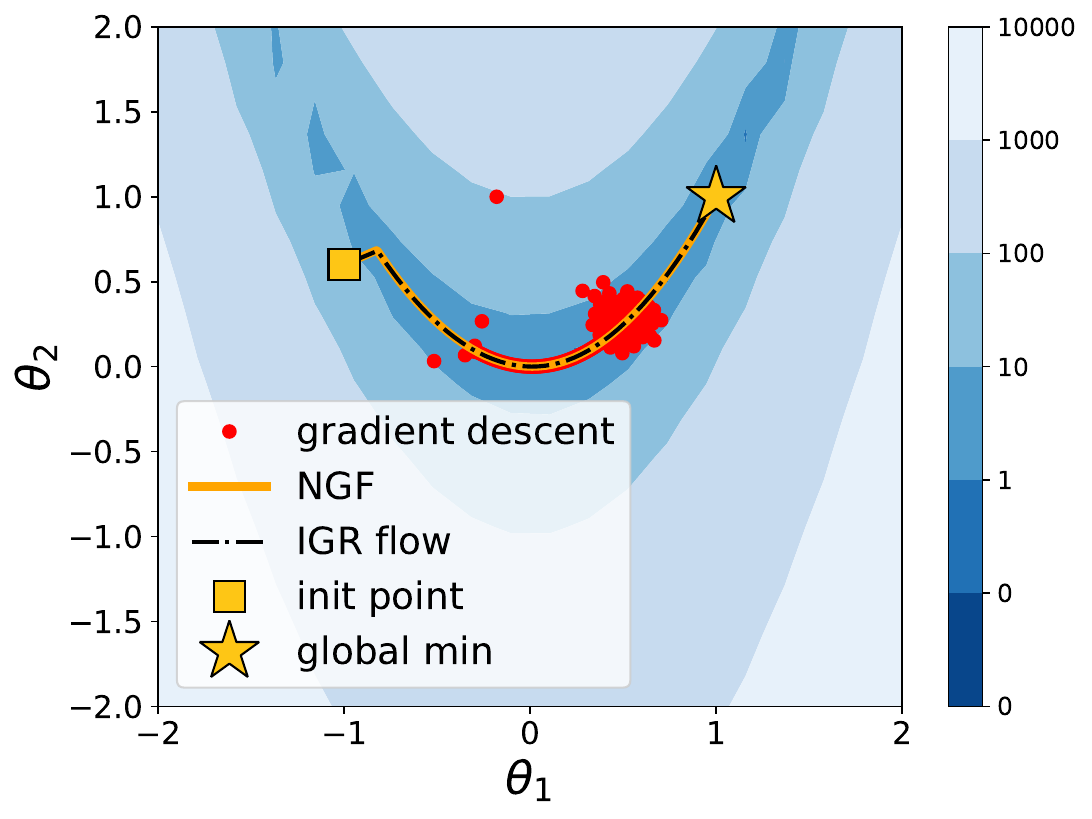}
 }\end{subfloat}
\caption[Motivation for the Principal Flow (PF): existing flows fail to capture the oscillatory or unstable behaviour of gradient descent.]{\textbf{Motivation}. Existing continuous-time flows fail to capture the oscillatory or unstable behaviour of gradient descent.}
\label{fig:motivation_small_examples}
\end{figure}

Like the NGF flow, however, the IGR flow does not explain the instabilities of gradient descent, as we illustrate in Figure~\ref{fig:motivation_small_examples}. Indeed,~\citet{igr} (their Remark 3.4) show that performing stability analysis around local minima using the IGR flow does not lead to qualitatively different conclusions from those using the NGF: both NGF and the IGR flow predict gradient descent to be always locally attractive around a local minimum (proofs in Section~\ref{sec:jacobian_igr_ngf}), contradicting the empirically observed behaviour of gradient descent.

\begin{figure}[tb]
\begin{subfloat}[Real flows.]{
  \includegraphics[width=0.44\columnwidth]{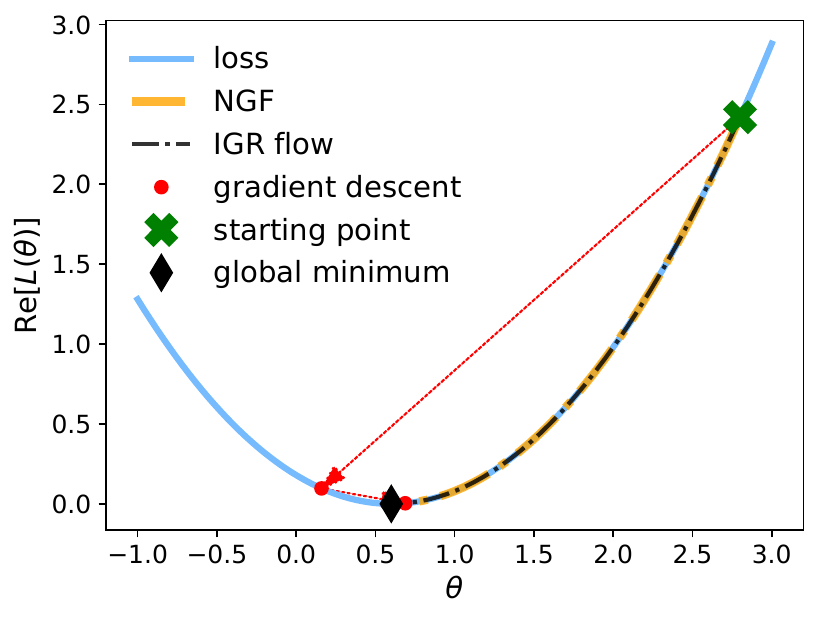}
  \label{fig:intuition_real}
 }\end{subfloat}
\begin{subfloat}[Complex flow.]{
 \includegraphics[width=0.45\columnwidth]{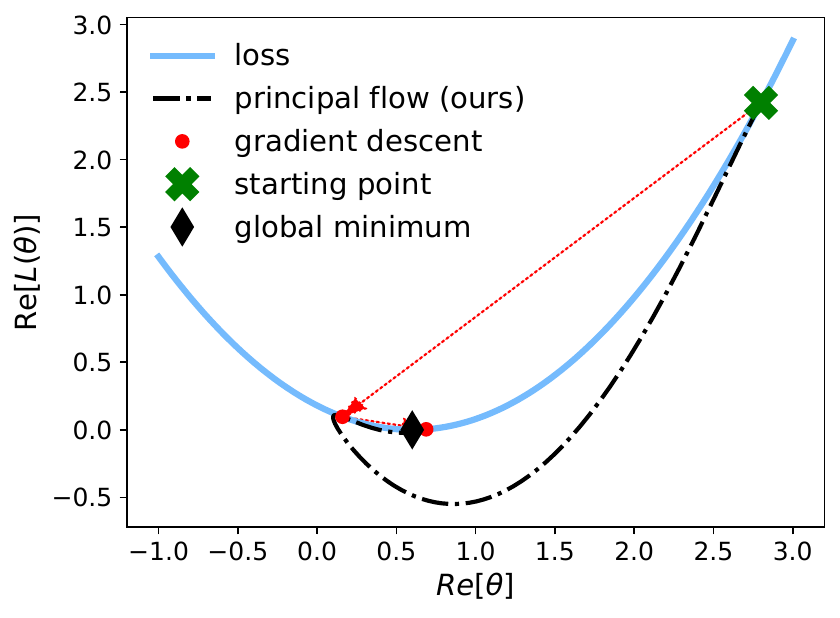}
  \label{fig:intuition_complex}
 }\end{subfloat}
\caption[Complex flows can capture oscillations and divergence of gradient descent around local minima.]{\textbf{Complex flows capture oscillations and divergence around local minima}. In the real space, the trajectory going from the starting point to the second gradient descent iterate goes through the global minima, and real flows stop there. In complex space however, that need not be the case.}
\label{fig:intuition_convex1d_complex_part_needed}
\end{figure}

\begin{figure}[th!]
\centering
\begin{subfloat}[MNIST.]{
 \includegraphics[width=0.41\columnwidth]{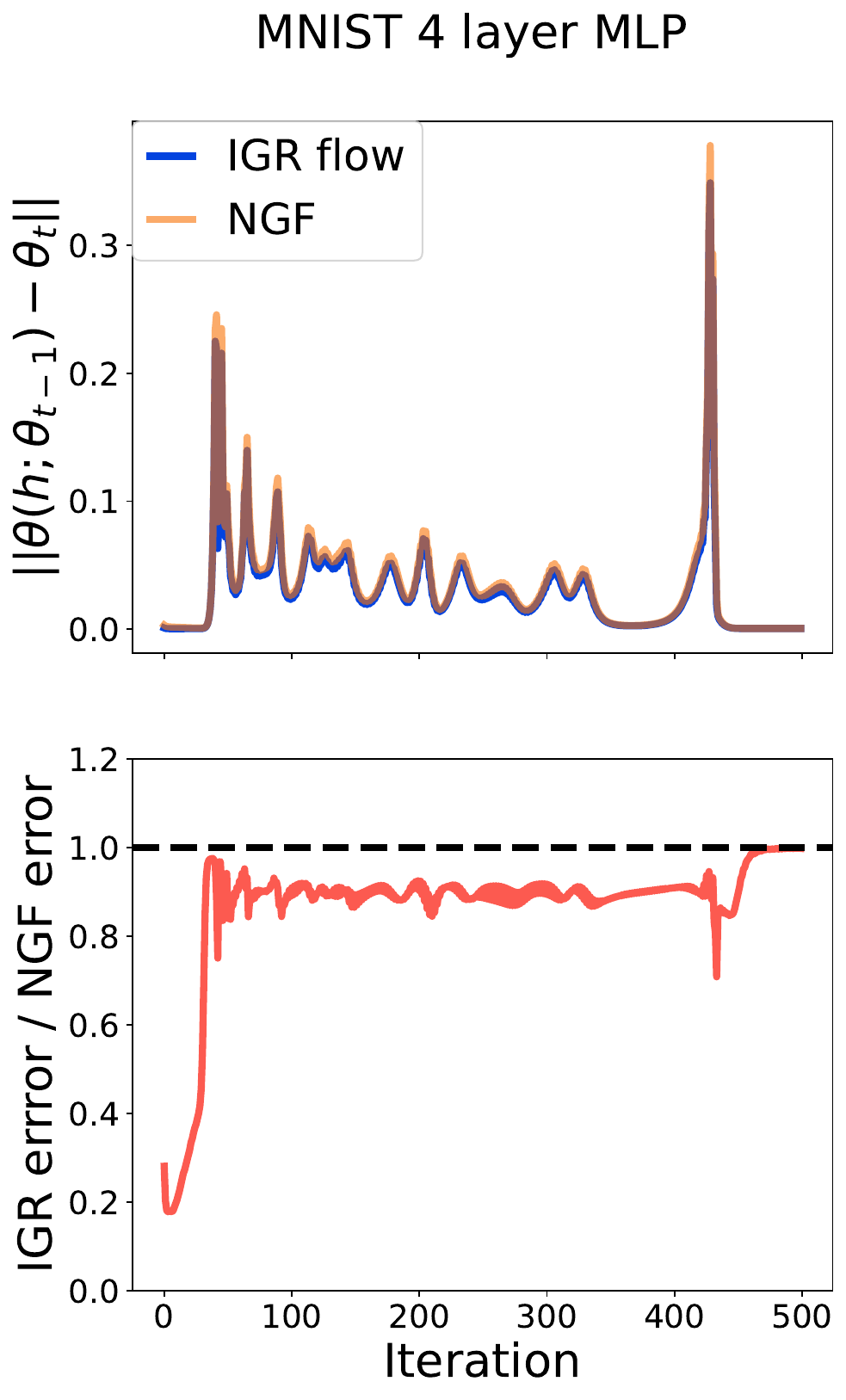}%
}\end{subfloat}\hspace{2em}
\begin{subfloat}[CIFAR-10.]{
 \includegraphics[width=0.42\columnwidth]{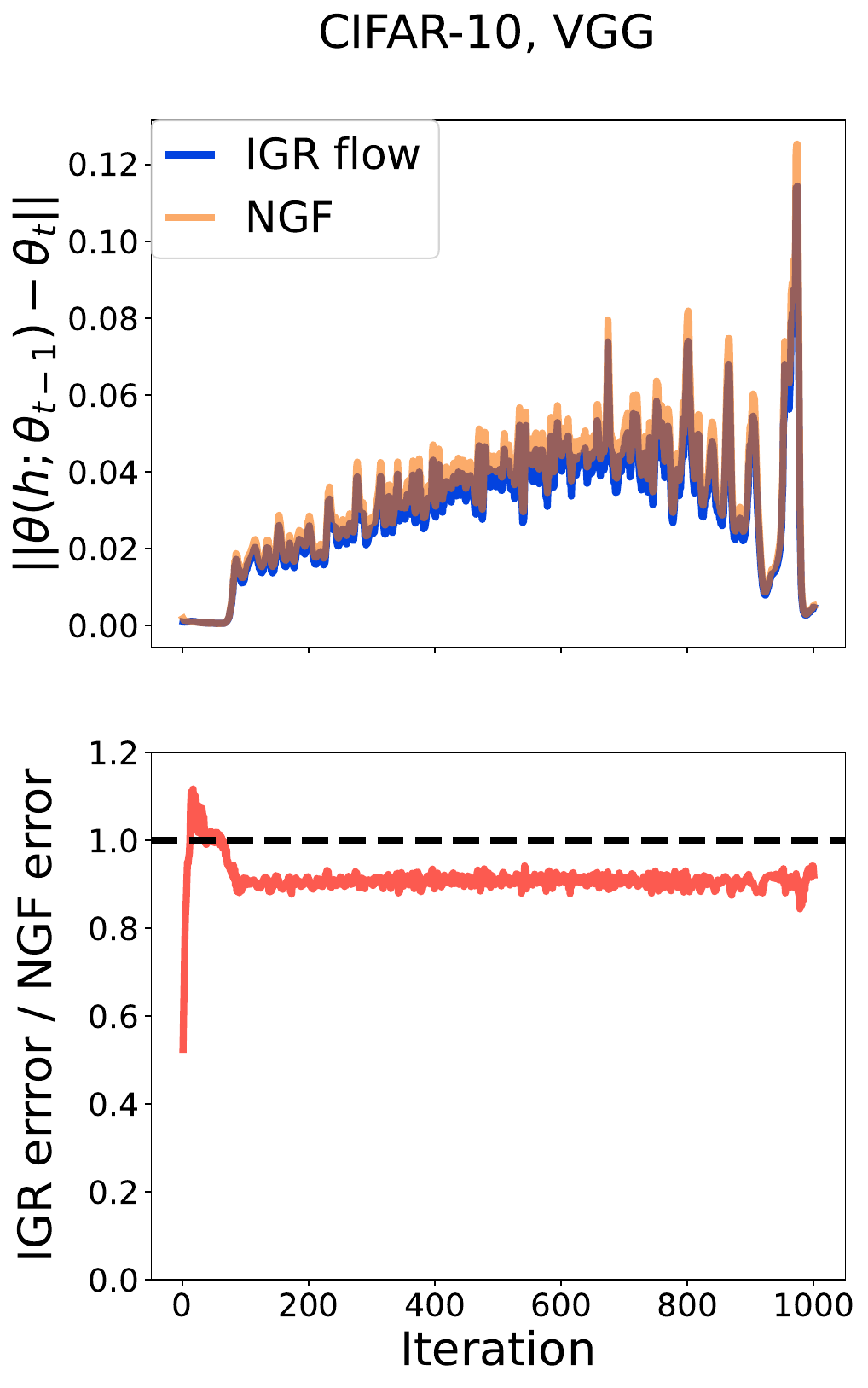}%
}\end{subfloat}
\caption[Per-iteration error between the NGF, IGR and gradient descent.]{\textbf{Motivation}. While the IGR flow captures aspects of the per-iteration discretisation drift for neural networks trained with gradient descent, a significant gap remains. To measure how well each flow captures gradient descent dynamics at one iteration, we train a model using gradient descent $\vtheta_t = \vtheta_{t-1} - h \nabla_{\vtheta} E(\vtheta_{t-1})$ and at each iteration $t$ we approximate the respective flows for time $h$. We then compute the difference in norms between the resulting flow parameters and the gradient descent parameters at the next iteration: $\norm{\vtheta_t - \vtheta(h)}$ with $\vtheta(0) = \vtheta_{t-1}$, which we report in the top row. The bottom row reports the ratio between these error norms obtained with the IGR and NGF flow, in order to provide a relative error scale. We observe that in the two experiments we perform here, the IGR flow captures approximatively $10\%$ of discretisation drift at the majority of training iterations.}
\label{fig:validating_bea}
\end{figure}

To understand why both the NFG and the IGR flow cannot capture oscillations and divergence around a local minimum, we note that stationary points $\nabla_{\vtheta}E = \mathbf{0}$ are fixed points for both flows.
We visualise an example in Figure~\ref{fig:intuition_real}: since to go from the initial point to the gradient descent iterates requires passing through the local minimum,  both flows would stop at the local minimum and never reach the following gradient descent iterates. In the case of neural networks we show in Figure~\ref{fig:validating_bea} that while the IGR flow is better than the NGF at describing gradient descent, a substantial gap remains.

\rebuttalrthree{The lack of ability of existing continuous-time flows to model instabilities empirically observed in gradient descent such as those shown in Figure~\ref{fig:motivation_small_examples} has been used as a motivation to use discrete-time methods instead \citep{yaida2018fluctuation,liu2021noise}. The goal of our work is to overcome this issue by introducing a novel continuous-time flow that captures instabilities observed in gradient descent. To do so, we follow the footsteps of \citet{igr} and use BEA. By using a continuous-time flow we can leverage the tools and advantages of continuous-time methods discussed earlier in this section; by incorporating discretisation drift into our model of gradient descent we can increase their applicability to explain unstable training behaviour.}
\rebuttalrthree{Indeed, we show in Figure~\ref{fig:intuition_complex} that the flow we propose captures the training instabilities; a key reason why is that, unlike existing flows, it operates in complex space. In Section~\ref{sec:principal_flow}, we show the importance of operating in complex space in order to understand oscillatory and instability behaviours of gradient descent. }

\section{The principal flow}
\label{sec:principal_flow}

In the previous section, we have seen how BEA can be used to define continuous-time flows that capture the dynamics of gradient descent up to a certain order in learning rate.
We have also explored the limitations of these flows, including the lack of ability to explain oscillations observed empirically when using gradient descent.
To further expand our understanding of gradient descent via continuous-time methods, we would like to get an intuition for the structure of higher order modified vector fields provided by BEA. We start with the following modified vector field, which we will call \textit{the third-order flow} (proof in Section~\ref{sec:third_order_flow_proof}):
\begin{align}
    \dot{\vtheta} = -\nabla_{\vtheta}E   -\frac{h}{2} \nabla_{\vtheta}^2E  \nabla_{\vtheta}E - h^2 \left( \frac{1}{3}  (\nabla_{\vtheta}^2E)^2 \nabla_{\vtheta} E + \frac {1} {12} \nabla_{\vtheta}E^T (\nabla_{\vtheta}^3E) \nabla_{\vtheta}E\right),
\label{eq:third_order_modified_vector_field}
\end{align}
with $\nabla_{\vtheta}^3E \in \mathbb{R}^{D \times D \times D}$ and $(\nabla_{\vtheta}E^T (\nabla_{\vtheta}^3E) \nabla_{\vtheta}E)_{k} = \sum_{i,j}{(\nabla_{\vtheta}E)_i} (\nabla_{\vtheta}^3E)_{i,k,j} {(\nabla_{\vtheta}E)_j}$.
The third-order flow tracks the dynamics of the gradient descent step ${\vtheta_t = \vtheta_{t-1} - h \nabla_{\vtheta} E(\vtheta_{t-1})}$ with an error of $\mathcal{O}(h^4)$, thus further reducing the order of the error compared to the IGR flow.
Like the IGR flow and the NGF, the third-order flow has the property that $ \dot{\vtheta} = \mathbf{0}$ if $\nabla_{\vtheta} E = \mathbf{0}$
and thus will exhibit the same limitations observed in Figure~\ref{fig:intuition_convex1d_complex_part_needed}.
The third-order flow allows us to spot a pattern:
the correction term of order $\mathcal{O}(h^n)$ in 
the BEA modified flow describing gradient descent contains the term $(\nabla_{\vtheta}^2E)^{n} \nabla_{\vtheta} E$ and terms that contain higher order derivatives with respect to parameters, terms which we will denote as $\mathcal{C}(\nabla_{\vtheta}^3 E)$. 

\rebuttalrtwo{\textbf{Our approach}}.
We will use the terms of the form $(\nabla_{\vtheta}^2E)^{n} \nabla_{\vtheta} E$ to construct a new continuous-time flow.
We will take a three-step approach. First, for an arbitrary order $\mathcal{O}(h^n)$ we will find the terms containing only first and second order derivatives in the modified vector field given by BEA and show they are of the form $(\nabla_{\vtheta}^2 E)^n \nabla_{\vtheta} E$ (Theorem \ref{thm:order_n_flow}). Second, we will use all orders to create a series (Corollary \ref{col:principal_series}). Third, we will use the series to find the modified flow given by BEA (Theorem \ref{thm:principal_ode}). All proofs are provided in Section~\ref{sec:all_proofs} of the Appendix.

\begin{theorem}
The modified vector field with an error of order $\mathcal{O}(h^{n+2})$ to the gradient descent update  $\vtheta_t = \vtheta_{t-1} -h \nabla_{\vtheta}E(\vtheta_{t-1})$
has the form:
\begin{align}
    \dot{\vtheta} = \sum_{p=0}^{n} \frac{-1}{p+1} h^p (\nabla_{\vtheta}^2 E)^p \nabla_{\vtheta} E + \mathcal{C}(\nabla_{\vtheta}^3 E),
\end{align}
\label{thm:order_n_flow}
where $\mathcal{C}(\nabla_{\vtheta}^3 E)$ denotes the family of functions that can be written as a sum of terms, each term containing a derivative of higher order than 3 with respect to parameters.
\end{theorem}

The result is proven by induction, with the full proof provided in Section~\ref{sec:higher_order_proofs} of the Appendix. The base cases for $n= 1, 2$, and $3$ follow from the NGF, IGR, and third-order flows. For higher order terms, the proof uses induction to find the term in $f_i$ depending on $\nabla_{\vtheta}^2 E $ and $\nabla_{\vtheta} E$ only and follows the BEA proof structure highlighted in Section \ref{sec:bea}, but Step 3 is modified to not account for terms in $\mathcal{C}(\nabla_{\vtheta}^3 E)$.  From the above, we can obtain the following corollary by using all orders $n$ and the eigen-decomposition of $\nabla_{\vtheta}^2 E$:
\begin{corollary}
\rebuttalrtwo{The full order modified flow obtained by performing BEA on gradient descent updates is of the form:}
\begin{align}
  \dot{\vtheta} 
   &=  \sum_{p=0}^{\infty} \frac{-1}{p+1} h^p (\nabla_{\vtheta}^2 E)^p \nabla_{\vtheta} E + \mathcal{C}(\nabla_{\vtheta}^3 E) \\
   &=  \sum_{p=0}^{\infty} \frac{-1}{p+1} h^p \left( \sum_{i=0}^{D-1} \lambda_i^p \vu_i \vu_i^T\right) \nabla_{\vtheta} E + \mathcal{C}(\nabla_{\vtheta}^3 E)\\
   &=  \sum_{i=0}^{D-1} \left(\sum_{p=0}^{\infty} \frac{-1}{p+1} h^p  \lambda_i^p  \right) (\nabla_{\vtheta} E^T \vu_i) \vu_i  + \mathcal{C}(\nabla_{\vtheta}^3 E),
\label{eq:principal_series}
\end{align}
where $\lambda_i$ and $\vu_i$ are the respective eigenvalues and eigenvectors of the Hessian $\nabla_{\vtheta}^2 E$.
\label{col:principal_series}
\end{corollary}

\rebuttalrtwo{If $\lambda_0 > 1/h$ the BEA series above diverges.
Generally BEA series are not convergent and approximate the discrete scheme only by truncation \citep{hairer2006geometric}.
When the series in Eq~\eqref{eq:principal_series} diverges, truncating it up to any order $n$, however, will result in a flow that will not be able to capture instabilities, even in the quadratic case. 
}
\rebuttalrtwo{Such flows (including the IGR flow) will always predict the loss function will decrease for a quadratic loss where a minimum exists, since:
$ {\frac{dE}{dt} = \nabla_{\vtheta}E ^T \left(\sum_{p=0}^{n} \frac{-1}{p+1} h^p (\nabla_{\vtheta}^2 E)^p \nabla_{\vtheta} E\right) = - \sum_{p=0}^{n} \frac{1}{p+1} h^p \sum_{i=0}^{D-1} (\lambda_i^p) (\nabla_{\vtheta} E ^T \vu_i)^2}$ which is never positive for any quadratic loss where a minimum exists (i.e. when $\lambda_i \ge 0, \forall i$). The above also entails that the flows always predict convergence around a local minimum, which is not the case for gradient descent which can diverge for large learning rates.}
\rebuttalrtwo{To further track instabilities we can use the BEA series to find the following flow:}
\begin{definition} We define the \textbf{principal flow} (PF) as
\begin{align}
  \dot{\vtheta} =  \sum_{i=0}^{D-1} \frac{\log(1 - h \lambda_i)}{h\lambda_i}(\nabla_{\vtheta} E^T \vu_i)  \vu_i.
\end{align}
\end{definition}
We note that $\lim_{\lambda \to 0} \frac{\log(1 - h \lambda)}{h\lambda} = -1$ and thus the PF is well defined when the Hessian $\nabla_{\vtheta}^2 E$ is not invertible. Unlike the NGF and the IGR flow, the modified vector field of the PF cannot be always written as the gradient of a loss function in $\mathbb{R}$, and can be complex valued.

\begin{theorem}
The Taylor expansion in $h$ at $h=0$ of the PF vector field coincides with the series coming from the BEA of gradient descent (Eq~\eqref{eq:principal_series}).
\label{thm:principal_ode}
\end{theorem}

\begin{proof}
Using the Taylor expansion at $z = 0$, $\textrm{Taylor}_{z=0} \frac{\log(1-z)}{z} = \sum_{p=0}^{\infty} \frac{-1}{p+1} z^p$ we obtain:
\begin{align}
\textrm{Taylor}_{h=0}  \sum_{i=0}^{D-1}  \frac{\log(1 - h \lambda_i)}{h\lambda_i} (\nabla_{\vtheta} E^T \vu_i) \vu_i = \sum_{i=0}^{D-1} \left(\sum_{p=0}^{\infty} \frac{-1}{p+1} h^p \lambda_i^p \right)(\nabla_{\vtheta} E^T \vu_i) \vu_i.
\end{align}
\end{proof}

We have used BEA to find the flow that when Taylor expanded at $h=0$ leads to the series in Eq~\eqref{eq:principal_series}.  
\rebuttalrtwo{When the BEA series in Eq~\eqref{eq:principal_series} converges, namely $\lambda_0 <1/h$, the PF and the flow given by the BEA series are the same. When $\lambda_0 > 1/h$, however, the PF is complex and the BEA series diverges. 
While in this case any BEA truncated flow will not be able to track gradient descent closely, we show that for quadratic losses the PF will track gradient descent exactly, and that it is a good model of gradient descent around fixed points. We show examples of the PF tracking gradient descent exactly in the quadratic case in Figures~\ref{fig:intuition_complex} and~\ref{fig:intuition_quadratic_2d}.}

\begin{remark} For quadratic losses of the form $E = \frac{1}{2}\vtheta^T \vA \vtheta + \vb^T \vtheta$, the PF captures gradient descent exactly. This case has been proven in \citet{hairer2006geometric}. The solution of the PF can also be computed exactly in terms of the eigenvalues of $\nabla_{\vtheta}^2 E$: $\vtheta(t) = \sum_{i=0}^{D-1} e^{\frac{\log(1 - h\lambda_i)}{h} t} \vtheta_0^T \vu_i \vu_i + t \sum_{i=0}^{D-1} \frac{\log (1 - h\lambda_i)}{h \lambda_i} b^T \vu_i$.
\label{remark:quadractic_remark}
\end{remark}

\begin{remark} In a small enough neighbourhood around a critical point (where higher-order derivatives can be ignored) the PF can be used to describe gradient descent dynamics closely. We show this also using a linearisation argument in Section~\ref{sec:linearisation} in the Appendix.
\end{remark}

\begin{definition} The terms $\mathcal{C}(\nabla_{\vtheta}^3 E)$ are called \textbf{non-principal terms}. The term $\frac {1} {12} \nabla_{\vtheta}E^T (\nabla_{\vtheta}^3E) \nabla_{\vtheta}E$ in Eq~\eqref{eq:third_order_modified_vector_field} is a non-principal term (we will call this term non-principal third-order term).
\end{definition}

\begin{definition} \label{def:non_princ} We define the \textbf{principal flow with third-order non- principal term} as
\begin{align}
  \dot{\vtheta} =  \sum_{i=0}^{D-1} \frac{\log(1 - h \lambda_i)}{h\lambda_i}(\nabla_{\vtheta} E^T \vu_i)  \vu_i
    - \underbrace{\frac{h^2}{12} \nabla_{\vtheta} E^T (\nabla^3_{\vtheta} E) \nabla_{\vtheta} E}_{\text{third-order non-principal term}}.
    \label{eq:pf_with_non_principal}
\end{align}
\end{definition}

\rebuttalrtwo{General theoretical bounds on the error between continuous time flows and gradient descent are challenging to construct in the case of a general parametrised $E(\vtheta)$ as the error will be determined by the shape of $E$. We know the conditions which determine when certain flows follow gradient descent exactly. The NGF and gradient descent will follow the same trajectory in areas where $\nabla^2_{\vtheta} E \nabla_{\vtheta} E = 0$ (see Theorem~\ref{thm:total_drift}) and thus $E$ has a constant gradient in time, since $\frac{d \nabla_{\vtheta} E}{d t} = \nabla^2_{\vtheta} E \nabla_{\vtheta} E$. The PF generalises the NGF, in that it follows the same trajectory as gradient descent not only for trajectories where  $\nabla^2_{\vtheta} E \nabla_{\vtheta} E = 0$, but also when $E$ is quadratic. Informally, we can state that the closer we are to these exact conditions, the more likely the flows are to capture the dynamics of gradient descent. Formally, bounds on the error between GD and NGF can be provided by the Fundamental Theorem (Theorem 10.6 in~\citet{wanner1996solving}) which has recently been adapted to a neural network parametrisation by~\citet{elkabetz2021continuous}; this bound depends on the magnitude of the smallest Hessian eigenvalue along the NGF trajectory. We hope that future work can expand the Fundamental Theorem such that error bounds between the PF and gradient descent can be constructed for deep neural networks. Here we take an empirical approach and show that although not exact outside the quadratic case the PF captures key features of the gradient descent dynamics in stable or unstable regions of training, around and outside critical points, for small examples or large neural networks.}

\subsection{The PF and the Hessian eigen-decomposition}
\label{sec:principal_flow_eigendecom}

\begin{table}[tb!]
\centering
\begin{tabular}{ c | l}
 \hline \hline
 \textbf{NGF} & $\dot{\vtheta} =  \sum_{i=0}^{D-1} \underbrace{- }_{\alpha_{NGF}(h \lambda_i) = -1} (\nabla_{\vtheta} E^T \vu_i)\vu_i$ \\
 \hline \hline
 \textbf{IGR Flow} & $\dot{\vtheta} =  \sum_{i=0}^{D-1} \underbrace{- (1 +\frac{h}{2} \lambda_i) }_{\alpha_{IGR}(h \lambda_i) = - (1 +\frac{h}{2} \lambda_i)}(\nabla_{\vtheta} E^T \vu_i) \vu_i$ \\
 \hline \hline 
 \textbf{PF} &  $\dot{\vtheta} =  \sum_{i=0}^{D-1} \underbrace{\frac{\log(1 - h \lambda_i)}{h\lambda_i}}_{\alpha_{PF}(h \lambda_i) = \frac{\log(1 - h \lambda_i)}{h\lambda_i}}  (\nabla_{\vtheta} E^T \vu_i) \vu_i$  
\end{tabular}
\caption[Writing the continuous-time flows discussed in terms of the eigen-decomposition of the Hessian.]{Understanding the differences between the flows discussed in terms of the eigen-decomposition of the Hessian. All flows have the form $\dot{\vtheta} = \sum_{i=0}^{D-1} \alpha(h \lambda_i) (\nabla_{\vtheta} E^T \vu_i) \vu_i$ with different $\alpha$ summarised here.}
\label{tab:principal_flow_vs_negative_grad_flow} 
\end{table}

All flows considered here have the form form $\dot{\vtheta} = \sum_{i=0}^{D-1} \alpha(h \lambda_i) (\nabla_{\vtheta} E^T \vu_i) \vu_i$, where $\alpha$ is a function computing the corresponding coefficient; we will denote the one associated with each flow as $\alpha_{NGF}$, $\alpha_{IGR}$ and $\alpha_{PF}$ respectively.
For a side-by-side comparison between the NGF, IGR flow, and the PF as functions of the Hessian eigen-decomposition see Table~\ref{tab:principal_flow_vs_negative_grad_flow}. 
 Since $\nabla_{\vtheta}E^T \vu_i \ge 0$, the $\alpha$ function determines the sign of a modified vector field in the direction $\vu_i$. For brevity it will be useful to define the coefficient of $\vu_i$ in the vector field of the PF:

\begin{definition} We call $sc_i = \frac{\log(1 - h \lambda_i)}{h\lambda_i}(\nabla_{\vtheta} E^T \vu_i) = \alpha_{PF}(h \lambda_i ) \nabla_{\vtheta} E^T \vu_i$ the \textbf{stability coefficient} for eigendirection $i$. $\sign(sc_i) = \sign (\alpha_{PF}(h \lambda_i))$.
\label{def:stab_coeff}
\end{definition}

\begin{figure}[tb!]
\begin{subfloat}[Real part.]{
 \includegraphics[width=0.45\columnwidth]{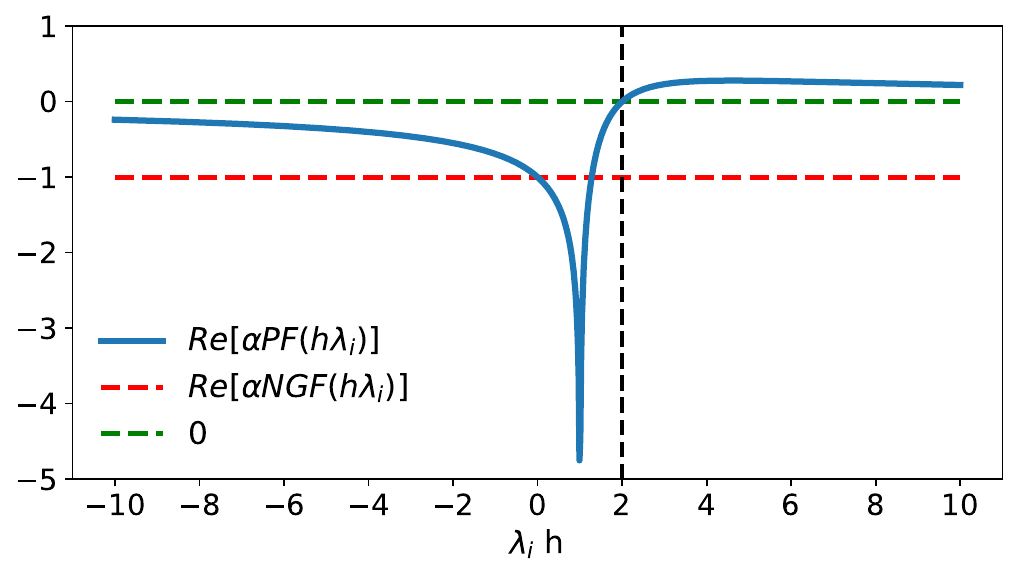}
 }\end{subfloat}
 \begin{subfloat}[Imaginary part.]{
 \includegraphics[width=0.45\columnwidth]{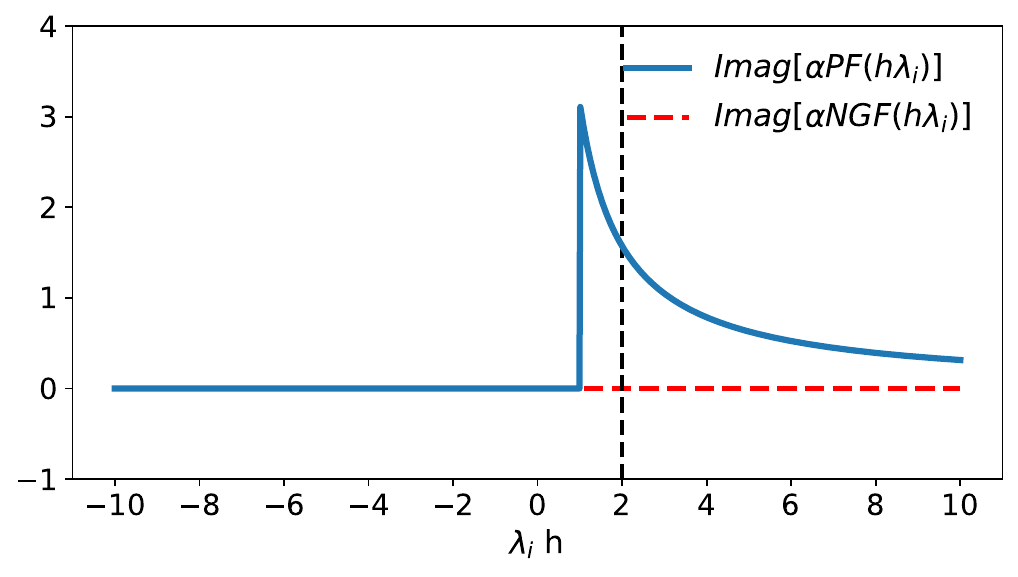}
 }\end{subfloat}
\caption{Comparing the coefficients $\alpha_{NGF}$ and $\alpha_{PF}$ based on the value of $\alpha h$. While $\alpha_{NGF}$ is always $-1$, the value of  $\alpha_{PF}$ depends on $\alpha h$ and can be complex, both with positive and negative real parts.}
\label{fig:r_plot}
\end{figure}

\begin{figure}[tb]
\captionsetup[subfigure]{justification=centering}
\begin{subfloat}[\hspace{2em} $\lambda_0< 1/h$, \newline \hspace{2em} $\alpha_{PF}(\lambda_0 h) < 0$]{
 \includegraphics[width=0.3\columnwidth]{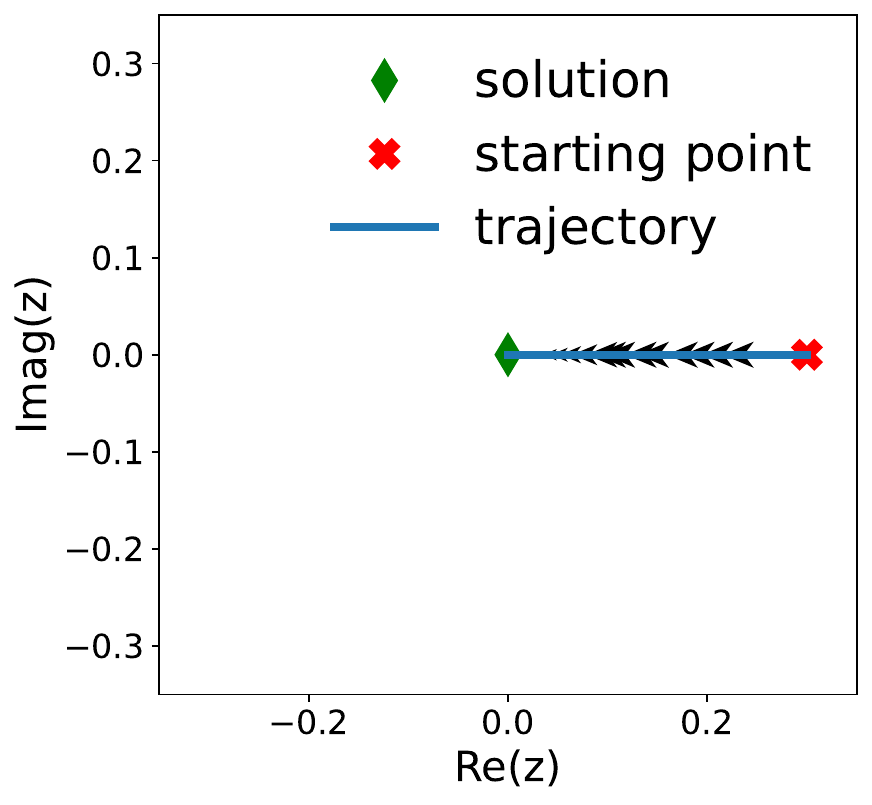}
 \label{fig:z_square_stable}
 }
\end{subfloat}%
\begin{subfloat}[ \hspace{1em} $1/h < \lambda_0 < 2/h, \newline \hspace{3em} \Re(\alpha_{PF}(\lambda_0 h)) < 0$ \hfill]{
 \includegraphics[width=0.3\columnwidth]{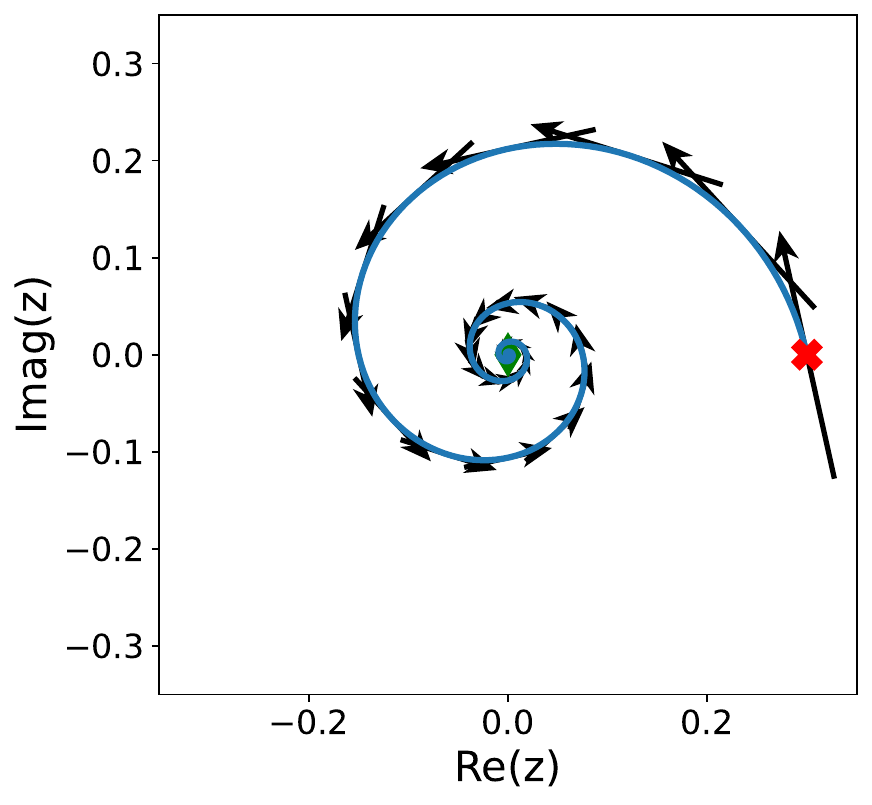}
 \label{fig:z_square_oscillation}
 } \end{subfloat}%
\begin{subfloat}[\hspace{2em} $2/h < \lambda_0, \hspace{0.5em} \newline \hspace{2em} \Re(\alpha_{PF}(\lambda_0 h)) > 0$]{
 \includegraphics[width=0.3\columnwidth]{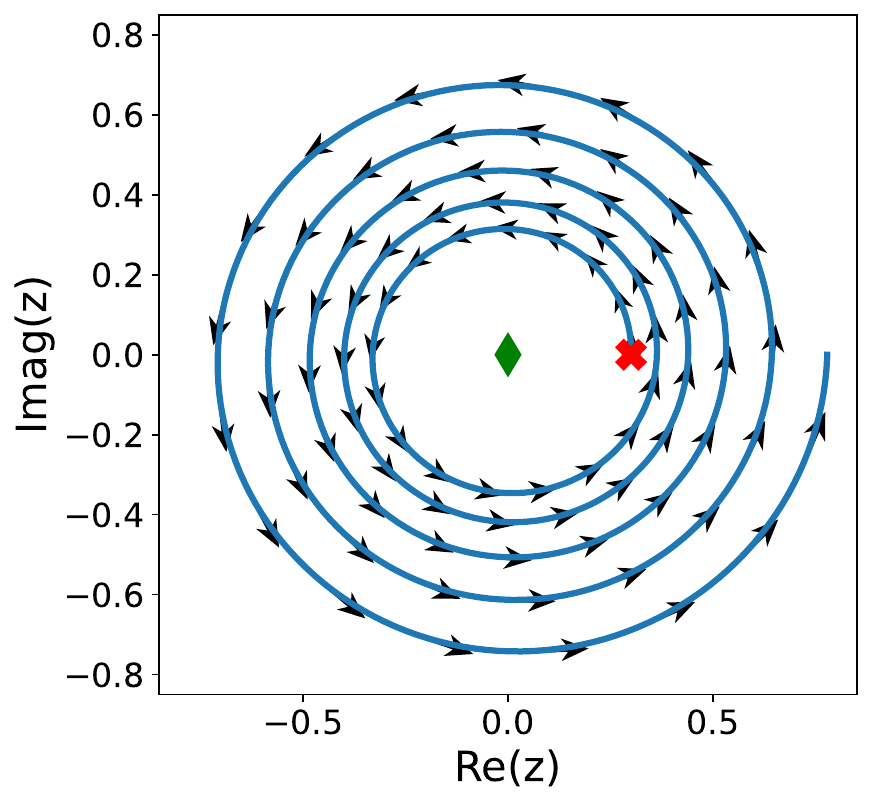}
 \label{fig:z_square_divergence}
 }\end{subfloat}
\caption[The behaviour of the PF on $E(z) = \frac 1 2 z^2$.]{The behaviour of PF on the quadratic function $E(z) = \frac 1 2 z^2$ with solution $z(t) = e^{\log(1 - h)/h}z(0)$. When $\lambda_0 < 1/h$,  $z(t) = (1-h)^{t/h} z(0)$ which is in real space and converges to the equilibrium. When $\lambda_0 > 1/h$, ${z(t) = (h-1)^{t/h} \left(\cos(\pi t/h) + i \sin(\pi t/h) \right)z(0)}$. This exhibits oscillatory behaviour, and when $\lambda_0 > 2/h$, diverges.}
\label{fig:z_square}
\end{figure}

In order to understand the PF and how it is different from the NGF we explore the change in each eigendirection $\vu_i$ and we perform case analysis on the relative value of the eigenvalues $\lambda_i$ and the learning rate $h$. To do so, we will compare $\alpha_{NGF}(h\lambda_i)$ and $\alpha_{PF}(h\lambda_i)$ since the sign of $\alpha_{NGF}(h\lambda_i)$ determines the direction that minimises $E$ given by $\vu_i$.
Since our goal is to understand the behaviour of gradient descent, we perform the case by case analysis of what happens at the start of a gradient descent iteration and thus use real values for $\lambda_i$ and $\vu_i$ even when the PF is complex valued.
We visualise $\alpha_{NGF}$ and $\alpha_{PF}$ in Figure~\ref{fig:r_plot} and we use Figure~\ref{fig:z_square} to show examples of each case using a simple function.

\noindent \textbf{Real stable case}: $\lambda_i< 1/h$. 
\begin{itemize}[topsep=0.1pt]
    \setlength\itemsep{0.05em}
    \item $\sign(\alpha_{NGF}(h\lambda_i)) = \sign(\alpha_{PF}(h\lambda_i)) = -1$.
    \item $\alpha_{NGF}(h\lambda_i) = -1$  and $\alpha_{PF}(h \lambda_i) = \frac{\log(1 - h \lambda_i)}{h \lambda_i} < 0$.
    \item The coefficients of the NGF and PF in eigendirection $\vu_i$ are negative and real.  
    \item This case is visualised in Figure~\ref{fig:z_square_stable}.
\end{itemize}

\noindent \textbf{Complex stable case}: $1/h < \lambda_i < 2/h$. 
\begin{itemize}[topsep=0.1pt]
    \setlength\itemsep{0.05em}
    \item $\sign(\alpha_{NGF}(h\lambda_i)) = \sign(\Re[\alpha_{PF}(h\lambda_i)]) = -1$. $\alpha_{PF}(h\lambda_i) \in \mathbb{C}$.
    \item $\alpha_{NGF}(h\lambda_i) = -1$ and $\alpha_{PF}(h \lambda_i) = \frac{\log(1 - h \lambda_i)}{h \lambda_i} = \frac{\log(-1 + h \lambda_i) + i \pi}{h\lambda_i} \in \mathbb{C}$ and  $\Re[\alpha_{PF}(h\lambda_i)] = \frac{\log(-1 + h \lambda_i)}{h\lambda_i} < 0$.
    \item  The real part of the coefficient of the NGF and PF in eigendirection $\vu_i$ are both negative. The imaginary part of $\alpha_{PF}$ can introduce instability and oscillations.
    \item This case is visualised in Figure~\ref{fig:z_square_oscillation}.
\end{itemize}

\noindent \textbf{Unstable complex case}: $ 2/h < \lambda_i$.
\begin{itemize}[topsep=0.1pt]
    \setlength\itemsep{0.05em}
    \item $\sign(\alpha_{NGF}(h\lambda_i)) \neq \sign(\Re[\alpha_{PF}(h\lambda_i)])$. $\alpha_{PF}(h\lambda_i) \in \mathbb{C}$. 
    \item $\alpha_{NGF}(h\lambda_i) = -1$ and $\alpha_{PF}(h \lambda_i) = \frac{\log(1 - h \lambda_i)}{h \lambda_i} = \frac{\log(-1 + h \lambda_i) + i \pi}{h\lambda_i} \in \mathbb{C}$ and  $\Re[\alpha_{PF}(h\lambda_i)] = \frac{\log(-1 + h \lambda_i)}{h\lambda_i} > 0$. 
    \item  The real part of the coefficient of the NGF in eigendirection $\vu_i$ is negative, while the real part of the coefficient of the PF is positive.
The PF goes in the opposite direction of the NGF which minimises $E$; this change in sign can cause instabilities.
 The imaginary component can still introduce oscillations, however, the larger $\lambda_i h$, the smaller the imaginary part of $\alpha_{PF}$.
 \item  This case is visualised in Figure~\ref{fig:z_square_divergence}.
\end{itemize}

\textbf{The importance of the largest eigenvalue $\lambda_0$}.
The largest eigenvalue $\lambda_0$ plays an important part in the PF. Since $h \lambda_0 \ge h \lambda_i \hspace{1em} \forall i$, $\lambda_0$ determines  where in the above cases the PF is situated and thus whether there are oscillations and unstable behaviour in training. 
For all flows of the form we consider we can write:
\begin{align}
\frac{d E(\vtheta)}{d t} = \nabla_{\vtheta} E^T \frac{d \vtheta}{dt} =  \nabla_{\vtheta} E^T \sum_{i=0}^{D-1} \alpha(h\lambda_i)  \nabla_{\vtheta} E^T \vu_i \vu_i =  \sum_{i=0}^{D-1} \alpha(h\lambda_i)  (\nabla_{\vtheta}E^T \vu_i)^2
\label{eq:changes_in_e}
\end{align}
and thus if $\alpha(h \lambda_i) \in \mathbb{R}$ and $\alpha(h \lambda_i) < 0$ $\forall i$ then $\frac{d E(\vtheta)}{d t} \le 0$ and following the corresponding flow minimises $E$.
In the case of the PF this gets determined by $\lambda_0$.
 If $\lambda_0 < \frac{1}{h}$ then $\alpha_{PF}(h\lambda_i) < 0 \; \forall i$  (real stable case above)  and the PF minimises $E$.
If  $1/h < \lambda_0 < \frac{2}{h}$ then $\Re[\alpha_{PF}(h\lambda_i)] < 0 \;\forall i$  (complex stable case above) close to a gradient descent iteration $\lambda_i, \vu_i \in \mathbb{R}$ we can write that $\frac{d \Re[E(\vtheta)]}{d t} = \sum_{i=0}^{D-1} \Re[\alpha_{PF}(h\lambda_i)]  (\nabla_{\vtheta}E^T \vu_i)^2$ and thus the real part of the loss function decreases. If $\lambda_0 > \frac{2}{h}$  then $ \Re[\alpha_{PF}(h\lambda_0)] > 0 $ (unstable complex case above) and if $(\nabla_{\vtheta}E^T \vu_0)^2$ is sufficiently large we can no longer ascertain the behaviour of $E$. We present a discrete-time argument for this observation in Section~\ref{sec:changes_in_loss_discrete}.

\begin{figure}[tb]
\captionsetup[subfigure]{justification=centering}
\begin{subfloat}[$\hspace{2em} \lambda_0 < 1/h$ \newline \hspace{4em}(stability)]{
 \includegraphics[width=0.33\columnwidth]{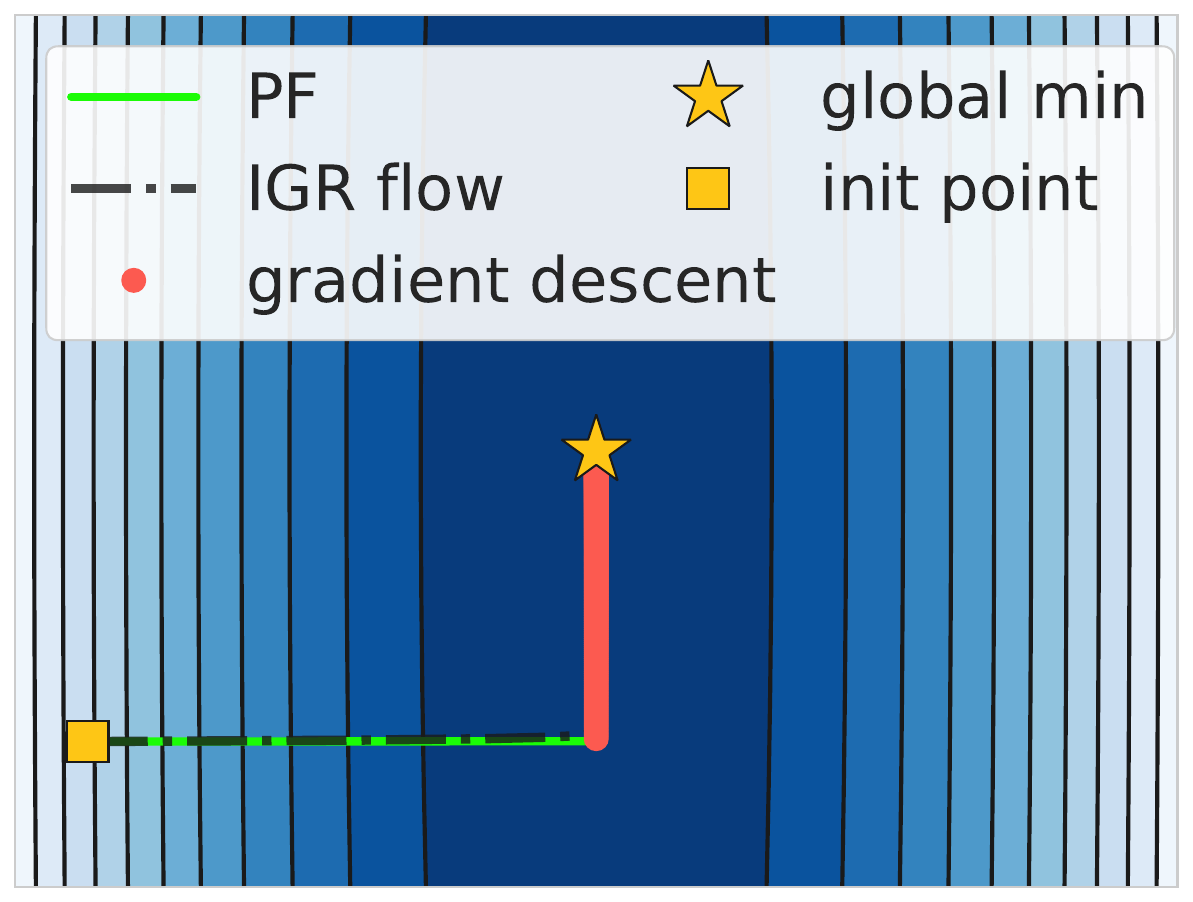}%
 }\end{subfloat}%
\begin{subfloat}[\hspace{2em} $1/h <\lambda_0 < 2/h$ \newline (oscillations)]{
 \includegraphics[width=0.33\columnwidth]{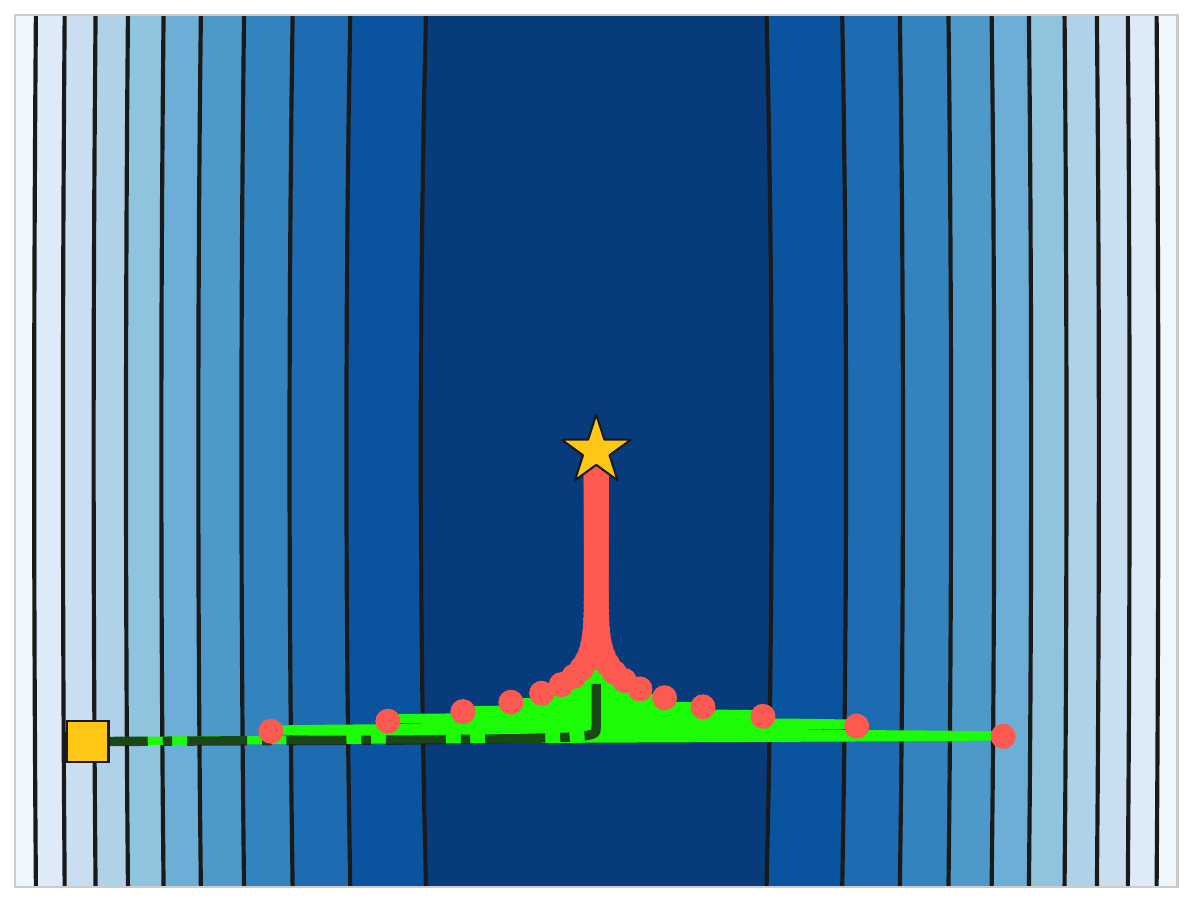}%
 \label{fig:2d_osc}
 }\end{subfloat}%
 \begin{subfloat}[\hspace{1.3em} $\lambda_0 > 2/h$ \newline \hspace{5em}(divergence)]{
 \includegraphics[width=0.33\columnwidth]{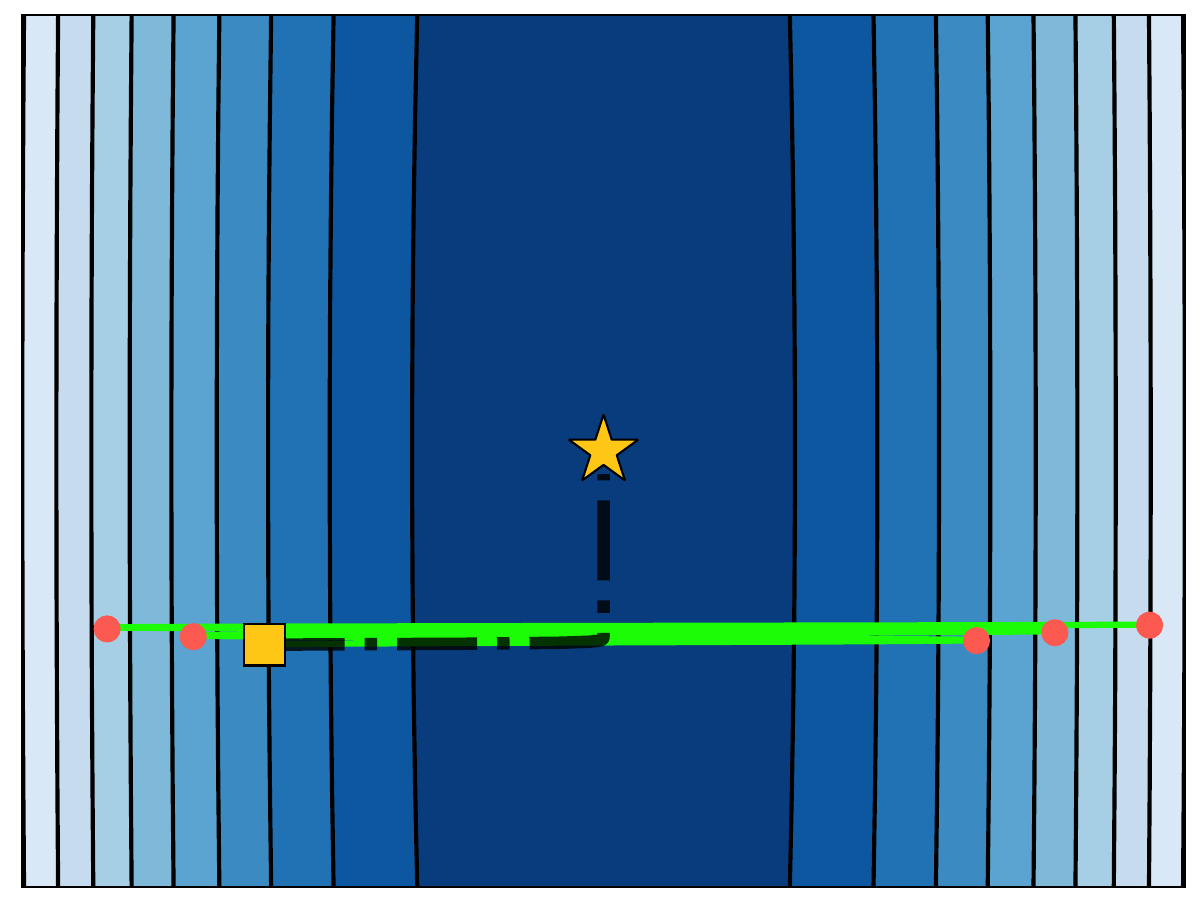}%
 \label{fig:2d_div}
 }\end{subfloat}
\caption[The PF captures the behaviour of gradient descent exactly for quadratic losses.]{\textbf{Quadratic losses in 2 dimensions}. The PF captures the behaviour of gradient descent exactly for quadratic losses, including oscillatory behaviour \subref{fig:2d_osc} and divergence \subref{fig:2d_div}. }
\label{fig:intuition_quadratic_2d}
\end{figure}

\begin{figure}[thb]
\centering
\begin{subfloat}[$\lambda_0 < 1/h$]{
 \includegraphics[width=0.291\columnwidth]{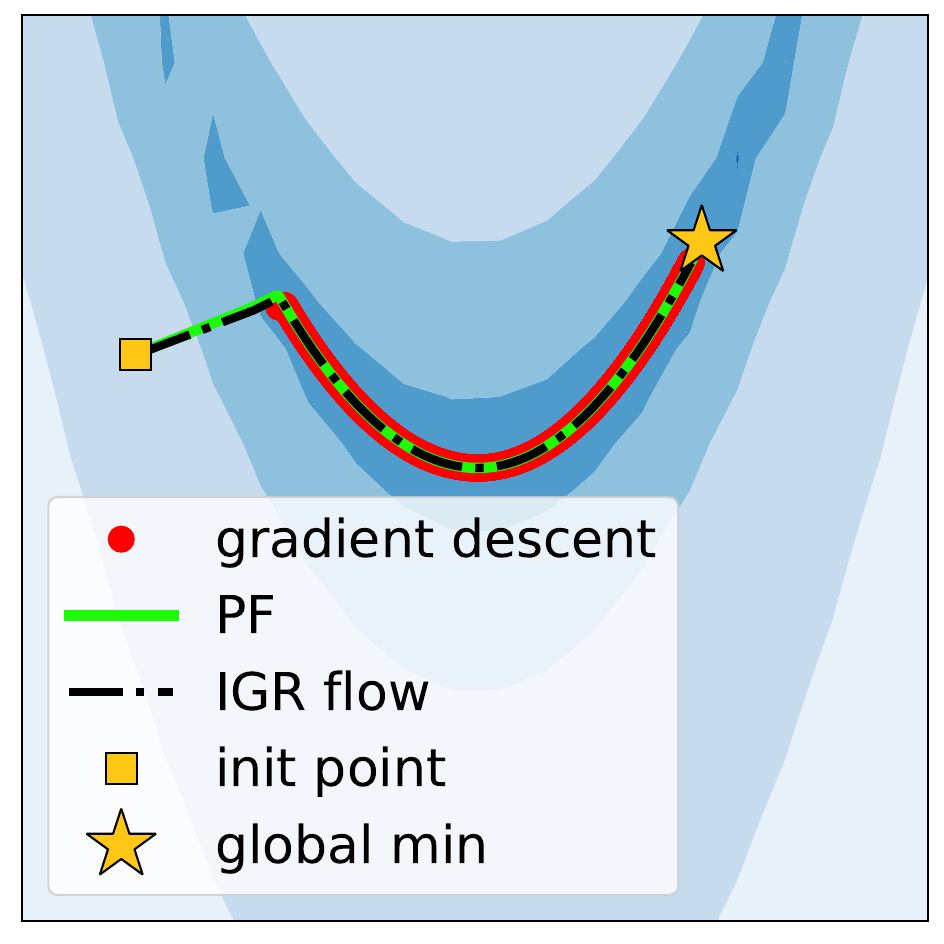}
 }\end{subfloat}%
 \begin{subfloat}[$\lambda_0 < 2/h (\lambda_0 \approx 1.9/h)$]{
 \includegraphics[width=0.291\columnwidth]{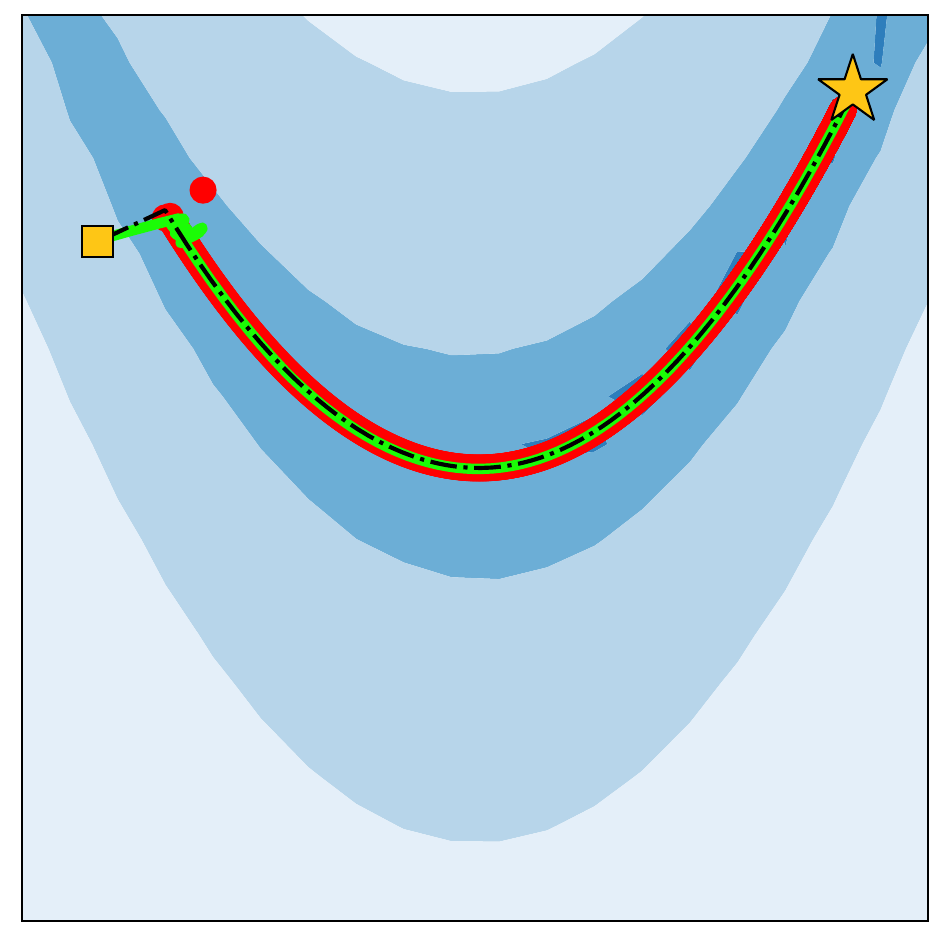}
 }\end{subfloat}%
\begin{subfloat}[$\lambda_0 >> 2/h (\lambda_0 \approx 5/h)$]{
 \includegraphics[width=0.36\columnwidth]{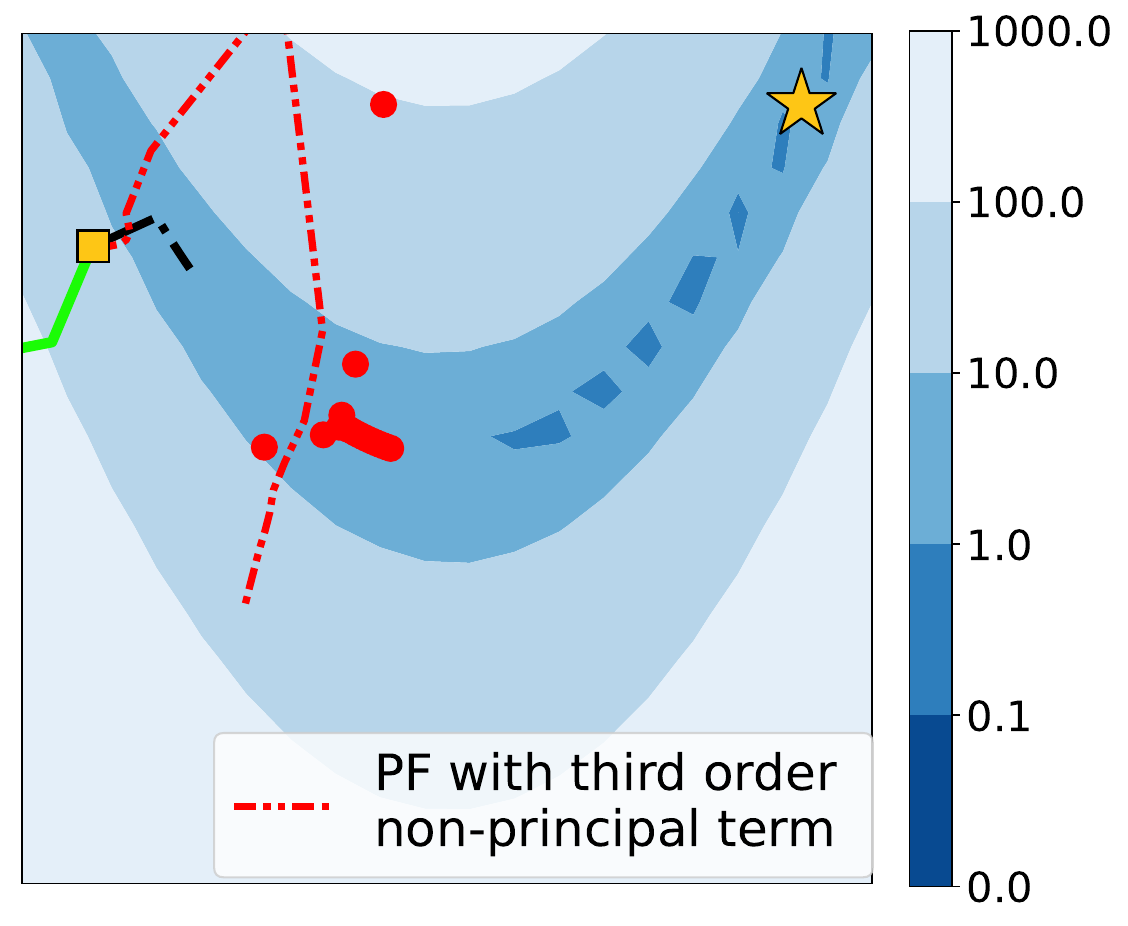}
 \label{fig:banana_unstable}
 }
 \end{subfloat}
\caption[The behaviour of the PF when the loss is the banana function.]{\textbf{Banana function}. The PF can capture instability and the gradient descent trajectory over many iterations  when $\lambda_0$ is close to $2/h$. When  $\lambda_0 >> 2/h$ \subref{fig:banana_unstable} the PF does not track the GD trajectory over many gradient descent steps, but when including a non-principal term the flow is able to capture the general trajectory of gradient descent and unstable behaviour of gradient descent.}
\label{fig:intuition_banana}
\end{figure}

\textbf{Building intuition}.
For a quadratic objective $E(\vtheta) = \frac{1}{2}\vtheta^T \vA \vtheta$ the  PF describes gradient descent exactly. We show examples Figures~\ref{fig:intuition_convex1d_complex_part_needed} and~\ref{fig:intuition_quadratic_2d}. 
Unlike the NGF or the IGR flow, the PF captures the oscillatory  and divergent behaviour of gradient descent. Importantly, to capture the unstable behaviour that occurs when $\lambda_0 > 1/h$ the imaginary part of the PF is needed.
To expand intuition outside the quadratic case, we show the PF for the banana function~\citep{rosenbrock1960automatic} in Figure~\ref{fig:intuition_banana} and an additional example in 1D with a non-quadratic function in Figure~\ref{fig:1d_cosine} in the Appendix. While in this case, the PF no longer follows the gradient descent trajectory exactly,  we still observe the importance of the PF in capturing instabilities of gradient descent. In Figure~\ref{fig:banana_unstable}, we also observe that adding non-principal terms---described in Definition~\ref{def:non_princ}---can restabilise the trajectory when  $\lambda_0 >> 2/h$.

\begin{remark} For the banana function, the principal terms have a destabilising effect when $h > 2/\lambda_0$ while the non-principal terms can have a stabilising effect.
\end{remark}

\subsection{The stability analysis of the PF}
\label{sec:pf_stability_analysis}

We now perform stability analysis on the PF, to understand how it can be used to predict certain behaviours of gradient descent around critical points of the loss function $E$; for an overview of stability analysis, see Section~\ref{sec:stability_analysis_overview}. Consider $\vtheta^*$ such a critical point, i.e. $\nabla_{\vtheta} E(\vtheta^*) = \mathbf{0}$. 
For a critical point $\vtheta^*$ to be exponentially asymptotically attractive, all eigenvalues of the Jacobian evaluated at $\vtheta^*$ need to have strictly negative real part.

The PF has the following Jacobian at critical points (proof in Section~\ref{sec:jacobian} in the Appendix):
\begin{align}
\vJ_{\text{PF}}(\vtheta^*) = \sum_{i=0}^{D-1} \frac{\log(1 - h \lambda_i^*)}{h} \vu_i^* {\vu_i^*}^T,
\label{eq:jac_pf_main}
\end{align}
where $\lambda_i^*$, $\vu_i^*$ are the eigenvalues and eigenvectors of the Hessian $\nabla_{\vtheta}^2E(\vtheta^*)$. 
We thus have that the eigenvalues of the Jacobian $\vJ_{\text{PF}}(\vtheta^*)$ at the critical point $\vtheta^*$ are $\frac 1h \log(1 - h\lambda_i^*)$ for $i=1,\dots, D$.

\textbf{Local minima}.
Suppose that $\vtheta^*$ is a local minimum. Then all Hessian eigenvalues are non-negative $\lambda_i^* \ge 0$.
We perform the stability analysis in cases given by the value of $\lambda_i^*$, corresponding to the cases in Section~\ref{sec:principal_flow_eigendecom}:
\begin{itemize}
\item $h < 1/\lambda_i^*$. The corresponding eigenvalue of the Jacobian $\frac 1h \log(1 - h\lambda_i^*)$ is negative, since ${0 < 1 - h\lambda_i^* < 1}$. The PF is attractive in the corresponding eigenvector direction.
\item $h \in [1/\lambda_i^*,\, 2/\lambda_i^*)$. The corresponding eigenvalue of the Jacobian $\frac 1h \log(1 - h\lambda_i^*) = \frac 1 h \log(h\lambda_i^* - 1) + i\frac \pi h$ is complex, with negative real part since since $h\lambda_i^* -1 < 1$. The PF is attractive in the corresponding eigenvector direction.
\item $h \ge 2/\lambda_i^*$.  The corresponding eigenvalue of the Jacobian $\frac 1h \log(1 - h\lambda_i^*) = \frac 1 h \log(h\lambda_i^* - 1) + i\frac \pi h$ is complex, with non-negative real part, since since $h\lambda_i^* - 1 \ge 1$. If $h > 2/\lambda_i^*$, the PF is repelled in the corresponding eigenvector direction.
\end{itemize}
The last case tells us that the PF is not always attracted to local minima, as it is repelled in eigen-directions where $h > 2/\lambda_i^*$. Thus \textbf{like gradient descent, the PF can be repelled around local minima for large learning rates}. This is in contrast to the NGF and the IGR flow, which always predict convergence around a strict local minimum: the eigenvalues of the NGF Jacobian are $-\lambda_i^*$, and for the IGR flow the eigenvalues are $-\lambda_i^* - \frac{h^2}{2} {\lambda_i^*}^2$, both are strictly negative when $\lambda_i^*$ is strictly positive. For derivations see Section~\ref{sec:jacobian_igr_ngf} in the Appendix.

\begin{remark} For quadratic losses, where the PF is exact, the results above recover the classical gradient descent result for quadratic losses namely that gradient descent convergences if $\lambda_0 < 2/h$, otherwise diverges.
\end{remark}

\textbf{Saddle points}.
Suppose that $\vtheta^*$ is a strict saddle point. In this case there exists $\lambda^*_s$, such that $\lambda^*_s < 0$. We want to analyse the behaviour of the PF in the direction of the corresponding eigenvector $\vu_s^*$. In that case, $\log( 1 - h\lambda_s^*) > 0$ which entails that the PF is repelled in the eigendirections of strict saddle points. Note that this is also the case for the NGF since the corresponding eigenvalues of the Jacobian of the NGF would be $-\lambda_s^*$,  also positive. Unlike the NGF, however, the subspace of eigendirections that the PF is repelled by can be larger since it includes also eigendirections where $\lambda_i^* > 2/h > 0$. 

\section{Predicting neural network gradient descent dynamics with the PF}
\label{sec:the_pf_and_nns}

\begin{figure}[tb]
 \begin{subfloat}[$\lambda_0 < 1/h$]{
 \includegraphics[width=0.33\columnwidth]{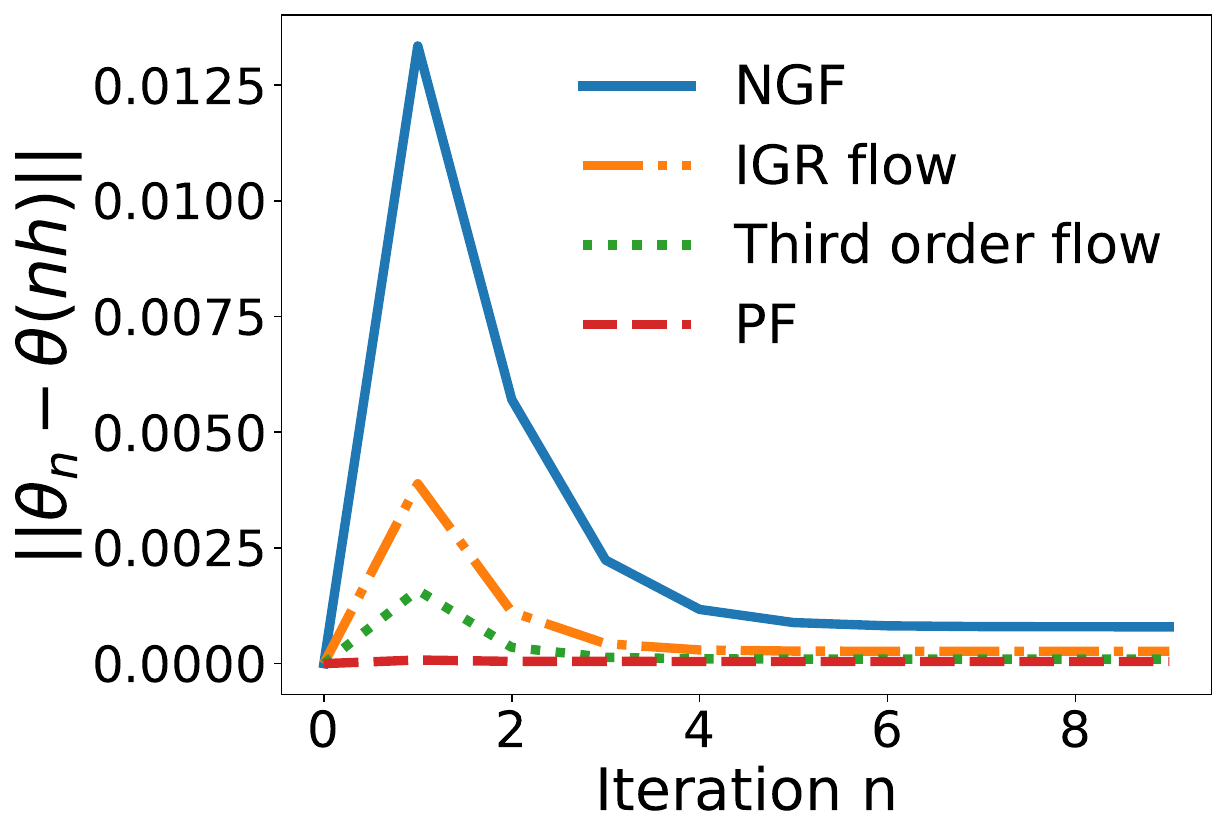}
 }\end{subfloat}%
\begin{subfloat}[$1/h < \lambda_0 < 2/h$]{
 \includegraphics[width=0.33\columnwidth]{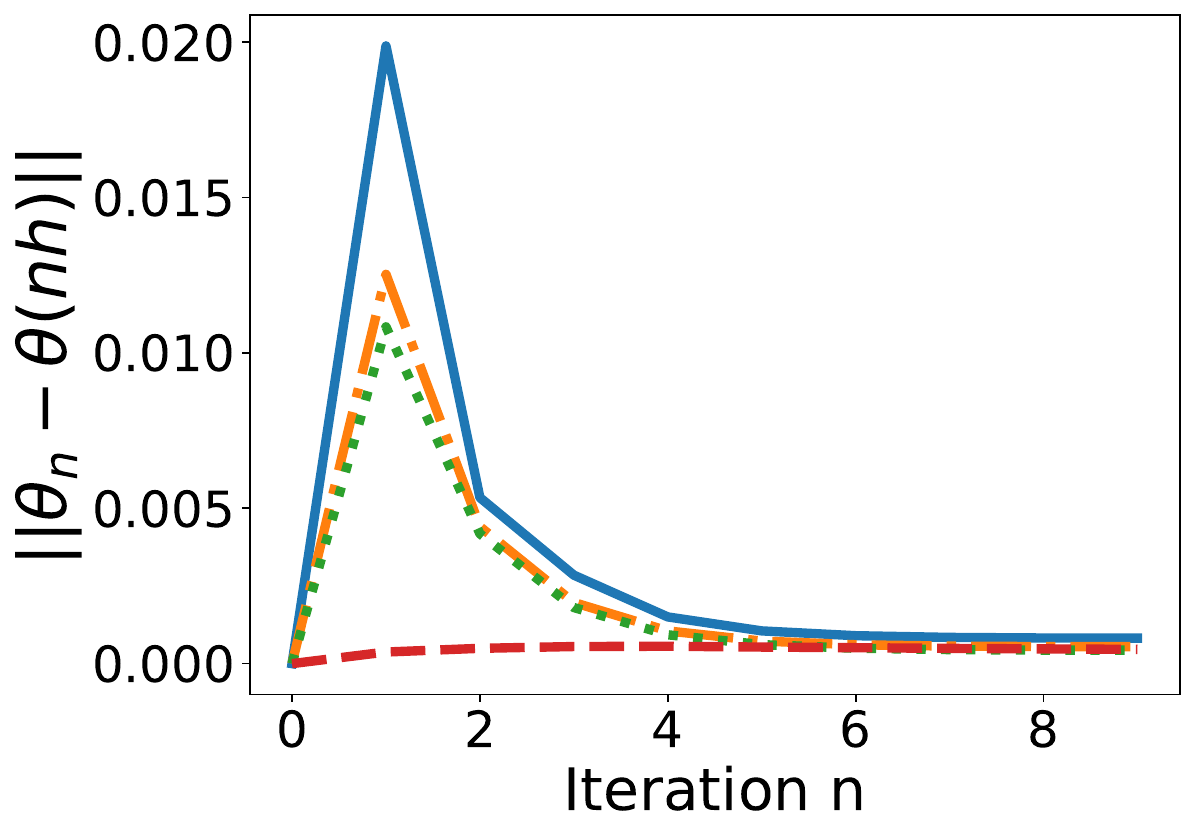}
 }\end{subfloat}%
\begin{subfloat}[$\lambda_0 > 2/h$]{
 \includegraphics[width=0.315\columnwidth]{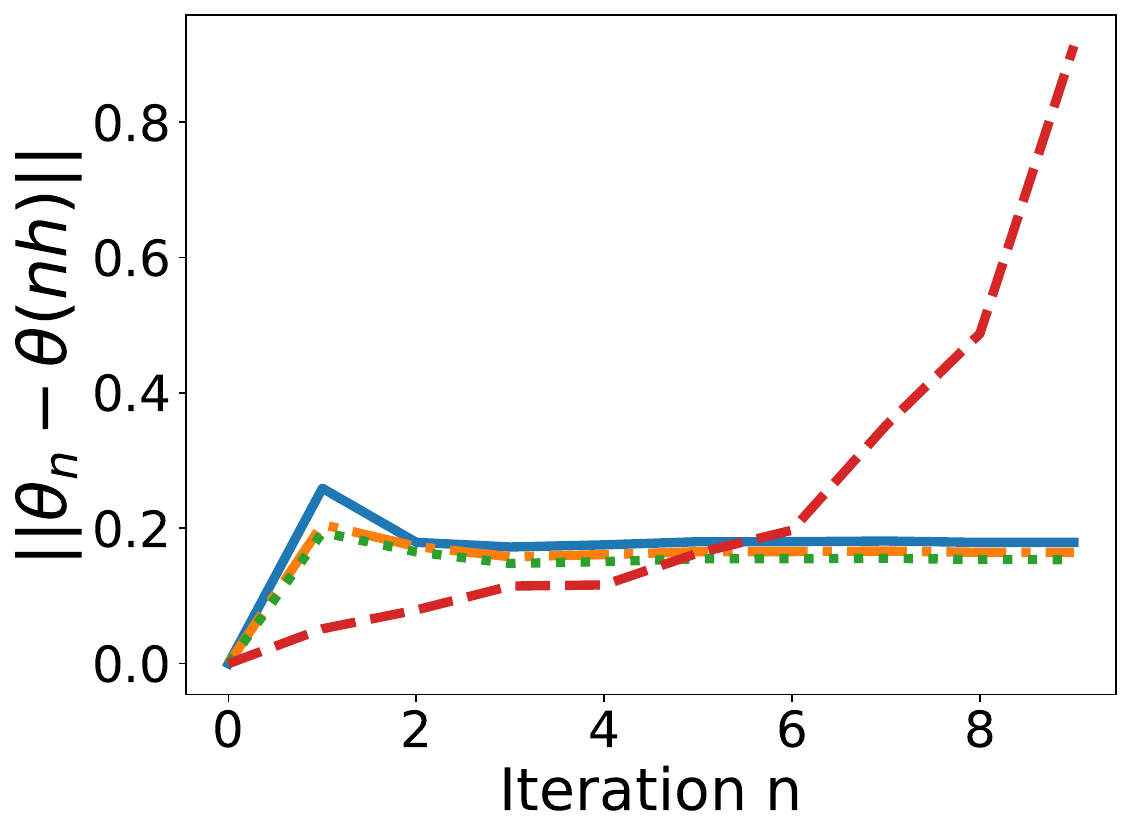}
 }\end{subfloat}%
\caption[Small neural networks: comparing continuous flows for multiple iterations.]{Error between gradient descent parameters and parameters obtained following continuous-time flows for multiple iterations: $\norm{\vtheta_{n} - \vtheta(nh)}$ with $\vtheta(0) = \vtheta_0$. For small $n$, the PF is better at capturing the behaviour of gradient descent across all cases.}
\label{fig:intuition_nn_function_delta}
\end{figure}

Computing the PF on large neural networks during training is computationally prohibitive, as it requires finding all eigenvalues of the Hessian matrix once for each step of the flow simulation, corresponding to many eigen-decompositions per gradient descent step. 
To build intuition about the PF for neural networks, we start with a small MLP for a two-dimensional input regression problem, with random inputs and labels. Here, we can compute the PF's vector field exactly and compare it with the behaviour of gradient descent. We show results in Figure~\ref{fig:intuition_nn_function_delta}, where we visualise the norm of the difference between gradient descent parameters at each iteration and the parameters produced by the continuous-time flows we compare with. We observe that \textit{short term the PF is better than all other flows at tracking the behaviour of gradient descent}. As the number of iterations increases, however, the PF accumulates error in the case of $\lambda_0 > 2/h$; this is likely due to the fact that while gradient descent 
parameters are real, this is not the case for the PF, as discussed in Remark~\ref{rem:pf_multiple_ints}. Since we are primarily concerned with using the PF to understand gradient descent for a small number of iterations this will be less of a concern in our experimental settings. Additional results that confirm the PF is better than the other flows at tracking gradient descent on a bigger network trained the UCI breast cancer dataset~\citep{asuncion2007uci} are shown in Figure~\ref{fig:breast_cancer_principal_flow} in the Appendix.

\begin{remark}
\textbf{Multiple iteration behaviour of the PF}. While gradient descent parameters are real for any iteration $\vtheta_t$, $\vtheta_{t+1}$, ... , $\vtheta_{t+n}$ when we approximate the behaviour of gradient descent by initialising $\vtheta(0) = \vtheta_t$ and running the PF for time $nh$, there is nothing enforcing that $\vtheta(h)$, ... $\vtheta(nh)$ will be real when the PF is complex valued ($\lambda_0 > 1/h$). Outside the quadratic case the Hessian is not Hermitian, the eigenvalues and eigenvectors of the Hessian will not be real, and the eigenvectors are no longer guaranteed to form a basis\footnote{To avoid the concern around the eigenvectors of the Hessian no longer forming a basis, one can use the Jordan normal form instead, as we will do later, in Section~\ref{sec:bea_pf_games}, where we expand the PF to games. We don't take this approach here as most of our following analysis is not affected, and is concerned with the behaviour of the PF around one gradient descent iteration. Furthermore, support of the Jordan normal form in code libraries is limited (especially for complex matrices), and we did not find this to be a significant issue in the experiments where we simulate the PF outside the quadratic case for a few iterations. We note, however, that mathematical analysis of long-term PF trajectories  for general functions should use the Jordan normal form.}. For long-term trajectories (larger $n$), this can have an effect on the long-term error between gradient descent and PF trajectories, through an accumulating effect of the imaginary part in the PF. This can be mitigated by using the PF to understand the short-term behaviour of gradient descent (small $n$).
\label{rem:pf_multiple_ints}
\end{remark}

\subsection{Predicting $\left(\nabla_{\vtheta} E\right)^T \vu_0$ using the PF}
\label{sec:nns_principal_flow}

For large neural networks,  instead of simulating the PF describing how the entire parameter vector changes in time we can use the PF to approximate changes in a scalar quantity only. This will allow us to compare the predictions of the PF against the predictions of the NGF and IGR flow on realistic settings, where the models have millions of parameters. To do so, we first have to compute how the gradient $\nabla_{\vtheta}E$ changes in time:
\begin{corollary} If $\vtheta$ follows the PF, then:
\begin{align}
\dot{\left({\nabla_{\vtheta}E}\right)} =  \sum_{i=0}^{D-1} \frac{\log(1 - h \lambda_i)}{h} (\nabla_{\vtheta} E^T \vu_i)  \vu_i.
\end{align}
\end{corollary}
This follows from applying the chain rule and using the definition of the PF. We now contrast this result with how the gradient evolves if the parameters follow the NGF or the IGR flow.
\begin{corollary} If $\vtheta$ follows the NGF, then:
\begin{align}
\dot{\left({\nabla_{\vtheta}E}\right)} =  \sum_{i=0}^{D-1} -  \lambda_i (\nabla_{\vtheta} E^T \vu_i)  \vu_i.
\label{col:grad_ngf_ode}
\end{align}
\end{corollary}

\begin{corollary} If $\vtheta$ follows the IGR flow, then:
\begin{align}
\dot{\left({\nabla_{\vtheta}E}\right)} =  \sum_{i=0}^{D-1} - \left(\lambda_i + \frac{h}{2} \lambda_i^2 \right)  (\nabla_{\vtheta} E^T \vu_i)  \vu_i.
\end{align}
\end{corollary}

We would like to use the above to assess how $\nabla_{\vtheta} E^T \vu_i$ changes in time under the above flows and check their predictions empirically against results obtained when training neural networks with gradient descent. Since $\vu_i$ is an eigenvector of the Hessian it also changes in time according to the changes given by the corresponding flow, making $\dot{\left(\nabla_{\vtheta}E^T \vu_i\right)} = \frac{d \left(\nabla_{\vtheta}E^T \vu_i\right)}{d t}$ difficult to calculate. Even when if we wrote an exact flow for $\frac{d \left(\nabla_{\vtheta}E^T \vu_i\right)}{d t}$,  it would be computationally challenging to simulate it since finding the new values of $\vu_i$ would depend on the full Hessian and would lead to the same computational issues we are trying to avoid in the case of large neural networks. In order to mitigate these concerns, we will make the additional approximation that $\lambda_i$ and $\vu_i$ do not change inside an iteration, which will allow us to  approximate changes to $\nabla_{\vtheta} E^T \vu_i$ and compare them against empirical observations. We note that we will not use this approximation for any other results.

\begin{remark}If we assume that $\lambda_i$, $\vu_i$ do not change between iterations, if $\vtheta$ follows the PF then $\frac{d \left(\nabla_{\vtheta}E^T \vu_i\right)}{d t} = \frac{\log(1 - h \lambda_i)}{h} \nabla_{\vtheta} E^T \vu_i$.
\end{remark}

\begin{remark} If we assume that $\lambda_i$, $\vu_i$ do not change between iterations, if $\vtheta$ follows the NGF we can write $\frac{d \left(\nabla_{\vtheta}E^T \vu_i\right)}{d t} = - \lambda_i \nabla_{\vtheta} E^T \vu_i$.
\end{remark} 

\begin{remark} If we assume that $\lambda_i$, $\vu_i$ do not change between iterations, if $\vtheta$ follows the IGR flow we can write $\frac{d \left(\nabla_{\vtheta}E^T \vu_i\right)}{d t} = - \left(\lambda_i + \frac{h}{2} \lambda_i^2 \right)  \nabla_{\vtheta} E^T \vu_i$.
\end{remark} 

\begin{figure}
\begin{subfloat}[$h=0.01$]{
 \includegraphics[width=0.32\columnwidth]{  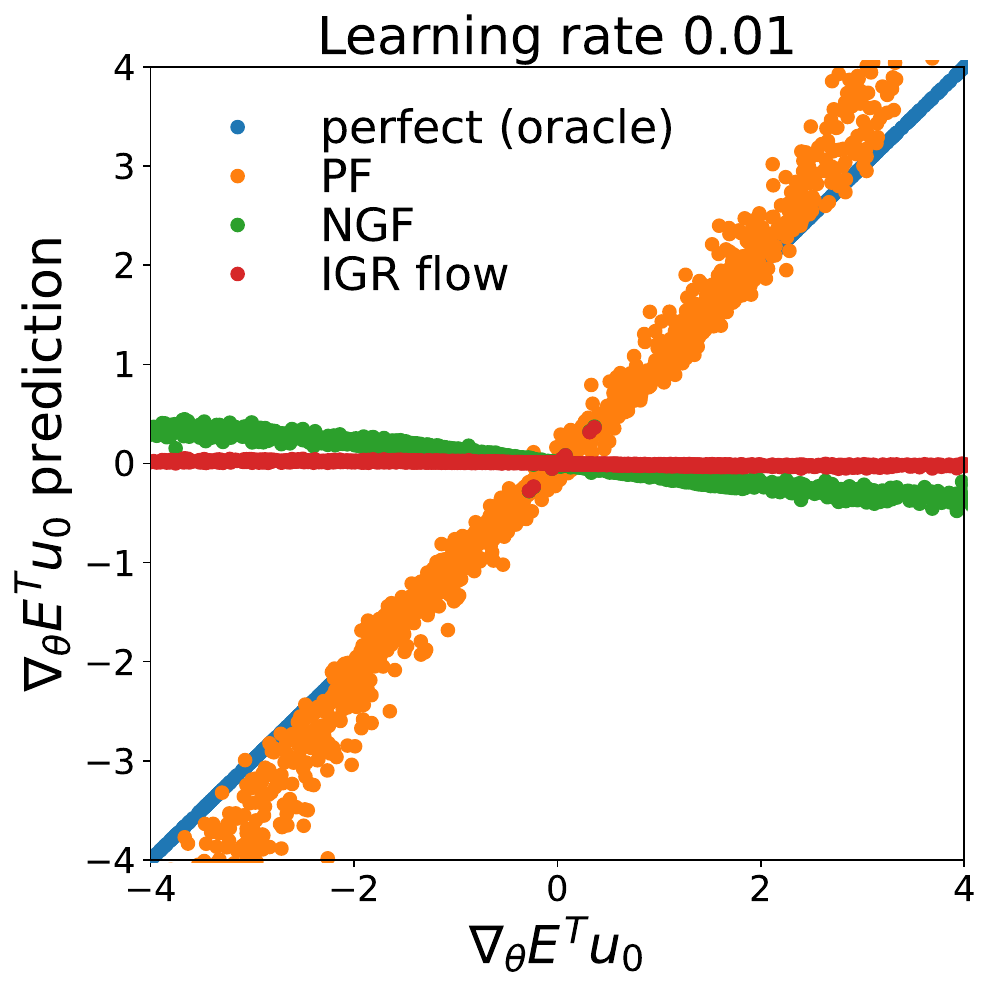}%
}\end{subfloat}%
\begin{subfloat}[$h=0.03$]{
 \includegraphics[width=0.32\columnwidth]{  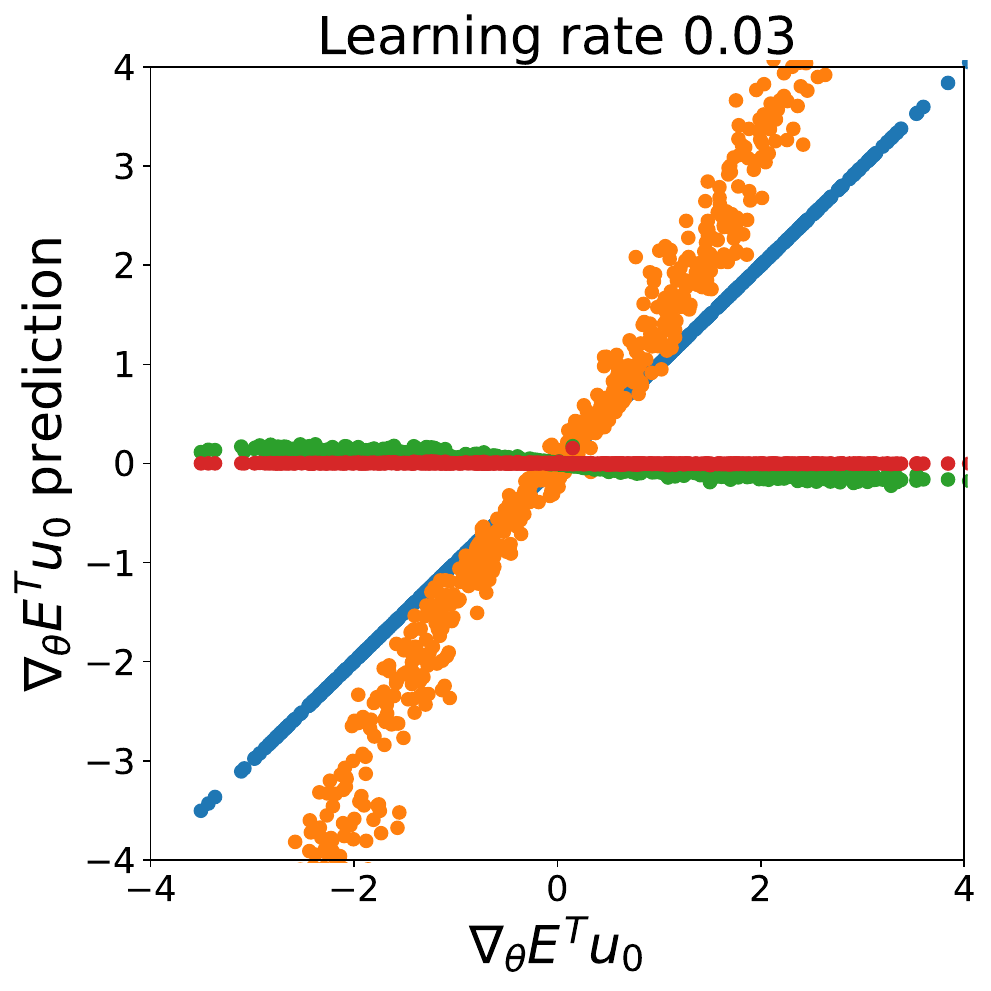}%
}\end{subfloat}%
\begin{subfloat}[$h=0.05$]{
 \includegraphics[width=0.32\columnwidth]{  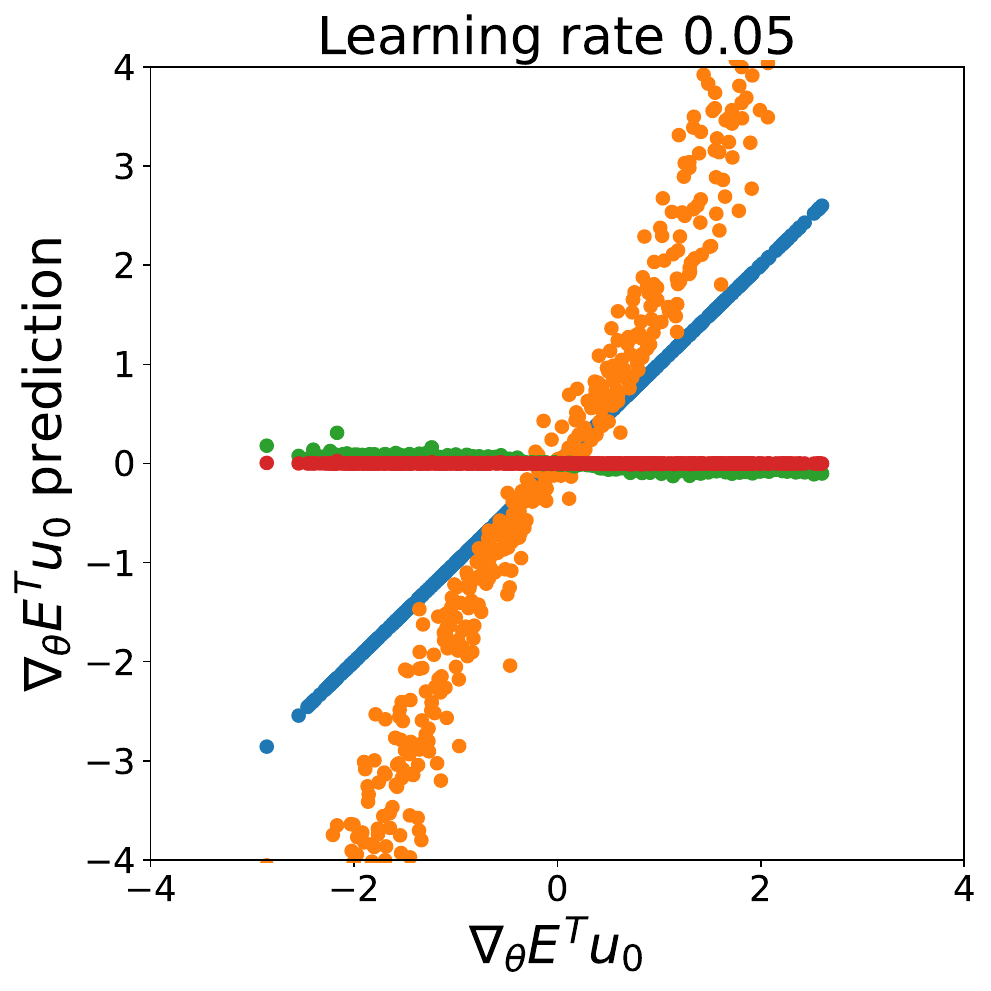}%
}\end{subfloat}%
\caption[Predictions of $\nabla_{\vtheta}E^T\vu_0$ according to continuous-time flows for a VGG model trained on CIFAR-10.]{Predictions of $\nabla_{\vtheta}E^T\vu_0$ according to the NGF, IGR flow and the PF on full-batch training of a VGG model on CIFAR-10. On the $x$ axis we plot the value of $\nabla_{\vtheta}E^T \vu_0$ as measured empirically in training, and on the $y$ axis we plot the corresponding prediction according to the flows from the value of the dot product at the previous iteration. The `exact match' line indicates a perfect prediction, the upper bound of performance. The PF performs best from all the compared flows, however, for higher learning rates its performance degrades when $\nabla_{\vtheta}E^T \vu_0$ is large; this is due to the fact that the higher the learning rate and the higher the gradient norm, the more likely it is that the additional assumption we used that $\lambda_i, \vu_i$ do not change does not hold.}
\label{fig:prediction_grad_u}
\end{figure}

The above flows have the form $\dot{x} = c x$, with solution $x(t) = x(0) e^{ct}$.
We can thus test these solutions empirically by training neural networks with gradient descent with learning rate $h$ and at each step compute $\nabla_{\vtheta}E(\vtheta_t)^T (\vu_i)_{t-1}$ and compare it with the prediction $x(h)$ obtained from the solution from each flow initialised at the previous iteration, i.e. $x(0) = \nabla_{\vtheta}E(\vtheta_{t-1})^T (\vu_i)_{t-1}$.  We show results with a VGG model trained on CIFAR-10 in Figure~\ref{fig:prediction_grad_u}. The results show that the PF is substantially better than the NGF and IGR flow at predicting the behaviour of $\nabla_{\vtheta} E^T \vu_0$. Since the NGF and the IGR flow solutions scale the initial value by the inverse of an exponential of magnitude given by $\lambda_0$, for large $\lambda_0$ this leads to a small prediction, which is not aligned with what is observed empirically. We also note that the higher the value of $\nabla_{\vtheta} E^T \vu_0$, the worse the prediction of the PF; these are the areas where the approximations made in the above remarks are likely not to hold due to large gradient norms.

\subsection{\rebuttalrone{Around critical points: escaping sharp local minima and saddles}}

\begin{figure}
\centering
\begin{subfloat}[$\lambda_0$.]{
  \includegraphics[width=0.49\columnwidth]{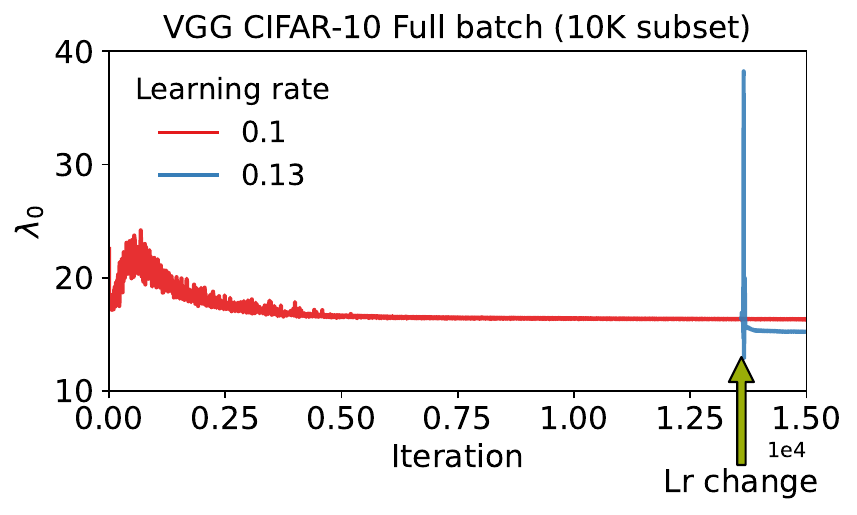}%
}
\end{subfloat}%
\begin{subfloat}[$E(\vtheta).$]{
 \includegraphics[width=0.49\columnwidth]{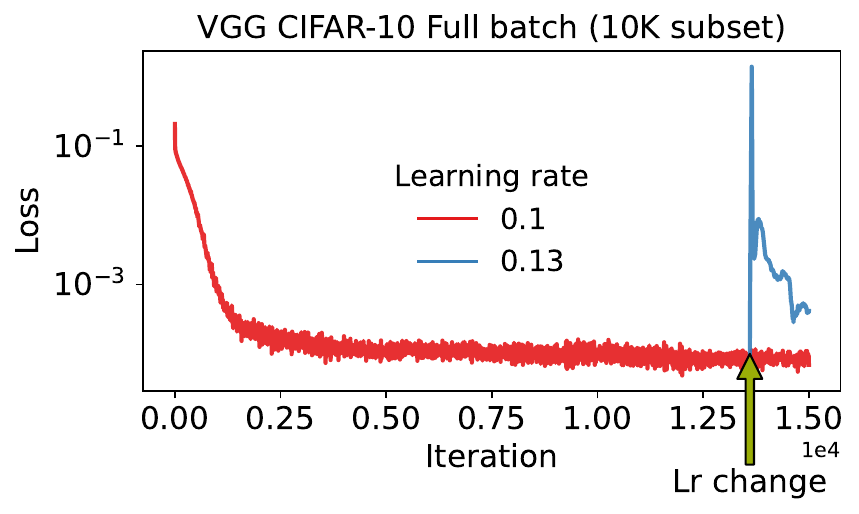}%
}
\end{subfloat}%
\caption[Sharp minima are not attractive to gradient descent, a neural network experiment.]{Local minima are not attractive to gradient descent if $\lambda_0^*> 2/h$. Since we cannot easily find local minima in neural network landscapes, we perform an experiment where we train a neural network with learning rate $h'<h$ close to convergence, and then change the learning rate to $h$, close to but above $2/\lambda_0^*$. Increasing the learning rate slightly leads to instabilities and exiting the stable training area, despite being in a late stage of training. We note that while this is \textit{not} the same as assessing local attraction in a stability sense, since we do not know if the point of the learning rate change is in the neighbourhood provided by the existence conditions of local stability, it is the closest practical approximation that we can provide given the complexity of neural network landscapes.}
\label{fig:stability_analysis_nn}
\end{figure}

\rebuttalrone{The stability analysis we performed in Section~\ref{sec:pf_stability_analysis} showed the PF is repelled by local minima where $\lambda_0^* >2/h$: that is, even if the model is close to a sharp local minima (with $\lambda_0^* > 2/h$), that local minima will not be attractive and training will continue until a shallow minima is reached. We provide experimental evidence to support that hypothesis in neural network training in Figure~\ref{fig:stability_analysis_nn}. We train a neural network close to convergence, and then increase the learning rate slightly to a learning rate that exceeds the sharpness threshold $h > 2/\lambda_0^*$, and observe that the model promptly exits the area, despite being in a late stage of training. These results are consistent with observations in the deep learning literature~\citep{jastrzkebski2018relation,cohen2021gradient}. Furthermore, while saddle points have long been considered a challenge with high dimensional optimisation~\citep{dauphin2014identifying} in practice gradient descent has not been observed to converge to saddles~\citep{lee2016gradient}. Our analysis suggests that saddles will be repelled not only in the direction of strictly negative eigenvalues, but also in the eigendirections with large positive eigenvalues when large learning rates are used; this can explain why neural networks do not converge to non-strict saddles which exist in deep neural landscapes~\citep{kawaguchi2016deep} but need not be repelling for the NGF and IGR flow (existing analyses of escaping saddle points by gradient descent apply only to strict saddles~\citep{du2017gradient,lee2016gradient}).}

\section{The PF, stability coefficients and edge of stability results}
\label{sec:instability_deep_learning}

\begin{figure}[tb]
\begin{subfloat}[Loss $E(\vtheta)$.]{
 \includegraphics[width=0.33\columnwidth]{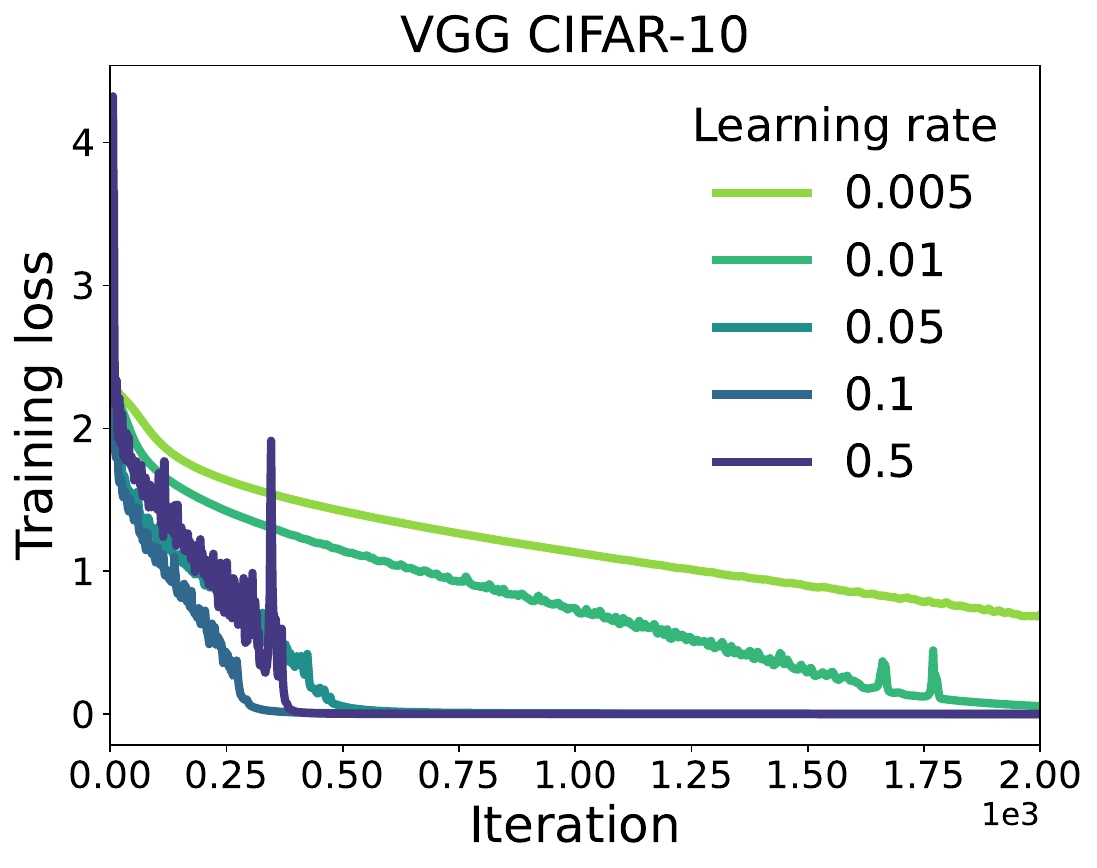}%
}
\end{subfloat}%
\begin{subfloat}[$\lambda_0$.]{
 \includegraphics[width=0.34\columnwidth]{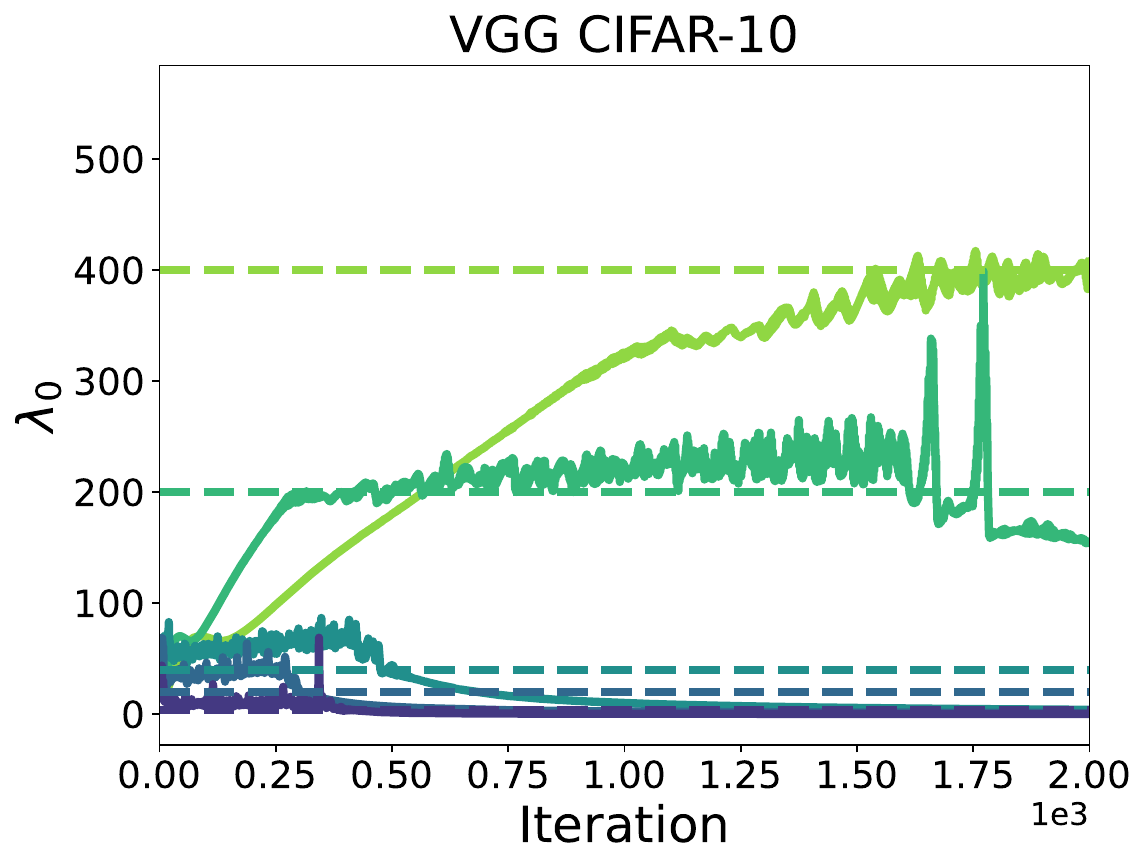}%
}
\end{subfloat}%
\begin{subfloat}[Increases in loss function overwhelmingly occur when $\lambda_0 > 2/h$.]{
 \includegraphics[width=0.33\columnwidth]{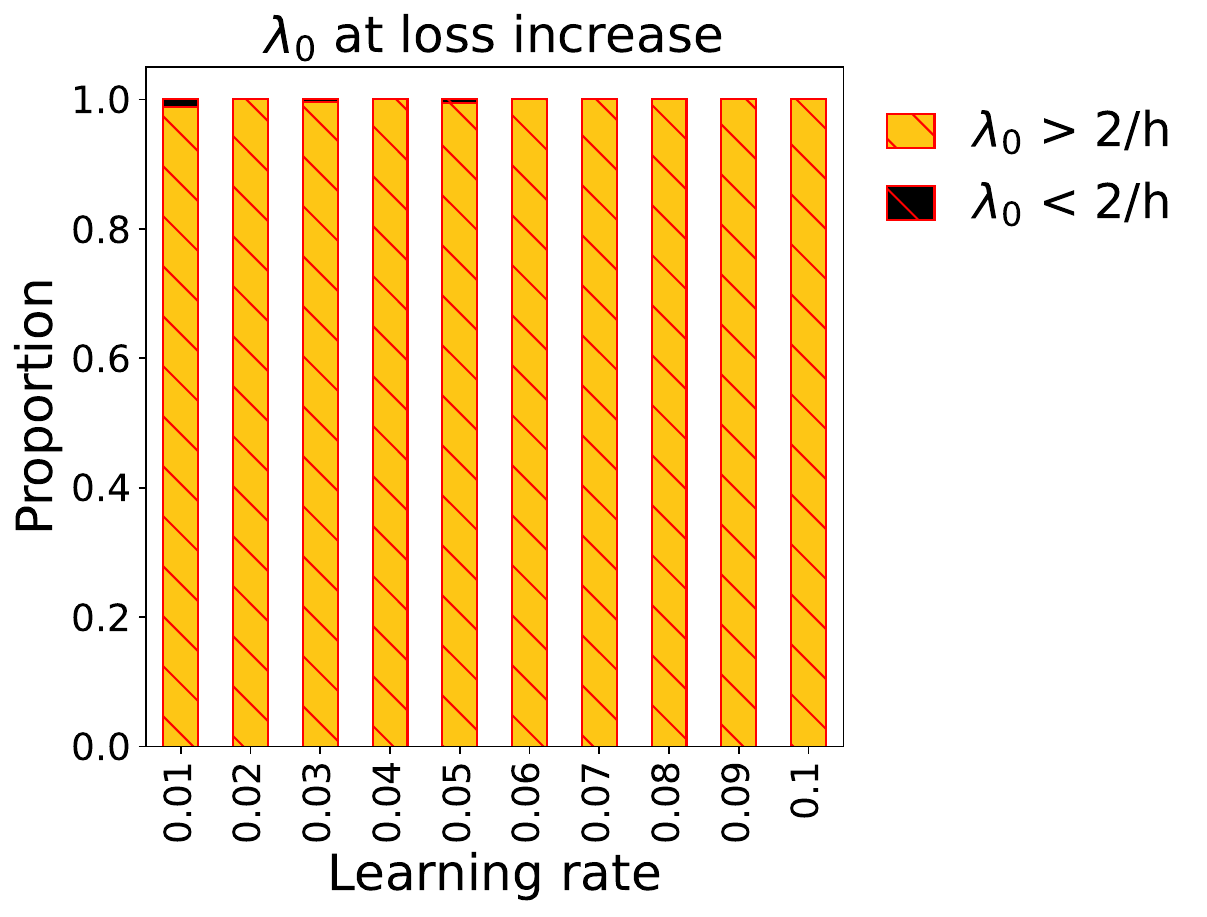}%
}
\end{subfloat}%
\caption[The edge of stability behaviour in neural networks.]{Edge of stability in neural networks. ~\citet{cohen2021gradient} observed that when training neural networks with full-batch gradient descent, training is stable and $\lambda_0$ increases until $\lambda_0 > 2/h$, when local loss instabilities occur and $\lambda_0$ starts oscillating around $2/h$. }
\label{fig:reproduce_edge_of_stability}
\end{figure}

\textbf{Edge of stability results.} 
 \citet{cohen2021gradient} did a thorough empirical study to show that when training deep neural networks with full-batch gradient descent the largest eigenvalue of the Hessian, $\lambda_0$, keeps growing until reaching approximately $2/h$ (a phase of training they call \textit{progressive sharpening}), after which it remains in that area; for mean-squared losses this continues indefinitely while for cross entropy losses they show it further decreases later in training. They also show that instabilities in training occur when $\lambda_0 >2/h$. Their empirical study spans neural architectures, data modalities, and loss functions. We visualise the edge of stability behaviour they observe in Figure~\ref{fig:reproduce_edge_of_stability}; since we use a cross entropy loss $\lambda_0$ decreases later in training.
 We also visualise that iterations where the loss increases compared to the previous iteration overwhelmingly occur when $\lambda_0 >2/h$.
 \citet{cohen2021gradient} also empirically observe that $\vtheta^T \vu_0$ has oscillatory behaviour in the edge of stability area but is 0 or small outside it.

\begin{figure}[tb]
\centering
\begin{subfloat}[Changes in parameters.]{
\includegraphics[width=0.49\columnwidth]{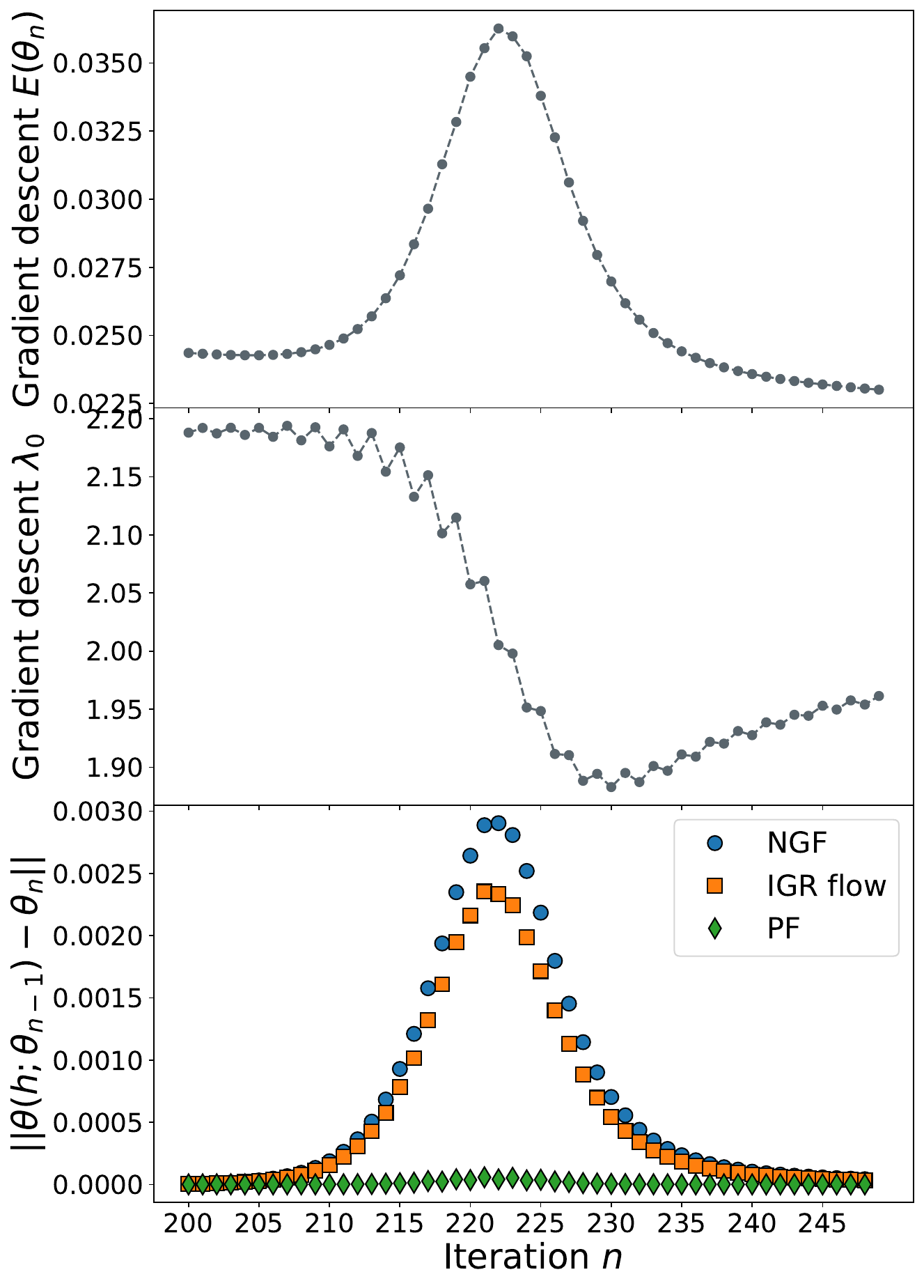}%
\label{fig:params_changes_edge_pf}
}
\end{subfloat}%
\begin{subfloat}[Changes in loss.]{
 \includegraphics[width=0.5\columnwidth]{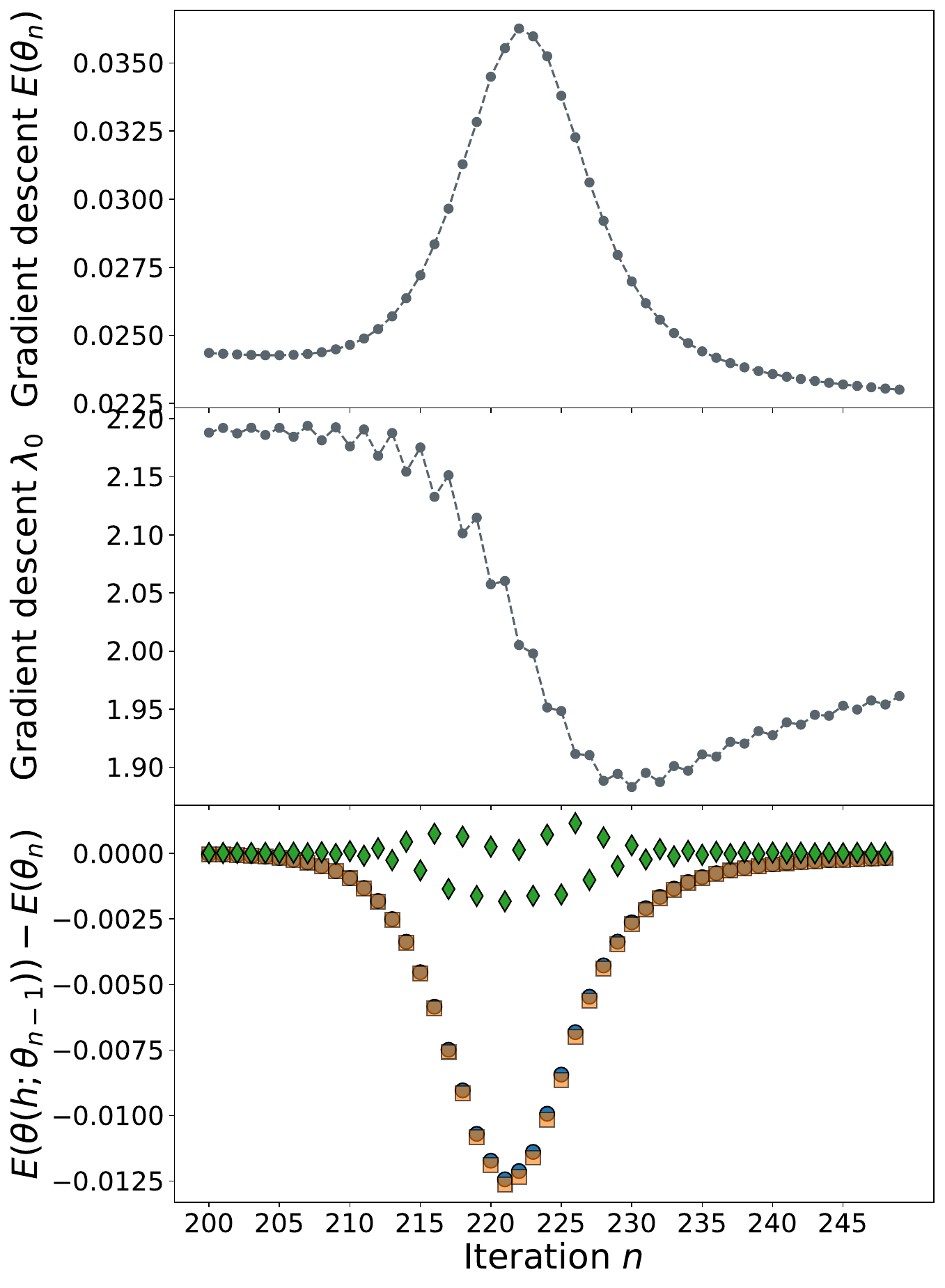}%
\label{fig:loss_changes_edge_pf}
}
\end{subfloat}%
\caption[Comparison of how continuous-time models capture the behaviour of gradient descent at edge of stability. ]{Comparing different continuous-time models of gradient descent at the edge of stability area on a small 5 layer MLP, with 10 units per layer. We show the local parameter prediction error $\norm{\vtheta_n - \vtheta(h;\vtheta_{n-1})}$ for the NGF, IGR and PF flows \subref{fig:params_changes_edge_pf}, as well as the loss prediction error $E(\vtheta(h;\vtheta_{n-1})) - E(\vtheta_n)$ \subref{fig:loss_changes_edge_pf}; when plotting the loss, for the PF we take the real part of the loss value.} 
\label{fig:model_of_gd_at_edge_of_stability}
\end{figure}

\rebuttalrone{\textbf{Continuous-time models of gradient descent at edge of stability.}} \rebuttalrone{To investigate if existing continuous-time flows and the PF capture gradient descent behaviour at the edge of stability we train a 5 layer MLP on the toy UCI Iris dataset~\citep{asuncion2007uci}; this simple setting allows for the computation of the full eigenspectrum of the Hessian. We show results in Figure~\ref{fig:model_of_gd_at_edge_of_stability}: 
the NGF and IGR flow have a larger error compared to the PF when predicting the parameters at the next gradient descent iteration in the edge of stability regime; the NGF and IGR flow  predict the loss will decrease, while the PF captures the loss increase observed when following gradient descent. 
As we remarked in Section~\ref{sec:motivation}, the NGF and the IGR flow do not capture instabilities when the eigenvalues of the Hessian are positive, which has been remarked to be largely the case for neural network training through empirical studies \citep{sagun2017empirical,ghorbani2019investigation,papyan2018full} and we observe here (Figure~\ref{fig:eigspectrum_small_mlp} in the Appendix).
We spend the rest of the section using the PF to understand and model edge of stability phenomena using a continuous-time approach.}

\textbf{Connection with the principal flow: stability coefficients.} The PF captures the key quantities 
observed in the edge of stability phenomenon: the eigenvalues of the Hessian $\lambda_i$ and the threshold $2/h$. These quantities appear in the PF via the stability coefficient $sc_i = \frac{\log(1-h\lambda_i)}{h\lambda_i}\nabla_{\vtheta} E^T \vu_i = \alpha_{PF}(\lambda_i h) \nabla_{\vtheta} E^T \vu_i$ of eigendirection $\vu_i$. Through the PF, by connecting the case analysis in Section~\ref{sec:principal_flow_eigendecom} with existing and new empirical observations, we can shed light on the edge of stability behaviour in deep learning. 

\begin{figure}[tb]
\centering
\begin{subfloat}[Early training.]{
 \includegraphics[width=0.48\columnwidth]{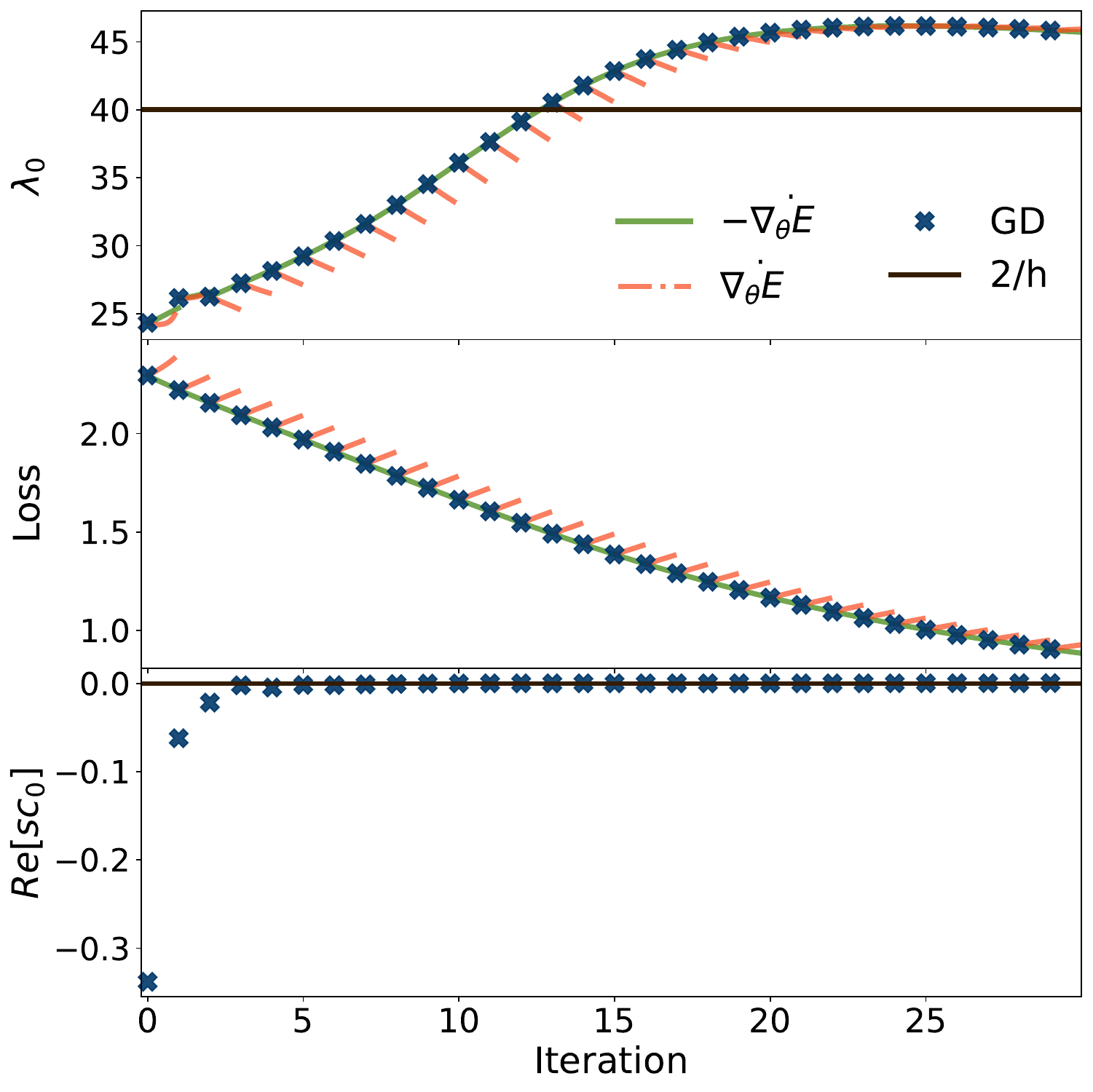}
 \label{fig:early_training_eigen_loss}
 }\end{subfloat}%
\begin{subfloat}[Edge of stability.]{
 \includegraphics[width=0.48\columnwidth]{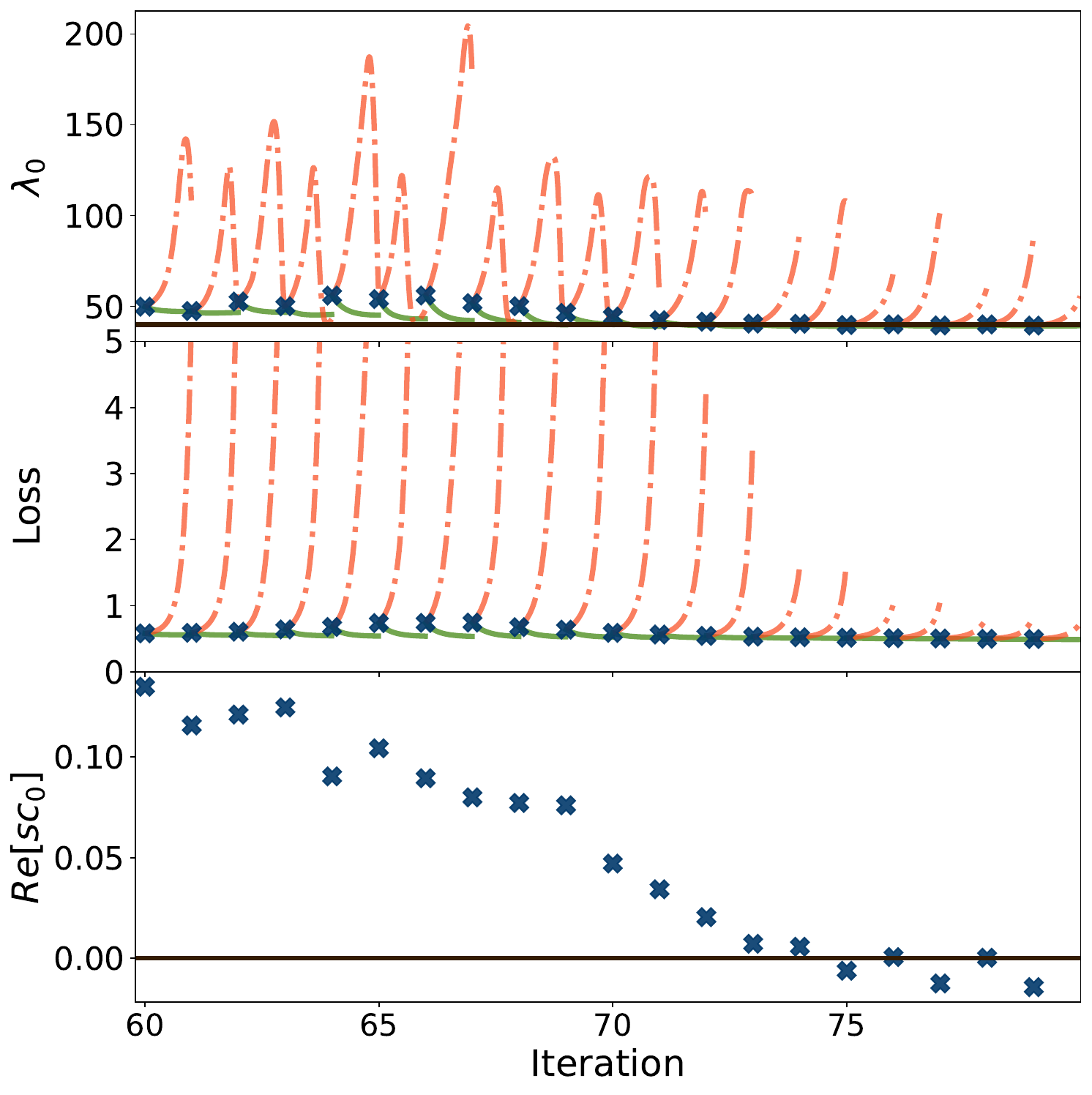}
 \label{fig:early_edge_eigen_loss}
}\end{subfloat}%
 \caption[Using the PF to understand stability and instability in a 4 layer MLP trained on MNIST.]{Understanding the edge of stability results using the PF on a 4 layer MLP: we plot the behaviour of the NGF $\dot{\vtheta} = - \nabla_{\vtheta} E$ and the positive gradient flow $\dot{\vtheta} = \nabla_{\vtheta} E$ initialised at each gradient descent iteration parameters, and see that the behaviour of gradient descent is connected to the behaviour of the respective flow through the stability coefficient $sc_0(\vtheta) = \alpha_{PF}(h\lambda_0)\nabla_{\vtheta} E^T \vu_0$. \rebuttalrone{\textit{\subref{fig:early_training_eigen_loss} shows that even when $\lambda_0 > 2/h$, if the real part of the stability coefficient $sc_0$ is negative or close to 0, there are no instabilities in the loss and the eigenvalue $\lambda_0$ keeps increasing, as it does when following the NGF in that region.}}}
\label{fig:edge_of_stability_results_local}
\end{figure}

\textit{First phase of training (progressive sharpening): $\lambda_0 < 2/h$}. This entails $\Re[sc_i] = \Re[\alpha_{PF}(h \lambda_i)] \le 0, \forall i$ (Real stable and complex stable cases of the analysis in Section~\ref{sec:principal_flow_eigendecom}). $\sign(\alpha_{NGF}) = \sign(\alpha_{PF}) =-1$ and following the PF minimises $E$ or its real part (Eq~\eqref{eq:changes_in_e}). 
To understand the behaviour of $\lambda_0$, we now have to make use of empirical observations about the behaviour of the NGF early in the training of neural networks.
It has been empirically observed that in early areas of training, $\lambda_0$ increases here when following the NGF~\citep{cohen2021gradient}; we further show this in Figure~\ref{fig:mnist_gradient_flow} in the Appendix. Since in this part of training gradient descent follows closely the NGF, it exhibits similar behaviour and $\lambda_0$ increases. We show this case in Figure~\ref{fig:early_training_eigen_loss}.

\textit{Second phase of training (edge of stability) $\lambda_0 \ge 2/h$}. This entails $\Re[sc_0(\vtheta)] = \Re[\alpha_{PF}(h \lambda_i)] \ge 0$. (Unstable complex case of the analysis in Section~\ref{sec:principal_flow_eigendecom}). We can no longer say that following the PF minimises $E$. $\sign(\alpha_{NGF}(h \lambda_0)) \neq \sign(\Re[(\alpha_{PF}(h \lambda_0)])$, since $\alpha_{NGF}(h \lambda_0) = -1$ and $\sign(\Re[(\alpha_{PF}(h \lambda_0)])>0$ meaning that in that direction gradient descent resembles the positive gradient flow $\dot{\vtheta} = \nabla_{\vtheta} E$ rather than the NGF. The positive gradient flow component can cause instabilities, and the strength of the instabilities depends on the stability coefficient $sc_0 = \alpha_{PF}(h \lambda_0) \nabla_{\vtheta} E^T \vu_0$. We show in Figures~\ref{fig:early_edge_eigen_loss} and \ref{fig:instabilities_resnet} how the behaviour of the loss and $\lambda_0$ are affected by the behaviour of the positive gradient flow when $\lambda_0 > 2/h$.

\begin{figure}[tb]
\centering
\begin{subfloat}[MNIST.]{
\includegraphics[width=0.475\columnwidth]{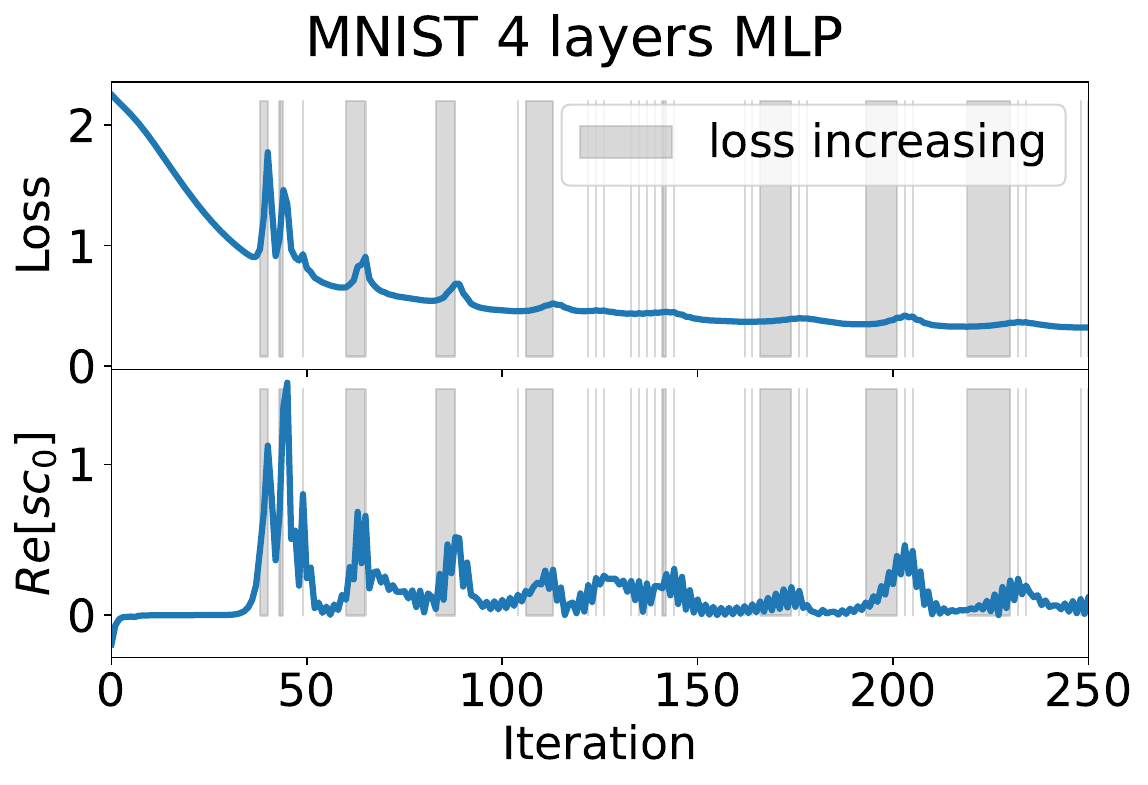}%
}\end{subfloat}%
\begin{subfloat}[CIFAR-10.]{
 \includegraphics[width=0.49\columnwidth]{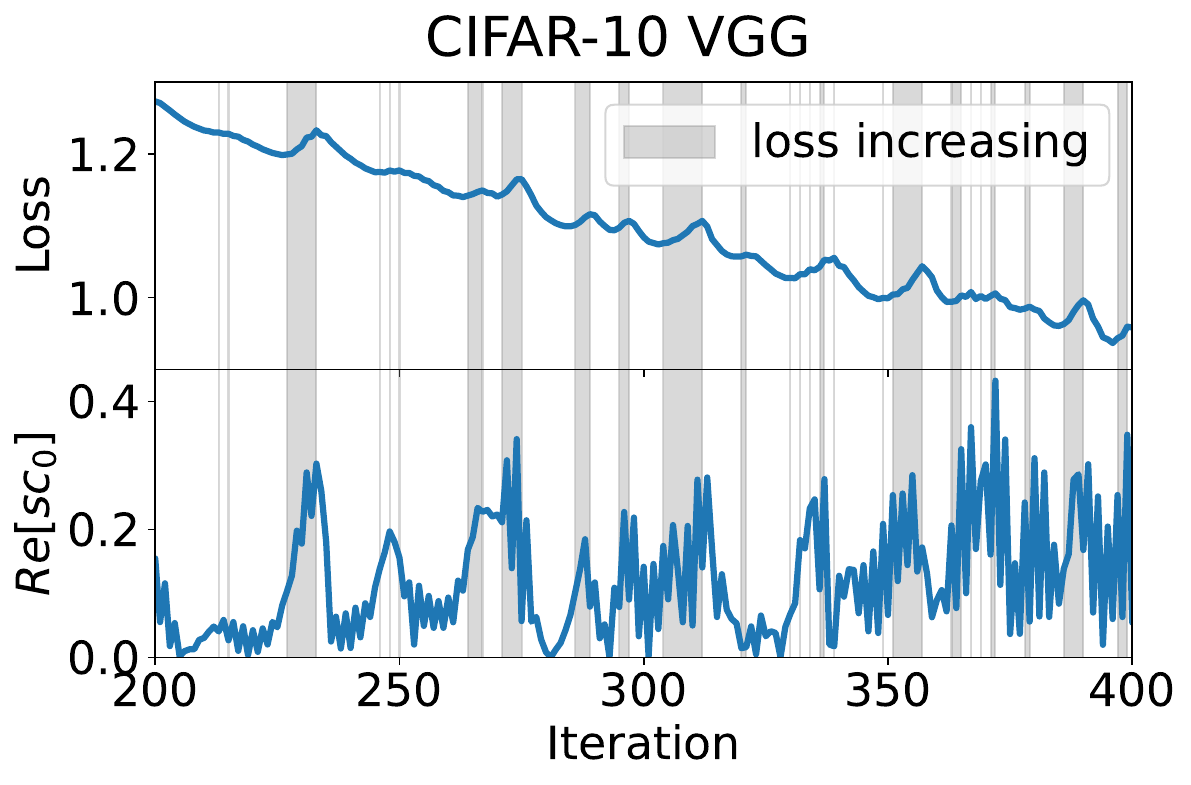}
}\end{subfloat}%
\caption[Using the PF and stability coefficient to understand instabilities in deep learning.]{\rebuttalrone{The value of the loss function and stability coefficients}: areas where the loss increases corresponds to areas where $sc_0$ is large. The highlighted areas correspond to regions where the loss increases.}
\label{fig:instabilities_short}
\end{figure}

\begin{figure}[tb!]
\centering
 \includegraphics[width=0.9\columnwidth]{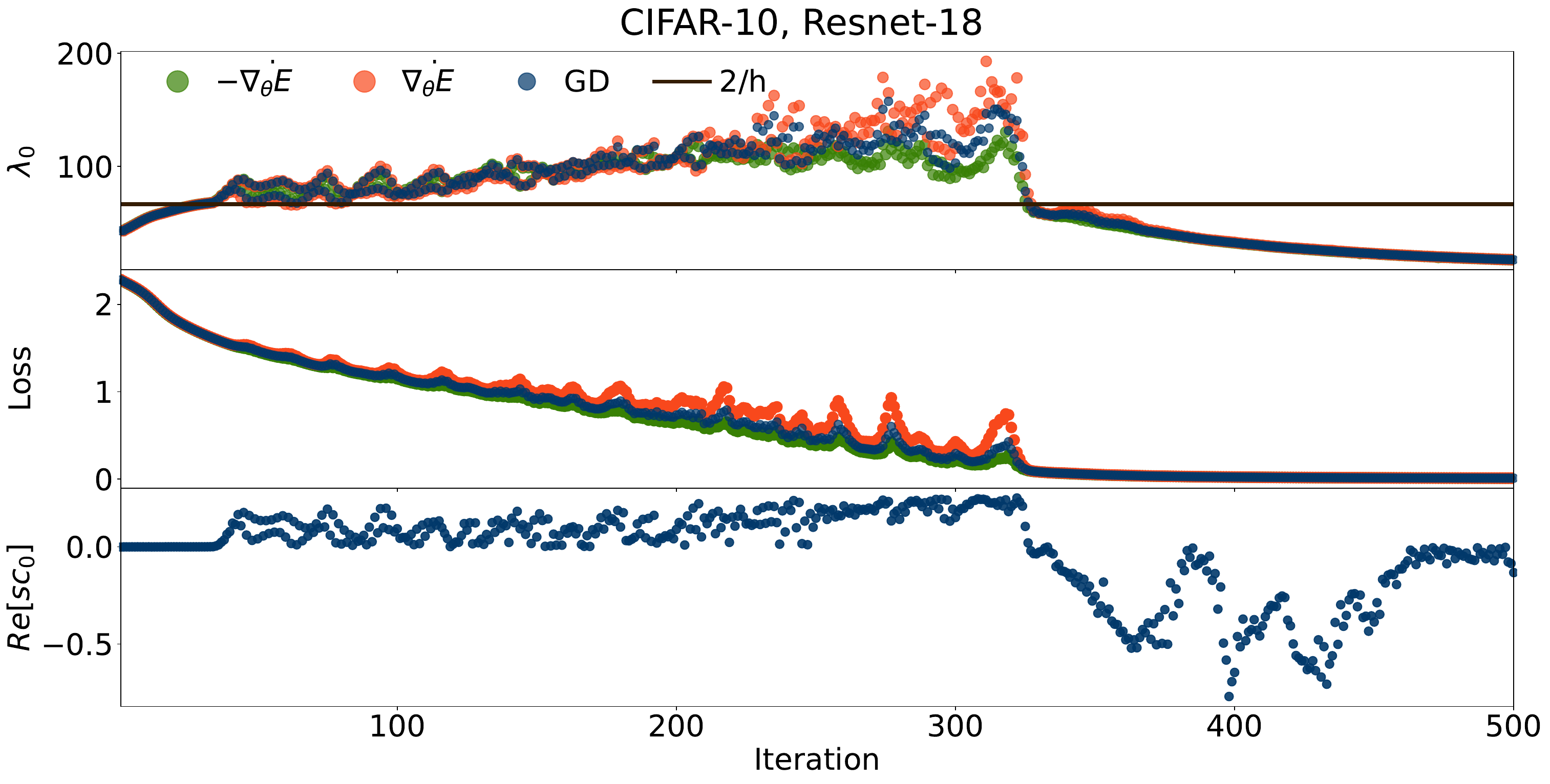}
\caption[Understanding the behaviour of the loss through the PF and the behaviour of $\lambda_0$.]{\rebuttalrone{Loss instabilities, $\lambda_0$ and stability coefficients for CIFAR-10.} Together with the behaviour of gradient descent, we plot the behaviour of the NGF and positive gradient flow initialised at $\vtheta_t$ and simulated for time $h$ for each iteration $t$. \rebuttalrone{The analysis we performed based on the PF suggests that when $\Re[sc_0] > 0$ and large we should expected gradient descent to exhibit behaviours close to those of the positive gradient flow. What we observe empirically is that increases in loss value of gradient descent are proportional to the increase of the positive gradient flow in that area (can be seen best between iterations 200 and 350); the same behaviour can be seen in relation to the eigenvalue $\lambda_0$. }}
\label{fig:instabilities_resnet}
\end{figure}

\textbf{More than $\lambda_0$: the importance of stability coefficients}. 
While the sign of the real part of the stability coefficient $sc_0$ is determined by $\lambda_0$, its magnitude is modulated by the dot product $\nabla_{\vtheta} E^T \vu_0$, since $sc_0 = \alpha_{PF}(h \lambda_0) \nabla_{\vtheta} E^T \vu_0$.
The magnitude of $\nabla_{\vtheta} E^T \vu_0$ plays an important role, 
since if $\lambda_0$ is the only eigenvalue greater than $2/h$ training is stable if $\nabla_{\vtheta}E ^T \vu_0 = 0$, as we observe in Figure~\ref{fig:edge_of_stability_results_local}. 
\textit{To understand instabilities, we have to look at stability coefficients, not only eigenvalues.}
 We show in Figure~\ref{fig:instabilities_short} how the instabilities in training can be related with the stability coefficient $sc_0$: the increases in loss occur when the corresponding $\Re[sc_0]$ is positive and large. 
 In Figure~\ref{fig:instabilities_resnet}, we show results with the behaviour of $\lambda_0$: $\lambda_0$ increases or decreases based on the behaviour of the corresponding flow and the strength of the stability coefficient and that gets reflected in instabilities in the loss function;
 specifically when $\lambda_0 > 2/h$, we use the positive gradient flow and see how the strength of its fluctuations affect the changes both in the loss value and $\lambda_0$ of gradient descent.
 We show additional results in Figures~\ref{fig:edge_of_stability_results_lambda} and~\ref{fig:edge_of_stability_results_cifar_resnet} in the Appendix.

\begin{figure}[tb]
\centering
\begin{subfloat}[Loss $E(\vtheta)$.]{
 \includegraphics[width=0.4\columnwidth]{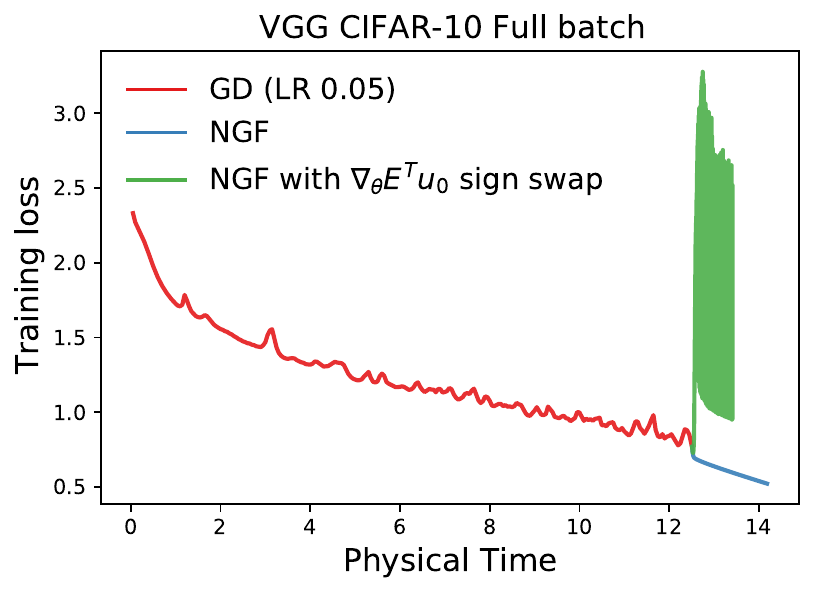}%
\label{fig:one_dim_sign_swap_loss}
}\end{subfloat}%
\hspace{3em}
\begin{subfloat}[$\lambda_0.$]{
 \includegraphics[width=0.4\columnwidth]{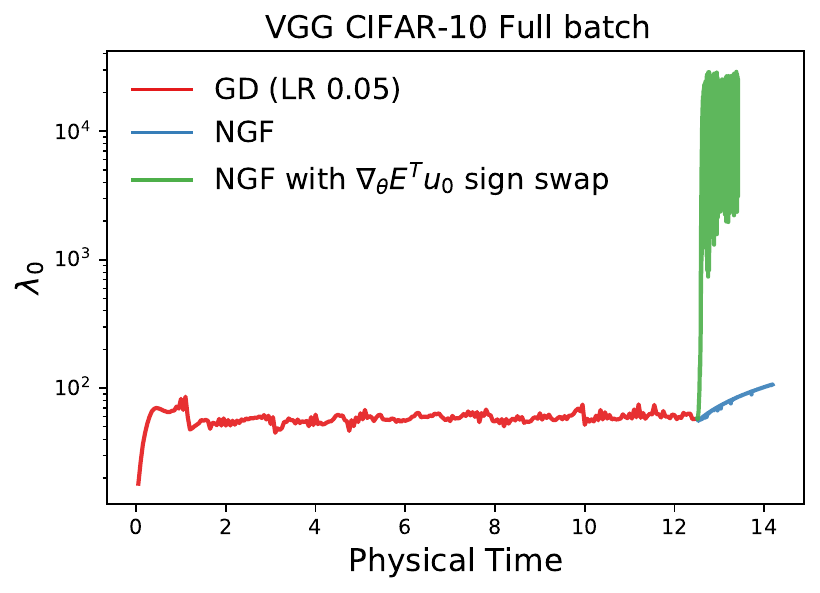}%
\label{fig:one_dim_sign_swap_lambda}
}\end{subfloat}%
\caption[Assessing whether 1 dimension is enough to cause instability in a continuous-time flow.]{One eigendirection is sufficient to lead to instabilities. To create a situation similar to that of the PF in neural network training, we construct a flow given by the NGF in all eigendirections but $\vu_0$; in the direction of $\vu_0$, we change the sign of the flow's vector field. This leads to the flow ${\dot{\vtheta} = \left(\nabla_{\vtheta}E^T \vu_0\right) \vu_0 + \sum_{i=1}^{D-1} -\left(\nabla_{\vtheta}E^T \vu_i\right) \vu_i}$. We show this flow can be very unstable when initialised in an edge of stability area, with increases in loss \subref{fig:one_dim_sign_swap_loss} and $\lambda_0$ \subref{fig:one_dim_sign_swap_lambda}.}
\label{fig:instabilities_change_learning_rate_loss}
\end{figure}

\textbf{Is one eigendirection enough to cause instability?} 
One question that arises from the PF is whether the leading eigendirection $\vu_0$ can be sufficient to cause instabilities, especially in the context of deep networks with millions of parameters. To assess this we train a model with gradient descent until it reaches the edge of stability ($\lambda_0 \approx 2/h$), after which we simulate the continuous flow ${\dot{\vtheta} = \left(\nabla_{\vtheta}E^T \vu_0\right) \vu_0 + \sum_{i=1}^{D-1} -\left(\nabla_{\vtheta}E^T \vu_i\right) \vu_i}$. 
The coefficients of the modified vector field of this flow are negative for all eigendirections except from $\vu_0$, which is positive; this is also the case for the PF when $\lambda_0$ is the only eigenvalue greater than $2/h$.
In Figure~\ref{fig:instabilities_change_learning_rate_loss} we empirically show that a positive coefficient for $\vu_0$ can be responsible for an increase in loss value and a significant change in $\lambda_0$ in neural network training.

\begin{figure}[tbh]
\centering
\begin{subfloat}[Loss $E(\vtheta)$.]{
  \includegraphics[width=0.45\columnwidth]{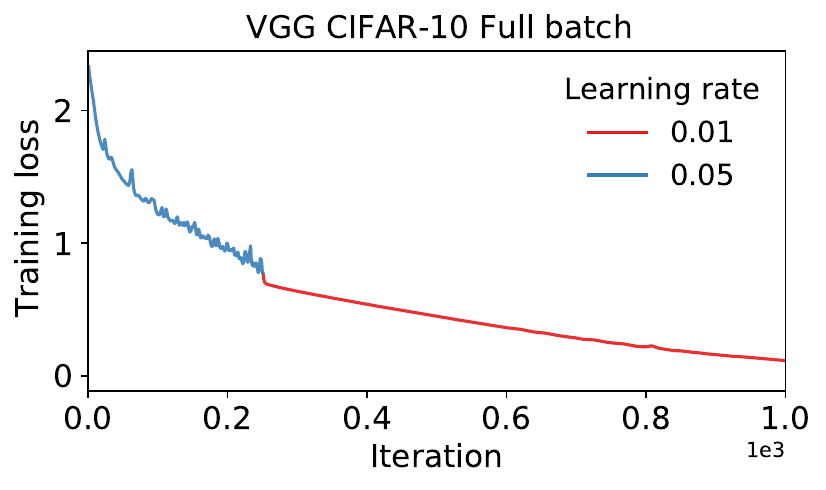}%
\label{fig:change_learning_rate_loss}
}\end{subfloat}%
\begin{subfloat}[$\lambda_0$.]{
  \includegraphics[width=0.47\columnwidth]{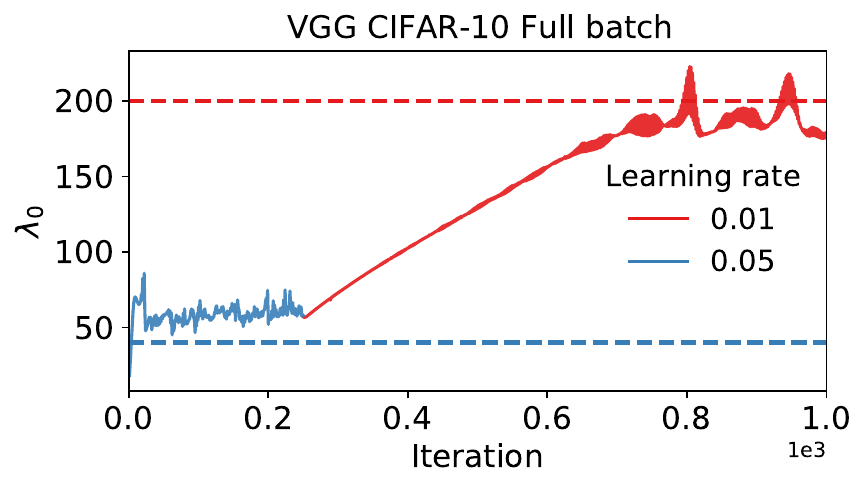}%
\label{fig:change_learning_rate_eigen}
}\end{subfloat}%
\caption[Decreasing the learning rate in the edge of stability area leads to increased stability.]{
Decreasing the learning rate in the edge of stability area leads to increased stability, as discovered by~\citet{cohen2021gradient}. The lack of loss instability with a smaller learning rate \subref{fig:change_learning_rate_loss} is explained by the PF now being in a situation where all stability coefficients are negative, and the increase in eigenvalue $\lambda_0$ \subref{fig:change_learning_rate_eigen} is consistent with what occurs when the PF more closely follows the NGF, since the NGF tends to increase eigenvalues in that part of training, as observed in Figure~\ref{fig:instabilities_change_learning_rate_loss}. }
\label{fig:instabilities_change_learning_rate_discrete}
\end{figure}

\textbf{Decreasing the learning rate.} \citet{cohen2021gradient} show that if the edge of stability behaviour is reached and the learning rate is decreased, the training stabilises and $\lambda_0$ keeps increasing; we visualise this phenomenon in Figure~\ref{fig:instabilities_change_learning_rate_discrete}. The PF tells us that decreasing the learning rate entails going from $\Re[sc_0] \ge 0$  to $\Re[sc_0] \le 0$ since $\lambda_0 <2/h$ after the learning rate change. Since all stability coefficients are now negative, this reduces instability. The increase in $\lambda_0$ is likely due to the behaviour of the NGF in that area (as can be seen in Figure~\ref{fig:instabilities_change_learning_rate_loss} when changing from gradient descent training to the NGF in an edge of stability area leads to an increase of $\lambda_0$).

\begin{figure}[thb]
\centering
\begin{subfloat}[$\nabla_{\vtheta}E^T \vu$ predicted behaviour \\ under the PF based on $h \lambda_0$.]{
 \includegraphics[width=0.33\columnwidth]{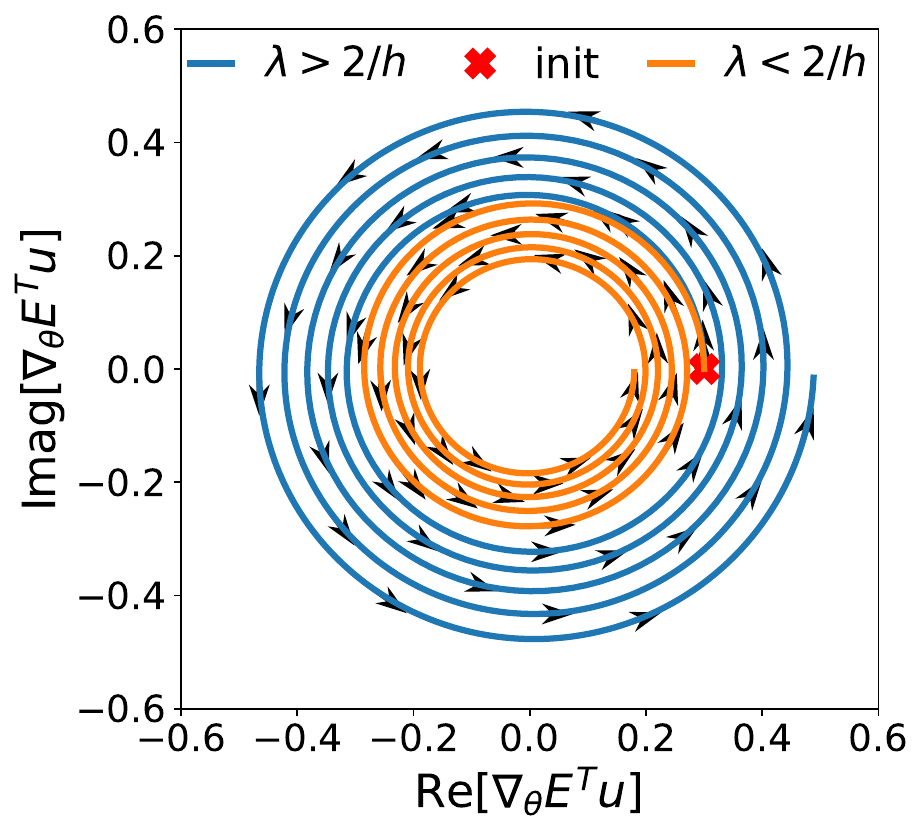}%
\label{fig:pf_grad_dot_prod_pred} 
}
\end{subfloat}%
\begin{subfloat}[$\nabla_{\vtheta}E^T \vu_0$ observed behaviour \\ in gradient descent training.]{
\includegraphics[width=0.322\columnwidth]{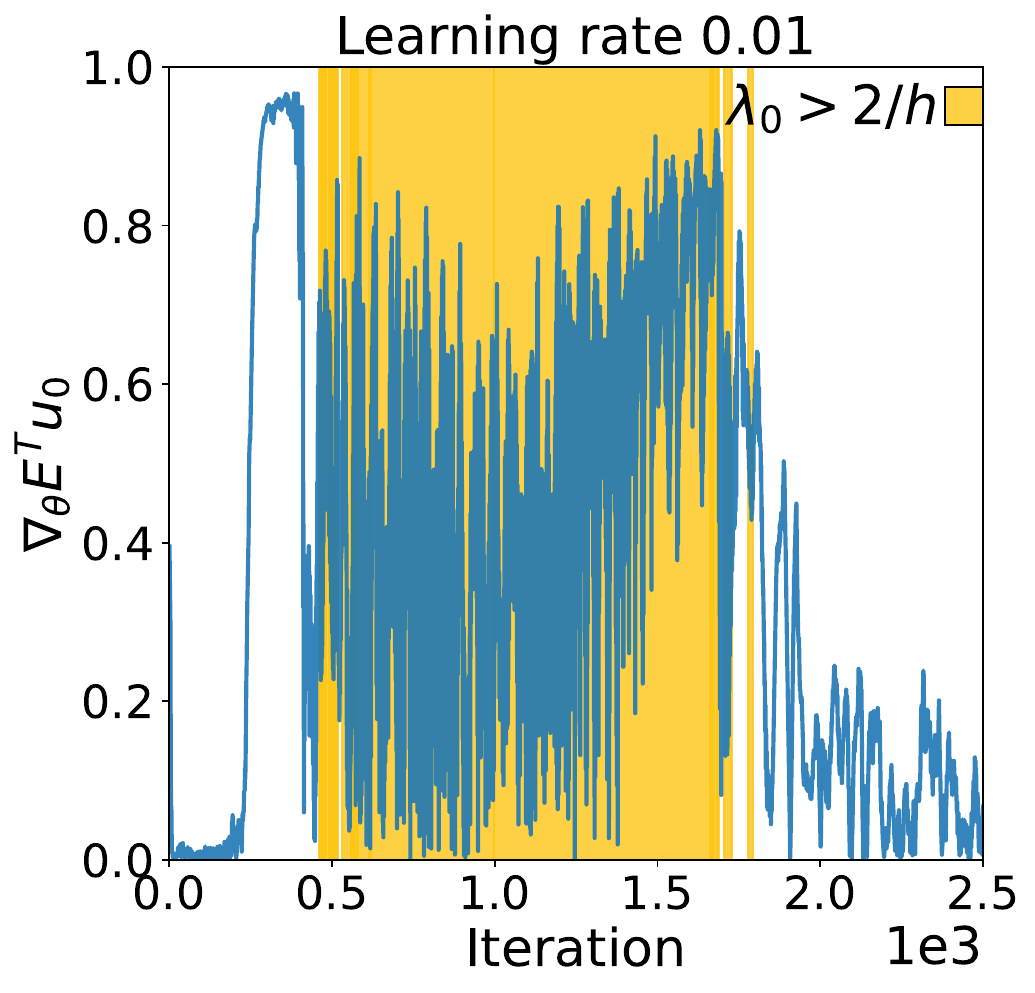}%
\label{fig:pf_grad_dot_prod_actual} 
}
\end{subfloat}%
\begin{subfloat}[Integration error in edge of stability area.]{
\includegraphics[width=0.33\columnwidth]{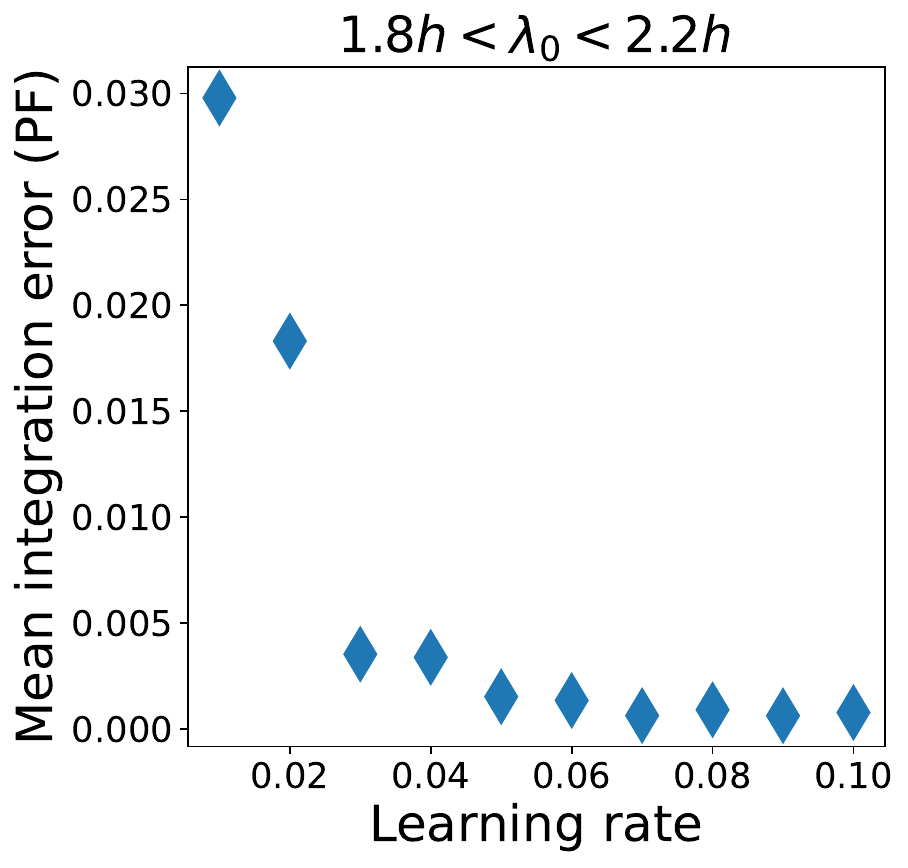}%
\label{fig:pf_grad_dot_prod_int_error} 
}
\end{subfloat}%
\caption[The unstable dynamics of $\nabla_{\vtheta}E^T \vu$ in the edge of stability area.]{\rebuttalrone{Predicting the unstable dynamics of $\nabla_{\vtheta}E^T \vu$  in the edge of stability area ($\lambda \approx 2/h)$ using the PF}. \subref{fig:pf_grad_dot_prod_pred}: the predicted behaviour of $\nabla_{\vtheta}E^T \vu$ under ${\frac{d \left(\nabla_{\vtheta}E^T \vu_i\right)}{d t} = \frac{\log(1 - h \lambda_i)}{h} \nabla_{\vtheta} E^T \vu_i}$, with an inflection point at $\lambda = 2/h$. \subref{fig:pf_grad_dot_prod_actual}: empirical behaviour of $\nabla_{\vtheta}E^T \vu$ for a model shows instabilities in the edge of stability area (highlighted). \subref{fig:pf_grad_dot_prod_int_error}: the approximation made to derive the flow is suitable around $\lambda \approx 2/h$. \subref{fig:pf_grad_dot_prod_actual} and \subref{fig:pf_grad_dot_prod_int_error} plots are obtained from training VGG models on CIFAR-10 with full-batch training.}
\label{fig:instability_largest_dot_prod}
\end{figure}

\textbf{The behaviour of $\nabla_{\vtheta}E^T \vu_0$.} The PF also allows us to explain the unstable behaviour of $\nabla_{\vtheta}E^T \vu_0$ around edge of stability areas. 
As done in Section~\ref{sec:nns_principal_flow}, we assume that $\lambda_i, \vu_i$ do not change substantially between iterations and write ${\frac{d \left(\nabla_{\vtheta}E^T \vu_i\right)}{d t} = \frac{\log(1 - h \lambda_i)}{h} \nabla_{\vtheta} E^T \vu_i}$ under the PF, with solution ${(\nabla_{\vtheta}E^T \vu_i)(t) = (\nabla_{\vtheta}E^T \vu_i)(0) e^{\frac{\log(1 - h \lambda_i)}{h}t}}$. 
This solution has different behaviour depending on the value of $\lambda_0$ relative to $2/h$: decreasing below $2/h$ and increasing above $2/h$. 
We show this theoretically predicted behaviour in Figure~\ref{fig:instability_largest_dot_prod}, 
alongside empirical behaviour showcasing the fluctuation of $\nabla_{\vtheta}E^T \vu_0$ in the edge of stability area, which confirms the theoretical prediction. We also compute the prediction error of the proposed flow and show it can capture the dynamics of $\nabla_{\vtheta}E^T \vu_0$ closely in this setting. We present a discrete-time argument for this observation in Section~\ref{sec:changes_in_dot_prod_discrete}. \rebuttalrthree{We note that the stable behaviour early in training together with the oscillatory behaviour of $\nabla_{\vtheta}E^T \vu_0$ in the edge of stability area, which we predict and observe can explain the results of~\citet{cohen2021gradient} on the behaviour of $\vtheta^T \vu_0$, since $\vtheta$ accumulates changes given by gradient updates.}

\rebuttalrthree{\textbf{Why not more instability?} To determine why there isn't more instability in the edge of stability area we have to consider that neural networks are not quadratic, which has two effects. Firstly, when following the PF the landscape changes slightly locally; this leads to changes in stability coefficients and thus the behaviour of gradient descent as we have consistently seen in the experiments in this section. Secondly, non-principal  terms can have an effect; while we do not know all non-principal terms in Section~\ref{sec:non_principal} we provide a justification for why the non-principal term we do know (Eq~\eqref{eq:pf_with_non_principal}) can have a stabilising effect by inducing a regularisation pressure to minimise $\lambda_i (\nabla_{\vtheta} E^T \vu_i)^2$ in certain parts of the training landscape.}

In this section we have shown the PF closely  predicts the behaviour of gradient descent in neural network training. \rebuttalrone{This has led to additional insights, including the importance of stability coefficients in determining instabilities in gradient descent (Figures~\ref{fig:edge_of_stability_results_local}, \ref{fig:instabilities_short}, \ref{fig:instabilities_resnet}), causally showing one eigendirection is sufficient to cause instability (Figure~\ref{fig:instabilities_change_learning_rate_loss}) and being able to closely predict the behaviour of the dot product between the gradient and the largest eigenvector (Figure~\ref{fig:instability_largest_dot_prod}). This evidence suggests that the PF captures significant aspects of the behaviour of gradient descent in deep learning; this is likely due to the specific structure of neural network models.}
While we take a continuous-time approach, a discrete-time approach can be used to motivate some of our observations (Section \ref{sec:discrete}); this is complementary to our approach but nonetheless related, since it also does not account for higher order derivatives of the loss and further suggests the strength of a quadratic approximation of the loss in the case of neural networks, as observed by \citet{cohen2021gradient}. 

\section{Non-principal terms can stabilise training}
\label{sec:non_principal}

This chapter focuses on understanding the effects of the PF on the behaviour of gradient descent.  
The principal terms, however, are not the only terms in the discretisation drift: we have found one non-principal term of the form $\nabla_{\vtheta} E^T (\nabla^3_{\vtheta} E) \nabla_{\vtheta} E$ (Eq~\eqref{eq:third_order_modified_vector_field}) and have seen that it can have a stabilising effect (Figure~\ref{fig:intuition_banana}). 
To provide some intuition, while $\nabla_{\vtheta} E^T (\nabla^3_{\vtheta}E) \nabla_{\vtheta} E$ cannot be written as a gradient operator we can write:
\begin{align}
\nabla_{\vtheta}E ^T \nabla_{\vtheta}^3 E \nabla_{\vtheta}E = \sum_{i=0}^{D-1} \left(\vu_i ^T \nabla_{\vtheta}^3 E \vu_i \right) (\nabla_{\vtheta}E ^T \vu_i) ^2  \approx \sum_{i=0}^{D-1}  \nabla_{\vtheta} \lambda_i (\nabla_{\vtheta}E ^T \vu_i) ^2,
\end{align}
where we used that $\vu_i$ does not change substantially around a GD iteration to write the derivative of $\lambda_i$, namely $\nabla_{\vtheta} \lambda_i = \vu_i ^T \nabla_{\vtheta}^3 E \vu_i$, since $\nabla_{\vtheta}^2 E$ is real and symmetric around a gradient descent iteration~\citep{petersen2008matrix}.
We can now write, again assuming $\vu_i$ is locally constant: 
\begin{align}
\sum_{i=0}^{D-1} \nabla_{\vtheta} \lambda_i (\nabla_{\vtheta}E ^T \vu_i) ^2  = \sum_{i=0}^{D-1} \nabla_{\vtheta} \left( \lambda_i (\nabla_{\vtheta}E ^T \vu_i) ^2 \right) - \sum_{i=0}^{D-1} 2 \lambda_i^2 (\nabla_{\vtheta}E ^T \vu_i) \vu_i.
\label{eq:non_principal_minimisation}
\end{align}
From here we can see that if $\lambda_i$ or $\nabla_{\vtheta}E ^T \vu_i$ are close to 0, the non-principal term leads to a pressure to minimise $ \lambda_i (\nabla_{\vtheta}E ^T \vu_i) ^2 $, since the non-gradient term $\lambda_i^2 (\nabla_{\vtheta}E ^T \vu_i) \vu_i$ in Eq~\eqref{eq:non_principal_minimisation} diminishes. This creates an incentive to keep the value of $ \lambda_i (\nabla_{\vtheta}E ^T \vu_i) ^2 $ around 0. We note that the work of~\citet{damian2022self} is what inspired us to write the third-order non-principal term in the form of Eq~\eqref{eq:non_principal_minimisation}, after we had previously noted its stabilising properties.

The incentive to keep $\lambda_i (\nabla_{\vtheta}E ^T \vu_i) ^2$ close to 0 can have a stabilising effect, since $\lambda_i = 0$ or $\nabla_{\vtheta} E ^T \vu_i=0$ results in $\alpha_{PF}(h\lambda_i) = \alpha_{NGF}(h \lambda_i) = -1$. Thus, minimising $\lambda_i (\nabla_{\vtheta} E ^T \vu_i)^2$ can reduce instability from the PF, since in those directions the PF has the same behaviour as the NGF.
We think this can shed light on observations about neural network training behaviour: the NGF tends to increase the eigenvalues, but as they reach close to 0 there is a pressure to minimise the above, which leads to $\lambda_i$ staying around 0; this is consistent with observations in the literature~\citep{sagun2017empirical,ghorbani2019investigation,papyan2018full}. This observation also explains the pressure for the dot product $\nabla_{\vtheta} E ^T \vu_0$ to stay small if it is initialised around 0, as we see is the case in the early in neural network training (see Figures~\ref{fig:early_training_eigen_loss} and~\ref{fig:instabilities_short}). Furthermore, as we show in Figure~\ref{fig:instability_largest_dot_prod} and previously argued, the dot product $\nabla_{\vtheta} E ^T \vu_0$ fluctuates around 0 in the edge of stability areas, making it likely for the minimisation effect induced by Eq~\eqref{eq:non_principal_minimisation} to kick in.

\begin{figure}[tb!]
\centering
\begin{subfloat}[$\frac{h^2}{12}\nabla_{\vtheta}E ^T \nabla_{\vtheta}^3 E \nabla_{\vtheta}E$]{
\includegraphics[width=0.45\columnwidth]{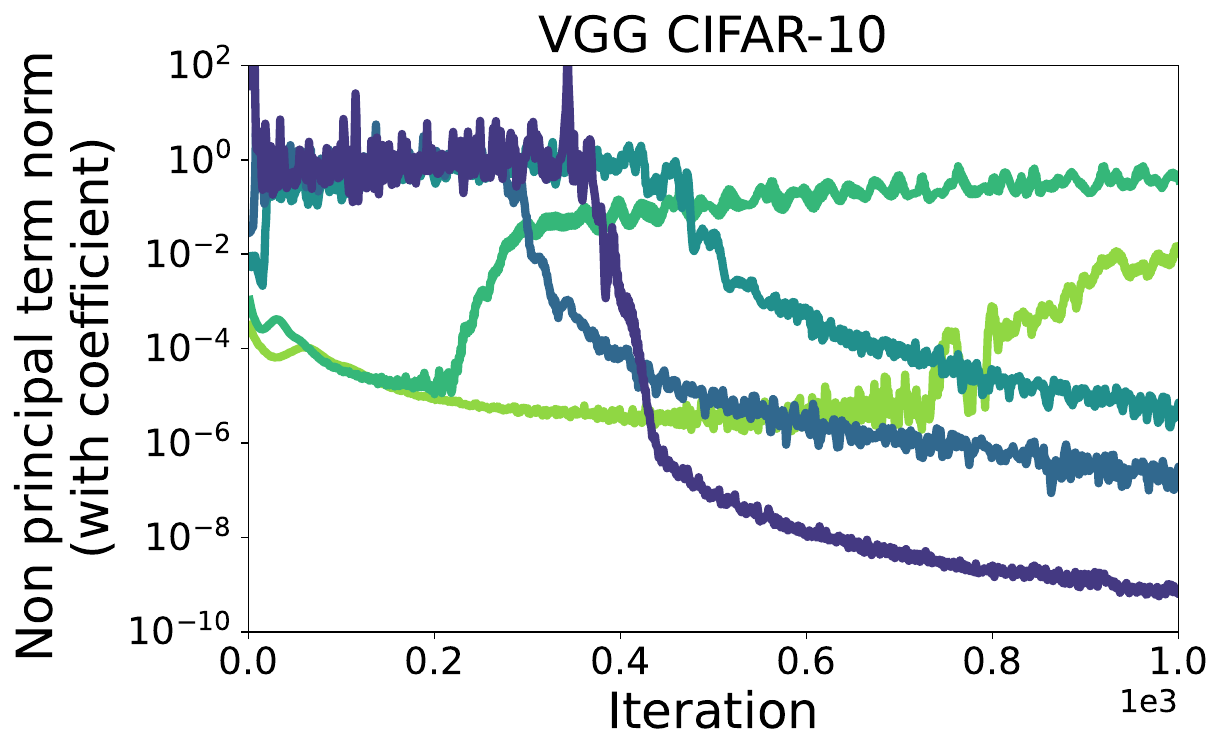}
}
\end{subfloat}%
\hspace{2em}
\begin{subfloat}[$\norm{\nabla_{\vtheta}E}$]{
\includegraphics[width=0.43\columnwidth]{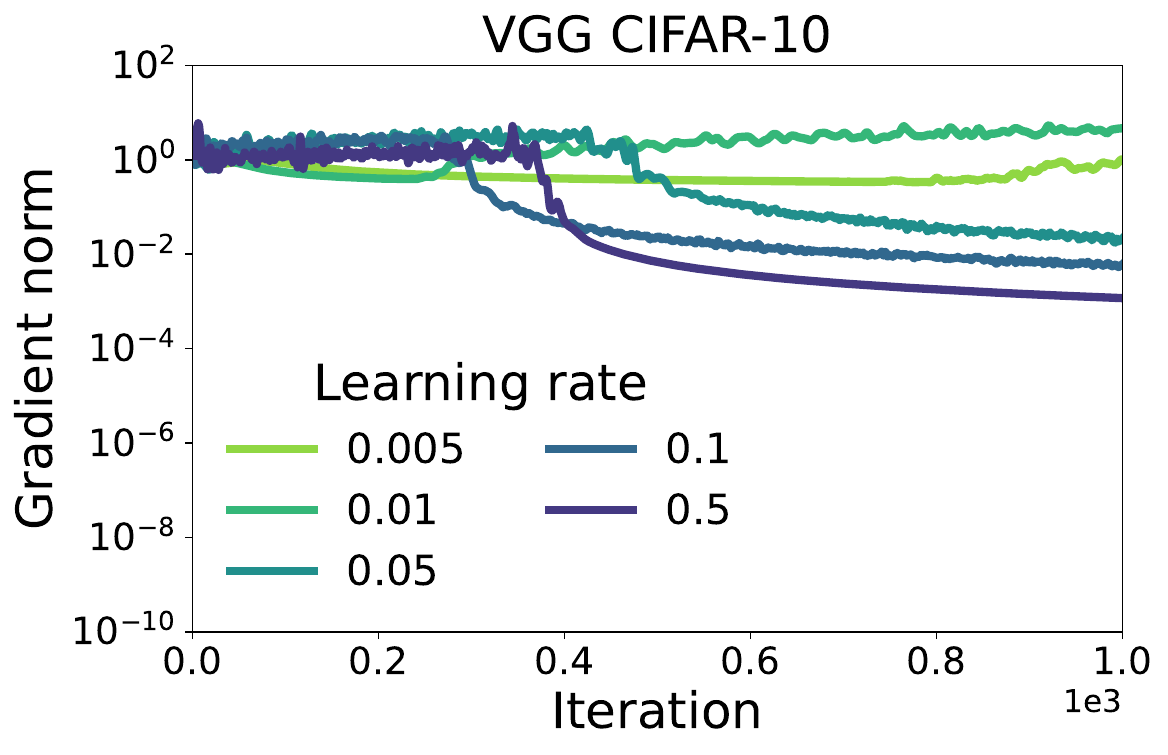}
}
\end{subfloat}%
\caption[The value of non-principal terms; a Resnet-18 trained on CIFAR-10 sweep.]{The value of the non-principal third-order term in training: outside the edge of stability areas (best seen for learning rates $0.005$ and $0.01$), the non-principal third-order term is very small compared to the gradient. Inside the edge of stability areas, the non-principal term has a higher magnitude. Corresponding loss and $\lambda_0$ plots are provided in Figure~\ref{fig:reproduce_edge_of_stability}.}
\label{fig:non_principal}
\end{figure}

We plot the value of this non-principal term in neural network training in Figure~\ref{fig:non_principal}.
These results show that the non-principal third-order term is very small outside the edge of stability area, but has a larger magnitude around the edge of stability, where the largest eigenvalues stop increasing but the magnitude of $(\nabla_{\vtheta} E(\vtheta_0)^T \vu_i)^2$ fluctuates and can be large  (Figure~\ref{fig:instability_largest_dot_prod}).
This observation is consistent with the above interpretation, but more theoretical and empirical work is needed to understand the effects of non-principal terms on training stability and generalisation.

\section{Stabilising training by accounting for discretisation drift}
\label{sec:stabilising_training}

The PF allows us to understand not only how gradient descent differs from the trajectory given by the NGF, but also when they follow each other closely. Understanding when gradient descent behaves like the NGF flow \rebuttalrone{reveals when the existing analyses of gradient descent using the NGF discussed in Section \ref{sec:motivation} are valid}. \rebuttalrone{It also has practical implications, since, in areas where gradient descent follows the NGF closely}, training can be sped up by increasing the learning rate. Prior works have empirically observed that gradient descent follows the NGF early in neural network training~\citep{cohen2021gradient}, and this observation can be used to explain why decaying learning rates~\citep{cosine_lr_decay} or learning rate warm up \citep{he2019bag} are successful when training neural networks: having a high learning rate in areas where the drift is small will not cause instabilities and can speed up training while decaying the learning rate avoids instabilities later in training when the drift is larger. 

\subsection{$\nabla_{\vtheta}^2 E \nabla_{\vtheta} E$ determines discretisation drift}

\label{sec:h_g}

In previous sections we have seen that the Hessian plays an important role in defining the PF \rebuttalrone{ and in training instabilities. We now want to quantify the difference between the NGF and the PF in order to understand when the NGF can be used as a model of gradient descent. We find that:} 
\begin{remark} In a region of the space where $\nabla_{\vtheta}^2 E \nabla_{\vtheta} E= \mathbf{0}$ the PF is the same as the NGF.
\end{remark}

 To see why, we can expand
\begin{align}
\nabla_{\vtheta}^2 E \nabla_{\vtheta} E= \sum_{i=0}^{D-1} \lambda_i \nabla_{\vtheta} E ^T \vu_i \vu_i.
\label{eq:expansion}
\end{align}
If $\nabla_{\vtheta}^2 E \nabla_{\vtheta} E = \mathbf{0}$ we have that $\lambda_j \nabla_{\vtheta} E ^T \vu_j = 0, \forall j \in \{1, \dots, D\}$, thus either $\lambda_j = 0$ leading to $\alpha_{NGF}(h\lambda_j) = \alpha_{PF}(h\lambda_j) = -1$ or $\nabla_{\vtheta} E ^T \vu_j = 0$. 
Then $\dot{\vtheta} =  \sum_{i=0}^{D-1} \alpha_{PF}(h\lambda_i)(\nabla_{\vtheta} E^T \vu_i)  \vu_i = \sum_{i=0}^{D-1} \alpha_{NGF}(h\lambda_i) (\nabla_{\vtheta} E^T \vu_i)  \vu_i$.

\rebuttalrone{Thus comparing the PF  with the NGF reveals an important quantity: $\nabla_{\vtheta}^2 E \nabla_{\vtheta} E$. Further investigating this quantity reveals it has a connection with the total drift, since:}

\begin{theorem} The discretisation drift (error between gradient descent and the NGF) after one iteration ${\vtheta_{t} = \vtheta_t - h \nabla_{\vtheta} E(\vtheta_{t-1})}$ is $\frac{h^2}{2} \nabla_{\vtheta}^2 E(\vtheta')  \nabla_{\vtheta} E(\vtheta')$ for a set of parameters $\vtheta'$ in the neighborhood of $\vtheta_{t-1}$.
\label{thm:total_drift}
\end{theorem}

This follows from the Taylor reminder theorem in mean value form (proof in Section~\ref{sec:proofs_total_per_iteration_drift}). This leads to:

\begin{corollary} In a region of space where $\nabla_{\vtheta}^2 E \nabla_{\vtheta} E=\mathbf{0}$ gradient descent follows the NGF.
\end{corollary}

\begin{figure}[tb!]
\centering
\captionsetup[subfigure]{justification=centering}
\begin{subfloat}[MNIST MLP.]{
\includegraphics[width=0.329\columnwidth]{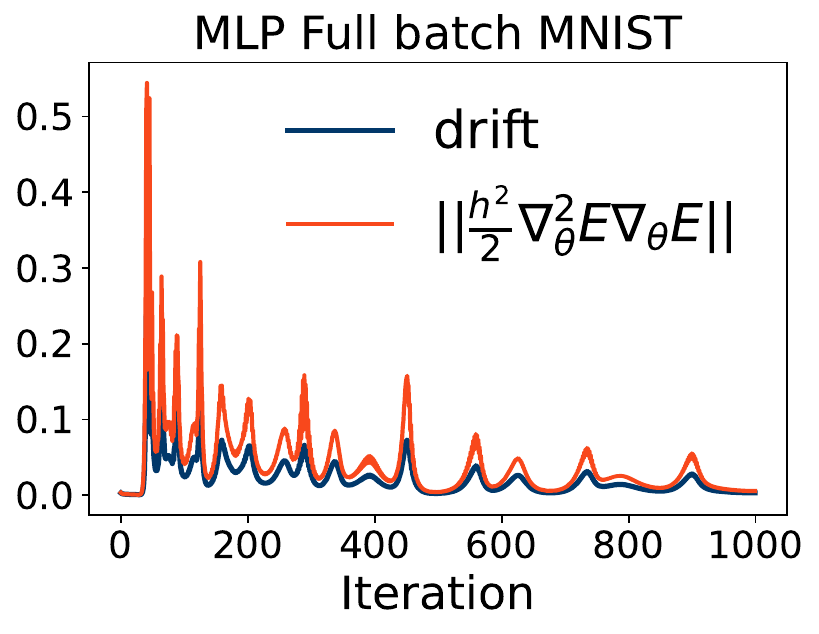}%
}\end{subfloat}%
\begin{subfloat}[CIFAR-10 VGG.]{
 \includegraphics[width=0.338\columnwidth]{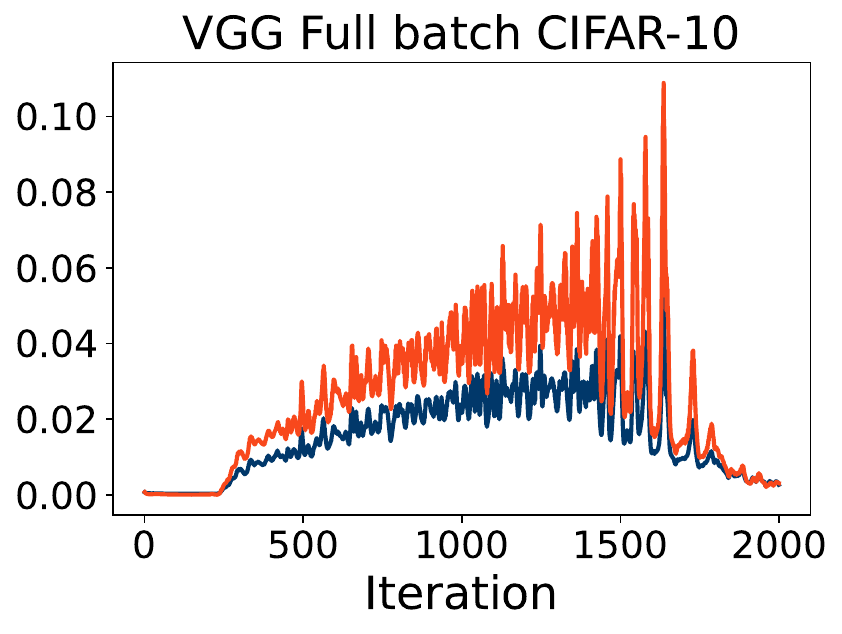}%
} \end{subfloat}%
\begin{subfloat}[CIFAR-10 VGG, \\ Batch Size 1024.]{
 \includegraphics[width=0.334\columnwidth]{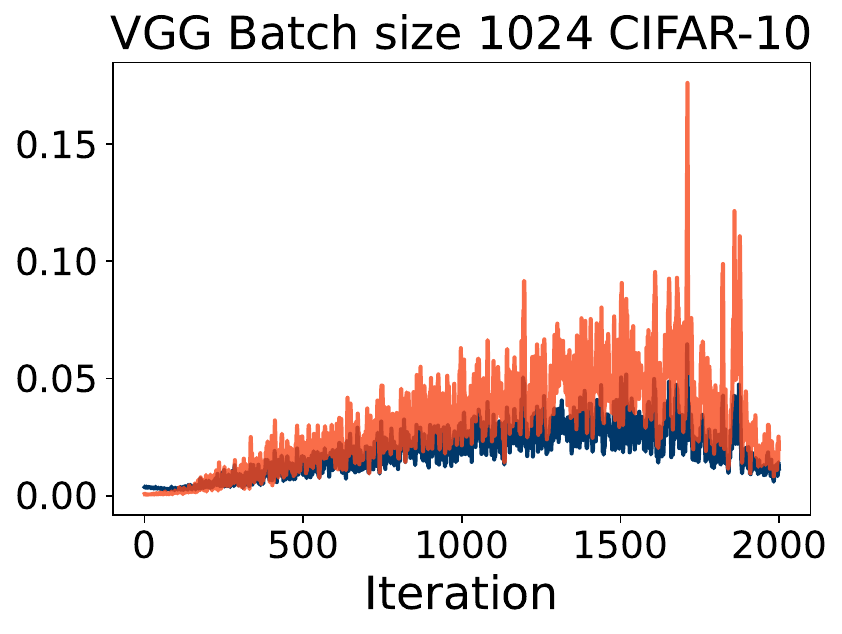}%
} \end{subfloat}%
\caption[$||\nabla_{\vtheta}^2 E \nabla_{\vtheta} E||$ and the per-iteration drift measured during training.]{ $||\nabla_{\vtheta}^2 E \nabla_{\vtheta} E||$ and the per-iteration drift have similar behaviour during training. As suggested by Theorem~\ref{thm:total_drift}, the norm of the per-iteration drift $\norm{\vtheta(h; \vtheta_{n-1}) - \vtheta_{n}}$ can be approximated by $||\nabla_{\vtheta}^2 E(\vtheta_{n-1}) \nabla_{\vtheta} E(\vtheta_{n-1})||$. The error in the approximation is due to the theorem providing an existence argument for a set of parameters $\vtheta'$ in the neighbourhood of $\vtheta_{n-1}$ such that $||\nabla_{\vtheta}^2 E(\vtheta') \nabla_{\vtheta} E(\vtheta')||$ is equal to the per-iteration drift; since we do not know those parameters we use the iteration parameters $\vtheta_{n-1}$ instead.}
\label{fig:h_g_drift}
\end{figure}
\begin{figure}[tb!]
\centering
\captionsetup[subfigure]{justification=centering}
\begin{subfloat}[MNIST MLP.]{
\includegraphics[width=0.333\columnwidth]{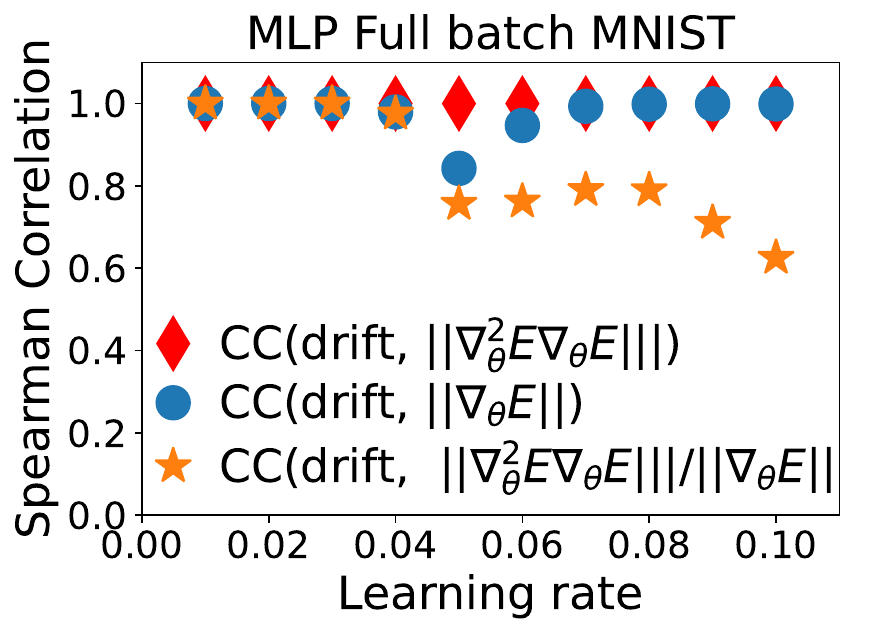}%
}\end{subfloat}%
\begin{subfloat}[CIFAR-10 VGG.]{
 \includegraphics[width=0.333\columnwidth]{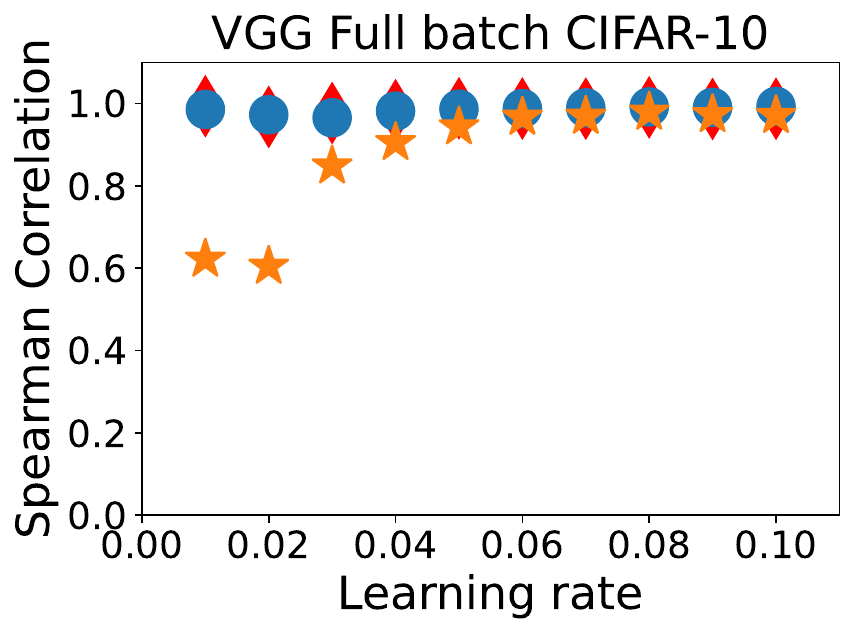}%
} \end{subfloat}%
\begin{subfloat}[CIFAR-10 VGG,\\ Batch Size 1024.]{
 \includegraphics[width=0.333\columnwidth]{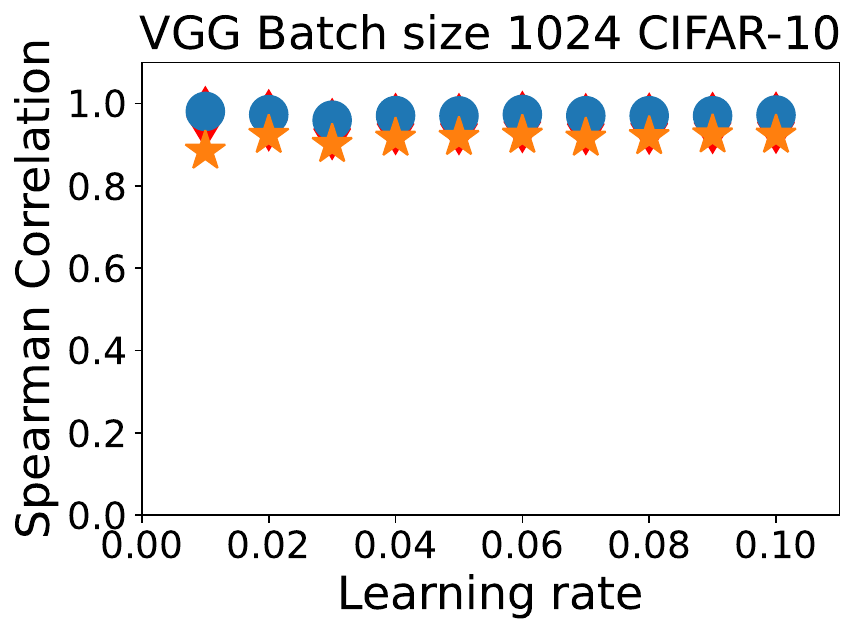}%
} \end{subfloat}%
\caption[Correlation between $||\nabla_{\vtheta}^2 E \nabla_{\vtheta} E||$ and the per-iteration drift.]{Correlation between $||\nabla_{\vtheta}^2 E \nabla_{\vtheta} E||$ and the per-iteration drift.  Since  $||\nabla_{\vtheta}^2 E \nabla_{\vtheta} E|| = \left(||\nabla_{\vtheta}^2 E \nabla_{\vtheta} E||\right) / || \nabla_{\vtheta} E|| || \nabla_{\vtheta} E||$, we measure the correlation between per-iteration drift and $\left(||\nabla_{\vtheta}^2 E \nabla_{\vtheta} E||\right) / || \nabla_{\vtheta} E||$ as well as $|| \nabla_{\vtheta} E||$. As Theorem~\ref{thm:total_drift} provides an existence argument for an unknown set of parameters $\vtheta'$ in the neighbourhood of $\vtheta_{n-1}$ such that $||\nabla_{\vtheta}^2 E(\vtheta') \nabla_{\vtheta} E(\vtheta')||$ is equal to the per-iteration drift $\norm{\vtheta(h; \vtheta_{n-1}) - \vtheta_{n}}$, we evaluate $||\nabla_{\vtheta}^2 E \nabla_{\vtheta} E||$ at parameters $\vtheta_{n-1}$.}
\label{fig:h_g_drift_spearman}
\end{figure}

Thus \rebuttalrone{the PF revealed $\nabla_{\vtheta}^2 E \nabla_{\vtheta} E$  as a core quantity in the discretisation drift of gradient descent. To further see the connection between with the PF consider that}
$\norm{\nabla_{\vtheta}^2 E \nabla_{\vtheta} E}^2= \norm{ \sum_{i=0}^{D-1}\lambda_i \nabla_{\vtheta} E ^T \vu_i \vu_i}^2 = \sum_{i=0}^{D-1}\norm{\lambda_i \nabla_{\vtheta} E ^T \vu_i}^2$; the higher each term in the sum, the higher the difference between the NGF and the PF.
To measure the connection between per-iteration drift and  $\norm{\nabla_{\vtheta}^2 E \nabla_{\vtheta} E}$  in neural network training we approximate it via $\norm{\vtheta_{t} - \widetilde{NGF}(\vtheta_{t-1}, h)}$ where $\widetilde{NGF}$ is the numerical approximation to the NGF initialised at $\vtheta_{t-1}$. 
Results in Figures~\ref{fig:h_g_drift} and~\ref{fig:h_g_drift_spearman} show the strong correlation between per-iteration drift and $\norm{\nabla_{\vtheta}^2 E \nabla_{\vtheta} E}$ throughout training and across learning rates. Since Theorem~\ref{thm:total_drift} tells us the form of the drift but not the exact value of $\vtheta'$, we have used $\vtheta_{t-1}$ instead to evaluate $\norm{\nabla_{\vtheta}^2 E \nabla_{\vtheta} E}$ and thus some error exists.

Understanding this connection is advantageous since computing discretisation drift is computationally expensive as it requires simulating the continuous-time NGF but computing $\norm{\nabla_{\vtheta}^2 E \nabla_{\vtheta} E}$ via Hessian-vector products is cheaper and approximations are available, such as 
$\nabla_{\vtheta}^2 E \nabla_{\vtheta} E \approx \frac{\nabla_{\vtheta} E(\vtheta + \epsilon \nabla_{\vtheta} E) - \nabla_{\vtheta} E(\vtheta)}{\epsilon}$
which only requires an additional backward pass~\citep{geiping2021stochastic}.

\subsection{Drift adjusted learning (DAL)}
\label{sec:dal}

A natural question to ask is how to use the correlation between  $\norm{\nabla_{\vtheta}^2 E \nabla_{\vtheta} E}$ and the per-iteration drift to improve training stability; $\norm{\nabla_{\vtheta}^2 E \nabla_{\vtheta} E}$ captures all the quantities we have shown to be relevant to instability highlighted by the PF: $\lambda_i$ and $\nabla_{\vtheta} E ^T \vu_i$ (Eq.~\eqref{eq:expansion}).
One way to use this information is to adapt the learning rate of the gradient descent update, such as using $\frac 2 {\norm{\nabla_{\vtheta}^2 E \nabla_{\vtheta} E}}$ as the learning rate. 
This learning rate slows down training when the drift is large---areas where instabilities are likely to occur---and it speeds up training in regions of low drift---areas where instabilities are unlikely to occur.
Computing the norm of the update provided by this learning rate shows a challenge, however, since $\frac{2 \norm{\nabla_{\vtheta} E}}{\norm{\nabla_{\vtheta}^2 E \nabla_{\vtheta} E}} \ge \frac{2 \norm{\nabla_{\vtheta} E}}{\lambda_0 \norm{\nabla_{\vtheta} E}} = \frac{2}{\lambda_0}$; this implies that when using this learning rate the norm of the gradient descent update will never be 0 and thus training will not result in convergence.
Furthermore, the magnitude of the parameter update will be independent of the gradient norm.
To reinstate the gradient norm, we propose using the learning rate
\begin{align}
h(\vtheta) = \frac{2}{\norm{\nabla_{\vtheta}^2 E \nabla_{\vtheta} E}/ \norm{\nabla_{\vtheta} E}} = \frac{2}{\norm{\nabla_{\vtheta}^2 E \hat{\vg}(\vtheta)}}, \label{eq:dal}
\end{align}
where $\hat{\vg}(\vtheta)$ is the unit normalised gradient $\nabla_{\vtheta} E / \norm{\nabla_{\vtheta} E}$. We will call this learning rate \textbf{DAL} (Drift Adjusted Learning).
As shown in Figure~\ref{fig:h_g_drift},  $\norm{\nabla_{\vtheta}^2 E \hat{\vg}(\vtheta)}$ has a strong correlation with the per-iteration drift.
 \rebuttalrthree{Another interpretation of DAL can be provided through a signal-to-noise perspective: the size of the learning signal obtained by minimising $E$ is that of the update $h \norm{\nabla_{\vtheta}E}$, while the norm of the noise coming from the drift can be approximated as $\frac{h^2}{2} \norm{\nabla_{\vtheta}^2E \nabla_{\vtheta}E}$. Thus, the `signal-to-noise ratio' can be approximated as $h \norm{\nabla_{\vtheta}E}/({\frac{h^2}{ 2} \norm{\nabla_{\vtheta}^2E \nabla_{\vtheta}E}}) =2/({h \norm{\nabla_{\vtheta}^2 E \hat{\vg}(\vtheta)}})$, which when using DAL (Eq~\eqref{eq:dal}) is 1; DAL can be seen as balancing the gradient signal and the regularising drift noise in gradient descent training.}
\begin{figure}[tb!]
\centering
\begin{subfloat}[VGG, CIFAR-10.]{
  \includegraphics[width=0.52\columnwidth]{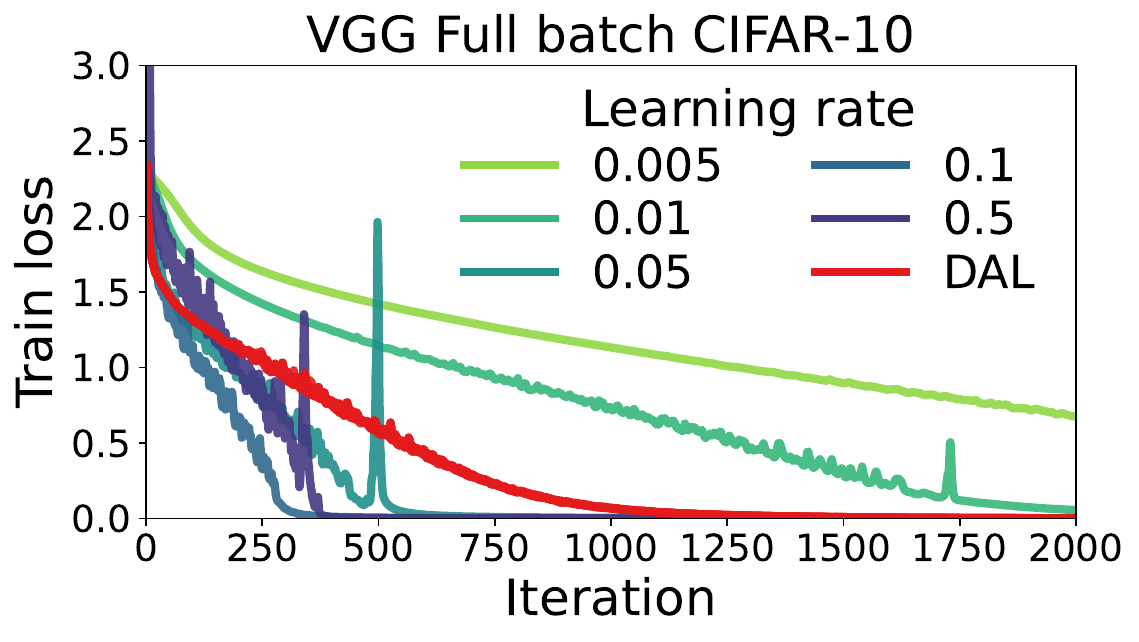}%
  }
\end{subfloat}%
\begin{subfloat}[Resnet-50, Imagenet.]{
  \includegraphics[width=0.48\columnwidth]{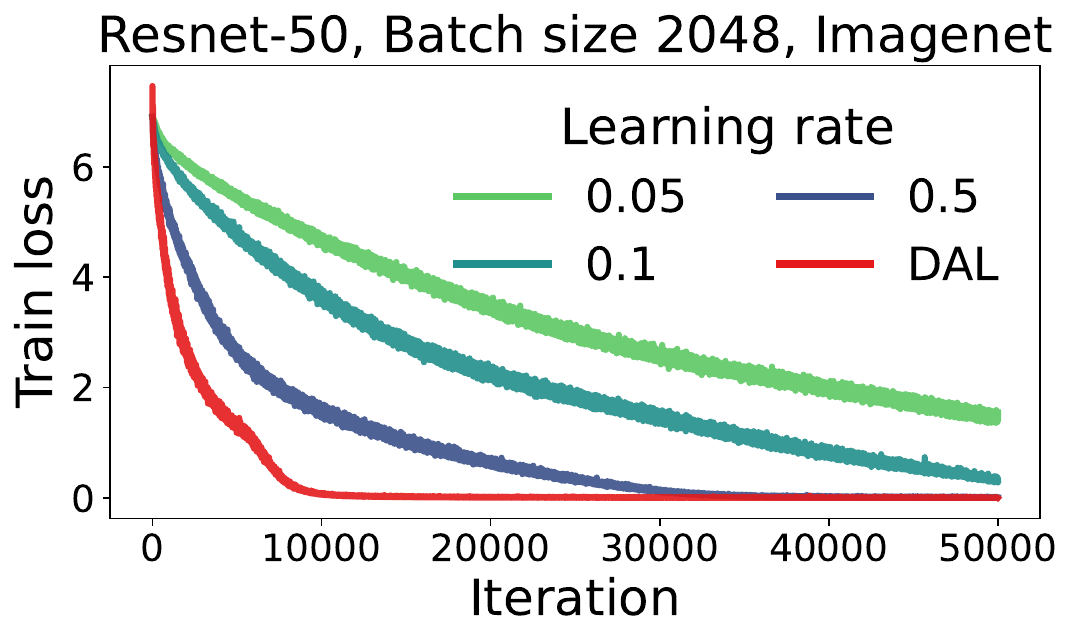}%
  }
\end{subfloat}%
\caption[Models trained using a learning rate sweep or DAL on CIFAR-10 and Imagenet. Models are VGG and Resnet-50.]{DAL: using the learning rate $\frac{2}{\norm{\nabla_{\vtheta}^2 E \hat{\vg}(\vtheta)}}$ results in improved stability without requiring a hyperparameter sweep; Resnet-18 results on CIFAR-10 in Figure~\ref{fig:dal_resnet} in the Appendix.}
\label{fig:dal}
\end{figure}

\begin{figure}[tb!]
\centering
\begin{subfloat}[Learning rate.]{
 \includegraphics[width=0.5\columnwidth]{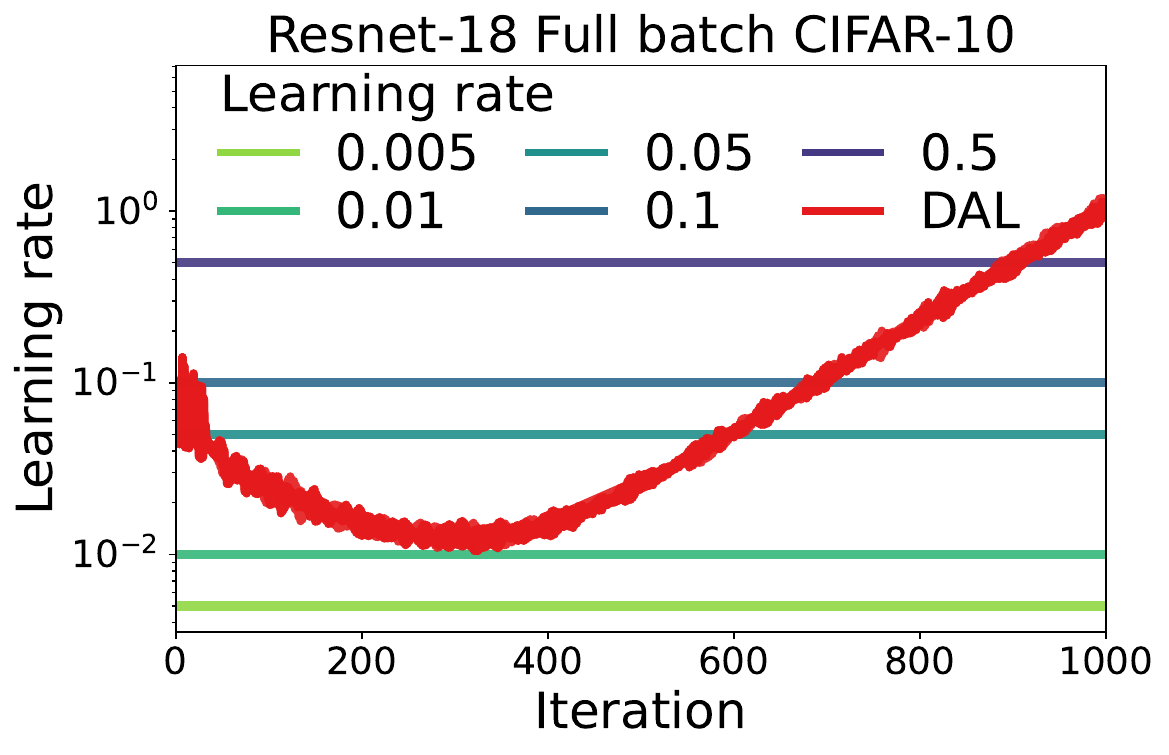}%
\label{fig:lr_dal_scaling}
}
\end{subfloat}%
\begin{subfloat}[Update norm.]{
 \includegraphics[width=0.5\columnwidth]{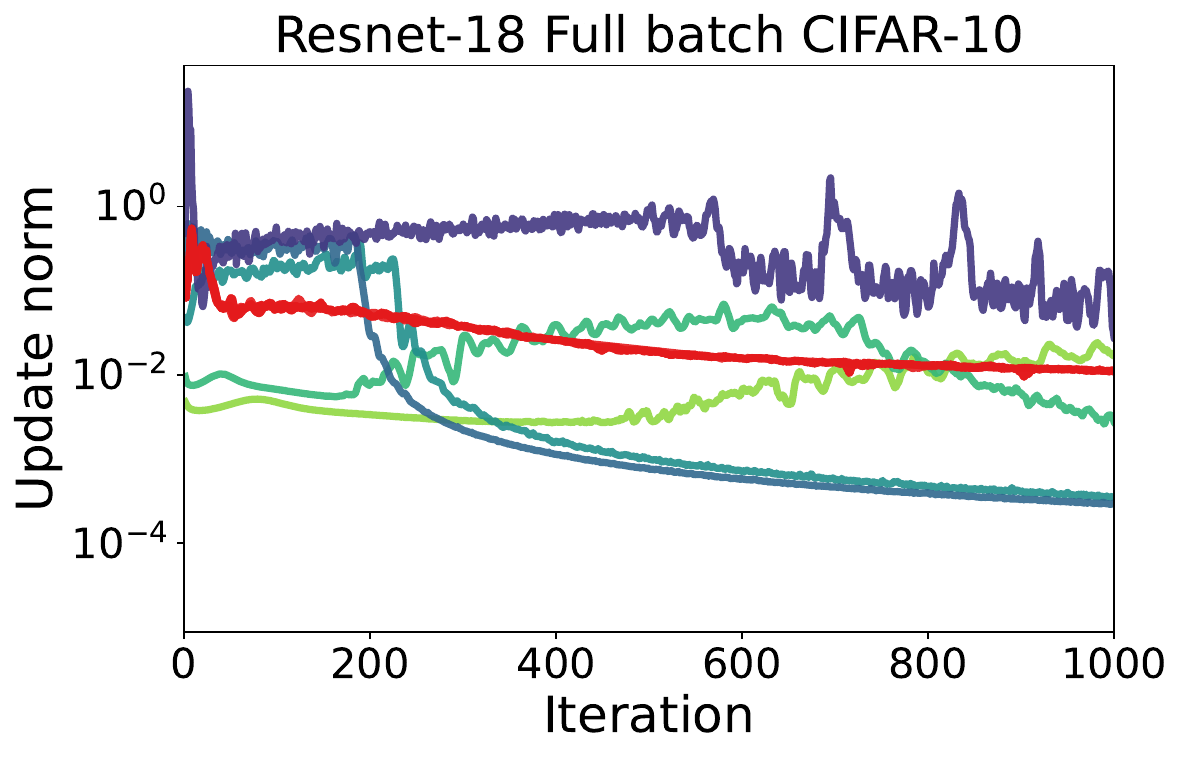}%
\label{fig:update_norm_dal_scaling}
 }
\end{subfloat}%
\caption[Learning rate and update norms in DAL compared to fixed learning rate training; a Resnet-18 model trained on CIFAR-10.]{Key quantities in DAL versus fixed learning rate training: \subref{fig:lr_dal_scaling} shows the learning rate $h(\vtheta) $ which is either constant or $\frac{2}{\norm{\nabla_{\vtheta}^2 E \hat{\vg}(\vtheta)}}$ for DAL, and \subref{fig:update_norm_dal_scaling} shows update norms $h(\vtheta) \norm{\nabla_{\vtheta} E}$. Unlike using a fixed learning rate, DAL adjusts to the training landscape and even though later in training the learning rate increases, this does not lead to an increase in the update norm.}
\label{fig:l_r_scaling_quantities}
\end{figure}

We use DAL to set the learning rate and show results  across architectures, models, and datasets in Figure~\ref{fig:dal} (with additional results in Figure~\ref{fig:imagenet_lr_scaling_across_batch_sizes} in the Appendix). \textit{Despite not requiring a learning rate sweep, DAL is stable compared to using fixed learning rates}. 
To provide intuition about DAL, 
we show the learning rate and the update norm in Figure~\ref{fig:l_r_scaling_quantities}:
for DAL the learning rate decreases in training after which it slowly increases when reaching areas with low drift. Compared to larger learning static learning rates where the update norm can increase in the edge of stability area with DAL the update norm steadily decreases in training.

\subsection{The trade-off between stability and performance}
\label{sec:trade_off}

\begin{figure}[tb!]
\begin{subfloat}[VGG, CIFAR-10.]{
  \includegraphics[width=0.5\columnwidth]{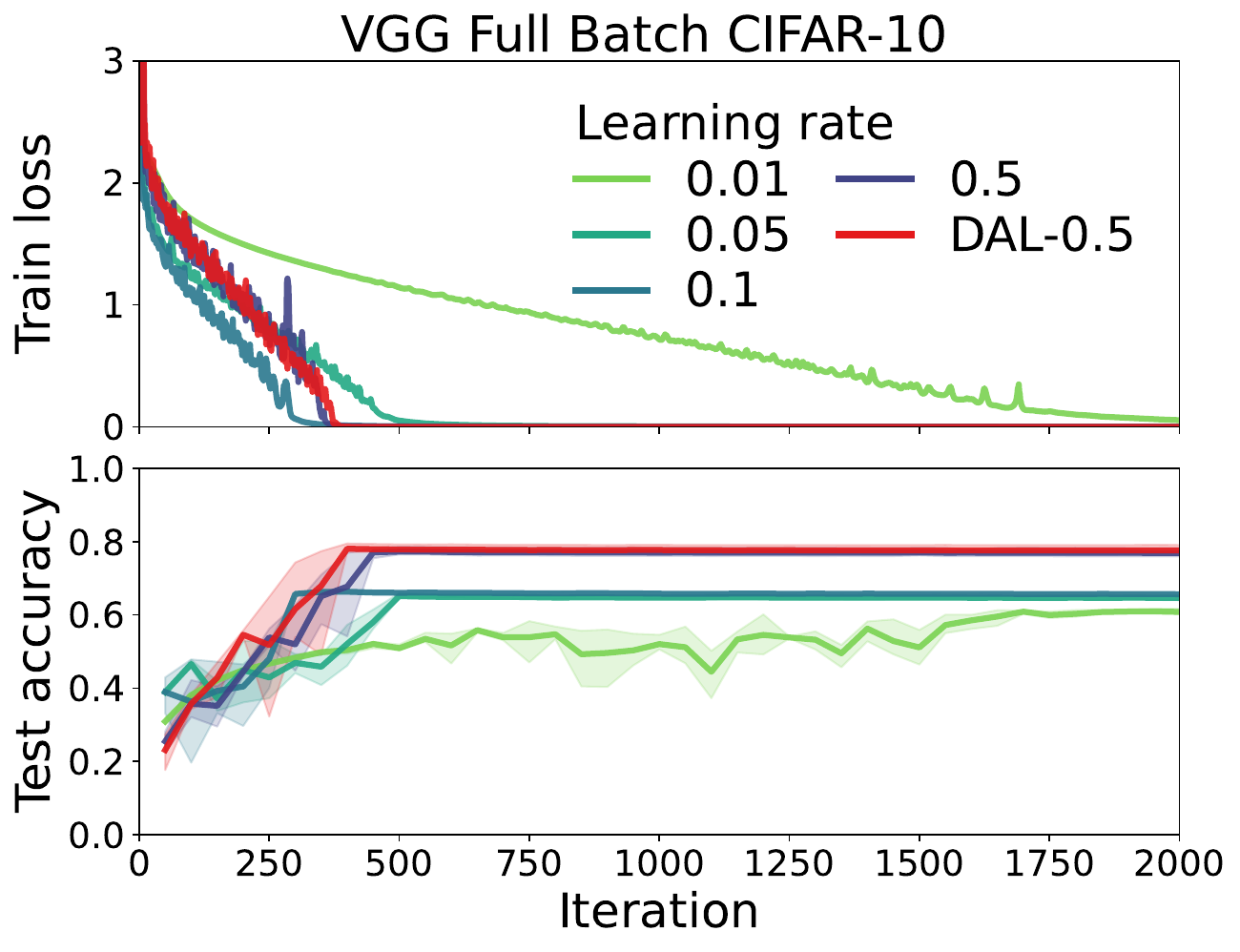}%
}
\end{subfloat}%
\begin{subfloat}[Resnet-50, Imagenet.]{
  \includegraphics[width=0.5\columnwidth]{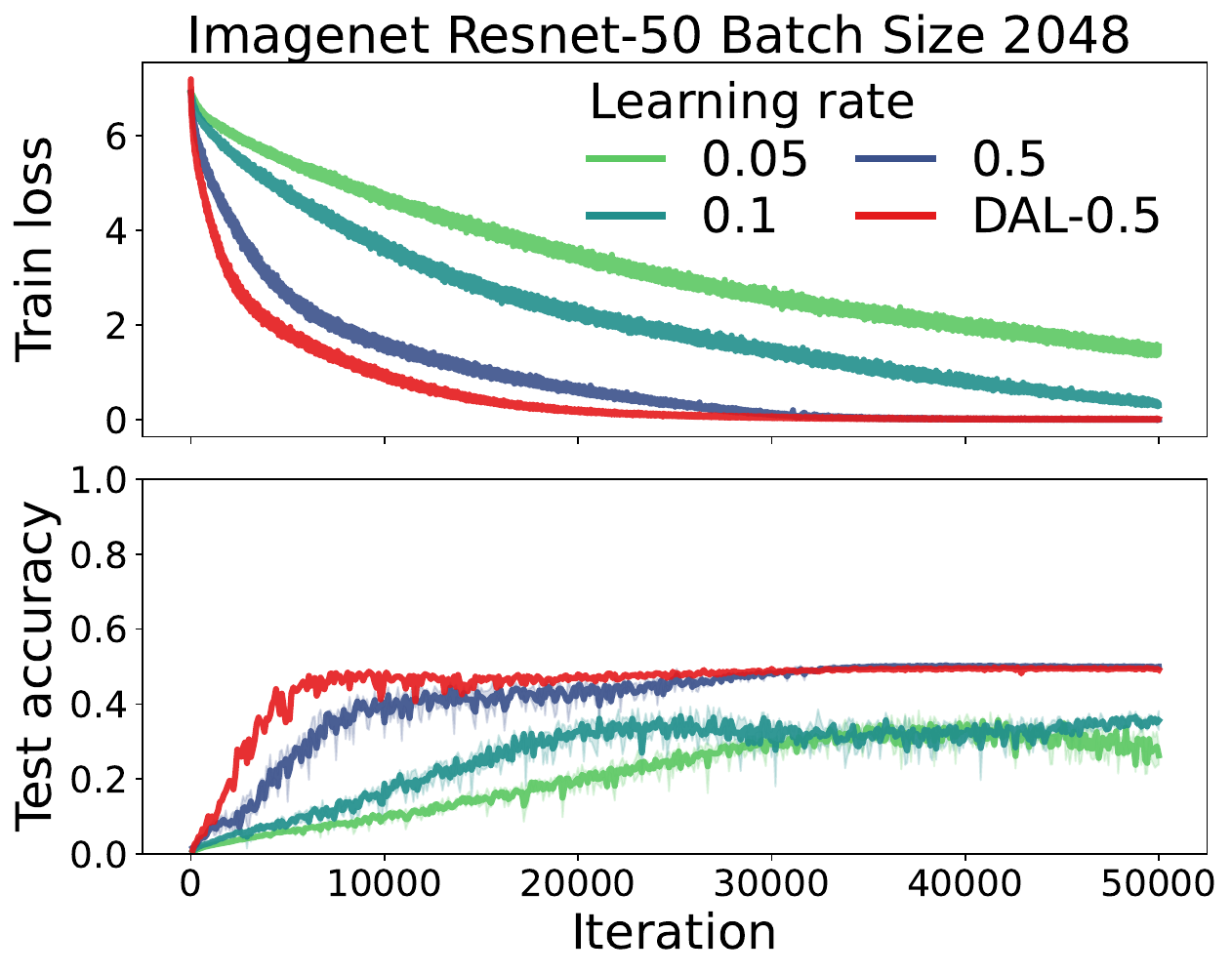}%
}
\end{subfloat}%
  \caption[DAL-0.5 results on a CIFAR-10 and Imagenet using  VGG, and Resnet-50 model respectively.]{DAL-0.5 leads to increased training speed and generalisation compared to a sweep of fixed learning rates. Resnet-18 results on CIFAR-10 in Figure~\ref{fig:dal_resnet} in the Appendix.}
\label{fig:l_r_scaling_sqrt}
\end{figure}

Since we are interested in understanding the  optimisation dynamics of gradient descent, we have so far focused on training performance. We now try to move our attention to test performance and generalisation. Previous works~\citep{li2019towards,igr,jastrzebski2019break} have shown that higher learning rates lead to better generalisation performance. We now try to further connect this information with the per-iteration drift and DAL.
To do so, we use learning rates with various degrees of sensitivity to per-iteration drift using DAL-$p$:
\begin{align}
 h_p(\vtheta) = \frac{2}{\left(\norm{\nabla_{\vtheta}^2 E \hat{\vg}(\vtheta)}\right)^p}.
\end{align}
The higher $p$, the slower the training and less drift there is; the lower $p$, there is more drift. We start with extensive experiments with $p=0.5$, which we show in Figure~\ref{fig:l_r_scaling_sqrt}, and show more results in Figure~\ref{fig:imagenet_lr_sqrt_scaling_across_batch_sizes}. Compared to $p=1$ (DAL), there is faster training but at times also more instability. Performance on the test set shows that DAL-$0.5$ performs as well or better than when using fixed learning rates.

\begin{remark}
We find that across datasets and batch sizes, DAL-$0.5$ performs best in terms of the stability generalisation trade-off and in these settings it can be used as a drop-in replacement for a learning rate sweep.
\end{remark}

\begin{figure}[tb!]
\begin{subfloat}[VGG, CIFAR-10.]{
  \includegraphics[width=0.51\columnwidth]{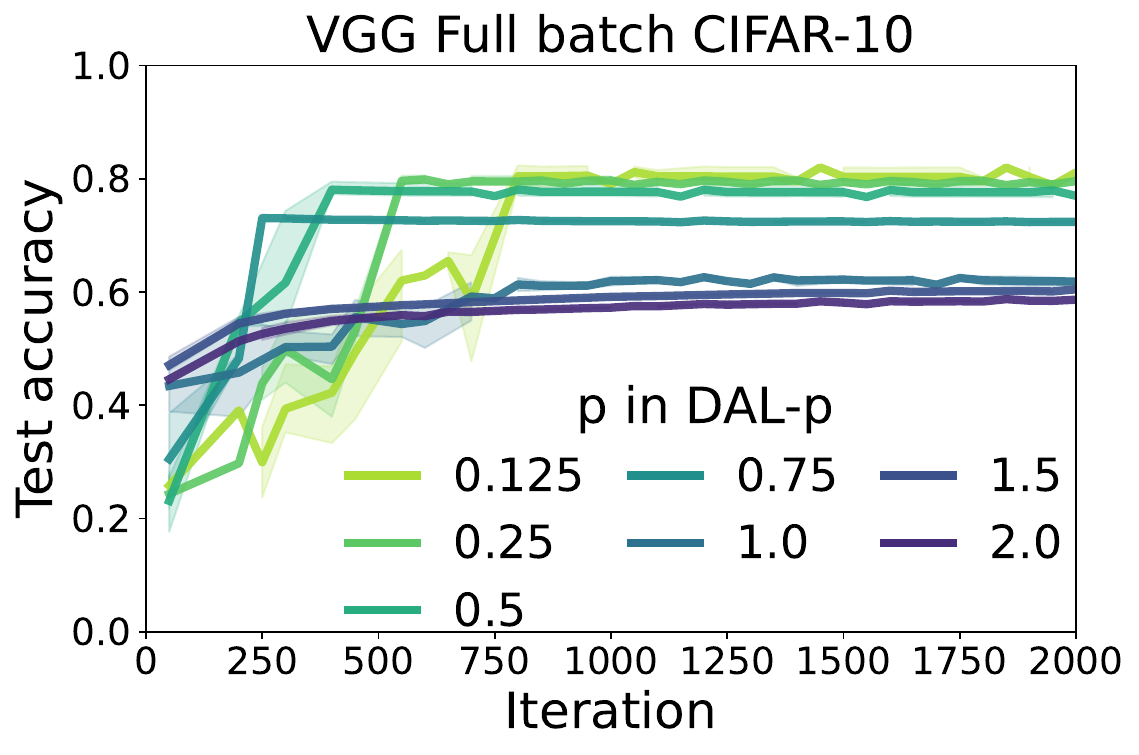}%
}%
\end{subfloat}%
\begin{subfloat}[Resnet-50, Imagenet.]{
 \includegraphics[width=0.49\columnwidth]{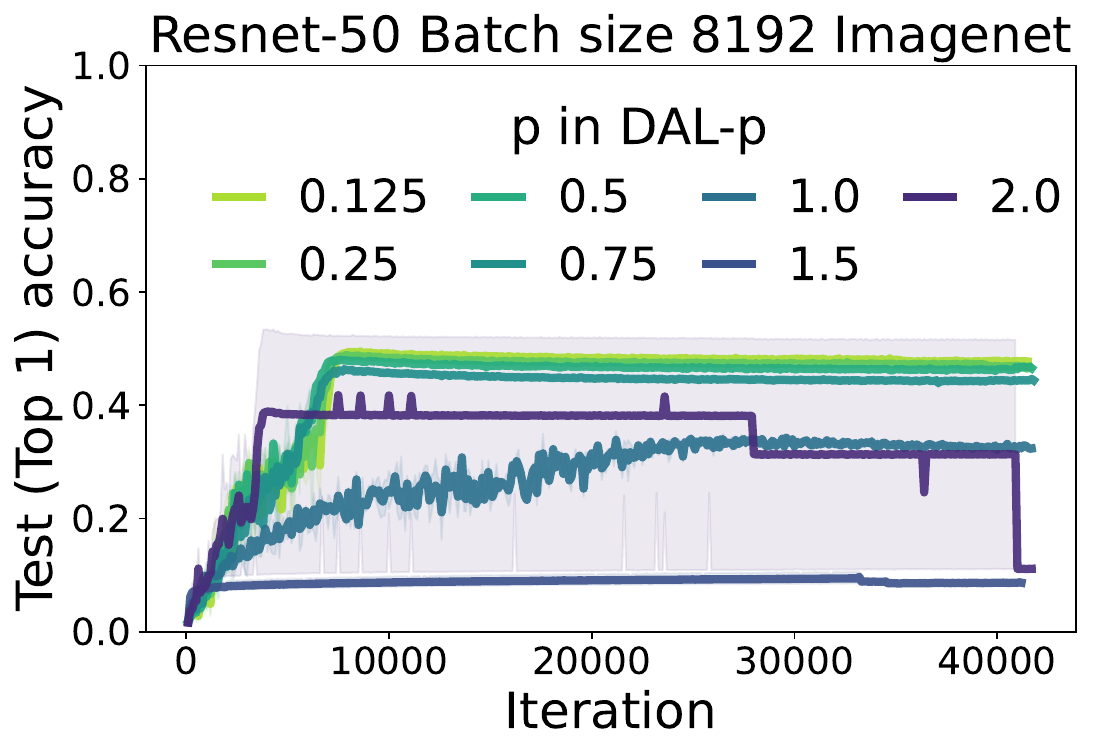}%
}%
\end{subfloat}%
\caption[DAL-$p$ sweep on full-batch training on CIFAR-10 with VGG and Resnet-18 models and a Resnet-50 model trained on Imagenet.]{DAL-$p$ sweep: discretisation drift helps test performance at the cost of stability. Corresponding training curves and loss functions are present in the Figure~\ref{fig:power_sweeps_all} in the Appendix; results showing the same trends across various batch sizes are shown in Figure~\ref{fig:power_sweeps_batch_sizes_vgg}.}
\label{fig:power_sweeps}
\end{figure}

\begin{figure}[tb!]
\centering
\begin{subfloat}[Fixed learning rate sweep.]{
 \includegraphics[width=0.45\columnwidth]{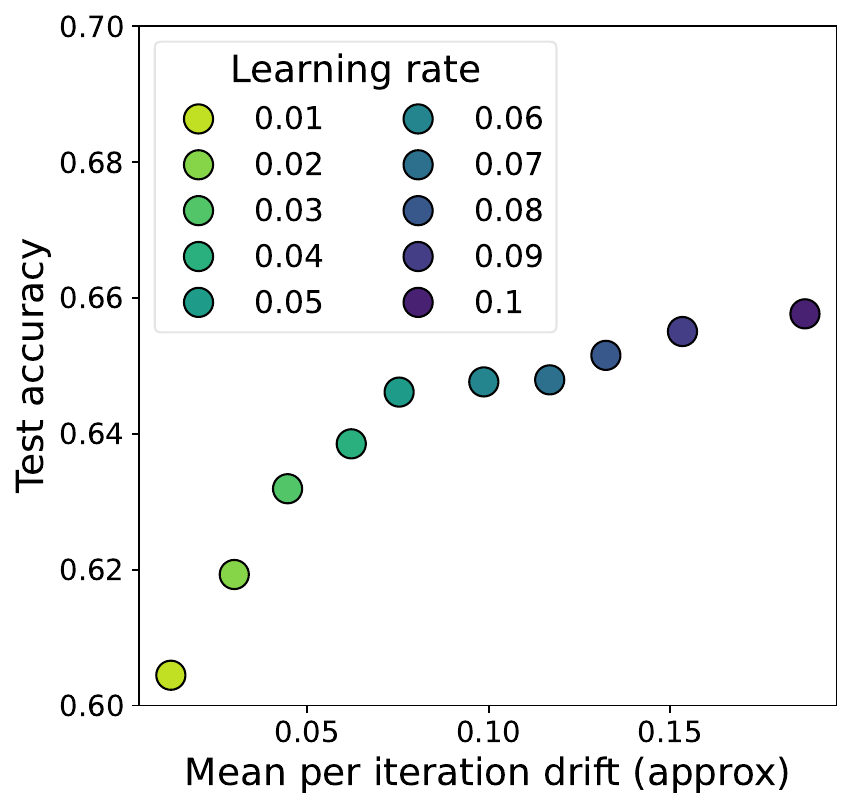}%
  \includegraphics[width=0.45\columnwidth]{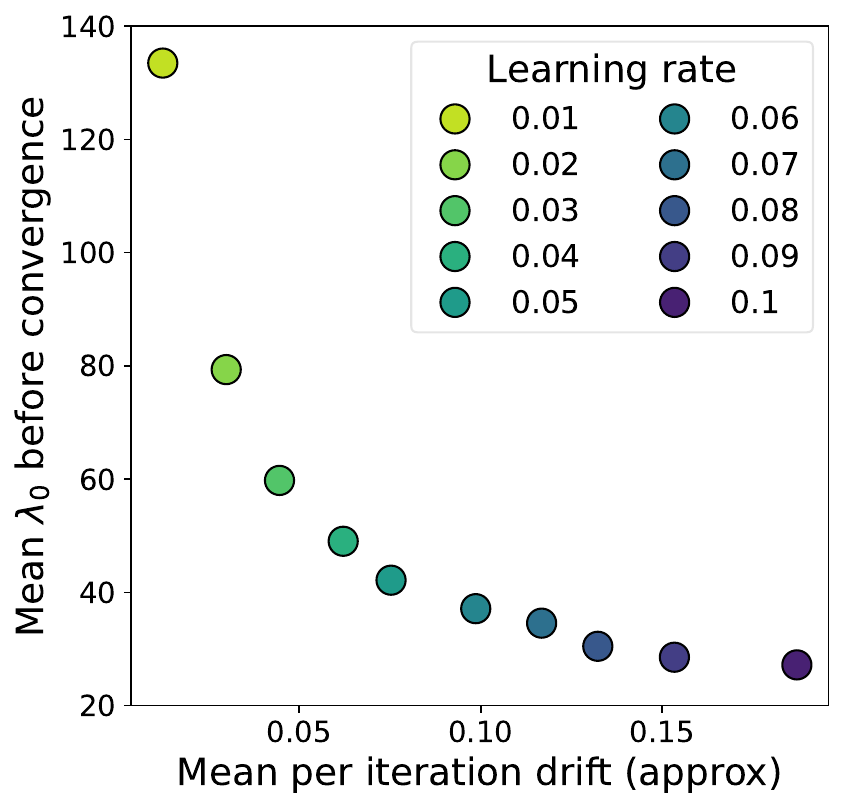}%
\label{fig:fixed_lr_sweep_dal_comp}
}\end{subfloat}
\begin{subfloat}[DAL-$p$ sweep.]{
  \includegraphics[width=0.45\columnwidth]{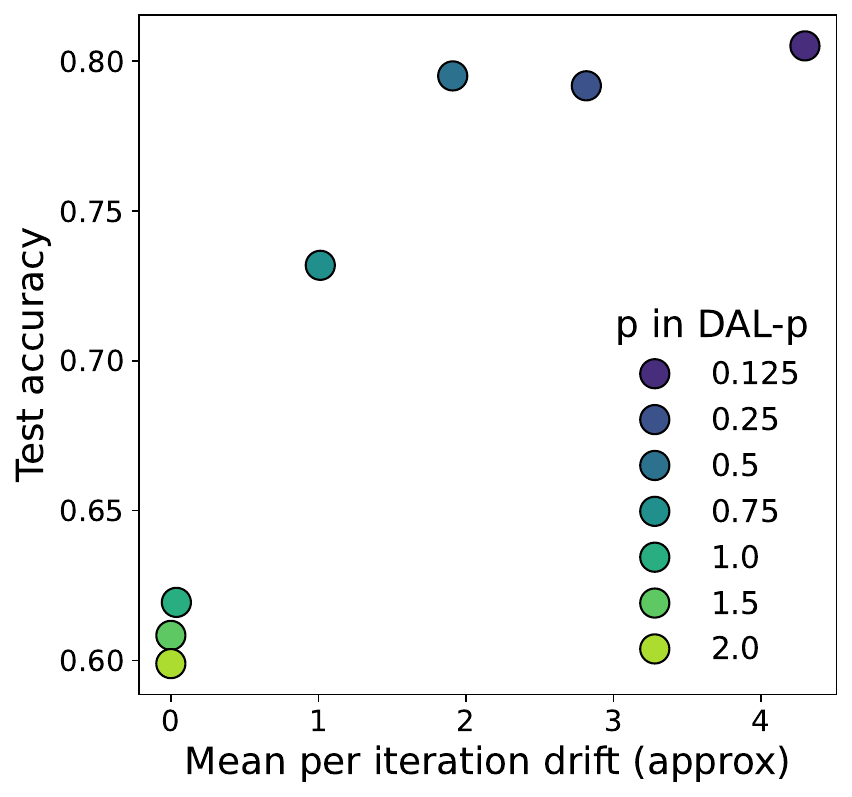}%
  \includegraphics[width=0.45\columnwidth]{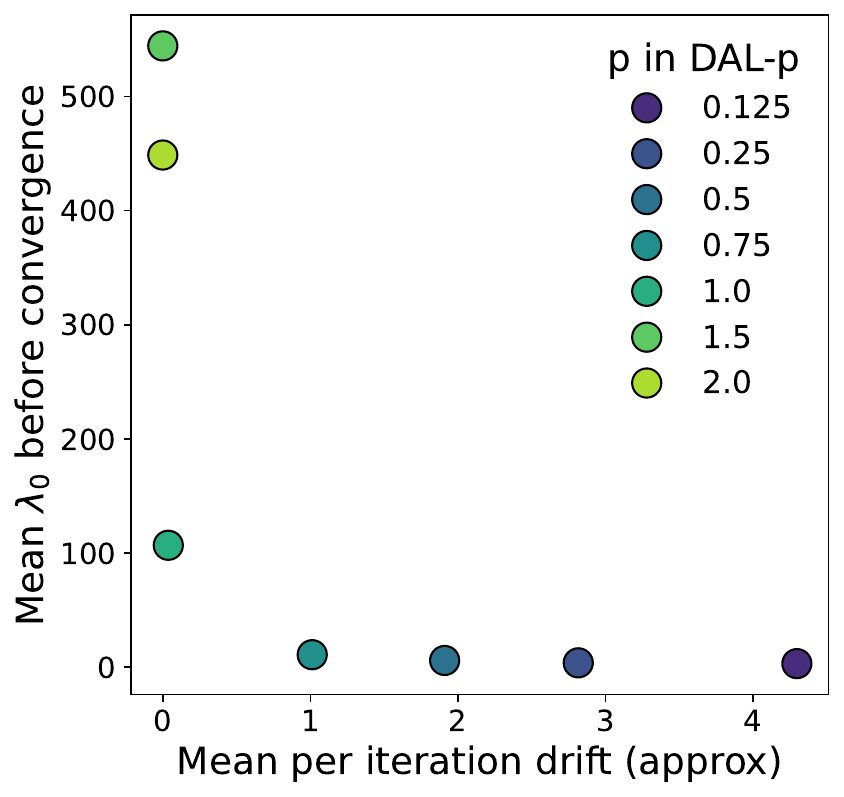}%
\label{fig:dal_sweep_drift}
}\end{subfloat}
\caption[VGG model trained on CIFAR-10 using full batch gradient descent, either with a fixed learning rate in a sweep or a DAL-$p$ sweep.]{The correlation between per-iteration drift, test set performance and $\lambda_0$ in full batch training on CIFAR-10. With fixed learning rates \subref{fig:fixed_lr_sweep_dal_comp}, the drift increases as the learning rate increases which leads to improved test performance and lower $\lambda_0$ on average in training. When using DAL \subref{fig:dal_sweep_drift}, the drift is controlled through the dynamic learning rate which adjusts to the magnitude of the drift; here too we observe that with high drift leads to increased performance and lower $\lambda_0$ on average in training. The same pattern can be seen in results with mini-batch training in Figure \ref{fig:drift_lambda_sgd_connection}.} 
\label{fig:stability_lambda_performance}
\end{figure}
\begin{figure}[tb!]
\centering
\begin{subfloat}[DAL]{
\includegraphics[width=0.49\columnwidth]{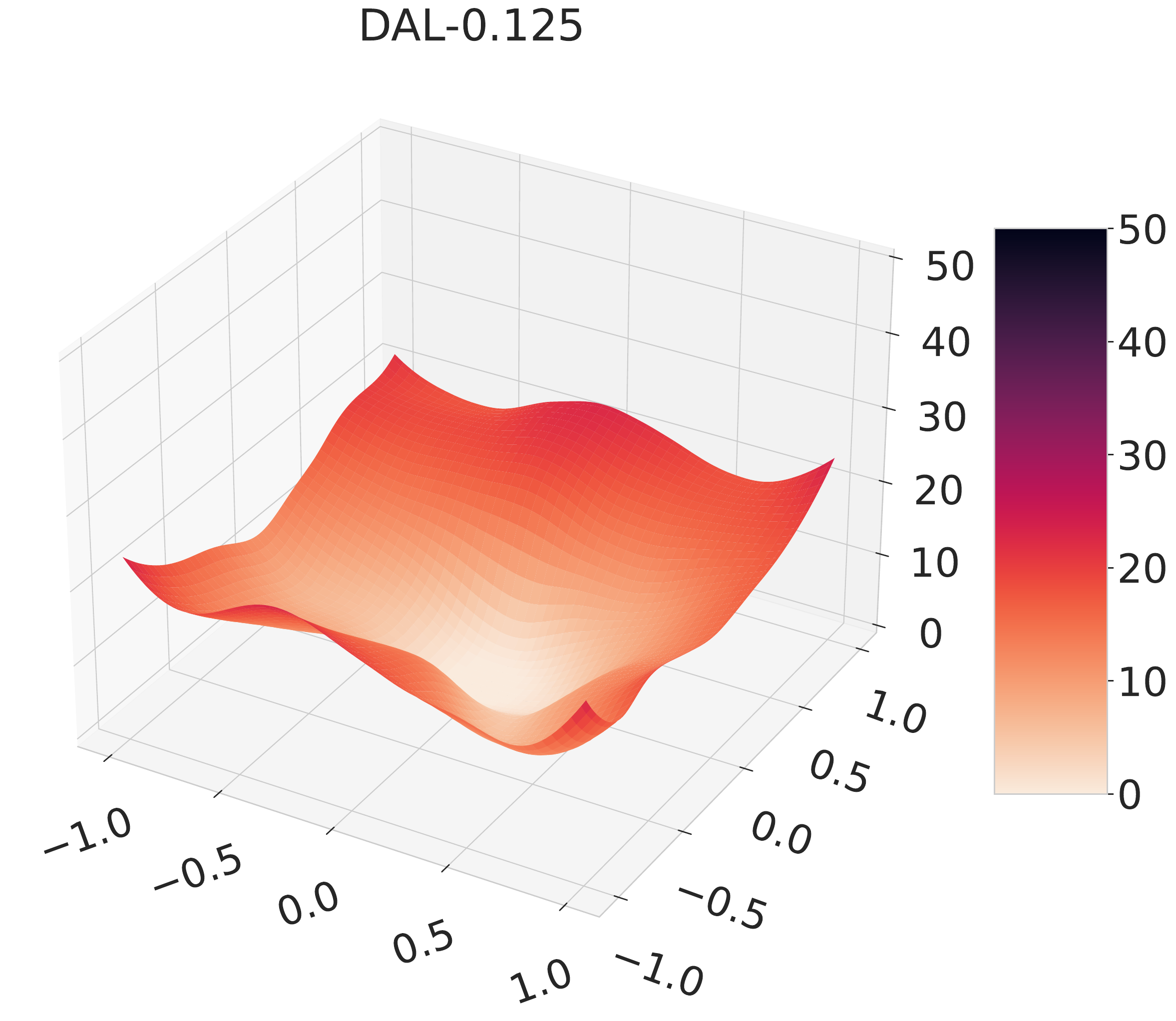}
\label{fig:landscape_dal_125bs_64}
}%
\end{subfloat}%
\begin{subfloat}[SGD]{
\includegraphics[width=0.49\columnwidth]{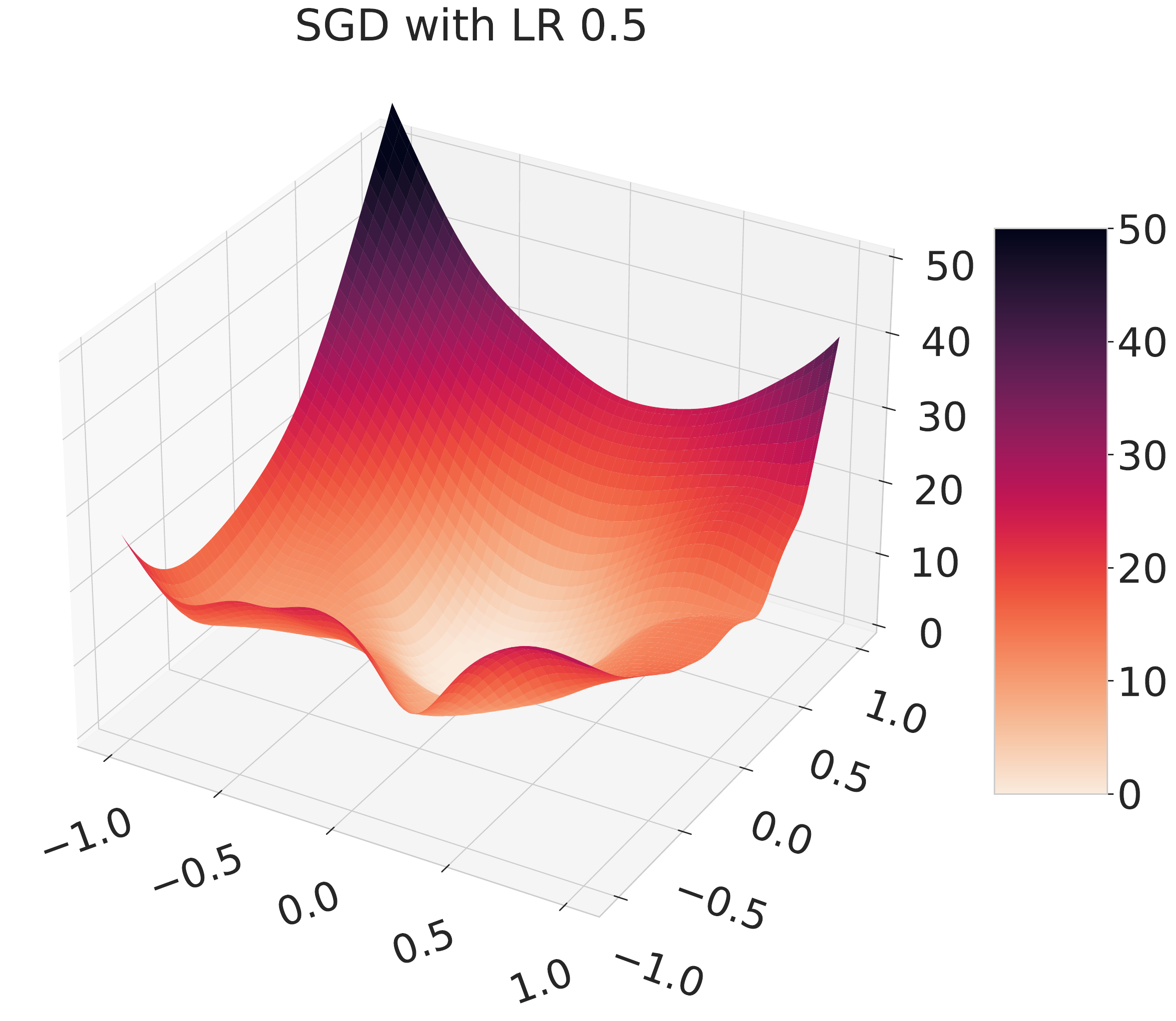}%
\label{fig:landscape_sgd_0_5_64}
}
\end{subfloat}
\begin{subfloat}[DAL]{
\includegraphics[width=0.4\columnwidth]{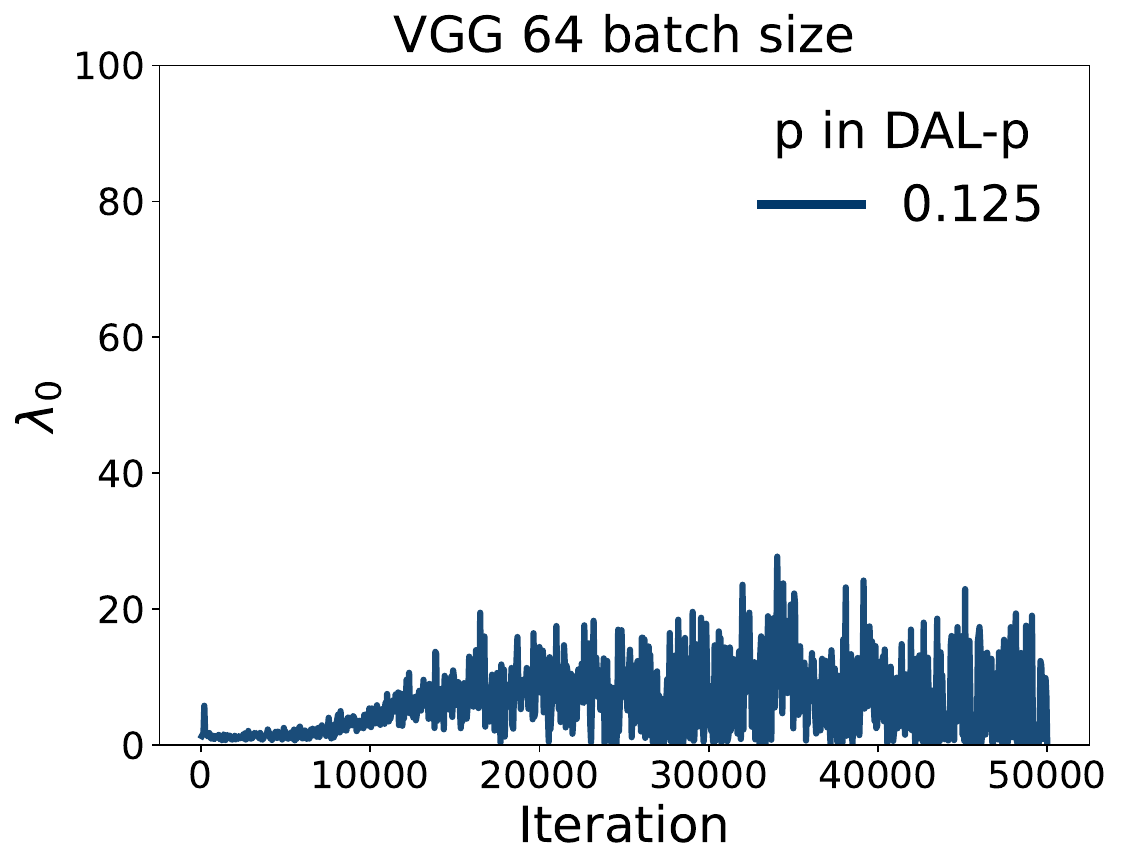}
\label{fig:landscape_dal_125bs_64_lambda_0}
}
\end{subfloat}\hspace{3em}
\begin{subfloat}[SGD]{
\includegraphics[width=0.4\columnwidth]{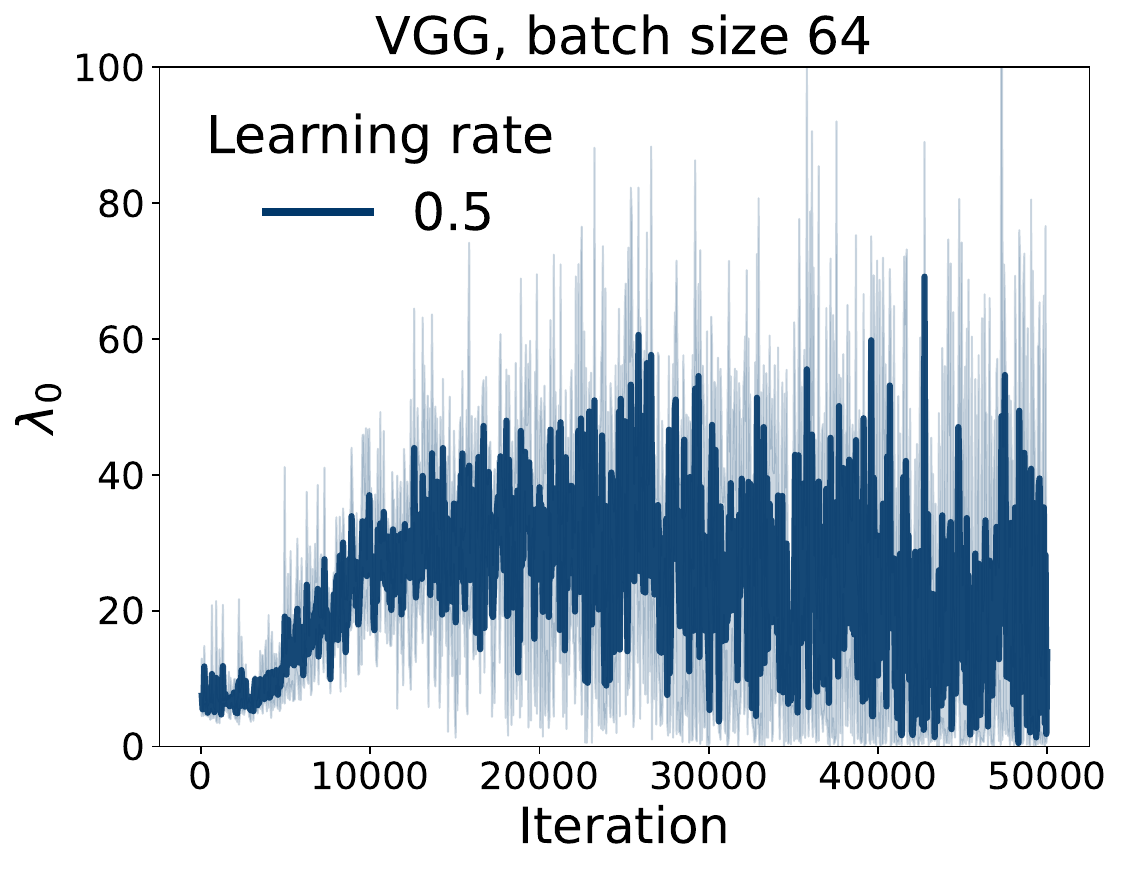}
\label{fig:landscape_sgd_0_5_64_lambda_0}
}
\end{subfloat}
\caption[DAL and gradient descent learned landscapes; batch size 64 on CIFAR-10.]{\textbf{CIFAR-10, batch size 64.} The 2D projection of the DAL-$p$ \subref{fig:landscape_dal_125bs_64} and SGD \subref{fig:landscape_sgd_0_5_64} learned landscapes on CIFAR-10 using a VGG model. The visualisation is made using the method of~\citet{li2018visualizing}. \textit{Though both models achieve a 86\% accuracy on the test dataset, DAL-$p$ has learned a flatter model.} Bottom row: We also show the trajectory of $\lambda_0$ for both models, which shows that throughout the learning trajectory, DAL \subref{fig:landscape_dal_125bs_64_lambda_0} travels through flatter landscapes compared to SGD \subref{fig:landscape_sgd_0_5_64_lambda_0}, as measured by $\lambda_0$.}
\label{fig:DAL_p_64_landscape}
\end{figure}
To further investigate the connection between drift and test set performance, we perform a set of sweeps over the power $p$ and show results in Figure~\ref{fig:power_sweeps}. These results show that the higher the drift (the smaller $p$), the more generalisation; additional results across batch sizes showing the same trend are shown in Figure~\ref{fig:power_sweeps_batch_sizes_vgg} in the Appendix. We also show in Figure~\ref{fig:stability_lambda_performance} the correlation between mean per-iteration drift and test accuracy both for learning rate and DAL-$p$ sweeps. The results consistently show that the higher the mean per-iteration drift, the higher the test accuracy. We also show that the mean per-iteration drift has a connection to the largest eigenvalue $\lambda_0$: the higher the drift, the smaller $\lambda_0$. These results add further evidence to the idea that discretisation drift is beneficial for generalisation performance in deep learning. We also notice that DAL-$p$ with smaller values of $p$ leads to a small $\lambda_0$ compared to vanilla gradient descent even when large learning rates are used for the latter; this could explain its generalisation capabilities as lower sharpness has been connected to generalisation in previous works~\citep{keskar2016large,jastrzkebski2018relation,foret2020sharpness}. \rebuttalrone{To consolidate these results, we use the method of~\citet{li2018visualizing} to visualise the loss landscape learned by DAL-$p$ compared to that learned using gradient descent in Figure~\ref{fig:DAL_p_64_landscape}, and observe that \textit{even when reporting similar accuracies, DAL-$p$ converges to a flatter landscape} and traverses a flatter area of space throughout training. These results are consistent across small and large batch sizes; we show additional results in Figures \ref{fig:DAL_p_full_batch_landscape} and \ref{fig:DAL_p_imagenet_landscape} in the Appendix.}

\rebuttalrthree{Inspired by understanding when the PF is close to the NGF, in this section we investigated the total discretisation drift of gradient descent. This led us to DAL-$p$, a method to automatically set the learning rate based on approximation to the per-iteration drift of gradient descent; we have seen that DAL produces stable training and further connected discretisation drift, generalisation, and flat landscapes as measured by leading Hessian eigenvalues.}

\section{Future work}
\label{sec:future_work}

\begin{figure}[tb]
\begin{subfloat}[Training loss.]{
 \includegraphics[width=0.49\columnwidth]{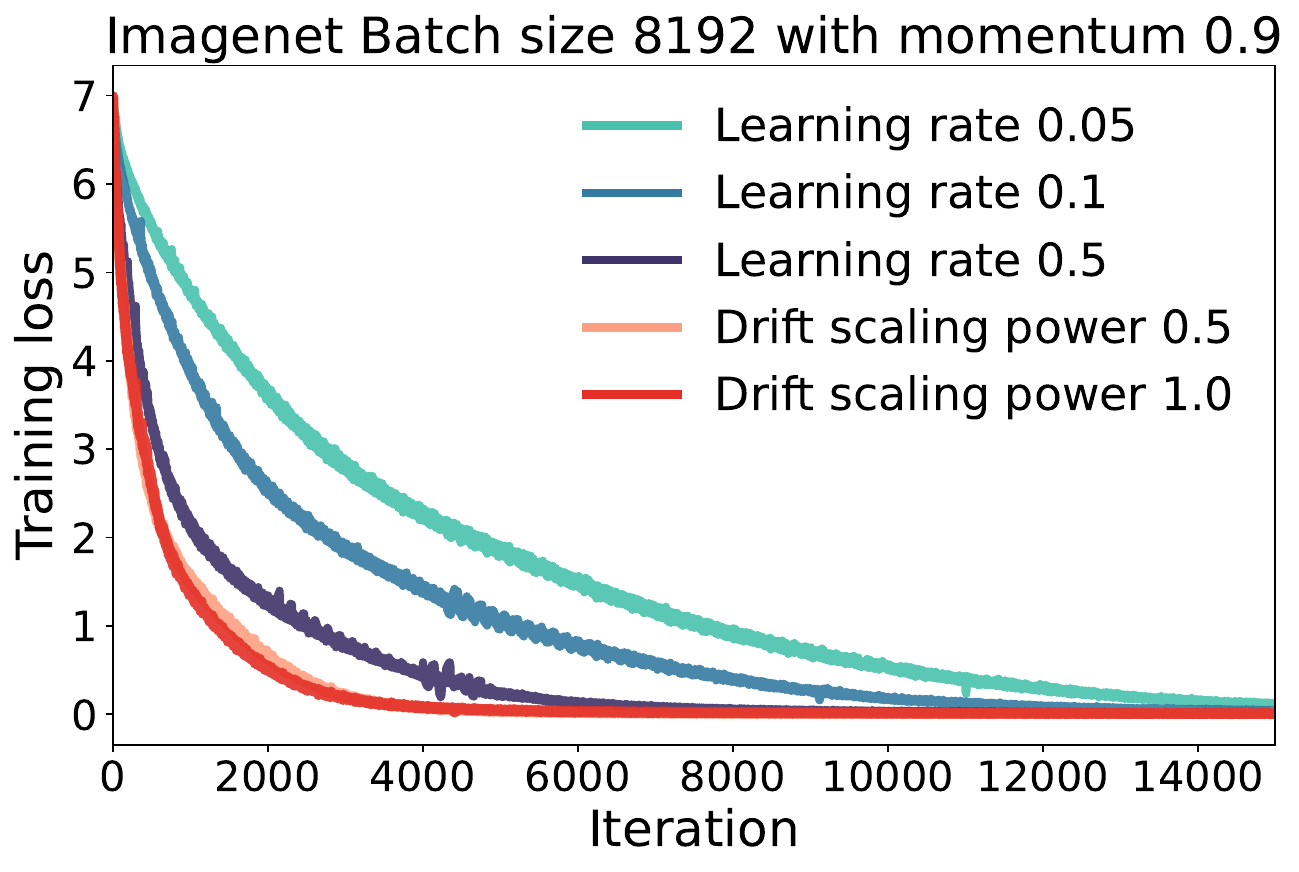}%
}\end{subfloat}%
\begin{subfloat}[Test accuracy.]{
  \includegraphics[width=0.51\columnwidth]{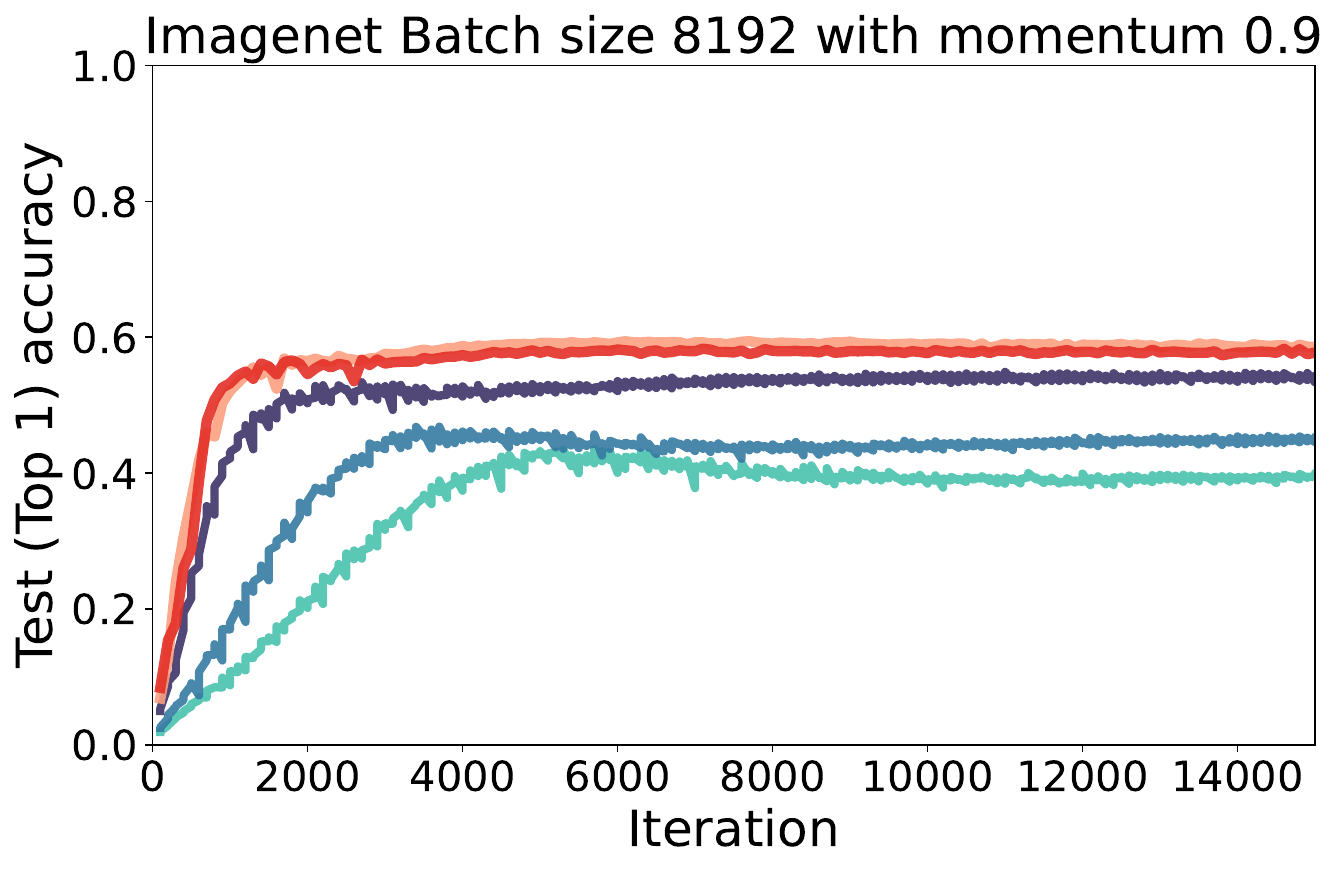}%
}\end{subfloat}%
\caption[DAL-$p$ and momentum Resnet-50; model trained on Imagenet.]{DAL with momentum: integrating drift information results in faster and more stable training compared to a fixed learning rate sweep. Compared to vanilla gradient descent there is also a significant performance and convergence speed boost.}
\label{fig:momentum_imagenet}
\end{figure}

\textbf{Beyond gradient descent}.
In this chapter, we focused on understanding vanilla gradient descent. 
Understanding discretisation drift via the PF can be beneficial for improving other gradient based optimisation algorithms as well,
as we briefly illustrate for momentum updates with decay $\beta$ and learning rate $h$ (for a discussion on momentum, see Section~\ref{sec:opt_algo}):
\begin{align}
\vv_t &= \beta \vv_{t-1} - h \nabla_{\vtheta} E(\vtheta_{t-1}); \hspace{10em}
\vtheta_t = \vtheta_{t-1} + \vv_t .
\end{align}
We can scale $\nabla_{\vtheta} E(\vtheta_{t-1})$ in the above not by a fixed learning rate $h$, but by adjusting the learning rate according to the approximation to the drift. This has two advantages: it removes the need for a learning rate sweep and it uses local landscape information in adapting the moving average, such that in areas of large drift the contribution is decreased, while it is increased in areas where the drift is small (a more formal justification is provided in Section~\ref{sec:proofs_total_per_iteration_drift}). This leads to the following updates:
\begin{align}
\vv_t &= \beta \vv_{t-1} - \frac{1}{2 ||{\nabla_{\vtheta}^2 E(\vtheta_{t-1})} \hat{\vg}(\vtheta)(\vtheta_{t-1})||}\nabla_{\vtheta} E(\vtheta_{t-1}); \hspace{5em}
\vtheta_t = \vtheta_{t-1} + \vv_t .
\end{align}
As with DAL-$p$, we can use powers to control the stability performance trade-off: the lower $p$, the more the current update contribution is reduced in high drift (instability) areas. We tested this approach on Imagenet and show results in Figure \ref{fig:momentum_imagenet}. The results show that integrating drift information improves the speed of convergence compared to standard gradient descent (Figure~\ref{fig:power_sweeps}), and leads to more stable training compared to using a fixed learning rate. We present additional experimental results in the Appendix.

\begin{figure}[th!]
\begin{subfloat}[Training loss.]{
  \includegraphics[width=0.495\columnwidth]{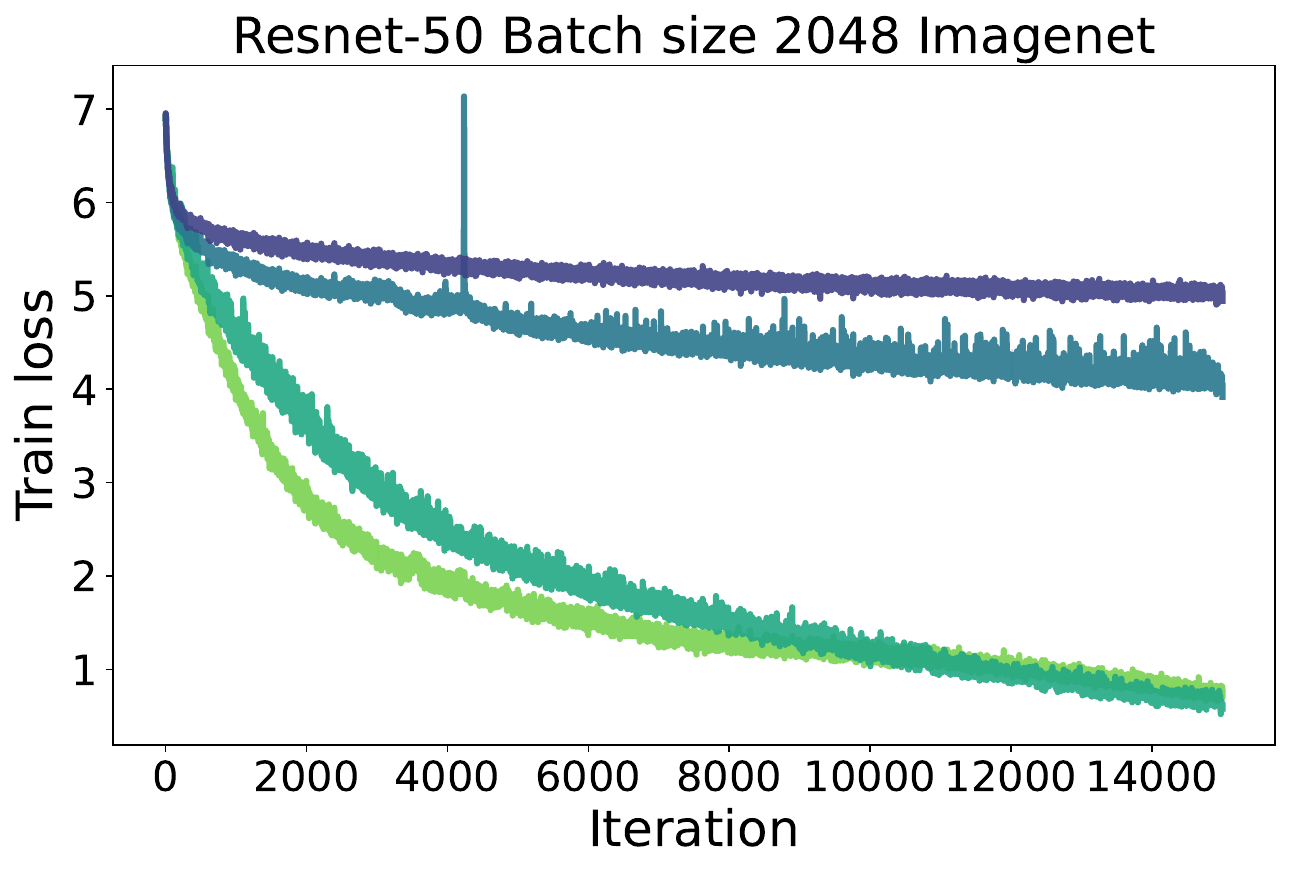}%
}
\end{subfloat}
\begin{subfloat}[Test accuracy.]{
  \includegraphics[width=0.505\columnwidth]{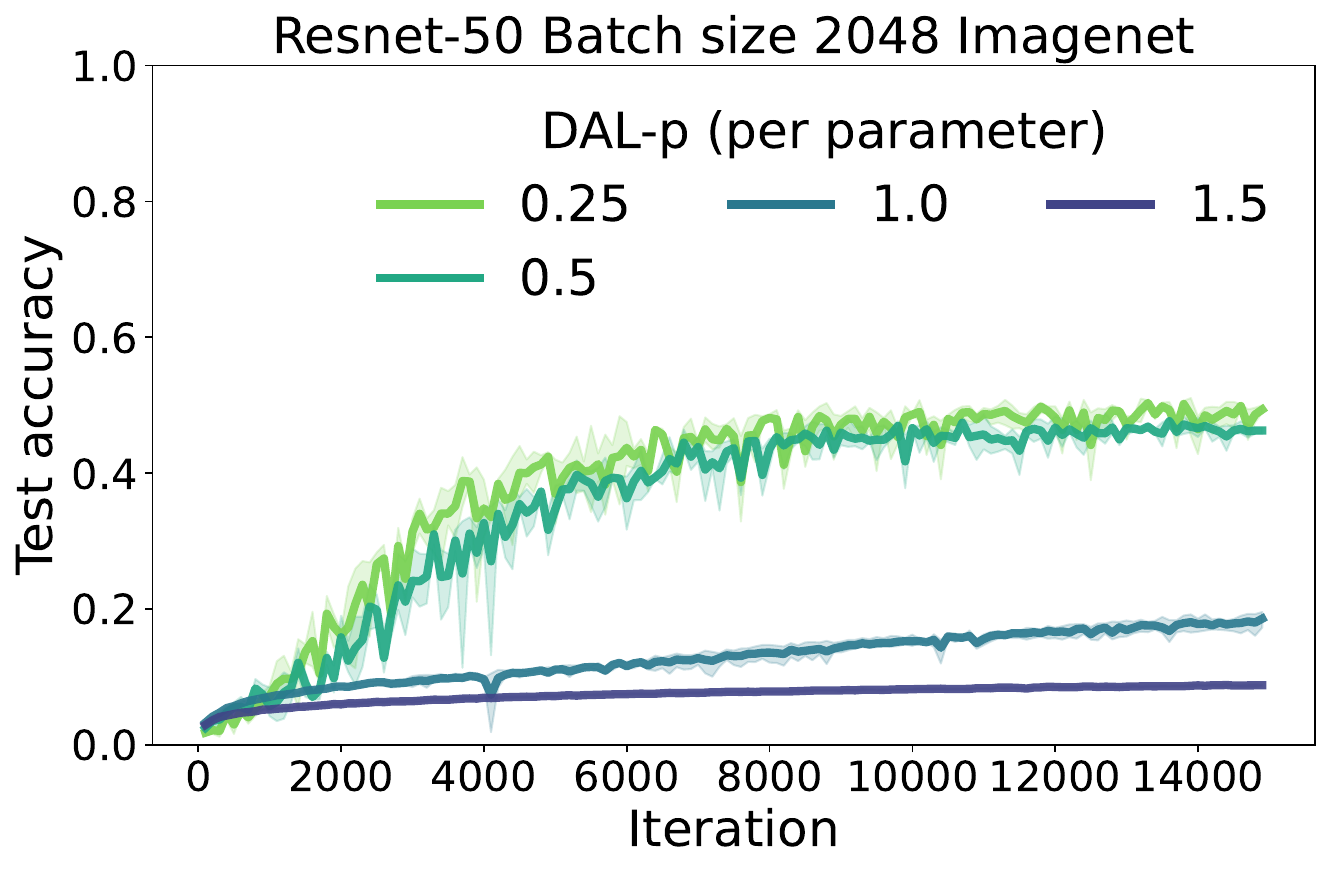}
}\end{subfloat}
\caption[Per-parameter DAL-$p$; batch size 8192 on Imagenet.]{Imagenet results when using a DAL like method of scaling per-parameter. Instead of using the global learning rate $2/||\nabla_{\vtheta} E^2 \widehat{g}(\vtheta)||$, using the per-parameter $\vtheta_i$ learning rate $h(\vtheta_i) = \frac{2}{\left(\nabla_{\vtheta} E^2 \widehat{g}(\vtheta)/\sqrt{D}\right)_i^p}$. These preliminary results show that using the per parameter information from discretisation drift can be used as an information to guide training.}
\label{fig:imagenet_lr_scaling_per_parameter}
\end{figure}

Just as momentum is a common staple of optimisation algorithms, so are adaptive schemes such as Adam~\citep{kingma2014adam} and Adagrad~\citep{duchi2011adaptive}, which adjust the step taken for each parameter independently. We can also use the knowledge from the PF to set a per-parameter learning rate: instead of using $\norm{{\nabla_{\vtheta}^2 E(\vtheta_{t-1})} \nabla_{\vtheta} E(\vtheta_{t-1})}$ to set a global learning rate, we can use the per-parameter information provided by $\nabla_{\vtheta}^2 E(\vtheta_{t-1}) \nabla_{\vtheta} E(\vtheta_{t-1})$ to adapt the learning rate of each parameter. 
We present preliminary results in the Figure~\ref{fig:imagenet_lr_scaling_per_parameter}. The above two approaches (momentum and per-parameter learning rate adaptation) can be combined, bringing us closer to the most commonly used deep learning optimisation algorithms, such as Adam (see Section~\ref{sec:opt_algo}).
While we do not explore this avenue here, we are hopeful that this understanding of discretisation drift can be leveraged further to stabilise and improve deep learning 
optimisation.

\textbf{Non-principal terms}.
This work focuses on understanding the effects of the PF on the behaviour of gradient descent.  Beyond the PF, we found one non-principal term (Eq~\eqref{eq:third_order_modified_vector_field}) and have seen that it can have a stabilising effect (Figure~\ref{fig:intuition_banana}).
We provided a preliminary explanation for the stabilising effect of this non-principal term in Section~\ref{sec:non_principal}. \rebuttalrone{One  promising avenue of non-principal terms is theoretically modelling the change of the eigenvalues $\lambda_i$ in time;} \rebuttalrthree{another promising direction is that of implicit regularisation: while existing work that uses BEA in deep learning has found important implicit regularisation effects~\citep{igr,igr_sgd,rosca2021discretisation}, we have shown here that effects of $\mathcal{O}(h^3)$ are not sufficient to capture the intricacies of gradient descent, suggesting that other implicit regularisation effects could be uncovered using the non-principal terms.}

\rebuttalrthree{\textbf{Neural network theory}}.
\rebuttalrthree{Many theoretical works studying at gradient descent in the neural network context use the NGF~\citep{du2018algorithmic,elkabetz2021continuous,symmetry,jacot2018neural}. We posit that replacing NGF in these theoretical contexts with PF may yield interesting results. In contrast to the NGF, the PF allows for the incorporation of the learning rate into the analysis, and unlike existing continuous-time models of gradient descent, it can model unstable behaviours observed in the discrete case.}
\rebuttalrthree{An example can be seen using the Neural Tangent Kernel: \citet{jacot2018neural} model gradient descent using the NGF to show that in the infinite-width limit gradient descent for neural networks follows kernel gradient descent. The PF can be incorporated in this analysis either by replacing the NGF with the PF as a model of gradient descent or by studying the difference in the PF for infinitely wide and finite width networks, since discretisation drift could be responsible for the observed gap between finite and infinite networks in the large learning rate case \citep{lee2020finite}.}

\section{Related work}

\textbf{Modified flows for deep learning optimisation}.
~\citet{igr} found the first order correction modified flow for gradient descent using BEA and uncovered its regularisation effects; they were the first to show the power of BEA in the deep learning context.~\citet{igr_sgd} find the first order error correction term in expectation during one epoch of stochastic gradient descent. 
 Modified flows have also been used for other optimisers than vanilla gradient descent: \citet{franca2020,shi2021understanding} compare momentum and Nesterov accelerated momentum; \citet{symmetry} study the symmetries of deep neural networks and use modified flows to show commonly used discrete updates break conservation laws present when using the NGF (for gradient descent they use the IGR flow while for momentum and weight decay they introduce different flows); \citet{kovachki2021continuous} use modified flows to understand the behaviour of momentum by approximating Hamiltonian systems; \citet{dissipative} construct optimisers controlling their stability and convergence rates while \citet{stochastic_adaptive_sgd} construct optimisers with adaptive learning rates in the context of stochastic differential equations. 

In concurrent work~\citet{miyagawatoward} use BEA to find a modified flow coined `Equations of Motion' (EOM) to describe gradient descent and find higher order terms, including non-principal terms; their focus is, however, on EOM(1), which is the IGR flow, which they use to understand scale and translation invariant layers. Their approach does not expand to complex space and does not capture the instabilities studied here (see also the discussion on the difference between the full modified flow provided by BEA and the PF in Section~\ref{sec:principal_flow}).

\textbf{Edge of stability and the importance of the Hessian}.
There have been a number of empirical studies on the Hessian in gradient descent. \citet{cohen2021gradient} observed the edge of stability behaviour and performed an extensive study which led to many empirical observations used in this work.~\citet{jastrzkebski2018relation} performed a similar study in the context of stochastic gradient descent.~\citet{sagun2017empirical,ghorbani2019investigation,papyan2018full} approximate the entire spectrum of the Hessian, and show that there are only a few negative eigenvalues, plenty of eigenvalues centered around 0, and a few positive eigenvalues with large magnitude. Similarly,~\citet{gur2018gradient} discuss how gradient descent operates in a small subspace.~\citet{lewkowycz2020large} discuss the large learning rate catapult in deep learning when the largest eigenvalue exceeds $2/h$.~\citet{gilmer2021loss} assess the effects of the largest Hessian eigenvalue in a large number of empirical settings. 

There have been a series of concurrent works aimed at theoretically explaining the empirical results above. \citet{ahn2022understanding} connect the edge of stability behaviour with what they coin as the `relative progress ratio':
$\frac{E(\vtheta - h \nabla_{\vtheta} E) - E(\vtheta)}{h \norm{\nabla_{\vtheta} E}^2}$, which they empirically show is 0 in stable areas of training and 1 in the edge of stability areas. To see the connection between the relative progress ratio and the quantities discussed in this chapter, one can perform a Taylor expansion on ${\frac{E(\vtheta - h \nabla_{\vtheta} E) - E(\vtheta)}{h \norm{\nabla_{\vtheta} E}^2} \approx \frac{- h \nabla_{\vtheta} E^T \nabla_{\vtheta} E + h^2/2 \nabla_{\vtheta} E^T \nabla_{\vtheta}^2 E \nabla_{\vtheta} E}{h \norm{\nabla_{\vtheta} E}^2} = -1 + \frac h 2 \frac{\nabla_{\vtheta} E^T \nabla_{\vtheta}^2 E \nabla_{\vtheta} E}{\norm{\nabla_{\vtheta} E}^2}}$. 
While this ratio is related to the quantities we discuss, we also note significant differences: it is a scalar, and not a parameter-length vector and thus does not capture per eigendirection behaviour as we see with the stability coefficients (Section~\ref{sec:instability_deep_learning}). \citet{arora2022understanding} prove the edge of stability result occurs under certain conditions either on the learning rate or on the loss function. \citet{ma2022multiscale} empirically observe the multi-scale structure of the loss landscape in neural networks and use it to theoretically explain the edge of stability behaviour of gradient descent. \citet{chen2022gradient} use low dimensional theoretical insights around a local minima to understand the edge of stability behaviour. \citet{damian2022self} use a cubic Taylor expansion to show that gradient descent follows the trajectory of a projected method which ensures that $\lambda_0 < 2/h$ and $\nabla_{\vtheta} E^T \vu = 0$; their work is what inspired us to write the third-order non-principal term in the form of Eq~\eqref{eq:non_principal_minimisation}, after we had previously noted its stabilising properties.
These important works are complementary to our own work; they do not use continuous-time approaches and tackle primarily the edge of stability problem or its subcases, while we focus on understanding gradient descent and applying that understanding broadly, including but not limited to the edge of stability phenomenon.

\rebuttalrthree{\textbf{Discrete models of gradient descent}. The desire to understand learning rate specific behaviour in gradient descent has been a motivation in the construction of discrete-time analyses. These analyses have provided great insights, from studying noise in the stochastic gradient descent setting~\citep{liu2021noise,ziyin2021strength}, the study of over-parametrised neural models and their convergence~\citep{gunasekar2018implicit,du2019gradient,allen2019convergence}, providing examples of when gradient descent can converge to local maxima~\citep{ziyin2021sgd}, the importance of width for proving convergence in deep linear networks~\citep{du2019width}. We differ from these studies both in motivation and execution: we are looking for a continuous-time flow that will increase the applicability of continuous-time analysis of gradient descent. We do so by incorporating discretisation drift using BEA and showing that the resulting flow is a useful model of gradient descent, which captures instabilities and escape of local minima and saddle points.}

\textbf{Understanding the difference between the negative gradient flow and gradient descent}. \citet{elkabetz2021continuous} recently examined the differences between gradient descent and the NGF in the deep learning context; their work examines the importance of the Hessian in determining when gradient descent follows the NGF. Their theoretical results show that neural networks are roughly convex and thus for reasonably sized learning rates one can expect that gradient descent follows the NGF flow closely. Their results complement ours and their approach might be extended to help us understand why the PF is sufficient to shed light on many instability observations in the neural network training.

\textbf{Second-order optimisation.} By using second order information (or approximations thereof) to set the learning rate, DAL is related to second-order approaches used in deep learning. 
Many second-order methods can be seen as approximates of Newton's method $\vtheta_t = \vtheta_{t-1} - \nabla^2_{\vtheta} E ^{-1}(\vtheta_{t-1}) \nabla_{\vtheta} E(\vtheta_{t-1})$.
Since computing the Hessian inverse, or storing a matrix of the size of the Hessian as required by Quasi-Netwon methods~\citep{fletcher2013practical}, can be prohibitively expensive for large models, many practical methods approximate it with tractable alternatives~\citep{martens2015optimizing}. 
\citet{foret2020sharpness} propose an optimisation scheme directly aimed at minimising sharpness and show this can improve generalisation.

\textbf{Connection between drift and generalisation}.
We have made the connection between discretisation drift and generalisation. This connection was first made by~\citet{igr} through the IGR flow. Generalisation has also been connected to the largest eigenvalue $\lambda_0$\citep{hochreiter1997flat,keskar2016large,jastrzkebski2018relation,lewkowycz2020large}; recently \citet{kaur2022maximum} however showed a more complex picture, primarily in the context of stochastic gradient descent. The largest eigenvalue could be a confounder to the drift as we have observed in Section~\ref{sec:trade_off}; we hope that future work can deepen these connections.

\section{Conclusion}

We have expanded on previous works that used backward error analysis in deep learning to find a new continuous-time flow, called the \textbf{Principal Flow} (PF), to analyse the behaviour of gradient descent. \rebuttalrthree{Unlike existing flows, the PF operates in complex space, which enables it to better capture the behaviour of gradient descent compared to existing flows, including but not limited to instability and oscillatory behaviour}. 
\rebuttalrthree{We use the form of the PF to find new quantities relevant to the stability of gradient descent, and shed light on newly observed empirical phenomena, such as the edge of stability results.}
After understanding the core quantities connected to instabilities in deep learning we  devised an automatic learning rate schedule, DAL, which exhibits stable training. We concluded by cementing the connection between large discretisation drift and increased generalisation performance.
\rebuttalrthree{We ended by highlighting future work avenues including incorporating the PF in existing theoretical analyses of gradient descent which use the negative gradient flow, incorporating our understanding of the drift of gradient descent in other optimisation approaches and specialising the PF for neural network function approximators.}

\chapter{Continuous time models of optimisation in two-player games}
\label{ch:dd_gans}

The fusion of deep learning with two-player games has produced a wealth of breakthroughs in recent years, from Generative Adversarial Networks (GANs)~\cite{goodfellow2014generative} through to model-based reinforcement learning~\cite{sutton2018reinforcement,rajeswaran2020game}. Gradient descent methods are widely used across these settings, partly because these algorithms scale well to high-dimensional models and large datasets. However, much of the recent progress in our theoretical understanding of two-player differentiable games builds upon the analysis of continuous differential equations that model the dynamics of training~\cite{singh2000nash,heusel2017gans,nagarajan2017gradient}, leading to a gap between theory and practice.
Our aim is to take a step forward in our understanding of two player games by finding continuous systems which better match the gradient descent updates used in practice.

Our work builds upon~\citet{igr}, who use backward error analysis (BEA) to quantify the discretisation drift induced by using gradient descent in supervised learning. We extend their work and use BEA to understand the impact of discretisation in the training dynamics of two-player games.
More specifically, we quantify \textit{Discretisation Drift} (DD), the difference between the solutions of the original flows defining the game and the discrete steps of the numerical scheme used to approximate them. To do so, we construct modified continuous systems that closely follow the discrete updates.
While in supervised learning DD has a beneficial regularisation effect~\citep{igr} (see also Section~\ref{sec:bea}), we find that the interaction between players in DD can have a destabilising effect in adversarial games.

\textbf{Contributions}: Our primary contribution, Theorems \ref{thm:2_players_sim} and \ref{thm:alt}, provides the continuous modified systems that quantify the discretisation drift in simultaneous and alternating gradient descent for general two-player differentiable games. Both theorems are novel in their scope and generality, as well as their application toward understanding the effect of the discretisation drift in two-player games parametrised with neural networks. Theorems \ref{thm:2_players_sim} and \ref{thm:alt} allow us to use dynamical system analysis to describe GD without ignoring discretisation drift, which we then use to:
\begin{itemize}
\setlength\itemsep{0.5em}
\item Provide new stability analysis tools (Section \ref{section:stability}). %
\item Motivate explicit regularisation methods that drastically improve the performance of simultaneous gradient descent in GAN training (Section \ref{section:explicit_regularisation}).
\item Pinpoint optimal regularisation coefficients and shed new light on existing explicit regularisers (Table \ref{tab:methods_comp}).
\item Pinpoint the best performing learning rate ratios for alternating updates (Sections \ref{sec:common_payoff} and \ref{section:learning_rate_ratios}).
\item Explain previously observed but unexplained phenomena such as the difference in performance and stability between alternating and simultaneous updates in GAN training (Section~\ref{section:zero_sum}).
\end{itemize}

\section{Background}
\label{sec:dd_gan_background}

A well-developed strategy for understanding two-player games in gradient-based learning is to analyse the continuous dynamics of the game~\cite{singh2000nash,heusel2017gans}. Tools from dynamical systems theory have been used to explain convergence behaviours and to improve training using modifications of learning objectives or learning rules~\citep{nagarajan2017gradient,balduzzi2018mechanics,mazumdar2019finding}. While many insights from continuous analysis carry over to two-player games trained with discrete updates, discrepancies are often observed between the continuous analysis conclusions and discrete game training~\citep{mescheder2018training}. In this chapter, we extend BEA from the single-objective case to the more general two-player game setting, thereby bridging the gap between the continuous systems that are often analysed in theory and the discrete 
numerical methods used in practice.  For an overview of optimisation in two-player games, we refer the reader to Section~\ref{sec:intro_multi_objective_optimisation}. We provided an overview of BEA in Section~\ref{sec:bea}, and used it in the previous chapter in the single-objective case.

\subsection{Generative adversarial networks}
Generative adversarial networks(GANs)~\citep{goodfellow2014generative} learn an implicit generative model through a two-player game. An implicit generative model does not have an explicit likelihood associated with it, as it is often the case with other machine learning models. Instead, an implicit model can be thought of as a simulator: it provides a sampling path that can be used to generate samples from its distribution. In the case of GANs the sampling path is defined via a latent variable model:
\begin{align}
	G(\vz; \vtheta), \hspace{3em} \vz \sim p(\vz)
\end{align}
where $p(\vz)$ is a prior distribution and $G(\vz; \vtheta)$ is the model with parameters $\vtheta$, often a deep neural network. This sampling process induces the implicit distribution
\begin{align}
	p(\vx; \vtheta) = \int_{G(\vz;\vtheta)= \vx} p(\vx| \vz; \vtheta) p(\vz)  d \vz = \int \mathbb{I} (G(\vz; \vtheta), \vx) p(\vz)  d \vz,
\end{align}
where $\mathbb{I} (G(\vz; \vtheta), \vx)$ is 1 if its two arguments are equal and 0 otherwise.
The above integral is intractable to compute or expensive to approximate via approaches such as annealed importance sampling~\citep{neal2001annealed} for many cases of interest, such as when $G(\cdot, \vtheta)$ is given by a deep neural network. This makes standard methods, such as maximum likelihood, unavailable to learn implicit models. To learn the parameters $\vtheta$, GANs introduce another model, the discriminator $D(\cdot, \vphi)$, which aims to distinguish between real data---samples from the dataset---and generated data---samples from the model $G(\cdot, \vtheta)$. The generative model $G(\cdot, \vtheta)$, called \textit{the generator}, aims to generate samples that cannot be distinguished from real data according to the discriminator. 

In the original GAN paper~\citep{goodfellow2014generative} the discriminator $D(\cdot, \vphi)$ is a binary classifier trained using the binary cross entropy loss; the generator's aim is to ensure that the discriminator does not classify generated data as fake.
This can be formalised as the zero-sum game:
\begin{align}
\min_{\vtheta} \max_{\vphi} E &= \frac{1}{2}\mathbb{E}_{p^*(\vx)} \log D(\vx; \vphi) +  \frac{1}{2}\mathbb{E}_{p(\vx; \vtheta)} \log( 1 - D(\vx; \vphi)) \\
&= \frac{1}{2}\mathbb{E}_{p^*(\vx)} \log D(\vx; \vphi) +  \frac{1}{2}\mathbb{E}_{p(\vz)} \log( 1 - D(G(\vz; \vtheta); \vphi).
\label{eq:original_gan}
\end{align}
where Eq~\eqref{eq:original_gan} uses the change-of-variable in the pathwise estimator discussed in Section~\ref{sec:stochastic_opt}.
To connect this zero-sum game to  generative modelling objectives used to learn probability distributions,~\citet{goodfellow2014generative} set the functional derivative  in Eq~\eqref{eq:original_gan} to 0 to find the optimal discriminator $D^*$:
\begin{align}
 \frac{1}{2}p^*(\vx) \frac{1}{D^*(\vx)} -  \frac{1}{2} p(\vx; \vtheta) \frac{1}{1 - D^*(\vx)} = 0 \implies D^*(\vx) = \frac{2 p^*(\vx)}{p^*(\vx) + p(\vx; \vtheta)}.
\end{align}
Replacing the above in Eq~\eqref{eq:original_gan} we have:
\begin{align}
 E &= \frac{1}{2}\mathbb{E}_{p^*(\vx)} \log  \frac{2 p^*(\vx)}{ (p^*(\vx) + p(\vx; \vtheta))} +  \frac{1}{2}\mathbb{E}_{p(\vx; \vtheta)} \frac{2 p(\vx; \vtheta)}{(p^*(\vx) + p(\vx; \vtheta))} \\
   &= \mathbb{KL}(p^*||\frac{1}{\frac{p^* + p(\cdot;\vtheta) } {2}}) +  \mathbb{KL}(p(\cdot;\vtheta)||\frac{1}{\frac{p^* + p(\cdot;\vtheta) } {2}})\\
    &= \mathbb{JSD}(p^*||p(\cdot;\vtheta)),
\end{align}
where $\mathbb{KL}$ and $\mathbb{JSD}$ denote the KL and Jensen--Shannon divergences respectively.
This entails that if the discriminator is optimal, the GAN generator is minimising the Jensen--Shannon
divergence between the model and data distribution.

The connection between distributional divergences and two-player games has fuelled a line of research motivated by finding the best objective to train GANs. This lead to new GANs motivated by Integral Probability Metrics, such as the Wasserstein distance or the Maximum Mean Discrepancy~\citep{gulrajani2017improved,binkowski2018demystifying}; $f$-divergences~\citep{nowozin2016f}; and others~\citep{jolicoeur2018relativistic}. Practitioners have observed, however, that what is most relevant in increasing GAN performance often comes from changes architectures, optimisation, and regularisation~\citep{dcgan,biggan,karras2019style}, rather than underlying divergence used.
An explanation for this observation is that the GAN discriminator is not optimal, both because it is modelled using a parametric model and because the discriminator is not optimised to convergence for each generator update, as required to obtain the best estimate of the underlying divergence~\citep{fedus2017many}. Instead, as discussed in Section~\ref{sec:intro_multi_objective_optimisation}, due to computational considerations the nested optimisation problem in Eq~\eqref{eq:original_gan} gets translated into the alternating updates described in Algorithm~\ref{alg:alt_updates}. 

Since GANs are implicit generative models, evaluating them via likelihoods is computationally intractable; approaches to approximate the likelihood do exist, but they have strong limitations~\citep{grover2018flow} and are rarely used in practice. Instead, the main avenue for GAN evaluation is to measure aspects of sample quality and diversity via a pretrained classifier, as done by the Inception Score~\citep{improved_techniques_gans} and Fréchet Inception Distance~\citep{heusel2017gans}. The Inception Score uses label information from the dataset, together with a pre-trained classifier, to assess a generative model. 
It does so by using the pre-trained classifier to obtain class posterior probabilities from samples $p(y|\vx)$, and from there the marginal distribution induced over class labels by the model distribution $p(y; \vtheta) = \int p(y|\vx) p(\vx;\vtheta) d \vx$, which is estimated using Monte-Carlo estimation. The Inception Score then measures average KL divergence between $p(y; \vtheta)$ and the distribution $p(y|\vx)$ obtained from model samples: $e^{{\mathbf{E}_{p(\vx; \vtheta)} \mathbb{KL}(p(y|\vx) || p_{\vtheta}(y))}}$. When a model produces high quality samples from all classes in the dataset $p(y; \vtheta)$ will be a broad distribution capturing the marginal distribution over classes in the dataset, while $p(y|\vx)$ will be a sharp distribution pre-training to the class associated with $\vx$; this leads to a high $\mathbb{KL}(p(y|\vx) || p(y; \vtheta)$ and thus a high Inception Score score. 

\section{Discretisation drift in two-player games}
\label{sec:general_theorems}

We denote by $\vphi \in \mathbb{R}^m$ and $\vtheta \in \mathbb{R}^n$ the parameters of the first and second player, respectively.
The players update functions will be denoted correspondingly by $f(\vphi, \vtheta): \mathbb{R}^m \times \mathbb{R}^n \rightarrow \mathbb{R}^m$ and by
$g(\vphi, \vtheta): \mathbb{R}^m \times \mathbb{R}^n \rightarrow \mathbb{R}^n$.
The Jacobian $\jacthetaf(\vphi, \vtheta)$ is the $m\times n$ matrix
$\jacthetaf(\vphi, \vtheta)_{i,j} = \left(\partial_{\theta_j}f_i\right)$ with $i = 1, \dots, m$ and  $j = 1, \dots, n$.

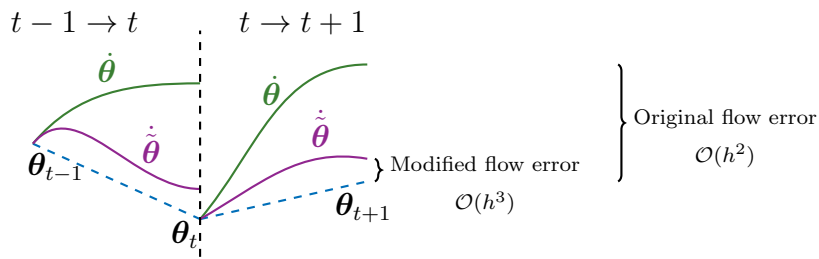
\begin{figure}[t]
\centering
\begin{tikzpicture}[every text node part/.style={align=center,inner sep=0,outer sep=0}][overlay]
\coordinate (theta_t_minus_1) at (0,0);
\coordinate (theta_t) at (2.2,-1);
\coordinate (theta_t_plus_1) at (4.4, -0.5);

\node(draw) at ($(theta_t_minus_1) + (+0.3,-0.32)$) {$\vtheta_{t-1}$};
\node(draw) at ($(theta_t) + (-0.2,-0.2)$) {$\vtheta_{t}$};
\node(draw) at ($(theta_t_plus_1) + (-0.05,-0.35)$) {$\vtheta_{t+1}$};

\coordinate (first_time_transition) at ($(theta_t_minus_1) + (+0.55,1.6)$);
\coordinate (second_time_transition) at ($(first_time_transition) + (3,0)$);

\node(draw) at (first_time_transition) {$t -1 \rightarrow t $};
\node(draw) at (second_time_transition) {$t \rightarrow t + 1$};

\coordinate (cont_theta_t) at (2.2,0.8);
\coordinate (cont_theta_t_plus_1) at (4.4, 1.05);

\coordinate (mod_cont_theta_t) at (2.2, -0.6);
\coordinate (mod_cont_theta_t_plus_1) at (4.4, -0.2);

\draw [NavyBlue,thick,dashed] (theta_t_minus_1) -- (theta_t);
\draw [NavyBlue,thick,dashed] (theta_t) -- (theta_t_plus_1);

\draw [OliveGreen,thick]  (theta_t_minus_1) to[out=50,in=180] node[midway,above] {$\dot{\vtheta}$} (cont_theta_t);
\draw [OliveGreen,thick]  (theta_t) to[out=50,in=180] node[midway,above]{$\dot{\vtheta}$} (cont_theta_t_plus_1);

\draw [Plum,thick]  (theta_t_minus_1) to[out=50,in=180]  node[near end,above] {$\dot{\tilde{\vtheta}}$} (mod_cont_theta_t);
\draw [Plum,thick]  (theta_t) to[out=30,in=170] node[near end,above] {$\dot{\tilde{\vtheta}}$} (mod_cont_theta_t_plus_1);

\draw [black,thick,dashed] ($(theta_t) + (0,-0.5)$) -- ($(cont_theta_t) + (0,0.7)$);

\draw [
    thick,
    decoration={
        brace,
        mirror,
        raise=0.1cm
    },
    decorate
] (theta_t_plus_1) -- (mod_cont_theta_t_plus_1)
node [pos=0.5,anchor=west,xshift=0.15cm,yshift=-0.2cm] {\scriptsize Modified flow error \\ \scriptsize $\mathcal{O}(h^3)$};

\draw [
    thick,
    decoration={
        brace,
        mirror,
        raise=3.3cm
    },
    decorate
] (theta_t_plus_1) -- (cont_theta_t_plus_1)
node [pos=0.5,anchor=west,xshift=3.35cm,yshift=-0.2cm] {\scriptsize Original flow error \\ \scriptsize $\mathcal{O}(h^2)$};
\vspace{-2em}
\end{tikzpicture}
 \caption[Visualisation of our approach using backward error analysis (BEA) to find new continuous-time flows describing the behaviour of gradient descent in two-player games.]{We use BEA to find the modified flow $\hatdotvtheta$ that captures the change in parameters introduced by the discrete updates with an error of $\mathcal{O}(h^3)$. The modified flow follows the discrete update more closely than the flow $\dot{\vtheta}$, which has an error of $\mathcal{O}(h^2)$. Here, we show only $\vtheta$ for simplicity.}
  \label{fig:idd_graphic}
\end{figure}

We aim to understand the impact of discretising the flows
\begin{align}
 \dot{\vphi} &=  f( \vphi, \vtheta)   \label{eq:ode1} \\
 \dot{\vtheta}  &= g( \vphi, \vtheta) \label{eq:ode2},
\end{align}
using either simultaneous  or alternating Euler updates. 
To do so, we derive a modified continuous system of the form
\begin{align}
 \dot{\vphi} &=  f( \vphi, \vtheta)  + h f_1( \vphi, \vtheta) \\
 \dot{\vtheta}  &= g( \vphi, \vtheta) + h g_1( \vphi, \vtheta),
\end{align}
which closely follows the discrete Euler update steps; the local error between a discrete update and the modified system is of order $\mathcal O(h^3)$ instead of $\mathcal O(h^2)$ as is the case for the continuous system given by Eqs \eqref{eq:ode1} and \eqref{eq:ode2}. We visualise our approach in Figure~\ref{fig:idd_graphic}.

We can specialise these modified equations using $f = -\nabla_{\vphi} E_{\vphi}$ and $g = -\nabla_{\vtheta} E_{\vtheta}$, where $E_{\vphi}$ and $E_{\vtheta}$ are the loss functions for the two players. We will use this setting later to investigate the modified dynamics of \textit{simultaneous} or \textit{alternating gradient descent}. We can further specialize the form of these updates for common-payoff games ($E_{\vphi} = E_{\vtheta}  = E$) and zero-sum games ($E_{\vphi} = -E$, $E_{\vtheta} = E$).

\subsection{DD for simultaneous Euler updates}

{The \textit{simultaneous Euler updates} with learning rates $\lrp h$ and $\lrt h$ respectively are given by
\begin{align}
\vphi_t &= \vphi_{t-1} + \lrp h f(\vtheta_{t-1}, \vphi_{t-1} ) \label{eq:simup1}\\
\vtheta_{t} &= \vtheta_{t-1} + \lrt h g(\vtheta_{t-1}, \vphi_{t-1}) \label{eq:simup2}.
\end{align}
By applying BEA to these discrete updates, we obtain:
\begin{theo} \label{thm:2_players_sim} The discrete \emph{simultaneous} Euler updates
in \eqref{eq:simup1} and \eqref{eq:simup2} follow the continuous system
\label{thm:sim}
\begin{align}
\dot{\vphi} &= f - \frac{\lrp h}{2} \left(\jacphif f + \jacthetaf g \right) \\
\dot{\vtheta} &= g - \frac{\lrt h}{2} \left(\jacphig f + \jacthetag g \right)
\end{align}
with an error of size $\mathcal O(h^3)$  after one update step.
\end{theo}

\begin{remark}
A special case of Theorem~\ref{thm:2_players_sim} for zero-sum games with equal learning rates can be found in~\citet{lu2021resolution}.
\end{remark}

\subsection{DD for alternating Euler updates}

For \textit{alternating Euler updates}, the players take turns to update their parameters, and can perform multiple updates each. We denote the number of alternating updates by $m$ and $k$ for the first  player and second player, respectively.
We scale the learning rates by the number of updates, leading to the following updates $ \vphi_{t} := \vphi_{m,t} $ and $\vtheta_{t} := \vtheta_{k, t}$ where
\begin{align}
 \vphi_{i, t} &=  \vphi_{i-1, t} + \frac{\lrp h}{m}   f(\vphi_{i-1,t}, \vtheta_{t-1}) , \hspace{1em} i = 1, \dots ,m, \label{eq:altup1} \\
 \vtheta_{j, t} &=  \vtheta_{j-1, t} + \frac{\lrt h}{k} g(\vphi_{m, t}, \vtheta_{j-1, t}), \hspace{1em} j = 1, \dots ,k. \label{eq:altup2}
\end{align}
By applying BEA to these discrete updates, we obtain:
\begin{theo} The discrete \emph{alternating} Euler updates in \eqref{eq:altup1} and \eqref{eq:altup2} follow the continuous system
\label{thm:alt}
\begin{align}
\dot{\vphi} &= f - \frac{\lrp h}{2} \left(\frac{1}{m}\jacphif f + \jacthetaf g \right) \\
\dot{\vtheta} &= g - \frac{\lrt h}{2} \left((1- \frac {2\lrp} {\lrt})\jacphig f + \frac{1}{k} \jacthetag g \right)
\end{align}
with an error of size $\mathcal O(h^3)$ after one update step.
\end{theo}
\begin{remark}
Equilibria of the original continuous systems (i.e., points where $f = \myvec{0}$ and $g=\myvec{0}$) remain equilibria of the modified continuous systems.
\end{remark}

\begin{definition}
\normalfont{The discretisation drift components we identify for each player have two terms: one term containing a player's own update function only---terms we will call \textit{self terms}---and a term that also contains the other player's update function---which we will call \textit{interaction terms}.}
\label{def:self_interaction_terms}
\end{definition}

\subsection{Sketch of the proofs}

Following BEA, the idea is to modify the original continuous system by adding corrections in powers of the learning rate: ${\tilde f = f + h f_1 + h^2 f_2 + \cdots}$ and ${\tilde g = g + h g_1 + h^2 g_2 + \cdots}$, where for simplicity in this proof sketch, we use the same learning rate for both players (detailed derivations can be found in the Appendix Section~\ref{sec:dd_gan_proofs}; the general structure of BEA proofs can be found in Section~\ref{sec:bea_proof_structure}). We want to find corrections $f_i$, $g_i$ such that the modified system $\dot \vphi = \tilde f$ and $\dot \vtheta = \tilde g$ follows the discrete update steps exactly. To do that we proceed in three steps:

{\bf Step 1:} We expand the numerical scheme to find a relationship between $\vphi_t$ and $\vphi_{t-1}$ and $\vtheta_t$ and $\vtheta_{t-1}$ up to order $\mathcal{O}(h^2)$.
In the case of the simultaneous Euler updates this does not require any change to Eqs~\eqref{eq:simup1} and~\eqref{eq:simup2}, while for alternating updates we have to expand the intermediate steps of the integrator using Taylor series.

{\bf Step 2:} We compute the Taylor series of the modified equations solution:
\begin{align}
\vphi(h)   &=  \vphi_{t-1}   + h f + h^2 \left(f_1 + \frac 12 (\jacphif f + \jacthetaf g)\right) + \mathcal{O}(h^3)\\
\vtheta(h) &=  \vtheta_{t-1} +  h g +  h^2 \left(g_1 + \frac 12 (\jacphig f + \jacthetag g)\right) + \mathcal{O}(h^3),
\end{align}
where all the $f$s, $g$s, and their derivatives are evaluated at ($\vphi_{t-1}$, $\vtheta_{t-1}$).

{\bf Step 3:} We match the terms of equal power in $h$ so that the solution of modified equations coincides with the discrete update after one step.
 This amounts to finding the corrections, $f_i$s and $g_i$s, so that all the terms of order higher than $\mathcal{O}(h)$ in the modified equation solution above will vanish; this yields the first order drift terms $f_1$ and $g_1$ in terms of $f$, $g$, and their derivatives. For simultaneous updates we obtain $f_1 =  - \tfrac{1}{2} (\jacphif f + \jacthetaf g)$ and $g_1 = - \tfrac{1}{2}(\jacphig f + \jacthetag g)$, and by construction, we have obtained the modified truncated system which follows the discrete updates exactly up to order $\mathcal O(h^3)$, leading to Theorem~\ref{thm:sim}.

\begin{remark} The modified equations in Theorems \ref{thm:2_players_sim} and \ref{thm:alt} closely follow the discrete updates only for learning rates where errors of size $\mathcal O(h^3)$ can be neglected. Beyond this, higher order corrections are likely to contribute to the DD.
\end{remark}

\subsection{Visualising trajectories}
To illustrate the effect of DD in two-player games, we use a simple example adapted from~\citet{balduzzi2018mechanics}:
\begin{align}
\dot{\phi} = f(\phi, \theta) =  - \epsilon_1 \phi  + \theta; \hspace{1em} \dot{\theta} = g(\phi, \theta) = \epsilon_2 \theta  - \phi.
\label{eq:simple_odes}
\end{align}
In Figure~\ref{fig:igr_dynamics_toy} we validate our theory empirically by visualising the trajectories of the discrete
Euler steps for simultaneous and alternating updates, and show that they closely match the trajectories of the
corresponding modified continuous systems that we have derived. To visualise the trajectories of the original unmodified continuous system, we use Runge--Kutta 4 (a fourth-order numerical integrator that has no DD up to $\mathcal{O}(h^5)$ in the case where the same learning rates are used for the two players---see \citet{backward_lifespan} and Appendix Section~\ref{sec:dd_rk}). We will use Runge--Kutta 4 as a low drift method for the rest of this chapter, and compare it with Euler updates in order to understand the effect of DD.

\begin{figure}[t]
  \centering
  \begin{subfloat}[Simultaneous updates.]{
  \includegraphics[width=0.48\columnwidth]{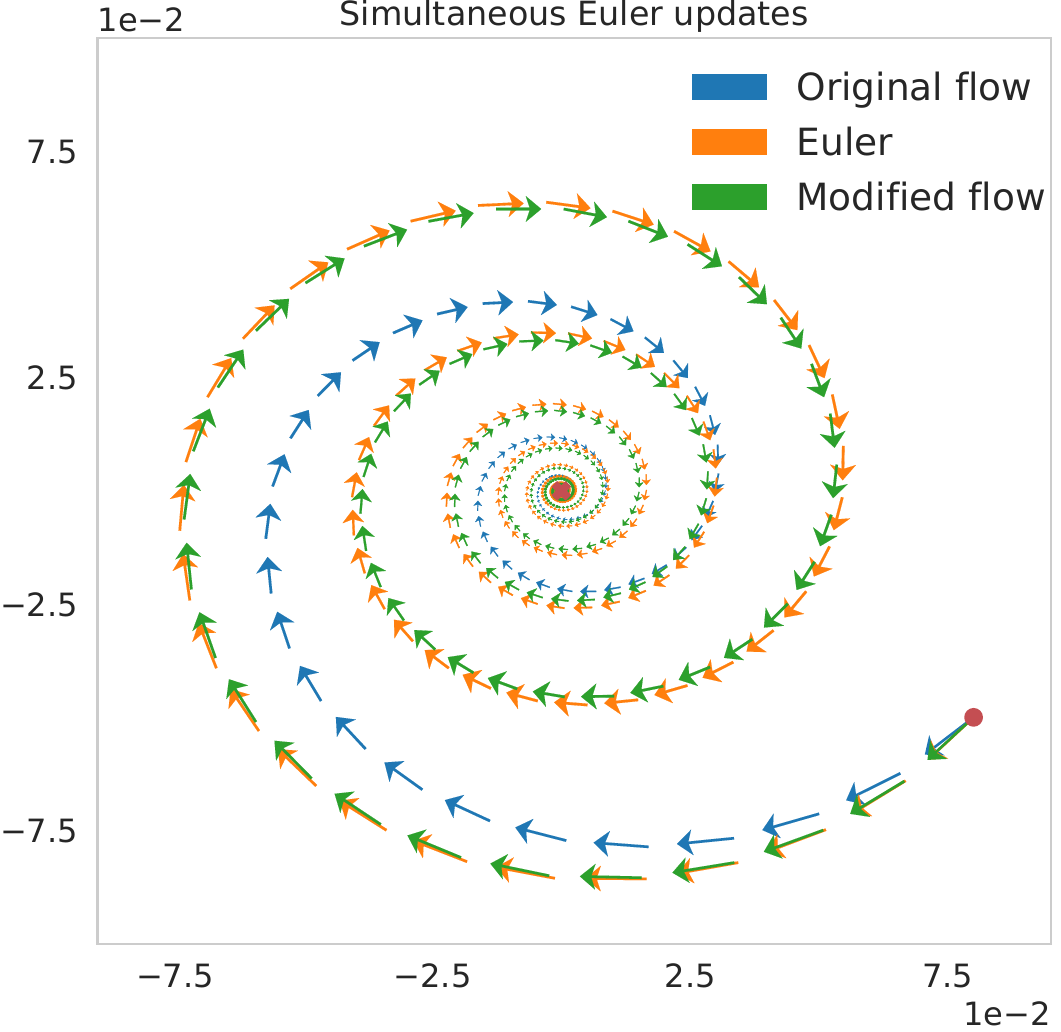}
  \label{fig:example_game_sim}
  }
  \end{subfloat}%
  \begin{subfloat}[Alternating updates.]{
    \includegraphics[width=0.48\columnwidth]{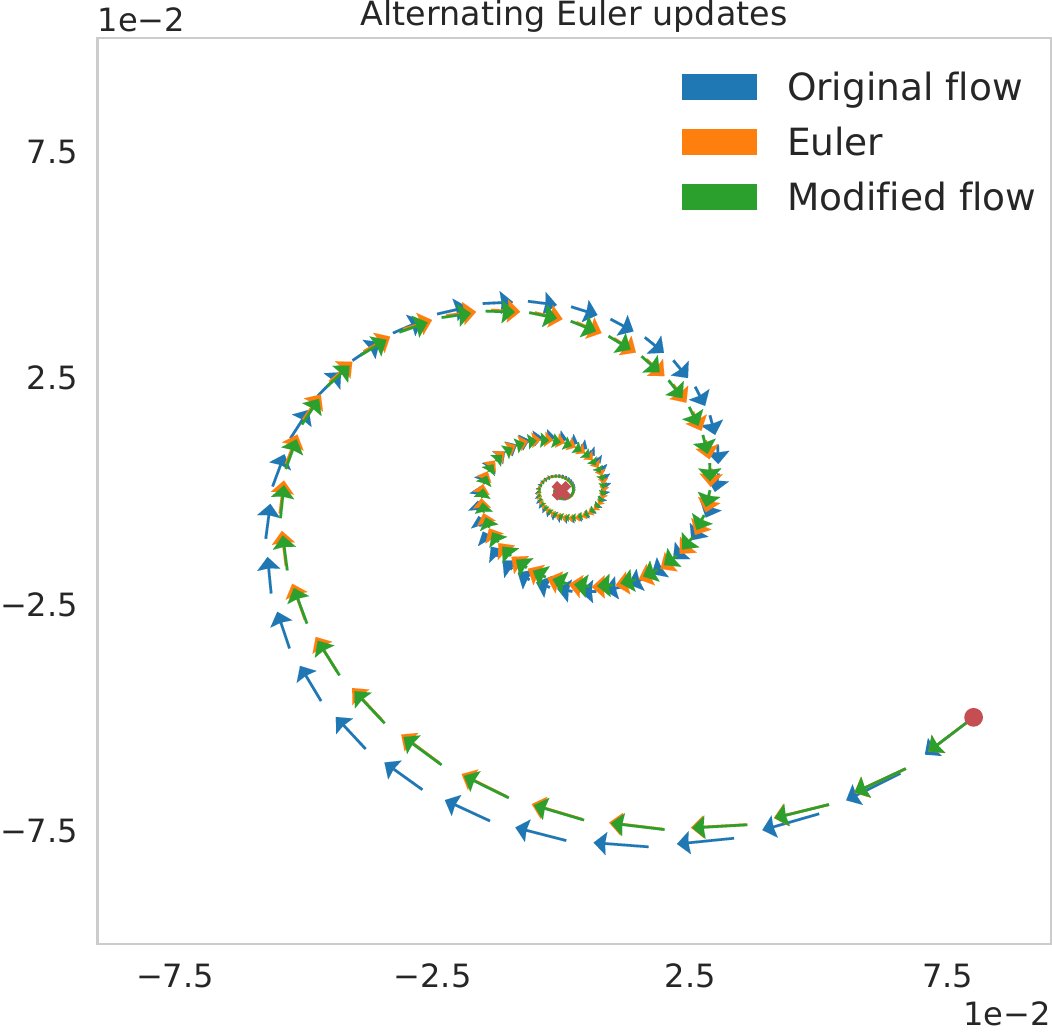}
  \label{fig:example_game_alt}
}\end{subfloat}%
  \caption[Modified flows given by BEA better capture discrete dynamics on a two dimensional example.]{By capturing the first order DD, the modified continuous flows we derive better capture the dynamics of their corresponding discrete update than the original game continuous dynamics. By doing so, the flows also capture the difference between simultaneous \subref{fig:example_game_sim} and alternating Euler updates \subref{fig:example_game_alt}.}
  \label{fig:igr_dynamics_toy}
\end{figure}

\section{The stability of DD}
\label{section:stability}

The long-term behaviour of gradient-based training can be characterised by the stability of its equilibria.
Using stability analysis of a continuous system to understand discrete dynamics in two-player games has a fruitful history; however, prior work ignored discretisation drift, since they analyse the stability of the unmodified game flows in Eq~\eqref{eq:ode1} and~\ref{eq:ode2}. A general overview of stability analysis is provided in Section~\ref{sec:stability_analysis_overview}.
 Using the modified flows given by BEA provides two benefits here: first, they account for the discretisation drift, and second, they provide \textit{different} flows for simultaneous and for alternating updates, capturing the specificities of the two optimisers.

The modified flow approach gives us a method to analyse the stability of the discrete updates: 1) Choose the system and the update type---simultaneous or alternating---to be analysed; 2) write the modified flows for the chosen system as given by Theorems~\ref{thm:sim} and~\ref{thm:alt}; 3) write the corresponding modified Jacobian, and evaluate it at an equilibrium; 4) determine the stability of the equilibrium by computing the eigenvalues of the modified Jacobian, using Remark~\ref{rem:stability_analysis}.

Steps 1 and 2 are easy, and step 4 is required in the stability analysis of any continuous system. For step 3, we provide a general form of the modified Jacobian at the equilibrium point of the original system, where $f=\myvec{0}$ and $g=\myvec{0}$
\begin{gather}
 \widetilde{\vJ}
 =
  \begin{bmatrix}
    \jacparam{\vphi}{\widetilde{f}} & \jacparam{\vtheta}{\widetilde{f}} \\
    \jacparam{\vphi}{\widetilde{g}}  &
   \jacparam{\vtheta}{\widetilde{g}}
   \end{bmatrix}
   = \vJ - \frac{h}{2} \mathbf{K},
\end{gather}
where $\vJ$ is the Jacobian of the un-modified system in Eqs~\eqref{eq:ode1}-\eqref{eq:ode2} and $\mathbf{K}$ depends on the update type; proofs are provided in Section~\ref{sec:dd_gan_stability_analysis}. For simultaneous updates 
\begin{gather}
 \mathbf{K}_{\text{sim}}= %
  \begin{bmatrix}
    \lrp ( \jacphif)^2 +  \lrp \jacthetaf \jacphig  & \lrp \jacphif \jacthetaf +  \lrp \jacthetaf \jacthetag\\
   \lrt \jacthetag  \jacphig +  \lrt \jacphig \jacphif&
   \lrt( \jacthetag)^2 +  \lrt \jacphig \jacthetaf  
   \end{bmatrix}.
\end{gather}
For alternating updates, we have:
\begin{gather}
 \mathbf{K}_{\text{alt}}
 =
  \begin{bmatrix}
    \frac {\lrp} {m} ( \jacphif)^2 + \lrp \jacthetaf \jacphig   & \frac{ \lrp} {m} \jacphif \jacthetaf +  \lrp \jacthetaf \jacthetag\\
   \frac{\lrt}{k} \jacthetag \jacphig +  \lrt (1 - \frac{2 \lrp}{\lrt}) \jacphig \jacphif&
   \frac {\lrt}{k}( \jacthetag)^2 +  \lrt (1 - \frac{2 \lrp}{\lrt}) \jacphig \jacthetaf
   \end{bmatrix}.
\end{gather}
Using the method above we show that, in two-player games, the drift term of order $\mathcal{O}(h^2)$ can change a stable equilibrium into an unstable one. This contrasts games and supervised learning, as the stability analysis of the IGR flow shows it is attracted to strict local minima (see~\citet{igr} and Section~\ref{sec:pf_stability_analysis}). For simultaneous updates with equal learning rates, this recovers a result of~\citet{daskalakis2018limit} derived for zero-sum games. We show this by example: consider the game given by the system of flows in Eq~\eqref{eq:simple_odes}.  By replacing the values of $f$ and $g$ into the results for the Jacobian of the modified flows obtain above, we obtain the corresponding Jacobians. For simultaneous updates
\begin{gather}
 \tilde{\vJ}_{\text{sim}}
 =
  \begin{bmatrix}
    -\epsilon_1 - h / 2 \epsilon_1^2 + h /2 & 1 + h / 2  \epsilon_1 - h / 2 \epsilon_2 \\
  -1 - h /2  \epsilon_1 + h /2  \epsilon_2 & \epsilon_2 + h / 2 - h/2  \epsilon_2^2
   \end{bmatrix},
\end{gather}
while for alternating updates
\begin{gather}
 \tilde{\vJ}_{\text{alt}}
 =
  \begin{bmatrix}
  - \epsilon_1 - h / 2   \epsilon_1^2 + h /2 & 1 + h / 2   \epsilon_1 - h / 2  \epsilon_2 \\
  -1 {\color{red}+} h /2   \epsilon_1 + h /2  \epsilon_2 & \epsilon_2 {\color{red}-} h / 2 - h/2  \epsilon_2 ^2
   \end{bmatrix}.
\end{gather}

 If we consider the system with $\epsilon_1 = \epsilon_2 = 0.09$, the stability analysis of its modified flows for the simultaneous Euler updates shows they diverge when $\lrp h = \lrt h =0.2$, since $\Tr({\tilde{\vJ}_{\text{sim}}}) = 0.19 > 0$, and thus at least one eigenvalue has positive real part. 
 For alternating updates, we obtain $\Tr(\tilde{\vJ}_{\text{alt}}) = -0.0016 < 0$ and 
 $|\tilde{\vJ}_{\text{alt}}| = 0.981 > 0$, leading to a stable system. In both cases, the results obtained using the stability analysis of the modified flows is consistent with empirical outcomes obtained by following the corresponding discrete updates, as shown in Figure~\ref{fig:igr_divergence}; this would not have been the case had we used the original system to do stability analysis, which would have always predicted convergence to an equilibrium and would not have been able to distinguish between simultaneous and alternating updates.
\begin{figure}[t]
  \centering
  \begin{subfloat}[Simultaneous updates.]{
  \includegraphics[width=0.48\columnwidth]{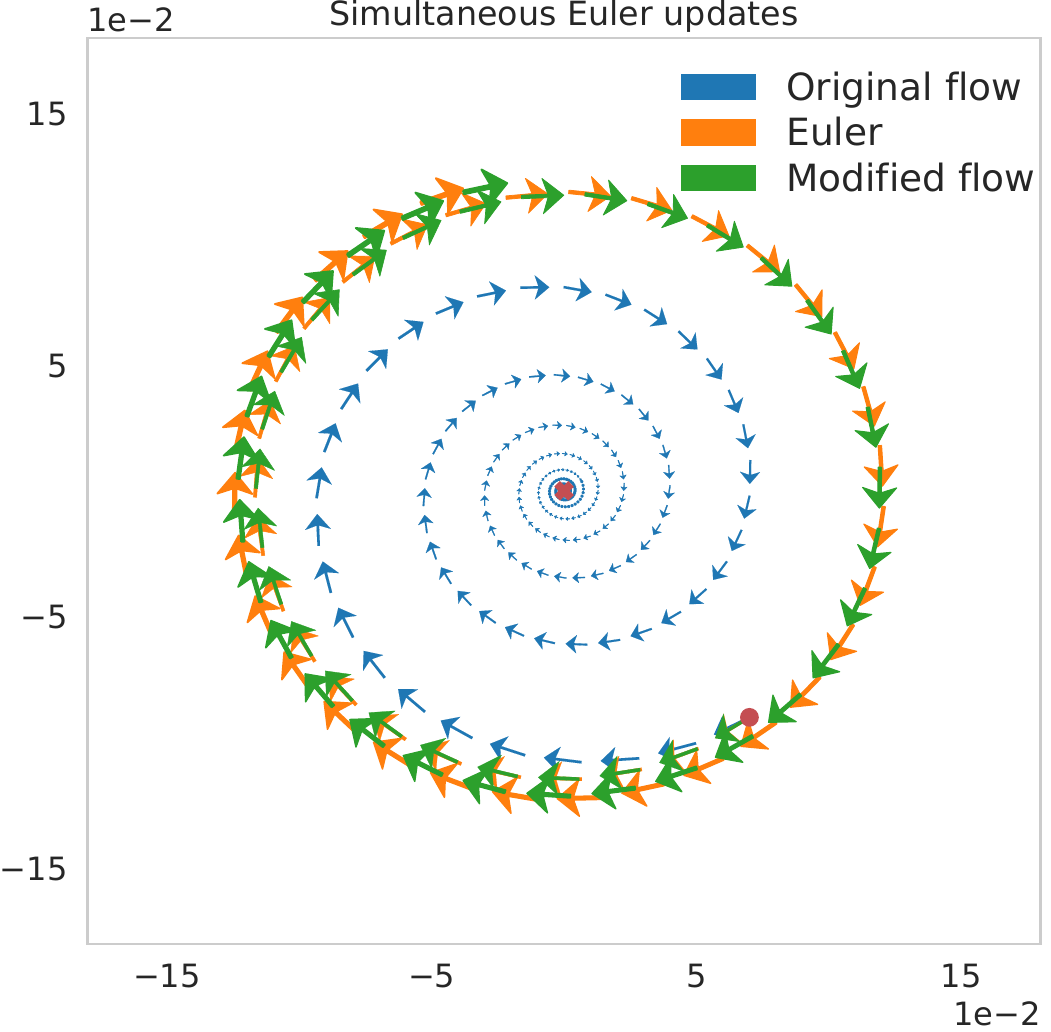}
  \label{fig:igr_div_sim}
  }
  \end{subfloat}%
  \begin{subfloat}[Alternating updates.]{
    \includegraphics[width=0.48\columnwidth]{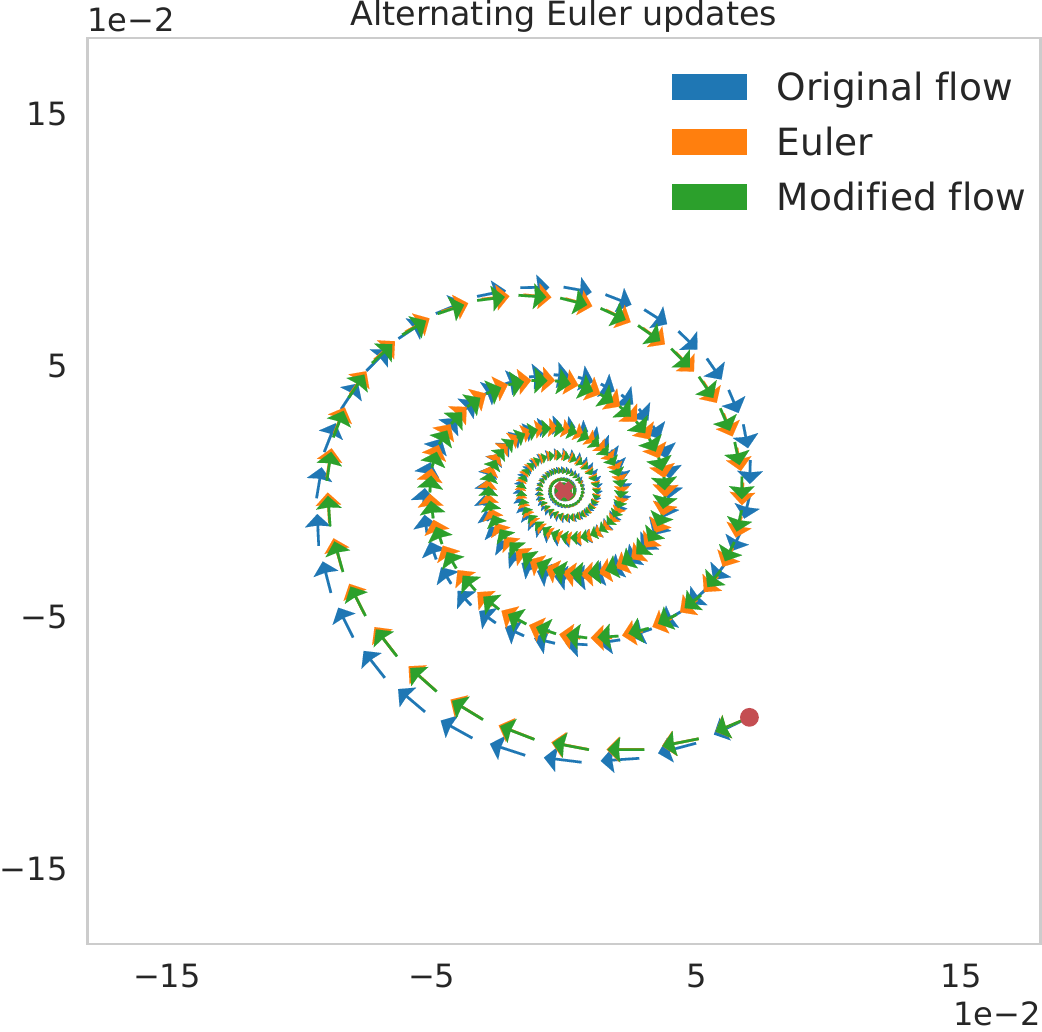}
  \label{fig:igr_div_alt}
}\end{subfloat}%
  \caption[Discretisation drift can change the stability of a game.]{Discretisation drift can change the stability of a game. We show here an example where the flow given by the original game dynamics converges, but simultaneous Euler updates do not \subref{fig:igr_div_sim}; for the same system, alternating updates converge \subref{fig:igr_div_alt}. In both cases, the modified flows capture the convergent or divergent behaviour of Euler updates. $\epsilon_1 = \epsilon_2 = 0.09$ and a learning rate $\lrp h = \lrt h = 0.2$.}
  \label{fig:igr_divergence}
\end{figure}

\textit{Benefits and caveats of using the modified flows for stability analysis}.
The modified flows help us bridge the gap between theory and practice: they allow us to extend the reach of stability analysis to a wider range of techniques used for training, such as alternating gradient descent. We hope the method we provide will be used in the context of GANs, to expand prior work, such as that of~\citet{nagarajan2017gradient}, to alternating updates. However, the modified flows we propose here are not without limitations: they ignore discretisation errors smaller than $\mathcal{O}(h^3)$, and thus they are not equivalent to the discrete updates. We have seen in Chapter~\ref{ch:pf} how in supervised learning higher order terms can be crucial in determining instabilities of gradient descent; we will briefly investigate the effect of these higher order terms in games in Section~\ref{sec:bea_pf_games}.
To fully account for DD, methods that directly assess the convergence of discrete updates (e.g.~\citet{mescheder2017numerics}) remain an indispensable tool for understanding discrete systems.

\section{Common-payoff games}
\label{sec:common_payoff}

When the players share a common loss, as in \emph{common-payoff games}, we recover supervised learning with a single loss $E$, but with the extra-freedom of training the weights corresponding to different parts of the model with possibly different learning rates and update strategies (see for instance \citet{imagenet_in_minutes} where a per-layer learning rate is used to obtain extreme training speed-ups at equal levels of test accuracy).
A special case occurs when two players with equal learning rates ($\lrp = \lrt$) perform simultaneous gradient descent. In this case, the modified losses recover the Implicit Gradient Regularisation flow in Eq~\eqref{eq:sup_learning_igr}. \citet{igr} argue that minimising the loss-gradient norm, in this case, has a beneficial effect.

In this section, we instead focus on alternating gradient descent. We partition a neural network into two sets of parameters, $\vphi$ for the parameters closer to the input and $\vtheta$ for the parameters closer to the output. This procedure freezes one part of the network while training the other part and alternating between the two parts.
This scenario may arise in a distributed training setting, as a form of block coordinate descent.
In the case of common-payoff games, we have the following as a special case of Theorem~\ref{thm:alt} by substituting $f = -\nabla_{\vphi} E$ and $g = -\nabla_{\vtheta} E$:

\begin{cor} \label{cor:common_payoff_alt} In a two-player common-payoff game with common loss $E$, \textit{alternating} gradient descent---as described in Eqs~\eqref{eq:altup1} and~\eqref{eq:altup2}---with one update per player follows a gradient flow given by the modified losses
\begin{align}
\tilde E_{\vphi}&= E + \frac{\lrp h}{4} \left(\norm{\nabla_{\vphi} E}^2  + \norm{\nabla_{\theta} E}^2\right) \\
\tilde E_{\vtheta}&=  E + \frac{\lrt h}{4} \left((1 - \frac {2 \lrp} {\lrt}) \norm{\nabla_{\vphi} E}^2  + \norm{\nabla_{\theta} E}^2\right)
\label{eq:cooperating_zero_sum}
\end{align}
with an error of size $\mathcal O(h^3)$  after one update step.
\end{cor}

The term $(1 - \frac {2 \lrp} {\lrt}) \norm{\nabla_{\vphi} E}^2$ in Eq.~\eqref{eq:cooperating_zero_sum} is negative when the learning rates are equal, seeking to maximise the gradient norm of player $\vphi$. According to \citet{igr}, we expect less stable training and worse performance in this case.
This prediction is confirmed in Figure~\ref{fig:alternating_updates_supervised_learning}, where we compare simultaneous and alternating gradient descent for a MLP trained on MNIST with a common learning rate.

\begin{figure}[t]
\centering
\begin{subfloat}[Gradient norm of the first player.]{
  \includegraphics[width=0.51\columnwidth]{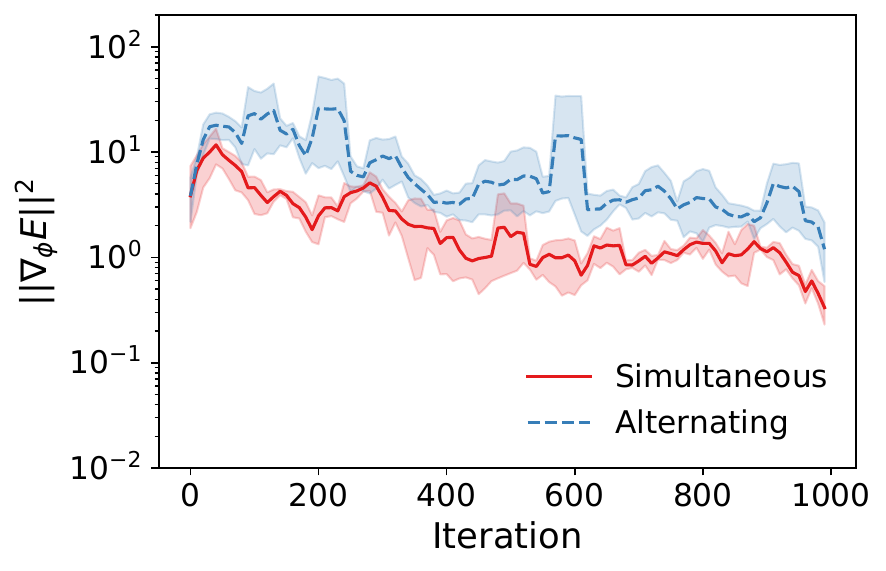}%
  \label{fig:gradient_norm_common_payoff}
}
\end{subfloat}%
\begin{subfloat}[Accuracy.]{
   \includegraphics[width=0.49\columnwidth]{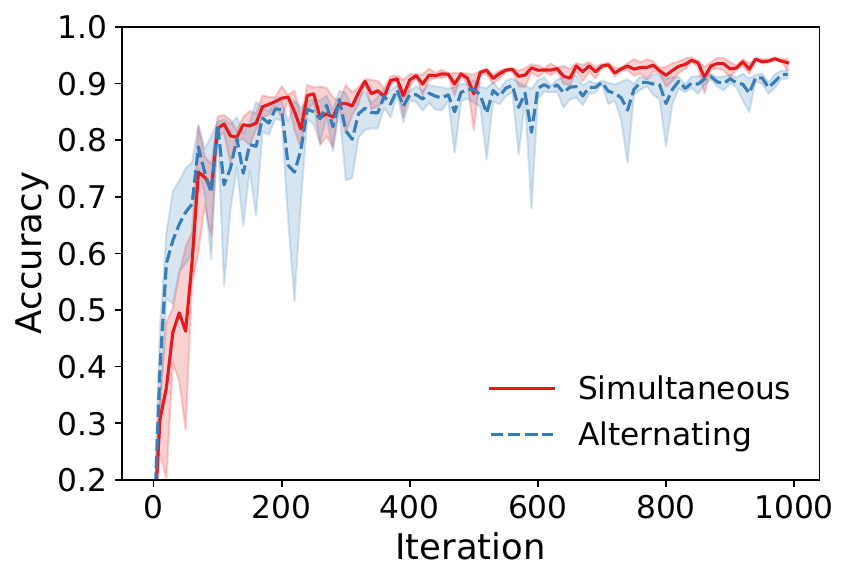}%
  \label{fig:accuracy_common_payoff}
}
\end{subfloat}%
  \caption[In common-payoff games alternating updates lead to higher gradient norms and unstable training when equal learning rates are used.]{In common-payoff games alternating updates lead to higher gradient norms and unstable training when equal learning rates $\lrp h= \lrt h$ are used. As predicted by Corollary~\ref{cor:common_payoff_alt}, the maximisation of the first player's gradient norm $\norm{\nabla_{\vphi} E}$ by the second player's modified loss in alternating updates leads to an increased gradient norm for the first player compared to simultaneous updates \subref{fig:gradient_norm_common_payoff}, which also results to loss instability and decrease in performance \subref{fig:accuracy_common_payoff}. The common-payoff objective here is classification of MNIST digits.}
  \label{fig:alternating_updates_supervised_learning}
\end{figure}

\section{Analysis of zero-sum games}
\label{section:zero_sum}

We now study zero-sum games, where the one player's gain is another player's loss: $ - E_{\vphi} = E_{\vtheta} = E$.
We substitute the updates $f = \nabla_{\vphi} E$ and $g = -\nabla_{\vtheta} E$ in the Theorems in Section~\ref{sec:general_theorems} and obtain:
\begin{cor} In a zero-sum two-player differentiable game, \textit{simultaneous} gradient descent updates---as described in Eqs~\eqref{eq:simup1} and~\eqref{eq:simup2}---follows a gradient flow given by the modified losses
\label{cor:zs-sim}
\begin{align}
\tilde E_{\vphi}&= - E + \frac{\lrp h}{4} \norm{\nabla_{\vphi} E}^2  -  \frac{\lrp h}{4} \norm{\nabla_{\theta} E}^2
\label{eq:min_min_simultaneous_zero_sum1} \\
\tilde E_{\vtheta}&= E -  \frac{\lrt h}{4} \norm{\nabla_{\vphi} E}^2  +  \frac{\lrt h}{4} \norm{\nabla_{\theta} E}^2,
\label{eq:min_min_simultaneous_zero_sum2}
\end{align}
with an error of size $\mathcal O(h^3)$ after one update step.
\end{cor}
 Corollary~\ref{cor:zs-sim} shows that the modified losses for simultaneous gradient descent preserve the adversarial structure of the game,
with the \textit{interaction terms} maximising the gradient norm of the opposite player. We note, however, that  the modified losses of zero-sum games trained with simultaneous gradient descent with different learning rates are no longer zero-sum, due to the different implicit regularisation coefficients introduced by DD.

\begin{cor} In a zero-sum two-player differentiable game, \textit{alternating} gradient descent---as described in Eqs~\eqref{eq:altup1} and~\eqref{eq:altup2}---follows a gradient flow given by the modified losses
\label{cor:zs-alt}
\begin{align}
\tilde E_{\vphi}&= -E + \frac{\lrp h}{4m} \norm{\nabla_{\vphi} E}^2  - \frac{\lrp h}{4} \norm{\nabla_{\theta} E}^2
\label{eq:min_min_alt_zero_sum1}
\\
\tilde E_{\vtheta}&= E - \frac{\lrt h}{4} (1 - \frac{2 \lrp}{\lrt}) \norm{\nabla_{\vphi} E}^2  + \frac{\lrt h}{4k}\norm{\nabla_{\theta} E}^2
\label{eq:min_min_alt_zero_sum2}
\end{align}
with an error of size $\mathcal O(h^3)$ after one update step.
\end{cor}
Corollary~\ref{cor:zs-alt} shows that the modified losses of zero-sum games trained with alternating gradient descent are not zero-sum.

\begin{remark} 
Since $(1 - \frac{2 \lrp}{\lrt})  < 1$ in the alternating case there is always less weight on the term encouraging maximizing $\norm{\nabla_{\vphi} E}^2$ compared to the simultaneous case under the same learning rates. For
$\frac{\lrp}{\lrt} > \frac{1}{2}$ both players minimise $\norm{\nabla_{\vphi} E}^2$.
\end{remark}

Corollaries~\ref{eq:min_min_simultaneous_zero_sum2} and~\ref{eq:min_min_alt_zero_sum2} show that in zero-sum games interaction terms can induce a pressure to maximise the gradient norm of the other player. We postulate that this maximisation effect can be a source of instability, something we will study throughout this chapter.

\subsection{Dirac-GAN: an illustrative example}\label{sec:DiracGAN}

\citet{mescheder2018training} introduce the Dirac-GAN as an example to illustrate the often complex training dynamics in zero-sum games. We follow this example to provide an intuitive understanding of DD.
Dirac-GAN aims to learn a Dirac delta distribution with mass at zero;
the generator is modelling a Dirac with parameter $\theta$: $G(z; \theta) = \theta$ and the discriminator is a linear model on the input with
parameter $\phi$: $D(x;\phi) = \phi x$. This results in the zero-sum game given by
\begin{align}
 E = l(\theta \phi) + l(0),
\label{eq:dirac_gan}
\end{align}
where $l$ depends on the GAN formulation used ($l(z) = - \log (1 + e^{-z})$ for instance).
As in~\citet{mescheder2018training}, we assume $l$ is continuously differentiable with $l'(x) \ne 0$ for all $x \in \mathbb{R}$. The partial derivatives of the loss function
\begin{align}
\frac{\partial E}{\partial\phi} = l'(\theta \phi) \theta, \hspace{3em}  \frac{\partial E}{\partial\theta} = l'(\theta \phi) \phi,
\end{align}
lead to the underlying continuous dynamics
\begin{align}
\dot{\phi} =  f(\theta, \phi) =  l'(\theta \phi) \theta, \hspace{3em} \dot{\theta} =  g(\theta, \phi) =  -l'(\theta \phi) \phi.
\label{eq:dirac_gan_vector_field}
\end{align}
The unique equilibrium point is $\theta = \phi = 0$. 

\textbf{Reconciling discrete and continuous updates in Dirac-GAN}. The original  continuous dynamics in Eq~\eqref{eq:dirac_gan_vector_field} preserve $\theta^2 + \phi^2$,
since
\begin{align}
\frac{d \left(\theta^2 + \phi^2\right)}{dt} = 2 \theta \frac{d \theta}{dt} + 2 \phi \frac{d \phi}{dt} = - 2 \theta l'(\theta \phi) \phi + 2 \phi l'(\theta \phi) \theta = 0.
\end{align}
\citet{mescheder2018training} show, however, that for simultaneous gradient descent $\theta^2 + \phi^2$ increases with each update: different conclusions are reached when analysing the dynamics of the original continuous system versus the discrete updates.
 We show that, by using the modified flows given by Eqs~\eqref{eq:min_min_simultaneous_zero_sum1} and~\eqref{eq:min_min_simultaneous_zero_sum2} instead of the original game dynamics, we can resolve this discrepancy. When following the modified flow induced by simultaneous gradient descent in DiracGAN, we have that (proof in Section~\ref{sec:dirac_gan_reconciling} in the Appendix), unless we start at the equilibrium $\theta = \phi = 0$, 
\begin{align}
&\frac{d \left(\theta^2 + \phi^2\right)}{dt} =   h \theta^2 l'(\phi, \theta) ^2 + h \phi^2 l'(\phi, \theta^2) > 0.
\label{eq:dirac_div}
\end{align}
We  thus obtain \textit{consistent conclusions from the modified continuous flows and the discrete dynamics used in practice}; we visualise this in Figure~\ref{fig:dirac_gan_reconciling}.

\begin{figure}[t]
  \centering
    \begin{subfloat}[Reconciling discrete and continuous dynamics.]{
  \includegraphics[width=0.485\columnwidth]{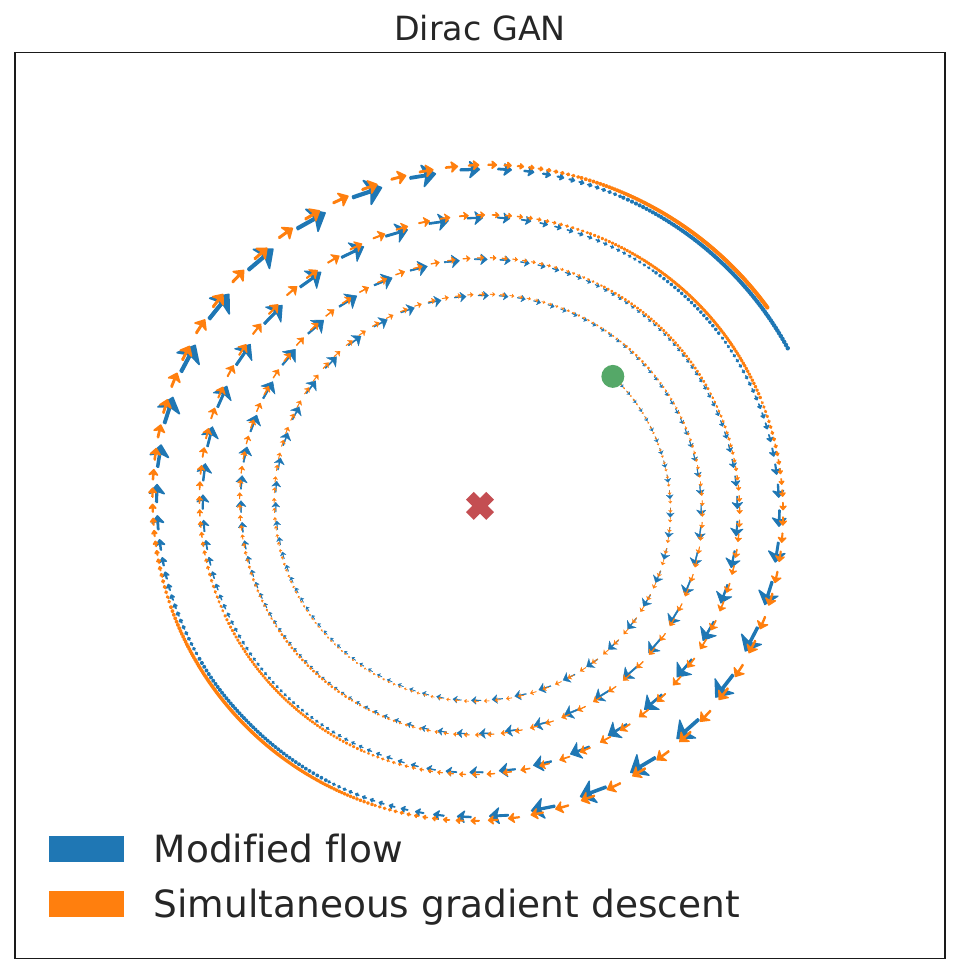}%
  \label{fig:dirac_gan_reconciling}
} \end{subfloat}%
    \begin{subfloat}[Explicit regularisation leads to convergence.]{
  \includegraphics[width=0.49\columnwidth]{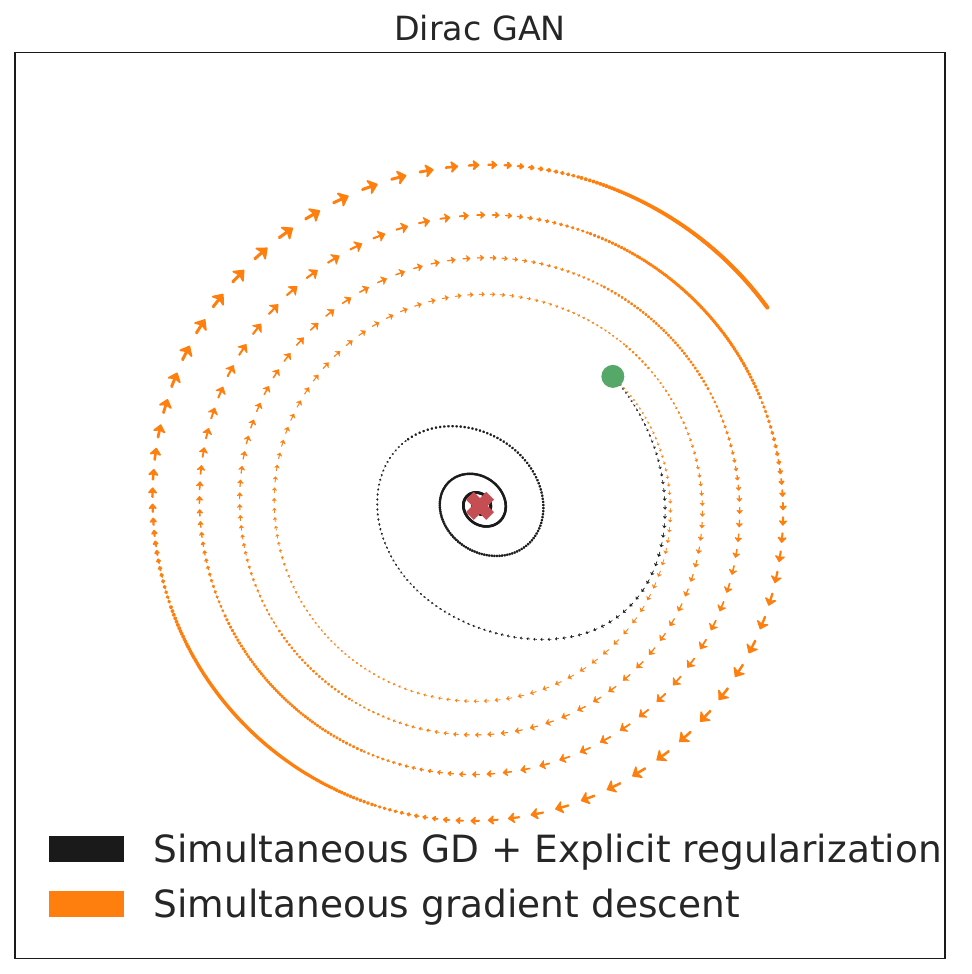}%
  \label{fig:dirac_gan_explicit}
} \end{subfloat}%
  \caption[Discretisation drift can cause instability in DiracGAN; cancelling the sources of instabilities leads to convergence.]{DiracGAN. \subref{fig:dirac_gan_reconciling}: The dynamics of simultaneous gradient descent updates match the continuous dynamics given by BEA for DiracGAN. \subref{fig:dirac_gan_explicit}: Explicit regularisation cancelling the interaction terms of DD stabilises the DiracGAN trained with simultaneous gradient descent.}
  \label{fig:dirac_gan}
\end{figure}

\textbf{Explicit regularisation stabilises Dirac-GAN}. We show that the instability of simultaneous updates in DiracGAN can be seen as a result of the effect of interaction terms in zero-sum games, as we postulated as a result of Corollary~\ref{eq:min_min_simultaneous_zero_sum2}. We counteract the norm maximisation effect of the interaction terms in simultaneous updates by cancelling the interaction terms using explicit regularisation: $E_{\vphi}  = - E + \gamma{\| \nabla_{\theta} E\|}^2$ and $E_{\vtheta}  = E + \zeta {\| \nabla_{\phi} E\|}^2$  where $\gamma, \zeta$ are of $\mathcal{O}(h)$. We find (proofs are provided in Section~\ref{sec:dirac-gan-exp-reg} of the Appendix) that the modified Jacobian of simultaneous updates induced by these new losses is negative definite if  $\gamma > h\lrp /4$ and $\zeta > h\lrt /4$---which are exactly the coefficients required to cancel the interaction terms shown in Corollary~\ref{eq:min_min_simultaneous_zero_sum2}---so the system is asymptotically stable and converges to the optimum. Unlike the modified flow of the original zero-sum dynamics, which diverge as we have seen in Eq~\ref{eq:dirac_div}, the flow of the system induced by cancelling the interaction terms converges. We visualise this behaviour in Figure~\ref{fig:dirac_gan_explicit}. Notably, by quantifying DD, we are able to find the regularisation coefficients that guarantee convergence and show that they depend on learning rates.

\section{Experimental analysis of GANs} \label{section:experimental}

To understand the effect of DD on more complex adversarial games, we analyse GANs trained for image generation on the CIFAR-10 dataset. We follow the model architectures from Spectral-Normalised GAN~\citep{miyato2018spectral}.
Both players have millions of parameters. Unlike Spectral-Normalised GAN, we employ the original GAN formulation, as described in Eq~\eqref{eq:original_gan}.

When it comes to gradient descent, GAN practitioners often use alternating, not simultaneous updates: the discriminator is updated first, followed by the generator; we described alternating updates and simultaneous updates in Algorithms~\ref{alg:alt_updates} and~\ref{alg:sim_updates}, respectively. Recent work, however, shows higher-order numerical integrators can work well with simultaneous updates~\citep{odegan}. We will show that DD can be seen as the culprit behind some of the challenges in simultaneous gradient descent in zero-sum GANs, indicating ways to improve training performance in this setting.
For a clear presentation of the effects of DD, we employ a minimalist training setup. Instead of using popular adaptive optimisers, such as Adam~\citep{kingma2014adam}, we train all models with vanilla stochastic gradient descent,
 without momentum or variance reduction methods.

We use the Inception Score (IS)~\citep{improved_techniques_gans}
for evaluation, and we report Fréchet Inception Distance (FID) results~\citep{heusel2017gans} in the Appendix; we described the computation of the Inception Score in Section~\ref{sec:dd_gan_background}.
Our training curves contain a horizontal line at the Inception Score of $7.5$, obtained with the same architectures we use, but with the Adam optimiser (the score reported by~\citet{miyato2018spectral} is $7.42$). 
We report box plots showing the performance quantiles across all hyperparameter and seeds, together with \textit{learning curves corresponding to the best $10\%$ models from a sweep over learning rates and 5 seeds}. This is due to the large variability observed across hyperparameters and seeds, but we observe the same relative results with the top $20\%$ and $30\%$ of models, as we show in the Appendix Section~\ref{sec:dd_gan_exp_res_app}, which also has additional discussions on robustness.
For SGD we use learning rates $\{ 5 \times 10^{-2}, 1\times 10^{-2}, 5 \times 10^{-3}, 1 \times 10^{-3}\}$ for each player; for Adam, we use learning rates  $\{ 1 \times 10^{-4}, 2\times 10^{-4}, 3 \times 10^{-4}, 4 \times 10^{-4}\}$, which have been widely used in the literature~\citep{miyato2018spectral}. When comparing to Runge--Kutta (RK4) to assess the effect of DD, we always use the same learning rates for both players. We present results using additional losses, via LS-GAN~\citep{ls_gan}, and experimental details in the Appendix. The code associated with this work can be found at \url{https://github.com/deepmind/discretisation_drift}.

\subsection{Does DD affect training?}

 \begin{figure}[t]
 \centering
  \begin{subfloat}[A sweep over a range learning rates.]{
   \includegraphics[width=0.49\columnwidth]{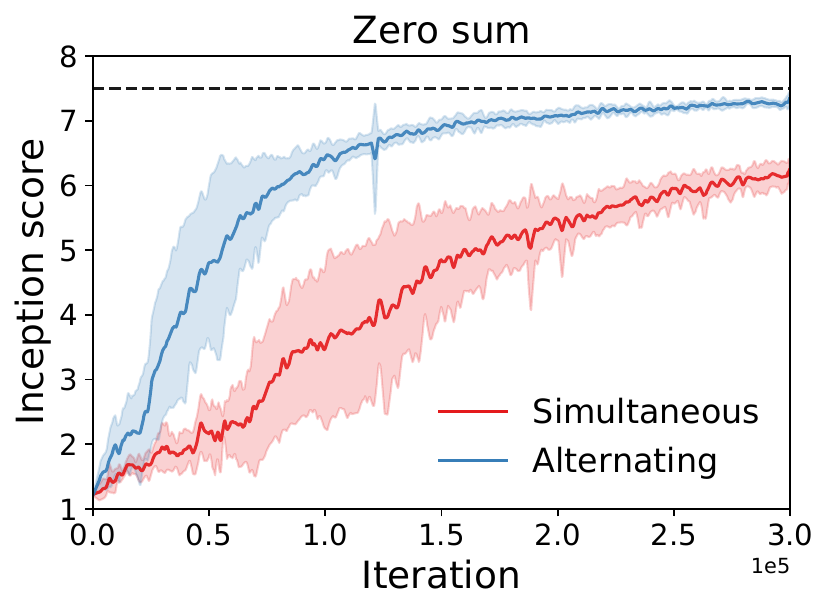}%
   \label{fig:is_zero_sum_diff_lr}
  }
  \end{subfloat}%
  \begin{subfloat}[Equal learning rates.]{
    \includegraphics[width=0.49\columnwidth]{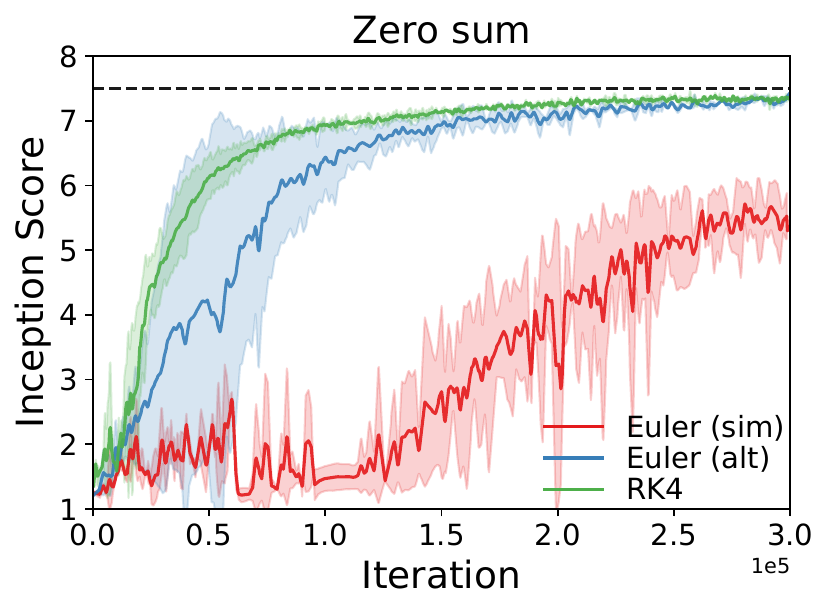}%
    \label{fig:is_zero_sum_equal_lr}
  }
  \end{subfloat}%
    \caption[The effect of discretisation drift on zero-sum games.]{The effect of DD on zero-sum games. Over a sweep of learning rates, alternating updates perform better than simultaneous updates \subref{fig:is_zero_sum_diff_lr}, and show lower variance. When using equal learning rates RK4 has $\mathcal{O}(h^5)$ drift, and thus we use it as a baseline to understand the effect of the drift in \subref{fig:is_zero_sum_equal_lr}; we observe that reducing drift leads to improved performance. These results show that DD has a strong effect on these games, with a substantial difference in performance between simultaneous and alternating updates, and an increased speed early in training when using RK4. While the modified flows we study only capture DD up to $\mathcal{O}(h^3)$, these results are consistent with what we expect from Corollaries~\ref{cor:zs-sim} and~\ref{cor:zs-alt}, which show that the drift of the first player leads to an incentive to maximise the second player's gradient norm; this can lead to instability compared to methods with reduced drift such as RK4 as we have seen in \subref{fig:is_zero_sum_equal_lr}. Similarity when comparing alternating and simultaneous updates, we expect less instability for alternating updates as the second player does not always have the incentive to maximise the first player's gradient norm, which is always the case for simultaneous updates; this is consistent with what we observe in \subref{fig:is_zero_sum_diff_lr}.}
   \label{fig:idd_zero_sum_games}
\end{figure}

We start our experimental analysis by showing the effect of DD on zero-sum games. We compare simultaneous gradient descent, alternating gradient descent, and Runge--Kutta 4 updates,
 since they follow different continuous dynamics given by the modified equations we have derived up to $\mathcal{O}(h^3)$ error. Figure~\ref{fig:idd_zero_sum_games} shows simultaneous gradient descent performs substantially worse than alternating gradient descent. When compared to Runge--Kutta 4, which has a DD error of $\mathcal{O}(h^5)$ when the two players have equal learning rates, we see that Runge--Kutta performs better, and that removing the drift improves training. Multiple updates also affect training, either positively or negatively, depending on learning rates---see Appendix Section~\ref{sec:dd_rk}.

\subsection{The importance of learning rates in DD}
\label{section:learning_rate_ratios}

While in simultaneous updates (Eqs~\eqref{eq:min_min_simultaneous_zero_sum1} and~\eqref{eq:min_min_simultaneous_zero_sum2}) the interaction terms of both players maximise the gradient norm of the other player, alternating gradient descent (Eqs~\eqref{eq:min_min_alt_zero_sum1} and~\eqref{eq:min_min_alt_zero_sum2}) exhibits less pressure on the second player (generator) to maximise the norm of the first player (discriminator). In alternating updates, when the ratio between the discriminator and generator learning rates exceeds $0.5$, both players are encouraged to minimise the discriminator's gradient norm. To understand the effect of learning rate ratios in training, we performed a sweep where the discriminator learning rates $\lrp$ are sampled uniformly between $[0.001, 0.01]$, and the learning rate ratios $\lrp / \lrt$ are in $\{0.1, 0.2, 0.5, 1, 2, 5\}$, with results shown in Figure~\ref{fig:sim_vs_alt_zero_sum_learning_rates}. Learning rate ratios greater than $0.5$ perform best for alternating updates
and enjoy a substantial increase in performance compared to simultaneous updates, which is consistent with our expectations based on the sign of the interaction terms.

\begin{figure}[t]
  \begin{subfloat}[Simultaneous updates.]{
   \includegraphics[width=0.49\columnwidth]{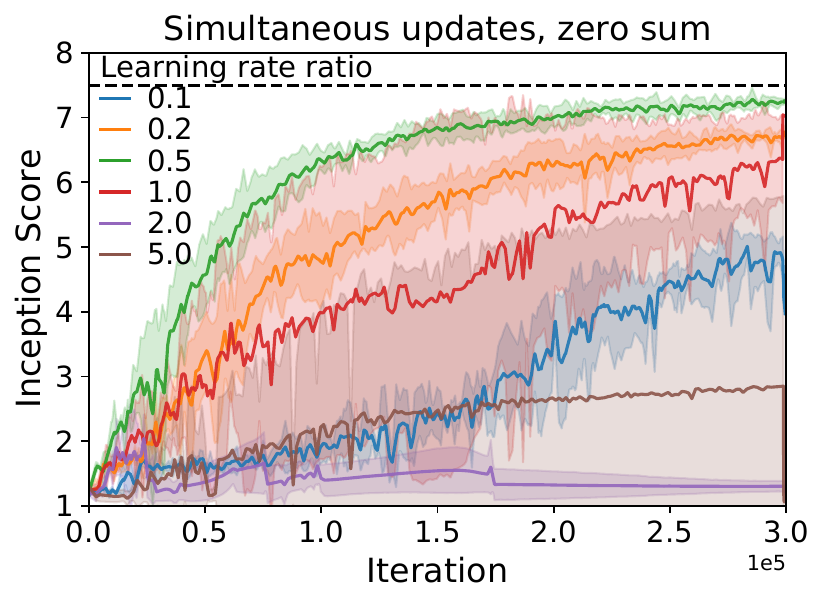}%
   \label{fig:sim_lr_ratios}
  }
  \end{subfloat}%
  \begin{subfloat}[Alternating updates.]{
  \includegraphics[width=0.49\columnwidth]{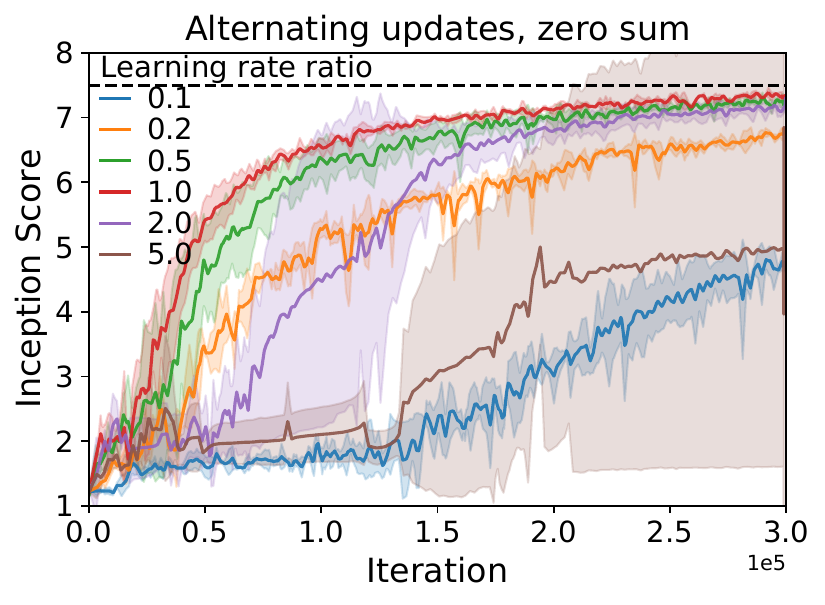}%
   \label{fig:alt_lr_ratios}
  }
  \end{subfloat}%
   \caption[Alternating gradient descent performs better for learning rate ratios which reduce the adversarial nature of discretisation drift.]{Alternating gradient descent \subref{fig:alt_lr_ratios} performs better for learning rate ratios which reduce the adversarial nature of DD, namely learning rates where $\lrp/\lrt \ge 0.5$ and thus the second player's (the generator) modified loss minimises, instead of maximises the discriminator's gradient norm. The same learning rate ratios show no advantage in the simultaneous case \subref{fig:sim_lr_ratios}.}
   \label{fig:sim_vs_alt_zero_sum_learning_rates}
\end{figure}

\subsection{Improving performance by explicit regularisation}
\label{section:explicit_regularisation}

We have postulated that the interaction terms can hinder performance and stability in zero-sum games, as they can lead to an incentive to maximise the gradient norm of one, or both players. Now, 
we investigate whether \textit{cancelling the interaction terms} between the two players can improve training stability and performance in zero-sum games trained using simultaneous gradient descent. We train models using the losses
\begin{align}
E_{\vphi} &= - E + c_1 \norm{\nabla_{\vtheta} E}^2 \\
E_{\vtheta} &= E + c_2 \norm{\nabla_{\vphi} E}^2.
\label{eq:explicit_regularisation_cancel_interaction}
\end{align}

\begin{figure}[t]
 \centering
  \begin{subfloat}{
  \includegraphics[width=0.48\columnwidth]{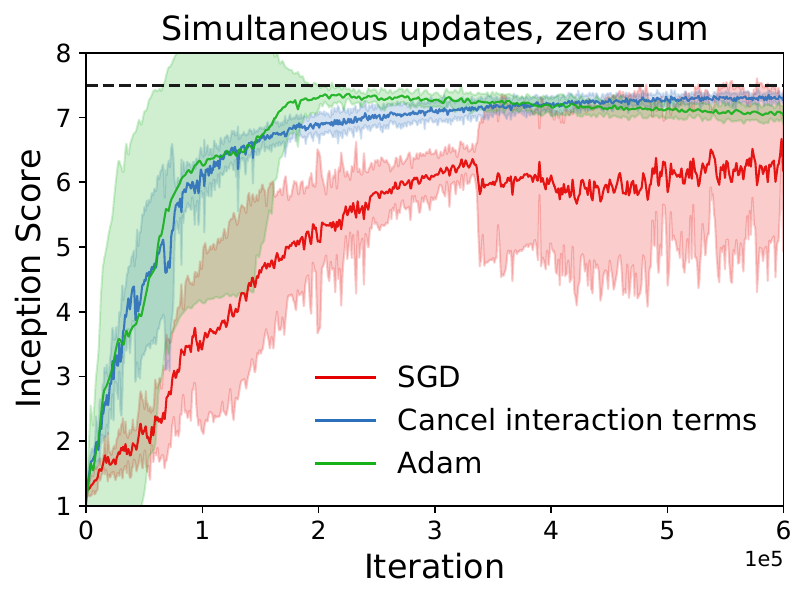}%
  \label{fig:sgd_comp_main_paper}
} \end{subfloat}%
 \begin{subfloat}{
  \includegraphics[width=0.51\columnwidth]{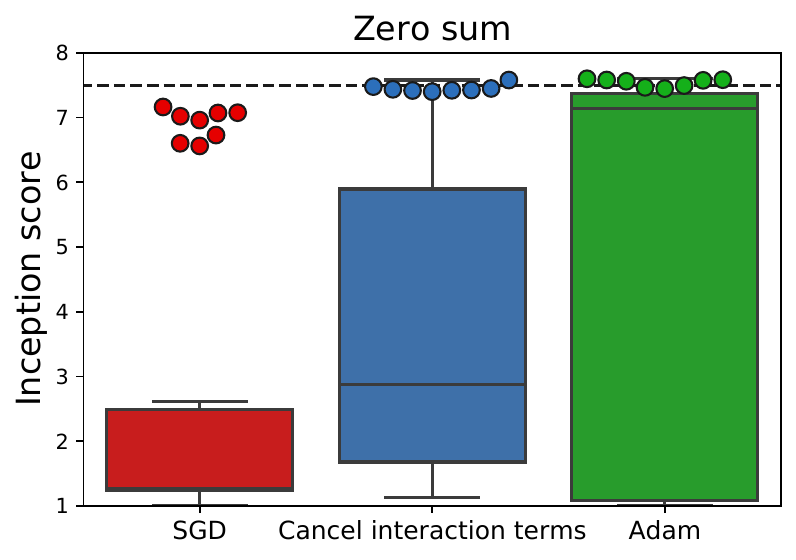}%
  \label{fig:cancel_explicit_sim_box}
} \end{subfloat}%
  \caption[Explicit regularisation cancelling the interaction terms of discretisation drift in simultaneous gradient descent improves performance in zero-games.]{Simultaneous updates in zero-sum games: Explicit regularisation cancelling the interaction terms of DD improves performance and stability, both for the best performing models \subref{fig:sgd_comp_main_paper} and across a learning rate sweep \subref{fig:cancel_explicit_sim_box}.}
  \label{fig:sgd_adam_sim_comparison}
\end{figure}

If $c_1, c_2$ are $\mathcal{O}(h)$ we can ignore the DD from these regularisation terms, since their effect on DD will be of order $\mathcal{O}(h^3)$.
We can set coefficients to be the negative of the coefficients present in DD, namely $c_1 = \lrp h/4$ and $c_2 = \lrt h/4$ to cancel the interaction terms, which maximise the gradient norm of the other player, while keeping the self terms, which minimise the player's own gradient norm.
We show results in Figure~\ref{fig:sgd_adam_sim_comparison}: cancelling the interaction terms leads to substantial improvement compared to SGD, obtains the same peak performance as Adam (though requires more training iterations) and recovers the performance of Runge--Kutta 4 (Figure~\ref{fig:idd_zero_sum_games}). Unlike Adam, we do not observe a decrease in performance later in training but report higher variance in performance across seeds---see Figure~\ref{fig:cancel_explicit_sim_box} and additional experiments in Appendix Section~\ref{sec:dd_gan_exp_res_app}.

\begin{figure}[t]
 \centering
  \begin{subfloat}[SGA comparison.]{
  \includegraphics[width=0.49\columnwidth]{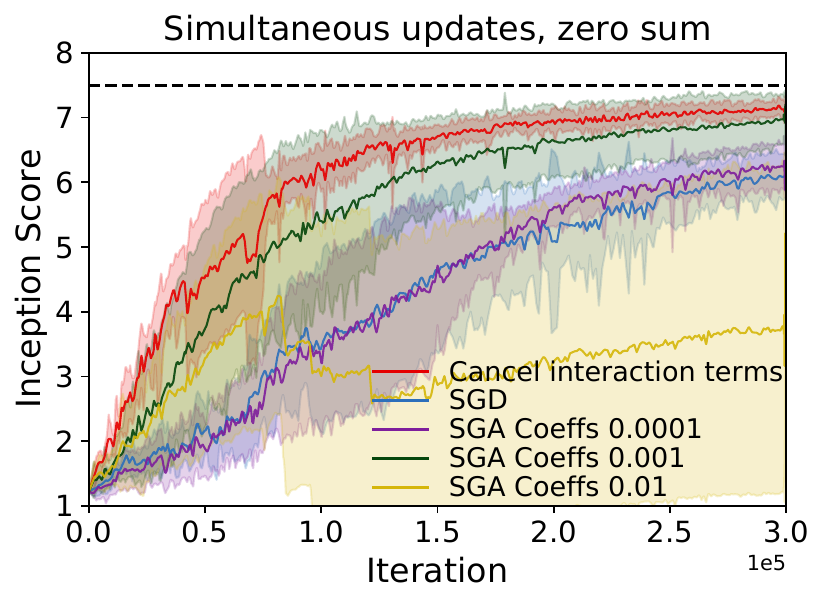}%
  \label{fig:sga_comp}
} \end{subfloat}%
 \begin{subfloat}[CO comparison.]{
 \includegraphics[width=0.49\columnwidth]{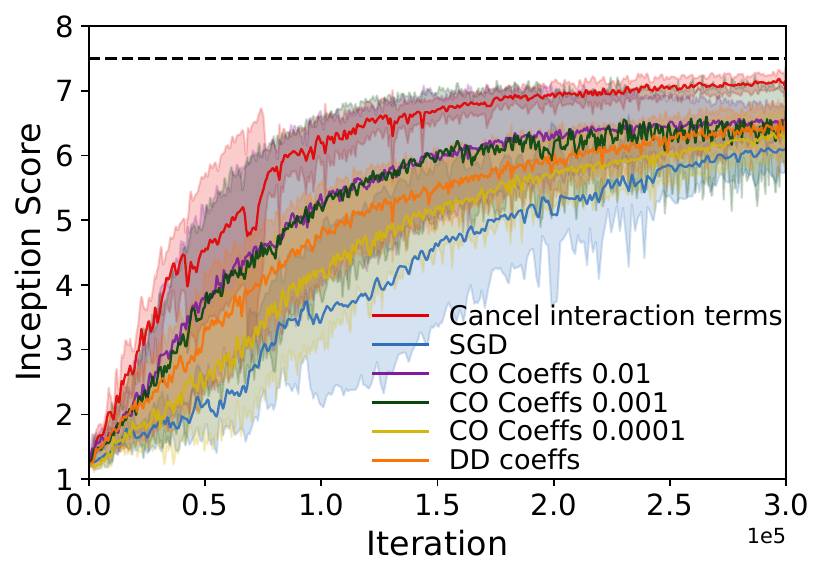}%
  \label{fig:co_comp}
} \end{subfloat}%
  \caption[Comparison with Symplectic Gradient Adjustment and Consensus Optimisation.]{Comparison with Symplectic Gradient Adjustment (SGA) and Consensus Optimisation (CO) in simultaneous gradient descent for zero-sum games: DD motivates  the form of explicit regularisation and provides does not require a larger hyperparameter sweep for the regularisation coefficient, as the explicit regularisation coefficients are determined by the player's learning rates. Cancelling the interaction terms leads to improved performance compared to both SGA and CO.}
  \label{fig:sga_consensus_opt_main_paper}
\end{figure}

\textit{Connection with Symplectic Gradient Adjustment (SGA)}: \citet{balduzzi2018mechanics} proposed SGA to improve the dynamics of gradient-based method for games, by counter-acting the rotational force of the vector field (see Section~\ref{app:sga} for a discussion). Adjusting the gradient field can be viewed as modifying the losses as in Eq~\eqref{eq:explicit_regularisation_cancel_interaction}; the modification from SGA cancels the interaction terms we identified. However, it is unclear whether the fixed coefficients of $c_1 = c_2 = \frac{1}{2}$ used by SGA are always optimal, while our analysis shows the strength of DD changes with the exact discretisation scheme, such as the step size. Indeed, as our experimental results in Figure~\ref{fig:sga_comp} show, adjusting the coefficient in SGA strongly affects training, and that cancelling the interaction terms outperforms SGA.

\textit{Connection with Consensus Optimisation (CO)}: \citet{mescheder2017numerics} analyse the discrete dynamics of gradient descent in zero-sum games and prove that, under certain assumptions, adding explicit regularisation that encourages the players to minimise the gradient norm of both players guarantees convergence to a local Nash equilibrium. Their approach includes cancelling the interaction terms, but also requires \textit{strengthening the self terms}, using losses:
\begin{align}
E_{\vphi} &= - E + c_1 \norm{\nabla_{\vtheta} E}^2 + s_1 \norm{\nabla_{\vphi} E}^2 \\
E_{\vtheta} &= E + s_2 \norm{\nabla_{\vtheta} E}^2 + c_2 \norm{\nabla_{\vphi} E}^2,
\label{eq:explicit_regularisation_self_terms}
\end{align}
where they use $s_1 = s_2 = c_1 = c_2 = \gamma$ where $\gamma$ is a hyperparameter. In order to understand the effect of the self and interaction terms, we compare to CO, as well as a similar approach where we use coefficients proportional to the drift, namely $s_1 = \lrp h/4$ and $s_2 = \lrt h/4$; this effectively doubles the strength of the self terms in DD. We show results in Figure~\ref{fig:co_comp}. We first notice that CO can improve results over vanilla SGD. However, similarly to what we observed with SGA, the regularisation coefficient is important and thus requires a hyperparameter sweep, unlike our approach, which uses the coefficients provided by the DD. We further notice that strengthening the norms using the DD coefficients can improve training, but performs worse compared to only cancelling the interaction terms. This shows the importance of finding the right training regime, and that strengthening the self terms does not always improve performance.

\textit{Alternating updates}: We perform the same exercise for alternating updates, where $c_1 = \lrp h/4$ and $c_2 = \lrt h / 4 ( 1 - \frac{2 \lrp}{\lrt}) $. We also study the performance obtained by only cancelling the discriminator interaction term, since when $\lrp/\lrt > 0.5$ the generator interaction term minimises, rather than maximises, the discriminator gradient norm and thus the generator interaction term might not have a strong destabilising force.
We observe that adding explicit regularisation only brings the benefit of reduced variance when cancelling the discriminator interaction term (Figure~\ref{fig:sgd_adam_sim_alt_updates}). As for simultaneous updates, we find that knowing the form of DD guides us to a choice of explicit regularisation: for alternating updates cancelling both interaction terms can hurt training, but the form of the modified losses suggests that we should only cancel the discriminator interaction term, with which we can obtain some gains.

\begin{figure}[t]
 \centering
  \begin{subfloat}{
  \includegraphics[width=0.49\columnwidth]{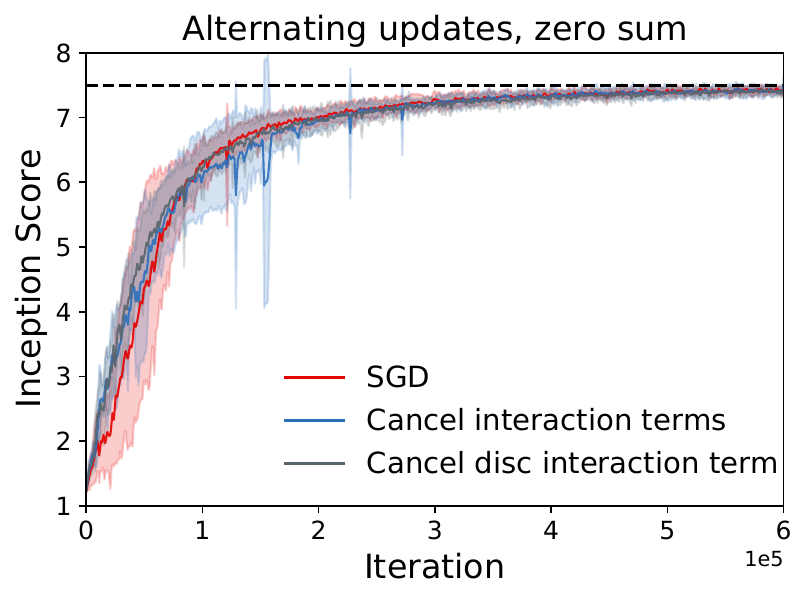}%
  \label{fig:alt_learning_curves}
} \end{subfloat}%
 \begin{subfloat}{
  \includegraphics[width=0.49\columnwidth]{
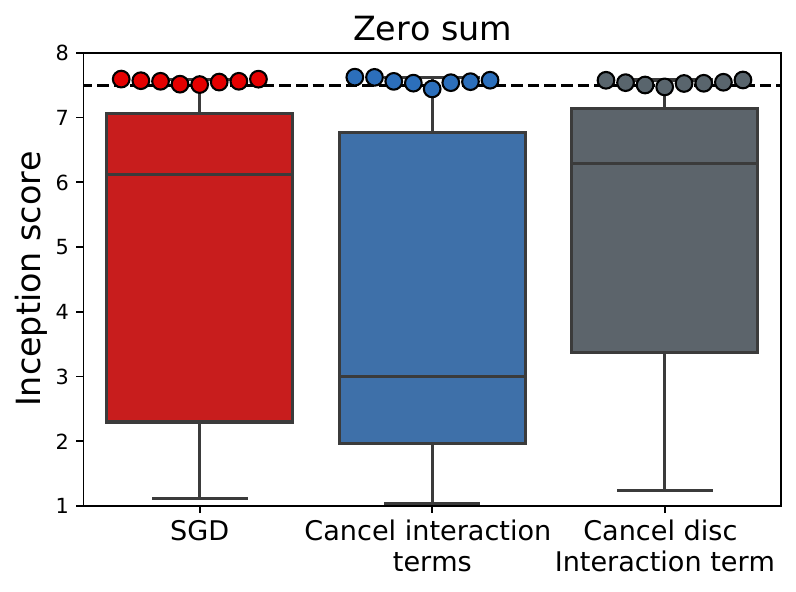}%
  \label{fig:alt_box_plot}
} \end{subfloat}%
  \caption[Explicit regularisation informed by discretisation drift for alternating updates in zero-sum games.]{Alternating updates in zero-sum games: the form of DD guides us in finding explicit regularisation terms. Since for alternating updates the generator's interaction term can be beneficial by minimising the discriminator's gradient norm, cancelling only the discriminator interaction term improves sensitivity across hyperparameters \subref{fig:alt_box_plot}. }
  \label{fig:sgd_adam_sim_alt_updates}
\end{figure}

\textit{Does cancelling the interaction terms help for every choice of learning rates?} The substantial performance and stability improvement we observe applies to the performance obtained across a learning rate sweep. For individual learning rate choices, cancelling the interaction terms is not guaranteed to improve learning.

\subsection{Extension to non-zero-sum GANs}
\label{sec:dd_non_zero_sum}

Finally, we extend our analysis to GANs with the non-saturating loss for the generator
\begin{align}
E_{\vtheta} = -\log D(G(\vz; \vtheta); \vphi)
\label{eq:non_sat_gen}
\end{align}
introduced by \citet{goodfellow2014generative}, while keeping the discriminator loss unchanged as that from the zero-sum formulation in Eq~\eqref{eq:original_gan}. The non-saturating loss has been observed to perform better than its zero-sum counterpart, and has often been used in practice.
\begin{figure}[t]
  \centering
    \begin{subfloat}[A sweep over a range learning rates.]{
\includegraphics[width=0.49\columnwidth]{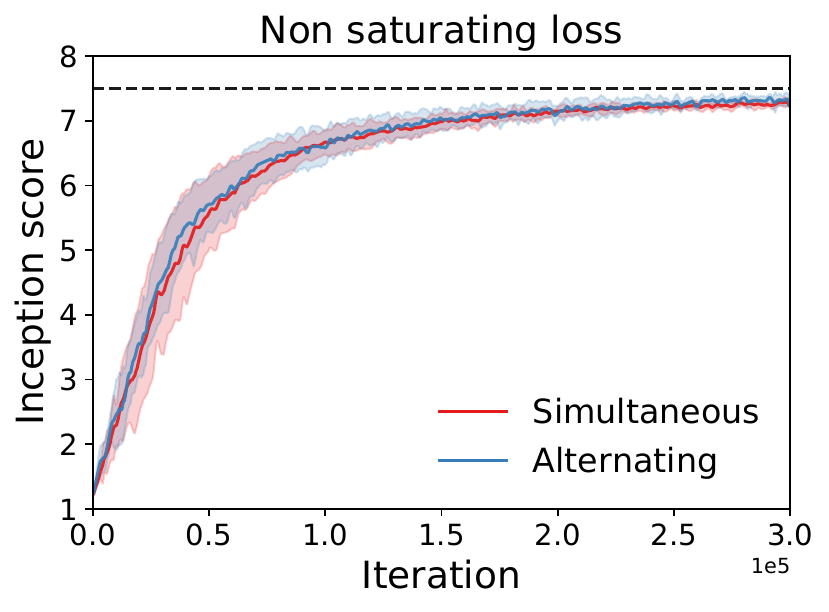}%
  }
  \end{subfloat}%
  \begin{subfloat}[Equal learning rates.]{
  \includegraphics[width=0.49\columnwidth]{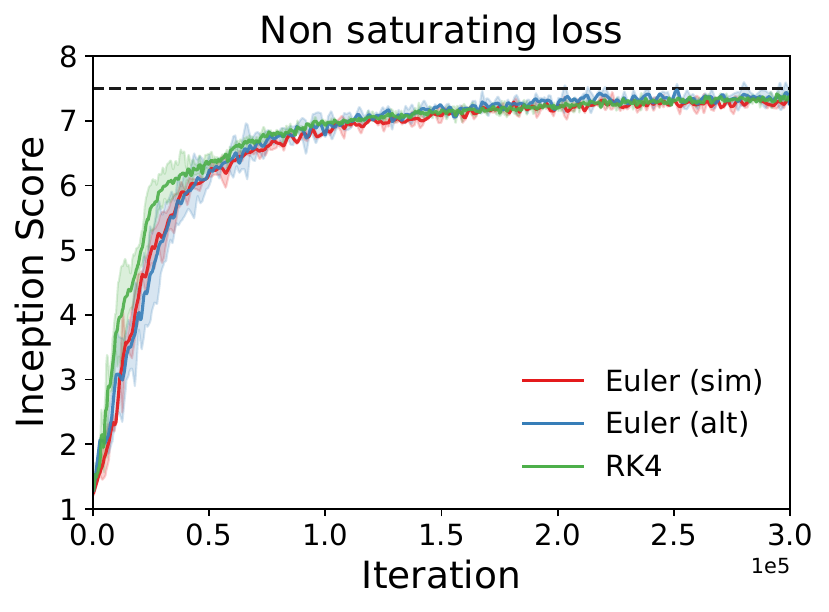}%
  \label{fig:equal_lr_rk_non_sat}
  }
  \end{subfloat}%
  \caption[The effect of discretisation drift depends on the game: its effect is less strong for the non-saturating loss.]{The effect of DD depends on the game: its effect is less strong for the non-saturating loss. Since when using the non-saturating loss GAN losses are no longer zero-sum, estimating the impact of the drift becomes more challenging, since we cannot easily write modified losses (though we provide novel insights into this problem in the next chapter, Chapter~\ref{ch:bea_stochasticity}). When using equal learning rates RK4 has $\mathcal{O}(h^5)$ drift, and thus we use it as a baseline to understand the effect of the drift in \subref{fig:equal_lr_rk_non_sat}.}
  \label{fig:non_saturating}
\end{figure}
In contrast with the dynamics from zero-sum GANs we analysed earlier, changing from simultaneous to alternating updates results in little change in performance, as can be seen in Figure~\ref{fig:non_saturating}. Despite having the same adversarial structure and the same discriminator loss, changing the generator loss changes the relative performance of the different discrete update schemes.

Since the effect of DD strongly depends on the game, we recommend analysing the performance of discrete numerical schemes on a case by case basis. Indeed, while for general two-player games we cannot always write modified losses as for the zero-sum case---see the Section~\ref{sec:dd_general_modified_losses} for a discussion---we can use Theorems~\ref{thm:sim} and~\ref{thm:alt} to understand the effect of the drift for specific choices of loss functions. We will tackle the question of modified losses for generally differentiable two-player games in the Chapter~\ref{ch:bea_stochasticity}, and further contrast the non-saturating and the saturating generator losses in GANs with Spectral Normalisation in Chapter~\ref{ch:rl}.

\section{A comment on different learning rates}
\label{sec:bea_diff_lr_games}

When considering different learning rates for the two players, we have thus far kept the consistency between physical time and learning rates. That is, we assumed the gradient descent updates are obtained by Euler discretisation with learning rates $\lrp h$ and $\lrt h$ of the flow
\begin{align}
 \dot{\vphi} &=  f( \vphi, \vtheta)    \label{eq:old_two_lr_ode1} \\
 \dot{\vtheta}  &= g( \vphi, \vtheta) \label{eq:old_two_lr_ode2}.
\end{align}
We then used BEA to fine modified flow such that after time $\lrp h$ and $\lrt h$ the error between the modified flow and the discrete updates is $\mathcal{O}(h^3)$. 

Alternatively, we can consider the gradient descent updates as the result of Euler discretisation with learning rate $h$ of the flow
\begin{align}
 \dot{\vphi} &=  \lrp f( \vphi, \vtheta)  \label{eq:new_two_lr_ode1}  \\
 \dot{\vtheta}  &= \lrt g( \vphi, \vtheta) .
 \label{eq:new_two_lr_ode2}
\end{align}
While in this approach the learning rate loses the connection with physical time for each player, it keeps the consistency of time between two players: after one iteration the same amount of time, $h$, has passed for both players. With this in mind, we will call this `the same-physical time between players' approach. 

Considering gradient descent as a discretisation of Eqs~\eqref{eq:new_two_lr_ode1} and~\eqref{eq:new_two_lr_ode2} instead of Eqs~\eqref{eq:old_two_lr_ode1} and~\eqref{eq:old_two_lr_ode2} has implications for the analysis of the behaviour of gradient descent using continuous-time tools such as stability analysis, as the two systems can lead to different results based on the two learning rates. Using Eqs~\eqref{eq:new_two_lr_ode1} and~\eqref{eq:new_two_lr_ode2} can also lead to a different perspective from the loss minimisation view. For the game with $f = - \nabla_{\vphi} E$ and $g = \nabla_{\vtheta} E$, the second system leads to the interpretation that the first player is minimising function $\lrp E(\vphi, \vtheta)$ while the second player is maximising function $\lrt E(\vphi, \vtheta)$. This is different than the zero-sum formulation of the first player minimising $E(\vphi, \vtheta)$ while the second player is maximising it, resulting from our previous interpretation.
The same-physical time between players approach also presents the following challenge:
\begin{remark}[The identifiability problem.] Given learning rates $\lrp h$ and $\lrt h$ for the two-players, $h$ cannot be identified exactly and has to be chosen by the user. The choice of $h$ leads to different flows describing the dynamics of the discrete updates. This is in contrast with the approach taken previously, which only depends on $\lrp$ and $\lrt$ through the learning rates $\lrp h$ and $\lrt h$.
\label{rem:id}
\end{remark}

Following the same-physical time between players approach also has implications for the use of BEA to find flows with error $\mathcal{O}(h^3)$ after one Euler update. Luckily, to find the modified flows using BEA under the same-physical time approach, we simply have to apply our results from Section~\ref{sec:general_theorems} for the system in Eqs~\eqref{eq:new_two_lr_ode1} and \eqref{eq:new_two_lr_ode2} by using their corresponding vector fields and learning rate $h$. We do so, and obtain:

\begin{theo}  The discrete \emph{simultaneous} Euler updates
in Eqs \eqref{eq:simup1} and \eqref{eq:simup2} follow the continuous system
\label{thm:sim_diff_lr}
\begin{align}
\dot{\vphi} &= \lrp f - \frac{h}{2} \left(\lrp^2 \jacphif f + \lrp \lrt \jacthetaf g \right) \\
\dot{\vtheta} &= \lrt g - \frac{h}{2} \left(\lrp \lrt \jacphig f + \lrt^2 \jacthetag g \right)
\end{align}
with an error of size $\mathcal O(h^3)$  after one update step.
\end{theo}

\begin{theo} The discrete \emph{alternating} Euler updates in \eqref{eq:altup1} and \eqref{eq:altup2} follow the continuous system
\label{thm:alt_diff_lr}
\begin{align}
\dot{\vphi} &= \lrp f - \frac {h}{2} \left(\frac{\lrp^2}{m}\jacphif f + \lrp \lrt \jacthetaf g \right) \\
\dot{\vtheta} &= \lrt g - \frac{h}{2} \left(-  \lrt \lrp \jacphig f + \frac{\lrt^2}{k} \jacthetag g \right)
\end{align}
with an error of size $\mathcal O(h^3)$ after one update step.
\end{theo}

We observe that both modified flows each player's vector field depends on the learning rate of the other player, unlike in our previous results where for simultaneous updates, the vector field for each player depended only on its learning rate.
When interpreting these results to obtain implicit regularisation effects as we have done previously in zero-sum and common-payoff games, the resulting coefficients of the implicit regularisers are different than those obtained in Section~\ref{section:zero_sum}. For zero-sum games, we obtain the following corollaries:
\begin{cor} In a zero-sum two-player differentiable game, \textit{simultaneous} gradient descent updates---as described in Eqs~\eqref{eq:simup1} and~\eqref{eq:simup2}---follows a gradient flow given by the modified losses
\label{cor:zs-sim-diff-lr}
\begin{align}
\tilde E_{\vphi}&= - \lrp E + \frac{\lrp^2 h}{4} \norm{\nabla_{\vphi} E}^2  -  \frac{\lrp \lrt h}{4} \norm{\nabla_{\vtheta} E}^2
\label{eq:min_min_simultaneous_zero_sum1-diff-lr} \\
\tilde E_{\vtheta}&= \lrt E -  \frac{\lrp \lrt h}{4} \norm{\nabla_{\vphi} E}^2 + \frac{\lrt^2 h}{4} \norm{\nabla_{\vtheta} E}^2 
\label{eq:min_min_simultaneous_zero_sum2-diff-lr}
\end{align}
with an error of size $\mathcal O(h^3)$ after one update step.
\end{cor}

\begin{cor} In a zero-sum two-player differentiable game, \textit{alternating} gradient descent---as described in Eqs~\eqref{eq:altup1} and~\eqref{eq:altup2}---follows a gradient flow given by the modified losses
\label{cor:zs-alt-diff-lr}
\begin{align}
\tilde E_{\vphi}&= -\lrp E + \frac{\lrp^2 h}{4m} \norm{\nabla_{\vphi} E}^2  - \frac{\lrp \lrt h}{4} \norm{\nabla_{\theta} E}^2
\label{eq:min_min_alt_zero_sum1_diff_lr}
\\
\tilde E_{\vtheta}&= \lrt E + \frac{\lrp \lrt h}{4} \norm{\nabla_{\vphi} E}^2  + \frac{\lrt^2 h}{4k}\norm{\nabla_{\theta} E}^2
\label{eq:min_min_alt_zero_sum2_diff_lr}
\end{align}
with an error of size $\mathcal O(h^3)$ after one update step.
\end{cor}

From Corollary~\ref{cor:zs-alt-diff-lr} one concludes for alternating updates that the second player's interaction \textit{always minimises the gradient norm of the first player}, which again helps explain the empirically observed stability of alternating updates over simultaneous updates; this is in contrast with our previous interpretation which suggested that this occurs only for certain learning rate ratios (Corollary~\ref{cor:zs-alt}). 
\begin{table*}[tb]
\centering
\begin{tabular}{ c |c|c|c| }
 \hline
 Simultaneous updates &   First player ($\vphi$) & Second player ($\vtheta$)\\
 \hline
 \hline
  Corollary~\ref{cor:zs-sim}  & $\frac{\lrp^2 h^2}{4} \nabla_{\vphi} \norm{\nabla_{\vtheta} E}^2$  & $\frac{\lrt^2 h^2}{4} \nabla_{\vtheta} \norm{\nabla_{\vphi} E}^2$ \\
      Corollary~\ref{cor:zs-sim-diff-lr}  & $\frac{\lrp \lrt h^2}{4} \nabla_{\vphi} \norm{\nabla_{\vtheta} E}^2$  & $\frac{\lrp \lrt h^2}{4} \nabla_{\vtheta} \norm{\nabla_{\vphi} E}^2$ \\
 \hline

 \hline
 Alternating updates &   First player ($\vphi$) & Second player ($\vtheta$)\\
 \hline
 \hline
   Corollary~\ref{cor:zs-alt}   & $\frac{\lrp^2 h^2}{4} \nabla_{\vphi} \norm{\nabla_{\vtheta} E}^2$  & $\left(\frac{\lrt^2 h}{4} - \frac{2 \lrp \lrt h^2}{4}\right) \nabla_{\vtheta} \norm{\nabla_{\vphi} E}^2$ \\
    Corollary~\ref{cor:zs-alt-diff-lr}   & $\frac{\lrp \lrt h^2}{4} \nabla_{\vphi} \norm{\nabla_{\vtheta} E}^2$  & $-\frac{\lrp \lrt h^2}{4} \nabla_{\vtheta} \norm{\nabla_{\vphi} E}^2$ \\
 \hline
\end{tabular}
\caption[Contrasting the implicit regularisation effects of the continuous-time flows proposed in zero-sum games.]{The strength of the interaction terms in DD under the different modified continuous-time flows we find for simultaneous and alternating gradient descent in zero-sum games. We obtained the different flows via different approaches of handling different learning rates for the two players, namely $\lrp h \ne \lrt h$. We do not write the self terms here as they are the same for both approaches. We write the implicit regularisation coefficients as observed in the continuous-time flow displacement for one gradient descent update, rather than the modified loss, in order to account for the different learning rates under the two interpretations (which are $\lrp h$ and $h$ for the first player, respectively). For simultaneous updates, the insight that the interaction terms for each player leads to a maximisation of the other player's gradient norm remains true regardless of interpretation, though the strength of the regularisation changes in the two interpretations. For alternating updates, the different interpretations suggest not only different regularisation strengths but also signs, with the result from Corollary~\ref{cor:zs-alt-diff-lr} showing that for alternating updates, the interaction term  of the second player always minimises the gradient norm of the first player, as opposed to the result in Corollary~\ref{cor:zs-alt}, for which this effect depends on the learning rate ratio $\lrp/\lrt$. Both results are consistent in explaining why alternating updates are more stable than simultaneous updates, since in both cases for alternating updates the second player's interaction term has a smaller coefficient when maximising the first player's gradient norm compared to that of simultaneous updates, or is minimising the first player's gradient norm.}
\label{tab:different_methods}
\end{table*}

If we would like to implement explicit regularisation methods to cancel the implicit regularisers obtained above using the same-physical time between players approach, we have to discretise the modified losses with learning rate $h$; this is opposed to using learning rates $\lrp h$ and $\lrt h$ to discretise the modified flows obtained in Section~\ref{section:zero_sum}. For the \textit{self terms, no difference is obtained compared to the previous interpretation}; under both approaches of tackling different learning rates, the strength of the self terms in the parameters update is proportional to $\lrp^2 h^2$ and $\lrt^2 h^2$, respectively. For interaction terms, however, the coefficients required in the parameter update are changed. We highlighted these changes in Table~\ref{tab:different_methods}; these coefficients also show that while in order to find the modified flow in the same-physical time between player approach we require to choose $\lrp$, $\lrt$ and $h$, for explicit regularisation \textit{we only need the learning rates of the two players}. 

We thus can investigate the effect of explicit regularisation to cancel the interaction terms under the same-physical time approach in zero-sum games. As before, we do so by using the generator saturating loss in GAN training. We show results in Figures~\ref{fig:new_int_lr_dd_sim} and~\ref{fig:new_int_lr_dd_alt} and compare with our previous results from Section~\ref{section:explicit_regularisation}. While for the simultaneous updates using the same physical time approach leads to more hyperparameters with increased performance, no significant difference is obtained for alternating updates.
\begin{figure}[tb!]
 \centering
  \begin{subfloat}{
  \includegraphics[width=0.46\columnwidth]{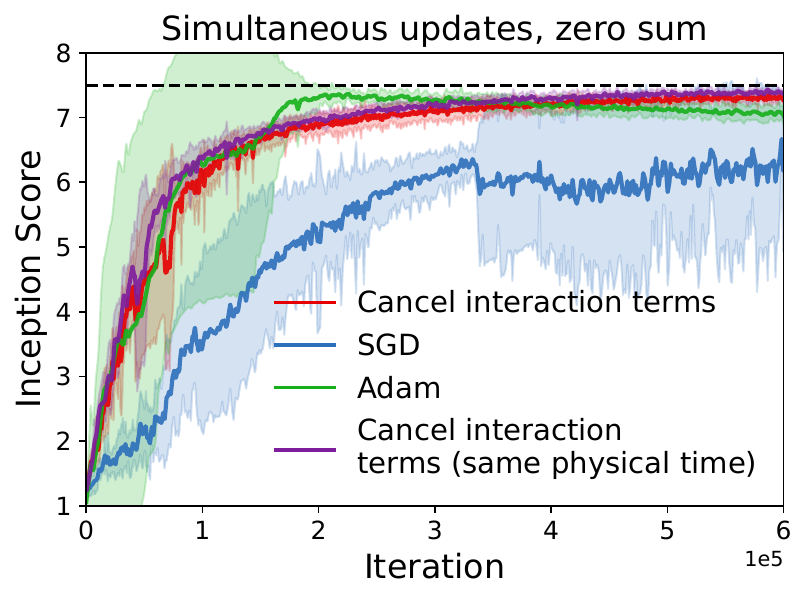}%
} \end{subfloat}%
 \begin{subfloat}{
  \includegraphics[width=0.54\columnwidth]{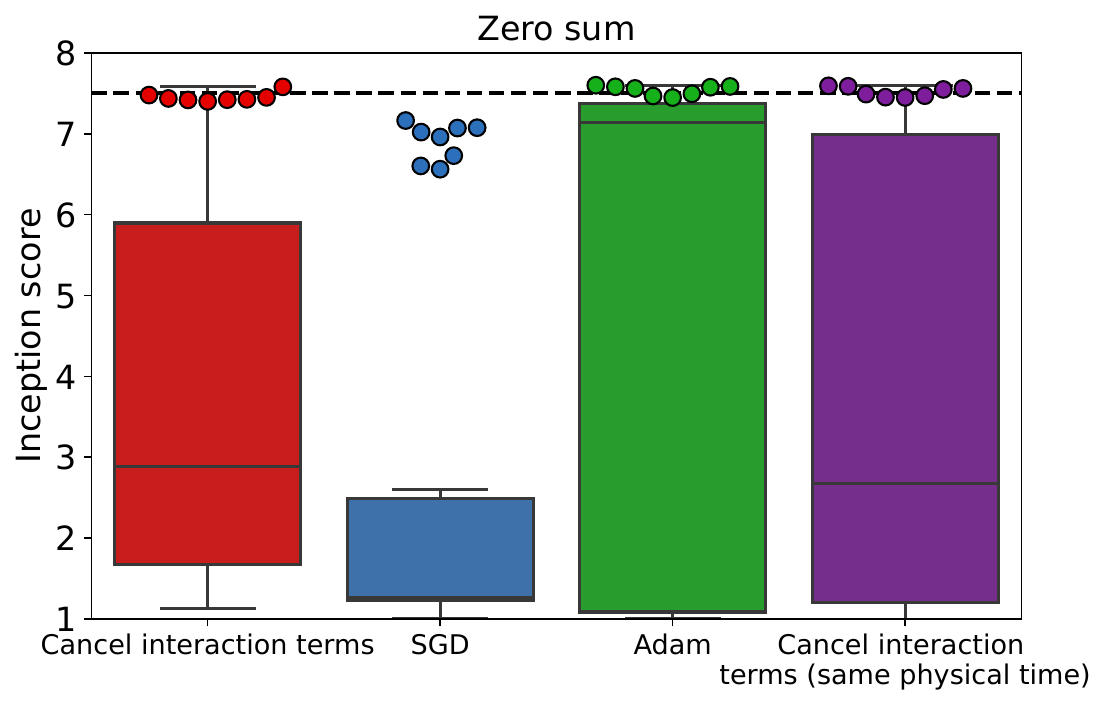}%
} \end{subfloat}%
  \caption[An alternative interpretation to different learning rates leads to a new set of flows for simultaneous gradient descent updates.]{Simultaneous updates in zero-sum games: cancelling the interaction terms under the different interpretations of tackling different learning rates (Corollaries~\ref{cor:zs-sim} and~\ref{cor:zs-sim-diff-lr}) does not result in a significant empirical difference. We denote the approach of Corollary~\ref{cor:zs-sim-diff-lr} as `same physical time', as per its motivation of using the same physical time between the two players.}
  \label{fig:new_int_lr_dd_sim}
\end{figure}
\begin{figure}[tb!]
 \centering
  \begin{subfloat}{
  \includegraphics[width=0.48\columnwidth]{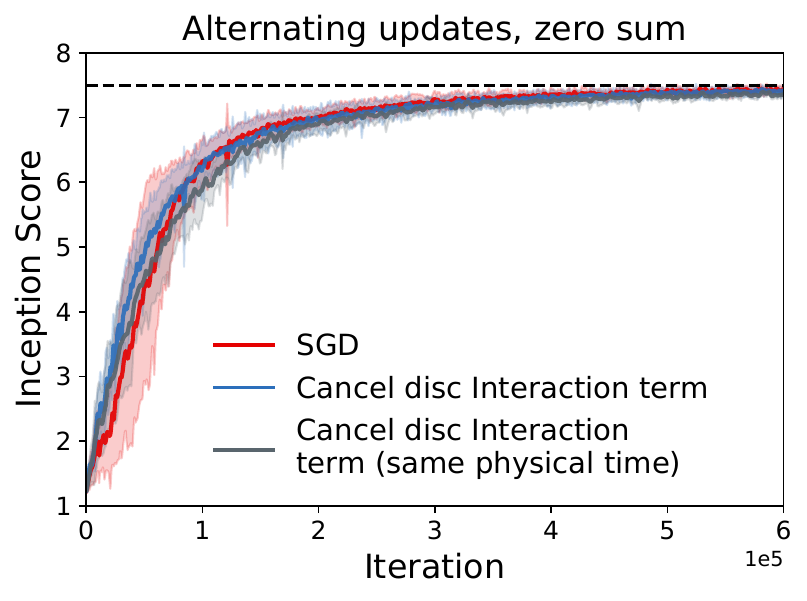}%
} \end{subfloat}%
 \begin{subfloat}{
  \includegraphics[width=0.52\columnwidth]{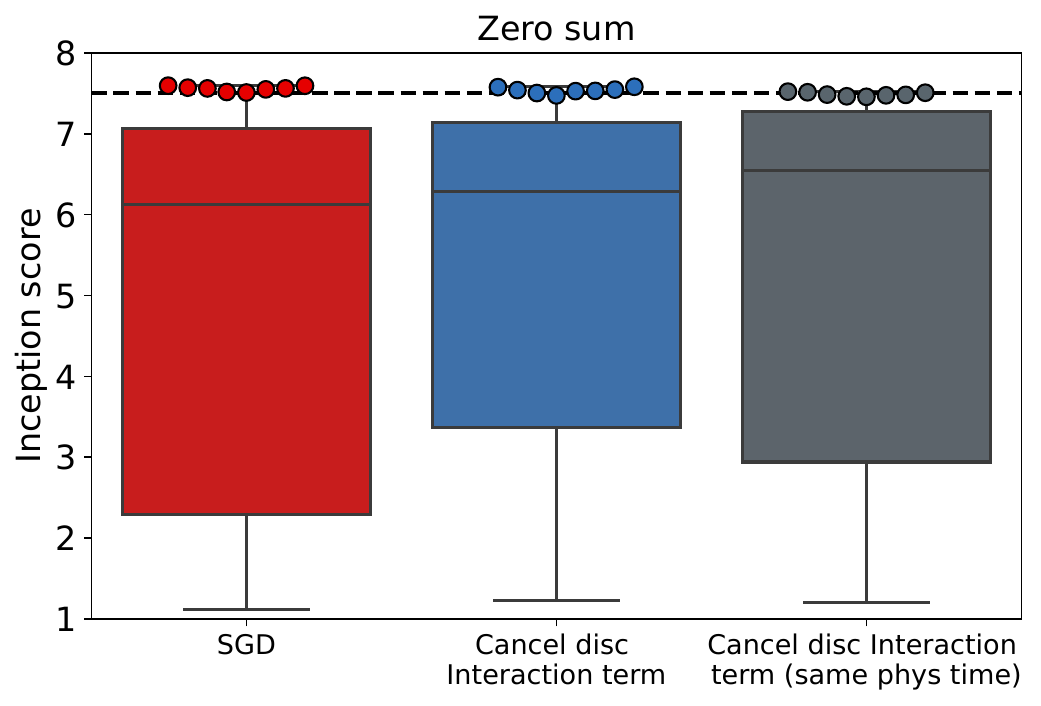}%
} \end{subfloat}%
  \caption[An alternative interpretation to different learning rates leads to a new set of flows for alternating gradient descent updates.]{Alternating updates in zero-sum games: cancelling the interaction terms under the different interpretations of tackling different learning rates (Corollaries~\ref{cor:zs-alt} and~\ref{cor:zs-alt-diff-lr}) does not result in a significant empirical difference. We denote the approach of Corollary~\ref{cor:zs-alt-diff-lr} as `same physical time', as per its motivation of using the same physical time between the two players.}
  \label{fig:new_int_lr_dd_alt}
\end{figure}

We conclude by noting that when studying two-player games trained with different learning rates, one can choose between two approaches of constructing corresponding continuous dynamics and modified losses. Both are valid, in that both lead to modified flows with the same error in learning rate, and can be used to understand implicit regularisation effects and the discrepancies between simultaneous and alternating Euler updates. Additional examination is needed to understand how these approaches explain continuous-time behaviour over a larger number of steps; we leave that as future work.

\section{Extending the PF for two-player games}
\label{sec:bea_pf_games}

In Chapter~\ref{ch:pf}, we introduced the principal flow (PF) as a model of single-objective gradient descent. We now briefly show how the same approach can be generalised to simultaneous Euler updates in two-player games, which will allow us to capture discretisation drift effects beyond those of order $\mathcal{O}(h^3)$ we explored in the rest of this chapter. For simplicity, we use assume equal learning rates for the two-players, though one can use the `same-physical time approach' from the previous section to adapt the results to different learning rates.

The approach we took in the single-objective setting had two steps: first, we developed the principal series with BEA (Theorem~\ref{thm:order_n_flow}) and second, we used the principal series given by BEA to find the PF based on the eigen-decomposition of the Hessian (Corollary~\ref{col:principal_series} and Theorem~\ref{thm:principal_ode}). The first step can be readily translated to games as it applies to any vector field, using  Theorem~\ref{thm:general_bea} in the Appendix, which Theorem~\ref{thm:order_n_flow} is a corollary of. For completeness, we reproduce Theorem~\ref{thm:general_bea} here:
\begin{theorem}
The modified flow given by BEA with an error of order $\mathcal{O}(h^{p+1})$ to the Euler update  ${\vpsi_t = \vpsi_{t-1} + h \gamesvf(\vpsi_{t-1})}$
has the form:
\begin{align}
  \dot{\vpsi} = \sum_{n=0}^{p} \frac{(-1)^{n}}{n+1} {(\jacparam{\vpsi}{\gamesvf})}^n \gamesvf + \csecondorderfamv
\end{align}
where $\csecondorderfamv$ denotes a class of functions defined as a sum of individual terms each containing higher than first order derivatives applied to $\gamesvf$.
\label{thm:general_bea_main_pf}
\end{theorem}

To use the above result in games, consider the two-player game with the original the dynamics in Eqs~\eqref{eq:ode1}-\eqref{eq:ode2}:
\begin{align}
  \dot{\left[ \begin{array}{c}
  \vphi \\
  \vtheta
\end{array}\right]}
   &=  \left[ \begin{array}{c}
  f \\
  g
\end{array}\right].
\end{align}
If we define
\begin{align}
\vpsi
   =  \left[ \begin{array}{c}
  \vphi \\
  \vtheta
 \end{array}\right] 
\hspace{3em}
u(\vpsi)
   =  \left[ \begin{array}{c}
  f(\vphi, \vtheta) \\
  g(\vphi, \vtheta)
\end{array}\right]
\end{align}
and the negative of the Jacobian as
\begin{align}
\vH(\vphi, \vtheta) = -
\begin{bmatrix}
    \jacparam{\vphi}{f} & \jacparam{\vtheta}{f} \\
   \jacparam{\vphi}{g} & \jacparam{\vtheta}{g}
\end{bmatrix},
\end{align}
we can replace this choice vector field and $\vH$ in Theorem~\ref{thm:general_bea_main_pf}, and obtain:
\begin{corollary}
The full series of BEA constructed from simultaneous Euler updates (Eqs~\ref{eq:simup1} and~\ref{eq:simup2}) with learning rate $h$ exactly is of the form
\begin{align}
  \dot{\left[ \begin{array}{c}
  \vphi \\
  \vtheta
\end{array}\right]}
   &=  \sum_{p=0}^{\infty} \frac{1}{p+1} h^p {\vH(\vphi, \vtheta)}^p \left[ \begin{array}{c}
  f \\
  g
\end{array}\right] + \csecondorderfam.
\end{align}
\end{corollary}
We can no longer use the eigen-decomposition of $\vH$, as unlike in the single-objective case ($f = - \nabla_{\vphi} E$ and $g = - \nabla_{\vtheta} E$)  we examined in Chapter~\ref{ch:pf}, $\vH(\vphi, \vtheta)$ need not be symmetric and thus we can no longer conclude Corollary~\ref{col:principal_series}. We instead consider the Jordan normal form of $\vH$
\begin{align}
\vH(\vphi, \vtheta) = \vP^{-1} \vJ \vP,
\end{align}
where $\vJ$ is a block diagonal matrix. We then have
\begin{align}
{\vH(\vphi, \vtheta)}^p = \vP^{-1} \vJ^p \vP,
\end{align}
leading to
\begin{align}
  \dot{\left[ \begin{array}{c}
  \vphi \\
  \vtheta
\end{array}\right]}
   &=  \sum_{p=0}^{\infty} \frac{-1}{p+1} \vP^{-1} (h\vJ)^p \vP \left[ \begin{array}{c}
  -f \\
  -g
\end{array}\right] + \csecondorderfam,
\end{align}
where we use the negative of the vector field in order to obtain consistency with the PF results in Chapter~\ref{ch:pf}, and recover single-objective results if $f = - \nabla_{\vphi} E$ and $g = - \nabla_{\vtheta} E$.
We are interested in finding a form for the series
\begin{align}
\sum_{p=0}^{\infty} \frac{-1}{p+1} (h \vJ)^p ,
\end{align}
which if converges can be written as another block diagonal matrix,
\begin{align}
\frac{1}{h} \log(\vI - h \vJ) \vJ^{-1}.
\end{align}
This result leads to the game equivalent of the PF:
\begin{definition} When expanding the PF to simultaneous Euler updates (Eqs~\ref{eq:simup1} and~\ref{eq:simup2}) with learning rate $h$ in two-player games, we obtain:
\begin{align}
\dot{\left[ \begin{array}{c}
  \vphi \\
  \vtheta
\end{array}\right]}
   &= \frac{1}{h}  \vP^{-1}  \log(\vI - h\vJ) \vJ^{-1} \vP \left[ \begin{array}{c}
  -f  \\
  -g
\end{array}\right] + \csecondorderfam, 
\label{eq:gen_pf_games}
\end{align}
where $\vH(\vphi, \vtheta) = \vP^{-1} \vJ \vP$ is the Jordan normal form of $\vH$.
\end{definition}
While the Jordan normal form of $\vH$ is not unique due to permutations of the order of the blocks in $\vJ$, $\vP^{-1}  \log(\vI - h\vJ) \vJ^{-1} \vP$ will be the same regardless of the block order.

\subsection{Stability analysis}

We now perform stability analysis on the modified flow obtained in Eq~\eqref{eq:gen_pf_games}, and hope it sheds light on the behaviour of Euler updates---and thus gradient descent---in games. We start by estimating the eigenvalues of the flow's Jacobian:
\begin{align}
\frac{1}{h} \vP^{-1}  \log(\vI - h \vJ) \vP.
\end{align}
As $\vP$ is invertible, the Jacobian's eigenvalues are the same as the eigenvalues of 
\begin{align}
\frac{1}{h}\log(\vI - h \vJ).
\end{align}
Since $\log(\vI - h \vJ)$ is block diagonal, its eigenvalues are the eigenvalues of its blocks $\frac{1}{h}\log(\vI_{n_k} - h \vJ_k)$, where $ \vI_{n_k}$ is the identity matrix of dimension $n_k$  and  $\vJ_k \in \mathbb{C}^{n_k, n_k}$.
Each block $\vJ_k$ corresponds to a unique eigenvalue $\lambda_k$ of $\vH$ with duplicity $n_k$. Given 
\begin{align}
\vZ_{n_k} = \begin{bmatrix}
0 & 1&  \hdots & 0 & 0 \\
\vdots & \vdots& \ddots & \ddots &  \vdots\\
0 & 0 & 0 & \hdots & 1 \\
0 & 0 & 0 & 0 & 0
\end{bmatrix},
\end{align}
 a matrix with all $0$s apart from the one off diagonal, $\vJ_{k}$ can be written as
\begin{align}
 \vJ_{k} = \lambda_k \vI_{n_k} + \vZ_{n_k}.
\end{align}
From here
\begin{align}
 \log (\vI - h \vJ_{k}) = \log(\vI - h\lambda_k \vI_{n_k} - h \vZ_{n_k}),
\end{align}
and since $\vI - h \vJ_{k}$ is upper triangular, its eigenvalues are its diagonal entries, which are $1 - h \lambda_k$ and the eigenvalues of $\log (\vI - h \vJ_{k})$ are $\log(1 - h \lambda_k)$. Thus, to perform stability analysis we have to assess whether $\Re{[\frac{1}{h}\log(1 - h \lambda_k)]} < 0$ for all $\lambda_k$. Since $h>0$, this is equivalent to $\Re{[\log(1 - h \lambda_k)]} < 0$ for all $\lambda_k$. Unlike the single-objective case, $\lambda_k$ might be complex.
If we write $\lambda_k = x_k + i y_k$ we have:
\begin{align}
\Re[\log(1 - h (x_k + i y_k))] = \log \sqrt{(1 - h x_k)^2 + h^2 y_k^2},
\end{align}
which is negative if $(1 - h x_k)^2 + (h y_k)^2 < 1$.

We now contrast this condition to the exponential stability convergence condition of the original continuous-time game dynamics. Under the original continuous-time dynamics, for a fixed point to be attractive, the real part of the $\vH$'s eigenvalues would need to be strictly positive since $\vH$ is the negative of the flow's Jacobian, i.e. $x_k > 0$. When accounting for DD, however, the importance of the imaginary part $y_k$ becomes apparent. 
Moreover, if we rewrite the condition  $(1 - h x_k)^2 + (h y_k)^2 < 1$ as a condition on the maximum learning rate to be convergent, we obtain 
\begin{align}
(1 - h x_k)^2 + (h y_k)^2 < 1 \implies h < \frac{1}{x_k} \frac{2}{1 + \left(\frac{y_k}{x_k}\right)^2},
\end{align}
which is the condition~\citet{mescheder2017numerics} (their Lemma 4) obtained from a discrete-time perspective. Thus in the game setting too, we recover existing fundamental results using a continuous-time perspective. We note that using this intuition~\citet{mescheder2017numerics} derive the algorithm Consensus Optimisation (CO), which we empirically assessed in Section~\ref{section:experimental}.

If $f = - \nabla_{\vphi} E$ and $g = - \nabla_{\vtheta} E$ we have a single objective, with $y_k = 0$ $\forall k$. In this case we recover the convergence conditions we are familiar with from the PF, which we derived in Section~\ref{sec:pf_stability_analysis}.

In the zero-sum case, where $f = - \nabla_{\vphi} E$ and $g = \nabla_{\vtheta} E$, we have seen in Section \ref{sec:intro_multi_objective_optimisation} that strict local Nash equilibria are local stable fixed points under the original game dynamics. That is, at a strict local Nash equilibrium we have that $x_k > 0, \forall k$. Under the continuous-time dynamics we derived in this section, which account for DD, this is no longer true, since $x_k$ being positive does not lead to $(1 - h x_k)^2 + (h y_k)^2 < 1$. We thus obtain an analogous result in zero-sum games to that of the single-objective case in Chapter~\ref{ch:pf}, where we saw that unlike the NGF, whether or not the PF is attracted to strict local minima depends on the learning rate $h$.

\subsection{An empirical example and different learning rates}

\begin{figure}[t]
\begin{subfloat}[Same learning rate.]{
\includegraphics[width=0.45\columnwidth]{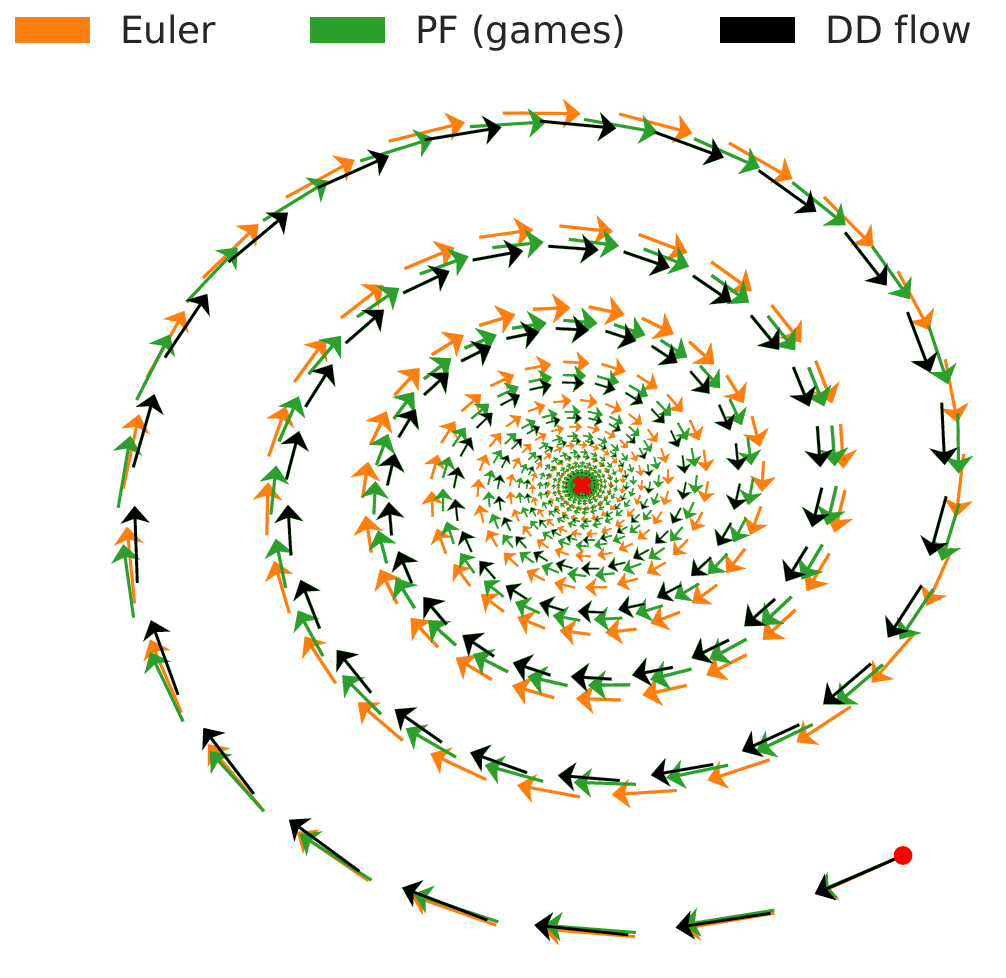}
\label{fig:pf_games_same_lr}
}
\end{subfloat}%
\hspace{2em}
\begin{subfloat}[Different learning rates.]{
\includegraphics[width=0.417\columnwidth]{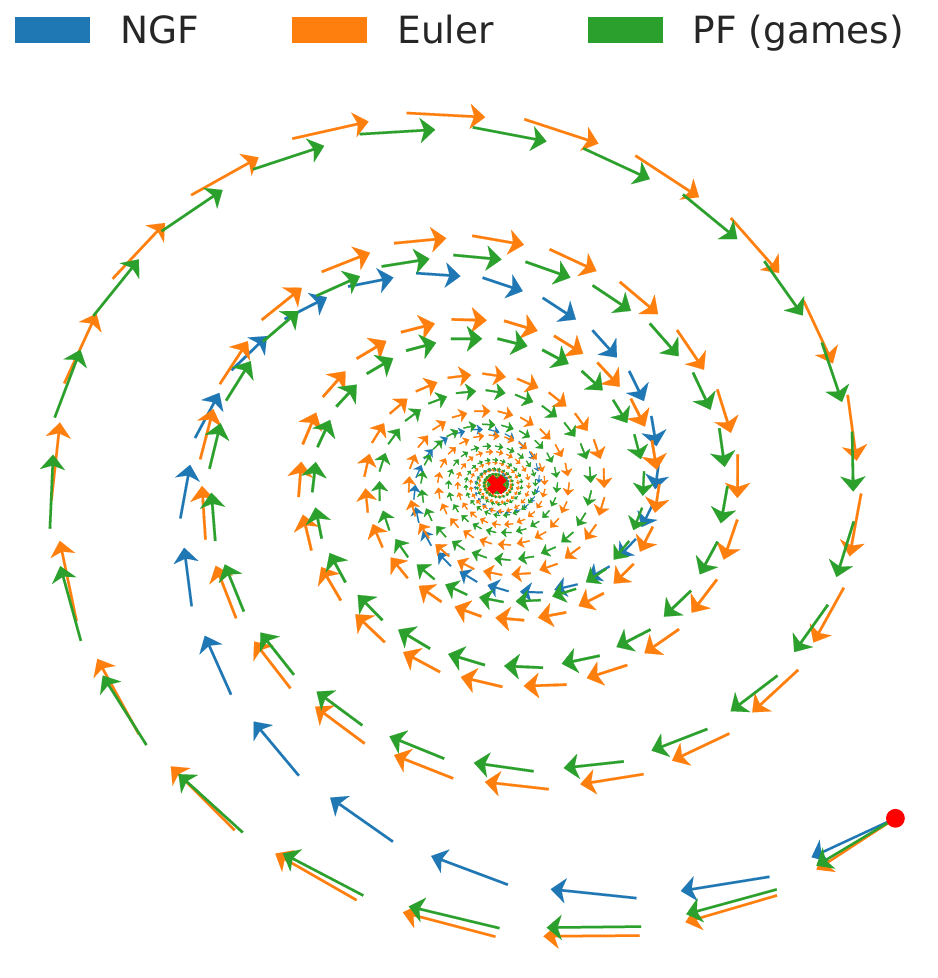}
\label{fig:pf_games_different_lr}
}
\end{subfloat}
\caption[A visualisation of a simple example of the extension of the PF to two-games.]{A visualisation of a simple example of the extension of the PF to two-games. Here we observe that the PF better tracks the gradient descent update compared to the flow obtained from using the $\mathcal{O}(h^3)$ error, and the NGF.}
\label{fig:pf_games}
\end{figure}

The game PF describes the Euler discretisation of the system in Eq~\eqref{eq:simple_odes} exactly (see Section~\ref{sec:pf_linearisation_games} for a full proof). We visualise this example in Figure~\ref{fig:pf_games_same_lr}, where we observe that the PF tracks the behaviour of the discrete updates better than the modified third-order correction previously derived.
We note that the Jordan normal form (here implemented using `sympy'~\citep{sympy}) is not always numerically stable and the observed (small) difference between the discrete updates and the PF is due to numerical issues.

If different learning rates are used for the two-players, we can use the approach outlined in Section~\ref{sec:bea_diff_lr_games} to construct vector fields that account for the each player's learning rate, and then apply the same analysis as above to the corresponding Jacobian to construct a flow and stability analysis conditions. We use this approach to visualise the same example from Eq~\eqref{eq:simple_odes} in Figure~\ref{fig:pf_games_different_lr}, and observe that here too the PF tracks the Euler updates closely.

\section{Related work}

{\bf Backward error analysis:} There have been a number of recent works applying BEA to machine learning in the one-player case, which can potentially be extended to the two-players settings. In particular, \citet{igr_sgd} extended the analysis of~\citet{igr} to stochastic gradient descent.
\citet{symmetry} used modified gradient flow equations to show that discrete updates break certain conservation laws present in the gradient flows of deep learning models. \citet{franca2020} compare momentum and Nesterov acceleration using their modified equations. \citet{dissipative} used BEA to help devise optimisers with a control on their stability and convergence rates. \citet{stochastic_adaptive_sgd} used modified equation techniques in the context of stochastic differential equations to devise optimisers with adaptive learning rates.
Other recent works such as \citet{grad_reg_avi} and \citet{fisher_explosion} noticed the strong impact of implicit gradient regularisation in the training of over-parametrised models with SGD using different approaches. These works provide evidence that DD has a significant impact in deep learning, and that BEA is a powerful tool to quantify it.
At last, note that a special case of Theorem~\ref{thm:2_players_sim} for zero-sum games with equal learning rates can be found in~\citet{lu2021resolution}.

\begin{table*}[t]
\centering
\begin{tabular}{ |c|c|c| }
 \hline
   & First player ($\vphi$) & Second player ($\vtheta$)\\
 \hline
 \hline
  \specialcell{Cancel DD \\ interaction terms (S)}  &  $\frac{\lrp h}{4}\norm{\nabla_{\vtheta} E}^2$  & $ \frac{\lrt h}{4}\norm{\nabla_{\vphi} E}^2$ \\ \hline
 \specialcell{Cancel DD \\ interaction terms (A)}  &  $\frac{\lrp h}{4}\norm{\nabla_{\vtheta} E}^2$  & $ -\frac{(2 \lrp - \lrt) h}{4}\norm{\nabla_{\vphi} E}^2$ \\ \hline
 SGA (S)   &$ \frac{1}{2} \norm{\nabla_{\vtheta} E}^2$ &  $ \frac{1}{2} \norm{\nabla_{\vphi} E}^2$ \\
 CO (S)  & $\zeta \norm{\nabla_{\vphi} E}^2 + \zeta \norm{\nabla_{\vtheta} E}^2$ & $\zeta \norm{\nabla_{\vphi} E}^2 + \zeta \norm{\nabla_{\vtheta} E}^2$ \\
Locally stable GAN (S)  & \text{\sffamily X} &  $\zeta \norm{\nabla_{\vphi} E}^2$ \\
 ODE-GAN (S) & $ \zeta \norm{\nabla_{\vtheta} E}^2$ &  \text{\sffamily X} \\
 \hline
\end{tabular}
\caption[Comparing cancelling discretisation drift terms with existing explicit regularisation methods in zero-sum games.]{Comparing cancelling discretisation drift terms with explicit regularisation methods in zero-sum games. SGA (without alignment, see the Appendix Section~\ref{app:sga} and \citealt{balduzzi2018mechanics}), Consensus Optimisation (CO)~\citep{mescheder2017numerics}, Locally Stable GAN~\citep{nagarajan2017gradient}, ODE-GAN~\citep{odegan}. We assume a $\min_{\vtheta} \min_{\vphi}$ game with learning rates $\lrp h$ and $\lrt h$  and number of updates $m$ and $k$, respectively. $\zeta$ denotes the regualrisation coefficient hyperparameter in methods which require one. S and A denote simultaneous and alternating updates. Unlike other approaches, the drift provides us with the coefficients to use to cancel the interaction terms and improve performance, and tackles both alternating and simultaneous updates.}
\label{tab:methods_comp}
\end{table*}
{\bf Two-player games:} As one of the best-known examples at the interface of game theory and deep learning, GANs have been powered by gradient-based optimisers as other deep neural networks. The idea of an implicit regularisation effect induced by simultaneous gradient descent in GAN training was first discussed in~\citet{schafer2019implicit}; the authors show empirically that the implicit regularisation induced by gradient descent can have a positive effect on GAN training, and take a game theoretic approach to strengthening it using competitive gradient descent~\citep{competitive_sgd}. By quantifying the DD induced by gradient descent using BEA, we shed additional light on the regularisation effect that discrete updates have on GAN training, and show it has both beneficial and detrimental components (as also shown in~\citet{daskalakis2018limit} with different methods in the case of simultaneous GD). Moreover, we show that the explicit regularisation inspired by DD results in drastically improved performance.
The form of the modified losses we have derived are related to explicit regularisation for GANs,
which has been one of the most efficient methods for stabilising GANs as well as other games. Some of the regularisers constrain the complexity of the players~\citep{miyato2018spectral,gulrajani2017improved,biggan}, while others modify the dynamics for better convergence properties~\citep{odegan,mescheder2018training,nagarajan2017gradient,balduzzi2018mechanics,wang2019solving,mazumdar2019finding}.
Our approach is orthogonal in that, we start from understanding the most basic gradient steps, underpinning any further modifications of the losses or gradient fields.
Importantly, we discovered a relationship between learning rates and the underlying regularisation.
Since merely cancelling the effect of DD is insufficient in practice (as also observed by \citet{odegan}), our approach complements regularisers that are explicitly designed to improve convergence.

We leave to future research to further study other regularisers in combination with our analysis. We summarise  these regularisers in Table~\ref{tab:methods_comp}, and contrast them to our results which cancel the interaction terms of DD, as suggested by Corollaries~\ref{cor:zs-sim} and~\ref{cor:zs-alt} respectively.

 To our knowledge, this is the first work towards finding continuous systems that better match the gradient descent updates used in two-player games. Studying discrete algorithms using continuous systems has a rich history in optimisation~\citep{su2016differential,wibisono2016variational}. Recently,
 models that directly parametrise differential equations demonstrated additional potential from bridging the discrete and continuous perspectives~\citep{chen2018neural,grathwohl2018ffjord}.

\section{Conclusion}
We have shown that using modified continuous systems to quantify the discretisation drift (DD) induced by gradient descent updates can help bridge the gap between the discrete optimisation dynamics and those of continuous methods. This allowed us to cast a new light on the stability and performance of games trained using gradient descent,
and guided us towards explicit regularisation strategies inspired by these modified systems.
We note however that DD merely modifies a game's original dynamics, and that the DD terms alone can not fully characterise a discrete scheme. In this sense, our approach complements works analysing the underlying game \citep{shoham2008multiagent}. 

We have focused our empirical analysis on a few classes of two-player games, but the effect of DD will be relevant for all games trained using gradient descent. Our method can be expanded beyond gradient descent to other optimisation methods such as Adam~\citep{kingma2014adam}, as well as to the stochastic setting as shown in~\citet{igr_sgd}.
We hope that our analysis provides %
a useful building block to further the understanding and performance of two-player games.

\chapter{Finding new implicit regularisers by revisiting backward error analysis}
\label{ch:bea_stochasticity}

In previous chapters, we used Backward Error Analysis (BEA) to construct flows that approximate a discrete optimiser update and shed light on optimisation dynamics, both for single-objectives and two-player games trained with gradient descent.  
The current usage of BEA is not without limitations, however, since not all the vector fields of continuous-time flows obtained using BEA can be written as a gradient, hindering the construction of modified losses revealing implicit regularisers.  Implicit regularisers uncover quantities minimised by following discrete updates and thus can be used to improve performance and stability across problem domains, from supervised learning to two-player games such as Generative Adversarial Networks~\citep{igr,igr_sgd,geiping2021stochastic,schafer2019implicit}; we have seen an example in common-payoff and zero-sum games in the previous chapter. In this chapter, we provide a novel approach to use BEA, and show how our approach can be used to construct continuous-time flows with vector fields that can be written as gradients. We then use this to find previously unknown implicit regularisation effects, such as those induced by multiple stochastic gradient descent steps while accounting for the exact data batches used in the updates, and in generally differentiable two-player games.

\section{Revisiting BEA}

In Section~\ref{sec:bea}, we described BEA and showed how \citet{igr} use BEA to find the IGR flow
\begin{align} 
\dot{\vtheta} = -\nabla_{\vtheta}E   -\frac{h}{2} \nabla_{\vtheta}^2 E \nabla_{\vtheta} E  = -\nabla_{\vtheta}\left(E(\vtheta) + \frac h4 \norm{\nabla_{\vtheta} E(\vtheta)}^2\right),
\end{align}
which has an error of $\mathcal{O}(h^3)$ after one gradient descent update ${\vtheta_t = \vtheta_{t-1} - h\nabla_{\vtheta} E(\vtheta)}$.
Gradient descent can thus be seen as implicitly minimising the modified loss $E(\vtheta) + \frac h4 \norm{\nabla_{\vtheta} E(\vtheta)}^2$. This showcases an implicit regularisation effect induced by the discretisation drift of gradient descent, dependent on learning rate $h$, which biases learning towards paths with low gradient norms.

We would like to go beyond full-batch gradient descent and model the implicit regularisation induced by the discretisation of one stochastic gradient descent (SGD). Given the SGD update $\vtheta_t = \vtheta_{t -1} - h \nabla_{\vtheta} E(\vtheta_{t -1}; \vX_{t})$---where we denoted $E(\vtheta_{t-1}; \vX) = \frac{1}{B}\sum_{i=1}^B E(\vtheta_{t-1}; \vx_i)$ as the average loss over batch $\vX$ and $B$ is the batch size---
we can use the IGR flow induced at this time step:
\begin{align}
  \dot{\vtheta} = - \nabla_{\vtheta} E(\vtheta; \vX_{t}) - \frac h 4 \nabla_{\vtheta} \norm{\nabla_{\vtheta} E(\vtheta; \vX_{t})}^2.
\label{eq:igr_stochastic}
\end{align}
Since the vector field in Eq~\eqref{eq:igr_stochastic} is the negative gradient of the modified loss ${E(\vtheta; \vX_{t}) +  \frac h 4 \norm{\nabla_{\vtheta} E(\vtheta; \vX_{t})}^2}$, this reveals a local implicit regularisation which minimises the gradient norm $ \norm{\nabla_{\vtheta} E(\vtheta; \vX_{t})}^2$. It is not immediately clear, however, how to combine the IGR flows obtained for each SGD update in order to model the \textit{combined} effects of multiple SGD updates, each using a different batch. What are, if any, the implicit regularisation effects induced by two SGD steps
 \begin{align}
&\vtheta_{t} = \vtheta_{t-1} - h \nabla_{\vtheta} E(\vtheta_{t-1}; \vX_{t}) \label{eq:sgd1} \\
&\vtheta_{t+1} = \vtheta_t - h \nabla_{\vtheta} E(\vtheta_t; \vX_{t+1})?
\label{eq:sgd2}
\end{align}
\citet{igr_sgd} find a modified flow in expectation over the shuffling of batches in an epoch, and use it to find implicit regularisation effects specific to SGD and study the effect of batch sizes in SGD. Since their approach works in expectation over an epoch, however, it does not account for the implicit regularisation effects of a smaller number of SGD steps, or account for the exact data batches used in the updates; we return to their results in the next section.
We take a different approach, and introduce a novel way to find implicit regularisers in SGD by revisiting the BEA proof structure and the  assumptions made thus far when using BEA.
Our approach can be summarised as follows:
\begin{remark}
Given the discrete update $\vtheta_t = \vtheta_{t-1} - h \nabla_{\vtheta}E (\vtheta_{t-1})$,
BEA constructs $\dot{\vtheta} = \nabla_{\vtheta} E + h f_1(\vtheta) + \cdots + h^p f_p(\vtheta)$, such that $\norm{\vtheta(h; \vtheta_{t-1}) - \vtheta_t} \in \mathcal{O}(h^{p+2})$ for a choice of $p \in \mathbb{N}, p \ge 1$. This translates into a constraint on the value of $\vtheta(h;\vtheta_{t-1})$.
Thus, BEA asserts only what the value of the correction terms $f_i$ in the vector field of the modified flow is at $\vtheta_{t-1}$. 
Given the constraints on $f_i(\vtheta_{t-1})$, we can \textit{choose}  $f_i: \mathbb{R}^D \rightarrow \mathbb{R}^D$ in the vector field of the modified flow $\dot{\vtheta}$ to depend on the initial condition $\vtheta_{t-1}$. 
\label{remark:bea_proofs}
\end{remark}

For example, in the proof by construction of the IGR flow (Section~\ref{sec:bea_proofs}), we obtained that if $\vtheta_t = \vtheta_{t-1} - h \nabla_{\vtheta}E (\vtheta_{t-1})$ and we want to find $\dot{\vtheta} = -\nabla_{\vtheta} E(\vtheta) + h f_1(\vtheta)$ such that $\norm{\vtheta(h; \vtheta_{t-1}) - \vtheta_t}$ is of order $\mathcal{O}(h^3)$, then $f_1({\color{red}{\vtheta_{t-1}}}) = - \frac{1}{2} \nabla_{\vtheta}^2 E({\color{red}{\vtheta_{t-1}}})\nabla_{\vtheta} E({\color{red}{\vtheta_{t-1}}})$ (Eq~\eqref{eq:igr_proof_value_iterate}). From there, following~\citet{igr}, we concluded that $f_1(\vtheta) = - \frac{1}{2} \nabla_{\vtheta}^2 E(\vtheta)\nabla_{\vtheta} E(\vtheta)$. But notice how BEA only \textit{sets a constraint on the value of the vector field at the initial point $\vtheta_{t-1}$}. If we allow the modified vector field to depend on the initial condition, an equally valid choice for $f_1(\vtheta) = - \frac{1}{2} \nabla_{\vtheta}^2 E(\vtheta)\nabla_{\vtheta} E({\color{red}{\vtheta_{t-1}}})$ or $f_1(\vtheta) =  -\frac{1}{2} \nabla_{\vtheta}^2 E({\color{red}{\vtheta_{t-1}}})\nabla_{\vtheta} E(\vtheta)$. \textit{By construction}, the above flows will also have an error of $\mathcal{O}(h^3)$ after one gradient descent step of learning rate $h$ with initial parameters $\vtheta_{t-1}$. The latter vector fields only describe the gradient descent update with initial parameters $\vtheta_{t-1}$ and thus they only apply to this \textit{specific gradient descent step}, though as previously noted that is also the case with the IGR flow due to the dependence on the data batch---see Eq~\eqref{eq:igr_stochastic}. The advantage of constructing modified vector fields that depend on initial parameters lies in the ability to write modified losses when a modified vector field depending only on $\vtheta$ cannot be written as a gradient operator, as we shall see in the next sections. 

This observation about BEA leads us to the following remarks:
\begin{remark}
 There are multiple flows that lead to the same order in learning rate error after one discrete update. Many of these flows depend on the \textit{initial conditions} of the system, i.e. the initial parameters of the discrete update. We visualise this approach in Figure~\ref{fig:idd_graphic_two_approaches} and contrast it with the existing BEA interpretation. 
\end{remark}

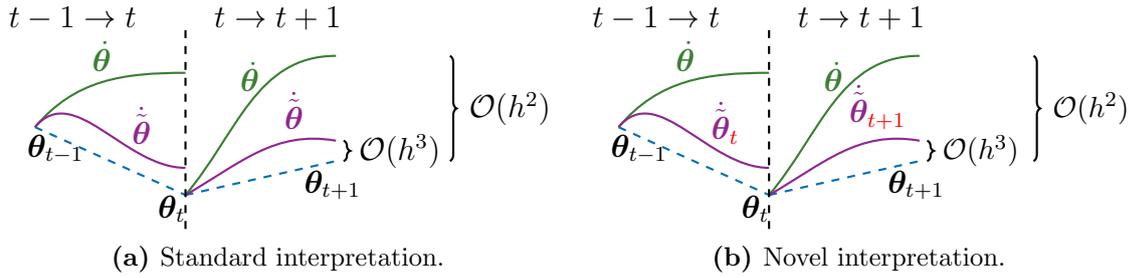
\begin{figure}[t]
\subfloat[Standard interpretation.]{
\begin{tikzpicture}[every text node part/.style={align=center,inner sep=0,outer sep=0}, scale=0.9][overlay]
\coordinate (theta_t_minus_1) at (0,0);
\coordinate (theta_t) at (2.2,-1);
\coordinate (theta_t_plus_1) at (4.4, -0.5);

\node(draw) at ($(theta_t_minus_1) + (+0.3,-0.32)$) {$\vtheta_{t-1}$};
\node(draw) at ($(theta_t) + (-0.2,-0.2)$) {$\vtheta_{t}$};
\node(draw) at ($(theta_t_plus_1) + (-0.05,-0.35)$) {$\vtheta_{t+1}$};

\coordinate (first_time_transition) at ($(theta_t_minus_1) + (+0.55,1.6)$);
\coordinate (second_time_transition) at ($(first_time_transition) + (3,0)$);

\node(draw) at (first_time_transition) {$t -1 \rightarrow t $};
\node(draw) at (second_time_transition) {$t \rightarrow t + 1$};

\coordinate (cont_theta_t) at (2.2,0.8);
\coordinate (cont_theta_t_plus_1) at (4.4, 1.05);

\coordinate (mod_cont_theta_t) at (2.2, -0.6);
\coordinate (mod_cont_theta_t_plus_1) at (4.4, -0.2);

\draw [NavyBlue,thick,dashed] (theta_t_minus_1) -- (theta_t);
\draw [NavyBlue,thick,dashed] (theta_t) -- (theta_t_plus_1);

\draw [OliveGreen,thick]  (theta_t_minus_1) to[out=50,in=180] node[midway,above] {$\dot{\vtheta}$} (cont_theta_t);
\draw [OliveGreen,thick]  (theta_t) to[out=50,in=180] node[midway,above]{$\dot{\vtheta}$} (cont_theta_t_plus_1);

\draw [Plum,thick]  (theta_t_minus_1) to[out=50,in=180]  node[near end,above] {$\dot{\tilde{\vtheta}}$} (mod_cont_theta_t);
\draw [Plum,thick]  (theta_t) to[out=30,in=170] node[near end,above] {$\dot{\tilde{\vtheta}}$} (mod_cont_theta_t_plus_1);

\draw [black,thick,dashed] ($(theta_t) + (0,-0.5)$) -- ($(cont_theta_t) + (0,0.7)$);

\draw [
    thick,
    decoration={
        brace,
        mirror,
        raise=0.1cm
    },
    decorate
] (theta_t_plus_1) -- (mod_cont_theta_t_plus_1)
node [pos=0.5,anchor=west,xshift=0.15cm] {$\mathcal{O}(h^3)$};

\draw [
    thick,
    decoration={
        brace,
        mirror,
        raise=1.5cm
    },
    decorate
] (theta_t_plus_1) -- (cont_theta_t_plus_1)
node [pos=0.5,anchor=west,xshift=1.6cm] {$\mathcal{O}(h^2)$};
\vspace{-2em}
\end{tikzpicture}
\label{tkz:orig_bea}
}
\subfloat[Novel interpretation.]{
\begin{tikzpicture}[every text node part/.style={align=center,inner sep=0,outer sep=0},scale=0.9][overlay]
\coordinate (theta_t_minus_1) at (0,0);
\coordinate (theta_t) at (2.2,-1);
\coordinate (theta_t_plus_1) at (4.4, -0.5);

\node(draw) at ($(theta_t_minus_1) + (+0.3,-0.32)$) {$\vtheta_{t-1}$};
\node(draw) at ($(theta_t) + (-0.2,-0.2)$) {$\vtheta_{t}$};
\node(draw) at ($(theta_t_plus_1) + (-0.05,-0.35)$) {$\vtheta_{t+1}$};

\coordinate (first_time_transition) at ($(theta_t_minus_1) + (+0.55,1.6)$);
\coordinate (second_time_transition) at ($(first_time_transition) + (3,0)$);

\node(draw) at (first_time_transition) {$t -1 \rightarrow t $};
\node(draw) at (second_time_transition) {$t \rightarrow t + 1$};

\coordinate (cont_theta_t) at (2.2,0.8);
\coordinate (cont_theta_t_plus_1) at (4.4, 1.05);

\coordinate (mod_cont_theta_t) at (2.2, -0.6);
\coordinate (mod_cont_theta_t_plus_1) at (4.4, -0.2);

\draw [NavyBlue,thick,dashed] (theta_t_minus_1) -- (theta_t);
\draw [NavyBlue,thick,dashed] (theta_t) -- (theta_t_plus_1);

\draw [OliveGreen,thick]  (theta_t_minus_1) to[out=50,in=180] node[midway,above] {$\dot{\vtheta}$} (cont_theta_t);
\draw [OliveGreen,thick]  (theta_t) to[out=50,in=180] node[midway,above]{$\dot{\vtheta}$} (cont_theta_t_plus_1);

\draw [Plum,thick]  (theta_t_minus_1) to[out=50,in=180]  node[near end,above] {$\dot{\tilde{\vtheta}}_{\color{red}{t}}$} (mod_cont_theta_t);
\draw [Plum,thick]  (theta_t) to[out=30,in=170] node[near end,above] {$\dot{\tilde{\vtheta}}_{{\color{red}{t + 1}}}$} (mod_cont_theta_t_plus_1);

\draw [black,thick,dashed] ($(theta_t) + (0,-0.5)$) -- ($(cont_theta_t) + (0,0.7)$);

\draw [
    thick,
    decoration={
        brace,
        mirror,
        raise=0.1cm
    },
    decorate
] (theta_t_plus_1) -- (mod_cont_theta_t_plus_1)
node [pos=0.5,anchor=west,xshift=0.15cm] {\small $\mathcal{O}(h^3)$};

\draw [
    thick,
    decoration={
        brace,
        mirror,
        raise=1.5cm
    },
    decorate
] (theta_t_plus_1) -- (cont_theta_t_plus_1)
node [pos=0.5,anchor=west,xshift=1.6cm] {\small $\mathcal{O}(h^2)$};
\vspace{-2em}
\end{tikzpicture}
\label{tkz:novel_bea}
}
   \caption[Visualising the standard approach to backward error analysis alongside an approach which constructs a different flow per gradient descent iteration.]{Visualising the standard approach to BEA alongside an approach which constructs a different flow per gradient descent iteration. In our previous use of BEA \subref{tkz:orig_bea}, we constructed modified flows $\dot{\tilde{\vtheta}}$ to capture the discretisation error of gradient descent; these flows did not depend on the initial iteration parameters. In this chapter, we take the second approach \subref{tkz:novel_bea}, allowing us to construct additional flows $\dot{\tilde{\vtheta}}_{\color{red}{t}}$, which depend on initial parameters, and showcase additional implicit regularisation effects.}
     \label{fig:idd_graphic_two_approaches}
\end{figure}

\begin{remark} Implicit in the choice of the IGR flow~\citep{igr}, as well as the flows we have introduced thus far in this work using BEA, namely the PF (Chapter~\ref{ch:pf}) and the modified flows of two-player games (Chapter~\ref{ch:dd_gans}), make an implicit assumption: that we are looking for the modified flows that hold for every training iteration and do not depend on the initial parameters. This is however challenged already in the case of stochastic gradient descent, where each implementation of the flows requires dependence on data batches, as shown in Eq~\eqref{eq:igr_stochastic}.
\end{remark}

\section{Implicit regularisation in multiple stochastic gradient descent steps}
\label{sec:modified_losses_sgd}

We now use the above observations to build modified flows that capture multiple gradient descent updates with an error of $\mathcal{O}(h^3)$ after  $n$ SGD steps. We are interested in modified flows 
that can be used to construct modified losses by writing the vector field of the flow as the negative gradient of a function; this enables us to capture the implicit regularisation effects of taking multiple SGD steps. 
We analyse $n$ SGD steps, starting at iteration $t$
\begin{align}
\vtheta_{t+\mu} = \vtheta_{t+\mu-1} - \nabla_{\vtheta} E(\vtheta_{t+\mu-1}; \vX^{t+\mu}), \hspace{3em} \mu \in \{0, \dots, n-1\}
\label{eq:sgd_multiple_updates_bea}
\end{align}
where for simplicity we denoted $E(\vtheta_{t-1}; \vX) = \frac{1}{B}\sum_{i=1}^B E(\vtheta_{t-1}; \vx_i)$, with elements $\vx_i$ forming the batch $\vX$.
We start with the following remark: 
\begin{lemma} 
Denote $E(\vtheta; \{\vX^{t}, \dots, \vX^{t+n-1} \}) = \frac{1}{n} \sum_{\mu=0}^{n-1} E(\vtheta; \vX^{t+\mu})$, i.e. the average loss obtained from the $n$ data batches. Then the trajectory obtained by taking $n$ steps of stochastic gradient descent follows the trajectory of minimising the loss in continuous-time with $\mathcal{O}(h^2)$ error:
\begin{align}
\tilde{E}(\vtheta) &= E(\vtheta; \{\vX^{t}, \dots, \vX^{t+n-1} \}).
\end{align}
\end{lemma}
Perhaps surprisingly, the above shows that effects from mini-batches appear only at higher-order terms in learning rate $h$. Thus, to capture implicit regularisation effects that account for the mini-batches used in the SGD updates, we use BEA. We find a modified flow that describes the SGD update with an error of  $\mathcal{O}(h^3)$,  with a vector field that can be written as a negative gradient. We provide proofs and the flows that construct the regularisers in the Appendix~\ref{ch:bea_stochasticity_supp}. 
We use the same steps used for our other BEA proofs: 1) we expand the $n$ discrete updates in Eq~\eqref{eq:sgd_multiple_updates_bea} to write a relation between the parameters at time $t-1$ and $t+n-1$ up to second order in learning rate; 2) expand the changes in continuous-time up to $\mathcal{O}(h^2)$; and 3) match the $\mathcal{O}(h^2)$ terms to find the terms which quantify the drift.
The significant difference with our previous approaches is that we allow the vector field to depend on the initial parameters, and choose a vector field that can be written as a gradient in order to construct a modified loss.

\begin{theorem} \label{thm:bea_sgd}
Denote $E(\vtheta; \{\vX^{t}, \dots, \vX^{t+n-1} \}) = \frac{1}{n} \sum_{\mu=0}^{n-1} E(\vtheta; \vX^{t+\mu})$, i.e. the average loss obtained from the $n$ data batches. Then the trajectory obtained by taking $n$ steps of stochastic gradient descent follows the trajectory of minimising the loss in continuous-time with $\mathcal{O}(h^3)$ error
\begin{align}
\tilde{E}(\vtheta) &= E(\vtheta; \{\vX^{t}, \dots, \vX^{t+n-1} \})
                  + \frac{n h}{4} \underbrace{\norm{ \nabla_{\vtheta}E(\vtheta; \{\vX^{t}, \dots, \vX^{t+n-1} \})}^2}_{\text{full batch norm regularisation}} \label{eq:sgd_norm_min}\\
                   & \quad - \frac{h}{n}\sum_{\mu =1 }^{n-1} \underbrace{\left[ \nabla_{\vtheta} E(\vtheta; \vX^{t+\mu})^T \left(\sum_{\tau = 0}^{\mu-1}    \nabla_{\vtheta}E({\color{red}{\vtheta_{t  -1}}}; \vX^{t+\tau})\right)\right]}_{\text{mini-batch gradient alignment}}.
\label{eq:modified_sup_learning}
\end{align}
\end{theorem}
We thus find the implicit regularisation effects induced by $n$ steps of SGD, capturing the \textit{importance of exact batches used and their order in Eq~\eqref{eq:modified_sup_learning}}. We note that without making use the observations regarding BEA in the previous section, and thus without the parameters $\vtheta_{t -1}$, a modified loss could not have been constructed outside of the full-batch case where one can recover the IGR flow.
The following remark immediately follows by setting $n=2$ in Eq~\eqref{eq:modified_sup_learning}:
\begin{remark} When taking a second stochastic gradient descent step, there is an implicit regularisation term maximising the dot product between the gradient at the current step and the gradient at the previous step: $\nabla_{\vtheta} E^T \nabla_{\vtheta} E(\vtheta_{t-1})$. This can be achieved by aligning the direction of the gradients between the two iterations or increasing the gradient norm, but we note that increasing gradient norm is counter other implicit regularisation effect in Eq~\eqref{eq:sgd_norm_min}.
\label{rem:2sgd_steps}
\end{remark}

We now compare this novel modified loss with the modified loss we obtained by ignoring stochasticity and assuming all $n$ updates have been done with a full-batch; this entails using the IGR loss (proof for multiple steps in Section~\ref{sec:igr_multiple_steps})
\begin{align}
\tilde{E} = E(\vtheta;\{\vX^{t}, \dots, \vX^{t+n-1} \}) + \frac{h}{4} \|\nabla_{\vtheta} E(\vtheta; \{\vX^{t}, \dots, \vX^{t+n-1} \}) \|^2 .
\label{eq:igr_flow_mul}
\end{align}
The above modified losses show that both full-batch gradient descent and SGD have a pressure to minimise the gradient norm  $\|\nabla_{\vtheta} E(\vtheta; \{\vX^{t}, \dots, \vX^{t+n-1} \})\|$. SGD leads to an additional regularisation effect capturing the importance of the order in which mini-batches are presented in training: maximising the dot product between gradients computed at the current parameters given a batch and the gradients computed at the initial parameters \textit{for all batches presented before the given batch}. While this can be achieved both by increasing the norm of the gradients or by aligning gradients with those at the initial iteration, we note that increasing the gradient norm is counter to the other regulariser induced by SGD, the gradient norm  minimisation effect shown in Eq~\eqref{eq:sgd_norm_min}.

The important role of learning rates in BEA appears here too: for our approximations to hold $nh$ has to be sufficiently small. If we adjust the learning rate by the number of updates, i.e. if we set the learning rate for SGD equal to $h/n$, where $h$ is the learning rate used by full-batch gradient descent, we obtain the same implicit regularisation coefficient  for the gradient norm $\|\nabla_{\vtheta} E(\vtheta;\{\vX^{t}, \dots, \vX^{t+n-1} \}) \|$ minimisation as the IGR flow in Eq~\ref{eq:igr_flow_mul}, and the main difference between the two modified losses is given by the mini-batch gradient alignment terms present in Eq~\eqref{eq:modified_sup_learning}. 

The number of updates, $n$, also plays an important role. While the mini-batch gradient regularisation term in Eq~\eqref{eq:modified_sup_learning} has a coefficient of $\frac{h}{n}$, there are $\frac{n(n-1)}{2}$ terms composing the term. Thus the magnitude of the mini-batch gradient regularisation term can grow with $n$, but its effects strongly depend on the distribution of the gradients computed at different batches. For example, if gradients $\nabla_{\vtheta} E(\vtheta_{t-1};\vX^{t +i})$ are normally distributed with the mean at the full-batch gradient, i.e. $\nabla_{\vtheta} E(\vtheta_{t-1};\vX^{t +i}) \sim \mathcal{N}(\nabla_{\vtheta} E(\vtheta_{t-1}), \sigma^2)$, as the number of updates grows the regularisation effect in Eq~\ref{eq:modified_sup_learning} will result in a pressure to align mini-batch gradients with the full-batch gradient at the initial parameters, $\nabla_{\vtheta} E(\vtheta_{t-1})$. Since our results hold for multiple values of $n$, empirical assessments need to be made to understand the interplay between the number of updates and the strength of  mini-batch gradient regularisation on training.

The approach provided here is complementary to that of that of~\citet{igr_sgd}, who obtain a relationship similar to Eq~\eqref{eq:modified_sup_learning} but in expectation---they describe expected value of the modified loss $\mathbb{E}_{\sigma} \left[E_{sgd}(\vtheta; \{\vX^{\sigma(t)}, \dots, \vX^{\sigma(t+n-1)} \}) \right]$ where the expectation is taken over all possible data batch shufflings $\sigma$, \textit{but not the elements in the batch}. 
As before we denote $E(\vtheta; \vX^{k})$ as the loss given by mini-batch $k$ (the equivalent to $\hat{C}_k$ in their notation, see their Eqs (1) and (2)), and write the modified loss they obtain in expectation as 
\begin{align}
\mathbb{E_\sigma} \left[E_{sgd}(\vtheta) \right] &= E(\vtheta;\{\vX^{0}, \dots, \vX^{n-1} \}) + \frac{h}{4n} \sum_{k=0}^{n-1} \norm{\nabla_{\vtheta} E(\vtheta;\vX^k)}^2 \\&= E(\vtheta; \{\vX^{0}, \dots, \vX^{n-1} \}) + \frac{h}{4}\|\nabla_{\vtheta} E(\vtheta; \{\vX^{0}, \dots, \vX^{n-1} \}) \|^2 \\ &\hspace{1em} + \frac{h}{4n} \sum_{k=0}^{n-1} \norm{\nabla_{\vtheta} E(\vtheta;\vX^k) - \nabla_{\vtheta} E(\vtheta; \{\vX^{0}, \dots, \vX^{n-1} \}}^2.
\end{align}
Our proposed approach does not require working in expectations and accounts for the exact batches sampled from the dataset; thus it directly describes SGD as used in practice. 
If we make the same assumptions as~\citet{igr_sgd} and take expectation over all possible batch shufflings in an epoch in Eq~\eqref{eq:modified_sup_learning} we obtain (proof in Section~\ref{sec:sgd_igr_comp}):
\begin{align}
\hspace{-0.7em}\mathbb{E}_{\sigma} \left[E_{sgd}(\vtheta) \right] 
                   &=  E(\vtheta; \{\vX^{t}, \dots, \vX^{t+n-1} \}) \\&\hspace{1em} + \frac{nh}{4}  \norm{\nabla_{\vtheta} E(\vtheta;\{\vX^{t}, \dots, \vX^{t+n-1} \})}^2 
                    \\&\hspace{1em}- \frac{h}{2n} \nabla_{\vtheta} E(\vtheta; \{\vX^{t}, \dots, \vX^{t+n-1} \})^T \nabla_{\vtheta} E({\color{red}\vtheta_{t-1}}; \{\vX^{t}, \dots, \vX^{t+n-1} \})
                   \\&\hspace{1em}  -  \frac{h}{2n}\left[\sum_{k=0}^{n-1}  \nabla_{\vtheta} E(\vtheta;\vX^{t+k}) ^T \nabla_{\vtheta} E({\color{red}\vtheta_{t-1}}; \vX^{t+k})\right].
\end{align}

As expected, we obtain a different result than that of~\citet{igr_sgd}: while the gradient norm minimisation is still present in our results, the per-batch regularisation in their formulation gets translated into a dot product term, where both the full-batch and per-batch gradients are regularised to be aligned with those at the beginning of the epoch.  Both approaches share the limitation that $nh$ needs to be suitably small for approximations to be relevant; we note that since we do not require $n$ to be the number of updates in an epoch---and we obtain interestingly regularisation effects for $n=2$, see Remark~\ref{rem:2sgd_steps}---this is less of an issue for our approach. Their approach has the advantage of finding an implicit regularisation effect that does not depend on the initial parameters. By depending on initial parameters, however, our approach does not require working in expectations and accounts for the exact batches used in the SGD updates. We hope this can be used to stabilise SGD over multiple steps, as we have done with a single GD step in Section~\ref{sec:stabilising_training}, that it can be used in continual and transfer learning~\citep{iman2022review,zhuang2020comprehensive,parisi2019continual}, as well as understanding the effects of the order of examples on model optimisation in online learning.

\section{Implicit regularisation in generally  differentiable two-player games}
\label{sec:dd_general_modified_losses}

In Chapter~\ref{ch:dd_gans}, we presented a framework for the quantification of discretisation drift in two-player games.  We found distinct modified flows that  describe \textit{simultaneous} Euler updates and \textit{alternating} Euler updates.
In both cases, we found corrections $f_1$ and $g_1$ to the original system
\begin{align}
 \dot{\vphi} &=  f( \vphi, \vtheta) \\
 \dot{\vtheta}  &= g( \vphi, \vtheta),
\end{align}
such that the modified continuous system
\begin{align}
 \dot{\vphi} &=  f( \vphi, \vtheta)  + h f_1( \vphi, \vtheta) \label{eq:general_bae_formulation1}\\
 \dot{\vtheta}  &= g( \vphi, \vtheta) + h g_1( \vphi, \vtheta), \label{eq:general_bae_formulation2}
\end{align}
follows the discrete steps of the method with a local error of order $\mathcal O(h^3)$.
More precisely, if $(\vphi_{t}, \vtheta_{t})$ denotes the discrete step of the method at time $t$ and $( \vphi(h),  \vtheta(h))$ corresponds to the continuous solution of the modified system above starting at $(\vphi_{t-1}, \vtheta_{t-1})$,  $\| \vphi_{t} -  \vphi( h) \| \textrm{ and }  \| \vtheta_{t} -  \vtheta( h) \|$
are of order $\mathcal O(h^3)$. In this section, we assume for simplicity that both players use the same learning rate $h$ and simultaneous updates, but the same arguments can be made when they use different learning rates or alternating updates.

Using this framework, we constructed \textit{modified loss functions} in the case of zero-sum (Section~\ref{section:zero_sum}) and common-payoff games (Section~\ref{sec:common_payoff}). However, using the aforementioned modified flows, \textit{we cannot always write the vector fields of the modified flows as a gradient for differentiable two-player games} and thus we cannot construct modified losses. To see why, consider the case where we have two loss functions for the two players respectively $E_{\vphi}(\vphi, \vtheta): \mathbb{R}^m\times \mathbb{R}^n \rightarrow \mathbb{R}$ and $E_{\vtheta}(\vphi, \vtheta): \mathbb{R}^m\times \mathbb{R}^n \rightarrow \mathbb{R}$.
This leads to the update functions $f = -\nabla_{\vphi} E_{\vphi}$ and $g = -\nabla_{\vtheta} E_{\vtheta}$.
By using $f = -\nabla_{\vphi} E_{\vphi}$ and $g = -\nabla_{\vtheta} E_{\vtheta}$ in Theorem~\ref{thm:sim}, we have that for simultaneous gradient descent 
\begin{align}
f_1 &= - \frac{1}{2}  \jacphif f- \frac{1}{2} \jacthetaf g\\
 &= \underbrace{- \frac{1}{4}  \nabla_{\vphi} \norm{ \nabla_{\vphi} E_{\vphi}}^2}_{\text{self term}} \underbrace{- \frac{1}{2} \jacparam{\vtheta}{\nabla_{\vphi} E_{\vphi}} \nabla_{\vtheta} E_{\vtheta}}_{\text{interaction term}}
\end{align}
and similarly
\begin{align}
g_1 &= \underbrace{- \frac{1}{4}  \nabla_{\vtheta} \norm{ \nabla_{\vtheta} E_{\vtheta}}^2}_{\text{self term}}  \underbrace{- \frac{1}{2}
\jacparam{\vphi}{\nabla_{\vtheta} E_{\vtheta}} \nabla_{\vphi} E_{\vphi}}_{\text{interaction term}}.
\end{align}
Thus, it is not always possible to write $f_1$ and $g_1$ as gradient functions, since the interaction terms---see Definition~\ref{def:self_interaction_terms} for the definition of self and interaction terms---cannot always be written as a gradient. Thus, the modified flows cannot be used to construct modified losses leading to implicit regularisers in generally differentiable two-player games, as we have done for zero-sum games in Chapter~\ref{ch:dd_gans}.

We now use the interpretation of BEA we provided in this chapter, to \textit{choose} two other functions $f_1$ and $g_1$ and we will use them to construct another set of modified flows. These flows will still satisfy the value constraints required by the BEA proofs and by construction after one gradient descent update the error between the modified flows and the discrete updates is $\mathcal{O}(h^3)$. Indeed, what the BEA proofs provide for simultaneous gradient descent is (see Eq~\eqref{eq:sim_geneal_eq} in the Appendix)
\begin{align}
f_1(\vphi_{t-1}, \vtheta_{t-1}) &= - \frac{1}{2}  \jacphif(\vphi_{t-1}, \vtheta_{t-1}) f(\vphi_{t-1}, \vtheta_{t-1}) - \frac{1}{2}  \jacthetaf(\vphi_{t-1}, \vtheta_{t-1}) g(\vphi_{t-1}, \vtheta_{t-1}) \\
g_1(\vphi_{t-1}, \vtheta_{t-1}) &= - \frac{1}{2} \jacphig(\vphi_{t-1}, \vtheta_{t-1}) f(\vphi_{t-1}, \vtheta_{t-1}) - \frac{1}{2}  \jacthetag(\vphi_{t-1}, \vtheta_{t-1}) g(\vphi_{t-1}, \vtheta_{t-1}).
\end{align}
From here, we can choose $f_1$ and $g_1$, now depending on the iteration number $t$:
\begin{align}
f_{1, t}(\vphi, \vtheta) &= - \frac{1}{2}  \jacphif(\vphi, \vtheta) f(\vphi, \vtheta) - \frac{1}{2}  \jacthetaf(\vphi, \vtheta) g({\color{red}\vphi_{t-1}, \vtheta_{t-1}}) \\
g_{1, t}(\vphi, \vtheta) &= - \frac{1}{2}  \jacphig(\vphi, \vtheta) f({\color{red}\vphi_{t-1}, \vtheta_{t-1}}) - \frac{1}{2}  \jacthetag(\vphi, \vtheta) g(\vphi, \vtheta).
\end{align}
 Here, we treat $g(\vphi_{t-1}, \vtheta_{t-1})$ and $f(\vphi_{t-1}, \vtheta_{t-1})$ as constants. We replace $f = - \nabla_{\vphi} E_{\vphi}$ and $g = - \nabla_{\vtheta} E_{\vtheta}$  in $f_{1, t}$ and $g_{1, t}$ and write the drift terms as gradient functions:
\begin{align}
f_{1, t}(\vphi, \vtheta) &= - \frac{1}{2} \jacparam{\vphi}{\nabla_{\vphi} E_{\vphi}} \nabla_{\vphi} E_{\vphi}  - \frac{1}{2} \jacparam{\vtheta}{\nabla_{\vphi} E_{\vphi}} \nabla_{\vtheta} E_{\vtheta}(\vphi_{t-1}, \vtheta_{t-1}) \\
&= - \frac{1}{2} \nabla_{\vphi} \norm{\nabla_{\vphi}E_{\vphi}}^2 - \frac{1}{2} \nabla_{\vphi}\left(\nabla_{\vtheta}  E_{\vphi} ^T \nabla_{\vtheta} E_{\vtheta}(\vphi_{t-1}, \vtheta_{t-1}) \right) \\
&= - \nabla_{\vphi} \left(\frac{1}{2} \norm{\nabla_{\vphi}E_{\vphi}}^2 + \frac{1}{2} \nabla_{\vtheta}  E_{\vphi}^T  \nabla_{\vtheta} E_{\vtheta}(\vphi_{t-1}, \vtheta_{t-1}) \right) \\
g_{1, t}(\vphi, \vtheta) &= - \nabla_{\vtheta} \left(\frac{1}{2} \norm{\nabla_{\vtheta}E_{\vtheta}}^2 + \frac{1}{2} \nabla_{\vphi}  E_{\vtheta}^T \nabla_{\vphi} E_{\vphi}(\vphi_{t-1}, \vtheta_{t-1}) \right).
\end{align}
Replacing the above in the modified flows in Eqs~\eqref{eq:general_bae_formulation1} and~\eqref{eq:general_bae_formulation2}), we obtain
\begin{align}
 \dot{\vphi} &=  f( \vphi, \vtheta)  + h f_1( \vphi, \vtheta) \\
              &= - \nabla_{\vphi} \left(E_{\vphi} +  h \left(\frac{1}{2} \norm{\nabla_{\vphi}E_{\vphi}}^2 + \frac{1}{2} \nabla_{\vtheta}  E_{\vphi}^T \nabla_{\vtheta} E_{\vtheta}(\vphi_{t-1}, \vtheta_{t-1}) \right)   \right) \label{eq:disc_int_app} \\
 \dot{\vtheta}  &= g( \vphi, \vtheta) + h g_1( \vphi, \vtheta) \\
                &= - \nabla_{\vtheta} \left(E_{\vtheta} +  h \left(\frac{1}{2} \norm{\nabla_{\vtheta}E_{\vtheta}}^2 + \frac{1}{2} \nabla_{\vphi}  E_{\vtheta} ^T \nabla_{\vphi} E_{\vphi}(\vphi_{t-1}, \vtheta_{t-1}) \right)\right).
\end{align}

 We can now write modified losses for each gradient descent iteration of a general two-player differentiable game (which depend on the iteration $t$), which describe the local trajectory of simultaneous gradient descent up to $\mathcal{O}(h^3)$:
\begin{align}
\tilde{E}_{\vphi, t} = E_{\vphi} +  h \left(\underbrace{\frac{1}{2} \norm{\nabla_{\vphi}E_{\vphi}}^2}_{\text{self term}} + \underbrace{\frac{1}{2} \nabla_{\vtheta}  E_{\vphi}^T \nabla_{\vtheta} E_{\vtheta}({\color{red}\vphi_{t-1}, \vtheta_{t-1}}) }_{\text{interaction term}} \right) \label{eq:int_term_d_first} \\
\tilde{E}_{\vtheta, t} = E_{\vtheta} +  h \left(\underbrace{\frac{1}{2} \norm{\nabla_{\vtheta}E_{\vtheta}}^2}_{\text{self term}} + \underbrace{\frac{1}{2}  \nabla_{\vphi}  E_{\vtheta}^T \nabla_{\vphi} E_{\vphi}({\color{red}\vphi_{t-1}, \vtheta_{t-1}}}_{\text{interaction term}})\right).
\end{align}

 The \textit{self terms} result in the implicit gradient regularisation found in supervised learning~\citep{igr} and  zero-sum games (Chapter~\ref{ch:dd_gans}): each player has an incentive to minimise its own gradient norm. The \textit{interaction term} for each player encourages minimising the dot product between the gradients of its loss with respect to the other player's parameters and the previous gradient update of the other player.
Consider the first player's interaction term, equal to $(-\nabla_{\vtheta} E_{\vphi}(\vphi_t, \cdot))^T (- \nabla_{\vtheta}E_{\vtheta}({\color{red}\vphi_{t-1}, \vtheta_{t-1}}))$, where $-E_{\vtheta}({\color{red}\vphi_{t-1}, \vtheta_{t-1}})$ is the previous update direction of $\vtheta$ aimed at minimising $E_{\vtheta}$. 
Its implicit regularisation effect depends on the functional form of $E_{\vphi}$ and $E_{\vtheta}$: if $\nabla_{\vtheta} E_{\vtheta}$ and $\nabla_{\vtheta} E_{\vphi}$ have aligned directions by construction, then the implicit regularisation effect nudges the first player's update towards a point in space where the second player's update changes direction; the opposite is true if $\nabla_{\vtheta} E_{\vtheta}$ and $\nabla_{\vtheta} E_{\vphi}$ are misaligned by construction.

\subsection{The effect of the interaction terms: a GAN example}
 We work through the regularisation effect of interaction terms for a pair of commonly used GAN losses that do not form a zero-sum game (using the generator non-saturating loss \citep{goodfellow2014explaining}) and contrast it with the zero-sum case (the saturating loss); we have previously investigated these losses in Chapter~\ref{ch:dd_gans}. As before, we denote the first player, the discriminator, as $D$, parametetrised by $\vphi$, and the generator as $G$, parametrised by $\vtheta$. We denote the data distribution as $p^*(\vx)$ and the latent distribution $p(\vz)$.
Given the non-saturating GAN loss \citet{goodfellow2014generative}:
\begin{align}
    E_{\vphi}(\vphi, \vtheta) &= \mathbb{E}_{p^*(\vx)} \log D(\vx; \vphi) + \mathbb{E}_{p(\vz)} \log (1 - D(G(\vz; \vtheta); \vphi)) \\
     E_{\vtheta}(\vphi, \vtheta) &=  \mathbb{E}_{p(\vz)} - \log D(G(\vz; \vtheta); \vphi),
\end{align}
we can then write the interaction term $\nabla_{\vtheta}  E_{\vphi}^T \nabla_{\vtheta} \nabla_{\vtheta} E_{\vtheta}({\color{red}\vphi_{t-1}, \vtheta_{t-1}})$ in Eq~\eqref{eq:int_term_d_first} at iteration $t$  as (derivation in Appendix~\ref{app:non_sat_gan})
\begin{align}
    \frac{1}{B^2}\sum_{i,j=1}^B &c^{non-sat}_{i,j}  \nabla_{\vtheta}  D(G(\vz^i_t; \vtheta); \vphi)^T   \nabla_{\vtheta}  D(G(\vz^j_{t-1}; {\color{red}\vtheta_{t-1}}); {\color{red}\vphi_{t-1})}), \hspace{2em} \text{with} \\
    &c^{non-sat}_{i,j} = \frac{1}{1 - D(G(\vz^i_t; \vtheta); \vphi)} \frac{1}{D(G(\vz^j_{t-1}; {\color{red}\vtheta_{t-1}}); {\color{red}\vphi_{t-1})}} \label{eq:c_i_j_non_sat},
\end{align}
where $\vz^i_t$ is the latent variable with index $i$ in the batch at time $t$ with batch size $B$.
Thus, the strength of the regularisation---$c^{non-sat}_{i,j}$---depends on how \textit{confident} the discriminator is. In particular, this implicit regularisation encourages the discriminator update into a new set of parameters where the gradient $\nabla_{\vtheta}  D(G(\vz^i_t; \vtheta); \vphi)$ \textit{points away} from the direction of  $\nabla_{\vtheta}  D(G(\vz^j_{t-1}; \vtheta_{t-1}); \vphi_{t-1})$ when $c_{i,j}$ is large. This occurs for $\vz^i_t$ where the discriminator is fooled by the generator---i.e. $1 - D(G(\vz_i; \vtheta); \vphi)$ is close to 0 and  $\frac{1}{1 - D(G(\vz^i_t; \vtheta); \vphi)}$ is large---and samples $\vz^j_{t-1}$ where the discriminator was correct at the previous iteration---$D(G(\vz^j_{t-1}; \vtheta_{t-1}); \vphi_{t-1})$ low and thus $\frac{1}{D(G(\vz^j_{t-1}; \vtheta_{t-1}); \vphi_{t-1})}$ is large. This can be seen as beneficial regularisation for the generator, by ensuring the update direction $- {\nabla_{\vtheta} E_{\vtheta} = \mathbb{E}_{p(\vz)} \frac{1}{D(G(\vz; \vtheta); \vphi)} \nabla_{\vtheta}  D(G(\vz; \vtheta); \vphi)}$ is adjusted accordingly to the discriminator's output. We note, however, that regularisation might not have a strong  effect, as gradients $\nabla_{\vtheta}  D(G(\vz; \vtheta); \vphi)$ that have a high weight in the generator's update are those where the discriminator is correct and classifies generated data as fake---i.e. $\frac{1}{D(G(\vz; \vtheta); \vphi)}$ is large---but there the factor $\frac{1}{1 - D(G(\vz; \vtheta); \vphi)}$ in the interaction term coefficient in Eq~\eqref{eq:c_i_j_non_sat} will be low. This is inline with our empirical results in Section~\ref{sec:dd_non_zero_sum}, where we saw that for the non-saturating loss discretisation drift does not have a strong effect on performance.

We can contrast this with what regularisation we obtain when using the saturating loss \citep{goodfellow2014generative}, where $E_{\vtheta}(\vphi, \vtheta) = - \mathbb{E}_{p(\vz)} \log (1 - D(G(\vz; \vtheta); \vphi))$. We used the saturating loss extensively in experiments in Section~\ref{section:experimental}, and we have shown it performs poorly in comparison to the non-saturating loss, in-line with the literature~\citep{goodfellow2014generative}. If we perform the same analysis as above for the saturating loss we obtain the following implicit regulariser
\begin{align}
   \frac{1}{B^2} \sum_{i,j=1}^B &c^{sat}_{i,j} \nabla_{\vtheta}  D(G(\vz^i_t; \vtheta); \vphi)^T   \nabla_{\vtheta}  D(G(\vz^j_{t-1}; {\color{red}\vtheta_{t-1}}); {\color{red}\vphi_{t-1})}),  \hspace{3em} \text{with}\\
    &c^{sat}_{i,j} = \frac{1}{1 - D(G(\vz^i_t; \vtheta); \vphi)} \frac{1}{1-D(G(\vz^j_{t-1}; {\color{red}\vtheta_{t-1}}); {\color{red}\vphi_{t-1})})} \label{eq:c_i_j_sat}.
\end{align}
Here, $c^{sat}_{i,j}$ is high for $\vz^i_t$ and $\vz^j_{t-1}$ where the generator was fooling the discriminator---i.e. low $1-D(G(\vz^j_{t-1}; \vtheta_{t-1}); \vphi_{t-1})$ and $1 - D(G(\vz^i_t; \vtheta); \vphi)$. Thus, instead of moving $\nabla_{\vtheta}  D(G(\vz^i_t; \vtheta); \vphi)$ away from directions where the generator was doing poorly previously as is the case for the non-saturating loss, \textit{it is moving it away from directions where the generator was performing well}, which could lead to instabilities or loss of performance. Moreover, unlike for the non-saturating loss, the implicit regularisation will have a strong effect on the generator's update $- {\nabla_{\vtheta} E_{\vtheta} = \mathbb{E}_{p(\vz)} \frac{1}{1-D(G(\vz; \vtheta); \vphi)} \nabla_{\vtheta}  D(G(\vz; \vtheta); \vphi)}$, since gradients $\nabla_{\vtheta}  D(G(\vz; \vtheta); \vphi)$ that have a high weight in the generator's update are those where the interaction term coefficient in Eq~\eqref{eq:c_i_j_sat} is high. This is inline with our results in Section~\ref{section:experimental}, showing that in zero-sum games (such as the one induced by the saturating loss), the regularisation induced by the discretisation error of gradient descent is strong, and can hurt performance and stability.

\section{Conclusion}
We provided a novel approach of interpreting backward error analysis and used it to find implicit regularisers induced by gradient descent in the single objective setting and in two-player games.
In the single objective case, we found implicit regularisation terms revealing importance of the alignment of gradients at the exact data batches used in multiple steps of stochastic gradient descent, while in two-player games we highlighted the need to examine the game structure in order to determine the effects of implicit regularisation.
We hope future work can empirically verify the effects of these implicit regularisers in deep learning.
We believe that our observations in this chapter might unlock future research, including verifying whether the dot product alignment regularisation we found explains the benefits of using stochastic gradient descent in deep learning, and what the empirical effect of the interaction terms is on generally differentiable two-player games.

\chapter{\smoothnesstitle}
\label{ch:smoothness}

In previous chapters, we used continuous-time approaches to model gradient descent optimisation dynamics and empirically assessed the effect of these dynamics on neural network training. 
We have seen how improvements to optimisation lead to increased stability and better performance across problem domains, from supervised learning to two-player games such as GANs.
Optimisation also interacts with the choice of model through neural architectures and model regularisation: some neural architectures are easier to optimise than others~\citep{resnets,liu2020understanding}, and model regularisation techniques can improve training stability~\citep{understanding_batch_norm,miyato2018spectral,vahdat2020nvae}. 
For the rest of this thesis, we explore interactions between optimisation and models, and in particular model smoothness and smoothness regularisers. While smoothness quantifies sensitivity to changes in model \textit{inputs} and optimisation is concerned with changes to model \textit{parameters}, we show that in neural network training smoothness regularisation can improve optimisation and that implicit regularisation effects induced by optimisation---such as those discussed in previous chapters---can increase model smoothness.

\section{Why model smoothness?}

How certain should a classifier be when it is presented with out of distribution data?
How much mass should a generative model assign around a datapoint?
How much should an agent's behaviour change when its environment changes slightly?
Answering these questions shows the need to quantify the manner in which the output of a
model varies with changes in its input, a quantity we will
intuitively call the smoothness of the model.
In deep learning,
neural network smoothness has been shown to boost generalisation and robustness of classifiers and
increase the stability and performance of generative models, as well as provide better priors for representation learning and reinforcement learning agents~\citep{bartlett2017spectrally,cisse2017parseval,novak2018sensitivity,sokolic2017robust,lassance2018laplacian,vahdat2020nvae,miyato2018spectral,d2020learn,on_mi_for_rep_learning}. After providing an overview of the benefits of smoothness in deep learning and motivated by its impact across problem domains, 
 later in the chapter we investigate some of the previously unexplored effects of smoothness regularisation: strong interactions with optimisation, reduced model capacity, and interactions with data scaling.

\section{Measuring smoothness}
\label{sec:measuring_smoothness}

Neural network ``smoothness'' is a broad, vague, catch all term.
 We use it to convey formal definitions, such as differentiable, bounded, Lipschitz,
 as well as intuitive concepts such as invariant to data dimensions or projections, robust to input perturbations, and others.
One definition states that a function $f: \mathcal{X} \rightarrow \mathcal{Y}$ is $n$ smooth if is $n$ times differentiable with the $n$-th derivative being continuous.
  The differentiability of a function
 is not a very useful inductive bias for a model, as it is both very local and constructed according to the metric of the space where limits are taken.
 What we are looking for is the ability to choose both the distance metric and how local or global our smoothness inductive biases are.
 With this in mind,
 Lipschitz continuity is appealing as it defines a \textit{global} property and provides the choice of distances in the domain and co-domain of $f$. It is defined as:
\begin{align}
  \norm{f(\vx_1) - f(\vx_2)}_{\mathcal{Y}} \le K \norm{\vx_1-\vx_2}_{\mathcal{X}}  \hspace{3em} \forall \vx_1, \vx_2 \in \mathcal{X},
  \label{eq:lip_def}
\end{align}
where $K$ is denoted as the Lipschitz constant of function $f$.
To avoid learning trivially smooth functions and maintain useful variability, it is often beneficial to constrain the function variation both from above and below. This leads to bi-Lipschitz continuity:
\begin{align}
  K_1 \norm{\vx_1-\vx_2}_{\mathcal{X}} \le \norm{f(\vx_1) - f(\vx_2)}_{\mathcal{Y}} \le K_2 \norm{\vx_1-\vx_2}_{\mathcal{X}}.
\end{align}

Enforcing Eq~\eqref{eq:lip_def} can be difficult, but according to Rademacher's theorem~\citep{federer2014geometric} if $\mathcal{X} \subset \mathbb{R}^I$ is an open set and $\mathcal{Y} = \mathbb{R}^O$ and $f$ is $K$-Lipschitz then $\norm{\frac{d f(\vx)}{d \vx}}_{op} \le K$ wherever the total derivative $D_{\vx}f(\vt) = \frac{d f(\vx)}{d \vx} \vt$ exists, which is almost everywhere. 
Conversely, a function $f$ that is differentiable everywhere with $\norm{\frac{d f(\vx)}{d \vx}}_{op} \le K$ is $K$-Lipschitz. 
Since the operator norm is not always easily computable, at times the Frobenius norm of the Jacobian $\frac{d f(\vx)}{d \vx}$ is constrained instead, since the Frobenius norm bounds the operator norm from above: $\norm{\frac{d f(\vx)}{d \vx}}_{op} \le \norm{\frac{d f(\vx)}{d \vx}}_{2}$; for $O=1$, $\norm{\nabla_\vx f(\vx)}_{op} = \norm{\nabla_\vx f(\vx)}_2$. Thus, a convenient strategy to make a differentiable function $K$-Lipschitz is to ensure $\norm{\frac{d f(\vx)}{d \vx}}_{2} \le K, \forall \vx \in \mathcal{X}$.

If $f$ and $g$ are Lipschitz with constants $K_f$ and $K_g$, $f \circ g$ is Lipschitz with constant $K_f K_g$. Since commonly used activation functions are 1-Lipschitz, a neural network can be made Lipschitz by constraining each learneable layer to be Lipschitz. Many neural networks layers are linear operators (linear and convolutional layers, BatchNormalisation~\citep{batch_norm}), and to compute their Lipschitz constant we can use that the Lipschitz constant of a linear operator $A$ under common norms such as $l_1, l_2, l_{\infty}$ is $\sup_{\vx \ne 0} \frac{\norm{A\vx}}{\norm{\vx}}$~\citep{petersen2008matrix}. For the $l_2$ norm, most commonly used in practice, this leads to finding the spectral norm of $A$.

\section{Smoothness regularisation in deep learning}
\label{sec:smoothness_techniques}

Smoothness regularisers have long been part of the toolkit of the deep learning practitioner:
early stopping encourages smoothness by stopping optimisation before the model overfits the training data;
 dropout~\citep{dropout} makes the network more robust to small changes in the input by randomly masking hidden activations; max pooling encourages smoothness with respect to local changes;
 $L_2$ weight regularisation and weight decay~\citep{loshchilov2017decoupled} discourage large changes in output by not allowing individual weight norms to grow;
 data augmentation allows us to specify what changes in the input should not result in large changes in the model prediction and thus is also closely related to smoothness and invariance to input transformations.
  These smoothness regularisation techniques are often introduced as methods that directly target generalisation and other beneficial effects of smoothness discussed in Section~\ref{sec:benefits}, instead of being seen through the lens of smoothness regularisation.

Methods that explicitly target smoothness on the entire input space focus on restricting the learned model family.
A common approach is to ensure Lipschitz smoothness with respect to the $l_2$ metric by individually restricting the Lipschitz constant of each layer; for layers that are linear operators, this entails restricting their spectral norm.
Spectral regularisation~\citep{yoshida2017spectral} uses the sum of the spectral norms---the largest singular value---of each layer as a regularisation loss to encourage Lipschitz smoothness. 
 Spectral Normalisation~\citep{miyato2018spectral} ensures the learned models are 1-Lipschitz by adding a node in the computational graph of the model layers by replacing the weights with their normalised version: 
\begin{align}
 \vW \rightarrow \vW/||\vW||_{op},
\end{align}
where
 $||\vW||_{op}$ is the spectral norm of $\vW$. Both Spectral Normalisation and Spectral Regularisation use power iteration~\citep{mises1929praktische} to compute the spectral norm of weight matrices $||\vW||_{op}$ for layers which are linear operators, such as convolutional or linear layers. ~\citet{gouk2018regularisation} use a projection method by dividing the weights by the spectral norm after a gradient update. This is unlike Spectral Normalisation, which backpropgates through the normalisation operation.
The majority of this line of work has focused on constraints for linear and convolutional layers, and only recently attempts to expand to other layers, such as self attention, have been made~\citep{lipschitz_constant_self_attention}.
Efficiency is always a concern and heuristics are often used even for popular layers such as convolutional layers~\citep{miyato2018spectral} despite more accurate algorithms being available~\citep{gouk2018regularisation,tsuzuku2018lipschitz}.
Parseval networks~\citep{cisse2017parseval} ensure weight matrices of linear and convolutional layers are 1-Lipschitz by enforcing a stronger constraint, an extension of orthogonality to non-square matrices.~\citet{bartlett2018representing} show that
any bi-Lipschitz function can be written as a composition of residual layers~\citep{resnets}.

Instead of restricting the learned function on the entire space, another approach of targeting smoothness constraints is to regularise the norm of the gradients of a predictor $f: \mathbb{R}^I \rightarrow \mathbb{R}$ with respect to inputs of the network $\norm{\nabla_{\vx}f(\vx;\vtheta)}$, at different \textit{regions of the space}~\citep{sokolic2017robust,gulrajani2017improved,fedus2017many,arbel2018gradient,kodali2017convergence}.
 This is often enforced by adding a gradient penalty to the loss function $\mathcal{L(\vtheta)}$:
\begin{align}
  \mathcal{L(\vtheta)} + \zeta \mathbb{E}_{p_{reg(\vx)}} \left(\norm{\nabla_{\vx}f(\vx;\vtheta)}^2 - K^2\right)^2,
\label{eq:grad_penalties}
\end{align}
where $\zeta$ is a regularisation coefficient,
$p_{reg(\vx)}$ is the distribution at which the regularisation is applied,
 which can either be the data distribution~\citep{sokolic2017robust,arbel2018gradient} or around it~\citep{kodali2017convergence,fedus2017many},
 or, in the case of generative models, at linear interpolations between data and model samples~\citep{gulrajani2017improved}.
 Gradient penalties encourage the function to be smooth around the support of $p_{reg(\vx)}$ either by encouraging
 Lipschitz continuity ($K \ne 0$) or by discouraging drastic changes of the function as the input changes ($K=0$).

Smoothness for classification tasks is defined by \citet{lassance2018laplacian} as preserving features similarities within the same class as we advance through the layers of the network.
The penalty used is $\sum_{l=1}^{L} \sum_{c=1}^{C}|\sigma^l(s_c) - \sigma^{l+1}(s_c)|$, where $\sigma^l(s_c)$ is the signal of features belonging to class $c$ computed using the Laplacian of layer $l$.
The Laplacian of a layer is defined by constructing a weighted symmetric adjacency matrix of the graph induced by the pairwise most similar layer features in the dataset.

\section{The benefits of learning smooth models}
\label{sec:benefits}

\textbf{Generalisation}. Learning models that generalise beyond training data is the goal of machine learning.
The study of generalisation has long been connected with the study of model complexity, as models with small complexity generalise better~\citep{vapnik2013nature}.
Despite this, we have seen that deep, over-parametrised neural networks tend to generalise better than their shallow counterparts~\citep{novak2018sensitivity} and that for Bayesian methods, Occam's razor does not apply to the number of parameters used, but to the complexity of the function~\citep{rasmussen2001occam}.
A way to reconcile these claims is to incorporate smoothness into definitions of model complexity and to show that smooth, over-parametrised neural networks generalise better than their less smooth counterparts.
Methods that encourage smoothness---such as weight decay, dropout, and early stopping ---have been long shown to aid generalisation~\citep{bartlett1997valid,golowich2018size,train_faster_generalize_better,deep_double_descent,dropout}.
Data augmentation
 has been shown to increase robustness to random noise or to modality specific transformations, such as image cropping and rotations~\citep{krizhevsky2012imagenet,simonyan2014very,data_augumentation_generative_modeling}.
~\citet{sokolic2017robust} show that the generalisation error of a network with linear, softmax, and pooling layers is bounded by the classification margin in input space.
Since classifiers are trained to increase classification margins in output space, smoothing by bounding the spectral norm of the model's Jacobian increases generalisation performance; this leads to empirical gains on standard image classification tasks. \citet{bartlett2017spectrally} provide a generalisation bound depending on classification margins and the product of spectral norms of the model's weights and show how empirically the product of spectral norms correlates with excess risk. 

\begin{figure}[t]
  \centering
  \begin{subfloat}[Traditional U-shaped curve. ]{
  \includegraphics[width=0.45\columnwidth]{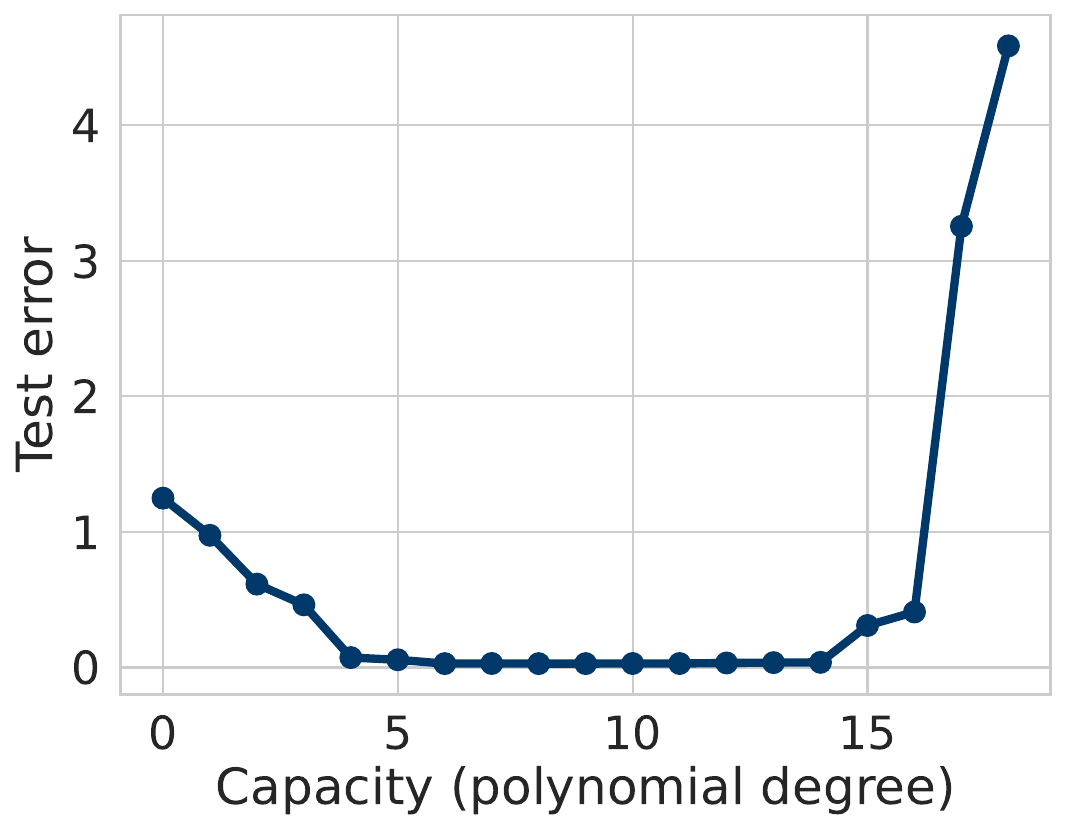}
  \label{fig:u_shape_p}
  } \end{subfloat}%
  \begin{subfloat}[Double descent in neural networks.]{
  \includegraphics[width=0.44\columnwidth]{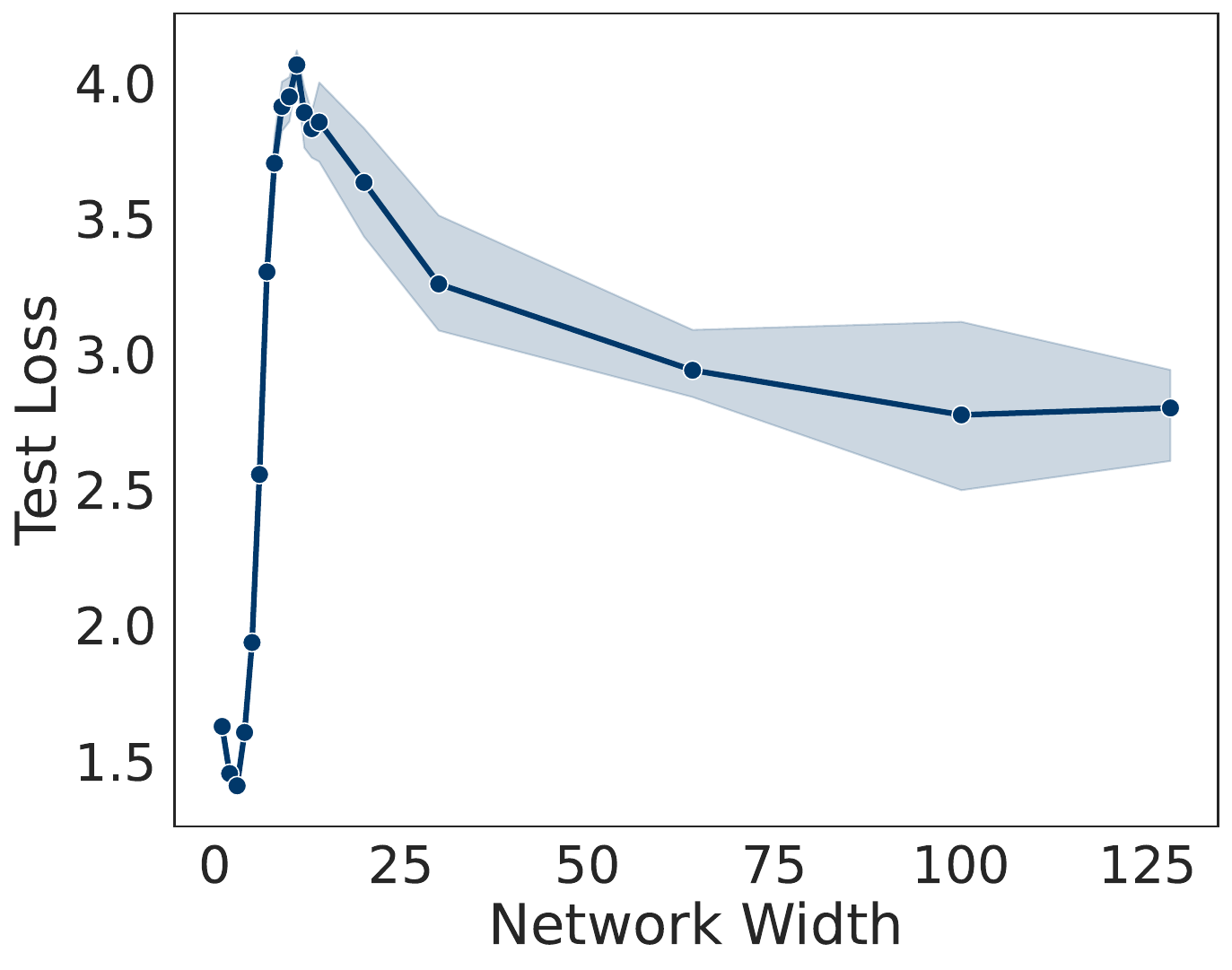}
  \label{fig:dd_shape_nets}
  }
  \end{subfloat}
  \caption[Double descent in deep learning compared to the traditional U-shaped complexity curve.]{Double descent in deep learning compared to the traditional U-shaped complexity curve. \subref{fig:u_shape_p}: traditional U-shaped curve associated with increased capacity; as we increase capacity, the test error first decreases after which it increases due to overfitting---the example we show here is that from Figure~\ref{fig:smooth_ex}. Figure~\ref{fig:dd_shape_nets}: when the complexity of neural networks is measured using the number of model parameters a double descent is observed, where a first descent is followed by a second descent.}
  \label{fig:smoothness_double_descent}
\end{figure}

Generalisation in deep learning has been recently re-examined under the light of double descent~\citep{deep_double_descent,double_descent},
a phenomenon named after the shape of the generalisation error
plotted against the size of a deep neural network: as the size of the network increases the generalisation error decreases (first descent), then increases, after which it decreases again (second descent). This is unlike the traditional U-shaped curve associated with non-neural models, where the second descent does not occur.
We plot the traditional U-shaped curve in Figure~\ref{fig:u_shape_p} and contrast it with the double descent phenomenon shown in Figure~\ref{fig:dd_shape_nets}.
We postulate there is a strong connection between double descent and smoothness:
in the first descent, the generalisation error is decreasing as the model is given extra capacity to
capture the decision surface; the increase happens when the model has enough capacity to fit the training data, but it cannot do so and retain smoothness; the second descent occurs as the capacity increases and smoothness can be retained.
 This view of double descent is supported by empirical evidence that shows that its effect
is most pronounced on clean label datasets and when early stopping and other regularisation techniques are not used~\citep{deep_double_descent}. Later in this chapter, we show that smoothness constraints heavily interact with optimisation. This further suggests that empirical investigations into the impact of smoothness on the double descent phenomenon are needed, since implicit regularisation effects induced by optimisation might act as  smoothness regularisers affecting model complexity. We will come back to this question in Chapter~\ref{ch:gc}.

\textbf{Reliable uncertainty estimates}. Neural networks trained to minimise classification losses provide notoriously unreliable uncertainty estimates;
an issue which gets compounded when the networks are faced with out of distribution data.
However, one can still leverage the power of neural networks to obtain reliable uncertainty estimates by combining smooth neural feature learners with non-softmax decision surfaces
~\citep{van2020simple,balaji_uncertanty_smoothness}.
The choice of smoothness regularisation or classifier can vary,
from using
 gradient penalties on the neural features with a Radial Basis Function classifier~\citep{van2020simple},
to using Spectral Normalisation on neural features
and a Gaussian Process classifier~\citep{balaji_uncertanty_smoothness}.
These methods
are competitive with standard techniques used for out-of-distribution detection~\citep{lakshminarayanan2017simple}
 on both vision and language understanding tasks.
The importance of smoothness regularising neural features indicates that having a smooth decision surface, such as a Gaussian Process, is not sufficient to compensate for sharp feature functions when learning models for uncertainty estimation~\citep{lakshminarayanan2017simple}.

\textbf{Robustness to adversarial attacks}. Adversarial robustness has become an active area of research in recent years~\citep{szegedy2013intriguing,goodfellow2014explaining,papernot2016limitations,chakraborty2018adversarial}.
Early works have observed that the existence of adversarial examples is related to the magnitude of the
gradient of the hidden network activation with respect to its input, and suggested that constraining
the Lipschitz constant of individual layers can make networks more robust to attacks~\citep{szegedy2013intriguing}.
Initial approaches to combating adversarial attacks, however, focused on data augmentation methods~\citep{kurakin2016adversarial,goodfellow2014explaining,moosavi2017universal,madry2017towards},
and only more recently smoothness has come into focus~\citep{cisse2017parseval,novak2018sensitivity,sokolic2017robust,lassance2018laplacian}.
We can see the connection between smoothness and robustness by looking at the desired robustness properties of classifiers, which aim to ensure that inputs in the same $\epsilon$-ball result in the same function output:
\begin{align}
 \norm{\vx - \vx'} \le \epsilon \implies \arg \max f(\vx) = \arg \max f(\vx').
 \label{eq:adv_examples}
\end{align}
The aim of adversarial defences and robustness techniques is
to have $\epsilon$ be as large as possible without affecting classification accuracy.
Robustness against adversarial examples has been shown to correlate with generalisation~\citep{gilmer2018adversarial}, and with the sensitivity of the network output with respect to the input as measured by the Frobenius norm of the Jacobian of the learned function~\citep{novak2018sensitivity,sokolic2017robust}.
~\citet{lassance2018laplacian} show that robustness to adversarial examples is enhanced when the function approximator is smooth as defined by the Laplacian smoothness signal discussed in Section~\ref{sec:smoothness_techniques}.
~\citet{tsuzuku2018lipschitz} show that  Eq~\eqref{eq:adv_examples}
holds when the $l_2$ norm is used if $\epsilon$ is smaller than the ratio of the classification margin and the
Lipschitz constant of the network times a constant, and thus they increase robustness by ensuring the margin is larger than the Lipschitz constant.

\textbf{Improved generative modelling performance}.
Smoothness constraints through gradient penalties or Spectral Normalisation have become a
recipe for obtaining state of the art generative models.
In Generative Adversarial Networks (GANs)~\citep{goodfellow2014generative}, smoothness constraints on
the discriminator and the generator have played a big part in scaling up training on large,
diverse image datasets at high resolution~\citep{biggan} and a combination of smoothness
constraints has been shown to be a requirement to get GANs to work on discrete data such as text~\citep{scratch_gan}.
The latest variational autoencoders~\citep{kingma2013auto,rezende2014stochastic} incorporate Spectral Regularisation to boost performance and stability~\citep{vahdat2020nvae}.
Explicit likelihood tractable models like normalizing flows~\citep{normalizing_flows} benefit from smoothness constraints through powerful invertible layers built using residual connections
$g(\vx) = \vx + f(\vx)$ where $f$ is Lipschitz
~\citep{invertible_residual_networks}.

\textbf{More informative critics}. Critics, learned approximators to intractable decision functions, have become a fruitful endeavour in generative modelling, representation learning, and reinforcement learning.

Critics are used in generative modelling to approximate divergences and distances between the learned model and the true unknown
data distribution, and have been mainly popularised by GANs.
A critic in a function class $\mathcal{F}$ can be used to approximate the KL divergence by minimizing the bound~\citep{nguyen2010estimating,f_gan}:
\begin{align}
\mathrm{KL}(p||q) &= \mathbb{E}_{p(\vx)} \log\frac{p(\vx)}{q(\vx)} = \sup_{f} \mathbb{E}_{p(\vx)} f(\vx) - \mathbb{E}_{q(\vx)} e^{f(\vx)-1} \\&\ge \sup_{f \in \mathcal{F}} \mathbb{E}_{p(\vx)} f(\vx) - \mathbb{E}_{q(\vx)} e^{f(\vx)-1}.
\label{eq:kl_bound}
\end{align}
Due to the density ratio $p(\vx)/q(\vx)$ in its definition, the KL divergence provides no learning signal when the model and data distributions do not have overlapping support. Choosing $\mathcal{F}$ to be a family of smooth functions in Eq~\eqref{eq:kl_bound}, however, results in a bound on the KL that provides useful gradients and can be used to train a model, even when the two distributions do not have overlapping support~\citep{fedus2017many,arbel2020kale}.
 We show an illustrative example in Figure~\ref{fig:kale_gan}: the true decision surface jumps from zero to infinity,
 while the approximation provided by the MLP is smooth.
Similarly, training the critic more and making it better at estimating the true decision surface but less smooth can hurt training~\citep{schafer2019implicit}. It's not surprising that imposing smoothness constraints on critics has become part of many flavours of GANs~\citep{arjovsky2017wasserstein,gulrajani2017improved,fedus2017many,biggan,arbel2018gradient,lipschitz_gan,yoshida2017spectral}.

The same conclusions have been reached in unsupervised representation learning, where
parametric critics are trained to approximate another intractable quantity, the mutual information, using
 the Donsker–Varadhan or similar bounds~\citep{belghazi2018mine,cpc}.
An extensive study on representation learning techniques based on mutual information showed that tighter bounds do not lead to better representations~\citep{on_mi_for_rep_learning}. Instead, the success of these methods is attributed to the inductive biases of the critics employed to approximate the mutual information.
In reinforcement learning, neural function approximators or neural ``critics'' approximate
state-value functions or action-state value functions and are then used to train a policy to maximise the expected reward. Directly learning a neural network parametric estimator of the action value \textit{gradients}
---the gradients of the action value with respect to the action---results in more accurate gradients (Figure 3 in \citep{d2020learn}),
 but also makes gradients smoother.
 This provides an essential exploration prior in continuous control,
 where similar actions likely result in the same reward and observing the same action twice is
 unlikely due to size of the action space;
 encouraging the policy network to extrapolate from the closest seen action
improves performance over both model free and model based continuous control approaches~\citep{d2020learn}.

\begin{figure}[t]
  \centering
  \begin{subfloat}[KL divergence optimal critic $f^* = \frac{p}{q}$ \newline and smooth critic estimate. ]{
  \includegraphics[width=0.47\columnwidth]{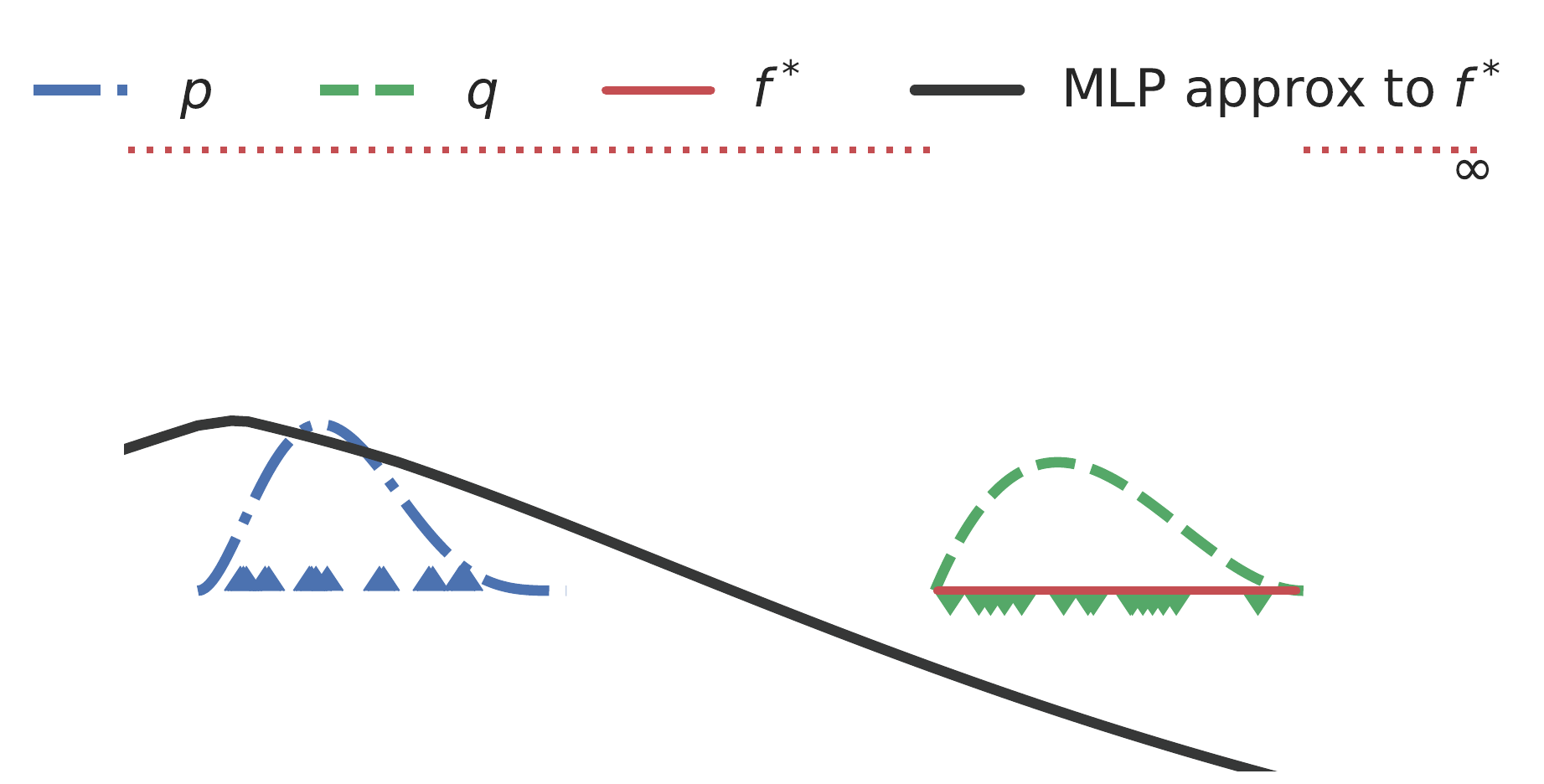}
  \label{fig:kale_gan}
  } \end{subfloat}%
  \begin{subfloat}[Optimal Wasserstein critic is smooth.]{
  \includegraphics[width=0.43\columnwidth]{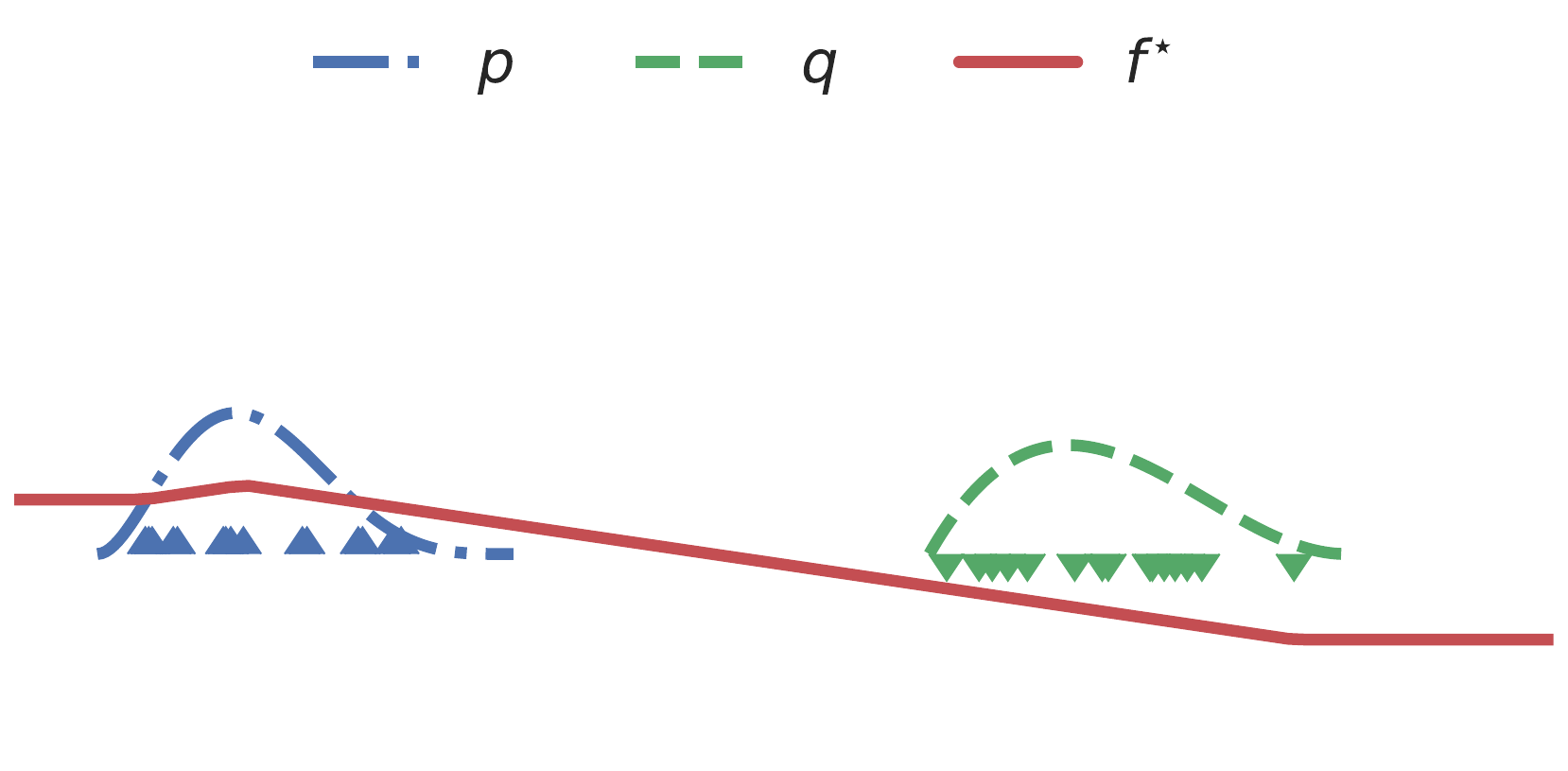}
  \label{fig:optimal_critic_wasserstein}
  }
  \end{subfloat}
  \caption[The importance of critic smoothness when estimating divergences and distances.]{The importance of critic smoothness when estimating divergences and distances. \subref{fig:kale_gan}: When the two distributions do not have overlapping support, the KL divergence provides no learning signal, while a smooth approximation via a learned critic does. \subref{fig:optimal_critic_wasserstein}: The optimal Wasserstein critic has a smoothness Lipschitz constraint in its definition.}
  \label{fig:smooth_wass_mlp}
\end{figure}

\textbf{Distributional distances}. Including smoothness constraints in the definition of distributional distances by using optimal transport has seen a uptake in machine learning applications in recent years,
from generative modelling~\citep{arjovsky2017wasserstein,gulrajani2017improved,wasserstein_autoencoders,ostrovski2018autoregressive}
to reinforcement learning~\citep{bellemare2017distributional,dabney2018implicit}, neural flows~\citep{finlay2020train}
 and fairness~\citep{chiappa2020general,jiang2020wasserstein}.
Optimal transport is connected to Lipschitz smoothness as the Wasserstein distance can be computed via the Kantorovich-Rubinstein duality~\citep{villani2008optimal}:
\begin{align}
W_1(p(\vx), q(\vx)) = \sup_{f: \norm{f}_{\text{Lip}} \le 1} \mathbb{E}_{p(\vx)} f(\vx) - \mathbb{E}_{q(\vx)} f(\vx).
\label{eq:wasserstein}
\end{align}

The Wasserstein distance finds the critic that can separate the two distributions in expectation,
but constraints that critic to be Lipschitz in order to avoid pathological solutions.
The importance of the Lipschitz constraint on the critic can be seen in Figure~\ref{fig:optimal_critic_wasserstein}:
unlike the KL divergence,
the optimal Wasserstein critic is well defined when the two distributions do not have overlapping support, and does not require an approximation to provide useful learning signal for a generative model.

\section{Challenges with smoothness regularisation}
\label{sec:downsides}

Despite the many benefits of learning smooth models we outlined in the previous section, many of the effects of smoothness regularisation have not been extensively studied. In this section, we lay the ground work for the next chapters by looking at some of the unexplored effects of smoothness regularisation, from reduced model capacity to interactions with optimisation.
Figure reproduction details for this chapter are available in Appendix Chapter~\ref{app:smoothness_exp}.

\subsection{Too much smoothness hurts performance}
\label{sec:too_much_smoothness}

Needlessly limiting the capacity of our models by enforcing smoothness constraints is a significant danger:
 a constant function is very smooth, but not very useful.
Beyond trivial examples, \citet{excessive_invariance_adv_vulnerability} show that one of the reasons  neural networks are vulnerable to adversarial perturbations is invariance to task relevant changes---too much smoothness with respect to the wrong metric. A neural network can be ``too Lipschitz'':
methods aimed at increasing robustness to adversarial examples do indeed decrease the Lipschitz constant of a classifier,
but once the Lipschitz constant becomes too low, accuracy drops significantly~\citep{fazlyab2019efficient}.

There are two main avenues for being too restrictive in the specification of smoothness constraints,
depending on \textit{where} and \textit{how} smoothness is encouraged.
  Smoothness constraints can be imposed on the entire input space or
only in certain pockets, often around the data distribution.
Methods that impose constraints on the entire space throw away useful information about the input distribution and restrict the learned function needlessly by forcing it to be smooth in areas of the space where there is no data. This is especially problematic when the input lies on a small manifold in a large dimensional space, such as in the case of natural images, which are a tiny fraction of the space of all possible images.
Model capacity can also be needlessly restrained by imposing strong constraints on the individual components of the model, often the network layers, instead of allowing the network to allocate capacity as needed.

\begin{figure}[t]
  \centering
  \captionsetup{justification=centering}
  \begin{subfloat}[2 layer MLP.]{
      \includegraphics[width=0.333\textwidth]{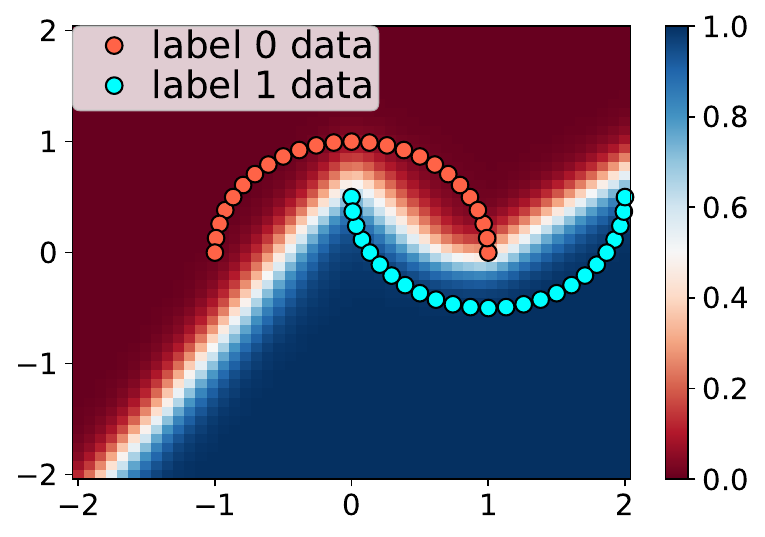}
    \label{fig:two_moons_decision_surface_shallow_mlp}
  }
  \end{subfloat}%
  \begin{subfloat}[4 layer MLP.]{
    \includegraphics[width=0.333\textwidth]{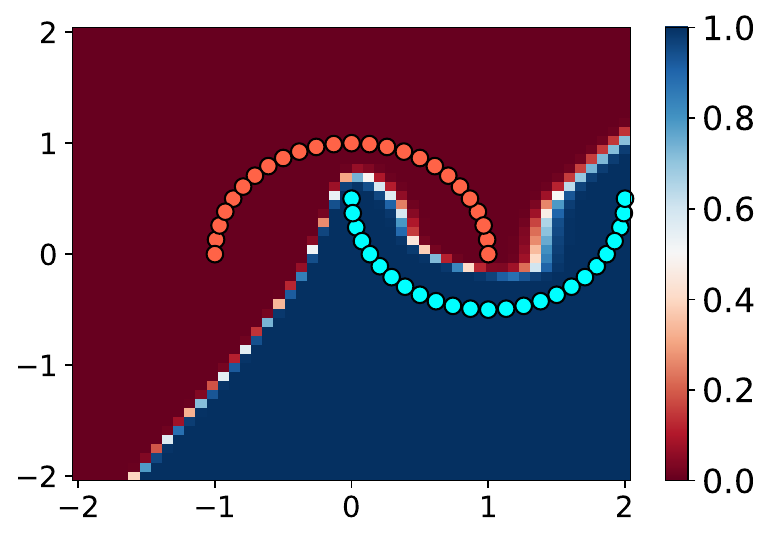}
  \label{fig:two_moons_decision_surface_deep_mlp}
  }
  \end{subfloat}%
  \begin{subfloat}[Gradient penalty at data; \\$K=1$.]{
    \includegraphics[width=0.333\textwidth]{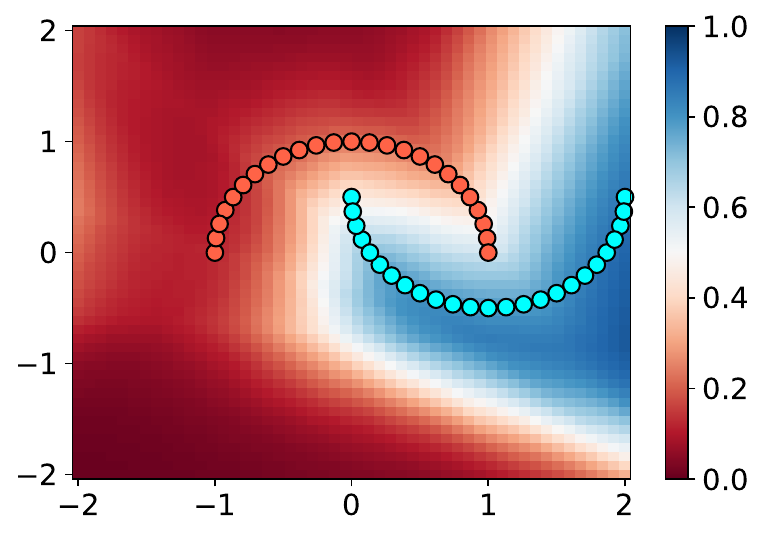}
    \label{fig:two_moons_decision_surface_grad_penalty}
  }
  \end{subfloat}\\
  \begin{subfloat}[Spectral norm; \\$K=1$.]{
    \includegraphics[width=0.333\textwidth]{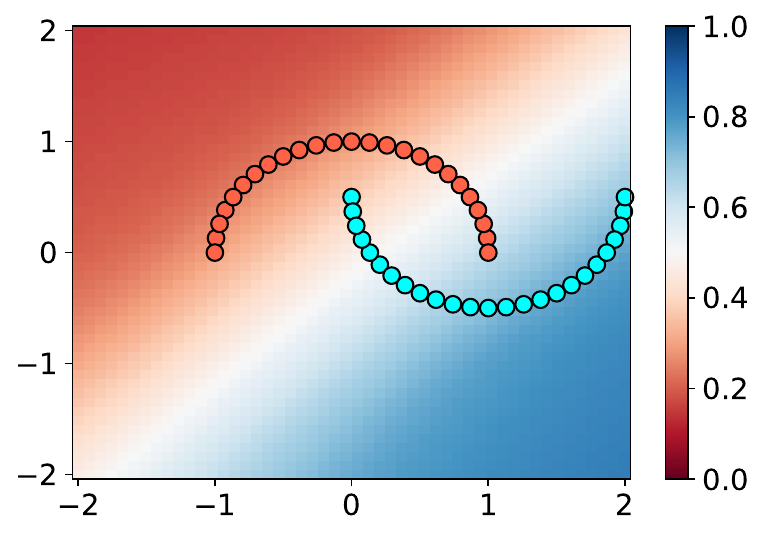}
  \label{fig:two_moons_decision_surface_sn}
  }
  \end{subfloat}%
  \begin{subfloat}[Spectral norm; \\$K=10$.]{
    \includegraphics[width=0.333\textwidth]{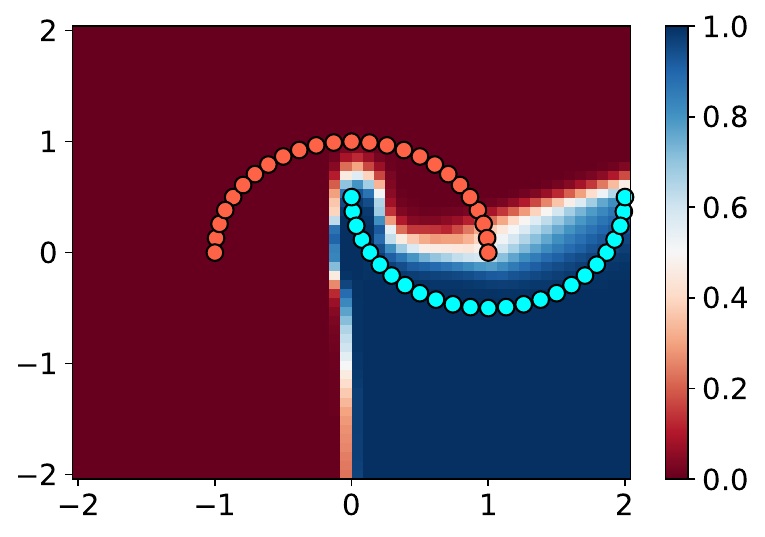}
  \label{fig:two_moons_decision_surface_sn_10}
  }
  \end{subfloat}%
  \begin{subfloat}[Spectral norm; \newline scaling the data by 10.]{
    \includegraphics[width=0.33\textwidth]{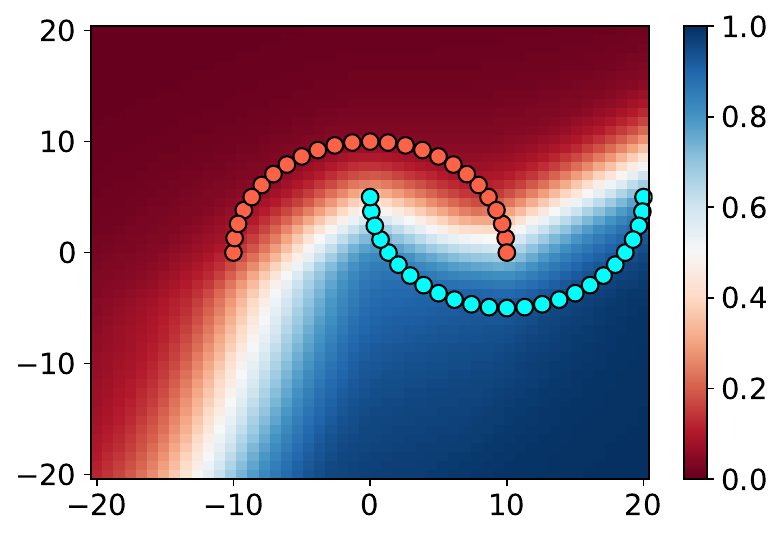}
    \label{fig:two_moons_decision_surface_sn_data_scaling}
    }
  \end{subfloat}
  \captionsetup{justification=justified}
  \caption[Decision surfaces on \textit{two moons} under different regularisation methods.]{Decision surfaces on \textit{two moons} under different regularisation methods.\\ Unless otherwise specified the model architecture is a 4 layer MLP. We observe that network depth can decrease smoothness \subref{fig:two_moons_decision_surface_deep_mlp}; that data dependent smoothness penalties can increase smoothness \subref{fig:two_moons_decision_surface_grad_penalty}; that hard smoothness constraints can lead to smooth surfaces but hurt performance \subref{fig:two_moons_decision_surface_sn}, but that this effect can be mitigated by increasing the Lipschitz constant of the constraint \subref{fig:two_moons_decision_surface_sn_10} or scaling the range of the input data \subref{fig:two_moons_decision_surface_sn_data_scaling}.}
  \label{fig:two_moons_decision_surface}
\end{figure}

\begin{figure}[t]
  \centering
  \captionsetup{justification=centering}
  \begin{subfloat}[Unregularised.]{
  \includegraphics[width=0.335\textwidth]{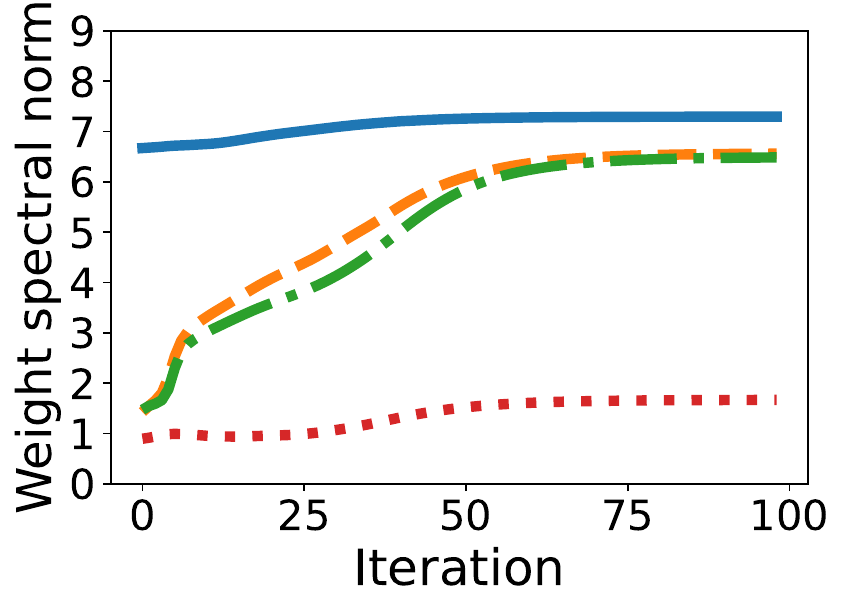}
  }
  \end{subfloat}%
  \begin{subfloat}[Gradient penalty. \newline $K=1$.]{
  \includegraphics[width=0.33\textwidth]{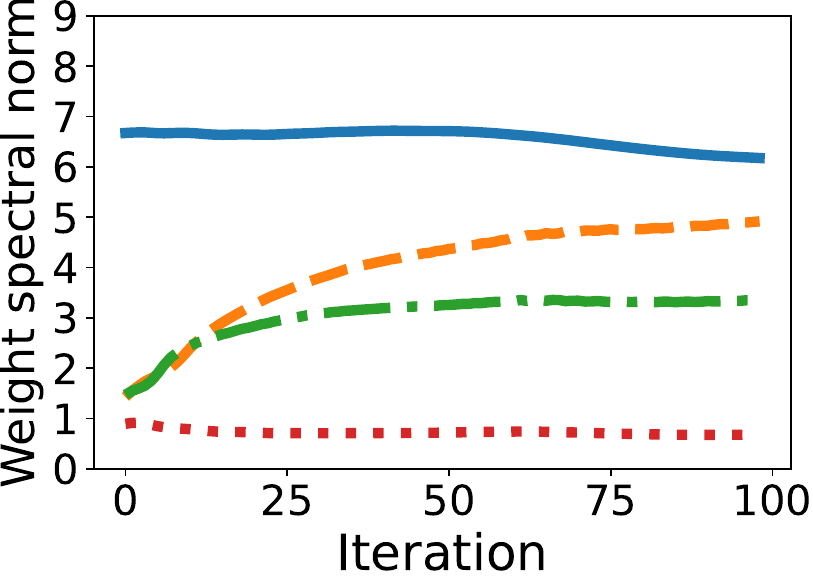}
  }
  \end{subfloat}%
  \begin{subfloat}[Spectral Normalisation. \newline $K=1$.]{
  \includegraphics[width=0.33\textwidth]{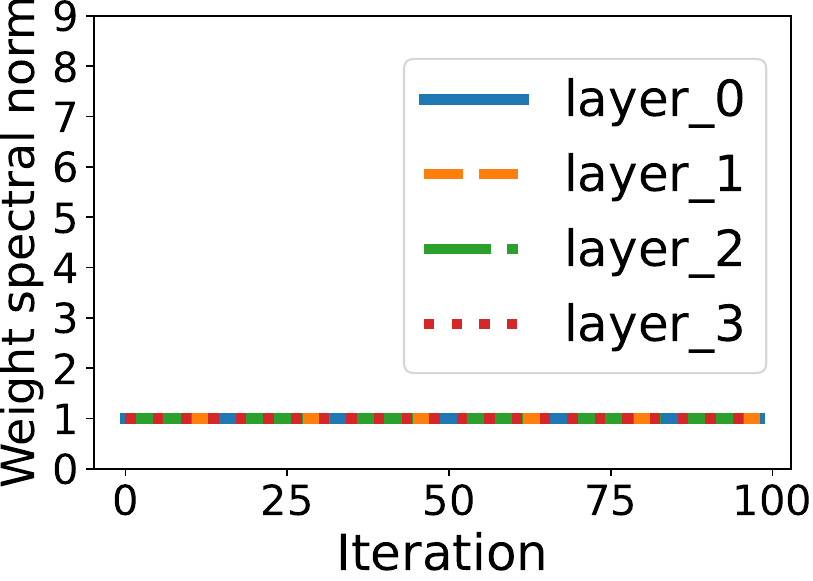}
  }
  \end{subfloat}%
  \captionsetup{justification=justified}
  \caption[Lipschitz constant of each layer of an MLP trained on the two moons dataset.]{Lipschitz constant of each layer of an MLP trained on the two moons dataset. \\The decision surfaces for the same models can be seen in Figure~\ref{fig:two_moons_decision_surface}. Smaller means smoother. Soft regularisation via the gradient penalty leads to a decrease in spectral norm of the  model's weight matrices, with each layer adjusting differently to the constraint. In contrast, for Spectral Normalisation the spectral norm of the weight matrix for every layer is equal to 1.}
  \label{fig:two_moons_layers_constant}
\end{figure}

\begin{figure}[t!]
  \centering
  \captionsetup{justification=centering}
  \begin{subfloat}[Unregularised.]{
  \includegraphics[width=0.333\textwidth]{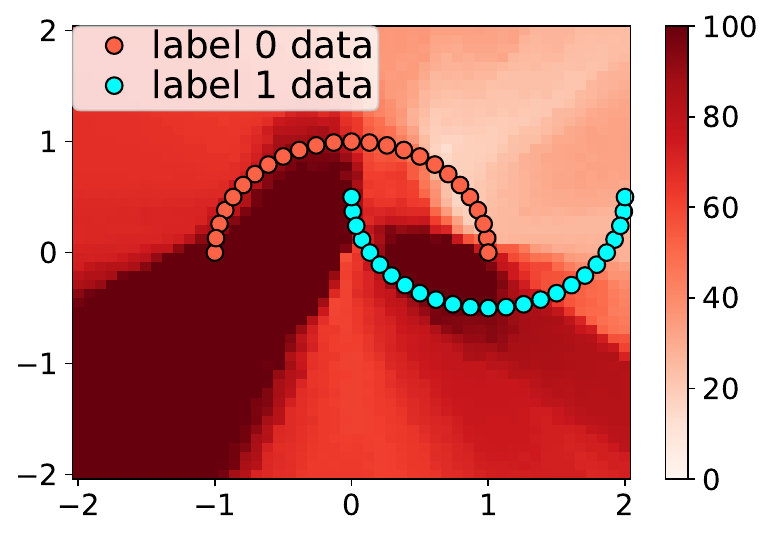}
  }
  \end{subfloat}%
  \begin{subfloat}[Gradient penalty. \newline $K=1$.]{
  \includegraphics[width=0.333\textwidth]{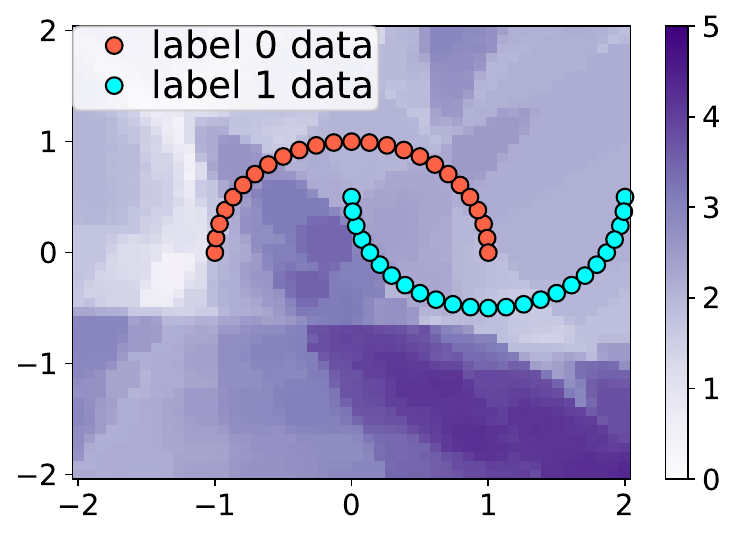}
  }
  \end{subfloat}%
  \begin{subfloat}[Spectral Normalisation. \newline $K=1$.]{
  \includegraphics[width=0.333\textwidth]{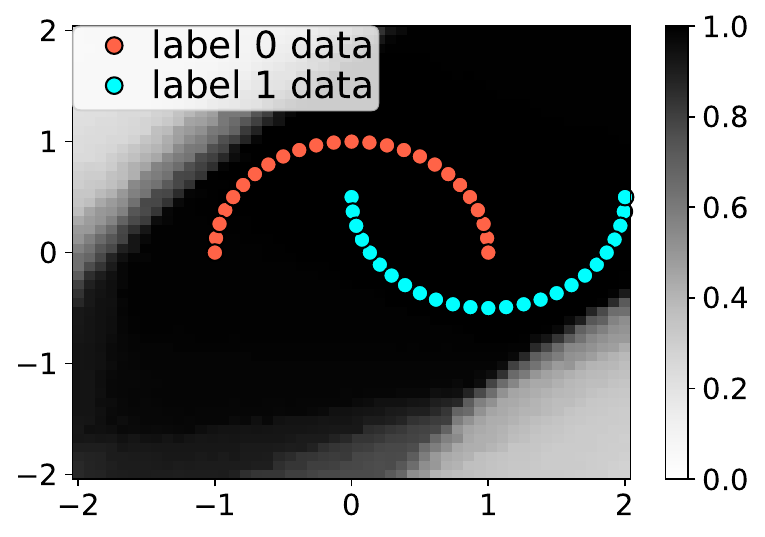}
  }
  \end{subfloat}%
  \captionsetup{justification=justified}
  \caption[The effect of regularisation on \textit{local} smoothness.]{The effect of regularisation on \textit{local} smoothness. We plot the local Lipschitz constants obtained using an exhaustive grid search in \textit{local neighbourhoods}, instead of loose bounds. We use different colours to emphasise the different scale of the constants for the different methods. While gradient penalties increase smoothness, they do so primarily around the data, where the constraint is active, while Spectral Normalisation affects a large part of the input space.}
  \label{fig:two_moons_layers_local_constant}
\end{figure}

We can exemplify the importance of where and how constraints are imposed with an example, by contrasting gradient penalties---end to end regularisation applied around the training data---and Spectral Normalisation---layer-wise regularisation applied to the entire space. Figure~\ref{fig:two_moons_decision_surface_sn} shows that using Spectral Normalisation to restrict the Lipschitz constant of an MLP to be 1 decreases the capacity of the network and severely affects accuracy compared to the
 baseline MLP---Figure~\ref{fig:two_moons_decision_surface_deep_mlp}---or the MLP regularised using gradient penalties---Figure~\ref{fig:two_moons_decision_surface_grad_penalty}.
 Further insight comes from Figure~\ref{fig:two_moons_layers_constant}, which shows that the gradient penalty only enforces a
weak constraint on the model and does not heavily restrict the spectral norms of individual layers; this is in stark contrast with Spectral Normalisation, which, by construction, ensures each network layer has spectral norm equal to 1. To show the effect of data dependent regularisation on \textit{local smoothness} we plot the Lipschitz constants of the model at neighbourhoods spanning the entire space in Figure~\ref{fig:two_moons_layers_local_constant}. Each Lipschitz constant is
computed using an exhaustive grid search inside each local neighbourhood rather than a bound---details are provided in Appendix~\ref{app:smoothness_exp}.
As expected, gradient penalties impose stronger constraints around the training data, while Spectral Normalisation has a strong effect on the smoothness around points in the entire space.
This simple example suggests that the search for better smoothness priors needs
to investigate \textit{where} we want functions to be smooth
and re-examine \textit{how} smoothness constraints should account for the compositional aspect of neural networks,
otherwise we run the risk of learning trivially smooth functions. In the next chapter, we will first hand encouter the need to avoid enforcing too strong smoothness constraints when applying Spectral Normalisation to reinforcement learning.

\subsection{Interactions between smoothness regularisation and optimisation}
\label{sec:opt_smoothness_int}

We show that viewing smoothness only through the lens of the model is misleading, as smoothness constraints have a strong effect on optimisation.
The interaction between smoothness and optimisation has been mainly observed when training generative models;
 encouraging the smoothness of the encoder through Spectral Regularisation increased the stability of hierarchical VAEs and led variational inference models to the state of the art of explicit likelihood non autoregressive models~\citep{vahdat2020nvae},
 while smoothness regularisation of the critic (or discriminator)
has been established as an indispensable stabiliser of GAN training, independently of the training criteria used~\citep{yoshida2017spectral,arbel2018gradient,fedus2017many,biggan,kurach2019large}.

Some smoothness regularisation techniques affect optimisation by
changing the loss function (gradient penalties, Spectral Regularisation) or the optimisation regime directly (early stopping, projection methods). Even if they don't explicitly change the loss function or optimisation regime,
smoothness constraints affect the path the model takes to reach convergence.
We use a simple example to show why smoothness regularisation interacts with optimisation in Figure~\ref{fig:mnist_classification}. We use different learning rates to train two unregularised MLP classifiers on MNIST~\citep{lecun1999object}  and observe that the learning rate used affects its smoothness throughout training, without changing testing accuracy. Here, we measured the model's smoothness as the upper bound on its Lipschitz constant given by product of spectral norms of learnable layers. These results show that
imposing similar smoothness constraints on two models that share the same architecture but are trained with  different learning rates would lead to very different strengths of regularisation and drastically change the trajectory of optimisation. This is particularly true for methods such as Spectral Normalisation, which directly impact the spectral norms of individual layers we measured here.

\begin{figure}[t]
\centering
  \includegraphics[width=0.5\textwidth]{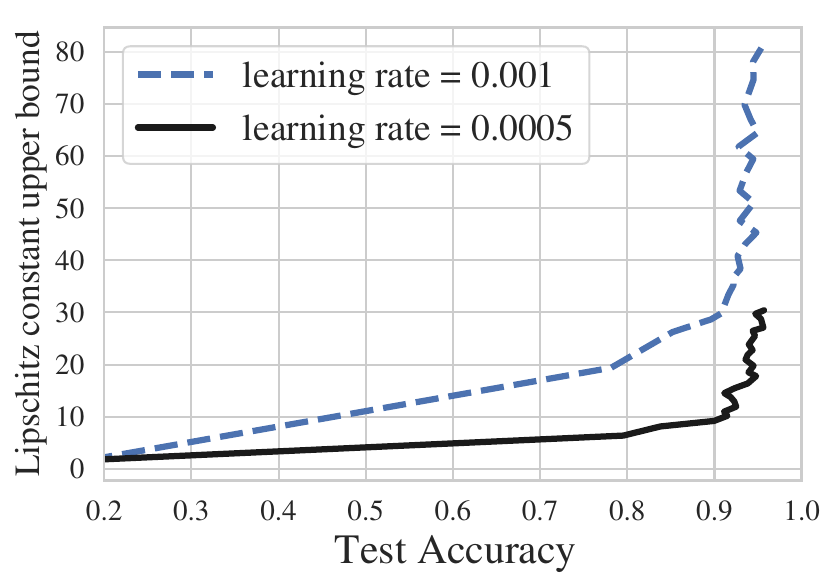}
  \caption[Smoothness constraints interact with learning rates.]{ Smoothness constraints interact with learning rates. MNIST classifiers trained with different learning rates exhibit different smoothness behaviour throughout training, even at the same test accuracy. We measured the model's smoothness as the upper bound on its Lipschitz constant given by product of spectral norms of learnable layers. This example shows that imposing the same smoothness constraint on the same model architecture trained with different learning rates will have different effects on optimisation dynamics. This can be particularly the case for methods that impact the spectral norms of individual layers we measured here.}
  \label{fig:mnist_classification}
\end{figure}
\begin{figure}[t]
  \centering
    \begin{subfloat}[Momentum decay rate $\beta_1 = 0.5$.]{
        \includegraphics[width=0.5\textwidth]{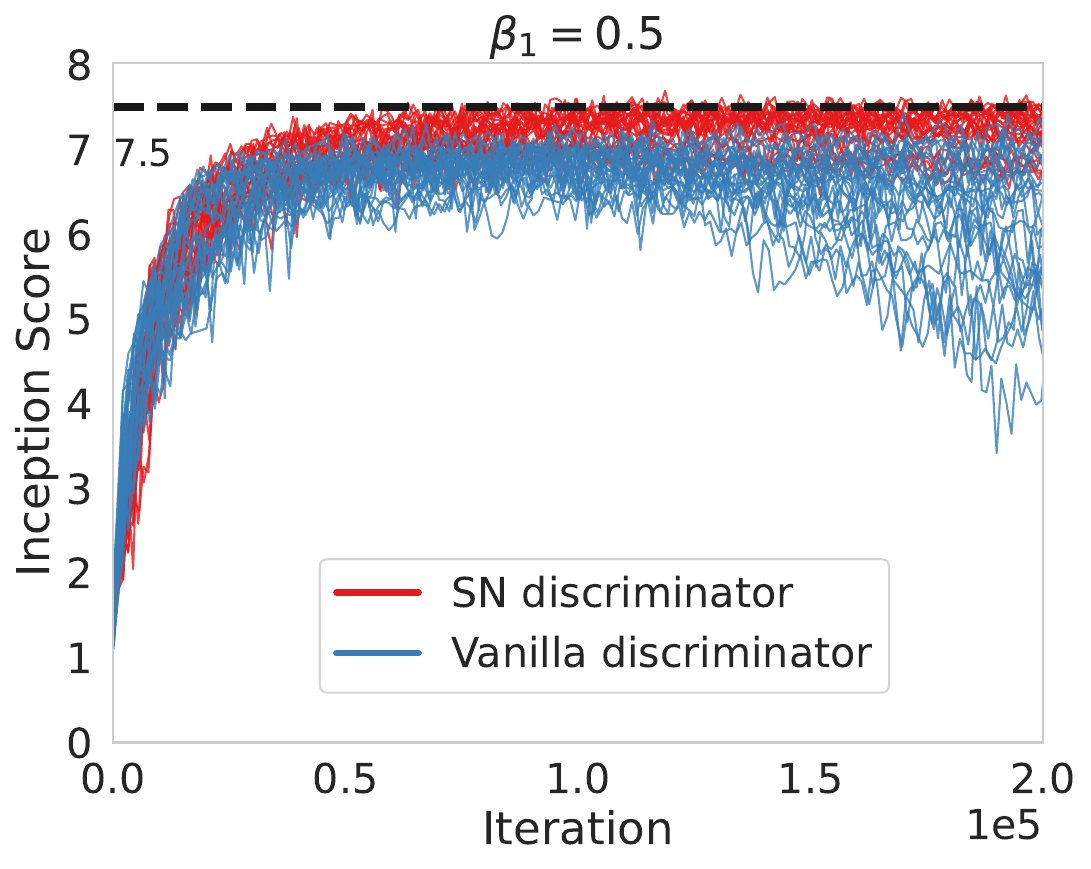}%
    }\end{subfloat}%
    \begin{subfloat}[Momentum decay rate $\beta_1 = 0.9$.]{
      \includegraphics[width=0.5\textwidth]{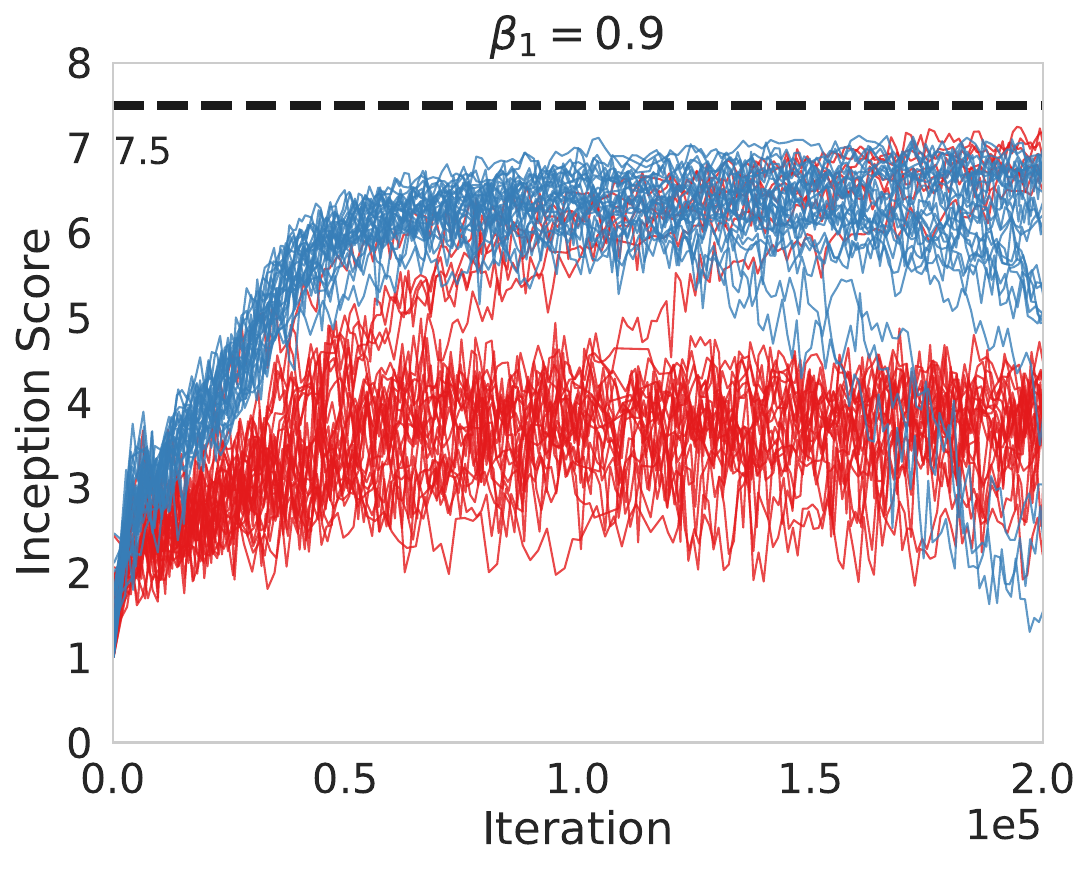}%
    }
    \end{subfloat}%
  \captionsetup{justification=justified}
  \caption[Smoothness constraints interact with momentum decay rates.]{Smoothness constraints interact with momentum decay rates: Spectral Normalisation requires low momentum in GAN training. The effects of momentum decay rate $\beta_1$ in the Adam optimiser on Spectral Normalisation applied to the GAN discriminator on CIFAR-10, over a sweep of learning rates. Higher is better. Individual learning curves are obtained from a sweep over seeds and learning rates.}
  \label{fig:sn_mom}
\end{figure}

Beyond learning rates, smoothness constraints also interact  decay rate hyperparameters in optimisers that use momentum. When the Adam optimiser is used, this is the $\beta_1$ hyperparameter; for an overview of Adam and momentum we refer the reader to Section~\ref{sec:opt_algo}.
When training GANs,~\citet{gulrajani2017improved} observed that weight clipping in the Wasserstein GAN critic requires low to no momentum. Weight clipping has since been abandoned in the favour of other methods, but as we show in Figure~\ref{fig:sn_mom}, current methods like Spectral Normalisation
applied to GAN critics trained with low momentum decrease sensitivity to learning rates but perform poorly in conjunction with high momentum, leading to slower convergence and higher hyperparameter sensitivity.

We have shown that smoothness constraints interact with optimisation parameters, such as learning rates and momentum, and argue that we need to reassess our understanding of smoothness constraints, not only as constraints on the final model, but as methods that influence the \textit{optimisation path}. We will study this in detail in the next chapter.

\subsection{Sensitivity to data scaling}

Sensitivity to data scaling of smoothness constraints can
make training neural network models sensitive to additional hyperparameters.
Let $f^*$ be the optimal decision function for a task obtained from using i.i.d. samples from random variable $X$, and
$f_c^*$ obtained similarly from i.i.d. samples obtained from $c X$.
Since $f^*$ and $f_c^*$ can be highly non linear, the relationship between the smoothness of the two functions is unclear.
 This gets further complicated when we consider their closest approximators under a neural family.
The effect of data scaling on the smooth constraints
 required to fit a model can be exemplified using the two moons dataset:
 with a Lipschitz constraint of 1 on the model the data is poorly fit---Figure~\ref{fig:two_moons_decision_surface_sn}---but a much better fit can be obtained by changing the Lipschitz constant to 10---Figure~\ref{fig:two_moons_decision_surface_sn_10}---or scaling the data---Figure~\ref{fig:two_moons_decision_surface_sn_data_scaling}.

\section{Conclusion}

In this chapter, we outlined the many benefits of neural network smoothness across multiple problem domains. Motivated by these benefits and the wide use of smoothness regularisation in deep learning, we highlighted areas where the effects of smoothness have not been fully understood, from connections to the double descent phenomenon to interactions between smoothness regularisation and optimisation. We continue this line of inquiry in the following chapters.

\chapter{Smoothness and optimisation effects of Spectral Normalisation}
\label{ch:rl}

Despite the many benefits of smoothness in supervised learning and generative modelling outlined in the previous chapter, smoothness regularisation has only been briefly studied in deep reinforcement learning (DRL). 
In this chapter, we study the effects of Spectral Normalisation, a common smoothness regularisation method described in Section~\ref{sec:smoothness_techniques}, on DRL.  We use Spectral Normalisation as it has shown tremendous success in GANs, and DRL and GANs share common characteristics, such as operating in non-stationary settings and challenges with optimisation~\citep{czarnecki2019distilling,bengio2020InterferenceAG,mescheder2017numerics,miyato2018spectral}. We find that Spectral Normalisation can lead to improved performance in DRL, as long as it is only applied selectively to a few layers in order to avoid the reduced capacity effects discussed in Section~\ref{sec:too_much_smoothness}. Building upon results in Section~\ref{sec:opt_smoothness_int}, we find interactions between Spectral Normalisation and optimisation dynamics, and show that in DRL the improvements brought by Spectral Normalisation can be obtained by optimisation only changes. 
Equipped with this knowledge, we reinvestigate the effect of Spectral Normalisation on GANs, and observe that there Spectral Normalisation plays a compounded role, with model and optimisation changes interacting with the choice of loss function 
in intricate ways.

\section{Introduction}

Reinforcement Learning (RL) has made astounding progress in recent years, showcasing the ability to play games where alternatives, such as handcrafted search or brute force approaches, are intractable~\citep{silver2017mastering,berner2019dota,vinyals2019grandmaster}.
While this progress could not be achieved without the feature learning aspect of DRL due to a prohibitively large state space or number of actions, research focus has been on RL-specific algorithmic improvements, such as objective functions, exploration strategies, and replay prioritisation \citep{Schaul2016PrioritizedER,Fortunato2018NoisyNF,Osband2016DeepEV,ostrovski2017count,bellemare2017distributional,dabney2017distributional,hessel2018RainbowCI}.
We take a different approach, and focus on the \textit{deep} aspect of DRL, by investigating the interplay between training neural networks and RL objectives. We do so by exploring the use Spectral Normalisation in DRL; as we have done in previous chapters, we hone in on interactions between smoothness regularisation and optimisation. To this effect, we make the following contributions:
\begin{itemize}
  \item We show that Spectral Normalisation, \textit{when applied to a subset of layers}, can improve performance of DRL agents and match the performance of Rainbow~\citep{hessel2018rainbow}, a competitive agent combining multiple RL algorithmic advances.
\item We show that the effect of selectively applying of Spectral Normalisation to a subset of layers is not solely a smoothness one, but an optimisation one, and can decrease sensitivity to optimisation hyperparameters.
  \item We introduce a set of adaptations to optimisation methods based on Spectral Normalisation, which \textit{without changing the model}, can recover and sometimes improve the results of Spectral Normalisation in DRL.
  \item We use our new insights to re-examine the effects of Spectral Normalisation in GANs, where we observe that normalising the discriminator avoids loss areas of low or high gradient magnitudes, and that Spectral Normalisation interacts with optimisation hyperparameters in important ways.
\end{itemize}

\section{Reinforcement learning}
RL translates the problem of acting sequentially in an environment in order to maximise reward into a formal optimisation problem, usually defined via Markov Decision Processes (MDP). An MDP is a tuple
$(\mathcal{S}, \mathcal{A}, P, r, \gamma)$ with $\mathcal{S}$ a set of available states,
$\mathcal{A}$ a set of available actions (which throughout this thesis we will assume to be discrete), $P$ a conditional transition probability or density 
$P: \mathcal{S} \times \mathcal{A} \times {S} \rightarrow \mathbb{R}^{+}$, such that 
$P(s'|s, a)$ defines the density to transition to state $s'$ from state $s$ if action $a$
is taken by the agent, as well as the unconditional $P(s_0)$ defining the density over initial states. For our use cases, we assume that $P$ is not available in closed form but can be sampled from by obtaining $s' \sim P(\cdot|s, a)$ from an environment simulator. The reward function $r: \mathcal{S} \times \mathcal{A} \rightarrow  \mathbb{R}$ defines the reward $r(s, a)$ obtained by the agent if it takes action $a$ from state $s$ and the discount factor $\gamma \in \mathbb{R}$, with $\gamma \in (0, 1)$.  The goal of RL is to find a policy $\pi$ defined as the conditional probability distribution $\pi: \mathcal{S}\times \mathcal{A} \rightarrow \mathbb{R}^{+}$, where  $\pi(a|s)$ encodes the probability of taking action $a$ if in state $s$ under policy $\pi$, such that the time discounted reward over an episode is maximised:
\begin{align}
\max_{\pi} \mathbb{E}_{P(s_0)} \mathbb{E}_{\pi(a_0|s_0)} \left[r(s_0, a_0) + \sum_{t=1}^{\infty} \mathbb{E}_{P(s_{t}|s_{t-1}, a_{t-1})} \mathbb{E}_{\pi(a_t|s_t)}   \gamma^t r(s_t, a_t) \right].
\label{eq:rl_def}
\end{align}
Directly optimising Eq~\eqref{eq:rl_def} via Monte Carlo estimation can be challenging as it can lead to high variance estimates of the objectives, as well as slow learning in the case of long episodes~\citep{sutton2018reinforcement}. Instead, it is common to define \textit{state-action value functions} as
\begin{align}
Q_{\pi}(s, a) = r(s, a) + \sum_{\substack{t=1,\\ a_0 = a, s_0=s}}^{\infty} \mathbb{E}_{P(s_{t}|s_{t-1}, a_{t-1})} \mathbb{E}_{\pi(a_t|s_t)}   \gamma^t r(s_t, a_t).
\end{align}
State-action value functions satisfy the recurrence relation
\begin{align}
Q_{\pi}(s, a) = r(s, a)+  \gamma \mathbb{E}_{P(s'|s, a)} \mathbb{E}_{\pi(a'|s')} Q_{\pi}(s', a'), \hspace{1em} \forall s \in \mathcal{S}, a \in \mathcal{A},
\label{eq:belman_eq}
\end{align}
known as the Bellman equation.
The objective in Eq~\eqref{eq:rl_def} can now be written as
\begin{align}
\max_{\pi} \mathbb{E}_{P(s_0)} \mathbb{E}_{\pi(a_0|s_0)} Q_{\pi}(s_0, a_0).
\end{align}
If we denote the optimal policy as $\pi^*$, another Bellman recurrence equation can also be used to find the $Q_\pi^*$, namely the Bellman optimality equation
\begin{align}
Q_{\pi^*}(s, a) = r(s, a)+  \gamma \mathbb{E}_{P(s'|s, a)} \max_{a'} Q_{\pi^*}(s', a'), \hspace{1em} \forall s \in \mathcal{S}, a \in \mathcal{A}.
\label{eq:belman_optim_eq}
\end{align}
Iterative updates ensuring Bellman optimality consistency of the current estimate $Q$ of $Q_{\pi^*}$ form the basis for $Q$-learning algorithms~\citep{watkins1992q}. Since $P(\cdot| s, a)$ is unknown, the expectation in Eq~\eqref{eq:belman_optim_eq} is replaced with a Monte Carlo estimate obtained by performing action $a$ in state $s$ and obtaining $s'$ from the environment; actions are often obtained using the $\epsilon$-greedy policy
\begin{align}
\pi(a|s) = \begin{cases}
              1 - \epsilon, \text{if } a = \arg\max_{a'} Q(s, a') \\
              \epsilon/(|\mathcal{A}| - 1), \text{otherwise}
           \end{cases}
\end{align}
and thus the Bellman optimality updates are evaluated at tuples \\ ${(s, a\sim \pi(\cdot|s), s' \sim P(\cdot| s, a), r = r(s, a))}$.

We note that while here we focus on value based, model-free, offline algorithms, other popular formulations of RL exist, and investigating the effect of smoothness for other algorithm families, including actor-critic and model-based RL, is a worthy area of study; for an overview of RL algorithms, see~\citet{sutton2018reinforcement}.

\subsection{Deep reinforcement learning}

While in traditional RL, often called tabular RL, when estimating a state-action value function individual values are approximated for each $s \in \mathcal{S}$ and $a \in \mathcal{A}$, in DRL $Q(s, a)$ is modelled using a neural network. The advantage of this approach is two fold: first, it makes large state and action spaces---which otherwise would be prohibitive due to memory requirements of storing a value for each state and each action---tractable; and second, it benefits from the generalisation provided by neural network features by learning a similarity measure between states.

One approach to DRL is DQN~\citep{mnih2015human}, which adapts the $Q$-learning algorithm to the neural setting, with the objective
\begin{align}
 \min_{\vtheta} E(\vtheta) = \mathbb{E}_{(s, a, s', r)\sim \mathcal{R}} \left(Q(s, a; \vtheta) -  \left(r+  \gamma \max_{a'} Q(s', a'; \vtheta)\right)\right)^2,
\label{eq:dqn_loss}
\end{align}
where $\mathcal{R}$ is a replay buffer that stores experiences obtained by acting according to an $\epsilon$ greedy policy induced by $Q$. Since computing the maximum operator requires computing $Q(s, a) \hspace{1em} \forall a \in \mathcal{A}$, the neural critic implementation takes as input a state $s$ and predicts using a forward pass all state-action value functions, thus implementing $Q(\cdot; \vtheta): \mathcal{S} \rightarrow \mathbb{R}^{|\mathcal{A}|}$.
 Due to instabilities in training, the target value $r+  \gamma \max_{a'} Q(s', a'; \vtheta)$ is often replaced with a target network $r+  \gamma \max_{a'} Q(s', a'; \vtheta_{old})$, where $\vtheta_{old}$ are a set of parameters obtained previously in training. The target network parameters $\vtheta_{old}$ get updated at regular intervals in training.

Many improvements in DRL have focused on changing RL specific aspects of training, such as finding better approaches to sample from the replay buffer $\mathcal{R}$, by weighing experiences either according to recency or estimated importance~\citep{Schaul2016PrioritizedER}~\citep{ostrovski2018autoregressive,Fortunato2018NoisyNF}. Significant progress has also been achieved by changes in loss functions and parametrisations, particularly through a distributional approach to value based learning~\citep{bellemare2017distributional,dabney2020TheVP}, which updates the Bellman operators to act on a distributional space rather than in expectation as shown in Eq~\eqref{eq:belman_eq}. The resulting DQN style agent obtained from the distributional perspective is called CategoricalDQN.
Rainbow~\citep{hessel2018RainbowCI} is an agent collating many of the above improvements, by combining categorical losses, improved prioritised replay strategies~\citep{Schaul2016PrioritizedER}, exploration bonuses~\citep{Fortunato2018NoisyNF}, and other RL specific advances such as double $Q$-learning~\citep{van2016deep} and disambiguation of state-action value estimates~\citep{wang2016dueling}. We will show how similar performance improvements can be obtained by focusing on the neural training aspects via changes to the critic modelling the state-action value,
instead of taking an RL centric view.

\subsection{Challenges with optimisation in DRL}

DRL presents its own unique set of optimisation challenges within the deep learning family.
One challenge stems from non-stationarity and the lack of a unique objective throughout training:
due to the expectation under the policy in loss functions, the loss function itself changes as training progresses; in this aspect, RL is similar to GANs, as the GAN loss also changes as the distribution induced by the generator changes.
In DRL the i.i.d. assumption required to obtain an unbiased estimate of the gradient (see Section~\ref{sec:stochastic_opt}) is also often violated, as data instances might come from the same episode; furthermore it can be shown that in RL the parameter updates might not form a gradient vector field~\citep{czarnecki2019distilling,bengio2020InterferenceAG}.
Both of these aforementioned issues are mitigated, but not fully resolved, through the use of replay buffers, which accumulate transition pairs required to perform the $Q$-learning update (Eq~\ref{eq:dqn_loss}) across a long horizon of agent experience. The loss formulation in RL can present an additional challenge: consistency losses where the target value of the function approximator depends on \textit{itself}, as can be seen in Eq~\eqref{eq:dqn_loss}, are harder to optimise than standard regression losses where the regression target is provided by the dataset, as is the case in supervised learning.

Despite these specificities of DRL, the optimisation algorithms used are those in standard deep learning and described in Section~\ref{sec:opt_algo}, which assume i.i.d. data and the existence of a unique function to optimise. Many a practical implication result as a consequence, including sensitivity to hyperparameterameters \citep{Henderson2018DeepRL}, the lack of performance of regularisation methods that have been shown to improve optimisation in supervised learning~\citep{understanding_batch_norm}, the heavy reliance on adaptive optimisers such as Adam and RMSprop compared to supervised learning where similar results can be obtained with gradient descent with momentum~\citep{hessel2018rainbow,hessel2019multi}.

Inspired by the certain similarities between DRL and GANs, such as non-stationarity, together with the observed common challenges such as sensitivity to hyperparameters, we investigate the use of Spectral Normalisation to alleviate some of the optimisation challenges in DRL. In GANs, Spectral Normalisation has been shown to improve sensitivity to hyperparameters such as learning rate ranges~\citep{miyato2018spectral}, and has been integrated in many state of the art algorithms; for a wider overview of the effects of Spectral Normalisation outside RL, we refer the reader to Chapter~\ref{ch:smoothness}.

\subsection{Reinforcement learning environments}
We evaluate agents on the Atari ALE benchmark \citep{bellemare2013arcade}, containing 51 Atari games. The Atari benchmark is often used in DRL, and thus provides useful baselines to serve as a comparison against our methods. Here we use as baselines C51---the CategoricalDQN agent introduced by~\citet{bellemare2017distributional} trained on Atari ALE---and the Rainbow agent~\citep{hessel2018rainbow}---which improves C51 with additional enhancements, providing a strong baseline on this environment. Unless otherwise specified, experiments on Atari use the standard DQN critic architecture and the experimental setting provided by the C51 baseline, shown in the Appendix in Table~\ref{tab:detault_atari_network} and Section~\ref{sec:app-atari}, respectively.

When performing ablation studies using a large setting such as ALE is computationally prohibitive, and we thus use the MinAtar environment \citep{young19minatar} instead. MinAtar reproduces five games from ALE without the visual complexity and has been found to be a good testing ground for reproducing ablations made using ALE \citep{ObandoCeron2020RevisitingRP}. Unless otherwise specified, MinAtar experiments use the critic architecture described in Table~\ref{tab:detault_minatar_network} in the Appendix.

\section{Spectral Normalisation in RL}

Our aim is to investigate whether the application of Spectral Normalisation in DRL can improve performance, as has been observed in other deep learning domains. To do so, we use Spectral Normalisation applied to neural critics in Q learning algorithms; for a description of Spectral Normalisation, see Section~\ref{sec:smoothness_techniques}. The Spectral Normalisation formulation we use is largely based on the one proposed in the original paper~\citet{miyato2018spectral}, with a few alterations: we do not backpropagate through the power iteration operation; we only apply the normalisation if the spectral norm of a layer is larger than 1; for convolutional layers we do not reshape the 4 dimensional filter tensors to a two-dimensional matrix and compute the spectral norm of the newly obtained linear operator but estimate the of the original convolutional operator instead~\citep{gouk2018regularisation}\footnote{This only applies to the RL experiments in this section, and not to any other Spectral Normalisation experiments in the thesis. We also note that the majority of the results in this section apply normalisation to linear layers, yielding them comparable with the other presented results.}. 
Code available to reproduce our results is available at \url{https://github.com/floringogianu/snrl}.

\subsection{Applying Spectral Normalisation to a subset of layers}

When we started experimenting with applying Spectral Normalisation to the DQN critic, we applied Spectral Normalisation to all layers, as is standard in GANs~\citep{miyato2018spectral}. We quickly noticed, however, that imposing the hard smoothness constraint on the state-action value function to be 1-Lipschitz can substantially decrease performance. As we have observed in Section~\ref{sec:too_much_smoothness}, enforcing a 1-Lipschitz constraint can reduce a model's capacity to capture the decision surface and decrease performance (Figure~\ref{fig:two_moons_decision_surface_sn});  relaxing the constraint by forcing the model to be 10-Lipschitz can mitigate this loss of performance (Figure~\ref{fig:two_moons_decision_surface_sn_10}). Similar observations have been made by \citet{gouk2020regularisation}, who propose introducing a hyperparameter that determines the strength of the Lipschitz constraint. 
We take a different approach here, where instead of changing the Lipschitz constant as a hyperparameter, we choose to only normalise certain layers of the network, most often just one. In particular we observe that normalising the last layer of the network is detrimental. This suggests that imposing a smoothness constraint on the $Q$ function is prohibiting the model from approximating the required decision surface. In the case of Atari games we study here  this is not entirely unexpected, since similar input states can lead to a drastically different game outcome. The expected reward of a particular action $a$ can be determined by a few pixels, such as the location of the ball and paddle in \textit{Pong}, and that of the ghost in \textit{MS-PacMan}. Thus, even for states $s, s'$ such that $||s - s'||$ is small, $||Q(s, a; \vtheta) - Q(s', a; \vtheta)||$ can be large, making the optimal $Q$ function non-smooth.

Instead of normalising the entire network, we instead normalise only a few layers.
Experimentally we found that Spectral Normalisation applied to a DRL critic performs best on the layer with the largest number of parameters; for the standard DQN architecture this is the penultimate layer~\citep{mnih2015human,hessel2018rainbow}. We thus denote the index of the Spectral Normalised layers from the bottom of the network, i.e. \texttt{SN[-1]} will refer to a critic where Spectral Normalisation is applied to the output layer and \texttt{SN[-2]} to the penultimate layer, while \texttt{SN[-2,-3]} refers to a network where Spectral Normalisation is applied both the penultimate layer and the one preceding it.

\subsection{Experimental results on the Atari benchmark}

\begin{figure}[tb!]
  \begin{subfloat}[Average HNS in training.]{
  \includegraphics[width=0.45\columnwidth,valign=b]{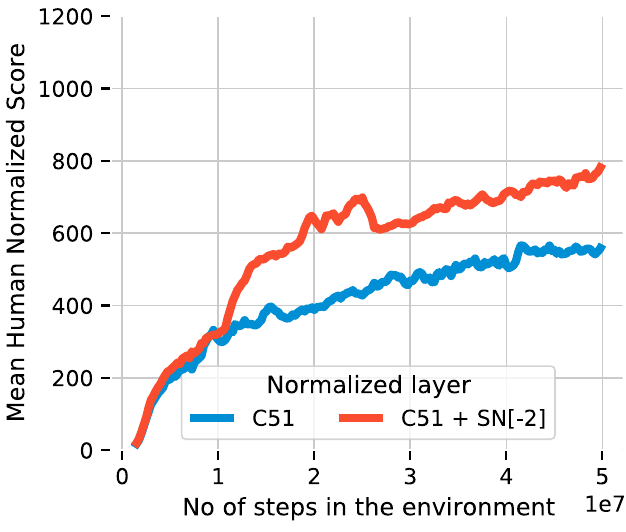}
    \label{fig:c_51_icml_ale_g54_hns_cropped}
  }    
  \end{subfloat}%
  \begin{subfloat}[Mean and Median HNS.]{
        \begin{small}
           \adjustbox{valign=b}{\begin{tabular}{lcccr}
                \toprule
                \textsc{Agent}                                  & \textsc{Mean}             & \textsc{Median} \\
                \midrule
                \textsc{DQN} \cite{wang2016dueling}             & \textsc{216.84}           & \textsc{78.37}  \\
                \textsc{DQN-Adam$^{\ast}$}                      & \textsc{358.45}           & \textsc{119.45} \\
                \textsc{DQN-Adam SN[-2]}                        & \textsc{\textbf{719.95}}  & \textsc{\textbf{178.18}} \\
                \hline%
                \rule{0pt}{2ex}%
                \textsc{C51} \cite{hessel2018RainbowCI}         & \textsc{523.06}           & \textsc{146.73} \\
                \textsc{C51} \cite{bellemare2017distributional} & \textsc{633.49}           & \textsc{174.84} \\
                \textsc{C51$^{\ast}$}                           & \textsc{778.68}           & \textsc{182.26} \\
                \textsc{Rainbow} \cite{hessel2018RainbowCI}     & \textsc{855.11}           & \textsc{227.05} \\
                \textsc{C51 SN[-2]}                             & \textsc{\textbf{1073.18}} & \textsc{\textbf{248.45}} \\
                \bottomrule
            \end{tabular}
          }
        \end{small}
    \label{fig:c_51_icml_ale_g54_hns_cropped_avg}
  }\end{subfloat}%
  \caption[Adding Spectral Normalisation to C51 improves the agent's performance on Atari.]{Selectivly adding Spectral Normalisation to RL agents improves their performance.
        \subref{fig:c_51_icml_ale_g54_hns_cropped}: \textit{Average} Human Normalised Score (HNS) per time-step for C51 with Spectral Normalisation.  
        \subref{fig:c_51_icml_ale_g54_hns_cropped_avg}: Mean and median Human Normalised Score on \dqngames{} Atari games with random starts evaluation. We observe that applying Spectral Normalisation to the penultimate layer of the critic leads to a significant performance boost both for DQN and C51, and rivals algorithm improvements obtained by \textsc{Rainbow}. References indicate the sources for the scores for each algorithm.
        We mark our own implementations of the baseline with $^*$. Agents are evaluated with the protocol in \citep{hessel2018RainbowCI}; all results use an average over \catgames{} ALE games. }
    \label{fig:rl_main_atari_results}
\end{figure}
We assess the effect of applying Spectral Normalisation to the critic both for the standard DQN as well as its distributional counterpart (C51) when trained on the ALE benchmark containing 54 Atari games and show results in Figure~\ref{fig:rl_main_atari_results}. 
We consistently observe that \texttt{SN[-2]} improves over the baseline significantly. Furthermore, \texttt{SN[-2]} provides a significant improvement over  Rainbow~\cite{hessel2018RainbowCI}, a competitive RL agent based on C51 combining multiple competitive RL specific techniques.
We further show in Figure~\ref{fig:atari_dqn_ale_g54_hns} that \texttt{SN[-2]} leads to better results when trained longer, as opposed to the un-normalised DQN which plateaus earlier in training, and towards the end of training has a decrease in performance. Figures~\ref{fig:atari_dqn_ale_g54_hns} and~\ref{fig:min_atar_dqn_ale_g54_hns} show that, while normalising the layer with the largest number of parameters---\texttt{SN[-2]}---performs best, normalising other layers also significantly improves performance over the baseline DQN agent. For all experiments in Figure~\ref{fig:rl_main_atari_results}, when applying Spectral Normalisation we use the \textit{same hyperparameters as the baseline}, without retuning hyperparameters. These results suggest that Spectral Normalisation can be added to existing performant DRL training configurations without requiring a new hyperparameter search.

\begin{figure}[tb!]
    \centering
    \begin{subfloat}[Atari environment.]{
    \includegraphics[height=0.38\columnwidth]{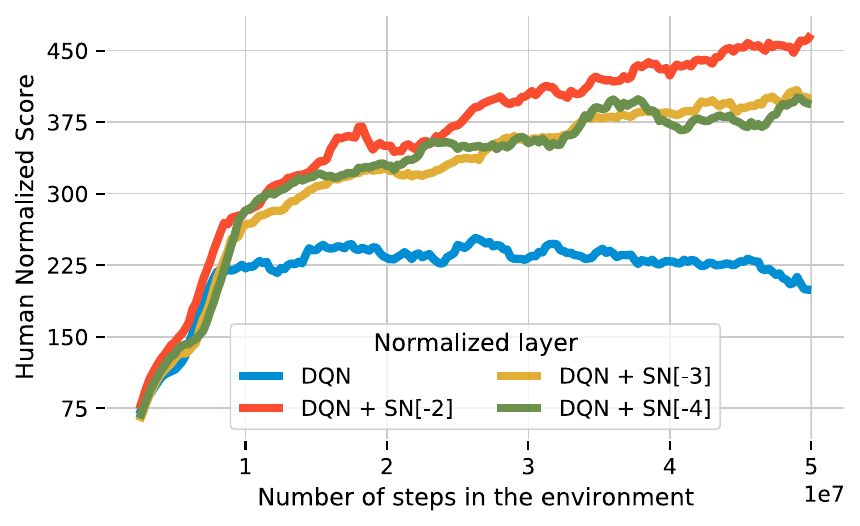}%
    \label{fig:atari_dqn_ale_g54_hns}
    }\end{subfloat}%
    \begin{subfloat}[MinAtar environment.]{
        \includegraphics[height=0.38\columnwidth]{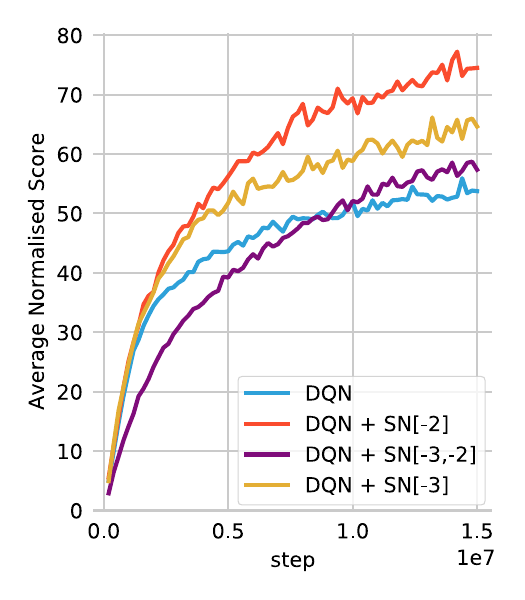}%
    \label{fig:min_atar_dqn_ale_g54_hns}
    }\end{subfloat}%
    \caption[Selectively applying Spectral Normalisation to one layer leads to significant improvements over a DQN agent trained with Adam.]{Selectively applying Spectral Normalisation to one layer leads to significant improvements over a DQN trained with Adam. While the baseline plateaus early in training, when applying Spectral Normalisation the agents continue to improve throughout training. \subref{fig:atari_dqn_ale_g54_hns}: Human Normalised Score for a DQN-Adam baseline with Spectral Normalisation applied on three different layers. Average over \dqngames{} Atari games.
        \subref{fig:min_atar_dqn_ale_g54_hns}: On a $15$M steps training on MinAtar; each line is an average of 10 seeds over four MinAtar games and four different model sizes.}
    \label{fig:atari_dqn_hns}
\end{figure}
\begin{figure}[tb!]
    \centering
    \includegraphics[width=\textwidth]{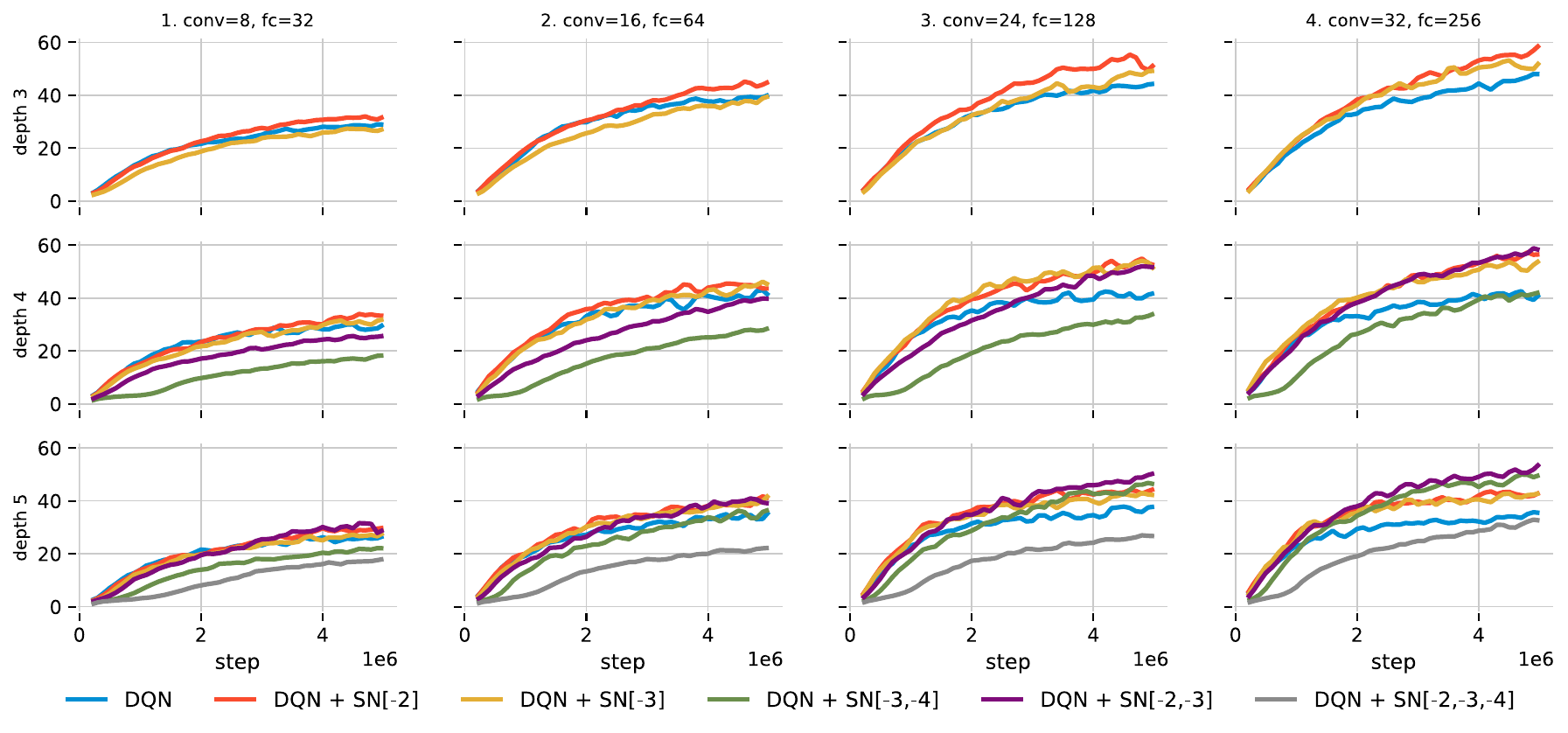}
    \caption[Spectral Normalisation shows gains for all model sizes when applied to a DQN agent on MinAtar.]{\textbf{Spectral Normalisation shows gains for all model sizes.} Looking at the baseline (\legend{solid,line width=1.5pt,538blue} DQN), we observe two performance regimes on MinAtar: for shallow, depth 3 models, performance increases with the width of the model; for deeper models performance generally stagnates with increasing depth and width. In both regimes applying SN on individual (\legend{solid,line width=1.5pt,538red}, \legend{solid,line width=1.5pt,538yellow}) or multiple (\legend{solid,line width=1.5pt,538purple}) layers improves upon the baseline suggesting a regularisation effect we could not reproduce with other regularisation methods. Notice that the strong regularisation resulted from applying Spectral Normalisation to input layers (\legend{solid,line width=1.5pt,538green}) or too many layers (\legend{solid,line width=1.5pt,538grey}) can however degrade performance. Each line is an average over normalised scores of 4 games $\times$ 10 seeds. }
    \label{fig:minatar_SN_depth_vs_width}
\end{figure}
In order to assess the performance of Spectral Normalisation beyond the standard DQN network, we perform a sweep over architectures and consistently observe that Spectral Normalisation improves performance when applied to a range of architectures, as shown in Figure~\ref{fig:minatar_SN_depth_vs_width}; here we use 12 different model sizes of three different model depths and four different layer widths.

\subsection{An optimisation effect}
\label{sec:rl_spectral_schedulers}

While selectively applying Spectral Normalisation to a subset of layers of a $Q$-learning critic is a simple intervention, we have seen in the previous section that it has a strong effect on performance. We now aim to understand \textit{why} Spectral Normalisation leads to these improvements. We first start with the smoothness hypothesis, considering whether the performance improvement is due to an increased smoothness of the critic. We note, however, that unlike other applications of Spectral Normalisation which normalise the entire network, selectively applying Spectral Normalisation to a subset of layers as we do here does not lead to any guarantee regarding the smoothness of the entire critic. Indeed, since there are no constraints on the other layers, the smoothness of those layers could increase throughout training to compensate for the constraint on the normalised layers. When aggregating over the games in MinAtar  we observe $75\%$ of applications of \texttt{SN[-2]} decrease the smoothness of the model---as approximated by the maximum norm of the critic's Jacobian with respect to inputs over a large set of state-action pairs, for a connection between input Jacobians and Lipschitz smoothness, see Section~\ref{sec:measuring_smoothness}---when controlling over hyperparameters; for certain games it is as high as 90\%, while for others as low as $58\%$. We also observe that applying Spectral Normalisation to more layers do not always increase smoothness in Table~\ref{tab:smoothness_sn_applied_performance_rl}.  Furthermore, we notice that while applying Spectral Normalisation increased performance for all games in MinAtar, there is not a strong effect between smoothness and performance as measured by the Spearman-rank correlation between the smoothness of the learned model and the best performance obtained by the agent in all games, as we show in Table~\ref{tab:smoothness_correlate_performance_rl}. To further assess the effect of the critic smoothness on performance, we used gradient penalties as a soft constraint to encourage smoothness (see Section~\ref{sec:smoothness_techniques} for more details on gradient penalties), which did not result in any significant benefit compared to the baseline, as we show in Figure~\ref{fig:rl_minatar_regularisation} in the Appendix.

\begin{table}[t]
    \vskip 0.15in
    \begin{center}
        \begin{small}
            \begin{tabular}{c cccc}
                    &    & \multicolumn{3}{c}{Less or no normalisation}  \\
                \cline{3-5}
                   &     & \multicolumn{1}{|c}{No normalisation} & \multicolumn{1}{|c}{\textsc{SN[-2]}}& \multicolumn{1}{|c|}{\textsc{SN[-3]}}   \\
                \cline{2-5}
                \cline{2-5}
                \parbox[t]{2mm}{\multirow{3}{*}{\rotatebox[origin=c]{90}{More SN}}}  &  \multicolumn{1}{|c}{\textsc{SN[-2]}} & \multicolumn{1}{|c|}{74.6\%} & \multicolumn{1}{c|}{\xmark} & \multicolumn{1}{c|}{\xmark}  \\
                \cline{2-5}
                & \multicolumn{1}{|c}{\textsc{SN[-3]}} &\multicolumn{1}{|c|}{69.2\%}  & \multicolumn{1}{c|}{\xmark} &  \multicolumn{1}{c|}{\xmark}\\
                \cline{2-5}
                & \multicolumn{1}{|c}{\textsc{SN[-2, -3]}} &  \multicolumn{1}{|c|}{75\%} & \multicolumn{1}{c|}{57\%} &  \multicolumn{1}{c|}{60\%}\\
                \cline{2-5}
                \cline{2-5}
            \end{tabular}
        \end{small}
    \end{center}
    \caption[The percentage of training settings for which applying Spectral Normalisation to more layers leads to smoother networks, as measured by the model's Jacobian at peak performance.]{The percentage of training settings (architectures and seeds) for which applying Spectral Normalisation to more layers leads to smoother networks, as measured by the model's Jacobian at peak performance. Selectively applying Spectral Normalisation to only a subset of layers does not consistently lead to an increase in smoothness, and neither does applying Spectral Normalisation to two layers compared to only one layer.
    Results are obtained from a sweep over architectures detailed in Table~\ref{tab:rl_architectures_sweep}, with 10 seeds each. 
    }
    \label{tab:smoothness_sn_applied_performance_rl}
\end{table}

\begin{table}[t]
    \vskip 0.15in
    \begin{center}
        \begin{small}
            \begin{tabular}{lc}
                \toprule
                \textsc{Game}   & \textsc{Spearman Rank}\\
                \midrule
                \rule{0pt}{2ex}%
                Asterix         & \textsc{0.129}\\
                Breakout        & \textsc{0.199}\\
                Seaquest        & \textsc{0.151}\\
                Space Invaders  & \textsc{0.453}\\
                \bottomrule
            \end{tabular}
        \end{small}
    \end{center}
    \caption[Spearman Correlation coefficient between the negative norm of the Jacobian and peak performance for each game.]{
        Spearman Correlation coefficient between the negative norm of the Jacobian (our measure of decreased smoothness) and peak performance for each game. Higher means more correlation. We don't consistently observe a strong correlation between performance and increased smoothness across games.
    }
    \label{tab:smoothness_correlate_performance_rl}
\end{table}

The above evidence suggests that the increased performance from the application of Spectral Normalisation to certain critic layers in DRL might not be fully explainable due to an increased smoothness effect. To better understand why Spectral Normalisation helps DRL, we turn our attention to the possibility of an increase in performance due to an improvement in optimisation; we have already seen in Section~\ref{sec:opt_smoothness_int} that Spectral Normalisation can interact with learning rate and momentum decay rates in supervised learning and in GANs. In RL, we show in Figure~\ref{fig:atari_dqn_hns} that using Spectral Normalisation allows for longer training without early plateaus, and that applying Spectral Normalisation to existing choice of hyperparameters leads to an increase of performance without requiring an additional hyperparameter sweep, as shown in the results in Figure~\ref{fig:rl_main_atari_results}. To investigate whether Spectral Normalisation has an effect  on hyperparameter sensitivity of DQN agents, we perform a large sweep over Adam hyperparameters learning rate $h$ and $\epsilon$ (for an overview of Adam see Section~\ref{sec:opt_algo}) and show that indeed, applying Spectral Normalisation increases the range of performant hyperparameters significantly; results are shown in Figure~\ref{fig:hypersensitivity_rl}.

\begin{figure}[tb]
    \centering
    \includegraphics[width=0.99\columnwidth]{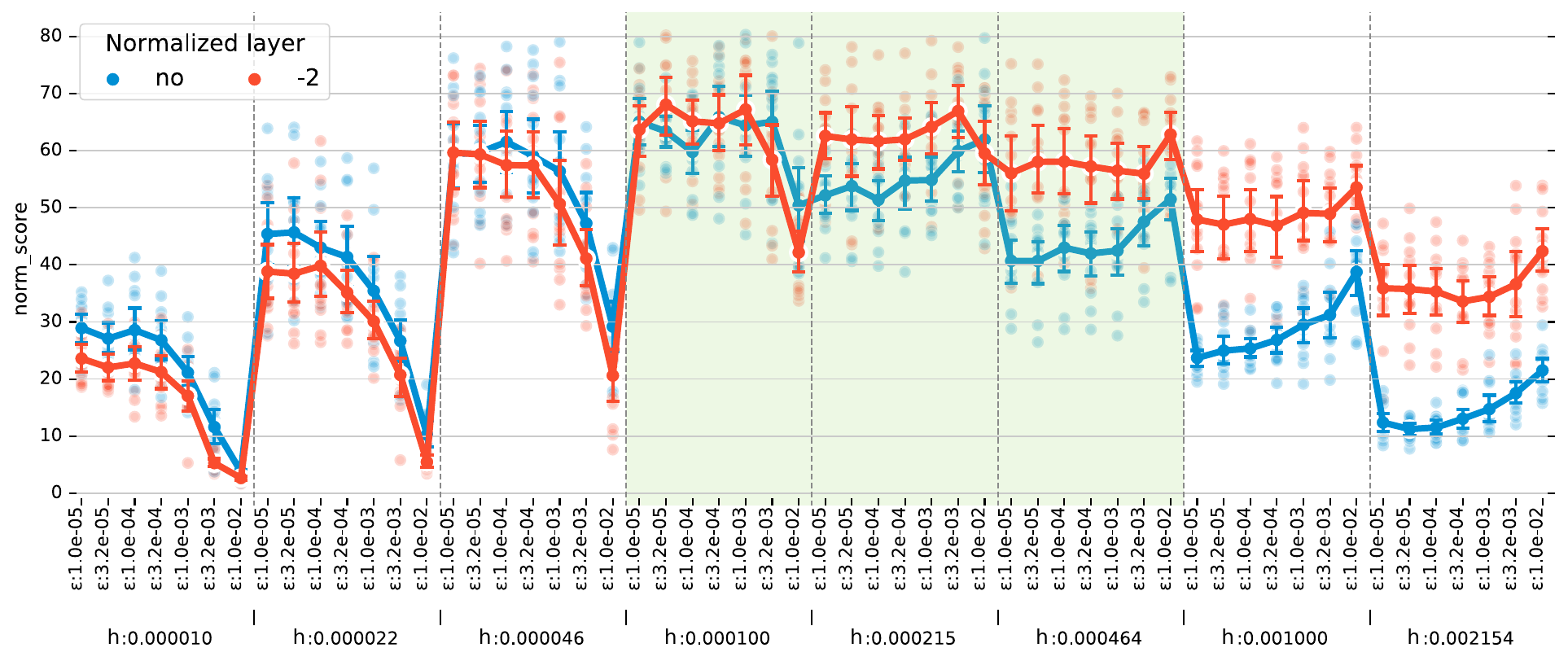}
    \caption[Spectral Normalisation can significantly decrease hyperparameter sensitivity of DQN when the Adam optimiser is used.] {\textsc{SN[-2]} applied to DQN, Spectral Normalisation applied to the largest layer in the DQN critic, can significantly decrease sensitivity to the learning rate and $\epsilon$ hyperparameter of Adam. The cross product of learning rates in $h \in \{0.00001, ..., 0.00215\}$ and in $\epsilon \in \{0.00001, ..., 0.01\}$ is explored.}
    \label{fig:hypersensitivity_rl}
\end{figure}

In order to further investigate the effect of Spectral Normalisation in DRL training, we develop a set of methods which independently assess its effects. To this end, consider the neural network to which Spectral Normalisation is applied, which we assume has $N$ layers, with weights $\vW_i \in \mathbb{R}^{d_{i+1} \times d_{i}}, \vb_i \in \mathbb{R}^{d_{i+1}}$:
\begin{align}
  \vo & = \vW_N \vh_{N-1} + \vb_{N} \\
  \vh_{N-1} &= a(\vz_{N-1}) = a(\vW_{N-1}\vh_{N-2} +  \vb_{N-1}) \\
  & \cdots \\
  \vh_{1} &= a (\vz_{1}) = a(\vW_{1} \vh_{0} +  \vb_{1}),
\end{align}
with $\vh_{0} = \vx$ and $a$ is an activation function applied element-wise to the input vector. The activation $a$ has to be a rectifier, such as ReLU~\citep{nair2010rectified} or Leaky ReLU~\citep{xu2015empirical}, since we rely on $a(\sigma_l \vx) = \sigma_l a(\vx)$, with $\sigma_l > 0$. We can write our optimisation problem as:
\begin{align}
\min_{\vtheta} E(\vtheta) = E(\vo;\vtheta = \{\vW_{1 \le i \le N}, \vb_{1 \le i \le N}\}).
\end{align}
We assess the effect of applying Spectral Normalisation to layer $l$,  but note the same derivation applies when multiple layers are normalised. We note that for simplicity we assumed above that the model is an MLP, though the same computation carries for convolutional layers.
The analysis that follows relies on the following observation:
\begin{remark} In a neural network with recitifer activation functions, if the bias term of a layer where Spectral Normalisation is applied gets adjusted to account for the normalisation, Spectral Normalisation can be seen as scaling the output of the layer by the inverse of the spectral norm $\sigma_l$.
\end{remark}

To see why, consider $\vh_{l+1} = a(\vW_{l} \vh_{l+1} + \vb_{l})$ and its normalised counterpart:  $\vh^{'}_{l+1} = a(\vW_{l}/\sigma_l \vh_{l+1} + \vb_{l})$. Since we chose the activations function such that $a(\vh/\sigma_l) = a (\vh)/\sigma_l$, we observe that if we also scale the bias $\vb_l$ by $1/\sigma_l$, we can obtain $\vh^{'}_{l+1} = \vh_{l+1}/\sigma_l$. Furthermore, if we apply the same bias scaling to subsequent layers $\vb_{i} \rightarrow \vb_{i} /\sigma_l, \forall i \ge l$, we obtain that $\vh'_i = \vh_i/\sigma_l$, including $\vo' = \vo/\sigma_l$. Thus \textit{if we assume that the bias vectors can be scaled by a constant in training}, the effect of Spectral Normalisation is to scale each activation following the normalisation by the inverse of the spectral norm of the normalised layer; if multiple layers are normalised, activations are scaled by inverse of the product of the spectral norm of the normalised layers. We note that the bias scaling assumption does not change how expressive the model can be, as the bias term could learn such as scaling through training. Indeed, we test this assumption empirically and use a procedure we term \textsc{divOut}, in which instead of applying Spectral Normalisation, we divide the model output by the product of the spectral norms of the normalised layers $\mathcal{\rho}$:
\begin{align}
\textsc{divOut}[\mathcal{\rho}]: \hspace{1em} \min_{\vtheta} E(\vtheta)= E(\frac{\vo}{\prod_{s \in \mathcal{\rho}}{\sigma_s}}; \vtheta ).
\end{align}

We investigate where \textsc{divOut} can perform similarly to its Spectral Normalisation counterparts and show in Figure~\ref{fig:spectral_schedulers_rl} that in offline reinforcement learning $\textsc{divOut}[\mathcal{\rho}]$ and $\texttt{SN}[\mathcal{\rho}]$ have similar behaviours.

\begin{figure}[t]
    \centering
    \begin{subfloat}[One normalised layer.]{
       \includegraphics[height=0.27\textwidth]{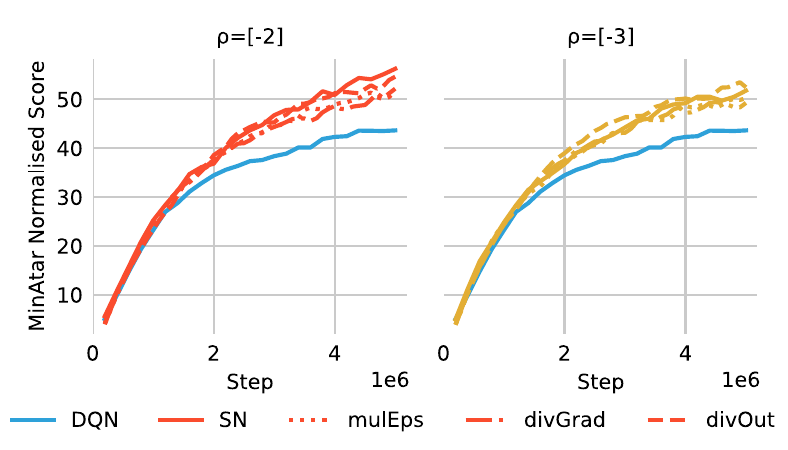}%
       \label{fig:mul_eps_1}
    }
    \end{subfloat}%
    \begin{subfloat}[Multiple normalised layers.]{
          \includegraphics[height=0.27\textwidth]{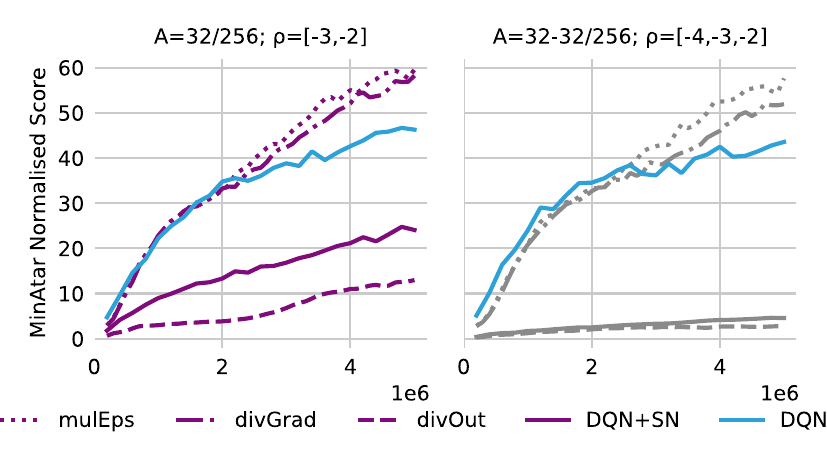}%
       \label{fig:mul_eps_2}
    }
    \end{subfloat}%
    \caption[The effect of \textsc{divOut}, \textsc{divGrad} and \textsc{mulEps} on RL performance on MinAtar.]{The effect of \textsc{divOut}, \textsc{divGrad} and \textsc{mulEps} on RL performance on MinAtar. \textit{Despite not changing the model, but only changing the optimisation procedure based on the effect Spectral Normalisation has on parameter updates}, \textsc{divGrad} and \textsc{mulEps} perform comparable to, or better than, Spectral Normalisation. \subref{fig:mul_eps_1} shows results obtained from simulating optimisation effects of Spectral Normalisation when applied to one layer, where \textsc{divGrad} and \textsc{mulEps} and \textsc{divOut} obtain similar performance performance to Spectral Normalisation, and significantly improve over the baseline. \subref{fig:mul_eps_2} shows results obtained when looking at the optimisation effects of applying Spectral Normalisation to multiple layers; here Spectral Normalisation hurts performance, and so does \textsc{divOut}, but \textsc{divGrad} and \textsc{mulEps} still significantly improve performance compared to the baseline. This suggests that large changes model output obtained by Spectral Normalisation and \textsc{divOut} can be detrimental, while optimisation only changes are beneficial in that setting.}
    \label{fig:spectral_schedulers_rl}
\end{figure}

We now turn our attention towards the effects that Spectral Normalisation has on the model gradients. To do so, we would like to compute $\frac{d E}{d \vW_i}$ and $\frac{d E}{d \vb_i}$ for normalised and un-normalised networks. As before, for simplicity we consider the situation of one normalised layer $l$ and \textit{we assume bias scaling}. We then have
\begin{align}
 \vo' = \vo/\sigma_l; \hspace{3em} \vh'_{i} = \vh_{i} /\sigma_l, \hspace{0.5em}\forall i \ge l; \hspace{3em} \vh'_{i} = \vh_{i}, \hspace{0.5em}\forall i < l.
 \label{eq:scaling_sn}
\end{align}
 While the computational graph only changes at layer $l$ where the ${\vh_{l} = a(\vW_{l} \vh_{l-1} +  \vb_{l})}$ is now replaced with
 \begin{align}
 \vh_{l}' = a(\vz_{l}') = a(\vW_{l}/\sigma_l \vh_{l-1} +  \vb_{l}/\sigma_l)= a(\vW_{l}\vh_{l-1} +  \vb_{l})/\sigma_l,
 \label{eq:normalisation_sn}
 \end{align}
the evaluation of the higher level derivatives will also change due to the scaling effect in Eq~\eqref{eq:scaling_sn}. To make the changes in the computational graph due to the normalisation explicit, we will call the loss with the normalisation $E'$. We are interested in computing $\frac{d E'}{d \vW_i}$ and $\frac{d E'}{d \vb_i}$, and comparing it with the backpropagation in an un-normalised model
\begin{align}
\frac{d E}{d \vW_N} &=   \frac{d E}{d \vo} \vh_{N-1} ^T \label{eq:backprop_first}\\
\frac{d E}{d \vh_{N-1}} &= \frac{d \vo}{d \vh_{N-1}}^T \frac{d E}{d \vo} = \vW_{N}^T \frac{d E}{d \vo}  \\
\frac{d E}{d \vW_i}
        &= \frac{d E}{d \vz_{i}} \vh_{i-1}^T =  \left(\frac{d E}{d \vh_{i}} \odot a'(\vz_{i})\right)\vh_{i-1}^T \hspace{2em} i < N \\
\frac{d E}{d \vh_{i-1}} &= \frac{d \vz_{i}}{d \vh_{i-1}}^T \frac{d E}{d \vz_{i}}   = \vW_i^T \left(\frac{d E}{d \vh_{i}} \odot a'(\vz_{i})\right) \hspace{2em} i < N. \label{eq:backprop_last}
\end{align}

When normalisation is applied, the computational graph does not change for the layers following the normalisation compared to the un-normalised case, but the evaluation of the derivatives changes due to the scaling  in Eq~\eqref{eq:scaling_sn}
\begin{align}
& \frac{d E'}{d \vo'}(\vo') = \frac{d E}{d \vo}(\vo / \sigma_l) \label{eq:jacobian_out_change} \\
&\frac{d E'}{d \vW_N} = \frac{d E'}{d \vo'} {\vh'_{N-1}} ^T =  \frac{d E'}{d \vo'} \left( \underbrace{\vh_{N-1}/\sigma_l}_{\eqref{eq:scaling_sn}} \right)^T =  \frac{1}{\sigma_l} \frac{d E'}{d \vo'} \vh_{N-1} ^T \\
&\frac{d E'}{d \vh'_{N-1}} = \vW_{N}^T \frac{d E}{d \vo}(\vo / \sigma_l)   \\
&\frac{d E'}{d \vW_{i}} =\left(\frac{d E'}{d \vh'_{i}} \odot a'(\vz'_{i}) \right){\vh'_{i-1}}^T = \frac{1}{\sigma_l}\left(\frac{d E'}{d \vh'_{i}} \odot a'(\vz_{i}) \right){\vh_{i-1}}^T, \hspace{1em} i > l. \label{eq:normalisation_sn_after}
\end{align}
For the normalised layer $l$, we have that due to Eq~\eqref{eq:normalisation_sn}
\begin{align}
&\frac{d E'}{d \vW_l}
        =  \frac{d E'}{d \vz'_{l}} \underbrace{{\vh'_{l-1}}^T/\sigma_l}_{normalisation, \eqref{eq:normalisation_sn}} = \frac{1}{\sigma_l} \left(\frac{d E'}{d \vh'_{l}} \odot a'(\vz_{l})\right)\vh_{l-1}^T  \label{eq:normalisation_sn_exact_layer}\\
&\frac{d E'}{d \vh'_{l-1}} = \underbrace{\frac{1}{\sigma_l} \vW_l^T}_{normalisation, \eqref{eq:normalisation_sn}} \left(\frac{d E'}{d \vh'_{l}} \odot a'(\vz'_{l})\right) .
\end{align}
For the layers before the normalisation, we have
\begin{align}
\frac{d E'}{d \vW_i}
        &=  \left(\frac{d E'}{d \vh'_{i}} \odot a'(\vz_{i})\right)\vh_{i-1}^T \hspace{2em} i < l \label{eq:eq:normalisation_sn_before}\\
\frac{d E}{d \vh'_{i-1}} &=  \vW_i^T \left(\frac{d E'}{d \vh'_{i}} \odot a'(\vz_{i})\right) \hspace{2em} i < l.
\end{align}
Thus, when comparing the gradients obtained with the normalised network with those in an un-normalised network, we observe two changes: first, a change in the output Jacobian (Eq~\eqref{eq:jacobian_out_change}), and second, a scaling effect which applies to \textit{all layers in the network}, either through the scaling of the activations for layers following the normalisation (Eq~\eqref{eq:normalisation_sn_after}), through the changes in the computational graph for the normalised layer (Eq~\eqref{eq:normalisation_sn_exact_layer}), or through the effect of backpropagation of the scaling of $\frac{d E'}{d \vh'_{l-1}}$, for the layers preceding the normalisation (Eq~\eqref{eq:eq:normalisation_sn_before}).  A similar argument can be made for biases. 
While the change in Jacobian is difficult to obtain without affecting the model as we have done with \textsc{divOut}, to capture the gradient scaling effect we propose \textsc{divGrad}, a normalisation scheme based on the effects of applying Spectral Normalisation which \textit{does not change the model, but only changes the optimisation procedure}, by scaling the gradients of the parameters:
\begin{align}
\textsc{divGrad}[\mathcal{\rho}]: \hspace{1em} \frac{d E}{d \vtheta_i} \rightarrow \frac{1}{\prod_{i \in \mathcal{\rho}} \sigma_i}\frac{d E}{d \vtheta_i}.
\label{eq:div_grad}
\end{align}
Results in Figure~\ref{fig:spectral_schedulers_rl} show that \textsc{divGrad} performs well and that it can outperform Spectral Normalisation or \textsc{divOut} on MinAtar; further results supporting this hypothesis are shown in Figure~\ref{fig:schedulers_detailed} in the Appendix.

The above observation regarding gradient scaling also allows us to intuit the effect of Spectral Normalisation in conjunction with optimisers like Adam (for a detailed description of the Adam optimiser, see Section~\ref{sec:opt_algo}). If we consider that the spectral norm of a weight matrices changes slowly in training (which we verify empirically in Figure~\ref{fig:dqn-minatar-all-radii-2}), \textit{under the bias scaling assumption}, the effect of Spectral Normalisation on the Adam update can be seen as
\begin{align}
\vm'_t &= \beta_1 \vm'_{t-1} + (1 - \beta_1) \nabla_{\vtheta}E(\vtheta_{t-1})/ \sigma \approx \vm'_t/\sigma \\
\vv'_t &= \beta_2 \vv_{t-1} + (1 - \beta_2) \nabla_{\vtheta}E(\vtheta_{t-1})^2 \approx \vv'_t/\sigma^2\\
\Delta \vtheta'_t &=  h \frac{\frac{1}{1 - \beta_1^t} \vm'_t}{\sqrt{\frac{1}{1 - \beta_2^t}\vv'_t} + \epsilon} =  h \frac{\frac{1}{1 - \beta_1^t} \vm_t}{\sqrt{\frac{1}{1 - \beta_2^t}\vv_t} + \sigma \epsilon},
\label{eq:mul_eps}
\end{align}
where $\sigma = \prod_{i \in \mathcal{\rho}} \sigma_i$, the product of the spectral norms of the normalised layers and we used element-wise operations. Thus, the effect on the Adam update is reflected in a scaling of the hyperparameter $\epsilon$ by $\sigma$.
To investigate this effect on the Adam optimiser, we introduce \textsc{mulEps}, which changes the Adam update by multiplying $\epsilon$ by the product of the spectral norms of the normalised layers. We show in Figure~\ref{fig:spectral_schedulers_rl} that \textsc{mulEps} performs similarly to selectively applying Spectral Normalisation in DRL, and can improve upon it. A strong effect is particularly observed when \textsc{mulEps} and Spectral Normalisation are applied to multiple layers (Figure~\ref{fig:mul_eps_2}), where Spectral Normalisation \textit{decreases} performance compared to the baseline, while \textsc{mulEps} increases it. Thus, \textsc{mulEps}, provides another example of how we can use Spectral Normalisation to derive optimisation only changes that lead to improvements in performance in DRL.

\section{Revisiting Spectral Normalisation in GANs}
\label{sec:sn_gans_after_rl}

By analysing the effect of Spectral Normalisation in the previous section, we have uncovered multiple effects: model effects such as output scaling (Eq~\eqref{eq:jacobian_out_change}), and optimisation effects such as gradient scaling (Eq~\eqref{eq:div_grad}) and interactions with optimisation hyperparameters such as $\epsilon$ in Adam (Eq~\eqref{eq:mul_eps}). We saw that in DRL the effects of Spectral Normalisation can be recovered through the use of optimisation schemes, such as  \textsc{divGrad} and \textsc{mulEps}, derived based on the optimisation only effects of Spectral Normalisation. Since Spectral Normalisation was first introduced for GANs, and indeed is a key component of many successful GANs~\citep{zhang2019self,biggan}, it is only natural to ask whether the same optimisation methods perform equally well when training GANs. We studied GANs previously and described them in Section~\ref{sec:dd_gan_background}, and the interactions between smoothness normalisation and optimisation in GANs have been discussed in Chapter~\ref{ch:smoothness}. The application of Spectral Normalisation is different in the GAN setting compared to RL, in that in GANs it is applied to \textit{every layer} of the model\footnote{When Spectral Normalisation was introduced, it was only applied to the discriminator, but since it has also been applied to the generator; here, for consistency with the Spectral Normalisation baseline in~\citet{miyato2018spectral}, we apply it only to the discriminator.}. This difference between the DRL and GANs is significant, since applying Spectral Normalisation on every layer ensures the model is Lipschitz (at least when Spectral Normalisation is applied on the linear operator given by convolutional layers and not on the reshaped filters), while applying it to only one layer leads to no such assurances, as we have seen in the previous section. This difference in application of Spectral Normalisation suggests perhaps that the change in the model is important in GANs. To test this intuition, we experimented with optimisation only methods \textsc{divGrad} and \textsc{mulEps} in GAN training. We note that as the generator's backpropagation requires a backward pass through the discriminator, any normalisation applied to the discriminator in \textsc{mulEps} and \textsc{divGrad} will also affect the generator, as was the case with un-normalised early layers in the RL critic (see Eqs~\eqref{eq:mul_eps} and~\eqref{eq:div_grad}).
Results in Figure~\ref{fig:mul_eps} confirm that indeed, methods that only make changes to optimisation dynamics and do not account for model changes induced by Spectral Normalisation do not recover its performance in GAN training.

\begin{figure}[t]
\begin{subfloat}[Spectral Normalisation.]{
\includegraphics[width=0.333\columnwidth]{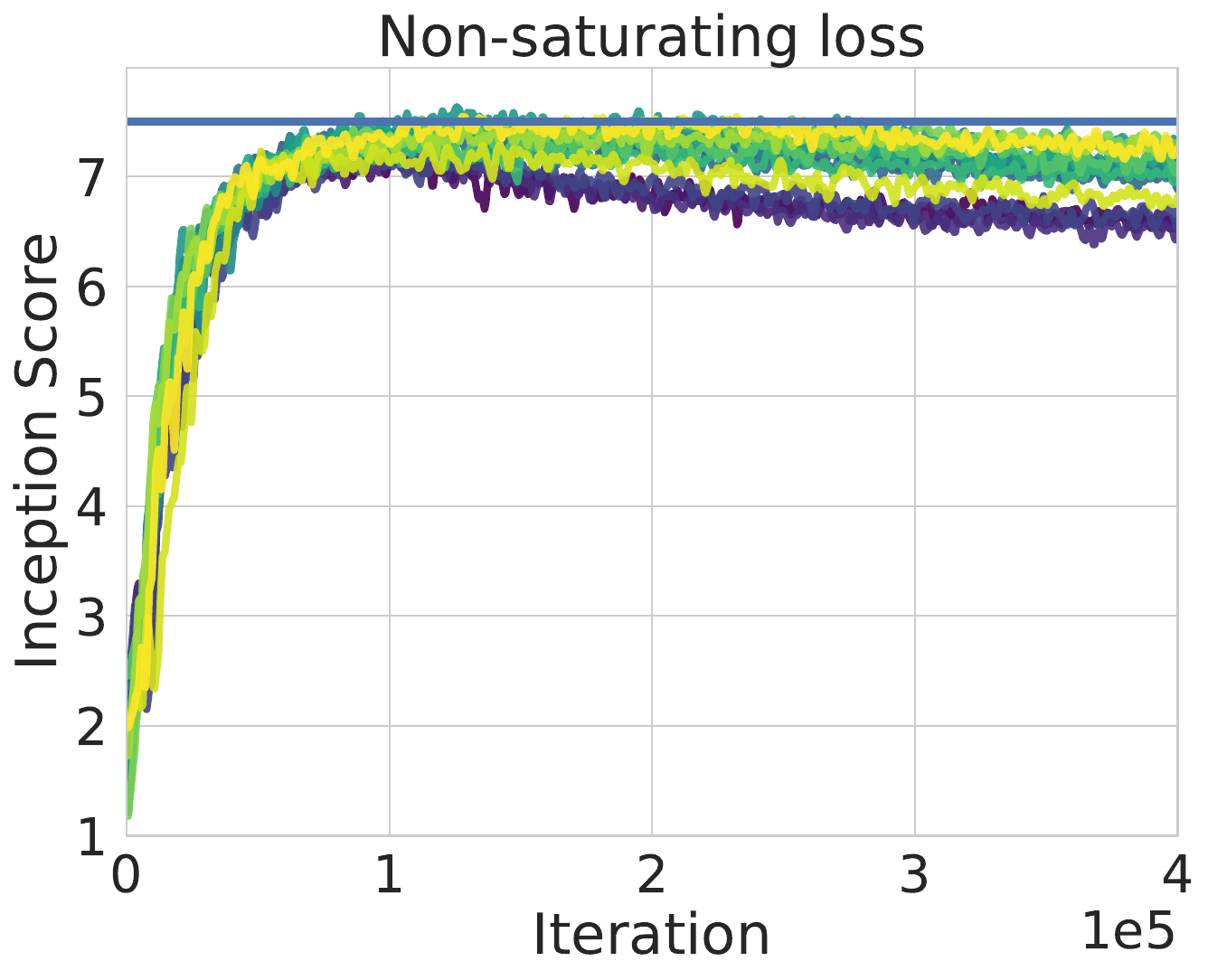}
}
\end{subfloat}%
\begin{subfloat}[No normalisation.]{
\includegraphics[width=0.333\columnwidth]{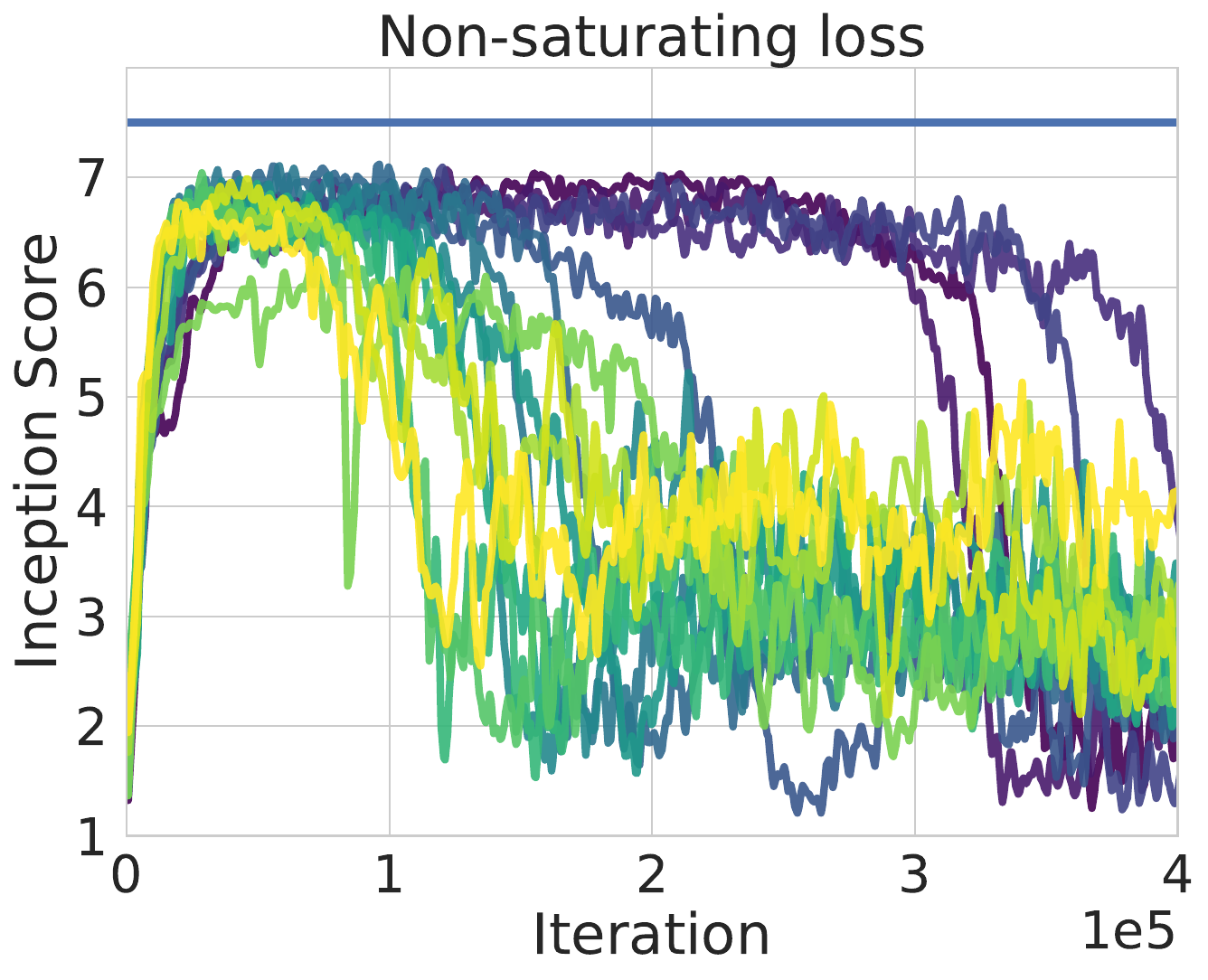}
}
\end{subfloat}%
\begin{subfloat}[MulEps.]{
\includegraphics[width=0.333\columnwidth]{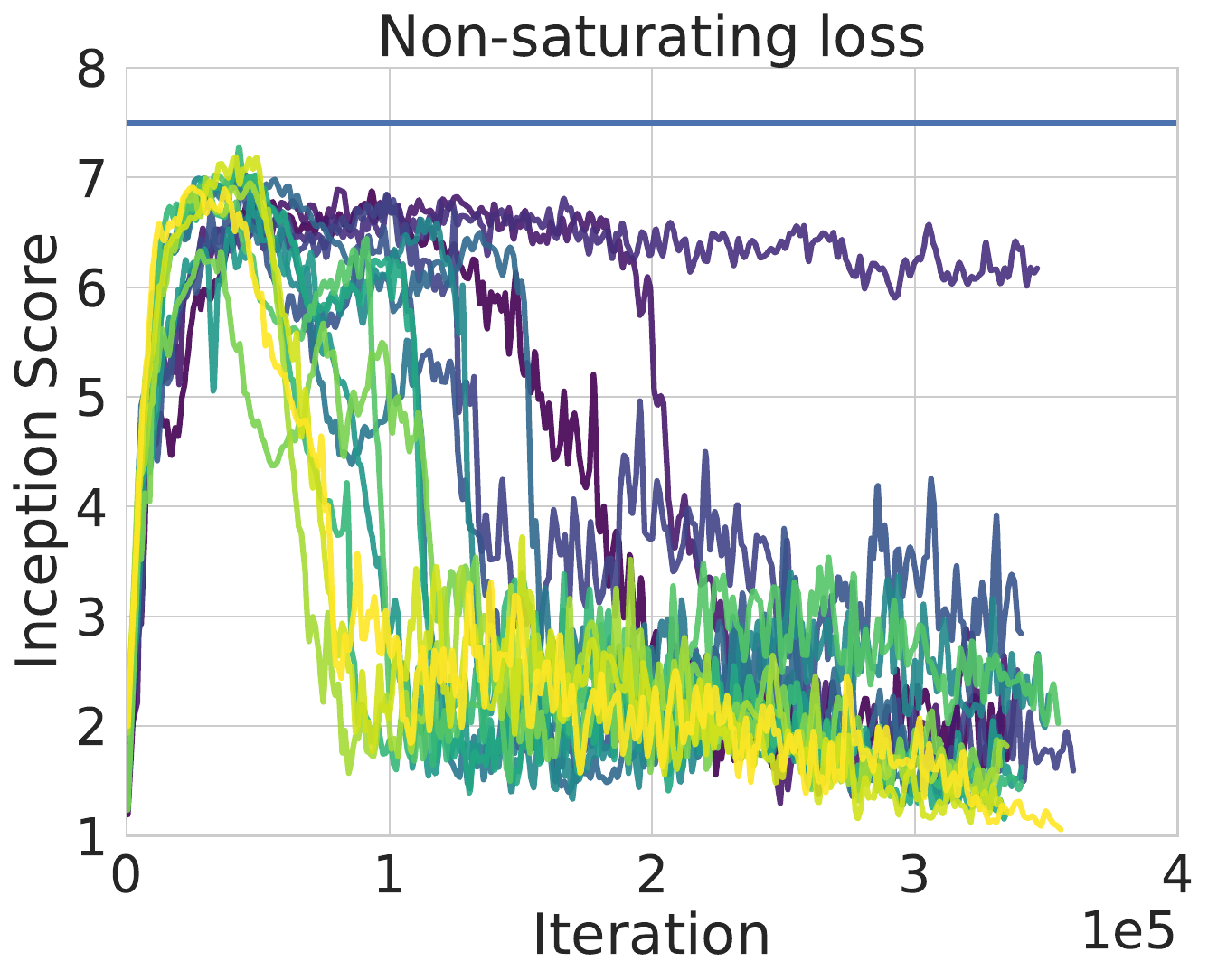}
}
\end{subfloat}
\caption[Optimisation changes based on Spectral Normalisation do not recover the performance of Spectral Normalisation in GANs.]{Optimisation changes based on Spectral Normalisation do not recover the performance of Spectral Normalisation in GANs. Different units indicate different learning rates for the discriminator and generator.}
\label{fig:mul_eps}
\end{figure}

\subsection{Spectral Normalisation and the discriminator output}
Since optimisation only changes do not recover the performance of Spectral Normalisation in GAN training, we now investigate whether Spectral Normalisation alters the loss prediction landscape in a substantial way.
We start by considering the effect of Spectral Normalisation on the range of discriminator outputs. Since Spectral Normalisation applies a \textit{global constraint} on the entire space, this applies 
between pairs of data instances, pairs of samples, and pairs of data and samples. 
If we write $D(\cdot;\vphi) = \varsigma(D_L(\cdot;\vphi))$, with $\varsigma$ being the sigmoid function, Spectral Normalisation ensures that $D_L(\cdot;\vphi)$ is 1-Lipschitz by normalising each layer in the discriminator's network. Given
 a generator sample $G(\vz;\vtheta)$ and a data instance $\vx$, $\norm{D_L(G(\vz;\vtheta);\vphi) - D_L(\vx;\vphi)}_2$ is bound by $\norm{G(\vz;\vtheta) - \vx}_2$.
Thus, while with no normalisation the discriminator has an easier task of separating the data and samples by using the saturating areas of the sigmoid function, this is harder to do when Spectral Normalisation is used;
we visualise this effect in Figure~\ref{fig:sigmoid_effect_sn_no_sn}.

 \begin{figure}[t]
\centering
\begin{subfloat}[No normalisation.]{
\includegraphics[width=0.45\columnwidth]{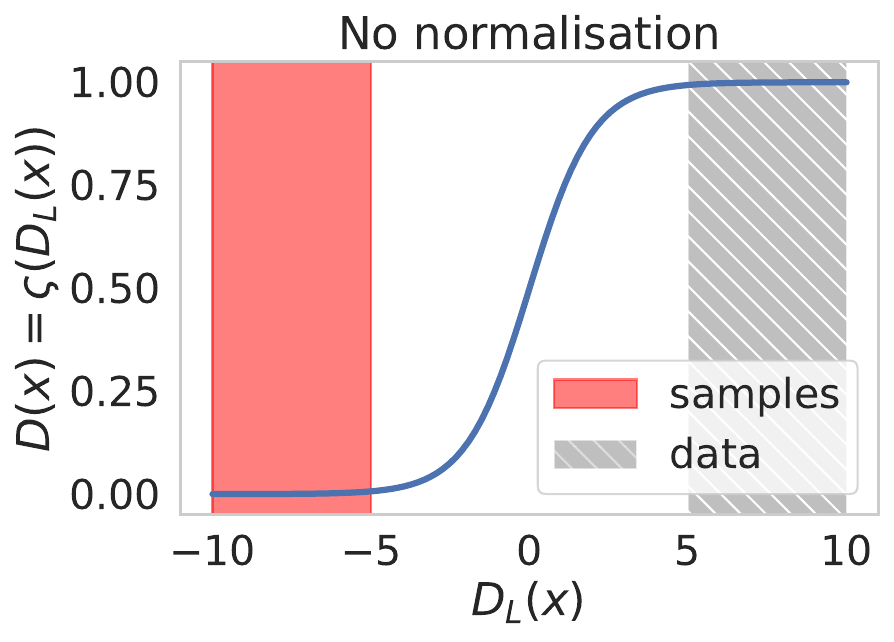}%
\hspace{2em}
}\end{subfloat}%
\begin{subfloat}[Spectral Normalisation.]{
\includegraphics[width=0.45\columnwidth]{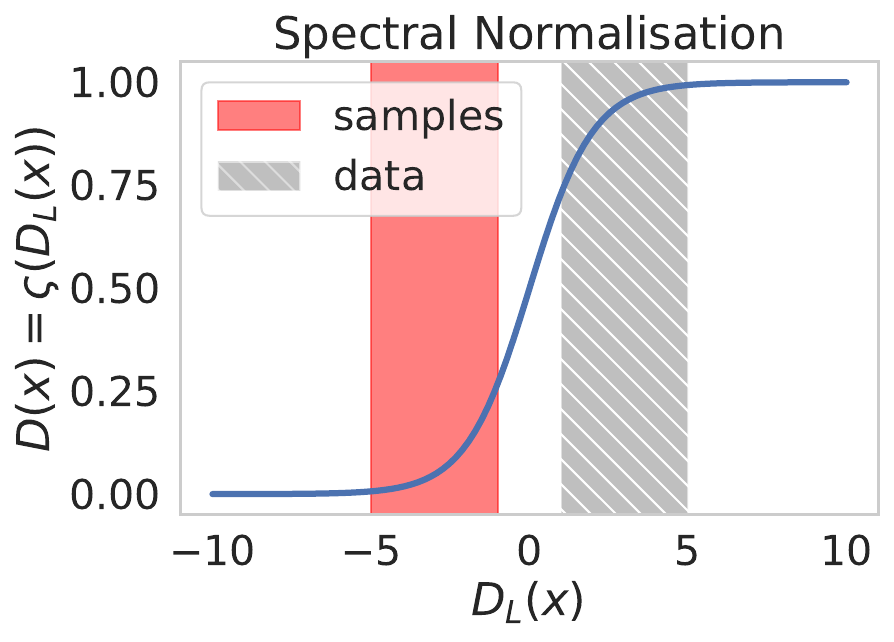}%
}
\end{subfloat}%
\caption[The effect of Spectral Normalisation on discriminator predictions.]{Visualisation of the effect of Spectral Normalisation on discriminator predictions: Spectral Normalisation is making the discriminator output less likely to reach the saturating areas of the sigmoid function, denoted here by $\varsigma$. }
\label{fig:sigmoid_effect_sn_no_sn}
\end{figure}

We now turn to empirically investigating this effect in GAN training. We measure the output of the discriminator throughout training both on samples and data batches and create a binned-histogram of the outputs obtained from a batch. That is, for each $\vx_i$ in a data batch with $D(\vx_i;\vphi) \in [0, 1]$ we assign it the interval $[0.1 p, 0.1 (p+1)]$ with $p \in \{0, ..., 9\}$ if $D(\vx_i;\vphi) \in [0.1 p, 0.1 (p+1)]$; we do the same for samples $D(G(\vz_i;\vtheta);\vphi)$ where $\vz_i$ are the latent vectors used to obtain the sample batch.
This allows us to measure the variability of the discriminator prediction inside data and sample batches; we measure the entropy of the discriminator output across a batch with and without Spectral Normalisation by measuring the entropy of the distribution induced by the histogram. We show aggregate results throughout training over a hyperparameter sweep in Figure~\ref{fig:spectral_norm_entropy}, and individual results with a fixed set of hyperparameters in Figures~\ref{fig:spectral_norm_histogram_examples_early} and~\ref{fig:spectral_norm_histogram_examples}.
 As we intuited, Spectral Normalisation increases the entropy of the discriminator output throughout training, both on data (Figure~\ref{fig:spectral_norm_entropy_data}) and samples (Figure~\ref{fig:spectral_norm_entropy_samples}). Early in training (Figure~\ref{fig:spectral_norm_histogram_examples_early}), discriminators trained with Spectral Normalisation are less likely to assign high probability to real data being real, or to samples being fake. Late in training (Figure~\ref{fig:spectral_norm_histogram_examples}), GANs trained with Spectral Normalisation also exhibit higher entropy of discriminator outputs; this end of training behaviour is more consistent to that of a Nash equilibrium (see Section~\ref{sec:intro_multi_objective_optimisation} for a definition and discussion on Nash equilibria), where 
the discriminator should not be able to distinguish between samples and data. 

Given the effect of Spectral Normalisation on the discriminator's predictions, it's natural to ask how this impacts training. We turn to this question next. 
\begin{figure}[t!]
\centering
\begin{subfloat}[Performance.]{
\includegraphics[width=0.32\columnwidth]{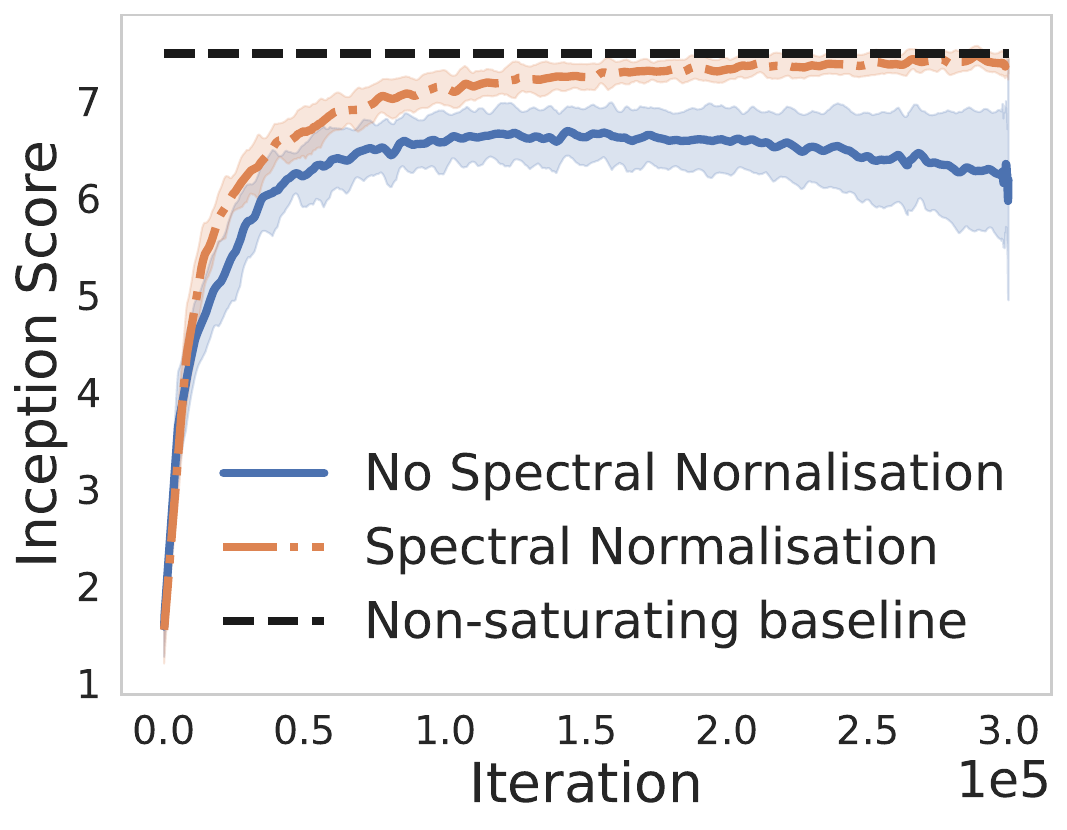}
\label{fig:perf}
}
\end{subfloat}%
\begin{subfloat}[Entropy on data.]{
\includegraphics[width=0.333\columnwidth]{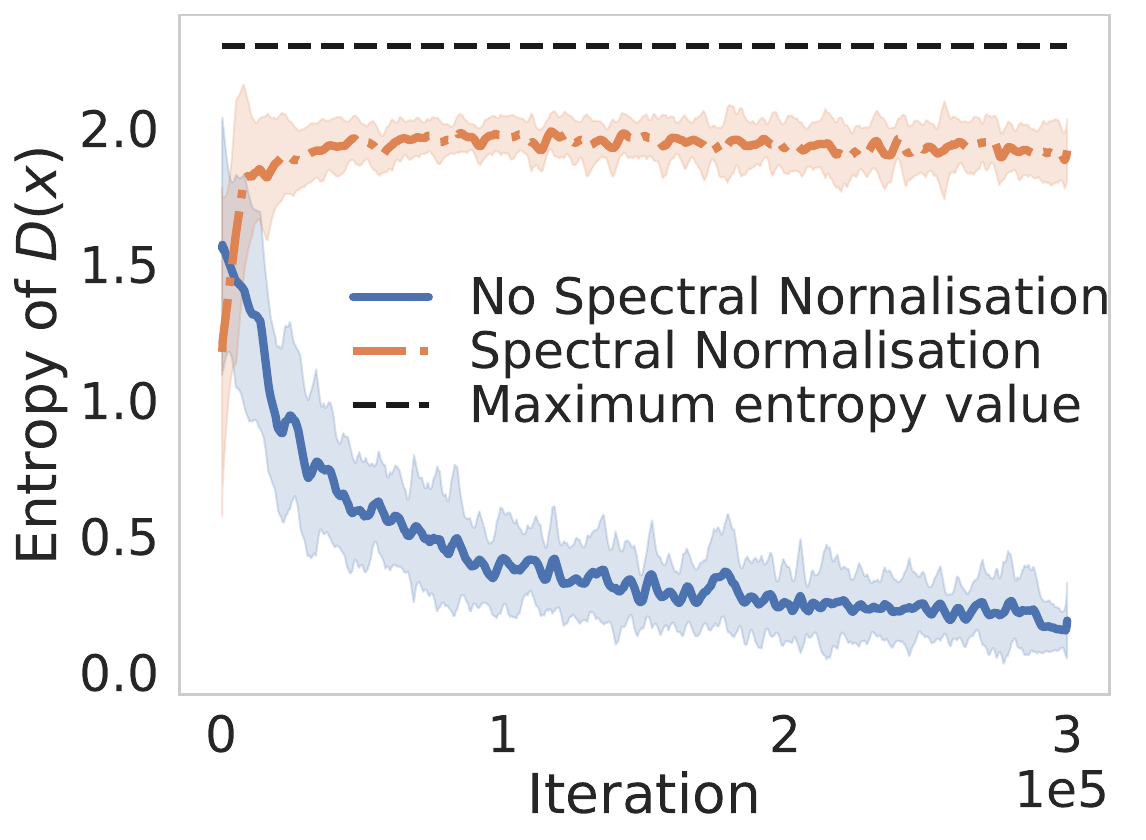}
\label{fig:spectral_norm_entropy_data}
}
\end{subfloat}%
\begin{subfloat}[Entropy on samples.]{
\includegraphics[width=0.333\columnwidth]{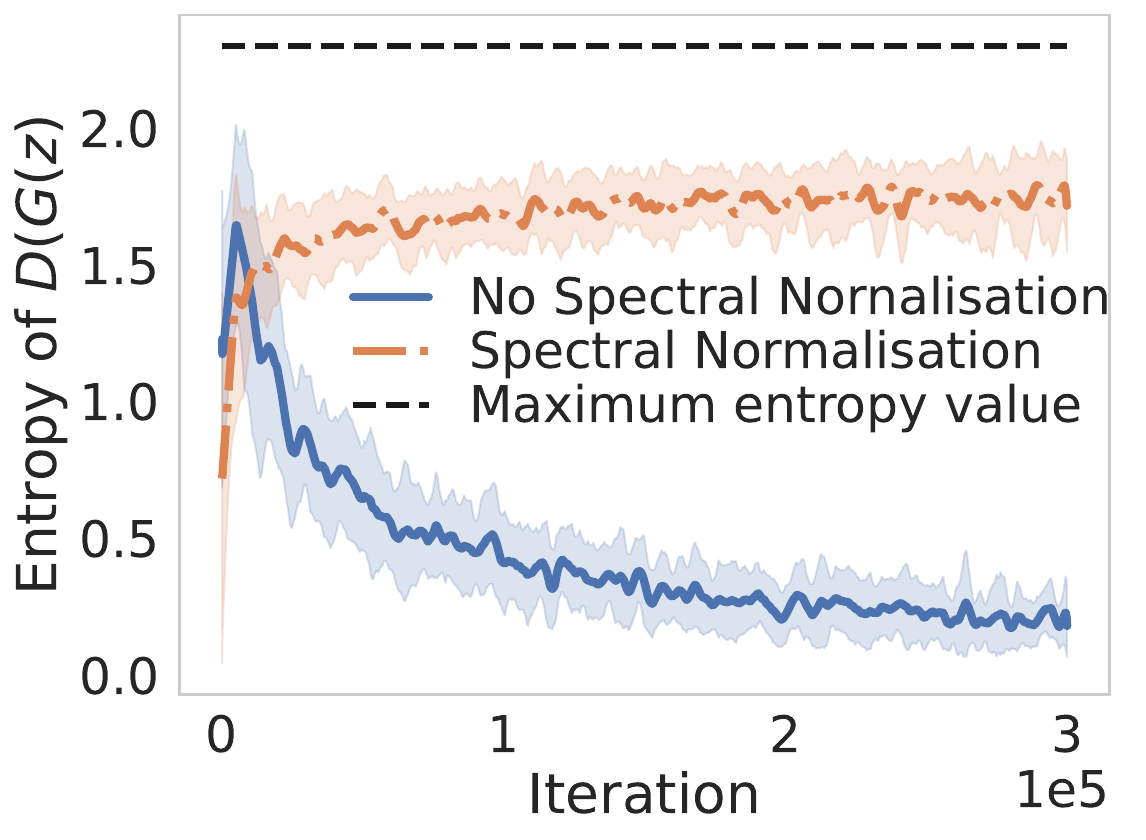}
\label{fig:spectral_norm_entropy_samples}
}
\end{subfloat}%
\caption[Spectral Normalisation increases the entropy of the discriminator's predictions on data and samples throughout training.]{Spectral Normalisation increases the entropy of the discriminator's predictions on data \subref{fig:spectral_norm_entropy_data} and samples \subref{fig:spectral_norm_entropy_samples}. The results are obtained over a sweep of learning rates.}
\label{fig:spectral_norm_entropy}
\end{figure}

\begin{figure}[t!]
\centering
\begin{subfloat}[Data.]{
\includegraphics[width=0.45\columnwidth]{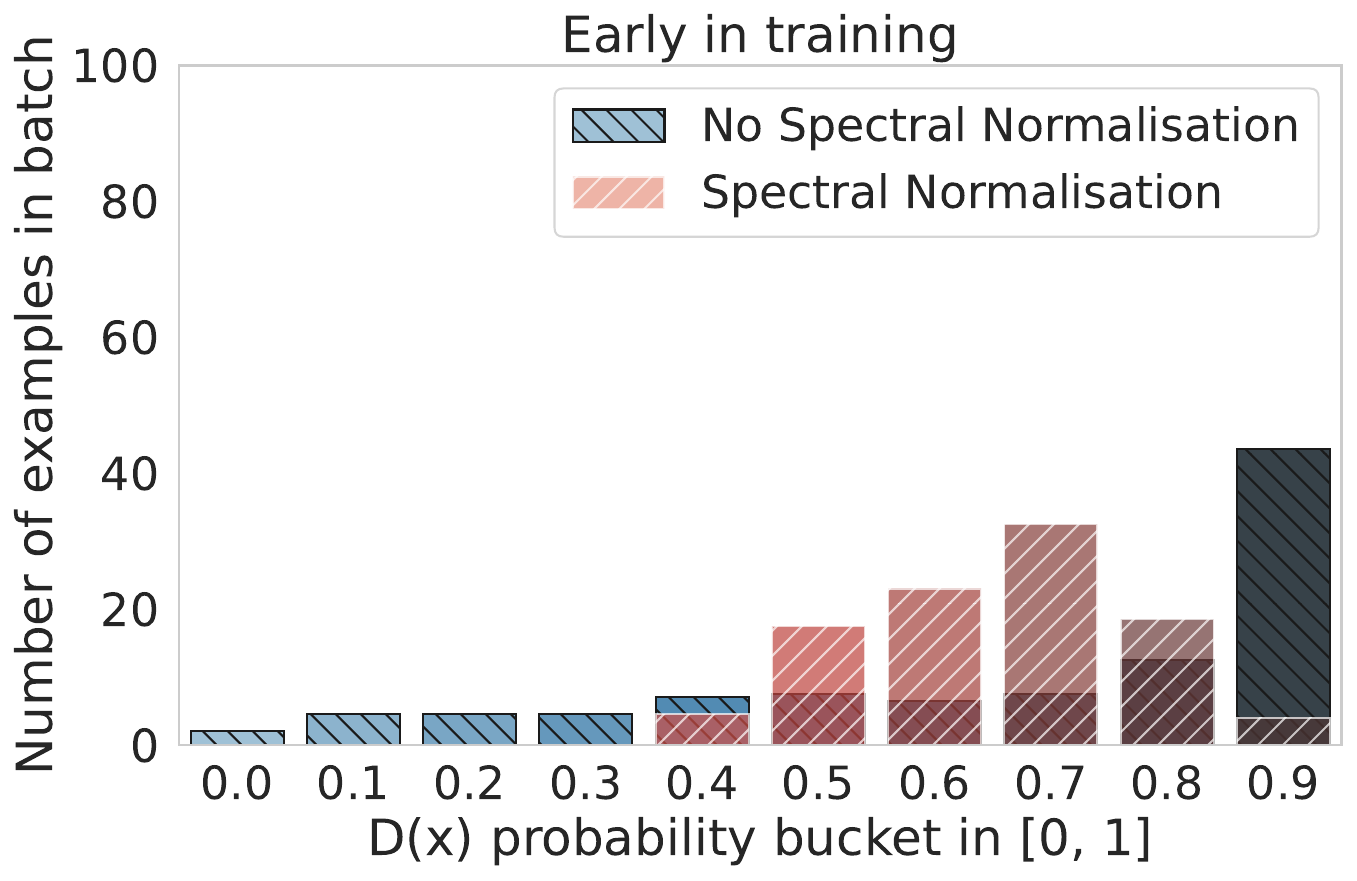}
\label{fig:data_hist_early}
}
\end{subfloat}%
\hspace{1em}
\begin{subfloat}[Samples.]{
\includegraphics[width=0.45\columnwidth]{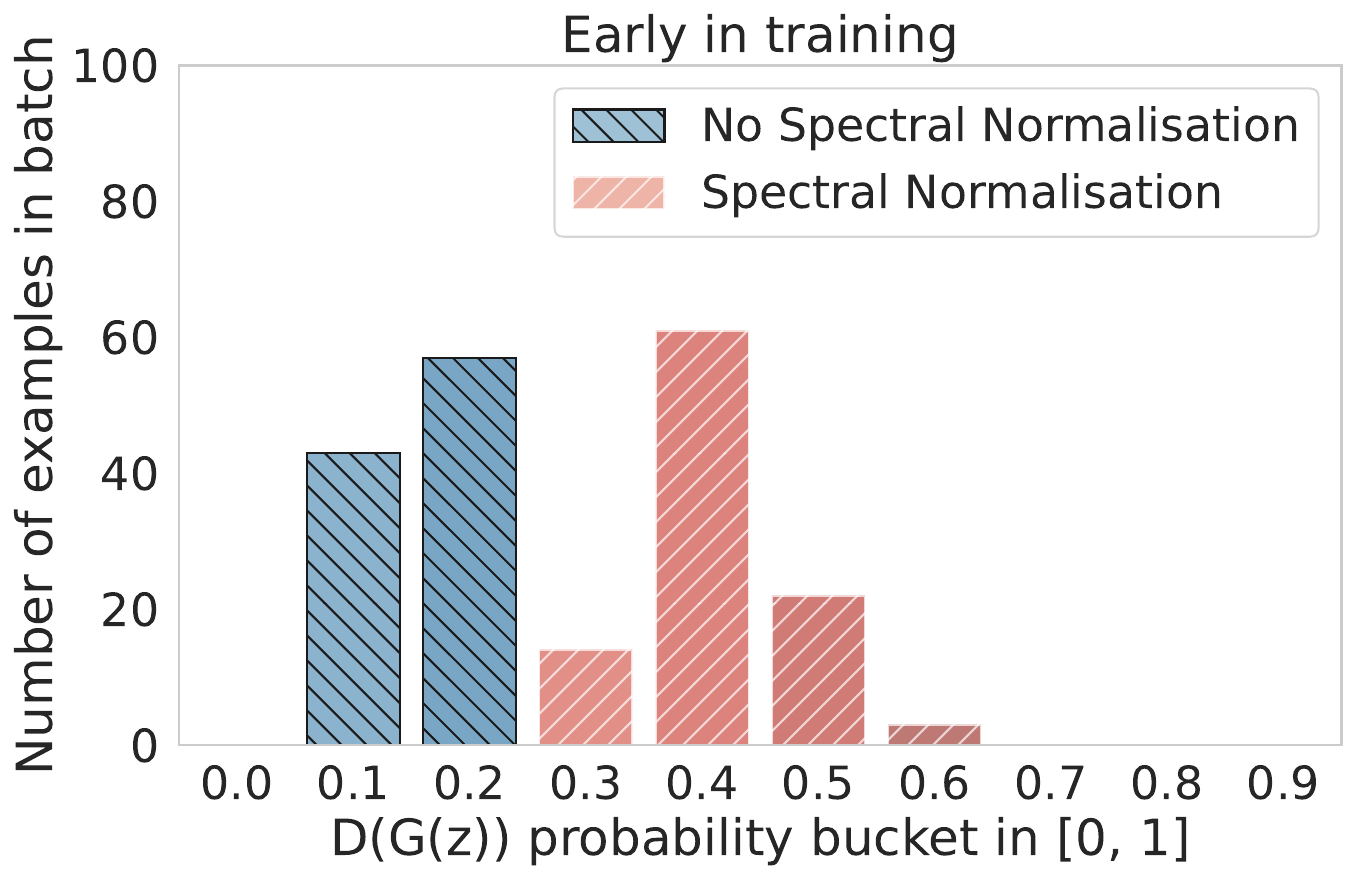}%
\label{fig:samples_hist_early}
}
\end{subfloat}%
\caption[Spectral Normalisation increases the entropy of the discriminator's predictions on data and samples in the early stages of training.]{\textbf{Early in training}. Spectral Normalisation increases the entropy of the discriminator's predictions on data \subref{fig:data_hist_early} and samples \subref{fig:samples_hist_early} in the early stages of training. Without Spectral Normalisation the discriminator is confident that the generator samples are fake, with most samples being assigned less than 0.3 probability of being real.}
\label{fig:spectral_norm_histogram_examples_early}
\end{figure}
\begin{figure}[bt!]
\centering
\begin{subfloat}[Data.]{
\includegraphics[width=0.45\columnwidth]{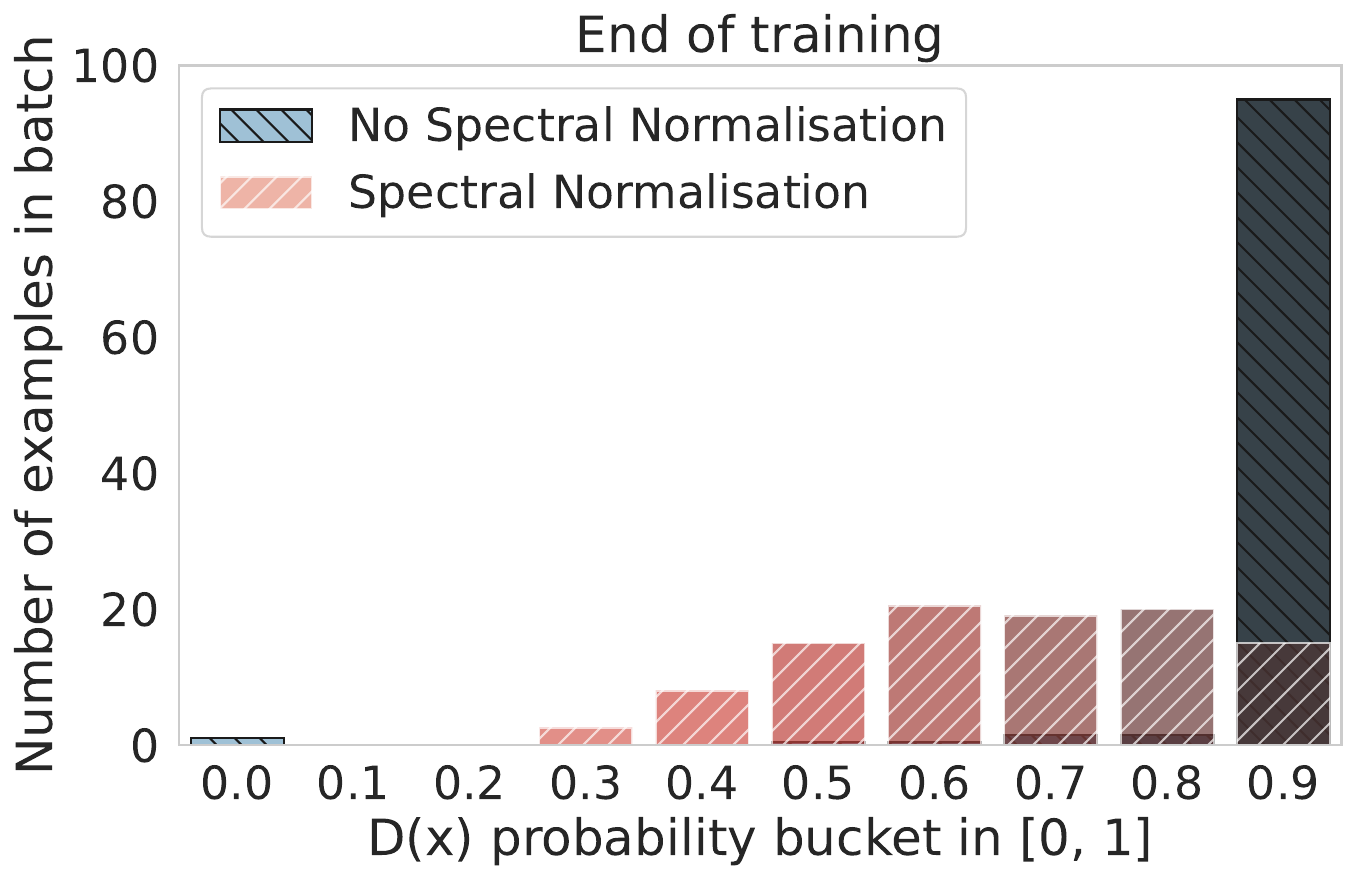}
\label{fig:data_hist_late}
}
\end{subfloat}%
\hspace{1em}
\begin{subfloat}[Samples.]{
\includegraphics[width=0.45\columnwidth]{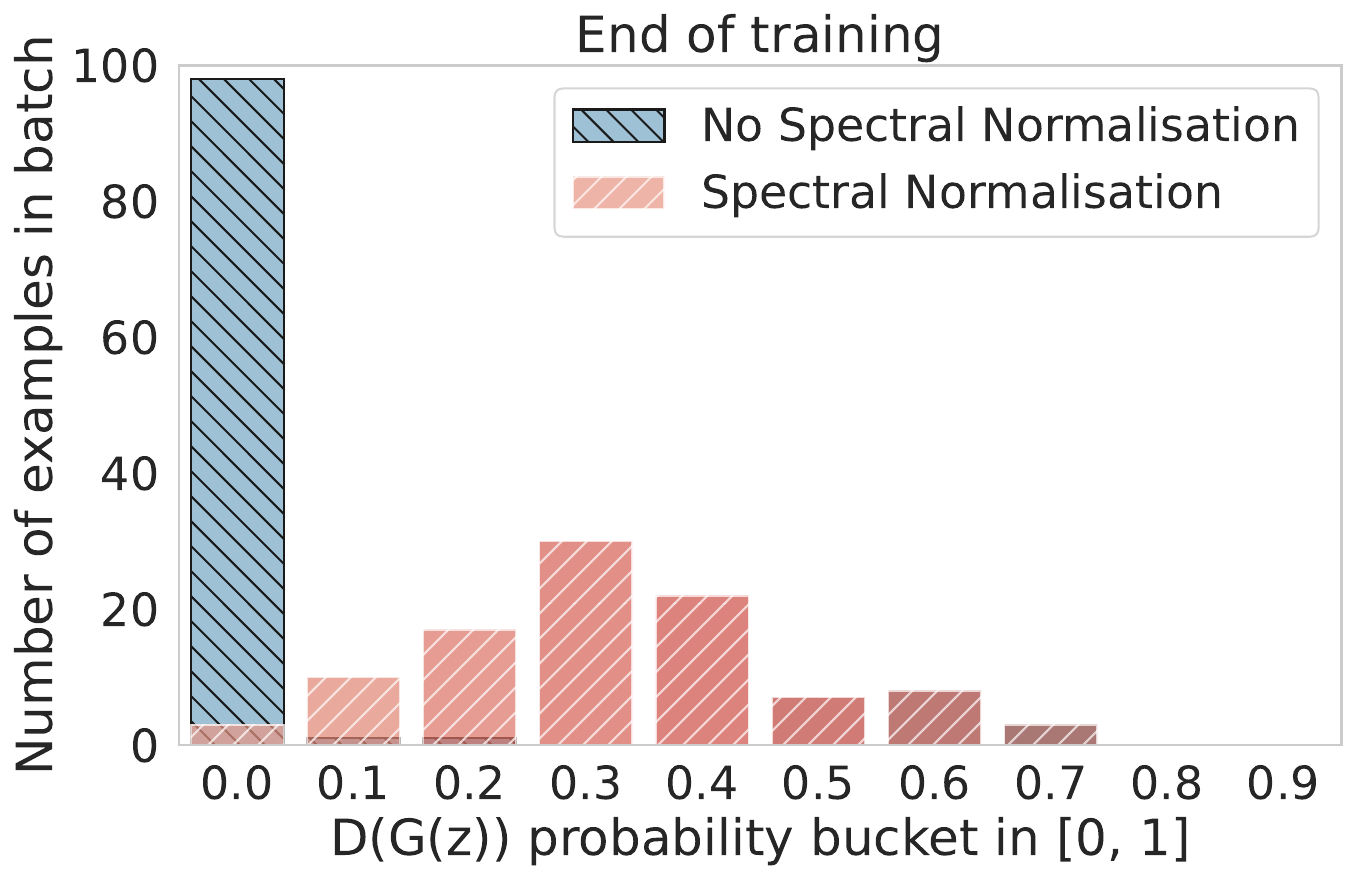}%
\label{fig:samples_hist_late}
}
\end{subfloat}%
\caption[Spectral Normalisation increases the entropy of the discriminator's predictions on data and samples in the late stages of training.]{\textbf{Late in training}. Spectral Normalisation increases the entropy of the discriminator's predictions on data \subref{fig:data_hist_late} and samples \subref{fig:samples_hist_late} in the late stages of training. Without Spectral Normalisation the discriminator is confident that real data is real and generator samples are fake. When Spectral Normalisation is used, the discriminator  assigns both data and samples a wide range of probability of being real.}
\label{fig:spectral_norm_histogram_examples}
\end{figure}

\subsection{Interactions with loss functions}

\begin{figure}[t]
\begin{subfloat}[Loss.]{
\includegraphics[width=0.49\columnwidth]{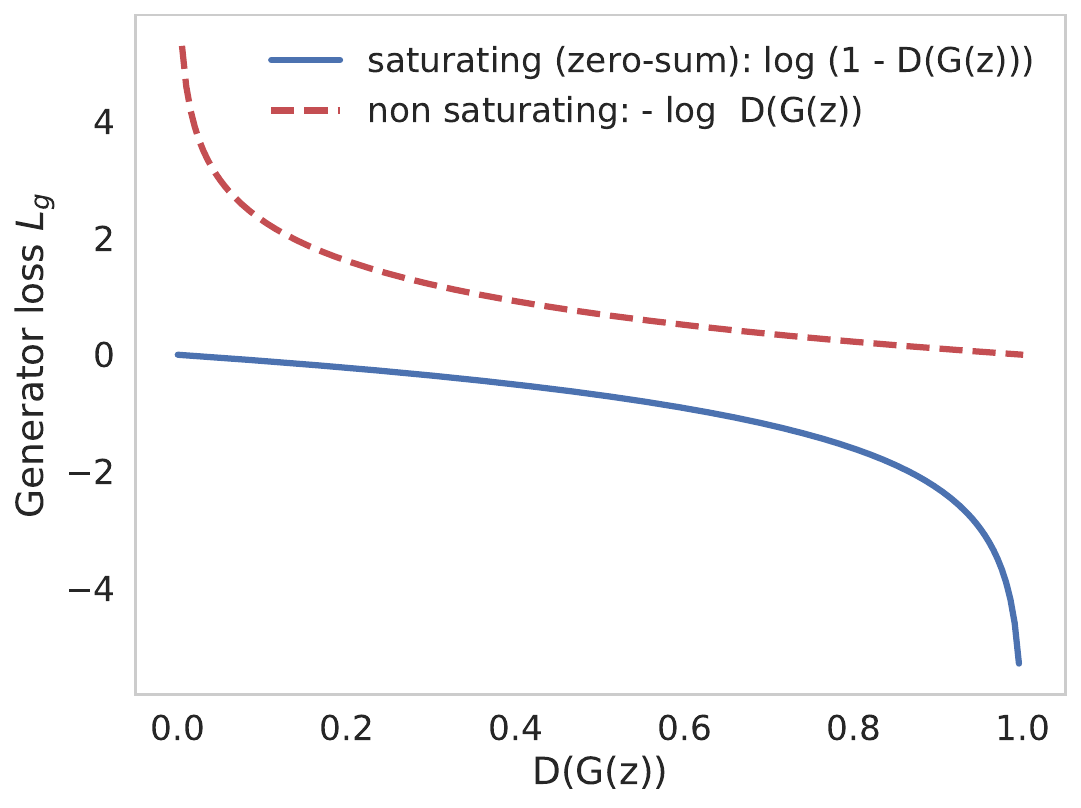}
\label{fig:losses_gan_losses}
}
\end{subfloat}%
\begin{subfloat}[Gradients.]{
\includegraphics[width=0.49\columnwidth]{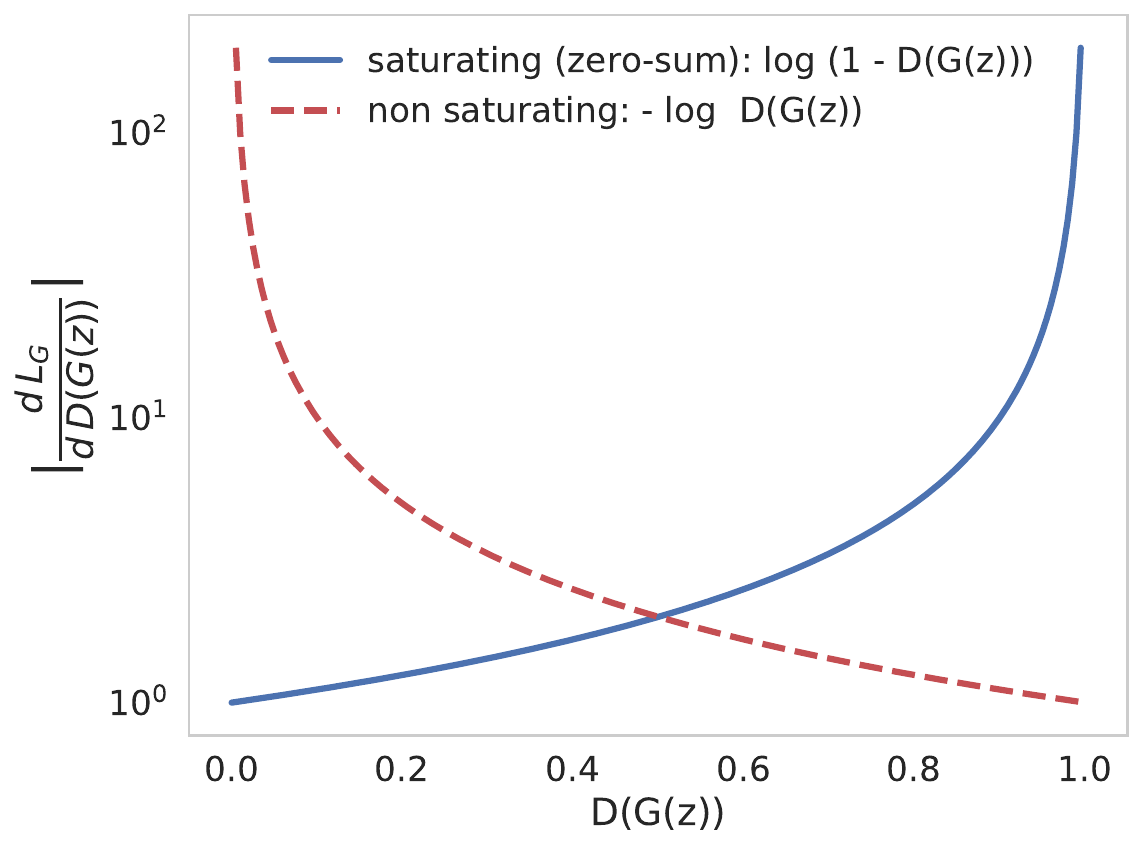}
\label{fig:grads_gan_losses}
}
\end{subfloat}
\caption[Understanding the landscape of GAN losses and gradients: comparing the saturating and non-saturating generator loss. ]{Understanding the landscape of GAN losses and gradients. \subref{fig:losses_gan_losses}: comparing the saturating and non-saturating generator loss. \subref{fig:grads_gan_losses}: unlike the saturating loss, the non-saturating loss provides stronger gradients when the discriminator can easily classify generated data as fake.}
\label{fig:sat_vs_non_sat}
\end{figure}

To examine the effect of the increased entropy on the discriminator's output, we use two formulations of GAN losses: the non-saturating loss (Eq~\eqref{eq:non_sat_gen}),  more commonly used in practice and in the above experiments; and the original zero-sum formulation of GANs (Eq~\eqref{eq:original_gan}), which has been known to be harder to train than other losses.
The low performance of this original loss has been attributed to the lack of learning signal the generator early in training, since when the generator performs poorly and the discriminator easily classifies its data as fake, the discriminator does not provide strong gradients for the generator to improve, as we visualise in Figure~\ref{fig:sat_vs_non_sat}; 
for this reason, the generator loss in this formulation is sometimes called the `saturating loss'.
 Moreover, strong gradients are provided when the generator is doing well, which can lead to unstable behaviour and exiting areas of good performance. Figure~\ref{fig:grads_gan_losses} also highlights why the increased entropy effect in the discriminator's output induced by Spectral Normalisation---and specifically the decreased probability of assigning outputs close to $0$ or $1$ observed in Figures~\ref{fig:spectral_norm_histogram_examples_early} and~\ref{fig:spectral_norm_histogram_examples}---can be beneficial, as \textit{for both losses, it avoids areas with large or small gradient magnitudes}.

By analysing the interaction between the loss function and the behaviour of the discriminator, we can now make predictions about the empirical behaviours of the two losses. For the saturating loss, without Spectral Normalisation the challenge is that early in training there is no learning signal for the generator; we thus would expect that without Spectral Normalisation the performance is low, and that Spectral Normalisation improves is significantly by avoiding the saturated areas early in training. We observe this in Figure~\ref{fig:div_out_gans_d_and_g_sat_main}, which shows that without Spectral Normalisation the saturating loss exhibits extremely poor performance, but that adding Spectral Normalisation significantly decreases the gap between the saturating and non-saturating losses. For the non-saturating loss, we expect that without Spectral Normalisation, the discriminator can be too confident in its prediction that generated data is fake late in the game, as we observed in Figure~\ref{fig:samples_hist_late}. This leads to large gradients---see Figure~\ref{fig:grads_gan_losses}---which can destabilise training and exit areas of good performance. This expectation is consistent with what we observe empirically in Figure~\ref{fig:div_out_gans}: adding Spectral Normalisation improves stability later in training.

\begin{figure}[t!]
\centering
\begin{subfloat}[Spectral Normalisation.]{
\includegraphics[width=0.4\columnwidth]{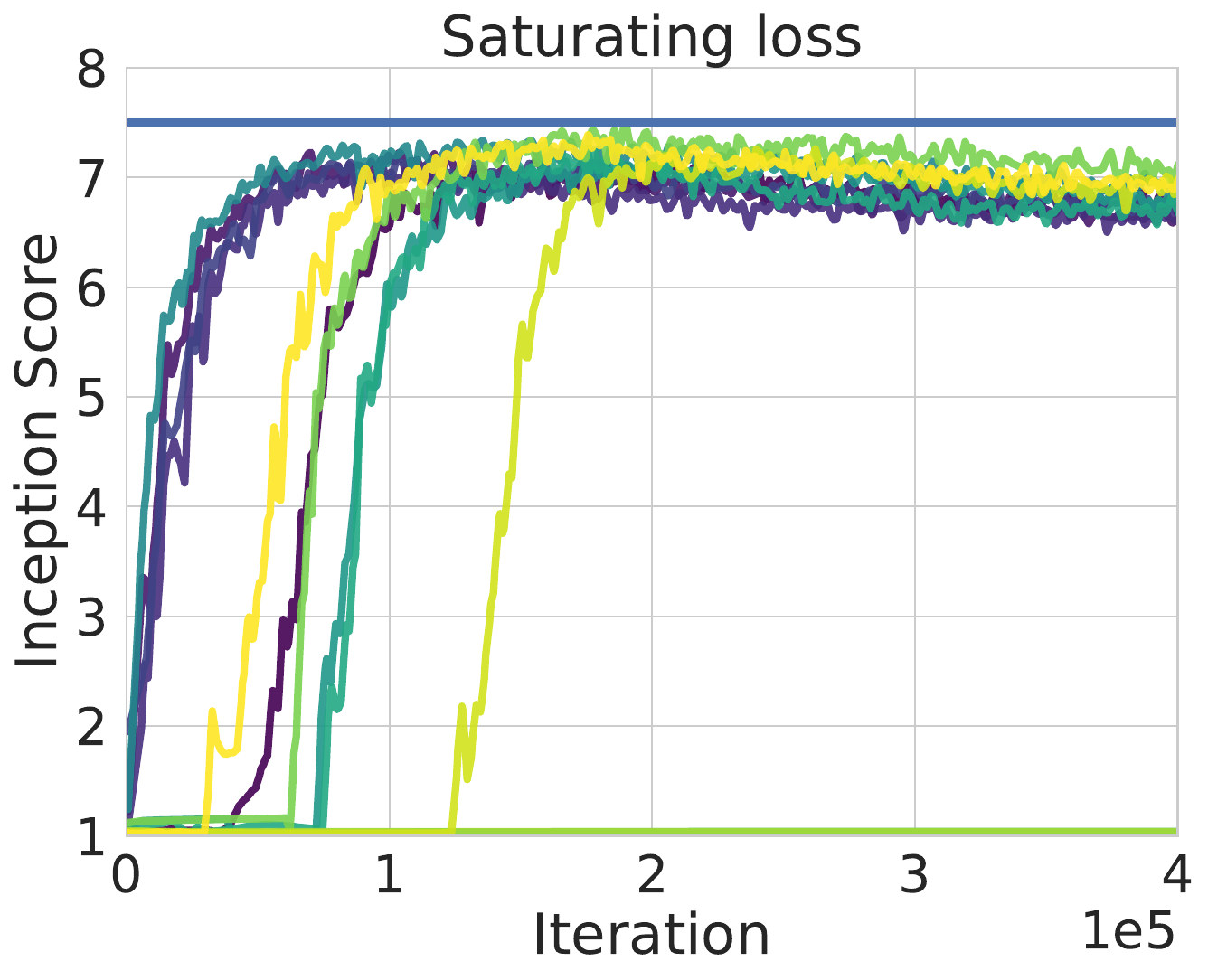}
}
\end{subfloat}
\hspace{2em}
\begin{subfloat}[No normalisation.]{
\includegraphics[width=0.4\columnwidth]{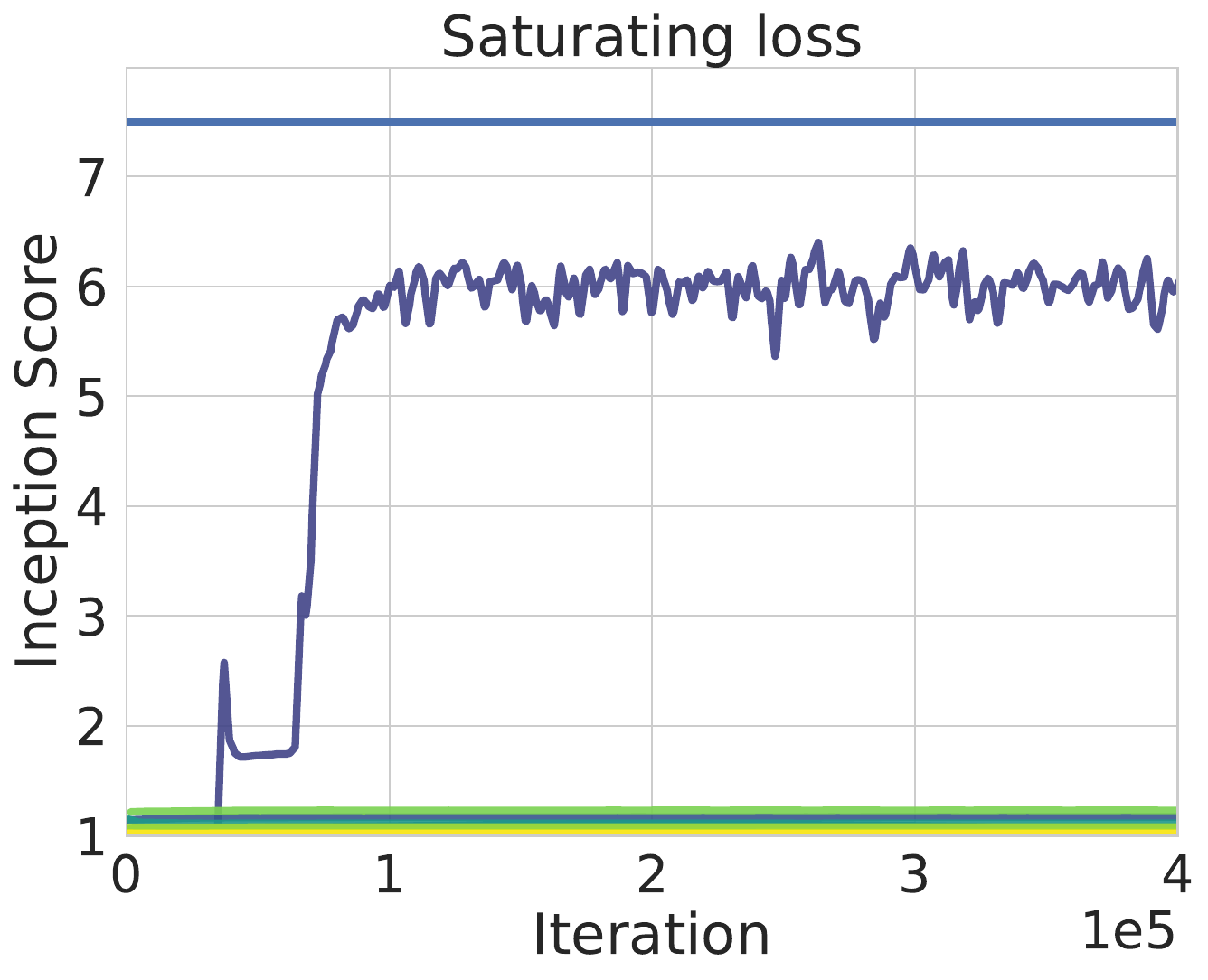}%
}
\end{subfloat}
\begin{subfloat}[\textsc{DivOut}, discriminator update only.]{
\includegraphics[width=0.4\columnwidth]{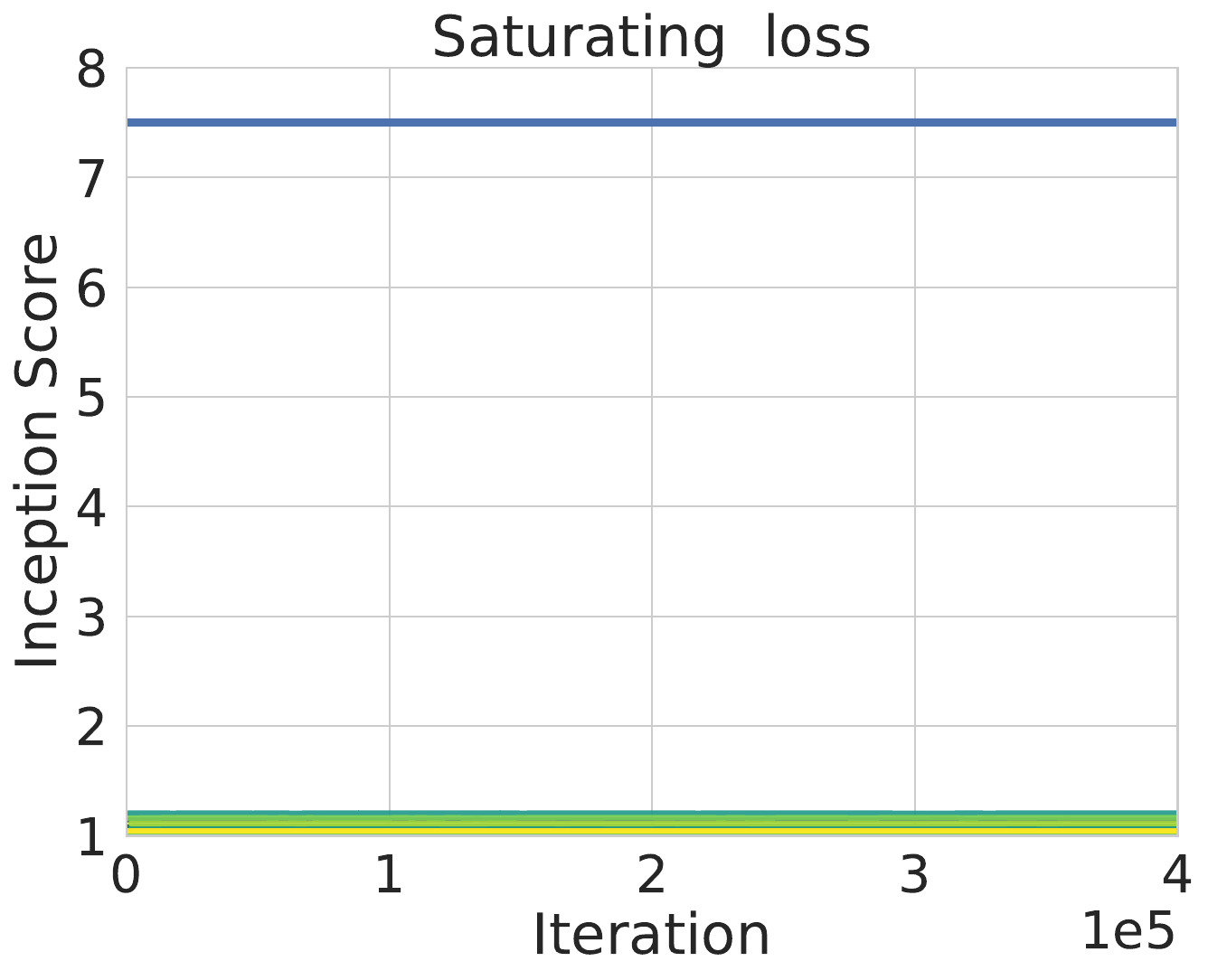}
\label{fig:sat_disc_only}
}
\end{subfloat}%
\hspace{2em}
\begin{subfloat}[\textsc{DivOut}.]{
\includegraphics[width=0.4\columnwidth]{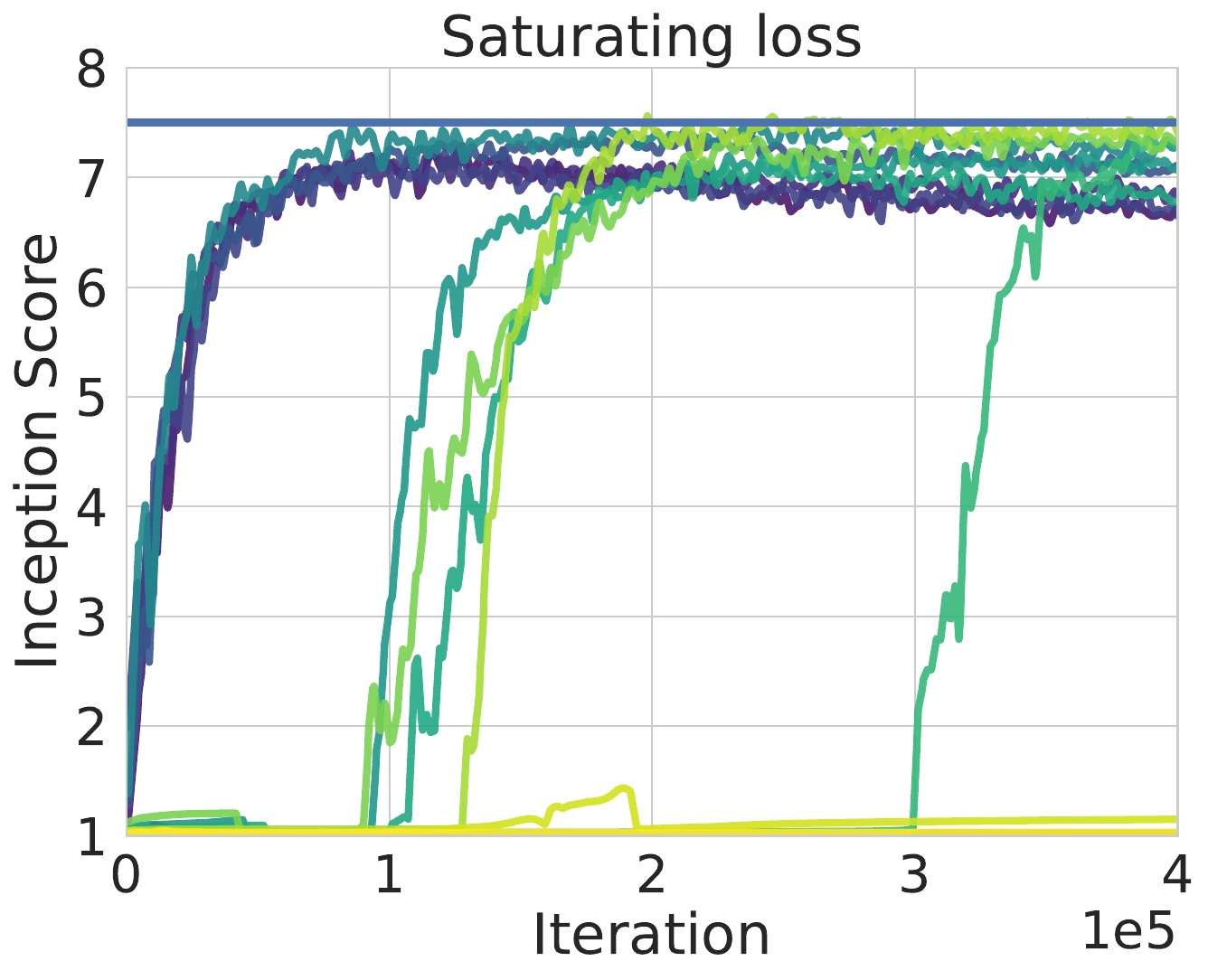}
\label{fig:sat_disc_and_gen}
}
\end{subfloat}
\caption[Only applying the Jacobian change to the discriminator only leads to extremely poor performance when the saturating loss is used.]{\textbf{Saturating loss.} Spectral Normalisation improves performance of GANs trained with the saturating loss. \textsc{divOut} obtains similar performance as Spectral Normalisation, but only when applied to both the discriminator and the generator. Only applying the output scaling from \textsc{divOut} to the discriminator only leads to extremely poor performance, as it does not affect the generator update, and thus does not avoid the areas with poorly conditioned generator gradients. Different units indicate different learning rates for the discriminator and generator. The training optimiser is Adam, used with simultaneous updates.}
\label{fig:div_out_gans_d_and_g_sat_main}
\end{figure}

\begin{figure}[t!]
\centering
\begin{subfloat}[Spectral Normalisation.]{
\includegraphics[width=0.41\columnwidth]{spectral_norm_div_comp}%
}
\end{subfloat}%
\hspace{2em}
\begin{subfloat}[No normalisation.]{
\includegraphics[width=0.41\columnwidth]{no_normalisation_div_mul_comp}%
}
\end{subfloat}
\begin{subfloat}[\textsc{DivOut}, discriminator update only.]{
\includegraphics[width=0.41\columnwidth]{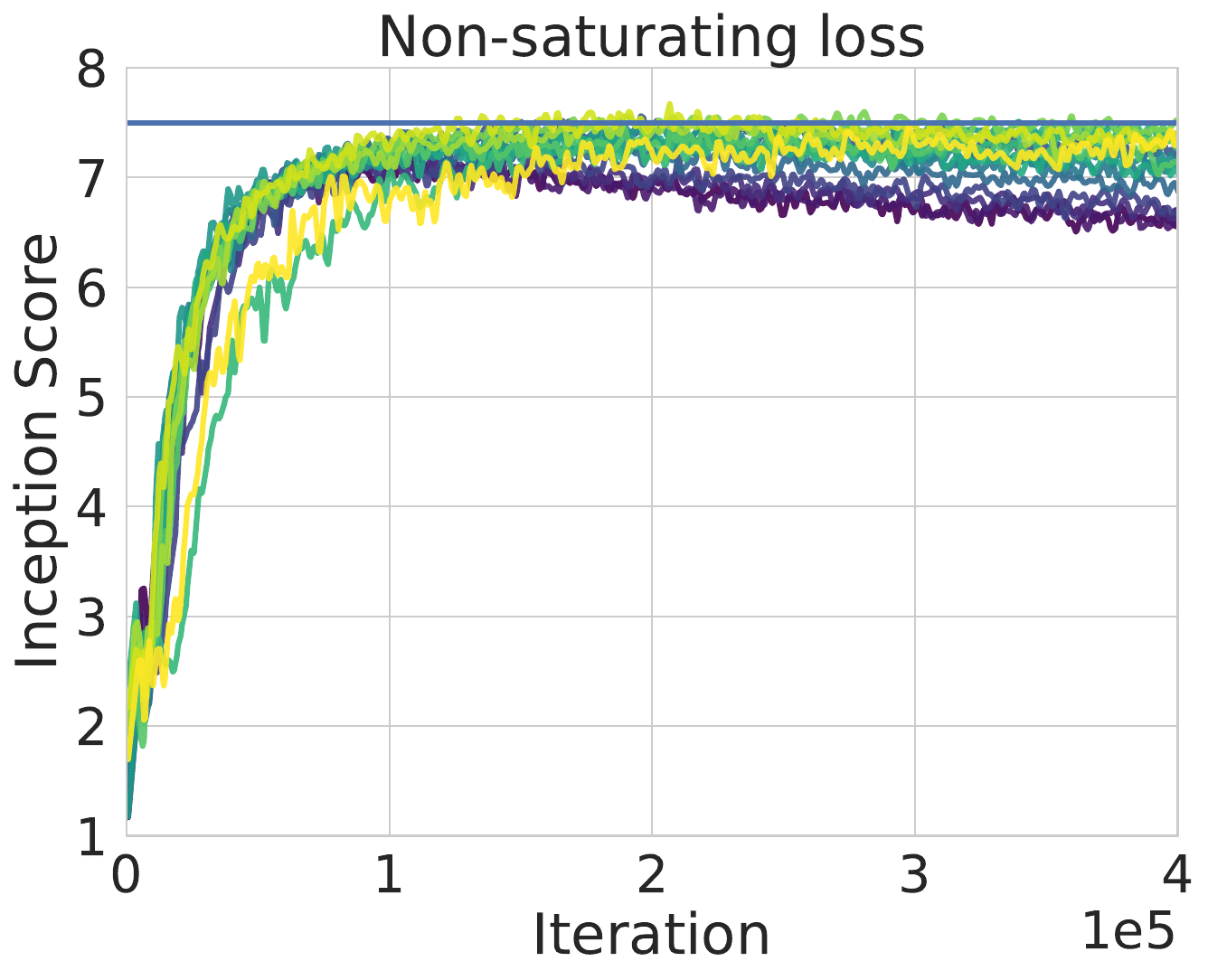}
\label{fig:non_sat_disc_only}
}
\end{subfloat}
\hspace{2em}
\begin{subfloat}[\textsc{DivOut}.]{
\includegraphics[width=0.41\columnwidth]{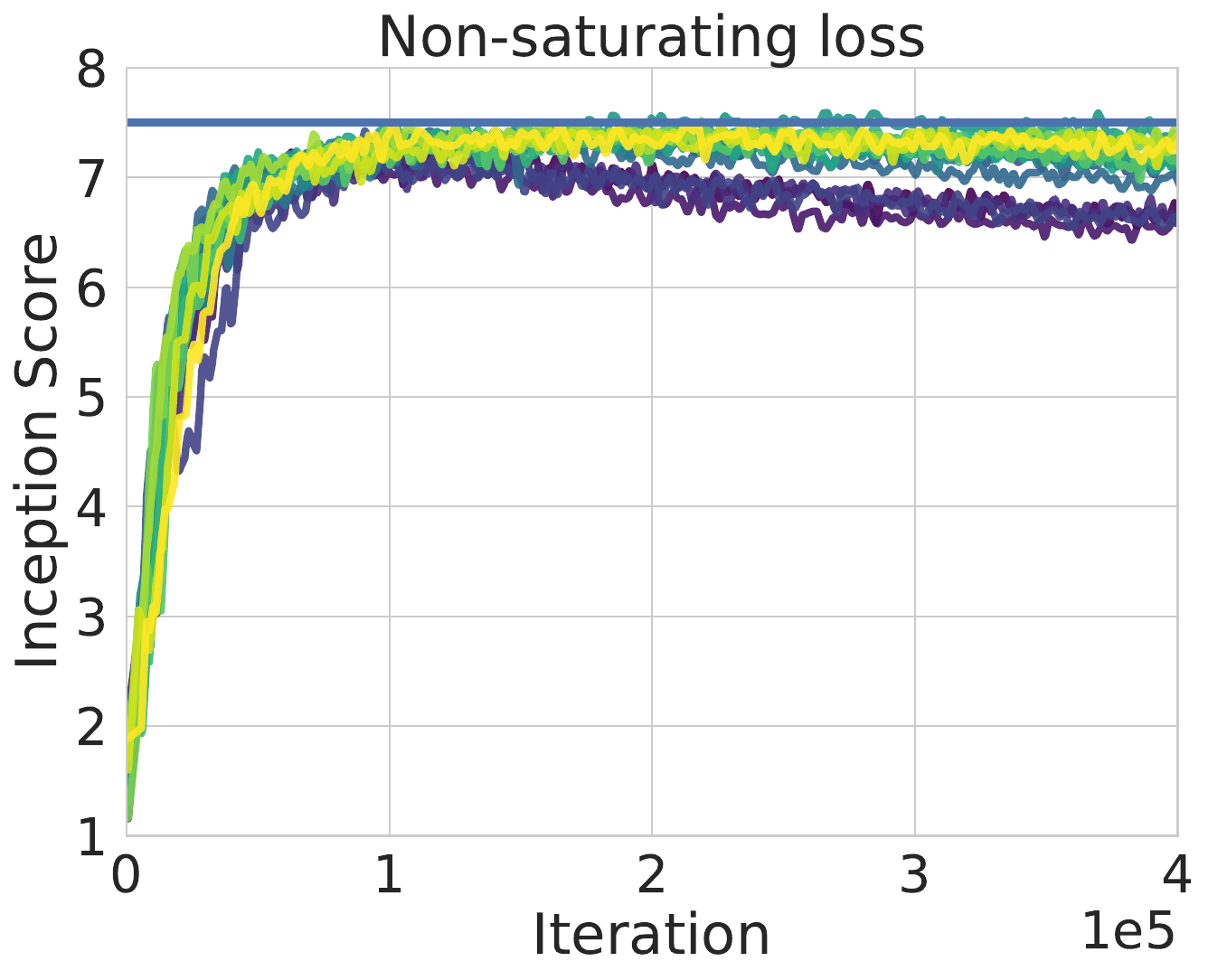}
\label{fig:non_sat_disc_and_gen}
}
\end{subfloat}
\caption[\textsc{DivOut} obtains similar performance to Spectral Normalisation for GANs on CIFAR-10 when the non-saturating loss is used.]{\textbf{Non-saturating loss.} \textsc{DivOut} obtains similar performance to Spectral Normalisation for GANs on CIFAR-10 when the non-saturating loss is used. Different units indicate different learning rates for the discriminator and generator. Only applying the output scaling from \textsc{divOut} to the discriminator leads to similar performance as when applying it to both the discriminator and the generator, since the non-saturating loss makes the generator less susceptible to the scale of the discriminator output, unlike for the saturating loss as we saw in Figure~\ref{fig:div_out_gans_d_and_g_sat_main}.  The training optimiser is Adam, used with simultaneous updates.}
\label{fig:div_out_gans}
\end{figure}

We are now ready to return to the optimisation methods we derived in Section~\ref{sec:rl_spectral_schedulers} based on Spectral Normalisation. Since \textsc{divOut} captures the changes in the model output induced by Spectral Normalisation, 
we would expect \textsc{divOut} to perform well for GANs trained on CIFAR-10. In particular, by dividing the discriminator output by the product of the spectral norms of the discriminator's weights, \textsc{divOut} captures the intuition present in Figure~\ref{fig:sigmoid_effect_sn_no_sn}: it ensures that data and generator samples are less likely to be separated and reach the saturating areas of the sigmoid function.  Indeed, as we show in Figures~\ref{fig:div_out_gans_d_and_g_sat_main} and~\ref{fig:div_out_gans}, \textsc{divOut} obtains a similar performance to Spectral Normalisation both for the saturating and non-saturating loss. Importantly, however, when applying \textsc{divOut} only to the discriminator update, \textsc{divOut} obtains extremely poor performance in the case of the saturating loss as we show in Figure~\ref{fig:sat_disc_only}, but 
still performs on par with Spectral Normalisation when the non-saturating loss is used, as seen in Figure~\ref{fig:non_sat_disc_only}. This further confirms that in the case of the saturating loss, one of the effects of Spectral Normalisation is to change the conditioning of the generator by avoiding the areas with low magnitude gradients early in training. We are also ready to revisit the \textsc{mulEps} results with the non-saturating loss presented in Figure~\ref{fig:mul_eps}; since \textsc{mulEps} does not capture the changes in discriminator output, it will not be able to avoid the decrease in performance observed when no Spectral Normalisation is used.

We thus see that in the GAN case, there is a strong effect in model training brought  by Spectral Normalisation, which does not come from an only optimisation change. We now turn our attention to the following question: even if in GANs Spectral Normalisation does not lead to a primarily optimisation effect, are GANs potentially benefiting from the optimisation changes brought by Spectral Normalisation?

\subsection{Hypothesis: Adam in the low-curvature regime}

We have seen in previous sections that Spectral Normalisation can be seen as acting to multiply the hyperparameter $\epsilon$ of Adam (Eq~\eqref{eq:mul_eps}). Since the spectral norms of weight matrices grow in training, this can be seen as an anneal of $\epsilon$ from small to large. In the previous section we have seen that in the GAN case the scheduler derived from this observation,  \textsc{MulEps}, does not recover the performance of Spectral Normalisation in GANs. Now we ask whether the $\epsilon$ annealing behaviour of Spectral Normalisation can still have a positive effect on GAN training, despite not being the only effect.

We start with a few observations. First, when a small $\epsilon$ is used, for dimensions of low-curvature Adam leads to a constant update given by the learning rate. Second, GANs trained with Adam exit the area of high performance, even when Spectral Normalisation and the non-saturating losses are used, as we show in Figure~\ref{fig:mul_eps}. Third, this behaviour is not observed when gradient descent or Runge--Kutta4 are used (see Figure~\ref{fig:sgd_adam_sim_comparison} and~\citet{odegan}), so this behaviour seems specific to the use of the Adam optimiser. These observations lead to us to postulate the following hypothesis: \textit{Is Adam exiting the areas of high performance in GAN training because many parameters are in a low-curvature regime, and taking a step of size given by the learning rate is then leading to an exit from the high performance area?}

To first understand the effect of $\epsilon$ in Adam in the low-curvature regime, we use the arguments from Section~\ref{sec:opt_algo}. If we are in a low-curvature regime for parameter $\vtheta_{[l]} \in \mathbb{R}$, then by definition its gradient $\nabla_{\vtheta_{[l]}} E$ does not change between iterations. We use this to expand $\vm_t$ in the Adam update, defined in Eq~\eqref{eq:adam_m_expanded}, at iteration $t$:
\begin{align}
{\left(\vm_t\right)}_{[l]} &= (1 - \beta_1) \sum_{i=1}^t \beta_1^{i-1} \nabla_{\vtheta_{[l]}}E(\vtheta_{t-i}) 
      \approx (1 - \beta_1) \sum_{i=1}^t \beta_1^{t-i} \nabla_{\vtheta_{[l]}} E \\
     & = (1 - \beta_1^{t}) \nabla_{\vtheta_{[l]}} E.
\label{eq:low_curvature_m}
\end{align}
With the same arguments we have
\begin{align}
{\left(\vv_t\right)}_{[l]} & \approx (1 - \beta_2^{t}) (\nabla_{\vtheta_{[l]}} E)^2.
\label{eq:low_curvature_v}
\end{align}
We thus have that the Adam update for parameter $\vtheta_{[l]}$ is
\begin{align}
\Delta \vtheta_{[l]} = h \frac{\frac{1}{1 - \beta_1^t} {\left(\vm_t\right)}_{[l]}}{\sqrt{\frac{1}{1 - \beta_2^t} {\left(\vv_t\right)}_{[l]}} + \epsilon} \approx h \frac{\nabla_{\vtheta_{[l]}} E}{\sqrt{(\nabla_{\vtheta_{[l]}} E)^2} + \epsilon}.
\label{eq:adam_low_curve}
\end{align}
This result shows the importance of $\epsilon$ in Adam in the low-curvature regime: if $\epsilon$ is small then the update is given by the learning rate $h$; this makes the update independent of the gradient magnitude. We plot this in Figure~\ref{fig:adam_eps_effect}, and show how increasing $\epsilon$ can reinstate the dependence on gradient magnitudes. Importantly, even if the model is in the neighbourhood of a local equilibrium and the gradient is close to $\mathbf{0}$, the size of the Adam update with default $\epsilon = 10^{-8}$ is given by the learning rate.  Thus, if there are many parameters in the low-curvature regime around an area of high performance, Adam can exit that area of high performance 
 by taking large steps when gradients are small; we have seen examples of this behaviour in a simple setting in Figure~\ref{fig:adam}. Since SGD and Runge--Kutta4 only use local information about the gradient, they would not exit the high performance area if gradients are small there.
\begin{figure}[t!]
\centering
\begin{subfloat}[$\epsilon=10^{-8}$.]{
\includegraphics[width=0.455\columnwidth]{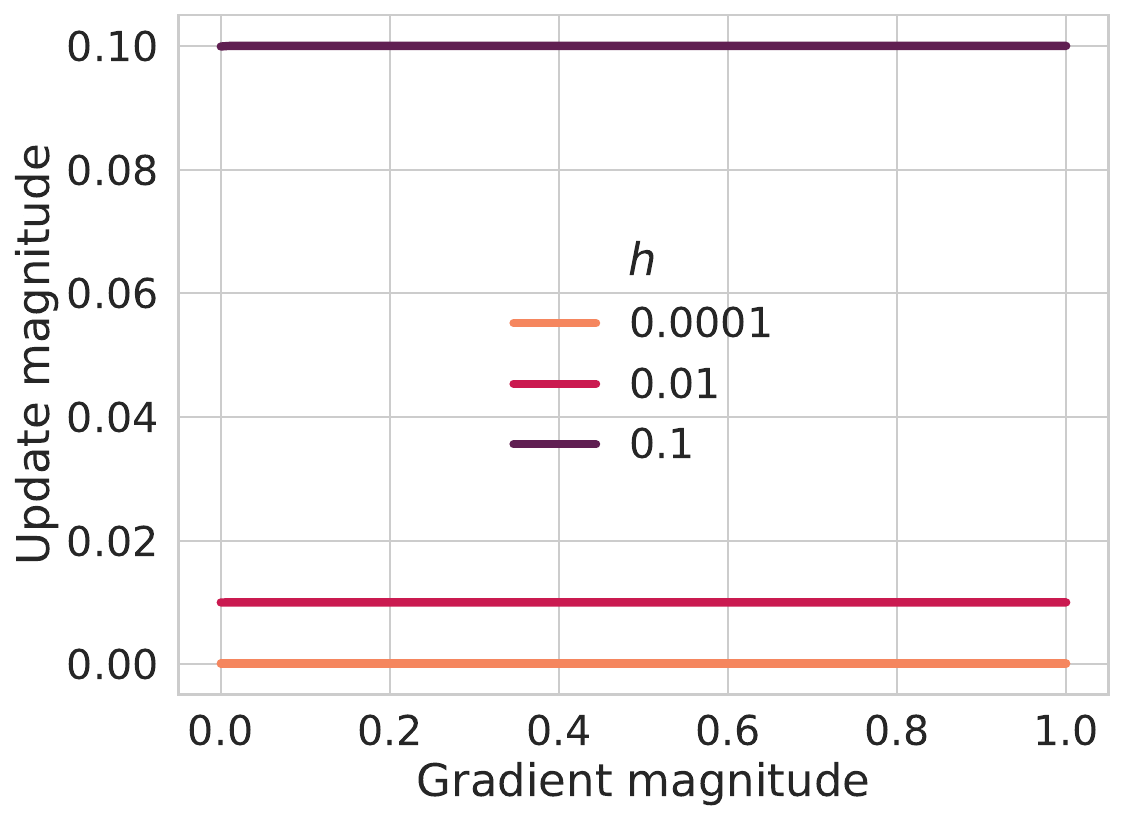}
\label{fig:adam_lr_effect}
}
\end{subfloat}
\begin{subfloat}[Before applying fixed $h$.]{
\includegraphics[width=0.445\columnwidth]{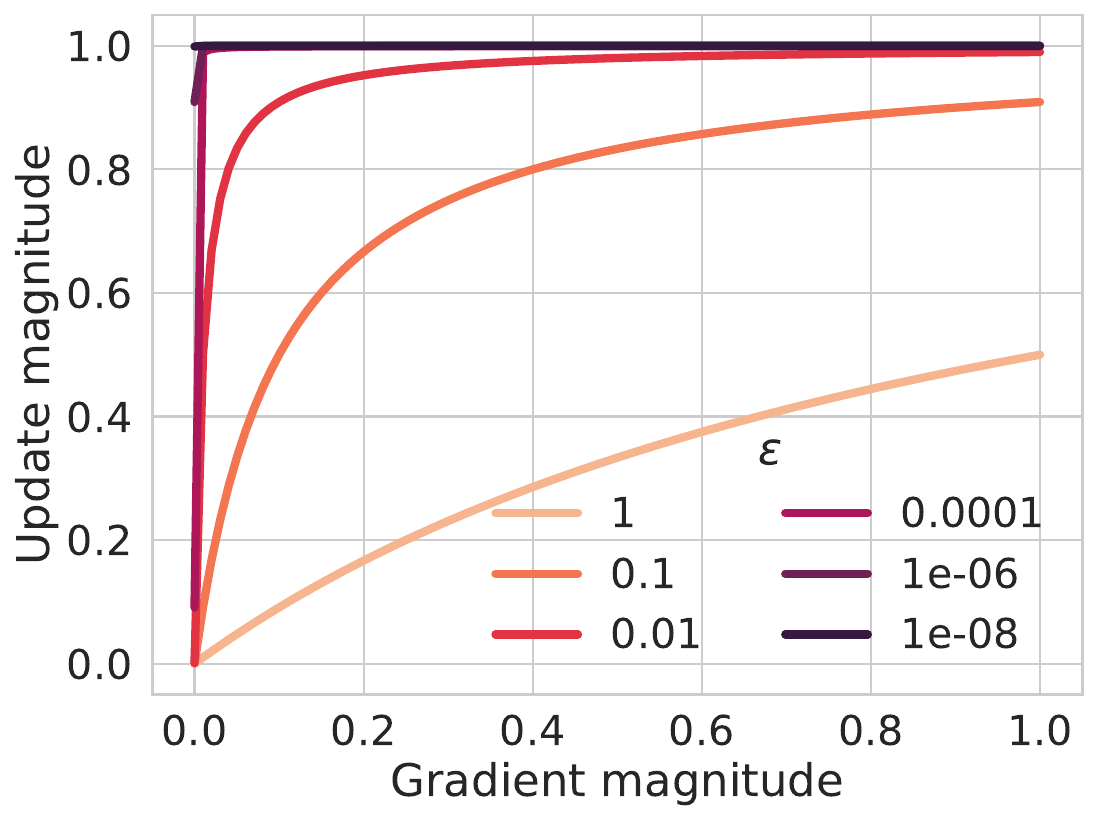}
\label{fig:adam_eps_effect}
}
\end{subfloat}
\caption[The effect of the learning rate and $\epsilon$  on the magnitude of the update in the low-curvature regime in Adam.]{The effect of the learning rate \subref{fig:adam_lr_effect} and $\epsilon$ \subref{fig:adam_eps_effect} on the magnitude of the update in the low-curvature regime in Adam: while the learning rate scales the gradient by a constant, the effect of $\epsilon$ strongly depends on the magnitude of the gradient, particularly for large $\epsilon$.}
\end{figure}
\begin{figure}[t!]
\centering
\begin{subfloat}[$\epsilon=10^{-8}$.\\ At peak: $\epsilon=10^{-8}$.]{
\includegraphics[width=0.332\columnwidth]{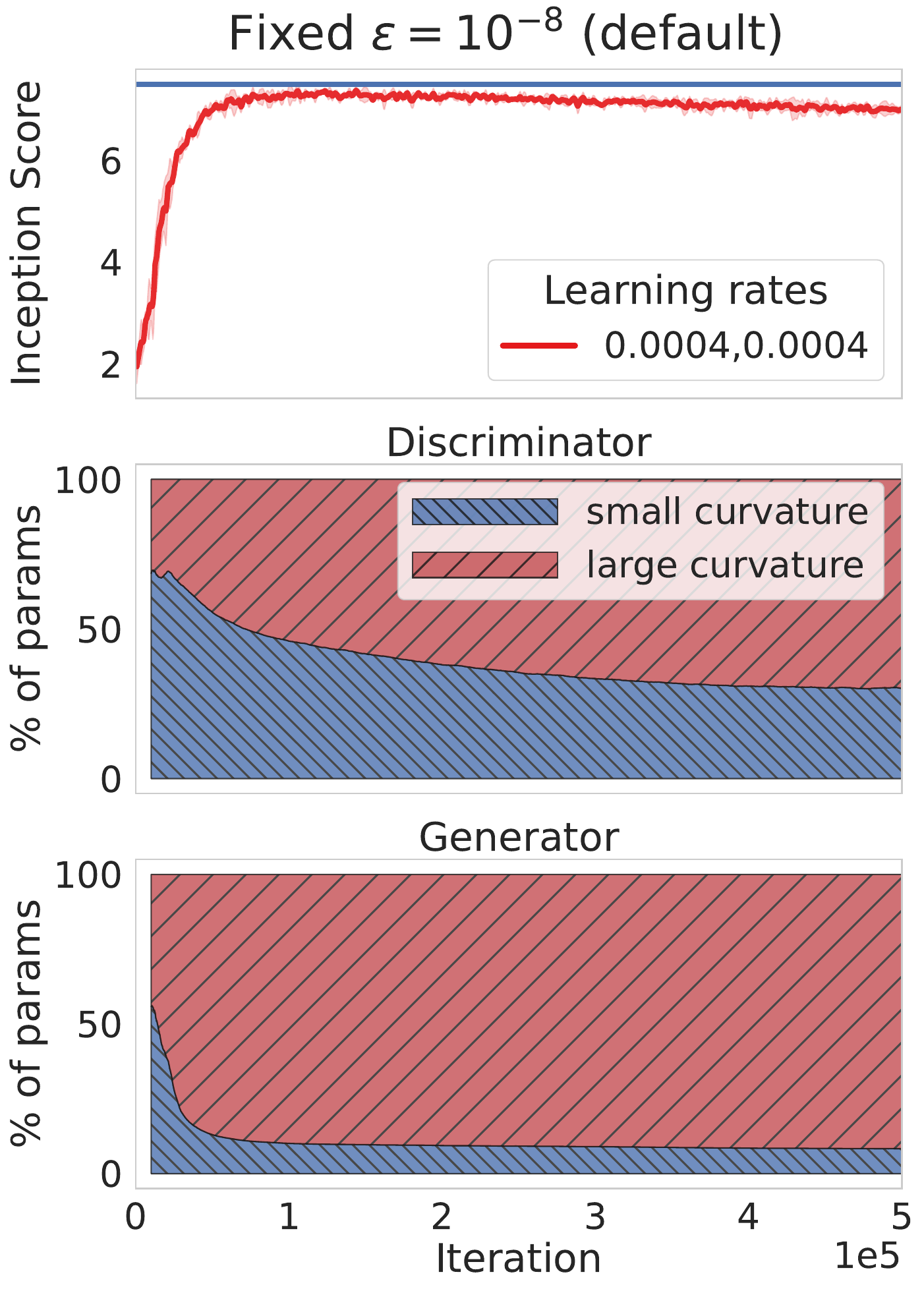}%
\label{fig:adam_curvature_small_eps}
}
\end{subfloat}%
\begin{subfloat}[$\epsilon$ annealed from $10^{-8}$ to $10^{-4}$.\\ At peak: $\epsilon\approx 10^{-5}$.]{
\includegraphics[width=0.333\columnwidth]{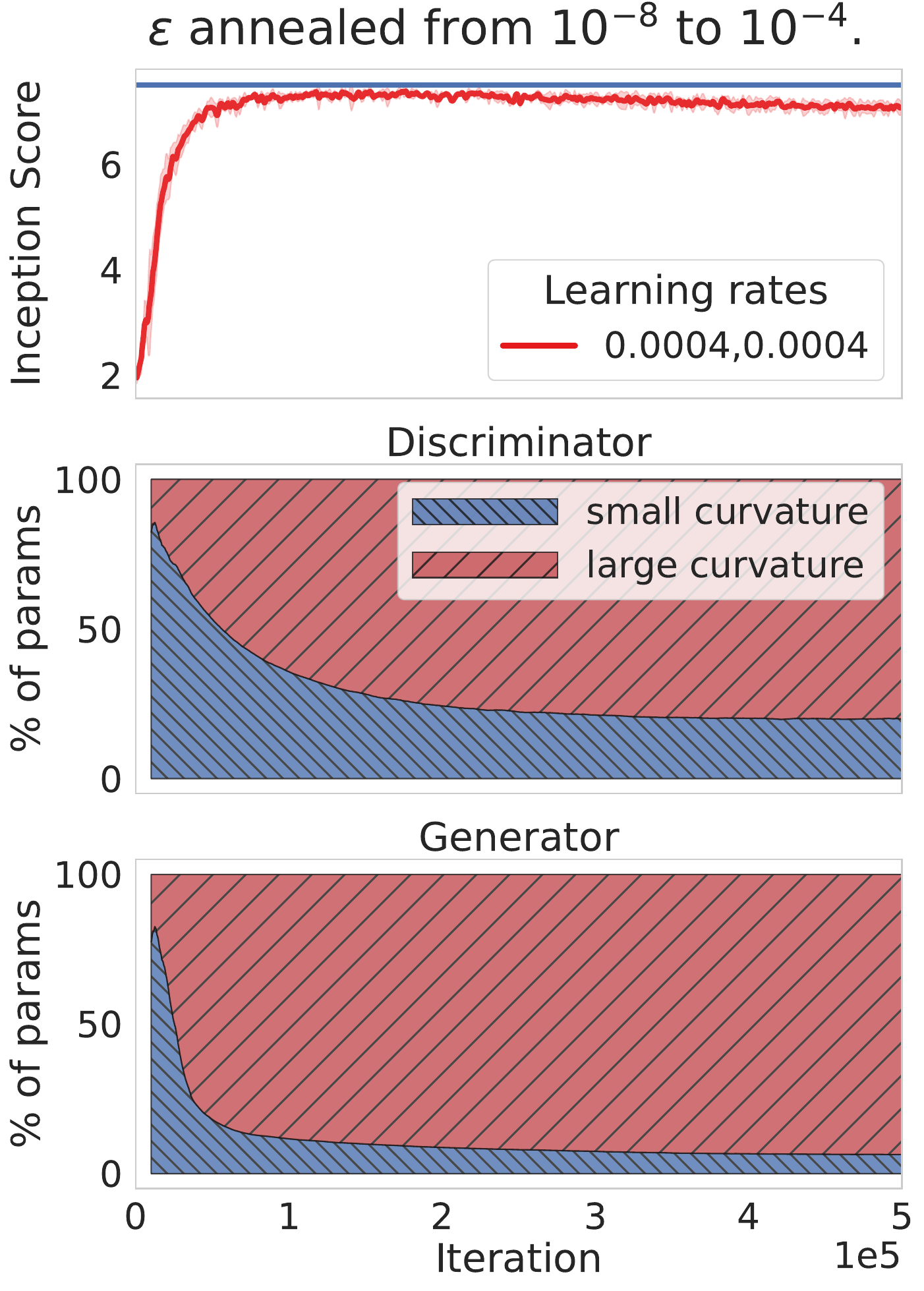}%
\label{fig:adam_curvature_anneal1}
}\end{subfloat}%
\begin{subfloat}[$\epsilon$ annealed from $10^{-8}$ to $10^{-2}$. \\ At peak: $\epsilon\approx 10^{-3}$.]{
\includegraphics[width=0.333\columnwidth]{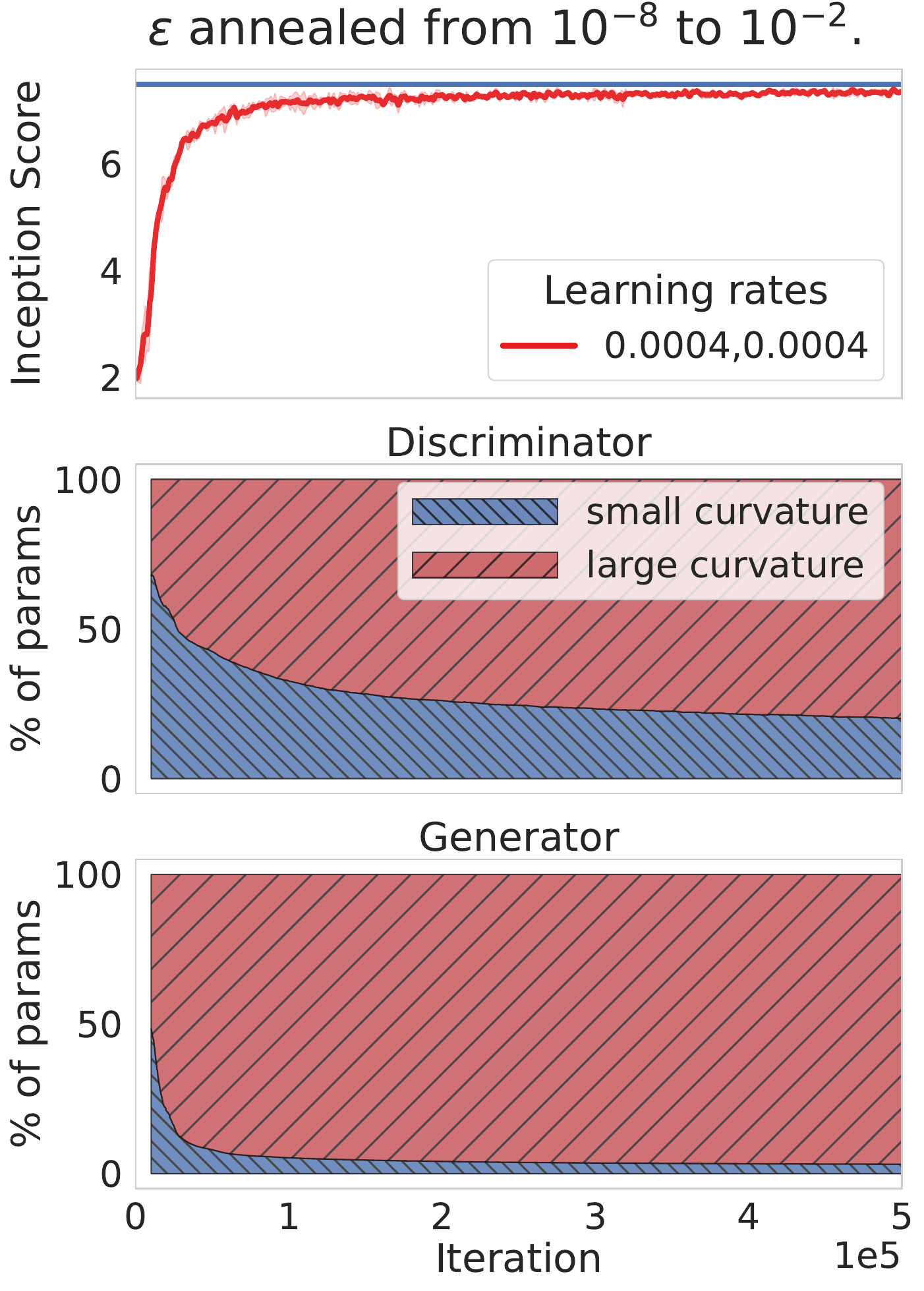}%
\label{fig:adam_curvature_anneal2}
}\end{subfloat}%
\caption[The effect of $\epsilon$ on Adam performance in GAN training.]{The effect of $\epsilon$ on Adam performance. Increasing $\epsilon$ in training ensures that Adam does not exit areas of good performance, even with many parameters in the low-curvature regime, particularly for the discriminator. We note that using a large $\epsilon$ from early in training leads to extremely low performance (likely due to small gradients training does not pick up) and we do not show it here.}
\label{fig:adam_curvature}
\end{figure}

In order to understand whether this effect occurs when training GANs in practice, we need to measure whether there are many parameters in a low-curvature regime, as well as investigate the effect of $\epsilon$ on GAN training.
To answer the first question, we measure
\begin{align}
  \left|\nabla_{\vtheta_{[l]}} E(\vtheta_t) - \frac{1}{1-\beta_1^t} {\left(\vm_t\right)}_{[l]} \right| \hspace{1em} \text{and}  \hspace{1em} \left|\left(\nabla_{\vtheta_{[l]}} E(\vtheta_t)\right)^2 - \frac{1}{1-\beta_2^t} {\left(\vv_t\right)}_{[l]}\right|
\end{align}
for all discriminator and generator parameters. These quantities are readily available when using Adam, we know they are $0$ for parameters in areas of low curvature (Eq~\ref{eq:low_curvature_m} and Eq~\ref{eq:low_curvature_v}), and we know how they affect the Adam update (Eq~\ref{eq:adam_low_curve}). We define a parameter as being in an area of low curvature at iteration $t$ if ${\left|\nabla_{\vtheta_{[l]}} E(\vtheta_t) - \frac{1}{1-\beta_1^t} {\left(\vm_t\right)}_{[l]} \right|< 10^{-5}}$ and ${\left|\left(\nabla_{\vtheta_{[l]}} E(\vtheta_t)\right)^2 - \frac{1}{1-\beta_2^t} {\left(\vv_t\right)}_{[l]}\right|< 10^{-5}}$. Results in Figure~\ref{fig:adam_curvature} show that when training GANs with Adam on CIFAR-10 many parameters are in the low-curvature regime when the high performance area is reached, particularly for the discriminator. When using $\epsilon=10^{-8}$, Figure~\ref{fig:adam_curvature_small_eps} shows that after the area of high performance is reached, the model exits it and performance degrades; this is consistent with our hypothesis that, due to large updates in areas of low curvature, Adam using a small $\epsilon$ can exit areas of high performance. 
By annealing $\epsilon$ throughout training we can ensure that a larger $\epsilon$ is used areas of high performance, and thus the Adam update is more sensitive to the gradient magnitude, especially for parameters with small gradients, as we show in Figure~\ref{fig:adam_eps_effect}. We show results in GAN training by annealing the Adam $\epsilon$ in Figures~\ref{fig:adam_curvature_anneal1} and~\ref{fig:adam_curvature_anneal2}:
the rate at which the high performance area is left decreases, and for $\epsilon$ annealed to as high as $10^{-2}$, the high performance area is no longer exited.

These results suggest that indeed, even though Spectral Normalisation does not have a purely optimisation effect, its annealing of $\epsilon$ when using the Adam optimiser has a beneficial effect, by avoiding large steps in areas of high performance. Furthermore, these results are consistent with what we observed in the DRL case, where Spectral Normalisation ensures that performance increases for longer in training (Figure~\ref{fig:atari_dqn_hns}).
While here we focused on the effect of $\epsilon$ in the low-curvature regime, we want to emphasise that this is not the only effect of $\epsilon$, but simply one easy to analyse. As $\epsilon$ dampens the size of parameter updates regardless of curvature, increasing $\epsilon$ has an effect on all parameters. From that perspective, increasing $\epsilon$ in training is akin to decreasing to learning rate. We do note significant differences, however: as we show in Figure~\ref{fig:adam_lr_effect}, a small $\epsilon$ in the low-curvature regime always leads to a lack of sensitivity to the gradient magnitude, regardless of learning rate, while changing $\epsilon$ can lead to increased sensitivity to gradient magnitudes. We believe more studies in the role of $\epsilon$ in Adam are needed, specifically in the context of DRL, where it has empirically been shown (here and elsewhere) to play an important role.

\section{Related work}

For an overview of the use of smoothness regularisation and normalisation in deep learning, together with benefits and downsides, see Chapter~\ref{ch:smoothness}. The relaxation of the 1-Lipschitz constraint in the application of Spectral Normalisation has been previously done by~ \citet{gouk2020regularisation}, who relax the Lipschitz constant but in contrast to our approach in DRL, still regularise the entire network. 

Other normalisation methods, such as Weight Normalisation and Batch Normalisation have been studied from an optimisation perspective~\citep{salimans2016weight,understanding_batch_norm}. Unlike Spectral Normalisation however, these normalisation techniques are not hard normalisations, as their parametrisation allows recovering the un-normalised approaches. Weight Normalisation~\citep{salimans2016weight} decouples the learning of the Frobenius norm of the matrix from its direction; in DRL, the authors show that applying Weight Normalisation improves the performance of DQN on a subset of Atari games. When introducing Spectral Normalisation, \citet{miyato2018spectral} compare Spectral Normalisation and Weight Normalisation empirically and in the GAN case obtain significant gains using Spectral Normalisation compared to Weight Normalisation (their Table 2).
The connection between optimisation hyperparameters in Adam and normalisation---as we have done with $\textsc{mulEps}$---has been previously made for Batch Normalisation and Weight Normalisation \citep{van2017l2,hoffer2018norm,arora2018theoretical}; in contrast to our approach, these works require $L_2$ regularisation and obtain the $\epsilon$ scaling effect jointly with a learning rate scaling effect, while we obtained no change on the learning rate (Eq~\eqref{eq:mul_eps}). The connection between $\epsilon$ and low-curvature is connected to existing works which track curvature during training~\citep{martens2015optimizing}.

While optimisation specific issues in DRL have long been overlooked, recent works have studied the negative interplay between temporal difference learning methods such as $Q$-learning and adaptive optimisation methods including RMSprop and Adam \citep{bengio2020InterferenceAG}, while 
others propose optimisers for DRL, but they are either restricted to linear estimators \citep{givchi2014QuasiNT,sun2020adaptive}, or they show no empirical advantages \citep{romoff2020TDpropDJ}.

Inspired by the insights obtained when investigating the effects of Spectral Normalisation in DRL, we also analysed its effects on GANs. A study of the effects of Spectral Normalisation was similarly performed recently, with~\citet{lin2021spectral} exploring the effects of Spectral Normalisation theoretically and uncovering its connections with exploding and vanishing gradients. Their work is related to our approach, which similarly uncovered the effects of Spectral Normalisation on training dynamics, as well as explored the effect of optimisation methods based on Spectral Normalisation, such as \textsc{divOut}, and the interaction with loss function choices and optimisers such as Adam. 

\section{Conclusion}

We investigated the use of Spectral Normalisation in DRL and showed that normalising one layer of the learned neural critic in $Q$-learning agents can substantially improve performance over methods which collate multiple RL specific improvements. We then investigated \textit{why} Spectral Normalisation helps DRL. By using the compositional structure of neural networks we uncovered strong optimisation effects,
including interactions with optimiser hyperparameters. Inspired by these optimisation effects, we developed novel optimisation updates that recover, or even outperform, the effects of Spectral Normalisation in DRL. Equipped with this knowledge we re-examined the effects of Spectral Normalisation in GANs, another domain where Spectral Normalisation has been applied with success, but where the normalisation procedure is applied to the entire network. We observed that for GANs optimisation only changes do not recover Spectral Normalisation performance, and that a strong effect of Spectral Normalisation is avoiding poor-conditioned areas of loss functions. Despite not recovering its performance however, optimisation changes due to Spectral Normalisation can help training when the Adam optimiser is used, particularly for parameters which exhibit low-curvature during training.

\chapter{Geometric Complexity: a smoothness complexity measure implicitly regularised in deep learning}
\label{ch:gc}

 Machine learning methods have long faced the challenge of finding the right balance between fitting the training data and generalising beyond said training data~\citep{fortmann2012understanding}. 
 The need to avoid overly complex models that overfit by memorising training datasets has led to the need to measure model complexity, closely followed by the desire to construct regularisation methods that control said model complexity.
 The aim of complexity measures is thus to capture generalisation (often measured via test set error, where lower is better) via a U-shaped curve: when complexity is low, the model is underfitting the training data and underperforming on the test data leading to a high test error (high bias), but as complexity increases the test error decreases, followed by an increase in test error once the model is overly complex and overfitting (high variance); we have observed this behaviour in Figure~\ref{fig:u_shape_p} of Chapter~\ref{ch:smoothness}.  While this traditional U-shaped curve has been long associated with complexity measures in machine learning, deep learning has led to a few puzzling observations. For deep or over-parametrised networks, overfitting is less of an issue than we would expect under the traditional U-shape curve, and the models both fit and the data and generalise~\citep{neyshabur2014search,zhang2021understanding,deep_double_descent}.  Deep neural networks present an additional challenge for complexity measures, as many existing measures are either too simplistic, such as number of learnable parameters or weight norms, or intractable to compute for the neural network class of models, such as classical measures like Rademacher \citep{Koltchinskii99rademacherprocesses,bartlett2002rademacher} complexity and VC dimensions~\citep{vapnik1971on,vapnik2013nature}.  

 Recent work has also uncovered connections between optimisation and generalisation in deep learning, while previously optimisation was viewed purely from a training objective minimisation perspective; these works broadly fall under the umbrella of implicit regularisation~\citep{neyshabur2017implicit,igr,igr_sgd,symmetry,ma2021sobolev}. Since many existing complexity measures measure the capacity of a family of functions~\citep{Koltchinskii99rademacherprocesses,vapnik1971on,vapnik2013nature}, they cannot capture implicit regularisation effects induced by optimisation inside a model class, such as the implicit regularisation induced by discretisation drift we studied in previous chapters.

 Motivated by the need to reconcile complexity measures and deep learning, and emboldened by our previous study of implicit regularisation and interactions between smoothness and optimisation,
  we make the following contributions:
\begin{itemize}
	\item We introduce a new measure of model complexity for deep neural networks, which we call \textit{Geometric Complexity} (GC). GC captures the complexity of a model rather than function class, can be computed exactly during training, and has connections to Lipschitz smoothness.
	\item We show that GC is connected with or implicitly regularised by many aspects of deep learning, including initialisation, explicit regularisation, and implicit regularisers induced by the discretisation drift of optimisation methods. 
    \item We show that through implicit regularisation induced by discretisation drift, the implicit minimisation of GC in neural network training can be strengthened through hyperparameter choices such as batch size or learning rate.
	\item We show that GC captures the double descent phenomenon, and restores the classical complexity versus performance U-shaped curve.
\end{itemize}

\section{Introduction}

\textbf{Existing complexity measures.} Many existing complexity measures target the complexity of a class of functions. These types of complexity measures range from the very simple, such as number of learned parameters, to more sophisticated, such as Rademacher complexity and VC dimensions~\citep{Koltchinskii99rademacherprocesses,vapnik1971on,vapnik2013nature,bartlett2002model,bartlett2002rademacher}.
While convenient and inexpensive, using the number of learned parameters as a complexity measure does not account for architectures and expressivity; when using the number of parameters to measure the capacity of a neural network, the now famous double descent phenomenon~\cite{belkin2021fear,Belkin15849,deep_double_descent} occurs in deep learning: as the number of parameters exceeds a critical threshold, the test error of the model decreases again, as opposed to increasing as expected under the U-shaped complexity curve. Rademacher complexity~\citep{Koltchinskii99rademacherprocesses,bartlett2002rademacher} and VC dimension do account for model class~\citep{vapnik1971on,vapnik2013nature}, but are geared towards measuring capacity of fitting \textit{random labels}. They thus do not account for the specific task and cannot leverage the inherent structure in the data present in many deep learning problems, and are also challenging to compute for specific deep learning architectures, though bounds do exist~\citep{neyshabur2018role,anthony1999neural}. Additional challenges get presented by the fact that neural networks have been shown to consistently be able to learn mappings to random labels~\citep{zhang2021understanding}, making it hard to distinguish between classes of neural networks.

Metrics for measuring the complexity of a single model rather than model family are available. They range from simple approaches, such as the norm of the learned parameters~\citep{neyshabur2017implicit}, to the more complex such as classification margins~\citep{jiang2018predicting} or flatness~\citep{keskar2016large,jiang2019fantastic}, Kolmogorov complexity \cite{Schmidhuber1997Discovering} and relatedly minimum description length \cite{NIPS1993_9e3cfc48,blier2018description}, the number of linear pieces for ReLU networks \cite{arora2018understanding,serra2018bounding}, but they are not without limitations. Metrics based on parameter norms, such as weight norms, are independent of model architectures and for large models it is not inherently clear how changing the parameter norm changes the model predictions. Norm-based metrics have also been shown to correlate poorly with generalisation, especially when stochastic optimisation is used~\citep{jiang2019fantastic}; classification margins can be challenging to compute for data with a high number of dimensions or classes; while flatness (or equivalently low sharpness) has been shown to be connected with generalisation~\citep{hochreiter1997flat,keskar2016large,jastrzkebski2018relation,igr,jiang2019fantastic}, limitations have been noted~\citep{dinh2017sharp,kaur2022maximum}. We previously discussed flatness---defined there as the leading eigenvalue of the Hessian---as a measure of generalisation in Chapter~\ref{ch:pf}. Complexity measures for distributions over models, such as PAC-Bayes~\citep{alquier2021user}, have also been adapted to neural networks~\citep{dziugaite2017computing}, but they are outside of the scope of this thesis.

\citet{neyshabur2018role} analysed many existing measures of complexity and showed that they increase alongside the number of the hidden units, and thus cannot explain behaviours observed in over-parametrised neural networks, where increasing the number of hidden units does not lead to increased overfitting (their Figure 5).~\citet{jiang2019fantastic} perform an extensive study of existing complexity measures and their correlation with generalisation as well as aim to find causal relationships and find that flatness has the strongest positive connection with generalisation.

\textbf{Explicit and implicit regularisation}. 
Explicit regularisation in deep learning takes many forms, either through adding terms to loss functions or regularising the model architecture~\citep{dropout,yoshida2017spectral,goodfellow2016deep,geiping2021stochastic,balduzzi2018mechanics, mescheder2017numerics, nagarajan2017gradient}. Explicit regularisation can affect model capacity in intricate ways and its effect can strongly depend on the magnitude of regularisation coefficients, with no  straightforward method of translating regularisation effects into restrictions on a model class.
Thus, despite their practical significance in obtaining increased generalisation performance, the effect of common explicit regularisers has not been unified by existing complexity measures.

Recent years have seen an expansion in the discovery of implicit regularisation effects in deep learning, from the implicit regularisation effects of learning rates~\citep{li2019towards,igr,jastrzebski2019break,neyshabur2017implicit,SeongLKHK18,igr_sgd,smith2018dont}, mini-batch optimisation noise~\citep{smith2020generalization,ziyin2021strength,sgd_reg} or initialisation  \citep{understanding2010glorot,gunasekar2018characterizing,he2015delving,li2018learning,nagarajan2019generalization,zhang2020type,zou2020gradient}. We previously explored implicit regularisation induced by the discretisation drift of optimisers in Chapters~\ref{ch:dd_gans} and~\ref{ch:bea_stochasticity}. 
Implicit regularisation effects have been largely unexplained by existing complexity measures; by construction complexity measures of function classes rather than individual models cannot capture the effects of optimisation inside a function class. Novel connections between optimisation hyperparameters such as learning rates and per-model complexity measures such as sharpness have been recently found~(\citet{cohen2021gradient} and Chapter~\ref{ch:pf}), but more investigation into its connection with stochasticity is needed~\citep{kaur2022maximum}. 
The role of network depth and overparametrisation as an implicit regulariser has also been studied~\citep{neyshabur2018role,li2018learning,du2018gradient}, and has been shown to play an important role in the generalisation properties of neural networks as well as the double descent phenomenon; in Chapter~\ref{ch:smoothness} we postulated a connection between depth, overparametrisation, and smoothness, which we will make concrete here.

\textbf{Smoothness}: We highlighted the benefits of smoothness for neural networks extensively in Section~\ref{sec:benefits} and intuited its deep connection with generalisation and double descent. Existing works have shown that generalisation bounds can be obtained from smoothness constraints~\citep{sokolic2017robust}. The connection between smoothness and many implicit and explicit regularisers has been highlighted in Chapter~\ref{ch:smoothness}; here we will define a smoothness based complexity measure and make this association concrete through theoretical arguments and experimentation. 

\section{Geometric Complexity}
\label{sec:gc_def}

We are now ready to define a novel metric of complexity for neural models, with the aim of finding a measure that can capture some of the implicit and explicit regularisation effects previously discussed.
\begin{definition}[Geometric Complexity] For a neural network $g(\cdot;{\vtheta}): \mathbb{R}^I \rightarrow \mathbb{R}^O$ with parameters $\vtheta \in \mathbb{R}^D$, we write $g(\cdot;{\vtheta})= \tau(f(\cdot;{\vtheta}))$ where $\tau$ is the last layer activation function (such as the softmax for multi-label classification). Given a dataset $\mathcal{D}$ with elements $\vx$ i.i.d. realisations of random variable $\mathcal{X}$, we define Geometric Complexity (GC) as
\begin{align}\label{eqn:geometric_complexity}
    GC(f;\vtheta, \mathcal{D}) = \frac{1}{|\mathcal{D}|}\sum_{\vx\in \mathcal{D}} \left\|\jacparam{\vx}{f(\vx; \vtheta)}\right\|_2^2,
\end{align}
where $||\vv||_2$ denotes the Frobenius norm.
\end{definition}

GC measures the complexity of a single model $f(\cdot; \vtheta)$, not a model class, can be computed efficiently, and depends on the task at hand through the training dataset $\mathcal{D}$. While GC is defined based on the training data, it is an estimate of an expectation:
\begin{remark} GC is the unbiased estimate of the underlying expectation under $p(\vx)$ the density of $\mathcal{X}$:
\begin{align}
    \mathbb{E}_{p(\vx)} \left[\left\| \jacparam{\vx}{f(\vx; \vtheta)}\right\|_2^2 \right] = \int \left\| \jacparam{\vx}{f(\vx; \vtheta)}\right\|_2^2 p(\vx) d \vx.
\end{align}
\end{remark}
GC can be connected to Lipschitz smoothness---which we discussed at length in Chapter~\ref{ch:smoothness}---using the following bound:
\begin{remark}[Connection to Lipschitz smoothness] If $K$ is the Lipschitz constant of $f(\cdot;\vtheta):\mathbb{R}^I \rightarrow \mathbb{R}^O$ w.r.t. the Frobenius norm then we have that:
\begin{align} 
GC(f;\vtheta, \mathcal{D}) &= \frac{1}{|\mathcal{D}|}\sum_{\vx\in \mathcal{D}} \left\|\jacparam{\vx}{f(\vx; \vtheta)}\right\|_2^2 \\ 
&\le \min(I, O) \frac{1}{|\mathcal{D}|}\sum_{\vx\in \mathcal{D}} \left\|\jacparam{\vx}{f(\vx; \vtheta)}\right\|_{op}^2 
\le \min(I, O) K^2,
\end{align}
where $\norm{A}_2$ and $\norm{A}_{op}$ denote the Frobenius and operator norms of $A$, respectively.
\label{rem:gc_lip}
\end{remark}
This result follows from $\norm{A}_2^2 \le \text{rank}(A) \norm{A}_{op}^2 \le \min(I, O) \norm{A}_{op}^2, \hspace{1em} \forall A \in \mathbb{R}^I \times \mathbb{R}^O$ and $\left\|\jacparam{\vx}{f(\vx; \vtheta)}\right\|_{op}^2  \le K^2$. For many problems of concern we have that $\min(I, O) = O$ and thus Remark~\ref{rem:gc_lip} exposes that the tightness of the bound between GC and Lipschitz smoothness depends on the number of outputs of the network.

While the Lipschitz smoothness of a function provides an upper bound on GC, there are a few fundamental differences between the two quantities. Firstly, GC is data dependent: a model can have low  GC while having high Lipschitz constant due to the model not being smooth in parts of the space where there is no training data. Measures on the entire space, such as Lipschitz constants, are unlikely to explain implicit and explicit regularisation effects when many of the regularisers are applied around data instances. 
Secondly, GC can be computed exactly given a model and dataset, while for when estimating the Lipschitz constant of neural networks loose upper and lower bounds that rely on function composition are often used~\citep{virmaux2018lipschitz,combettes2019lipschitz}.
~\citet{fazlyab2019efficient} provide an algorithm with tighter bounds by leveraging that activation
functions are derivatives of convex functions and cast finding the Lipschitz constant as the result of a convex
optimisation problem;
however, their most accurate approach scales quadratically with the number of neurons and only applies to feed forward networks.
~\citet{sokolic2017robust} provide an upper bound for the Lipschitz constant of a neural network with linear,
softmax and pooling layers restricted to input space $\mathcal{X}$ via $\norm{f(\vx) - f(\vy)}_{2} \le \sup_{\vz \in \text{convex\_hull}(\mathcal{X})} \norm{\frac{d f}{d\vz}}_2 \norm{\vx-\vy}_2,  \hspace{1em} \forall \vx, \vy \in \mathcal{X}$,
but empirically resort to layer-wise bounds.
In contrast, GC can be computed exactly regardless of network architecture and can be easily implemented using modern machine learning software frameworks.

\subsection{GC for deep linear models}

Consider a deep linear network $lin(\vx, \vtheta) = \vW_n \dots \vW_1 \vx + \vb$, with $\vtheta = \{\vb\}  \union \{\vW_i, i \in \{1, \dots, n\}\}$ We then have that
\begin{align}
GC(lin;\vtheta, \mathcal{D}) = \| \vW_n \dots \vW_1\|_2^2.
\end{align}
This shows that the GC of an affine transformation does not depend on the data and recovers the standard Frobenius norm, which is often used as a regulariser, particularly in ridge regression.

\subsection{GC for rectifier feed-forward networks}
\label{sec:rectifier_nets_gc}

Networks with rectifier activations, such as ReLu~\citep{agarap2018deep} and Leaky Relu~\citep{maas2013rectifier}, are piecewise affine models, that is for each $\vx_i$ we have that $f(\vx_i;\vtheta) = \vA_i \vx_i + \vb_i$, where $\vA_i$ and $\vb_i$ depend on where in the partition of the input space input $\vx_i$ falls in. We can then write
\begin{align}
GC(f;\vtheta, \mathcal{D}) &= \frac{1}{|\mathcal{D}|}\sum_{\vx\in \mathcal{D}} \left\|\jacparam{\vx}{f(\vx; \vtheta)}\right\|_2^2 = \sum_{p=1}^P \frac{n_p}{|\mathcal{D}|} \|\vA_p\|_2^2, 
\label{eq:relu_gc}
\end{align}
where $n_p$ is the number of times an input is in the partitioning corresponding to the linear piece $\vA_p \vx + \vb_p$ in the piecewise model. Eq~\eqref{eq:relu_gc} shows that for models with rectifier activations, GC can be approximated by using batches, instead of the entire dataset, as long as the batch size is large enough to include representation from many piecewise branches of the network. 

To further understand the form of GC for rectifier models, let's consider a simple two-layer network, where we have $f(\vx; \vtheta) = \vW_2 a(\vW_1 \vx + \vb_1) + \vb_2$ with $\vtheta = \{\vW_1, \vW_2, \vb_1, \vb_2\}$. Denote $\vh = a(\vz) = a(\vW_1 \vx + \vb_1)$, where $a$ is the rectifier activation function. We then have
\begin{align}
\jacparam{\vx}{f} = \vW_2 \diag(a'(\vz)) \vW_1 ,
\label{eq:gc_two_layers}
\end{align}
where $\diag(\vv)$ is the diagonal matrix with diagonal entries given by $\vv$. Since $\left(\diag(a'(\vz)) \vW_1\right)_{i,j} = a'(\vz_i) \vW_{i,j}$ and since in the case of rectifiers $a'(\vz_i)$ is a piecewise constant function (such as $a'(x) = 0$, if $x < 0$ or $1$ otherwise for ReLu activations), we see the connection between  $L_2$ regularisation and weight decay in rectifier networks, and how decreasing the norm of the weight matrices will decrease GC. 
We observe that unlike weight norms, GC is sensitive to the impact the weight has on the output of the network. This can be most clearly seen for the ReLU activation, where the activation derivative $a'$ can be $0$ and thus for units that do not contribute to the output of a data instance, no norm penalty is imposed.
Consider the case of unit $\vh_i = 0$, where $\vz_i  = {\left(\vW_1\right)}_{i}^T \vx + {(\vb_1)}_i < 0$. Then the entries of row ${\left(\vW_1\right)}_{i}$, which don't contribute to the model output, will not be penalised by $\norm{\jacparam{\vx}{f}}$ in GC, since they are modulated by $a'(\vz_i) = 0$ in Eq~\ref{eq:gc_two_layers}. The same can be said about the column $i$ of $\vW_2$, which does not contribute to the model output as $\vo = \vW_2 \vz + \vb_2$. Thus, GC is robust to weights that do not contribute to model output, such as when a weight matrix has an entry with negative but large absolute value, which leads to a negative pre-activation and a zero hidden unit for all inputs in the dataset. In contrast, the weight norm will be heavily skewed by such a weight due to its large absolute value; this insight might help explain why parameter norm measures are not indicative of generalisation performance~\citep{jiang2019fantastic}.

\subsection{GC for convolutional and residual layers}

Since convolutional layers are linear operators, the form of GC when convolutional layers are used is that of Eq~\eqref{eq:relu_gc}, where the linear operator $\vA_p$ will be determined by the linear operators of the convolutional layers. A significant difference between convolutional and linear layers will be that each parameter of a convolutional filter will appear repeatedly in the linear operator present in GC, which further highlights the difference between GC and weight norms discussed above.

Residual layers~\citep{he2016deep} have the form of $a(\vM \vx + \vx)$, where $\vM$ is a composition of convolutional layers followed by the activation function $a$, usually a rectifier~\citep{nair2010rectified,maas2013rectifier}. Alternatively, they can use an additional projection $\vR$ leading to  $a(\vM \vx + \vR \vx)$.  Thus, networks rectifier activations with residual layers will also be piecewise linear models and Eq~\eqref{eq:relu_gc} holds. 

\section{GC increases as training progresses}

We now focus on the value of GC at network initialisation, and we show that for deep networks with rectifier activations and commonly used initialisers, the GC of the model is close to $0$ at initialisation. To see why, we again leverage the piecewise nature of networks with rectifier activations, together with the fact that network biases are initialised to be $0$: at initialisation rectifier MLPs can be written as $f(\vx; \vtheta) = \vW_n \vP_{n-1} \vW_{n-1} \dots  \vP_1 \vW_1 \vx$ where $\vP_i = \diag(a'(\vz_i)) = \diag(a'(\vW_i \vz_{i-1}))$.
For ReLu networks, since the diagonal entries of $\vP_i$ will be 0 if $\vz_{i-1}$ is negative, as the depth of the network increases, the chance that the output is not 0 diminishes. 
Consider the 1-dimensional case where $f(x; \vtheta) = w_n a(w_{n-1} \dots a(w_1 x))$ and $a$ is the ReLu activation function. If we assume standard Gaussian initialisations of the form $w_i \sim \mathcal{N}(0, \sigma_i^2)$ we have that $P(w_1 x > 0) = \tfrac{1}{2}$, thus $P(a(w_1 x) \ne 0) = \tfrac{1}{2}$. With similar arguments, $P(a(w_2 z_1) \ne 0| z_1 \ne 0) = \tfrac{1}{2}$ leading to $P(a(w_2 z_1) \ne 0) = \frac{1}{2}^2$. By induction we can then show that for $n$ layers
 $P(f(x; \vtheta) \ne 0) = \frac{1}{2^n}$; this shows that as we add one layer, the probability of the output of the network not being $0$ is exponentially decreasing.
When we consider higher dimensional weight matrices, the chance of hitting exactly $0$ decreases, but the obtained function values will be small due to the decreased chance of consistently obtaining positive numbers under the random initialisation. Thus, we expect GC, as the average norm  of $\jacparam{\vx}{f(\vx; \vtheta)} = \vW_n \vP_{n-1} \vW_{n-1} \dots  \vP_1 \vW_1$, to be low at initialisation and decrease as the number of layers in the network increases. In Figure~\ref{fig:initialization_output} we confirm this by looking at the effect of depth on the output of an MLP with input dimensions given by those of input dimension $I=150528 = 224 \times 224 \times 3$ and $O=1000$ (as from the Imagenet dataset), and show how both the mean and the maximum output obtained from 100 samples are decreasing as the number of layers increases.
 \begin{figure}[t]
  \centering
  \begin{subfloat}[Mean model output.]{
 \includegraphics[width=0.45\textwidth]{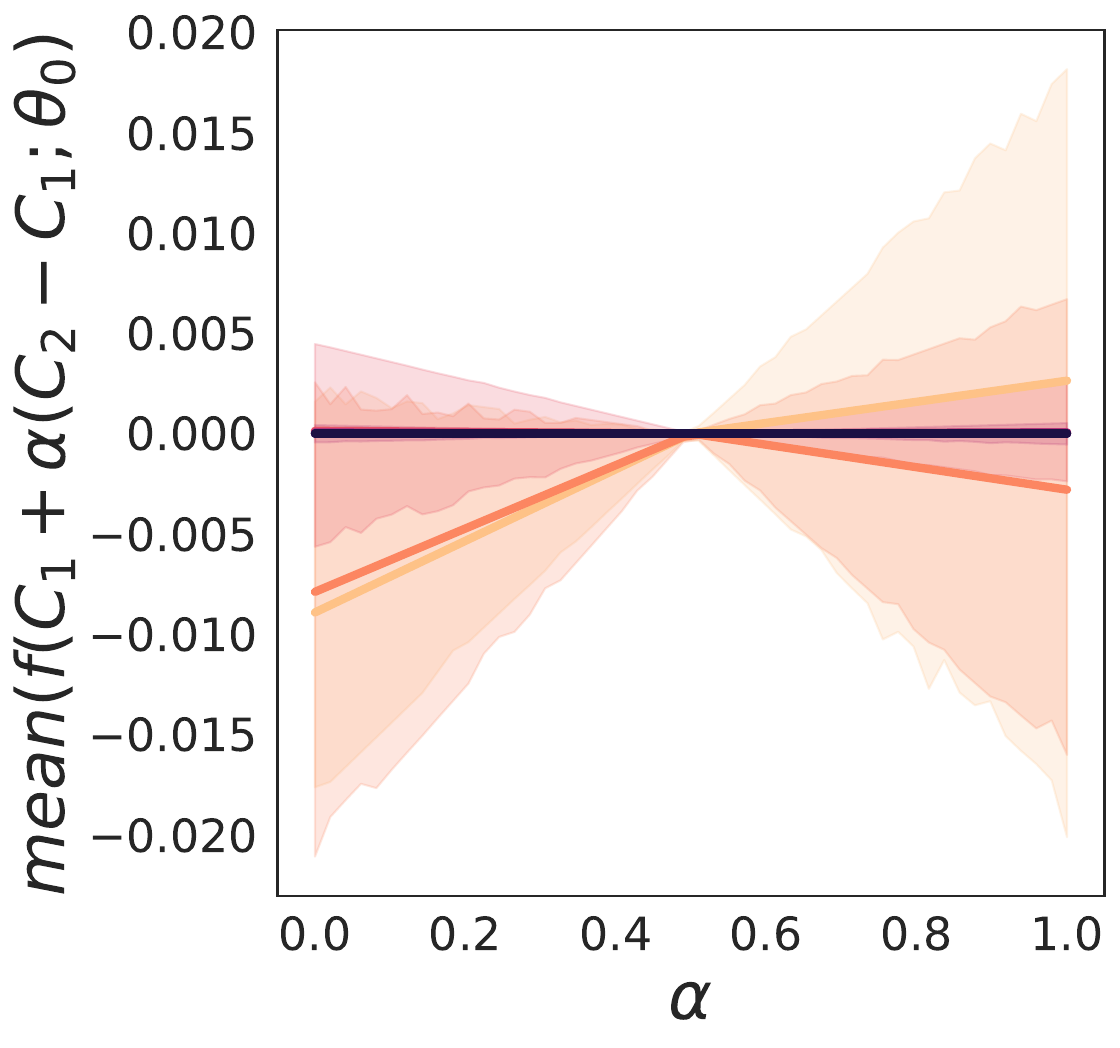}
 \label{fig:init_mlp_mean}
  }
  \end{subfloat}
    \begin{subfloat}[Max model output.]{
 \includegraphics[width=0.45\textwidth]{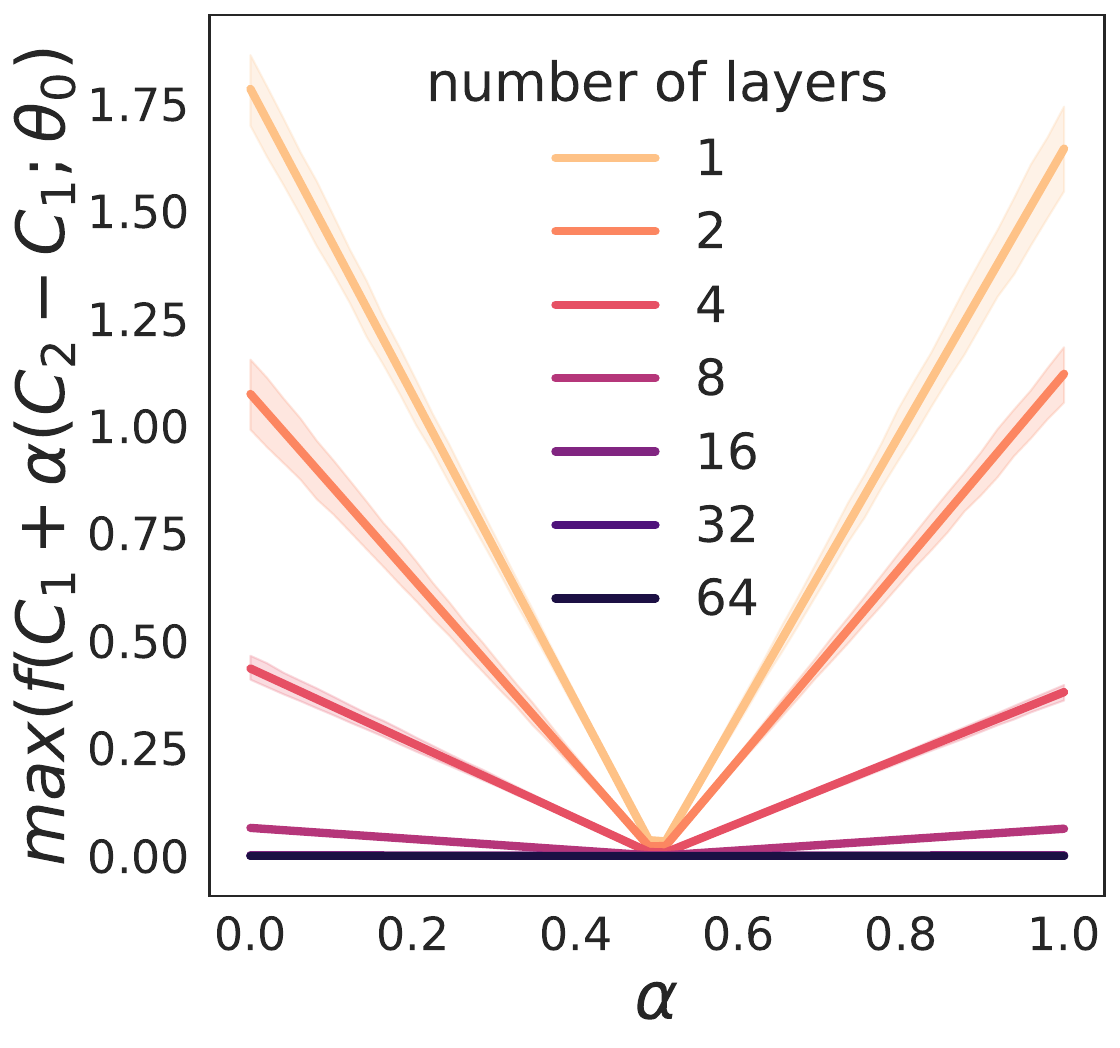}
 \label{fig:init_mlp_max}
  }
  \end{subfloat}
  \caption[Neural networks initialise close to the 0 constant function.]{Initialisation of MLPs. Outputs given by $f(\vC_1 + \alpha(\vC_2 - \vC_1); \vtheta)$ where $\vC_1 \in \mathbb{R}^I$ with every entry equal to 1 and $\vC_2 = - \vC_1$ where $f$ is an MLP with 500 neurons per layer; the input dimension $I=150528 = 224 \times 224 \times 3$. The mean model outputs~\subref{fig:init_mlp_mean} and max model outputs~\subref{fig:init_mlp_max} are closer to 0 at initialisation as depth increases. Uncertainty is obtained by using multiple seeds to initialise the network.}
  \label{fig:initialization_output}
\end{figure}
To empirically assess the effect of the number of layers on GC at initialisation, we use MLPs initialised either with the default initialisation in many deep learning packages, which we will call `standard initialisation' (with weights sampled from a truncated normal distribution bounded between -2 and 2 and scaled by the square root of number of incoming units), and Glorot initialisation~\citep{glorot2010understanding}. We show in Figure~\ref{fig:gc_initialisation} that the value of GC at initialisation decreases as the number of layers increases.
These results show deep MLPs are smoother at initialisation, in that they are less sensitive to the value of the input and are close to the constant function $0$; this is a useful initialisation to start with as otherwise the training procedure might have to undo the effects of a bad initialisation (where for example inputs from the same class might be initialised to have very different output logits), and furthermore increases reliability as training schemes become less dependent on initialisations. We also observe that GC captures this effect at initialisation, and helps explain why deeper models can better model the training data.

 \begin{figure}[t]
  \centering
  \includegraphics[width=0.5\textwidth]{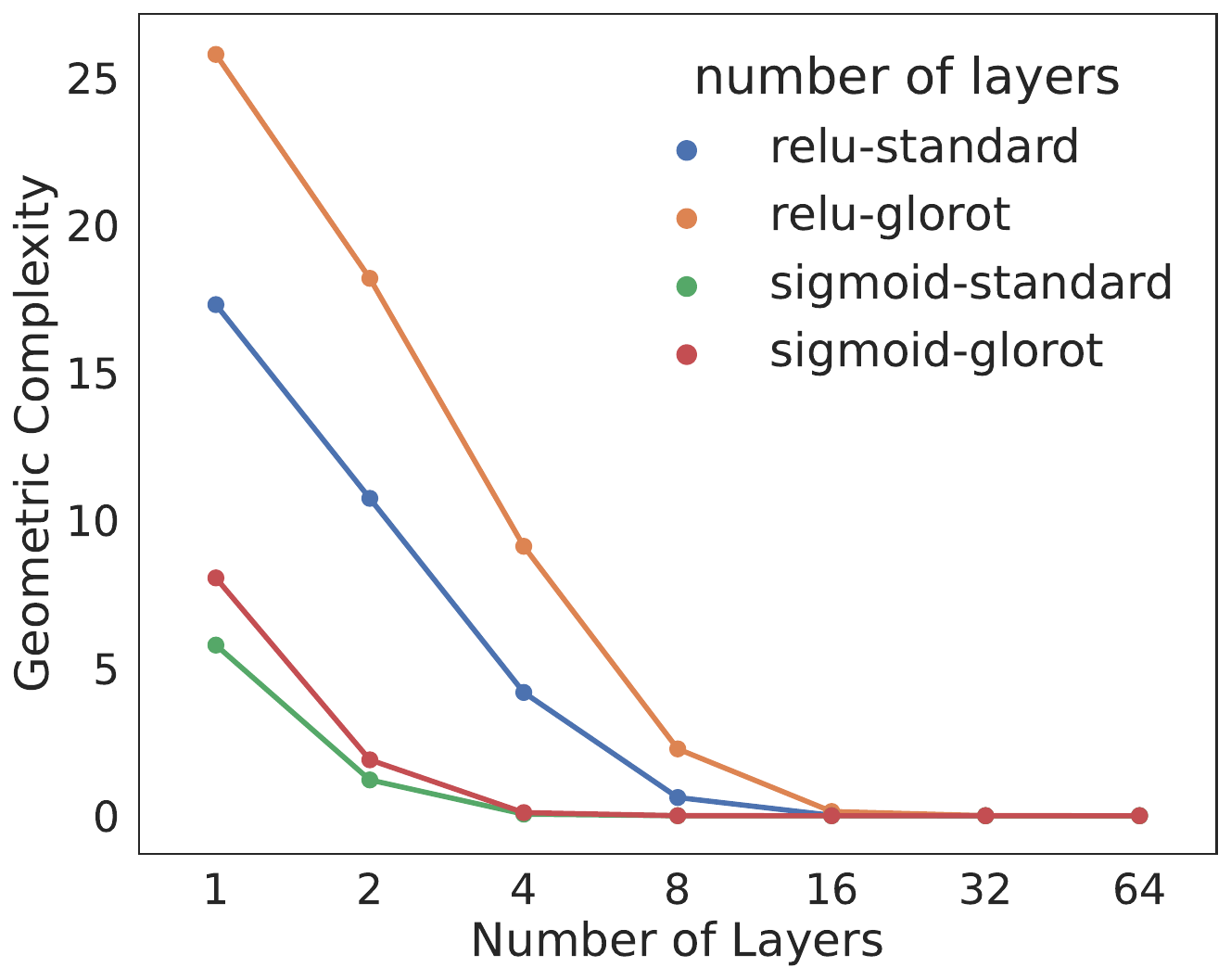}
  \caption[Geometric Complexity (GC) at initialisation decreases as the number of layers in the network increases.]{Geometric Complexity (GC) at initialisation decreases as the number of layers in an MLP increases; this reflects the initialisation phenomenon showcased in Figure~\ref{fig:initialization_output}. This effect shows that the deeper the MLP, the smoother it is at initialisation.}
  \label{fig:gc_initialisation}
\end{figure}

We have seen how deep MLPs initialised with commonly used initialisation schemes start training at low (close to 0) GC. We also observe that as training progresses and the model captures the training data, its GC increases; we see this in Figure~\ref{fig:training_sequence} and we will consistently observe this when training larger models as well (see Figures~\ref{fig:gc_learning_rate} and~\ref{fig:gc_batch_sizes}). During training however, deep networks do not learn overly complex functions, even as GC increases; as shown in Figure~\ref{fig:training_sequence}, even when using a large MLP to train 10 data points, the model fits the data without overfitting. We now turn to explore the potential mechanisms for this observation.

\begin{figure}[t]
  \centering
  \begin{subfloat}[0 epochs.]{
    \includegraphics[width=0.24\linewidth]{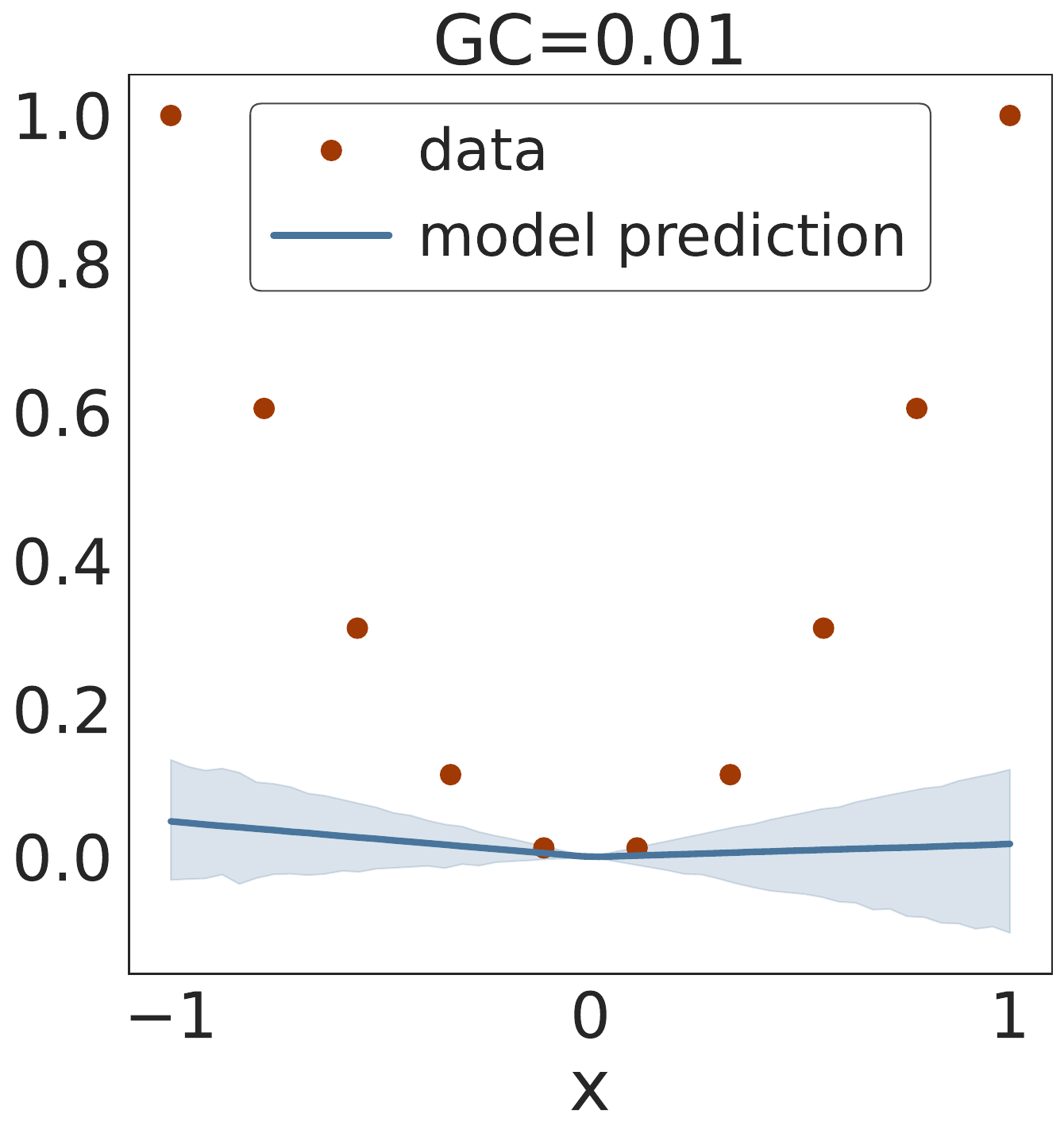}%
  }
  \end{subfloat}%
  \begin{subfloat}[10 epochs.]{
    \includegraphics[width=0.24\linewidth]{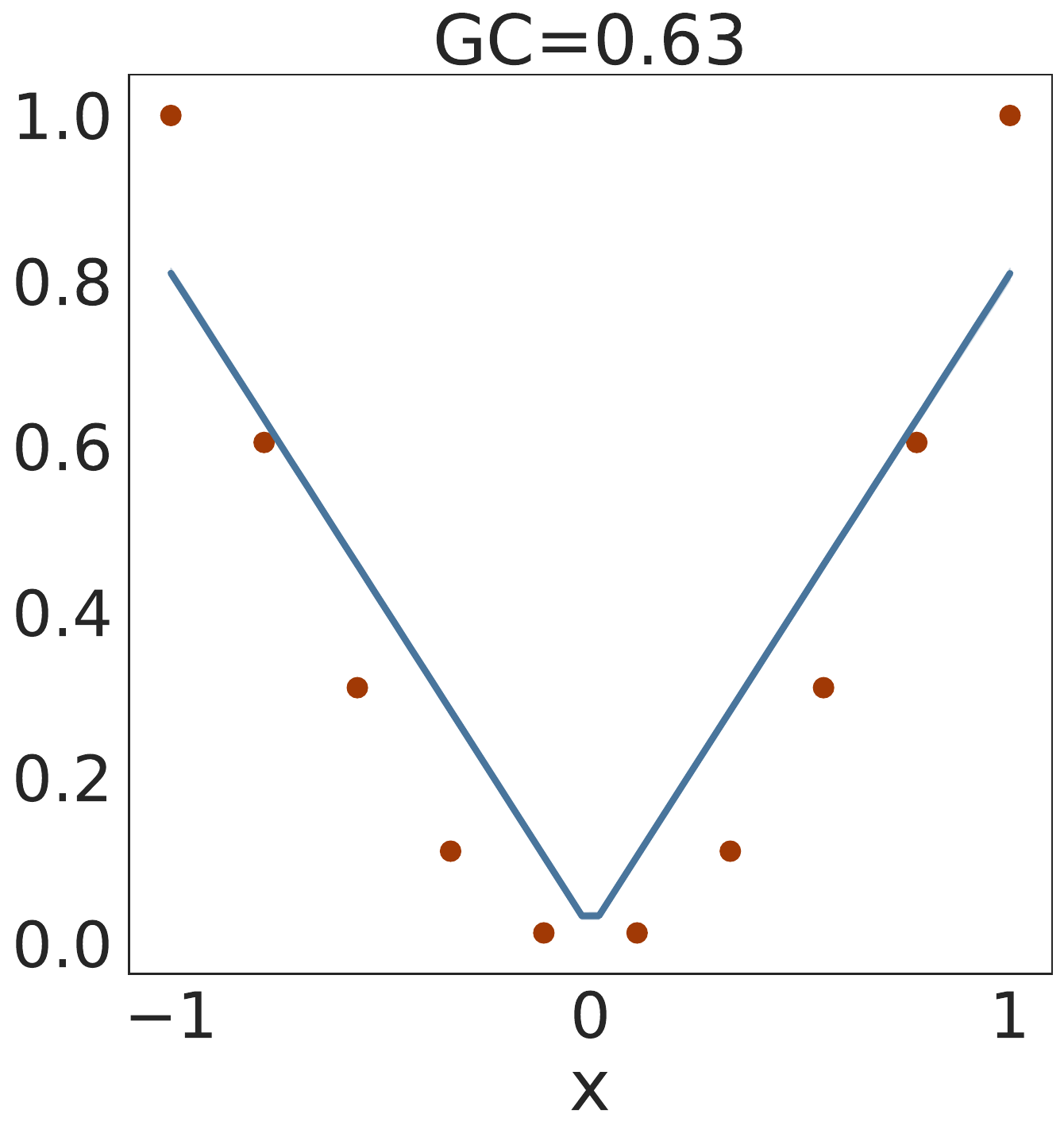}%
  }
  \end{subfloat}%
   \begin{subfloat}[1000 epochs.]{
    \includegraphics[width=0.25\linewidth]{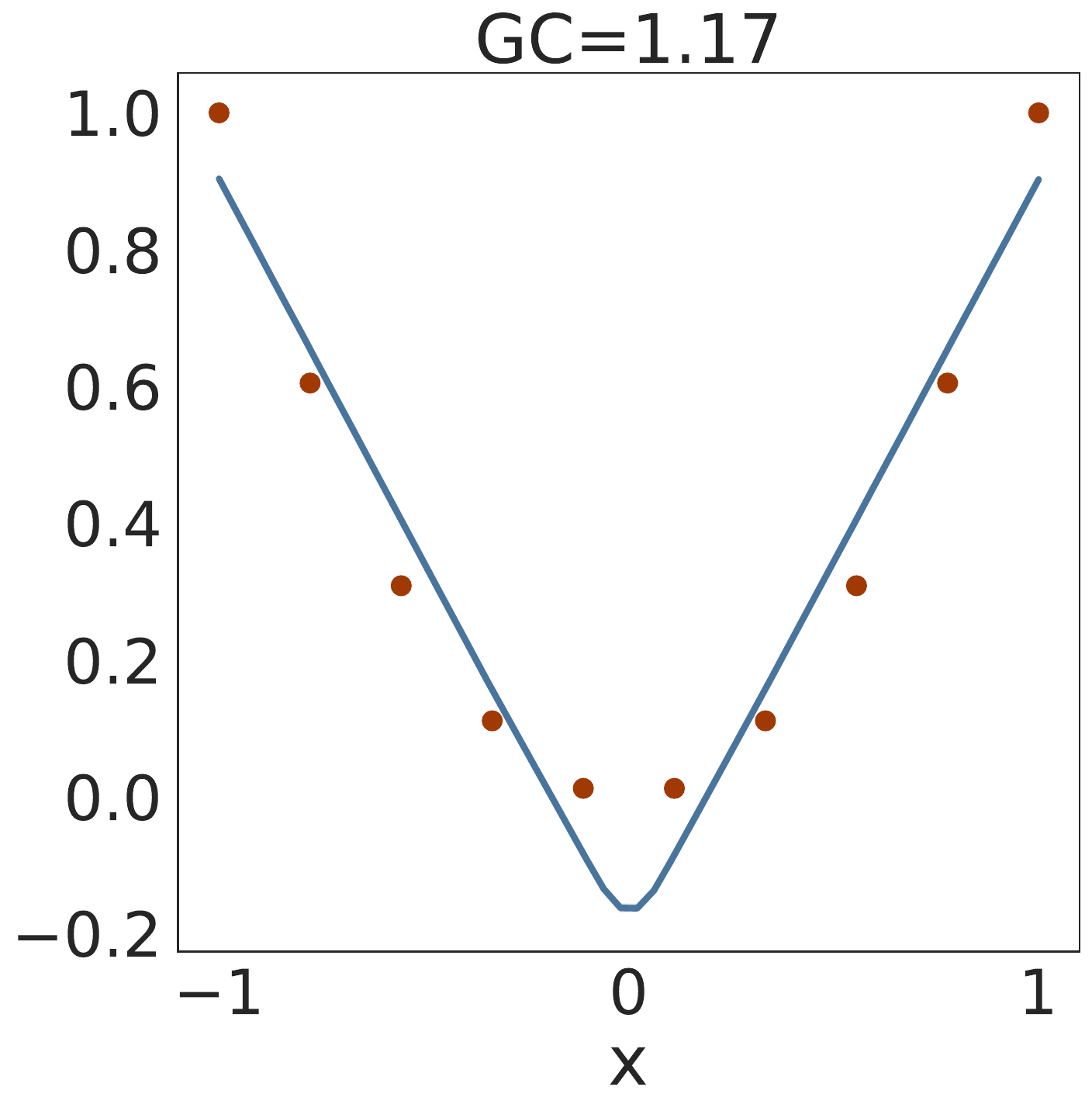}%
  }
  \end{subfloat}%
  \begin{subfloat}[10000 epochs.]{
    \includegraphics[width=0.24\linewidth]{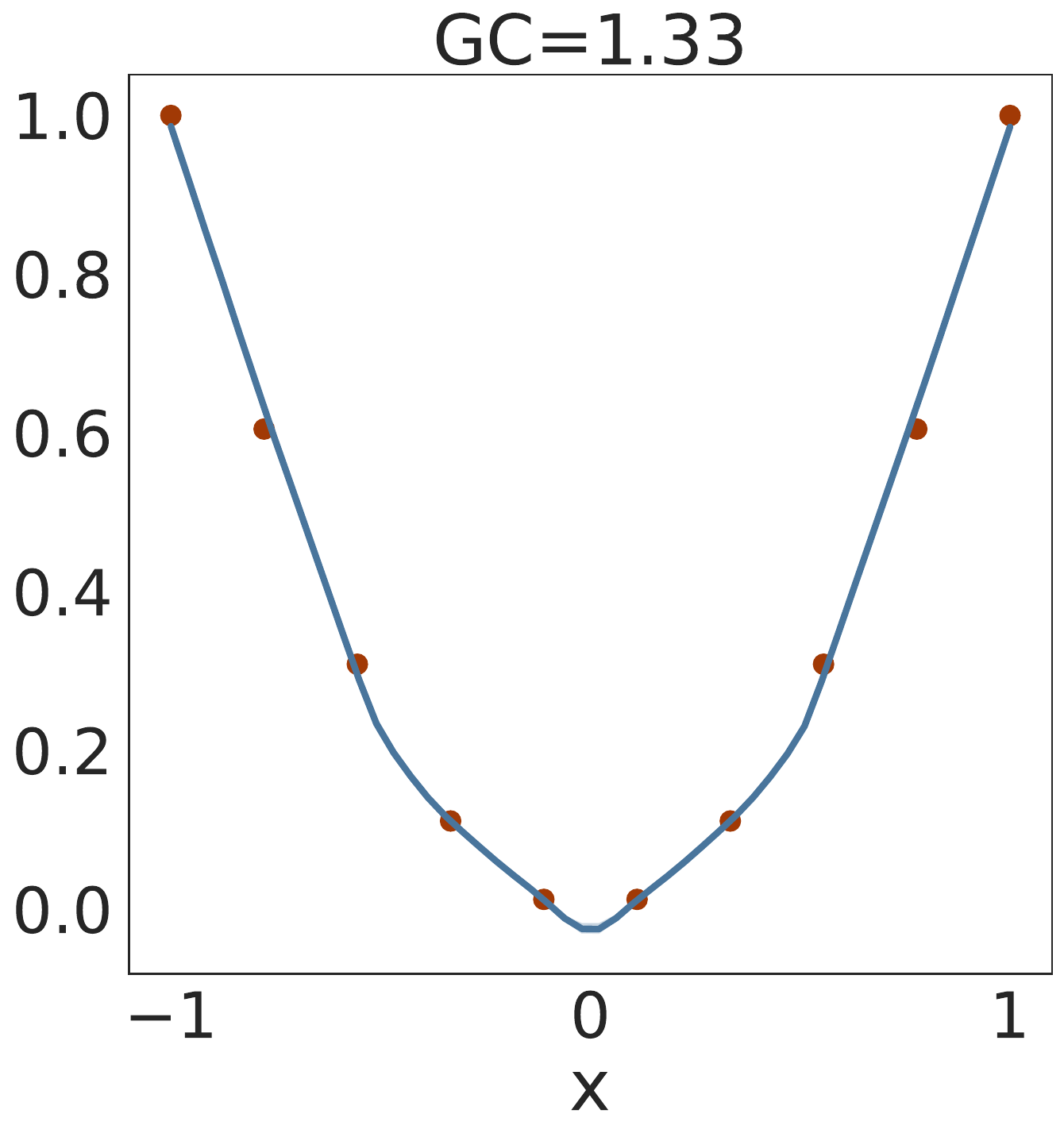}%
  }
  \end{subfloat}%
  \caption[Geometric Complexity (GC) increases as training progresses.]{For a large MLP fitting 10 points in 1 dimension, the GC of the function being learned gradually grows in training. Despite having capacity to overfit, the model learns a simple function while fitting the data.}
  \label{fig:training_sequence}
\end{figure}

\section{GC and regularisation}

We now highlight how many implicit and explicit regularisation methods regularise GC, despite not targetting it directly. To disentangle the effects studied
here from optimisation choices, we use stochastic gradient descent without momentum to train all
models unless otherwise specified. We also study the impact of a given training heuristic on GC in isolation
of other techniques to avoid masking effects. For this reason, we do not use data augmentation or
learning rate schedules. 

\subsection{Explicit regularisation}

\textbf{$L_2$ norm regularisation}. Section~\ref{sec:rectifier_nets_gc} has shown the connection between GC and weight norms in the case of rectifier neural networks. $L_2$ norm regularisation adds the regulariser $\zeta \sum_{l=1}^N \norm{\vW_l}_2^2$ with $\zeta > 0$ to the loss function, minimising the magnitude of each weight parameter. This results in a decreased GC in the case of rectifier neural networks where the GC can be written using the form in Eq~\eqref{eq:relu_gc}, where the value of each piecewise linear function will be determined by the magnitude of the network's weights. While this is not a formal argument, we confirm in Figure~\ref{fig:l2_explicit_gc} that as the $L_2$ regularisation strength increases, GC decreases.
 \begin{figure}[t!]
  \centering
    \begin{subfloat}[$L_2$.]{
    \includegraphics[height=0.37\linewidth]{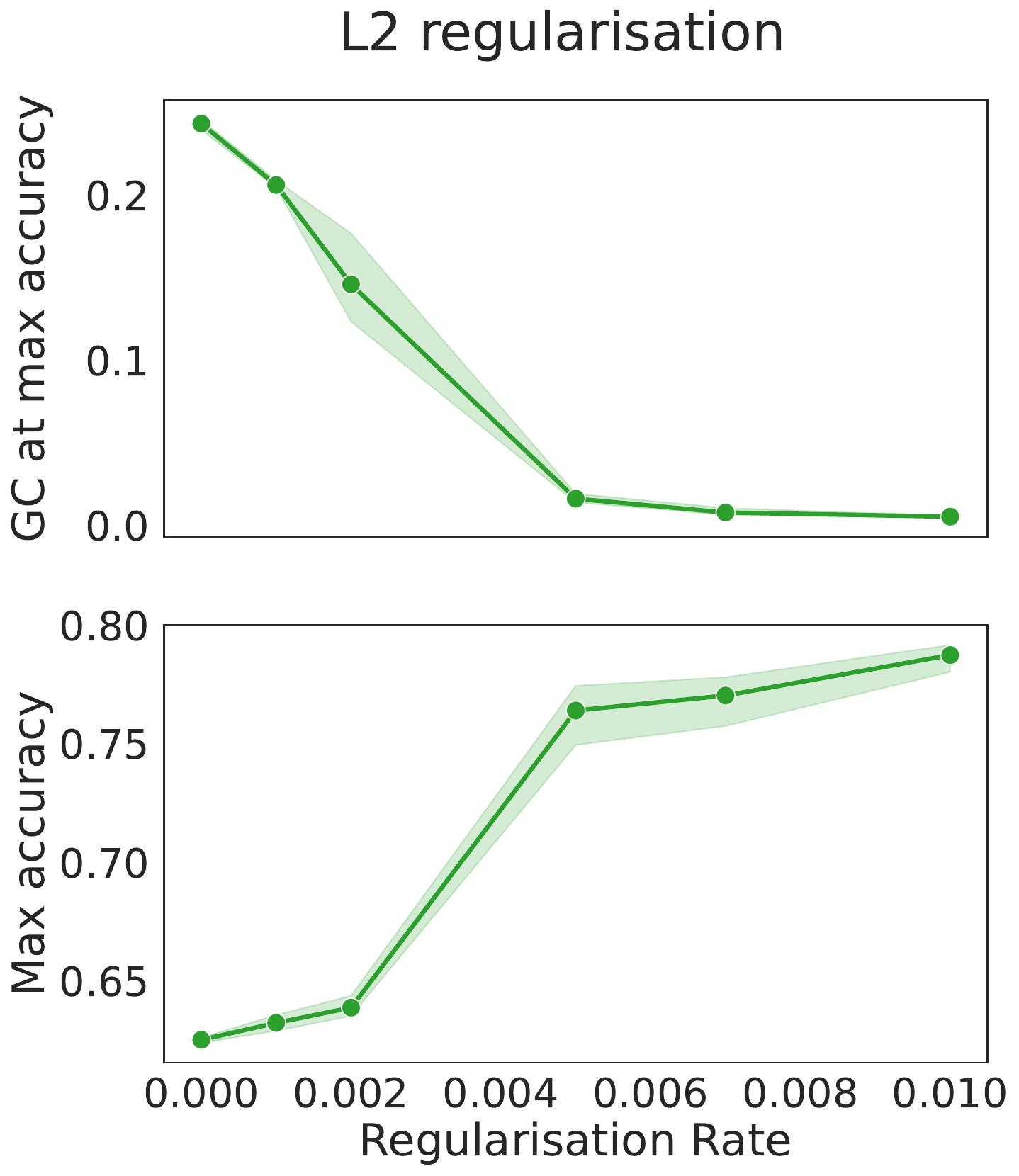}
    \label{fig:l2_explicit_gc}
  }
  \end{subfloat}%
  \begin{subfloat}[Gradient norm.]{
  \includegraphics[height=0.37\linewidth]{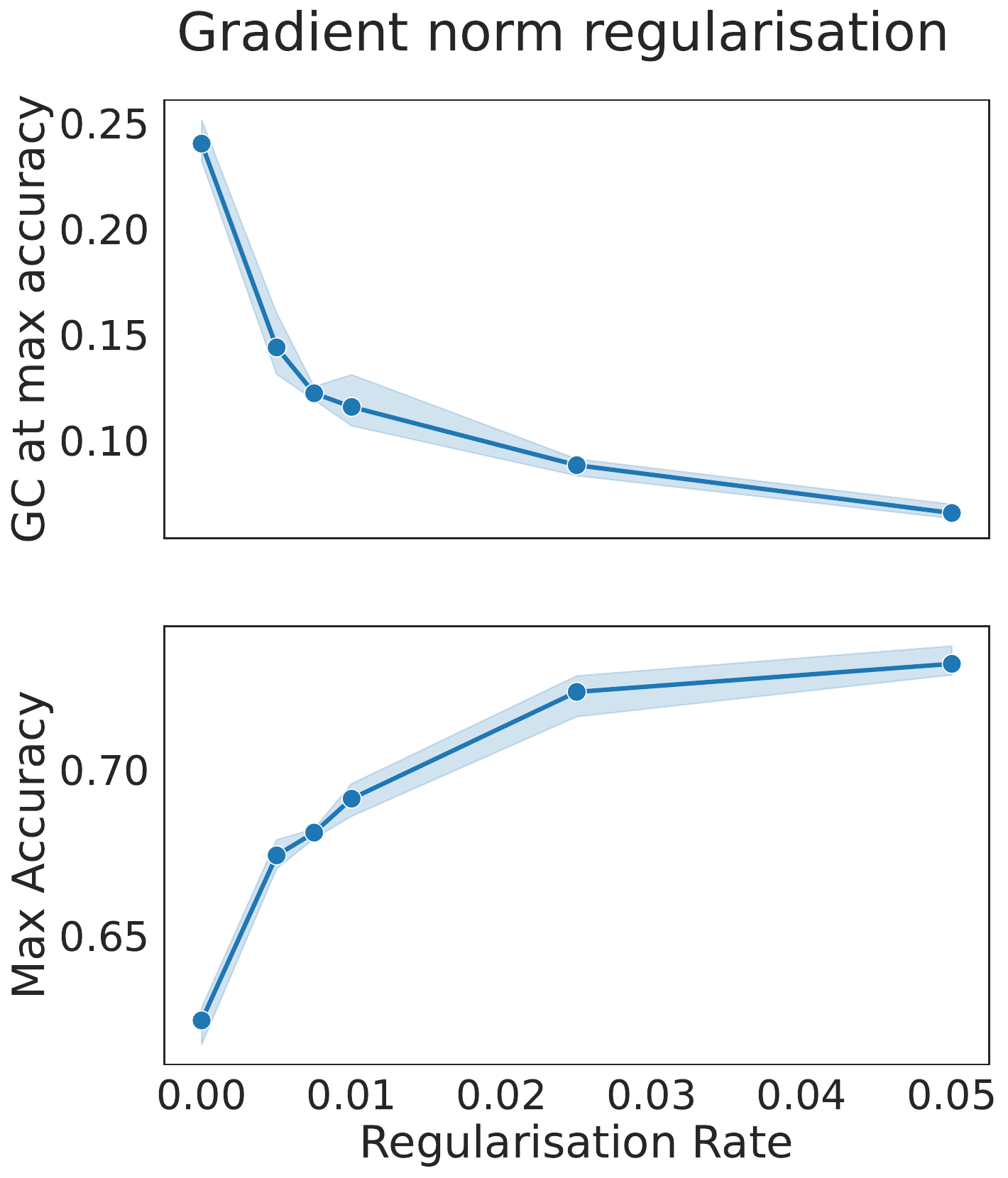}
      \label{fig:gradient_norm_explicit}
  }
  \end{subfloat}%
  \begin{subfloat}[Spectral.]{
    \includegraphics[height=0.37\linewidth]{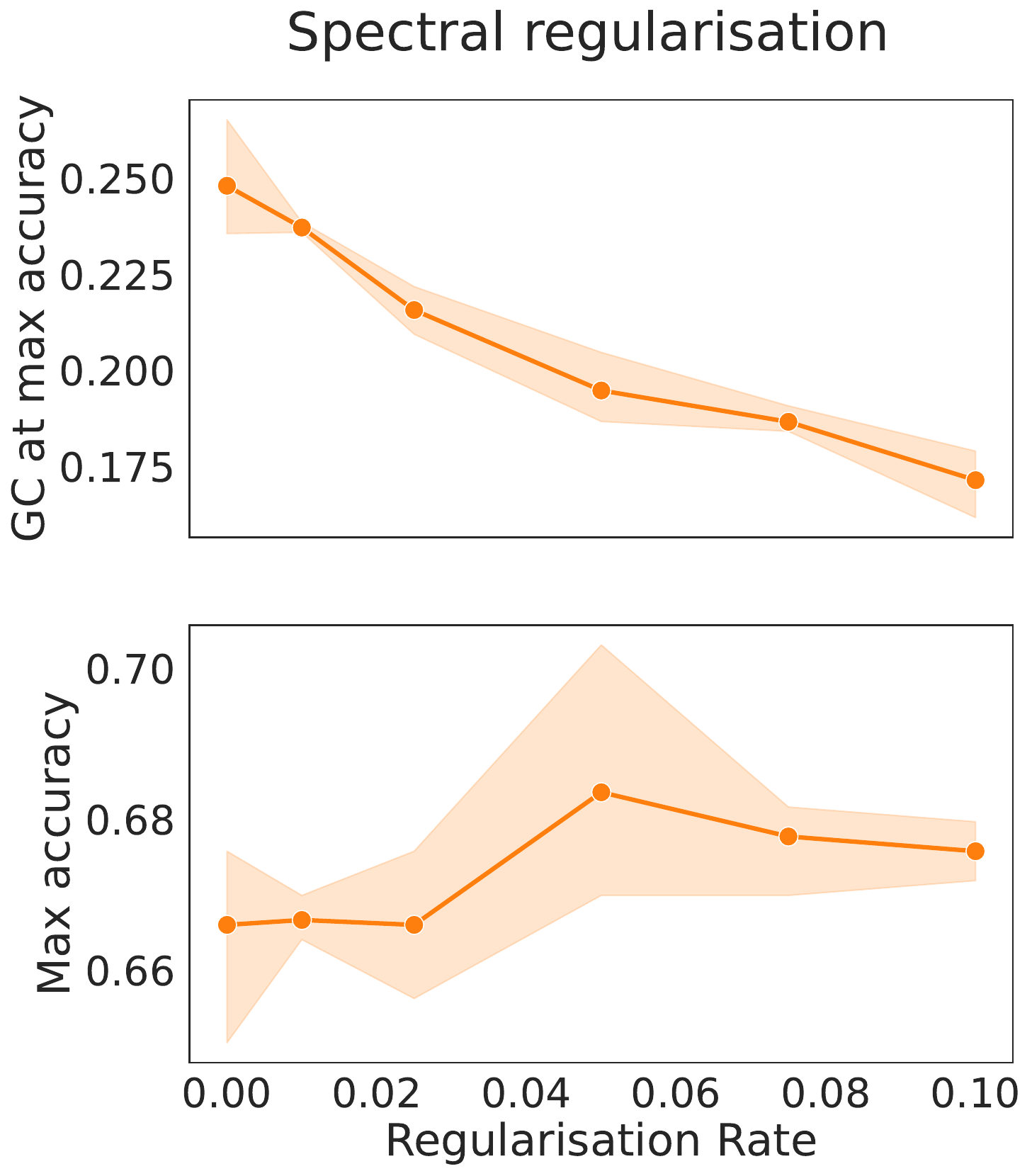}
     \label{fig:spectral_explicit}
  }
  \end{subfloat}%
  \caption[Common explicit regularisation methods lead to decreased GC when using vanilla stochastic gradient descent.]{Common explicit regularisation methods lead to decreased GC when using vanilla stochastic gradient descent to train a Resnet-18 model on CIFAR-10. GC decreases with increased regularisation strength. Similar trends are observed when momentum is used, as shown in Section~\ref{sec:gc_exp_app} of the Appendix, where we also show loss curves.}
  \label{fig:gc_explicit_regularisation}
\end{figure}

\noindent \textbf{Gradient norm regularisation}. Another popular form of regularisation is to add an explicit regulariser with the norm of the gradient $\norm{\nabla_{\vtheta} E(\vtheta)}_2^2$ to the loss function \cite{geiping2021stochastic,igr_sgd,balduzzi2018mechanics, mescheder2017numerics, nagarajan2017gradient,odegan}. While GC is concerned with gradients with respect to inputs, and the gradient norm regulariser penalises the gradient norm with respect to parameters, the two are connected via backpropagation; we previously used neural network structure and backpropagation to connect the effect of an input smoothness regularisation technique---Spectral Normalisation---to scaling of gradients with respect to parameters in Chapter~\ref{ch:rl}. As we see in backpropagation Eqs~\eqref{eq:backprop_first}---\eqref{eq:backprop_last}, decreasing the gradient norm $\norm{\nabla_{\vtheta} E(\vtheta)}_2^2$ can be achieved by decreasing the magnitude of the weight matrix entries or that of the activation derivative $a'(\vz)$, which for ReLu networks can be achieved with increased sparsity; both of these will lead to a decrease in GC. We show in Figure~\ref{fig:gradient_norm_explicit} that indeed, increased gradient norm regularisation consistently leads to a decrease in GC.

\noindent\textbf{Spectral Normalisation/regularisation}. GC is deeply connected to Lipschitz smoothness; as we have shown in Section~\ref{sec:gc_def}, the Lipschitz constant of the model is an upper bound on its GC. Hard Lipschitz normalisation methods, such as Spectral Normalisation, will thus impose a fixed upper bound on the GC of the model, while regularisation methods such as Spectral Regularisation induce a regularisation pressure to decrease that bound; for an overview of these methods as well as a contrast between soft and hard constraints, see Chapter~\ref{ch:smoothness}. 
We show in Figure~\ref{fig:spectral_explicit} that minimising the spectral norm of each layer, as done by Spectral Regularisation, leads to a decrease in GC. Furthermore, as the strength of Spectral Regularisation increases---by increasing the regularisation coefficient---GC decreases.

\noindent \textbf{Jacobian regularisation}. Jacobian regularisation is an \textit{existing} form of explicit regularisation which \textit{directly minimises GC}, i.e. it adds $\zeta GC(f;\vtheta, \mathcal{D}) = \frac{\zeta}{|\mathcal{D}|}\sum_{\vx\in \mathcal{D}} \|\jacparam{\vx}{f(\vx; \vtheta)}\|_2^2$ to the loss function. Jacobian regularisation has been correlated with increased generalisation but also robustness to input distributional shift \cite{hoffman2020robust,sokolic2017robust,varga2018gradient,yoshida2017spectral}. \citet{sokolic2017robust}  show that adding Jacobian regularisation to the loss function can lead to an increase in test set accuracy (their Tables III, IV, and V). We confirm their findings, namely an increase in test accuracy and a decrease in GC with Jacobian/GC regularisation in Figure~\ref{fig:explicit_gc_regularisation_cifar}. As we have noted previously, too much smoothness can hurt model capacity and decrease performance, which is something that we observe here too for large regularisation coefficients.
 \begin{figure}[t]
  \centering
  \begin{subfloat}[GC decreases with increased regularisation strength.]{
  \includegraphics[height=0.42\linewidth]{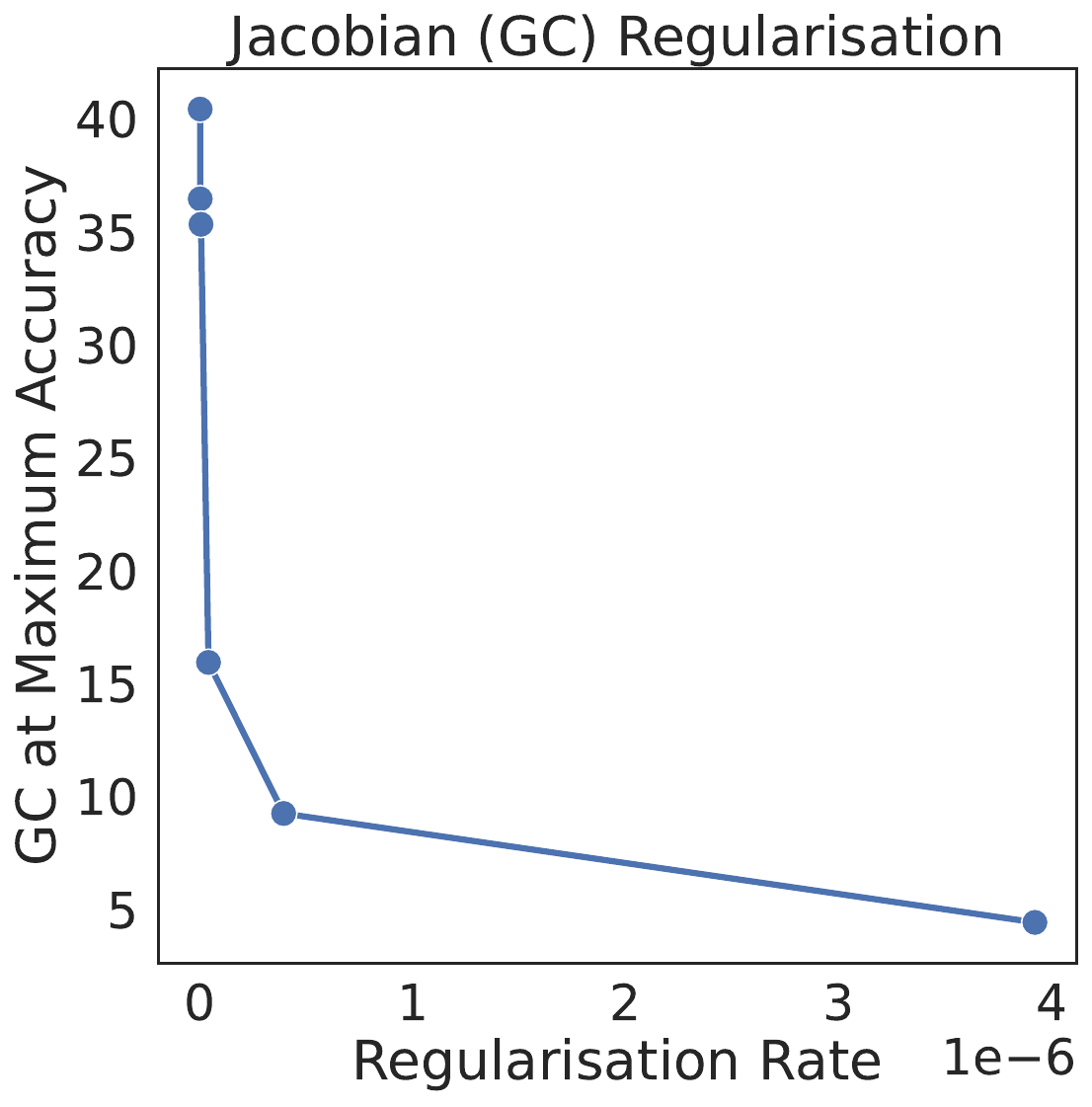}%
  }
  \end{subfloat}%
  \hspace{1.5em}
    \begin{subfloat}[Accuracy increases with GC regularisation strength, unless too much regularisation is used.]{
    \includegraphics[height=0.42\linewidth]{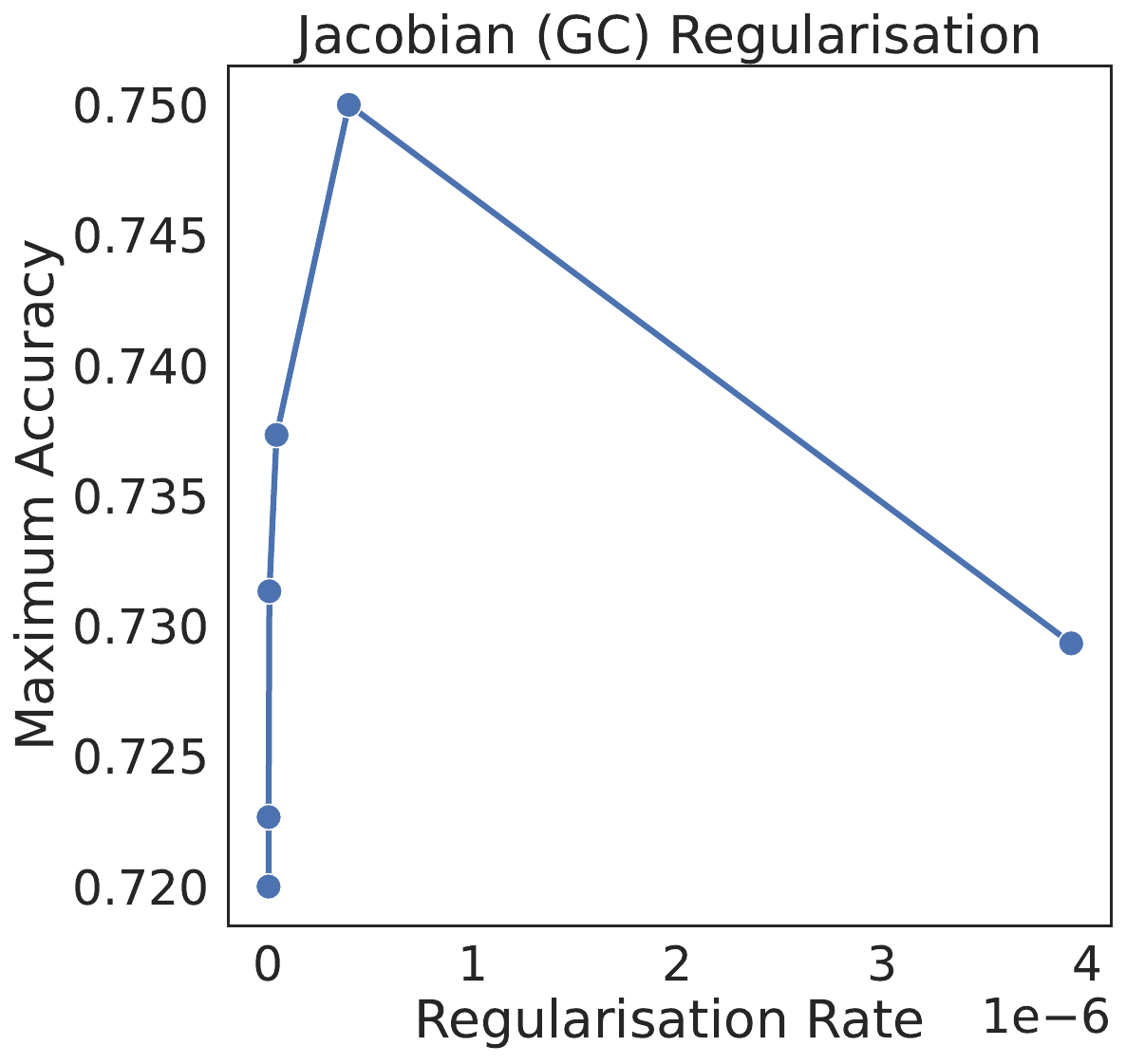}%
  }
  \end{subfloat}%
  \caption[Explicit minimising GC leads to increased generalisation.]{Jacobian (GC) regularisation on CIFAR-10. Explicit regularisation minimising GC (already existing in literature as Jacobian regularisation) leads to increased generalisation, unless the regularisation coefficient is too large. This effect reflects what we have observed in other settings, where too much smoothness can hurt performance, from an illustrative example in Figure~\ref{fig:smooth_ex}, to the Spectral Normalisation example in Figure~\ref{fig:two_moons_decision_surface_sn} and the reinforcement learning examples in Figure~\ref{fig:spectral_schedulers_rl}. Results obtained using 1 seed. }
  \label{fig:explicit_gc_regularisation_cifar}
\end{figure}

\subsection{Implicit regularisation via discretisation drift}

We now show how the implicit regularisation induced by discretisation drift in gradient descent optimisation, previously discussed in Chapters~\ref{ch:opt_intro},~\ref{ch:dd_gans} and~\ref{ch:bea_stochasticity}, can lead to a decrease in GC due the connection between gradient norm minimisation and GC in neural networks. Since the strength of implicit regularisation is affected by optimisation hyperparameters such as learning rate and batch size, we empirically observe how changing these hyperparameters leads to direct effect on GC.

\textbf{Learning rate}. Large learning rates have been consistently shown to have an implicit regularisation effect and lead to increased generalisation~\citep{li2019towards,igr,igr_sgd,hoffer2017train}. Attempts at formalising this include the Implicit Gradient Regularisation (IGR) flow discussed in Chapter~\ref{ch:opt_intro}, which suggests that gradient descent implicitly minimises the loss function $\tilde{E}(\vtheta) = E(\vtheta) + \frac h 4 \norm{\nabla_{\vtheta} E}^2$, where $h$ is the learning rate. In the previous section, we described how backpropagation reveals shared components between gradients w.r.t. parameters---$\nabla_{\vtheta} E$ present in IGR---and  gradients w.r.t. inputs---$\nabla_{\vx}f$ present in GC. We then showed that explicitly regularising the gradient norm $\norm{\nabla_{\vtheta} E}^2$ leads to a decrease in GC. Based on these results, we expect that increasing the learning rate $h$ leads to a decrease in GC due to increased implicit gradient regularisation pressure $\frac h 4 \norm{\nabla_{\vtheta} E}^2$, which we  empirically observe in Figure~\ref{fig:gc_learning_rate}.

\begin{figure}[t]
  \begin{subfloat}[Train loss.]{
    \includegraphics[width=0.33\linewidth]{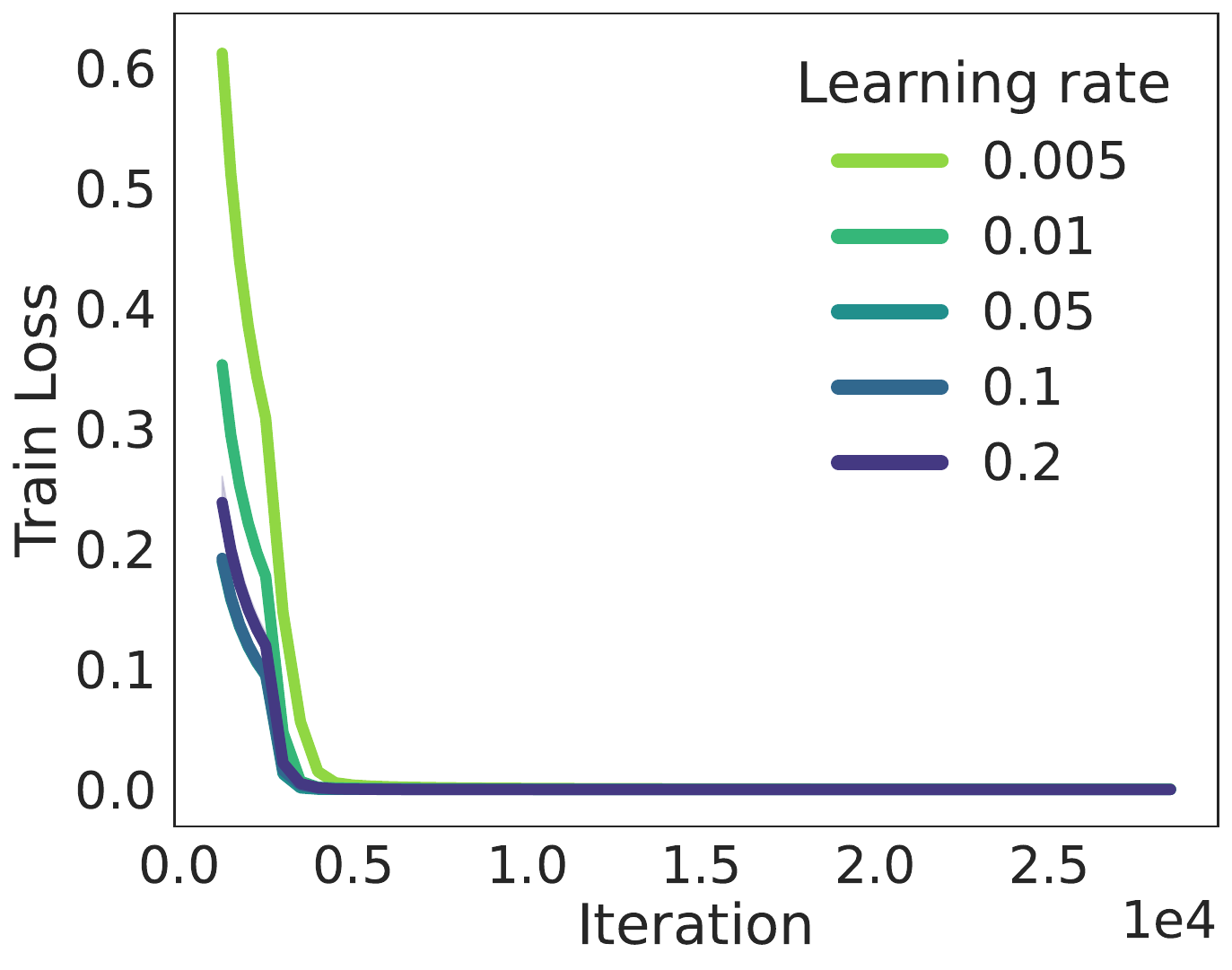}%
  }
  \end{subfloat}%
    \begin{subfloat}[GC.]{
    \includegraphics[width=0.33\linewidth]{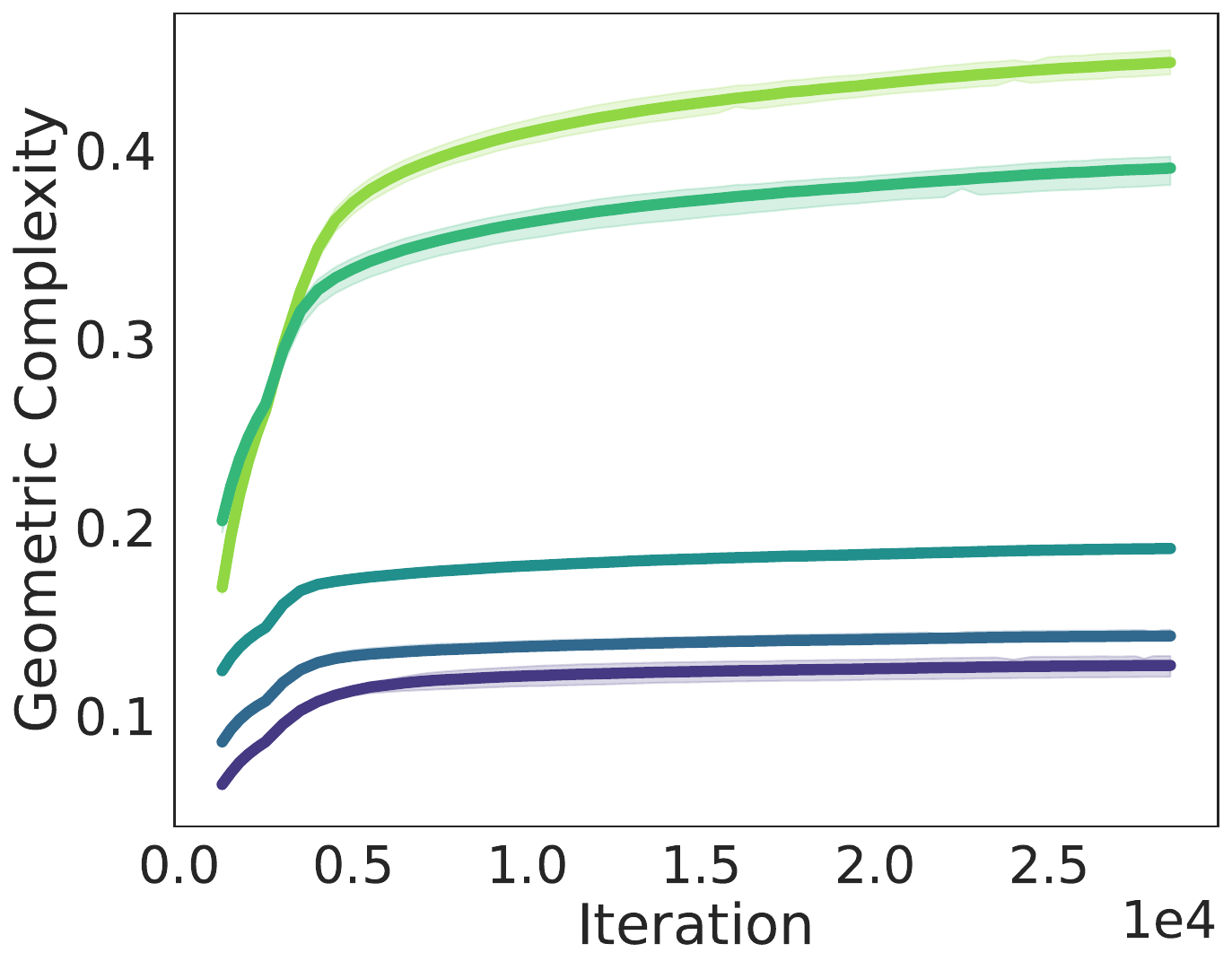}%
  } 
  \end{subfloat}%
   \begin{subfloat}[Test accuracy.]{
    \includegraphics[width=0.33\linewidth]{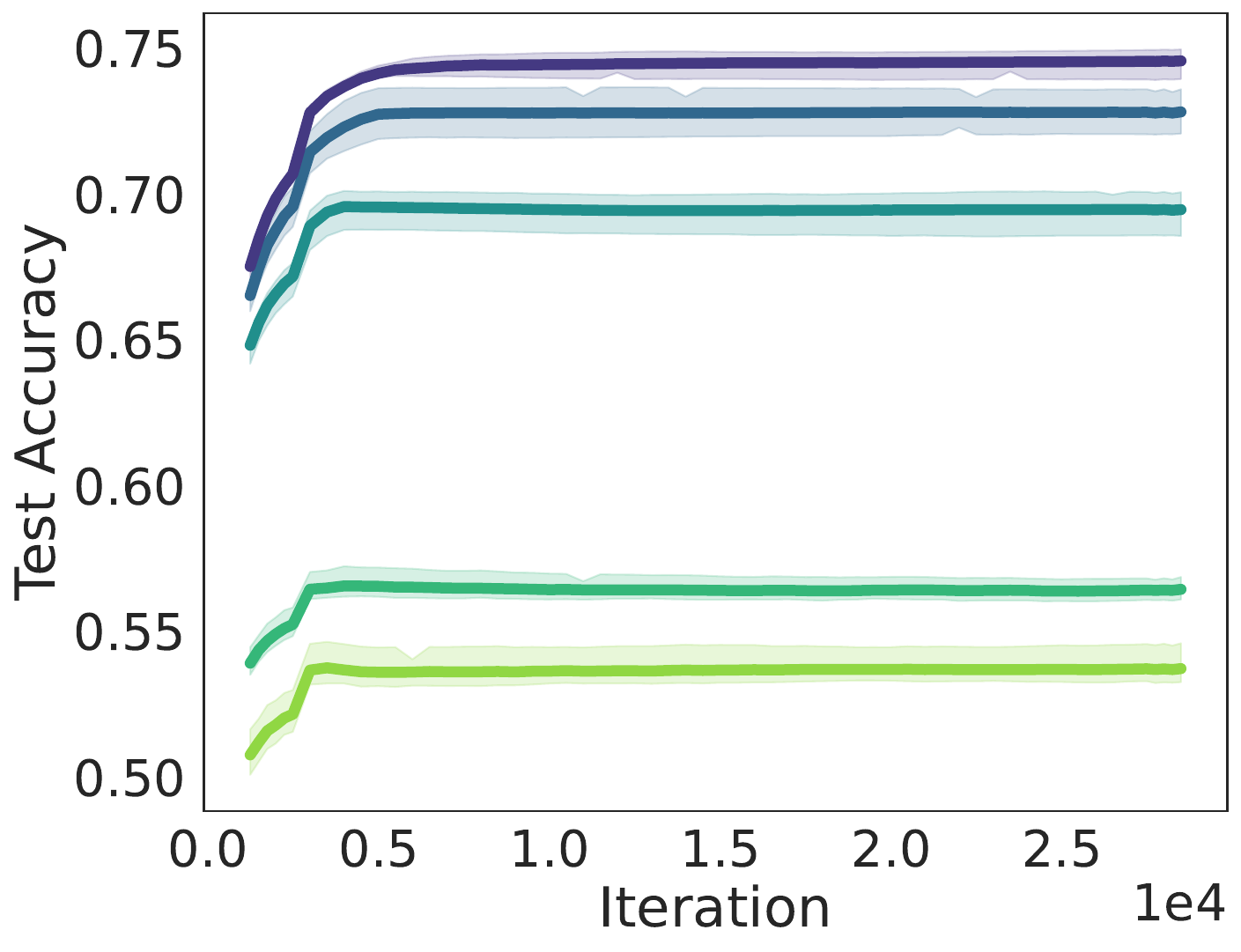}%
  }
  \end{subfloat}%
  \caption[The implicit regularisation effect of large learning rates on GC.]{The implicit regularisation effect of large learning rates on GC: increasing the learning rate results in decreased GC and increased performance. The results here use vanilla gradient descent; a similar trend is observed when momentum is used,  as shown in Section~\ref{sec:gc_exp_app} of the Appendix.}
  \label{fig:gc_learning_rate}
\end{figure}

\noindent \textbf{Batch size.}
The increased generalisation performance of lower batch sizes has been long studied in deep learning, often under the name `the generalisation gap'~\cite{lecun2012efficient,hoffer2017train,keskar2016large,smith2020generalization,masters2018revisiting}. While traditionally this increased generalisation effect has been attributed to gradient noise due to the sampling process, recent work has revisited this assumption:
the generalisation gap was recently bridged with~\citet{geiping2021stochastic} showing that full-batch training can achieve the same test set performance as stochastic gradient descent. \citet{geiping2021stochastic} use an explicit regulariser inspired by IGR, with the intuition that the strength of the implicit gradient norm regulariser induced by discretisation drift $\norm{\nabla_{\vtheta} E}^2$ is stronger for stochastic gradient descent compared to full-batch gradient descent, as described by~\citet{igr_sgd}; we previously discussed the implicit regularisation induced by discretisation drift in stochastic gradient descent and the results of~\citet{igr_sgd} in Chapter~\ref{ch:bea_stochasticity}. Thus, \citet{geiping2021stochastic} use $\frac{h}{4 |\mathcal{B}|}\sum_{B \in \mathcal{B}} \norm{\frac{1}{B}\sum_{\vX \in B} \nabla_{\vtheta} E(\vX;\vtheta)}^2$ instead, where $\mathcal{B}$ is the set of batches used for training, in contrast to $\frac{h}{4} \norm{\frac{1}{|\mathcal{D}|}\sum_{\vX \in \mathcal{D}} \nabla_{\vtheta} E(\vX;\vtheta)}^2$, the IGR regulariser in the full-batch case. Since the gradient norm is computed over a batch  rather than the entire dataset, outliers, such as examples with a large gradient, are less likely to be averaged out and this leads to a stronger regularisation pressure. 
Their result further strengthens the connection between small batch sizes and increased gradient norm regularisation pressure, which we have consistently seen to lead to smaller GC. \
Coupling these results, suggesting that for smaller batch sizes there is a stronger effect due to IGR, with what we have previously observed, namely that minimising $\norm{\nabla_{\vtheta} E}^2$ leads to a decrease in GC, we should expect that decreasing batch sizes leads to a decrease in GC. We empirically assess this question and show results in Figure~\ref{fig:gc_batch_sizes}, observing a consistent decrease in GC with smaller batch sizes. These results show the implicit regularisation effect of batch sizes on model smoothness.
\begin{figure}[t]
  \begin{subfloat}[Train loss.]{
    \includegraphics[width=0.33\linewidth]{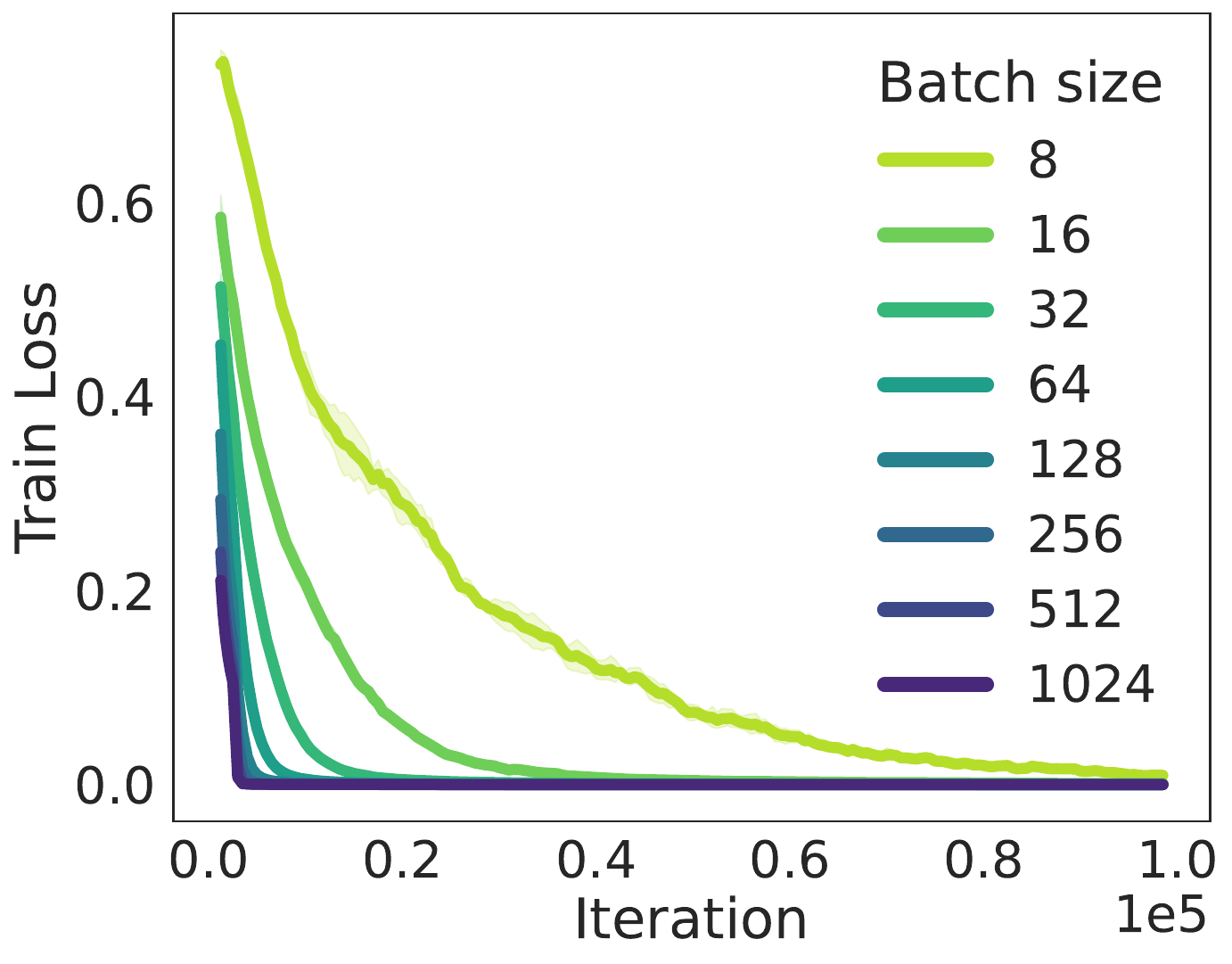}%
  }
  \end{subfloat}%
  \begin{subfloat}[GC.]{
    \includegraphics[width=0.33\linewidth]{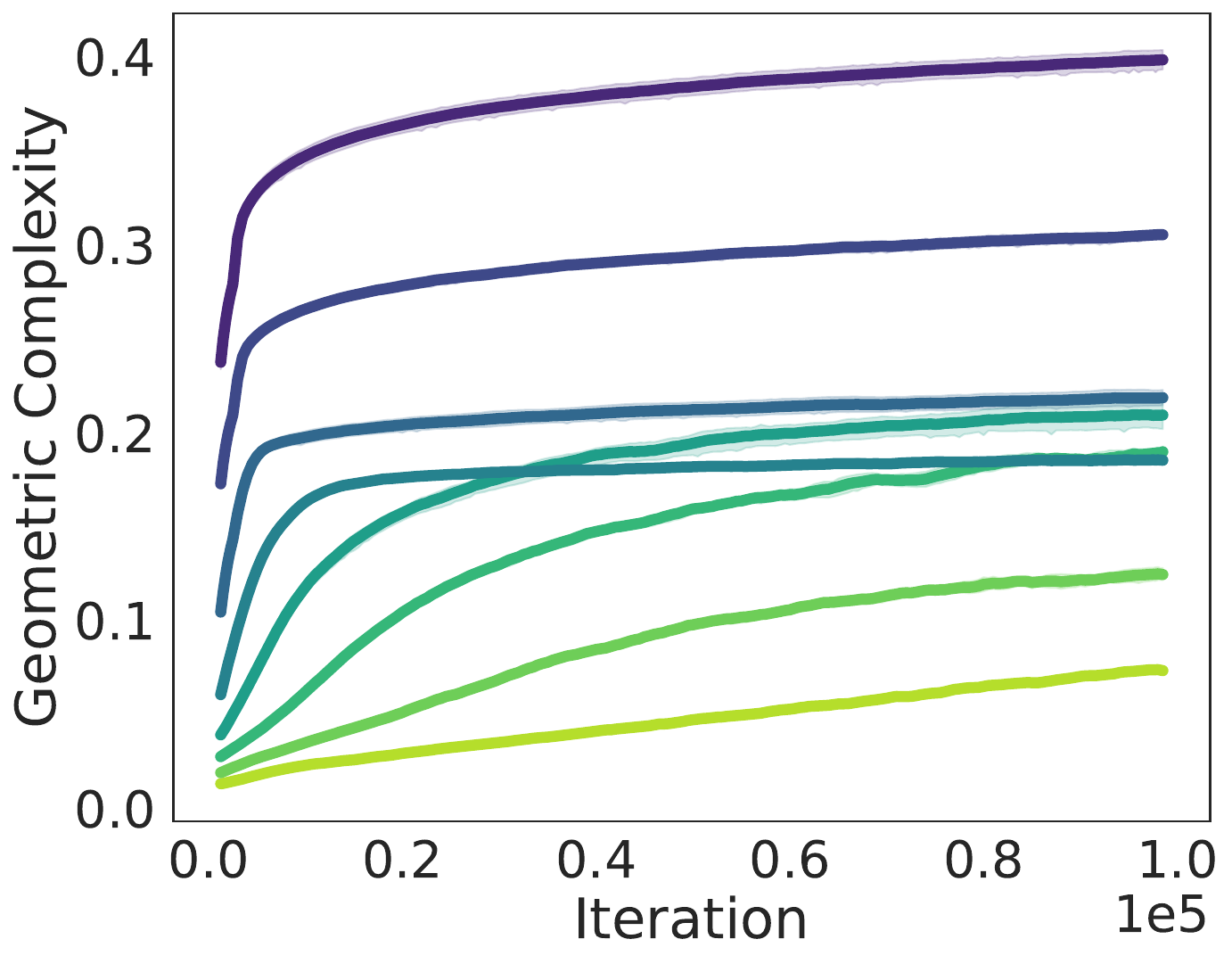}%
  }
  \end{subfloat}%
   \begin{subfloat}[Test accuracy.]{
    \includegraphics[width=0.33\linewidth]{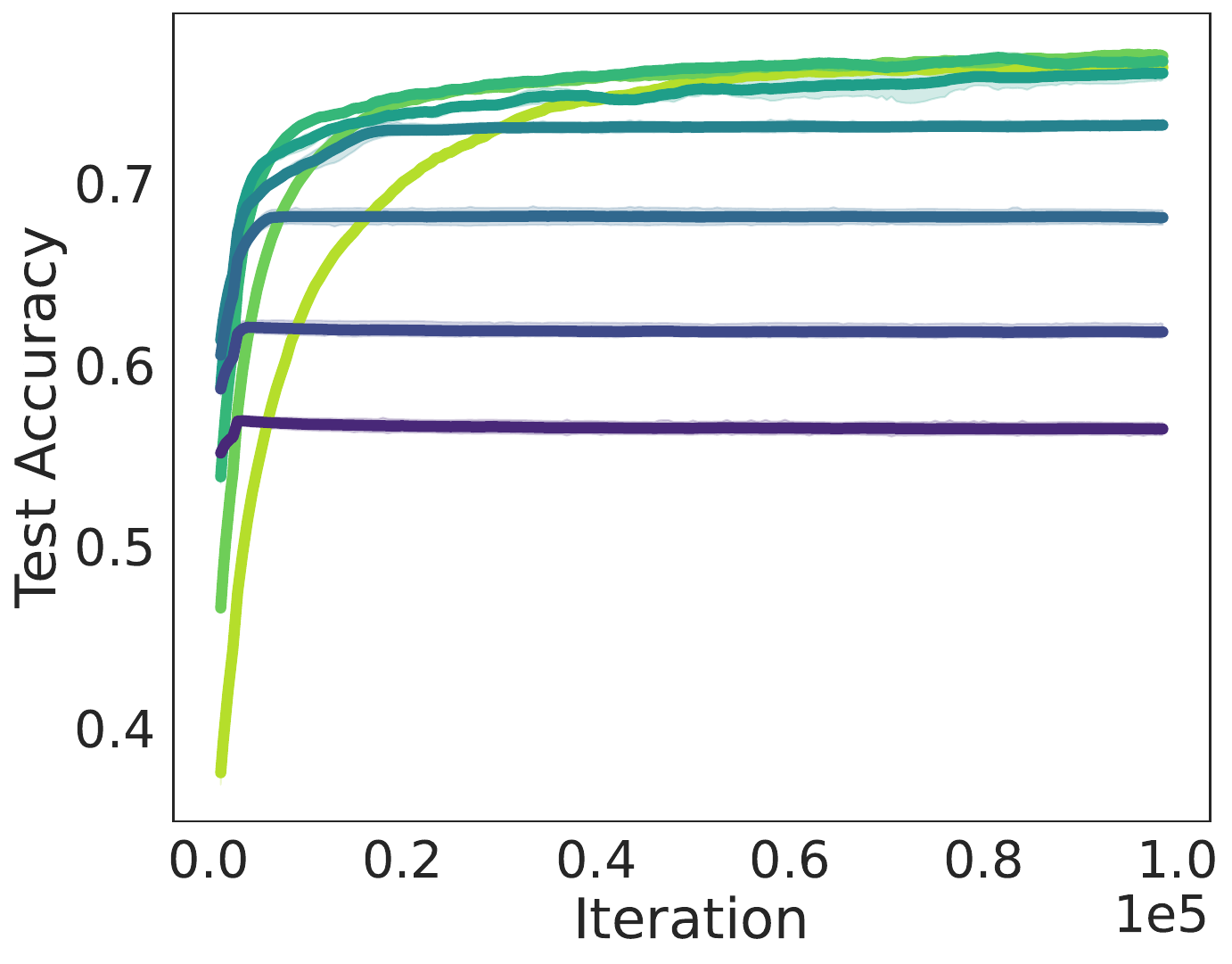}%
  }
  \end{subfloat}%
  \caption[The implicit regularisation effect of small batch sizes on GC.]{The implicit regularisation effect of small batch sizes on GC: decreasing the batch size results in decreased GC and increased performance. The results here use vanilla gradient descent; a similar trend is observed when momentum is used,  as shown in Section~\ref{sec:gc_exp_app} of the Appendix.}
  \label{fig:gc_batch_sizes}
\end{figure}

Results and justifications in this section use vanilla gradient descent, but we note that the gradient norm minimisation effect induced by the discretisation drift of momentum has recently been mathematically formalised by \citet{ghosh2023implicit}, and thus we expect our arguments to hold for gradient descent with momentum too. We confirm this with empirical results in Appendix~\ref{ch:gc_appendix}.

\section{Double descent and GC}

 When neural network complexity is measured using a simple parameter count, a double descent phenomenon has been consistently observed: as the number of parameters increases, the test loss decreases at first, before increasing again, followed by a second descent toward a low test error value~\cite{belkin2021fear,Belkin15849,deep_double_descent}.  An excellent overview of the double descent  phenomena in deep learning can be found in~\citet{belkin2021fear}, together with connections to smoothness as an inductive bias of deep networks and the role of optimisation. 
We previously discussed the double descent phenomenon in Section~\ref{sec:benefits}, where we contrasted it with the U-shaped curve associated with standard machine learning methods and complexity measures in Figure~\ref{fig:smoothness_double_descent}. We discussed the double descent phenomenon from the smoothness perspective and intuited that smoothness likely plays an important role in the second descent: we postulated that in the first descent, the generalisation error is decreasing as the model is given extra capacity to
capture the decision surface; the increase happens when the model has enough capacity to fit the training data, but it cannot do so and retain smoothness, and thus overfits; the second descent occurs as the capacity increases and smoothness can be retained.

 To explore whether GC as a measure of model complexity based on input smoothness can capture the double descent phenomena we follow the set up introduced in~\citet{deep_double_descent}: we train multiple ResNet-18 networks on CIFAR-10 with increasing layer width, and show results in Figure~\ref{fig:double_descent}. We make two observations: first, like the test loss, GC follows a double descent curve as width increases (Figure~\ref{fig:gc_train_test_dd}); second, when plotting GC against the test loss, we observe a U-shape curve, recovering the traditional expected behaviour of complexity measures (Figure~\ref{fig:gc_vs_train}). Importantly, we observe the connection between the generalisation properties of over-parametrised models and GC: after the critical region, increase in model size leads to a decrease in GC. These results provide further evidence that GC is able to capture model capacity in a meaningful way, suggestive of a reconciliation of traditional complexity theory and deep learning, as well as provide an empirical basis for our previously highlighted connection between smoothness and double descent.

 \begin{figure}[tb!]
\centering
\begin{subfloat}[GC and train/test losses as the network width increases.]{
\includegraphics[width=0.45\columnwidth]{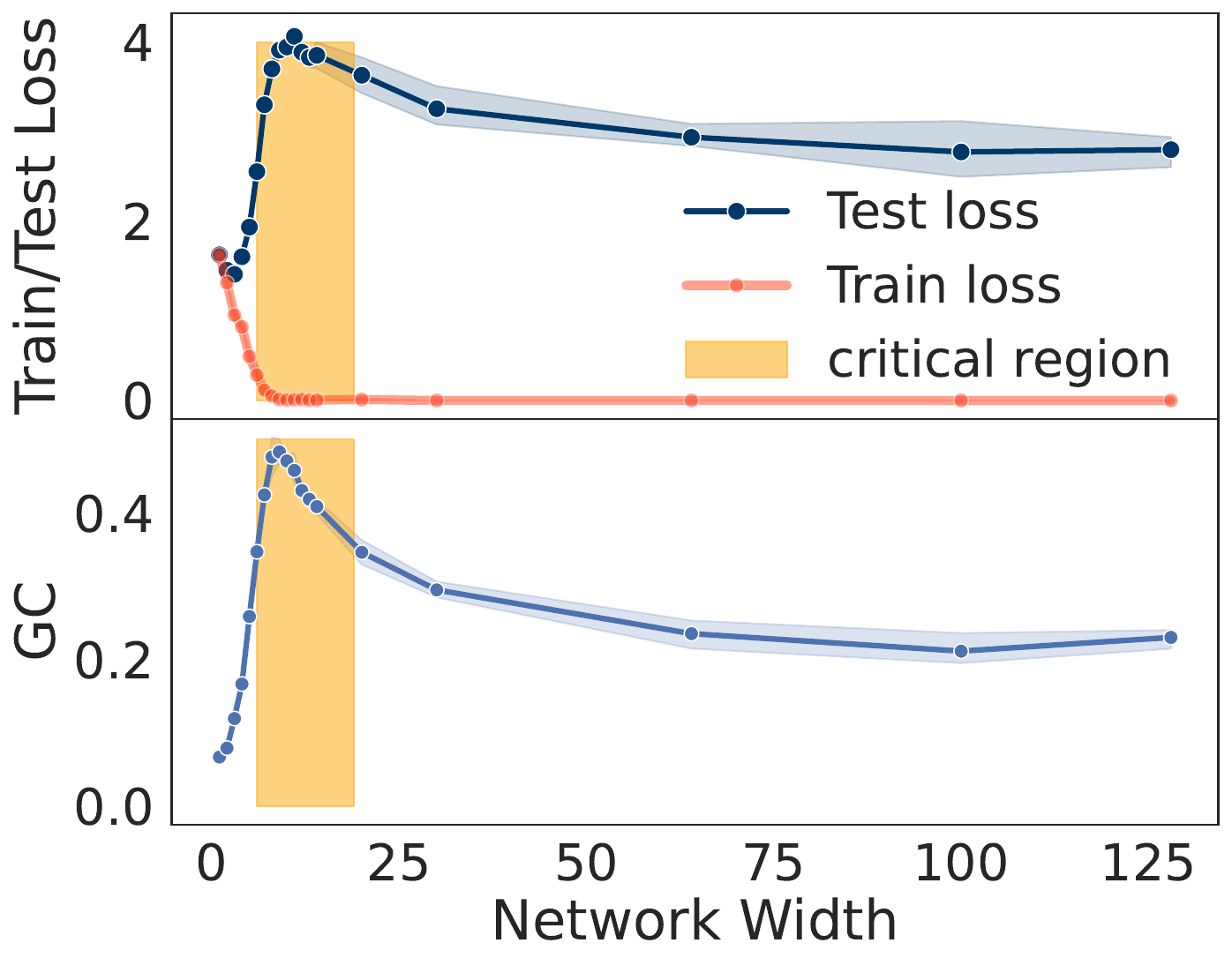}
\label{fig:gc_train_test_dd}
}
\end{subfloat}
\begin{subfloat}[Train and test losses as GC increases.]{
\includegraphics[width=0.43\columnwidth]{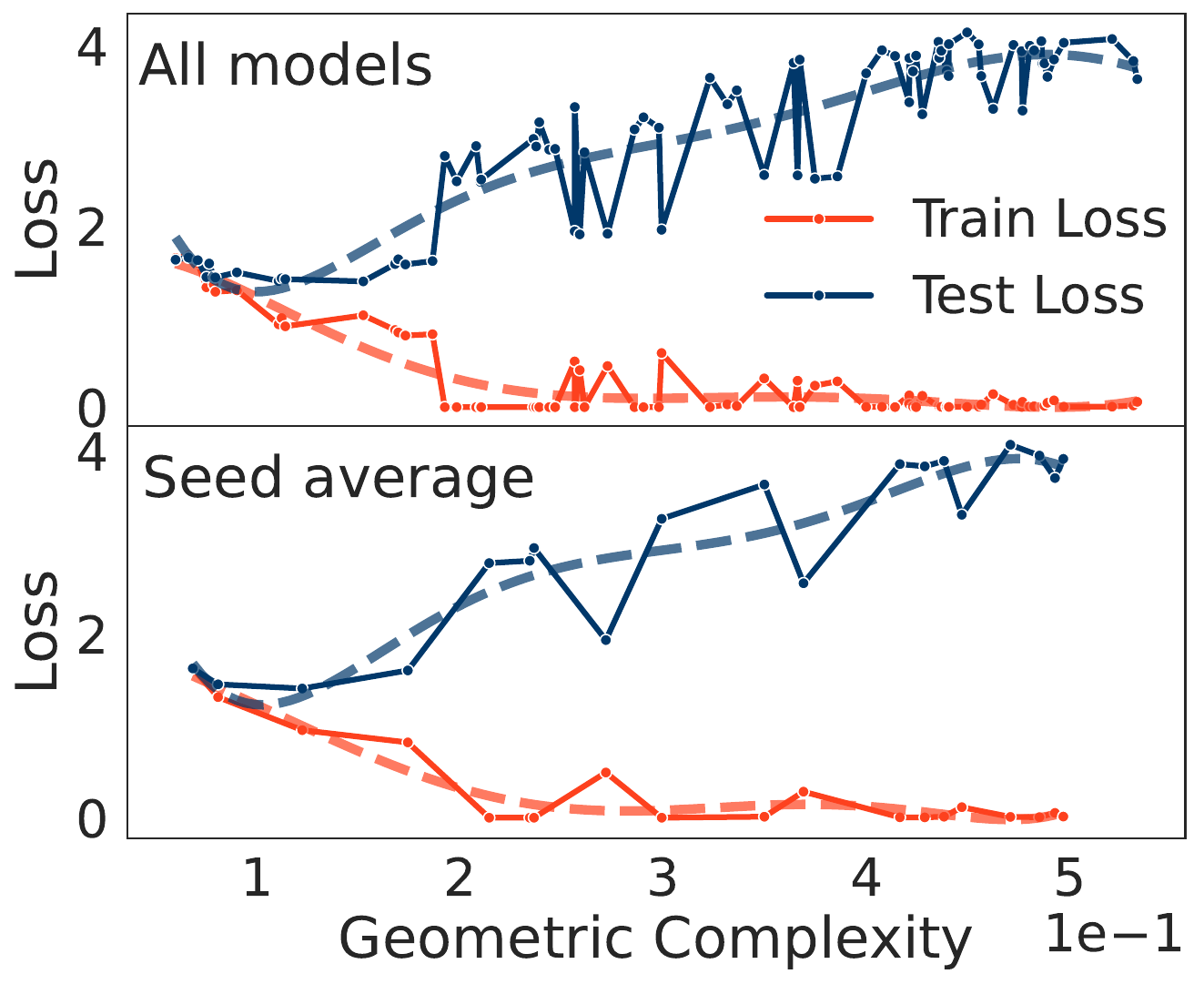}
\label{fig:gc_vs_train}
}
\end{subfloat}
\caption[Double descent and GC.]{Double descent and GC. \subref{fig:gc_train_test_dd}: GC captures the double descent phenomenon. \subref{fig:gc_vs_train}: GC captures the traditional U-shape curve, albeit with some noise, showing (top) GC compared to train and test losses for all models and (bottom)  GC compared to train and test losses averaged across different seeds. We fit a 6 degree polynomial to the curves to showcase the trend.}
\label{fig:double_descent}
\end{figure}

\section{Related work}

\noindent \textbf{Complexity measures for deep neural networks}. There has been a recent effort to create complexity measures that capture both the generalisation phenomenon of deep learning and are computationally tractable.~\citet{lee2020neural} introduce `Neural Complexity Measures' in which they use a neural network to approximate the excess risk---the difference between empirical and true risk, which the author call `the generalisation gap'---from validation data and use it as a regulariser for the model;
~\citet{nagarajan2019generalization} argue for an initialisation dependent measure of complexity;~\citet{jiang2018predicting} approximate classification decision margins for neural networks and show their predictive generalisation gap correlates to the true generalisation gap.~\citet{maddox2020rethinking} use the effective dimensionality~\citep{mackay1991bayesian} of the Hessian matrix $\nabla_{\vtheta}^2 E$ and show that it can capture the double descent phenomenon; since effective dimensionality depends on the eigenspectrum of the Hessian which is computationally expensive to compute for large networks, the authors select the leading eigenvalues to estimate the metric. 

Most recently,~\citet{grant2022predicting} modernised the Generalised Degrees of Freedom 
(GDof)~\citep{efron1983estimating,ye1998measuring} complexity measure by providing a tractable approach to estimate it online during training. GDof is a similar measure to GC in that it computes the model sensitivity with respect to the dataset labels (GC instead computes the sensitivity to inputs).~\citet{grant2022predicting} showed that GDof can resolve the double descent phenomenon; we hope that future work will assess the connection between GC and GDof both empirically and theoretically.

\citet{huh2021low} argue that deep neural networks exhibit a low-rank bias, measured as the rank of the Gram matrix of feature embeddings. 
By measuring a relationship between data points, a low-rank measure is less local than GC, which assess how sensitive the model is to individual input changes. 
\citet{huh2021low} connect low-rank bias with depth and initialisation, but contend that depth plays a more significant role than initialisation and optimisation. Instead, we have associated the implicit bias induced by GC with optimisation and optimisation hyperparameters. 

\noindent \textbf{Jacobian regularisation}. Using GC as an explicit regulariser has existed previously as Jacobian regularisation, used to induce model robustness with respect to inputs and increase generalisation~\citep{hoffman2020robust,sokolic2017robust,varga2018gradient,yoshida2017spectral}. Our goal here was not to introduce a new regulariser, but to show that many existing implicit or explicit regularisers minimise GC, and that GC can be used as a neural network complexity measure.

\noindent \textbf{Gradient penalties}. Gradient penalties are used to regularise $f: \mathbb{R}^I \rightarrow \mathbb{R}$ using a penalty of the form  $\sum_{\vx} (||\nabla_{\vx}{f(\vx; \vtheta)}||- K)^2$ with $K$ a hyperparameter; instead of minimising the Jacobian norm, these methods ensure it is close $K$. For additional discussion on gradient penalties, see Section~\ref{sec:smoothness_techniques}. Gradient penalties have been primarily used for GAN training~\citep{fedus2017many,gulrajani2017improved,kodali2017convergence}; 
we leave the investigation of the importance of GC outside supervised learning for future work.

\noindent \textbf{Smoothness and generalisation}. 
\citet{novak2018sensitivity} is perhaps the closest work to ours: they show the correlation across multiple datasets between generalisation and 
$\frac{1}{\mathcal{D}_{test}} \sum_{\vx_i \in \mathcal{D}_{test}} \norm{\evaljac{\vx}{(\tau \circ f(\cdot; \vtheta))}{\vx_i}}_2$  where $\tau$ is the last layer activation function (such as the softmax). They also 
briefly show that their Jacobian norm metric captures the regularisation effect of stochastic gradient descent compared to full-batch training using L-BFGS~\citep{liu1989limited}. Since our goal is to find a complexity measure we do not use the test set but the training set instead; our main focus is to show the connections between GC implicit and explicit regularisation, as well as initialisation in deep learning.

\citet{bartlett2017spectrally} provide a generalisation bound depending on classification margins and the product of spectral norms of the model's weights and show empirically that the product of spectral norms correlates with excess risk. \citet{bubeck2021universal} connect Lipschitz smoothness with over-parametrisation, by providing a probabilistic upper bound on the model's Lipschitz constant based on its training performance and the inverse of its number of parameters.

\noindent \textbf{Smoothness and double descent}. 
In concurrent work, \citet{gamba2022deep} discuss the connection between the Jacobian and the Hessian of the \textit{loss} with respect to inputs, namely the empirical average of $||\nabla_{\vx} E||_2$ and $||\nabla_{\vx}^2 E||_2$ over the training set and show that they follow a double descent curve. Like our work with GC, \citet{gamba2022deep} mention how their work suggests our postulate in Chapter 3: `\textit{Finally, Rosca et al. (2020) investigate how to encourage smoothness, either on the entire input space or around training
data points. Interestingly, they postulate a connection between model-wise double descent and smoothness: during
the first ascent the model size is large enough to fit the training data at the expense of smoothness, while the second
descent happens as the model size becomes large enough for smoothness to increase. [...] Our work provides empirical
evidence supporting the postulate of Rosca et al. (2020).}'

\section{Limitations and future work}

Our experiments use the Resnet-18 and MLP architectures, and the ReLu activation function. We hope that future work will further validate these results in a variety of settings, including for other architectures, activation functions, and input domains.

Our results suggest that GC is able to capture a model's complexity and its generalisation abilities on a test set obtained using i.i.d. sampling from the same underlying distribution as the training set. We hope future work will investigate the properties of GC when the test time distribution differs from the training distribution, as is the case in transfer learning~\citep{zhuang2020comprehensive,iman2022review}, or when the training set gradually changes in time, as in continual learning~\citep{parisi2019continual}. We intuit that the results will depend on the distance between the training and test distribution in the case of transfer learning, or the rate of change of the input in the case of continual learning.

 We believe incorporating GC for model selection has the potential to construct criteria analogous to the Bayesian information criterion~\citep{bishop2006pattern}, adapted to the deep learning domain.
For architecture search GC could prove to be useful by excluding architectures that have high GC at initialisation on the training data. While we did not investigate this here, we hope this is a fruitful avenue for future work.

\section{Conclusion}

We introduced Geometric Complexity (GC), a measure of model complexity defined as the average Jacobian norm of a network's logits with respect to inputs from the training dataset. We showed that GC is closely connected to Lipschitz smoothness, but unlike the Lipschitz constant of a deep network, GC is tractable and can be measured during training. We have empirically shown how common explicit regularisers that target other quantities, such as weight norm regularisation and gradient norm regularisation, implicitly regularise GC.  We then connected GC with the implicit regularisation induced by the discretisation drift of gradient descent. We saw how mechanisms that induce higher implicit regularisation, such as decreased batch size and increased learning rate, lead to a decrease in GC. Last but not least, we showed how GC can resolve the double descent challenge associated with traditional complexity measures and recovers the expected U-shaped curve.

\chapter{Conclusion}

Throughout the thesis we have seen that despite being fundamental ingredients in the melting pot behind deep learning's recipe for success, the effects of gradient-based optimisation and smoothness regularisation are far from being understood. Our aim was to take a few steps towards a practical theory of deep learning, inspired by great recent works making strides in that direction. 
To this end, we started with investigating gradient descent, a perhaps deceptively simple algorithm, which forms the basis of all modern deep learning optimisation. 
We took a continuous-time perspective to understanding gradient descent, either by quantifying discretisation drift between existing continuous-time flows and gradient descent or by finding new flows that better describe the gradient descent trajectory.
We then constructed novel continuous-time analyses to determine sources of instability in  supervised learning and two-player games such as Generative Adversarial Networks, which led to mitigation strategies either in the form of adaptive learning rates or explicit regularisers. 
We continued our investigation by studying interactions between optimisation and models, in particular model smoothness with respect to inputs and related regularisers.
We expanded the use of smoothness regularisation to a new domain, reinforcement learning, where we observed that Spectral Normalisation leads to a consistent gain in agent performance that can be recovered through changes to optimisation.  When revisiting the use of Spectral Normalisation in Generative Adversarial Networks, however, we encountered a mixture of model regularisation and optimisation effects. We concluded by introducing a measure of model complexity based on smoothness and explored its interactions with other ingredients of the deep learning melting pot, such as implicit regularisation effects induced by discretisation drift,  initialisation, and explicit regularisation.

Motivated by our desire to dig deeper into why deep learning works so well, we employed theoretical and empirical tools to  perform an investigation into individual deep learning components and their many interactions. Using our novel intuitions and understanding of deep learning, we constructed methods that led to increased training stability or improved evaluation performance across multiple problem domains.
We hope both the approach and the results presented here can form the fruitful basis of further lines of inquiry, for the author as well as the wider community.

\clearpage
\bibliography{references}

\clearpage
\appendix

\clearpage
\appendix
\appendixpage

\startcontents[chapters]
\printcontents[chapters]{l}{0}{\setcounter{tocdepth}{1}}

\chapter{On a new continuous-time model of gradient descent }
\section{Proofs}
\label{sec:all_proofs}

\subsection{BEA proof structure}
\label{sec:bea_proof_structure}

\begin{figure}[h]
\begin{tikzpicture}[every text node part/.style={align=center,inner sep=0,outer sep=0}][overlay]
\coordinate (theta_t_minus_1) at (0,0);
\coordinate (theta_t) at (2.2,-1);

\node(draw) at ($(theta_t_minus_1) + (+0.3,-0.32)$) {$\vtheta_{t-1}$};
\node(draw) at ($(theta_t) + (0.5,-0.2)$) {$\vtheta_{t}$};
\node(draw) at ($(theta_t) + (4.2,-0.2)$) {$= {\color{purple} \vtheta_{t-1}}  {\color{blue}- h \nabla_{\vtheta} E(\vtheta_{t-1})}$};

\coordinate (first_time_transition) at ($(theta_t_minus_1) + (+0.55,1.6)$);

\coordinate (mod_cont_theta_t) at (2.2,0.8);
\node(draw) at ($(mod_cont_theta_t) + (3,-0.)$) {${\vtheta}(h) \hspace{0.1em}= \sum_{p=0}^{\infty} \frac{h^p}{p!} {\vtheta}^{(p)} $};
\node(draw) at ($(mod_cont_theta_t) + (6.75,-1.)$) {$= {\color{purple} \vtheta_{t-1}} {\color{blue}- h \nabla_{\vtheta} E(\vtheta_{t-1})} +  \sum_{i=2}^{n+1} h^i \underbrace{l_i(\vtheta_{t-1})}_{{\color{red}{\mathbf{0}}}} + {\color{red}\mathcal{O}(h^{n+2})} $};

\draw [thick,dashed] (theta_t_minus_1) -- (theta_t);

\draw [thick]  (theta_t_minus_1) to[out=50,in=180]  node[near end,above] {$\dot{\vtheta}$} (mod_cont_theta_t);

\draw [
    thick,
    decoration={
        brace,
        mirror,
        raise=0.1cm
    },
    decorate
] (theta_t) -- (mod_cont_theta_t)
node [pos=0.5,anchor=west,xshift=0.15cm,yshift=-0.0cm] { \scriptsize ${\color{red}\mathcal{O}(h^{n+2})}$};
\end{tikzpicture}
 \caption[Backward error analysis proof structure.]{BEA finds continuous modified flows that describe the gradient descent update with learning rate $h$ with an error of $\mathcal{O}(h^{n+2})$. We identify $f_1, \cdots f_n$ so that terms of order $\mathcal{O}(h^p), 2 \le  p \le n+1$ in ${\vtheta}(h)$ are $\mathbf{0}$.}
  \label{fig:bea_proof_structure}
\end{figure}

\textbf{General structure.}
The goal of BEA is to find the functions $f_1$, \dots $f_{n}$ such that the flow
\begin{align}
    \dot{\vtheta} =  - \nabla_{\vtheta} E + h f_1(\vtheta) + \cdots + h^n f_n(\vtheta)
\end{align}
has an error $\| \vtheta_t - {\vtheta}(h)\|$ of order $\mathcal{O}(h^{n+2})$ after 1 gradient descent step of learning rate $h$. To do so requires multiple steps (visualized in Figure~\ref{fig:bea_proof_structure}): 
\begin{enumerate}
    \item Expand ${\vtheta}(h)$ via a Taylor expansion in $h$: ${\vtheta}(h) = \sum_{p=0}^{\infty} \frac{h^p}{p!} {\vtheta}^{(p)}$;
    \item Expand each ${\vtheta}^{(p)}$ up to order $\mathcal{O}(h^{n+2-p})$ as a function of $f_1$, \dots $f_{n}$ via the chain rule;
    \item Group together terms of the same order in $h$ in the expansion, up to order $n+2$. 
    \begin{align}
    {\vtheta}(h) &= \sum_{i=0}^{n+1} h^i l_i(\vtheta_{t-1}) + \mathcal{O}(h^{n+2}) \\&= \vtheta_{t-1} - h \nabla_{\vtheta} E(\vtheta_{t-1}) + \sum_{i=2}^{n+1} h^i l_i(\vtheta_{t-1}) + \mathcal{O}(h^{n+2})
    \end{align}
    \item Compare the above update with the gradient descent update $\vtheta_t = \vtheta_{t-1} - h \nabla_{\vtheta} E(\vtheta_{t-1})$ and conclude that $l_i=\mathbf{0}$, $\forall i \in \{2, n + 1\}$. Use this to identify $f_1$, \dots $f_n$.
\end{enumerate}

\textbf{Notation and context}: all proofs below apply to general Euler updates not only gradient descent. We thus assume an update function $f$ with the Euler step $\vtheta_t = \vtheta_{t-1} + h f(\vtheta_{t-1})$. We can then use BEA to find the higher order correction terms describing the Euler discrete update up to a certain order, and replace $f = - \nabla_{\vtheta} E$ to obtain the corresponding results for gradient descent. When we perform a Taylor expansion in $h$ we often drop in notation the evaluation at $h=0$ and we make that implicit.

\subsection{Third-order flow}
\label{sec:third_order_flow_proof}

\begin{theorem} The modified flow 
\begin{align}
 \dot{\vtheta} = f - h \frac{1}{2} \jacparam{\vtheta}{f} f + h^2 \left(\frac{1}{3}  \left(\jacthetaf\right)^2 f  + \frac {1} {12} f^T \secondorderjac{\vtheta}{f} f\right)
 \label{eq:third_order_euler_app}
\end{align}
with $\vtheta(0) = \vtheta_{t-1}$ follows an Euler update $\vtheta_{t} = \vtheta_{t-1} + hf(\vtheta_{t-1})$ with an error of $\mathcal{O}(h^4)$ after 1 gradient descent step.
\end{theorem}
\begin{proof}

We are looking for functions $f_1$ and $f_2$ such that the modified flow
\begin{align}
 \dot{\vtheta} = f + h f_1 + h^2 f_2
 \label{eq:bea_structure_third_order}
\end{align}
follows the steps of GD with an error up to $\mathcal{O}(h^4)$.
We perform a Taylor expansion of step size $h$ of the above modified flow to be able to see the displacement in that time up to order $\mathcal{O}(h^4)$. 
We obtain (all function evaluations of $f$  and $f_i$ are at $\vtheta_{t-1}$ which we omit for simplicity, and annotate proof steps, CR denotes Chain Rule): 
\begin{align} 
   \vtheta&(h) = \vtheta(0) + h \dot{\vtheta}(0) + \frac{1}{2} h^2 \dot{\dot{\vtheta}}(0) + \frac{1}{6} h^3 \dot{\dot{\dot{\vtheta}}}(0) + \mathcal{O}(h^4)\\
                     &= \vtheta_{t-1} + h \left(f + h f_1 + h^2 f_2\right) + \frac{1}{2} h^2 \frac{d}{dt}(f + h f_1 + h^2 f_2)  \\&\hspace{1em}+ \frac{1}{6} h^3 \dot{\dot{\dot{\vtheta}}} + \mathcal{O}(h^4)  \owntag{using Eq \eqref{eq:bea_structure_third_order}} \\
                     &= \vtheta_{t-1} + h f + h^2 f_1 + h^3 f_2 \\&\hspace{1em}+ \frac{1}{2} h^2 \raw{\vtheta}{f + h f_1 + h^2 f_2}(f + h f_1 + h^2 f_2)  \\&\hspace{1em}+ \frac{1}{6} h^3 \dot{\dot{\dot{\vtheta}}} + \mathcal{O}(h^4) \owntag{CR} \\
                    &= \vtheta_{t-1} + h f + h^2 f_1 + h^3 f_2 + \frac{1}{2} h^2 (\jacthetaf f + h  \jacparam{\vtheta}{f }f_1 + h  \jacparam{\vtheta}{f_1} f)  \\&\hspace{1em}+ \frac{1}{6} h^3 \dot{\dot{\dot{\vtheta}}} + \mathcal{O}(h^4) \\
                     &= \vtheta_{t-1} + h f + h^2 f_1 + h^3 f_2 + \frac{1}{2} h^2 (\jacthetaf f + h  \jacparam{\vtheta}{f }f_1 + h  \jacparam{\vtheta}{f_1} f) \\&\hspace{1em}+ \frac{h^3}{6} \frac{d}{dt} \left[\jacthetaf f + h  \jacparam{\vtheta}{f }f_1 + h  \jacparam{\vtheta}{f_1} f\right] + \mathcal{O}(h^4) \\
                    &= \vtheta_{t-1} + h f + h^2 f_1 + h^3 f_2 + \frac{1}{2} h^2 (\jacthetaf f + h  \jacparam{\vtheta}{f }f_1 + h  \jacparam{\vtheta}{f_1} f) \\&\hspace{1em}+ \frac{1}{6} h^3 \frac{d}{dt} \left[\jacthetaf f\right] + \mathcal{O}(h^4) \\
                    &= \vtheta_{t-1} + h f + h^2 f_1 + h^3 f_2 + \frac{1}{2} h^2 (\jacthetaf f + h  \jacparam{\vtheta}{f }f_1 + h  \jacparam{\vtheta}{f_1} f) \\&\hspace{1em}+ \frac{1}{6} h^3  \raw{\vtheta}{\jacthetaf f}(f + h f_1 + h^2 f_2) + \mathcal{O}(h^4)  \owntag{CR} \\
                    &= \vtheta_{t-1} + h f + h^2 f_1 + h^3 f_2 + \frac{1}{2} h^2 (\jacthetaf f + h  \jacparam{\vtheta}{f }f_1 + h  \jacparam{\vtheta}{f_1} f) \\&\hspace{1em}+ \frac{1}{6} h^3  \raw{\vtheta}{\jacthetaf f} f + \mathcal{O}(h^4)
\label{eq:modified_flow_exp}
\end{align}

If we match the terms with the Euler update, we obtain that:
\begin{align}
f_1 = -\frac{1}{2} \jacparam{\vtheta}{f} f
\label{eq:f_1_third_order}
\end{align}
and
\begin{align}
f_2 &= - \frac{1}{2}  \left[\jacthetaf  f_1  + \jacparam{\vtheta}{f_1} f \right] - \frac{1}{6} \raw{\vtheta}{\jacthetaf f}f \\
f_2 &= - \frac{1}{2}  \left[\jacthetaf (-\frac{1}{2} \jacparam{\vtheta}{f} f)  + \raw{\vtheta} {-\frac{1}{2} \jacparam{\vtheta}{f} f} f\right] - \frac{1}{6} \raw{\vtheta}{\jacthetaf f} f \\
f_2 &=  \frac{1}{4}  {(\jacthetaf)}^2 f + \frac 1 4 \raw{\vtheta}{\jacthetaf f} f - \frac{1}{6} \raw{\vtheta}{\jacthetaf f} f \\
f_2 &=  \frac{1}{4}  {(\jacthetaf)}^2 f + \frac {1} {12}  \raw{\vtheta}{\jacthetaf f} f \\
f_2 &=  \frac{1}{4}  {(\jacthetaf)}^2  f + \frac {1} {12}  \left({(\jacthetaf)}^2 + f^T \secondorderjac{\vtheta}{f}  \right) f\\
f_2 &=  \frac{1}{3}  {(\jacthetaf)}^2  f + \frac {1} {12} f^T \secondorderjac{\vtheta}{f}  f
\label{eq:f_2_third_order}
\end{align}
where $f^T \secondorderjac{\vtheta}{f}$ is a matrix in $\mathbb{R}^{D\times D}$, with $\left(f^T \secondorderjac{\vtheta}{f}\right)_{i,j} = \left(\jacparam{\vtheta}{\jacparam{\vtheta_j}{f_i}}\right)^T f$ and $f^T \secondorderjac{\vtheta}{f}  f$ is a vector in $\mathbb{R}^{D}$ with $\left(f^T \secondorderjac{\vtheta}{f}  f\right)_{k} = \sum_{i, j} \frac{d f_k}{d \vtheta_i d\vtheta_j} f_i f_j$.

Replacing $f_1$ and $f_2$ in Eq~\eqref{eq:bea_structure_third_order} leads to the desired Eq~\eqref{eq:third_order_euler_app}.

\end{proof}

In the case of gradient descent, $f = -\nabla_{\vtheta} E$, and thus $\jacthetaf = \nabla_{\vtheta}^2E$ and we write $\secondorderjac{\vtheta}{f} = \nabla_{\vtheta}^3E \in \mathbb{R}^{D \times D \times D}$ and $(\nabla_{\vtheta}E^T (\nabla_{\vtheta}^3E) \nabla_{\vtheta}E)_{k} = \sum_{i,j}{(\nabla_{\vtheta}E)_i} (\nabla_{\vtheta}^3E)_{i,k,j} {(\nabla_{\vtheta}E)_j}$. Substituting in $f_1$ and $f_2$, we obtain (Eq~\eqref{eq:third_order_modified_vector_field}):
\begin{align}
\dot{\vtheta} = -\nabla_{\vtheta}E   -\frac{h}{2} \nabla_{\vtheta}^2E  \nabla_{\vtheta}E - h^2 \left( \frac{1}{3}  (\nabla_{\vtheta}^2E)^2 \nabla_{\vtheta} E + \frac {1} {12} \nabla_{\vtheta}E^T (\nabla_{\vtheta}^3E) \nabla_{\vtheta}E\right).
\end{align}

\subsection{Higher order terms of the form ${(\jacthetaf)}^n f$ or $\left(\nabla_{\vtheta}^2 E\right)^n \nabla_{\vtheta} E$}
\label{sec:higher_order_proofs}

\begin{theorem}
The modified flow with an error of order $\mathcal{O}(h^{p+1})$ to the Euler update  ${\vtheta_t = \vtheta_{t-1} + h f(\vtheta_{t-1})}$
has the form:
\begin{align}
    \dot{\vtheta} = \sum_{n=0}^{p} \frac{(-1)^{n}}{n+1} {(\jacthetaf)}^n f + \csecondorderfam
\end{align}
where $\csecondorderfam$ denotes a class of functions defined as a sum of individual terms each containing higher than first order derivatives applied to $f$.
\label{thm:general_bea}
\end{theorem}
\begin{proof}
Consider the modified flow given by BAE:
\begin{align}
\dot{\vtheta} = \tilde{f} = f + h f_1 + h^2 f_2 + \dots + h^n f_n + \dots
\label{eq:modified_bea_app}
\end{align}
 For the proof we need to show that the only terms not involving higher order derivatives are of the form ${(\jacthetaf)}^{n}f $  and we need to find the coefficients of ${(\jacthetaf)}^{n} f$ in $f_n$, which we will call $c_n$. We already know that $c_0 = 1$, $c_1 = -\frac{1}{2}$ (Eq.~\eqref{eq:f_1_third_order}) and $c_3 = \frac 1 3$ (Eq.~\eqref{eq:f_2_third_order}).
We use the general structure provided in Section~\ref{sec:bea_proof_structure}.

We start by  expanding $\vtheta(h)$ via Taylor expansion (Step 1 in the proof structure in Section~\ref{sec:bea_proof_structure}):
    \begin{align}
\vtheta(h) = \vtheta(0) + h \dot{\vtheta} + \sum_{k=2}^{\infty} \frac{1}{k!} \vtheta^{(k)} 
                 = \vtheta_{t-1} +  h f(\vtheta_{t-1}) + \sum_{n=1}^{\infty} h^{n+1} f_n + \sum_{k=2}^{\infty} \frac{h^{k}}{k!}  \vtheta^{(k)} 
\label{eq:modified_ode_expansion}
\end{align}
We now express in ${\vtheta}^{(k)}$ in terms of the wanted quantities $f_n$ via  Lemma~\ref{lemma:theta_n_expansion}:
    \begin{align}\vtheta^{(k)} = \sum_{m \ge 0}h^m  \frac{d^{k-2}}{dt^{k-2}} \sum_{i+j=m}  \jacparam{\vtheta}{f_j} f_i \hspace{1em} k\ge2
    \label{eq:theta_k_expansion_result}
    \end{align}
By replacing Eq~\eqref{eq:theta_k_expansion_result} in Eq~\eqref{eq:modified_ode_expansion} we obtain (Step 2 in the proof structure in Section~\ref{sec:bea_proof_structure}):
       \begin{align}
       \vtheta(h) &= \vtheta_{t-1} +  h f(\vtheta_{t-1}) + \sum_{n=1}^{\infty} h^{n+1} f_n(\vtheta_{t-1})  \\&\hspace{1em}+ \sum_{k=2}^{\infty} \frac{h^k}{k!} \sum_{m \ge 0}h^m  \frac{d^{k-2}}{dt^{k-2}} \sum_{i+j=m}  \evaljac{\vtheta}{f_j}{\vtheta_{t-1}} f_i(\vtheta_{t-1})  \\
       & =  \vtheta_{t-1} +  h f(\vtheta_{t-1}) + \sum_{n=1}^{\infty} h^{n+1} f_n(\vtheta_{t-1}) \\&\hspace{1em}+ \sum_{m \ge 0} \sum_{k=2}^{\infty} \frac{h^{k+m }}{k!}  \frac{d^{k-2}}{dt^{k-2}} \sum_{i+j=m}  \evaljac{\vtheta}{f_j}{\vtheta_{t-1}} f_i(\vtheta_{t-1})
       \label{eq:theta_h_exp_f_i}
\end{align}
 We note that due to the derivative w.r.t. $t$ and a dependence of $h$ in the modified vector field, there will be a dependence of $h$ in $\frac{d^{k-2}}{dt^{k-2}} \sum_{i+j=m}  \jacparam{\vtheta}{f_j} f_i$, since the chain rule will expose $\frac{d \vtheta}{d t}$ which depends on $h$ (Eq~\eqref{eq:modified_bea_app}). To highlight this dependence on $h$ we introduce the notation
       \begin{align} 
        \frac{d^{k-2}}{dt^{k-2}} \sum_{i+j=m}  \jacparam{\vtheta}{f_j} f_i = \sum_p h^p A_{k,m}^{p}
        \label{eq:f_n_based_on_a}
     \end{align}
where $A_{k,m}^{p}: \mathbb{R}^D \rightarrow \mathbb{R}^D$ is defined to be the term of order $h^p$ in $\frac{d^{k-2}}{dt^{k-2}} \sum_{i+j=m}  \jacparam{\vtheta}{f_j} f_i$; thus by definition $A_{k,m}^{p}$ does not depend on $h$. Replacing the above in Eq~\eqref{eq:theta_h_exp_f_i}, we conclude Step 3 outlined in the proof structure in Section~\ref{sec:bea_proof_structure}:
    \begin{align}
       \vtheta(h)  =  \vtheta_{t-1} +  h f(\vtheta_{t-1}) + \sum_{n=1}^{\infty} h^{n+1} f_n(\vtheta_{t-1}) +  \sum_{m \ge 0} \sum_{k=2}^{\infty} \frac{h^{k+m }}{k!} \sum_p  h^p A_{k,m}^{p}(\vtheta_{t-1})
\end{align}
Since the Euler update $\vtheta_t = \vtheta_{t-1} +hf(\vtheta_{t-1})$ does not involve terms of order higher than $h$ the $f_n$ terms are obtained by cancelling the terms of order $\mathcal{O}(h^{n+1})$  in $\vtheta(h)$ (Step 4 outlined in the proof structure in Section~\ref{sec:bea_proof_structure}), we have that for $k+m+p = n+1$:
     \begin{align}
        f_n = - \sum_{m \ge 0} \sum_{k=2}^{\infty} \frac{1}{k!}A_{k,m}^{n+1-(k+m)}
     \end{align}
   Since $A_{k,m}^{n+1-(k+m)}$ depends on the functions $f_n$, we have found a recursive relation for $f_n$. We now use induction to show that
     \begin{align}
      f_n = c_n  {(\jacthetaf)}^{n} f + \csecondorderfam
     \end{align}
    where $\csecondorderfam$ denotes a class of functions defined as a sum of individual terms each containing higher than first order derivatives applied to $f$  (Lemma~\ref{lemma:induction}). The same proof shows the recurrence relation for $c_n$:
    \begin{align}
        c_n = - \sum_{k=2}^{n+1} \frac{1}{k!} \sum_{l_1+ l_2+ \dots+l_k = n - k + 1} c_{l_1} c_{l_2} \dots c_{l_k}
    \end{align}
with solution $c_n = \frac{(-1)^{n}}{n+1}$ (Lemma~\ref{lemma:c_n_recurrence_proof}).

We have found that each $f_n$ can be written as $\frac{(-1)^{n}}{n+1} {(\jacthetaf)}^n f + \csecondorderfam$. Replacing this in Eq.\eqref{eq:modified_bea_app} concludes the proof.
\end{proof}

In the case of gradient descent, where $f = - \nabla_{\vtheta} E$ and  $\jacthetaf = \nabla_{\vtheta}^2E$, we have that the modified flow has the form 
\begin{align}
    \dot{\vtheta} &= \sum_{n=0}^{p} \frac{(-1)^{n}}{n+1}  (-\jacthetaf)^n (- \nabla_{\vtheta} E) + \csecondorderfam \\
                &= \sum_{n=0}^{p} \frac{(-1)^{n}}{n+1}  (-\nabla_{\vtheta}^2 E)^n (- \nabla_{\vtheta} E) + \csecondorderfam \\
                &= \sum_{n=0}^{p} \frac{-1}{n+1}  (\nabla_{\vtheta}^2 E)^n (\nabla_{\vtheta} E) + \csecondorderfam,
\end{align}
which is the statement of Theorem~\ref{thm:order_n_flow} in the main thesis.

\begin{lemma} 
    \begin{align}
    \vtheta^{(k)} = \sum_{m \ge 0}h^m  \frac{d^{k-2}}{dt^{k-2}} \sum_{i+j=m}  \jacparam{\vtheta}{f_j} f_i \hspace{1em} k\ge2
    \end{align}
\label{lemma:theta_n_expansion}
\end{lemma}

\begin{proof}

\begin{align}
\dot{\vtheta} &= \tilde{f} \\
\dot{\dot{\vtheta}} &= \jacparam{\vtheta}{\tilde{f}}  \tilde{f} \\
{\vtheta}^{(k)} &= \frac{d^{k-2}}{dt^{k-2}} \dot{\dot{\vtheta}}= \frac{d^{k-2}}{dt^{k-2}}  \jacparam{\vtheta}{\tilde{f}} \tilde{f}
\end{align}

We have that
\begin{align}
 \jacparam{\vtheta}{\tilde{f}} \tilde{f} = \left( \sum_i h^i \jacparam{\vtheta}{f_i} \right) \left( \sum_j h^j f_j \right)= \sum_{m \ge 0} h^m \sum_{i+j=m}  \jacparam{\vtheta}{f_i} f_j
\end{align}
which concludes the proof.
We note that this is an adaptation of Lemma A.2 from~\citep{igr}. We need this adaptation to avoid assuming that $f$  is a symmetric vector field---i.e. that $\jacparam{\vtheta}{\tilde{f}} \tilde{f}= \frac{1}{2} \nabla_{\vtheta} \norm{\tilde{f}}^2$.
\end{proof}

\begin{lemma} By induction we have that:
    \begin{align}
      f_n = c_n  {(\jacthetaf)}^{n} f + \csecondorderfam
     \end{align}
with 
\begin{align}
c_n = - \sum_{k=2}^{n+1} \frac{1}{k!} \sum_{l_1+ l_2+ \dots+l_k = n - k + 1} c_{l_1} c_{l_2} \dots c_{l_k}
\end{align}
\label{lemma:induction}
\end{lemma}

\begin{proof}

\textbf{Base step}: This is true for $n=1, 2$ and $3$, for which we computed all terms in the BEA expansion. 

\textbf{Induction step}:
We know that (Eq~\eqref{eq:f_n_based_on_a}):
    \begin{align}
        f_n = - \sum_{m \ge 0} \sum_{k=2}^{\infty} \frac{1}{k!}A_{k,m}^{n+1-(k+m)}
        \label{eq:f_n_as_function_a}
     \end{align}
where  $A_{k,m}^{p}$ is defined as the coefficient of $h^p$ in:
\begin{align}
        \frac{d^{k-2}}{dt^{k-2}} \sum_{i+j=m}  \jacparam{\vtheta}{f_j} f_i = \sum_p h^p A_{k,m}^{p}
\end{align}
 Under the induction hypothesis we can use that 
 \begin{align}
      f_i = c_i  {(\jacthetaf)}^{i} f + \csecondorderfam \hspace{2em} \forall i < n
\end{align}
And thus that the flow is of the form:
 \begin{align}
\dot{\vtheta} =  f + \sum_{i=1}^{n-1} c_i{(\jacthetaf)}^{i} f + \csecondorderfam + \mathcal{O}(h^n)
\label{eq:flow_under_ih}
 \end{align}
To find $f_n$ (Eq~\eqref{eq:f_n_as_function_a}) we need to find $A_{k,m}^{n+1-(k+m)}$ with $k\ge 2$ and $m\ge 0$. We thus need to find  $A_{k,m}^{p}$ with $p \le n-1$ (since $p \le (n+1) - (k+m) \text{ and } k\ge 2, m\ge 0 \implies p \le n -1$). We do so expanding $\frac{d^{k-2}}{dt^{k-2}} \sum_{i+j=m}  \jacparam{\vtheta}{f_j} f_i$ under the induction step. Since $i+j = m$ and $k+m \le n+1$ and $k\ge 2$, we have that $m \le n-1$ and thus $i, j \le n-1$ and thus we can apply the induction hypothesis for $f_i$ and $f_j$.

We annotate the steps with IH (Induction Hypothesis, often by using Eq. \eqref{eq:flow_under_ih}) and CR (Chain Rule) and S (Simplifying or grouping terms of $\csecondorderfam)$ by using that products and sum of terms in this function class also belong in this function class).

\begingroup
\allowdisplaybreaks
\begin{align}
&\frac{d^{k-2}}{dt^{k-2}} \sum_{i+j=m}  \jacparam{\vtheta}{f_j} f_i \\
&= 
 \frac{d^{k-2}}{dt^{k-2}} \sum_{i+j=m} \raw{\vtheta} {c_j {(\jacthetaf)}^{j} f + \csecondorderfam}(c_i {(\jacthetaf)}^{i} f+ \csecondorderfam)  \owntag{IH}\\
&= \frac{d^{k-2}}{dt^{k-2}} \sum_{i+j=m} \left(c_j {(\jacthetaf)}^{j+1} +c_j  f^T \raw{\vtheta}{{(\jacthetaf)}^{j}}  +   \csecondorderfam\right)(c_i {(\jacthetaf)}^{i} f + \csecondorderfam) \owntag{CR} \\
&= \frac{d^{k-2}}{dt^{k-2}} \sum_{i+j=m} \left(c_j {(\jacthetaf)}^{j+1} +   \csecondorderfam\right)(c_i  {(\jacthetaf)}^{i} f + \csecondorderfam) \owntag{S} \\
&= \frac{d^{k-2}}{dt^{k-2}} \sum_{i+j=m} c_i c_j {(\jacthetaf)}^{m+1} f+ 
    \csecondorderfam  \owntag{S} \\
    &= \frac{d^{k-3}}{dt^{k-3}} \sum_{i+j=m} \frac{d}{dt} c_i c_j {(\jacthetaf)}^{m+1} f+ 
    \csecondorderfam \\
        &= \frac{d^{k-3}}{dt^{k-3}} \sum_{i+j=m} \frac{d}{d\vtheta} \left(c_i c_j {(\jacthetaf)}^{m+1} f +  
    \csecondorderfam \right)\frac{d\vtheta}{d t} \owntag{CR}\\
&= \frac{d^{k-3}}{dt^{k-3}}  \sum_{i+j=m}  \frac{d}{d \vtheta} \left(c_i c_j {(\jacthetaf)}^{m+1} f + \csecondorderfam \right) \left(\sum_{l_1=0}^{n-1} h^{l_1} f_{l_1} + \mathcal{O}(h^n)\right)\owntag{Eq. \eqref{eq:flow_under_ih}} \\
&= \frac{d^{k-3}}{dt^{k-3}}  \sum_{i+j=m}  \frac{d}{d \vtheta} \left(c_i c_j {(\jacthetaf)}^{m+1} f + \csecondorderfam \right) (\sum_{l_1=0}^{n-1} h^{l_1} f_{l_1})   + \mathcal{O}(h^n)  \\
&= \frac{d^{k-3}}{dt^{k-3}}  \sum_{i+j=m}   \left(c_i c_j {(\jacthetaf)}^{m+2} + \csecondorderfam \right) (\sum_{l_1=0}^{n-1} h^{l_1} f_{l_1})  + \mathcal{O}(h^n) \owntag{CR, S}\\
&= \frac{d^{k-3}}{dt^{k-3}}  \sum_{i+j=m}   \left(c_i c_j {(\jacthetaf)}^{m+2} + \csecondorderfam\right) \left(\sum_{l_1=0}^{n-1} h^{l_1} c_{l_1}  (\jacthetaf) ^{l_1} f + \csecondorderfam \right) \nonumber\\ &\hspace{2em} + \mathcal{O}(h^n) \owntag{IH}
\end{align}
\endgroup
Continuing by applying the induction hypothesis and the chain rule we obtain:
\begin{align}
&\frac{d^{k-2}}{dt^{k-2}} \sum_{i+j=m}  \jacparam{\vtheta}{f_j} f_i \\
&= \frac{d^{k-3}}{dt^{k-3}}  \sum_{i+j=m} \sum_{l_1=0}^{n-1} h^{l_1} c_i c_j c_{l_1} {(\jacthetaf)}^{l_1+m+2} f + \csecondorderfam + \mathcal{O}(h^n)
\owntag{S} \\
&= \frac{d^{k-4}}{dt^{k-4}}  \sum_{i+j=m} \left(\sum_{l_1=0}^{n-1} h^{l_1} c_i c_j c_{l_1}  \frac{d}{d \vtheta} \left( {(\jacthetaf)}^{l_1+m+2} f + \csecondorderfam\right)\right)(\sum_{l_2=0}^{n-1} h^{l_2} f_{l_2})  + \mathcal{O}(h^n) \owntag{CR, Eq. \eqref{eq:flow_under_ih}} \\
&= \frac{d^{k-4}}{dt^{k-4}}  \sum_{i+j=m} \left(\sum_{l_1=0}^{n-1} h^{l_1} c_i c_j c_{l_1} \left( {(\jacparam{\vtheta}{f})}^{l_1+m+3} + \csecondorderfam\right)\right) \\ &\hspace{6.1em} \left(\sum_{l_2=0}^{n-1} h^{l_2} c_{l_2}  {(\jacthetaf)}^{l_2} f + \csecondorderfam\right)  + \mathcal{O}(h^n) \owntag{CR, IH}\\
&= \frac{d^{k-4}}{dt^{k-4}}  \sum_{i+j=m} \sum_{l_1=0}^{n-1} \sum_{l_2=0}^{n-1} h^{l_1+l_2} c_i c_j c_{l_1} c_{l_2}  {(\jacthetaf)}^{l_1+ l_2+ m+3} f + \csecondorderfam + \mathcal{O}(h^n) \owntag{S}\\
&= \sum_{i+j=m} \sum_{l_1=0}^{n-1} \sum_{l_2=0}^{n-1}  \dots \sum_{l_{k-2}=0}^{n-1} h^{l_1+l_2 + \dots+l_{k-2}} c_i c_j c_{l_1} c_{l_2} \dots c_{l_k-2} {(\jacthetaf)}^{l_1+ l_2+ \dots+{l_{k-2}} + m+k-1} f \nonumber \\&\hspace{5em} + \csecondorderfam  + \mathcal{O}(h^n) \\
&= \sum_{i+j=m} \sum_{\substack{0 \le l_1,\dots, l_{k-2}\le n-1,\\ l_1 + l_2 +\dots l_{k-2} = p}}  h^{p} c_i c_j c_{l_1} c_{l_2} \dots c_{l_k-2}  {(\jacthetaf)}^{p+  m+k-1} f+ \csecondorderfam   + \mathcal{O}(h^n)
\end{align}
By extracting the terms of $\mathcal{O}(h^p)$ with $p \le n -1$ from the above we can now conclude that (note that since we are now only concerned with $p \le n -1$ we drop the $l_i \le n-1$ since it is implied from $ l_1 + l_2 +\dots l_{k-2} = p$):
\begin{align}
A_{k,m}^{p} = \sum_{i+j=m} \sum_{\substack{l_1,\dots, l_{k-2}\ge 0,\\ l_1 + l_2 +\dots l_{k-2} = p}}  c_i c_j c_{l_1} c_{l_2} \dots c_{l_k-2}  {(\jacthetaf)}^{p+ m+k-1} f+ \csecondorderfam \\
 \hspace{10em} \forall p \le n-1
\end{align}
Replacing this in $f_n$:
     \begin{align}
       &f_n = - \sum_{m \ge 0} \sum_{k=2}^{\infty} \frac{1}{k!}A_{k,m}^{n+1-(k+m)} \\
            &= - \sum_{m \ge 0} \sum_{k=2}^{\infty} \frac{1}{k!} \sum_{i+j=m} \sum_{\substack{l_1,\dots, l_{k-2}\ge 0,\\ l_1 + l_2 +\dots l_{k-2} = n+1-(k+m)}} \hspace{-2.5em} c_i c_j c_{l_1} c_{l_2} \dots c_{l_k-2}  {(\jacthetaf)}^{n+1-(k+m)+ k+m-1} f \nonumber \\& \hspace{3em}+ \csecondorderfam\\
             &= - \sum_{m \ge 0} \sum_{k=2}^{\infty} \frac{1}{k!} \sum_{i+j=m}  \sum_{\substack{l_1,\dots, l_{k-2}\ge 0,\\ l_1 + l_2 +\dots l_{k-2} = n+1-(k+m)}} \hspace{-2.8em} c_i c_j c_{l_1} c_{l_2} \dots c_{l_k-2}  {(\jacthetaf)}^n f+ \csecondorderfam\\
               &= -  \sum_{k=2}^{\infty} \frac{1}{k!} \sum_{m \ge 0} \sum_{\substack{l_1,\dots, l_{k}\ge 0,\\ l_1 + l_2 +\dots l_k = n+1-k}}  c_{l_1} c_{l_2} \dots c_{l_k} {(\jacthetaf)}^n f + \csecondorderfam
     \end{align}
We can now say that $f_n = c_n  {(\jacthetaf)}^{n} f + \csecondorderfam$. Not only that, we have now found the recurrence relation we were seeking:
\begin{align}
c_n = - \sum_{k=2}^{n+1} \frac{1}{k!} \sum_{l_1+ l_2+ \dots+l_k = n - k + 1} c_{l_1} c_{l_2} \dots c_{l_k}
\label{eq:rec_coeffs}
\end{align}
\end{proof}

\begin{lemma} The solution to the recurrence relation
\begin{align}
c_n = - \sum_{k=2}^{n+1} \frac{1}{k!} \sum_{l_1+ l_2+ \dots+l_k = n - k + 1} c_{l_1} c_{l_2} \dots c_{l_k}
\end{align}
with $c_0 = 1$, is $c_n = \frac{(-1)^{n}}{n+1}$.
\label{lemma:c_n_recurrence_proof}
\end{lemma}
\begin{proof}

We will use generating functions to solve the recurrence. Let $c(x)$ be the generating function of $c_n$:
\begin{align}
c(x) = c_0  + \sum_{n=1}^{\infty} c_n x^n
\end{align}

We have that:
\begin{align}
c(x) &= 1 - \sum_{n=1}^{\infty} x^n \left(\sum_{k=2}^{n+1}\frac{1}{k!} \sum_{l_1+ l_2+ \dots+l_k = n - k + 1} c_{l_1} c_{l_2} \dots c_{l_k}\right)\\
     &= 1 -\frac{1}{x} \sum_{n=1}^{\infty} x^{n+1} \left(\sum_{k=2}^{n+1}\frac{1}{k!} \sum_{l_1+ l_2+ \dots+l_k = n - k + 1} c_{l_1} c_{l_2} \dots c_{l_k}\right)\\
        &= 1 -\frac{1}{x} \sum_{n=1}^{\infty}  \left(\sum_{k=2}^{n+1}\frac{x^k}{k!} x^{n+1-k} \sum_{l_1+ l_2+ \dots+l_k = n - k + 1} c_{l_1} c_{l_2} \dots c_{l_k}\right)\\
     &= 1 - \frac{1}{x} \sum_{n=1}^{\infty} \left(\sum_{k=2}^{n+1}\frac{x^k}{k!} \sum_{l_1+ l_2+ \dots+l_k = n - k + 1} c_{l_1}x^{l_1} c_{l_2}x^{l_2} \dots c_{l_k} x^{l_k}\right)\\
     &= 1 - \frac{1}{x} \sum_{k=2}^{\infty}\frac{x^k}{k!}\left( \sum_{l_1=0}^{\infty}\sum_{l_2=0}^{\infty}\sum_{l_k=0}^{\infty} c_{l_1}x^{l_1} c_{l_2}x^{l_2} \dots c_{l_k} x^{l_k}\right) \label{eq:sum_transition}\\
     &= 1 - \frac{1}{x} \sum_{k=2}^{\infty}\frac{x^k}{k!}\left( \sum_{l_1=0}^{\infty} c_{l_1}x^{l_1} \right)\left(\sum_{l_2=0}^{\infty}c_{l_2}x^{l_2}\right)  \dots \left( \sum_{l_k=0}^{\infty}  c_{l_k} x^{l_k}\right)\\
      &= 1 - \frac{1}{x} \sum_{k=2}^{\infty}\frac{x^k}{k!} c(x)^k\\
      &= 1 - \frac{1}{x} (e^{x c(x)} -1 - x c(x))
      \label{eq:c_x_rec}
\end{align}
Where Eq \eqref{eq:sum_transition} can be obtained via the following reasoning:
\begin{align}
& \sum_{k=2}^{\infty}\frac{x^k}{k!}\left( \sum_{l_1=0}^{\infty}\sum_{l_2=0}^{\infty}\sum_{l_k=0}^{\infty} c_{l_1}x^{l_1} c_{l_2}x^{l_2} \dots c_{l_k} x^{l_k}\right) \\
& = \sum_{k=2}^{\infty}\frac{x^k}{k!}\left(\sum_{m=0}^{\infty} \sum_{l_1 + l_2 +\dots l_k = m} c_{l_1}x^{l_1} c_{l_2}x^{l_2} \dots c_{l_k} x^{l_k}\right)  \\
& = \sum_{k=2}^{\infty}\frac{x^k}{k!}\left(\sum_{n=k-1}^{\infty} \sum_{l_1 + l_2 +\dots l_k = n+1-k} c_{l_1}x^{l_1} c_{l_2}x^{l_2} \dots c_{l_k} x^{l_k}\right)  \\
& = \sum_{k=2}^{\infty}\sum_{n=k-1}^{\infty}\frac{x^k}{k!} \sum_{l_1 + l_2 +\dots l_k = n+1-k} c_{l_1}x^{l_1} c_{l_2}x^{l_2} \dots c_{l_k} x^{l_k}  \\
& = \sum_{n=1}^{\infty}\sum_{k=2}^{n+1}\frac{x^k}{k!} \sum_{l_1 + l_2 +\dots l_k = n+1-k} c_{l_1}x^{l_1} c_{l_2}x^{l_2} \dots c_{l_k} x^{l_k}
\end{align}

Solving for $c(x)$ in Eq~\eqref{eq:c_x_rec} we obtain: 
\begin{align}
c(x) &=  1 - \frac{1}{x} (e^{x c(x)} -1 - x c(x)) \\
c(x) &=  1 - \frac{1}{x} e^{x c(x)} + \frac 1 x + c(x) \\
0 &=  1 - \frac{1}{x} e^{x c(x)} + \frac 1 x\\
 e^{x c(x)} &=  x + 1\\
 x c(x) &=  \log(x + 1)\\
 c(x) &=  \frac{\log(x + 1)}{x}
\end{align}

We now perform the series expansion for $c(x)$:
\begin{align}
 c(x) =  \frac{\log(x + 1)}{x}
      =  \frac{ \sum_{n=1}^{\infty} \frac{(-1)^{n-1}}{n} x^n}{x}
      =  \sum_{n=1}^{\infty} \frac{(-1)^{n-1}}{n} x^{n-1}
      =  \sum_{n=0}^{\infty} \frac{(-1)^{n}}{n+1} x^{n}
\end{align}

We thus have that $c_n = \frac{(-1)^{n}}{n+1}$ which finishes the proof.

\end{proof}

\subsection{Linearisation results around critical points}
\label{sec:linearisation}

Assume we are around a critical point $\vtheta^*$, i.e. $\nabla_{\vtheta} E(\vtheta^*)= \mathbf{0}$. Via the Hartman–Grobman theorem~\citep{grobman,hartman1960lemma} we can write:
\begin{align}
 \nabla_{\vtheta} E \approx  \nabla_{\vtheta} E(\vtheta^*) + \nabla_{\vtheta}^2 E(\vtheta^*) (\vtheta - \vtheta^*) = \nabla_{\vtheta}^2 E(\vtheta^*) (\vtheta - \vtheta^*)
\label{eq:critical_point_approx}
\end{align}
Replacing this in the gradient descent updates:
\begin{align}
\vtheta_t &= \vtheta_{t-1} - h \nabla_{\vtheta} E(\vtheta_{t-1}) \implies \vtheta_t - \vtheta^* \approx  (I - h \nabla_{\vtheta}^2 E(\vtheta^*)) (\vtheta_{t-1} - \vtheta^*) \\& \implies \vtheta_t - \vtheta^* \approx  (I - h \nabla_{\vtheta}^2 E(\vtheta^*))^t (\vtheta_{0} - \vtheta^*)
\end{align}
Since $\left(I - h \nabla_{\vtheta}^2 E(\vtheta^*)\right)^t (\vtheta_{0} - \vtheta^*) = \sum_{i=0}^{D-1} (1 - h \lambda_i^*)^t {\vu_i^*} {\vu_i^*}^T (\vtheta_{0} - \vtheta^*)$, gradient descent converges in the limit of $t \to \infty$ when $|1 - h \lambda_i^*|< 1$, i.e. $\lambda_i < 2/h$. Thus, we obtain the same conclusion as obtained with the PF around a local minima.

We now also show how we can use these results to derived a specialised version of the PF, one that is only valid around critical points and requires the additional assumption that the Hessian of the critical point is invertible. 
We start with
\begin{align}
\vtheta_t - \vtheta^*  \approx (I - h \nabla_{\vtheta}^2 E(\vtheta^*))^t (\vtheta_{0} - \vtheta^*) = \sum_{i=0}^{D-1} (1 - h \lambda_i^*)^t {\vu_i^*} {\vu_i^*}^T (\vtheta_{0} - \vtheta^*).
\end{align}
Under the above approximation we have:
\begin{align}
\vtheta_t = \vtheta^* + \sum_{i=0}^{D-1} (1 - h \lambda_i^*)^t {\vu_i^*} {\vu_i^*}^T  (\vtheta_{0} - \vtheta^*)
\label{eq:linearisation_expansion_theta_t}
\end{align}
which we can write in continuous form as (we use that iteration $n$ is at time $t =nh$): 
\begin{align}
\vtheta(t) &= \vtheta^* + \sum_{i=0}^{D-1} (1 - h \lambda_i^*)^{t/h} {\vu_i^*} {\vu_i^*}^T (\vtheta_{0} - \vtheta^*) \\          & =\vtheta^* + \sum_{i=0}^{D-1} e^{\log(1 - h \lambda_i^*){t/h}} {\vu_i^*} {\vu_i^*}^T (\vtheta_{0} - \vtheta^*).
\end{align}
Taking the derivative wrt to $t$:
\begin{align}
\dot{\vtheta} = \sum_{i=0}^{D-1} \frac{\log(1 - h \lambda_i^*)}{h} \underbrace{e^{\log{(1 - h \lambda_i^*)} {t/h}} {\vu_i^*} {\vu_i^*}^T (\vtheta_{0} - \vtheta^*)}_{Eq. \eqref{eq:linearisation_expansion_theta_t}} = \sum_{i=0}^{D-1} \frac{\log(1 - h \lambda_i^*)}{h} (\vtheta - \vtheta^*)
\end{align}
From here, if we assume the Hessian at the critical point is invertible we can write from the approximation above (Eq. \eqref{eq:critical_point_approx}) that 
$\vtheta - \vtheta^* = (\nabla_{\vtheta}^2 E)^{-1} \nabla_{\vtheta} E = \sum_{i=0}^{D-1} \frac{1}{\lambda_i^*} \nabla_{\vtheta} E^T {\vu_i^*} {\vu_i^*}^T$ and replacing this in the above we the PF:
\begin{align}
\dot{\vtheta} = \sum_{i=0}^{D-1} \frac{\log(1 - h \lambda_i^*)}{h \lambda_i^*}  \nabla_{\vtheta} E^T {\vu_i^*} {\vu_i^*}^T.
\end{align}

\subsection{The solution of the PF for quadratic losses}

We derive the solution of the PF in the case of quadratic losses - Remark~\ref{remark:quadractic_remark} in the main thesis. We have $E(\vtheta) = \frac{1}{2} \vtheta^t \vA \vtheta + \vb^T \vtheta$  with $\vA$ symmetric. $\nabla_{\vtheta} E = \vA \vtheta + \vb$ and $\nabla_{\vtheta}^2 E = \vA$. Thus $\lambda_i$ and $\vu_i$ are the eigenvalues and eigenvectors of $\vA$ and do not change based on $\vtheta$. Replacing this in the PF leads to:
\begin{align}
\dot{\vtheta} &= \sum_{i=0}^{D-1} \frac{\log (1 - h\lambda_i)}{h \lambda_i} \left(\vA \vtheta + b \right)^T \vu_i  \vu_i \\
 &= \sum_{i=0}^{D-1} \frac{\log (1 - h\lambda_i)}{h \lambda_i} (\vA \vtheta )^T \vu_i \vu_i + \sum_{i=0}^{D-1} \frac{\log (1 - h\lambda_i)}{h \lambda_i} b^T \vu_i  \vu_i \\
  &= \sum_{i=0}^{D-1} \frac{\log (1 - h\lambda_i)}{h \lambda_i} \vtheta^T \vA \vu_i \vu_i + \sum_{i=0}^{D-1} \frac{\log (1 - h\lambda_i)}{h \lambda_i} b^T \vu_i  \vu_i \\
    &= \sum_{i=0}^{D-1} \frac{\log (1 - h\lambda_i)}{h \lambda_i}\lambda_i \vtheta^T  \vu_i \vu_i + \sum_{i=0}^{D-1} \frac{\log (1 - h\lambda_i)}{h \lambda_i} b^T \vu_i  \vu_i \\
    &= \sum_{i=0}^{D-1} \frac{\log (1 - h\lambda_i)}{h } \vtheta^T  \vu_i \vu_i + \sum_{i=0}^{D-1} \frac{\log (1 - h\lambda_i)}{h \lambda_i} b^T \vu_i  \vu_i
\end{align}
From here
\begin{align}
\dot{(\vtheta^T \vu_i)} &= \frac{\log (1 - h\lambda_i)}{h } \vtheta^T  \vu_i +  \frac{\log (1 - h\lambda_i)}{h \lambda_i} b^T \vu_i.
\end{align}

The solution of this flow is $(\vtheta^T \vu_i)(t)  = e^{\frac{\log (1 - h\lambda_i)}{h } t} \vtheta_0^T  \vu_i + t \frac{\log (1 - h\lambda_i)}{h \lambda_i} \vb^T \vu_i$.

Since $\vu_i$ form a basis of $\mathbb{R}^D$, we can now write
\begin{align}
\vtheta(t) &= \sum_{i=0}^{D-1} \vtheta(t)^T \vu_i \vu_i  \\
          &= \sum_{i=0}^{D-1} \left(e^{\frac{\log (1 - h\lambda_i)}{h } t} \vtheta_0^T  \vu_i + t \frac{\log (1 - h\lambda_i)}{h \lambda_i} b^T \vu_i \right) \vu_i .
\end{align}

We thus have obtained the solution described in Remark~\ref{remark:quadractic_remark} in the main text of the thesis.

\subsection{The Jacobian of the PF at critical points}
\label{sec:jacobian}

We compute Jacobian of the PF at a critical point $\vtheta^*$ with the eigenvalues and eigenvectors of $\nabla_{\vtheta}^2E(\vtheta^*)$ denoted as $\lambda_i^*$ and $\vu_i^*$. We will use that $\nabla_{\vtheta}E(\vtheta^*) = \myvecsym{0}$.
 
\begin{align}
\vJ_{\text{PF}}(\vtheta^*) &= \sum_{i=0}^{D-1} \frac{d (\nabla_{\vtheta} E^T \vu_i)  \frac{\log(1 - h \lambda_i)}{h \lambda_i} \vu_i}{d\vtheta} (\vtheta^*) \\
            &= \sum_{i=0}^{D-1} \frac{d (\nabla_{\vtheta} E^T \vu_i)  \frac{\log(1 - h \lambda_i)}{h \lambda_i}}{d\vtheta} (\vtheta^*) {\vu_i^*}^T \\&\hspace{1em} + \sum_{i=0}^{D-1} \underbrace{ \nabla_{\vtheta} E(\vtheta^*)^T}_{\mathbf{0}}  \vu_i^*  \frac{\log(1 - h \lambda_i^*)}{h \lambda_i^*} \frac{d \vu_i}{d\vtheta}\\
              &= \sum_{i=0}^{D-1} \frac{d (\nabla_{\vtheta} E^T \vu_i)  \frac{\log(1 - h \lambda_i)}{h \lambda_i}}{d\vtheta} (\vtheta^*) {\vu_i^*}^T \\
              &= \sum_{i=0}^{D-1} \frac{\log(1 - h \lambda_i^*)}{h \lambda_i^*} \frac{d (\nabla_{\vtheta} E^T \vu_i) }{d\vtheta}(\vtheta^*) {\vu_i^*}^T \\&\hspace{1em}  + \sum_{i=0}^{D-1} (\underbrace{\nabla_{\vtheta} E(\vtheta^*)^T}_{\mathbf{0}} \vu_i^*) \frac{d   \frac{\log(1 - h \lambda_i)}{h \lambda_i}}{d\vtheta} {\vu_i^*}^T \\
                &= \sum_{i=0}^{D-1} \frac{\log(1 - h \lambda_i^*)}{h \lambda_i^*} \frac{d (\nabla_{\vtheta} E^T \vu_i) }{d\vtheta}(\vtheta^*) {\vu_i^*}^T\\
                &= \sum_{i=0}^{D-1} \frac{\log(1 - h \lambda_i^*)}{h \lambda_i^*} \left(\frac{d \nabla_{\vtheta} E}{d\vtheta}(\vtheta^*)\right)^T \vu_i^* {\vu_i^*}^T \\&\hspace{1em}+  \sum_{i=0}^{D-1} \frac{\log(1 - h \lambda_i^*)}{h \lambda_i^*} \left(\frac{d \vu_i }{d\vtheta}(\vtheta^*)\right)^T  \underbrace{\nabla_{\vtheta} E(\vtheta^*)}_{\mathbf{0}}  {\vu_i^*}^T \\
                &= \sum_{i=0}^{D-1} \frac{\log(1 - h \lambda_i^*)}{h \lambda_i^*} \nabla_{\vtheta}^2E(\vtheta^*) \vu_i^* {\vu_i^*}^T \\
                &= \sum_{i=0}^{D-1} \frac{\log(1 - h \lambda_i^*)}{h \lambda_i^*} \lambda_i^* \vu_i^* {\vu_i^*}^T \\
                &= \sum_{i=0}^{D-1} \frac{\log(1 - h \lambda_i^*)}{h} \vu_i^* {\vu_i^*}^T
\end{align}

We thus arrived at the result used in Equation~\ref{eq:jac_pf_main} of the thesis, where we used to perform stability analysis on the PF.
This results shows that the eigenvalues of the Jacobian for the PF are $\frac{\log(1 - h \lambda_i^*)}{h}$, thus local minima where $\lambda_i^* \ge 0 $ are only attractive if $h \lambda_i^* <2$.

\subsection{Jacobians of the IGR flow and NGF at critical points}
\label{sec:jacobian_igr_ngf}
For the NGF, we have
\begin{align}
\vJ_{\text{NGF}}(\vtheta^*) &= - \nabla_{\vtheta}^2 E(\vtheta^*) = - \sum_{i=0}^{D-1} \lambda_i^* \vu_i^* {\vu_i^*}^T
\end{align}
and thus the eigenvalues of the Jacobian for the NGF are $-\lambda_i^*$. Thus local minima where $\lambda_i^* > 0 \implies -\lambda_i^* < 0$  are attractive for the NGF.

For the IGR flow, we have
\begin{align}
\vJ_{\text{IGR}}(\vtheta^*) &= -\frac{d \left(\nabla_{\vtheta} E + \frac{h^2}{2} \nabla_{\vtheta}^2 E \nabla_{\vtheta} E \right)}{d\vtheta}\left( \vtheta^*\right) \\
                  &= -\frac{d \nabla_{\vtheta} E}{d \vtheta}\left( \vtheta^*\right) - \frac{h^2}{2} \frac{d \nabla_{\vtheta}^2 E \nabla_{\vtheta} E }{d\vtheta}\left( \vtheta^*\right) \\
                  &= - \sum_{i=0}^{D-1} \lambda_i^* \vu_i^* {\vu_i^*}^T - \frac{h^2}{2} \left( \nabla_{\vtheta} E^T \nabla_{\vtheta}^3 E   + \nabla_{\vtheta}^2 E \frac{d \nabla_{\vtheta} E}{d\vtheta} \right) \left( \vtheta^*\right)\\ 
                 &= - \sum_{i=0}^{D-1} \lambda_i^* \vu_i^* {\vu_i^*}^T  - \frac{h^2}{2}  \underbrace{ \nabla_{\vtheta} E(\vtheta^*)^T}_{\mathbf{0}}\nabla_{\vtheta}^3 E(\vtheta^*)   - \frac{h^2}{2}  \nabla_{\vtheta}^2 E \left( \vtheta^*\right) \nabla_{\vtheta}^2 E\left( \vtheta^*\right) \\
                  &= - \sum_{i=0}^{D-1} \lambda_i^* \vu_i^* {\vu_i^*}^T - \frac{h^2}{2}  \nabla_{\vtheta}^2 E \left( \vtheta^*\right) \nabla_{\vtheta}^2 E\left( \vtheta^*\right) \\
                  &= - \sum_{i=0}^{D-1} \lambda_i^* \vu_i^* {\vu_i^*}^T - \sum_{i=0}^{D-1} \frac{h^2}{2} {\lambda_i^*}^2 \vu_i^* {\vu_i^*}^T\\
                  &= - \sum_{i=0}^{D-1} \left(\lambda_i^* + \frac{h^2}{2} {\lambda_i^*}^2\right) \vu_i^* {\vu_i^*}^T
\end{align}
and thus the eigenvalues of the Jacobian for the IGR flow are $- \left(\lambda_i^* + \frac{h^2}{2} {\lambda_i^*}^2\right)$. Thus local minima where $\lambda_i^* > 0 \implies - \left(\lambda_i + \frac{h^2}{2} {\lambda_i^*}^2\right) < 0$  are attractive for the IGR flow.

\subsection{Multiple gradient descent steps}
\label{sec:multiple_steps_proof}

We now show that if we use BEA to construct a flow that has an error of order $\mathcal{O}(h^p)$
after 1 gradient descent step, then the error will be of the same order after 2 steps. The same argument can be applied to multiple steps, though the error scales proportionally to the number of steps and the bound becomes vacuous as the number of steps increases.
We have $\dot{\vtheta}$ with $\vtheta(0) = \vtheta_{t-1}$ tracks $\vtheta_{t} = \vtheta_{t-1} - h \nabla_{\vtheta} E(\vtheta_{t-1})$  with an error $\mathcal{O}(h^p)$ (i.e $\norm{\vtheta(h) - \vtheta_t}$ is $\mathcal{O}(h^p)$),  then $\dot{\vtheta}$ tracks two steps of gradient descent $\vtheta_{t} = \vtheta_{t-1} - h \nabla_{\vtheta} E(\vtheta_{t-1})$ and $\vtheta_{t+1} = \vtheta_t - h \nabla_{\vtheta} E(\vtheta_t)$  with an error of the same order $\mathcal{O}(h^p)$ (i.e. $\norm{\vtheta(2h) - \vtheta_{t+1}}$ is $\mathcal{O}(h^p)$). We prove this below. For the purpose of this proof we use the following notation $\vtheta(h; \vtheta')$ is the value of the flow at time $h$ with initial condition $\vtheta'$. Making the initial condition explicit is necessary for the proof.
\begin{align}
 \norm{\vtheta(2h; \vtheta_{t-1}) - \vtheta_{t+1}} 
 &= \norm{\vtheta(2h; \vtheta_{t-1}) - \vtheta(h; \vtheta_{t}) + \vtheta(h; \vtheta_{t}) - \vtheta_{t+1}} \\
 & \le \norm{\vtheta(2h; \vtheta_{t-1}) - \vtheta(h; \vtheta_{t})} + \norm{\vtheta(h; \vtheta_{t}) - \vtheta_{t+1}}\\
 &\le \norm{\vtheta(2h; \vtheta_{t-1}) - \vtheta(h; \vtheta_{t})} + \mathcal{O}(h^p) \\
 & =  \norm{\vtheta(h; \vtheta(h; \vtheta_{t-1})) - \vtheta(h; \vtheta_{t})} + \mathcal{O}(h^p)
\end{align}

We thus have to bound how the flow changes after time $h$ when starting with two different initial conditions $\vtheta(h; \vtheta_{t-1})$ and $\vtheta_{t})$. We also know that $\norm{\vtheta(h; \vtheta_{t-1}) - \vtheta_{t}}$ is $\mathcal{O}(h^p)$. Thus by expanding the Taylor series and the mean value theorem we obtain:
\begin{align}
 &\norm{\vtheta(h; \vtheta(h; \vtheta_{t-1})) - \vtheta(h; \vtheta_{t})} \\& = \norm{\sum_{i=0}^\infty \frac{h^i}{i!} \vtheta^{(i)}\left(\vtheta(h; \vtheta_{t-1})\right) -\sum_{i=0}^\infty \frac{h^i}{i!} \vtheta^{(i)}(\vtheta_t))} \\
  & = \norm{\sum_{i=0}^\infty \frac{h^i}{i!} \left(\vtheta^{(i)}\left(\vtheta(h; \vtheta_{t-1})\right) - \vtheta^{(i)}(\vtheta_t))\right)} \\
  &= \norm{\sum_{i=0}^\infty \frac{h^i}{i!} \frac{d}{d\vtheta}\vtheta^{(i)}(\vtheta') \left(\vtheta(h; \vtheta_{t-1}) - \vtheta_t\right)} \\
    &\le \sum_{i=0}^{\infty} \frac{h^i}{i!} \left|\frac{d}{d\vtheta}\vtheta^{(i)}(\vtheta')\right| \underbrace{\norm{(\vtheta(h; \vtheta_{t-1})) - \vtheta_t}}_{\mathcal{O}(h^p)} = \mathcal{O}(h^p)
\end{align}
This tells us that we can construct a bound:
\begin{align}
 \norm{\vtheta(2h; \vtheta_{t-1}) - \vtheta_{t+1}} \le \mathcal{O}(h^p) + \mathcal{O}(h^p) = \mathcal{O}(h^p)
\end{align}

\textbf{Dependence on the number of steps}.
While the order in learning rate is the same, 
we note that as the number of steps increases the errors are likely to accumulate with the number of steps (for $n$ discrete steps we will sum $n$ terms of order $\mathcal{O}(h^p)$ in the above bound). For example when taking the number of steps $n \rightarrow \infty$ the above no longer provides a bound.

\subsection{Approximations to per-iteration drift for gradient descent and momentum}
\label{sec:proofs_total_per_iteration_drift}

We now prove Thm~\ref{thm:total_drift}, on the per-iteration drift of gradient descent. We apply the Taylor reminder theorem on the NGF initialised at gradient descent iteration parameters $\vtheta_{t-1}$. We have that there exists $h'$ such that 
\begin{align}
\vtheta(h)   = & \vtheta(0) + h \dot{\vtheta}(0) + \frac{h^2}{2}\dot{\dot{\vtheta}}(h') \\
           =&  \vtheta_{t-1} - h \nabla_{\vtheta} E (\vtheta_{t-1}) + \frac{h^2}{2} \nabla_{\vtheta}^2 E(\vtheta') \nabla_{\vtheta} E(\vtheta')
\end{align}
with $h' \in (0, h)$. The above proof measures the drift of gradient descent, which we then used to construct DAL.

In Section~\ref{sec:future_work}, we also used DAL for momentum, and used an intuitive justification. We now show a more theoretical justification, by measuring the total per-iteration drift of momentum. 
To do so, we use the following flow to describe momentum, provided by~\citet{symmetry}
\begin{align}
\dot{\vtheta} = - \frac{1}{1-\beta}\nabla_{\vtheta} E,
\end{align}
and we measure the discretisation drift of a momentum update compared to this flow.
If we apply the same strategy as above, we have that 
\begin{align}
\vtheta(h) &= \vtheta(0) + h \dot{\vtheta}(0) + \frac{h^2}{2}\dot{\dot{\vtheta}}(h') \\
        &= \vtheta_{t-1} - \frac{h}{1-\beta} \nabla_{\vtheta} E (\vtheta_{t-1}) + \frac{h^2}{2 (1-\beta)^2}  \nabla_{\vtheta}^2 E(\vtheta') \nabla_{\vtheta} E(\vtheta')
\end{align}
 When we compare this with the discrete update 
\begin{align}
    \vtheta_t = \vtheta_{t-1} + \beta \vv_{t-1} - h \nabla_{\vtheta} E(\vtheta_{t-1})
\end{align}
and obtain that the per-iteration drift is
\begin{align}
& \left(\vtheta_{t-1} - \frac{h}{1-\beta} \nabla_{\vtheta} E (\vtheta_{t-1}) + \frac{h^2}{2 (1-\beta)^2}  \nabla_{\vtheta}^2 E(\vtheta') \nabla_{\vtheta} E(\vtheta') \right) \\ & \hspace{3em}- \left( \vtheta_{t-1} + \beta \vv_{t-1} - h \nabla_{\vtheta} E(\vtheta_{t-1})\right) \\
& =\frac{h^2}{2 (1-\beta)^2}  \nabla_{\vtheta}^2 E(\vtheta') \nabla_{\vtheta} E(\vtheta') + \beta \vv_{t-1} + h \nabla_{\vtheta} E(\vtheta_{t-1}) -  \frac{h}{1-\beta} \nabla_{\vtheta} E (\vtheta_{t-1}) \\
& =\frac{h^2}{2 (1-\beta)^2}  \nabla_{\vtheta}^2 E(\vtheta') \nabla_{\vtheta} E(\vtheta') + \beta \vv_{t-1} -\frac{h \beta}{1-\beta} \nabla_{\vtheta} E(\vtheta_{t-1})
\end{align}
 Thus there is one part of the drift that comes from the same source as gradient descent and another part of the drift which comes from the alignment of the current gradient and the moving average obtained using momentum. Thus when using DAL-$p$ with momentum, we only focus on one of the sources of drift (the first term).

\section{Comparison with a discrete-time approach}
\label{sec:discrete}

\subsection{Changes in loss function}
\label{sec:changes_in_loss_discrete}

We aim to obtain similar intuition to what we have obtained from the PF by discretising the NGF using Euler steps. We have:
\begin{align}
 E(\vtheta_{t+1}) &\approx  E(\vtheta_{t}) - h \nabla_{\vtheta} E(\vtheta_t)^T\nabla_{\vtheta} E(\vtheta_t) + \frac{h^2}{2} \nabla_{\vtheta} E(\vtheta_t)^T \nabla_{\vtheta}^2 E(\vtheta_t) \nabla_{\vtheta} E(\vtheta_t) \\
                 &\approx  E(\vtheta_{t}) - h \nabla_{\vtheta} E(\vtheta_t)^T\nabla_{\vtheta} E(\vtheta_t) + \frac{h^2}{2} \sum_i \lambda_i  \left(\nabla_{\vtheta} E(\vtheta_t)^T \vu_i \right)^2\\
                  &\approx  E(\vtheta_{t}) - h \sum_i \left(\nabla_{\vtheta} E(\vtheta_t)^T \vu_i \right)^2 + \frac{h^2}{2} \sum_i \lambda_i  \left(\nabla_{\vtheta} E(\vtheta_t)^T \vu_i \right)^2\\
                   &\approx  E(\vtheta_{t}) +  \sum_i (1 - h/(2\lambda_i))\left(\nabla_{\vtheta} E(\vtheta_t)^T \vu_i \right)^2
\end{align}
thus under this approximation loss to decrease between iterations one requires $1 - h/(2\lambda_i) \le 0$. This is consistent with the observations of the loss obtained using the PF in Eq~\eqref{eq:changes_in_e}.

\subsection{The dynamics of $\nabla_{\vtheta} E^T \vu_i$}
\label{sec:changes_in_dot_prod_discrete}

We now show to obtain the approximated dynamics of $\nabla_{\vtheta} E^T \vu_i$ obtained from the PF in Section~\ref{sec:instability_deep_learning} using a discrete-time approach.

We have that:
\begin{align}
\nabla_{\vtheta} E(\vtheta_{t+1}) &\approx \nabla_{\vtheta} E\left(\vtheta_t - h \nabla_{\vtheta} E(\vtheta_t)\right) \\
                        &\approx \nabla_{\vtheta} E(\vtheta_t) - h \nabla_{\vtheta}^2 E(\vtheta_t)\nabla_{\vtheta} E(\vtheta_t)
\end{align}
If we assume that the Hessian eigenvectors do not change between iterations, we have:
\begin{align}
\nabla_{\vtheta} E(\vtheta_{t+1}) ^T {(\vu_{i})}_{t+1} &\approx \nabla_{\vtheta} E(\vtheta_t)^T {(\vu_{i})}_{t+1}  - h \nabla_{\vtheta}^2 E(\vtheta_t)\nabla_{\vtheta} E(\vtheta_t) {(\vu_{i})}_{t+1}  \\&=   ( 1 - h \lambda_i )\nabla_{\vtheta} E(\vtheta_t)^T {(\vu_{i})}_{t+1} 
\end{align}
and obtain the same behaviour predicted in Section~\ref{sec:instability_deep_learning}.

\subsection{\rebuttalrthree{The connection between DAL and Taylor expansion optimal learning rate}}

\rebuttalrthree{In Section~\ref{sec:dal} introduced DAL as a way to control the drift of gradient descent; to do so, we set the learning rate of gradient descent to be inverse proportional to $\norm{\nabla_{\vtheta}^2 E\nabla_{\vtheta}E}/\norm{\nabla_{\vtheta}E} = \norm{\nabla_{\vtheta}^2 E \hat{\vg}}$}.

\rebuttalrthree{The learning rate set by DAL is similar, but distinct to that obtained by using a Taylor expansion of the loss $E$, as follows:
\begin{align}
E(\vtheta - h \nabla_{\vtheta}E) = E(\vtheta) - h \nabla_{\vtheta}E^T \nabla_{\vtheta}E + \frac{h^2}{2} \nabla_{\vtheta}E ^T \nabla_{\vtheta}^2 E \nabla_{\vtheta}E + \mathcal{O}(h^3)
\end{align}
}
\rebuttalrthree{
solving for the optimal $h$ in the above leads to:
\begin{align}
h = \frac{\nabla_{\vtheta}E^T \nabla_{\vtheta}E}{\nabla_{\vtheta}E ^T \nabla_{\vtheta}^2 E \nabla_{\vtheta}E} = \frac{1}{\hat{\vg}^T \nabla_{\vtheta}^2 E \hat{\vg}}
\end{align}
}
\rebuttalrthree{
While this learning rate contains similarities to DAL, since
 ${\hat{\vg}^T \nabla_{\vtheta}^2 E \hat{\vg} = \sum_i \lambda_i (\hat{\vg}^T \vu_i)^2}$ and $\norm{\nabla_{\vtheta}^2 E \hat{\vg}} = |{\sum_i \lambda_i \hat{\vg}^T \vu_i}|$,
}
\rebuttalrthree{
we also note a few significant differences:
\begin{itemize}
    \item Since DAL is derived from discretisation drift, we can use it in a variety of settings. We have shown in Section~\ref{sec:future_work} how DAL can be adapted to be used with momentum, where the main idea is to increase the contribution of local gradients to the momentum moving average when the drift is large, and decrease it otherwise. 
    \item While we focused on using one learning rate for all parameters in the main thesis, and thus use the norm operator in DAL, this is not necessary. Instead we can use the \textit{per-parameter drift} to construct per-parameter adaptive learning rates. We show preliminary results in Figure~\ref{fig:imagenet_lr_scaling_per_parameter}.
    \item The aim of DAL is to control the discretisation drift of gradient descent and show that this can affect both the stability and performance of gradient descent, through DAL-p. Beyond providing a training tool, DAL-p is another tool to use to understand gradient descent. Importantly, the ability to control the drift also enhances the generalisation capabilities of the method.
\end{itemize}
}

\section{Experimental details}

\textbf{Estimating continuous-time trajectories of flows.} To estimate the continuous-time trajectories we use Euler integration with a small learning rate $5 \times 10^{-5}$. We reached this learning rate through experimentation: further decreasing it did not change the obtained results. We note that this approach can be computationally expensive: to estimate the trajectory of the NGF of time corresponding to one gradient descent step with learning rate $10^{-2}$, we need to do 5000 gradient steps. It is common in the literature to use Runge--Kutta4 to approximate continuous-time flows, but we noticed that approximating a flow for time $h$ using Runge--Kutta4 with learning rate $h$ still introduced significant drift: if the learning rate was further reduced and multiple steps were taken the results were significantly different. Thus Runge--Kutta4 also needs multiple steps to estimate continuous trajectories. Given that one Runge--Kutta4 update requires computing 4 gradients, we found that using Euler integration with small learning rates is both sufficient and more efficient in practice.

\noindent \textbf{Datasets}. When training neural networks with primarily used three standard datasets: MNIST \citep{lecun1995learning}, CIFAR-10 \citep{cifar10} and Imagenet \citep{deng2009imagenet}. On the small NN example in Section \ref{sec:principal_flow}, we used a dataset of 5 examples, where the input is 2D sampled from a Gaussian distribution and the regression targets are also sample randomly. We used the UCI breast cancer dataset \citep{asuncion2007uci} in Figure \ref{fig:breast_cancer_principal_flow}.

\noindent \textbf{Architectures}. We use standard architectures: MLPs for MNIST, VGG \citep{simonyan2014very} or Resnet-18 (Version 1) \citep{he2016deep} for CIFAR-10, Resnet-50 for Imagenet (Version 1). We do not use any form of regularisation or early stopping. We use the Elu activation function \citep{elu} to ensure that the theoretical setting we discussed applies directly (we thus avoid discontinuities caused by ReLus \citep{agarap2018deep}). We note that in the CIFAR-10 experiments, we did not adapt the Resnet-18 architecture to the dataset; doing so will likely increase performance.

\noindent \textbf{Losses}. Unless otherwise specified we used a cross entropy loss.

\noindent \textbf{Computing eigenvalues and eigenvectors.} We use the Lanczos algorithm to compute $\lambda_i$ and $\vu_i$.

\noindent \textbf{Seeds}. All test accuracies are shown averaged from 3 seeds. For training curves, we compare models across individual training runs to be able to observe the behaviour of gradient descent. We did not observe variability across seeds with any of the behaviours reported in this thesis.

\noindent \textbf{DAL.} When using DAL we set a maximum learning rate of 5 to avoid any potential instabilities. We did not experiment with other values.

\noindent \textbf{Computing $\norm{\nabla_{\vtheta}^2 E \nabla_{\vtheta} E}$}. We use Hessian vector products to compute $\norm{\nabla_{\vtheta}^2 E \nabla_{\vtheta} E}$. We experimented with  using the approximation $\nabla_{\vtheta}^2 E \nabla_{\vtheta} E = \frac{1}{2}\nabla_{\vtheta} \norm{\nabla_{\vtheta} E}^2 \approx \frac{\nabla_{\vtheta}E(\vtheta + \epsilon \nabla_{\vtheta} E) - \nabla_{\vtheta}E(\vtheta)}{\epsilon}$ with $\epsilon = 0.01/\norm{\nabla_{\vtheta} E}$ as suggested by~\citet{geiping2021stochastic} and saw no decrease in performance in the full-batch setting when using it in DAL, but a slight decrease in performance when using it in stochastic gradient descent. We show experimental results in Figures~\ref{fig:dal_cifar_full_batch}, \ref{fig:dal_cifar_sgd},~\ref{fig:dal_approx_comp} and~\ref{fig:dal_0_5_approx_comp}. More experimentation is needed to see how to best leverage this approximation,  either by trying other values of $\epsilon$ or using larger batches to approximate $\norm{\nabla_{\vtheta}^2 E \nabla_{\vtheta} E}$. We note however that all the conclusions we observed in the thesis regarding the relative performance of $p$ values of DAL still hold, with and without the approximation and that indeed we see less of a difference when lower values of $p$ are used.

\begin{figure}[tbh!]
\begin{subfloat}[DAL-1]{
   \includegraphics[width=0.333\columnwidth]{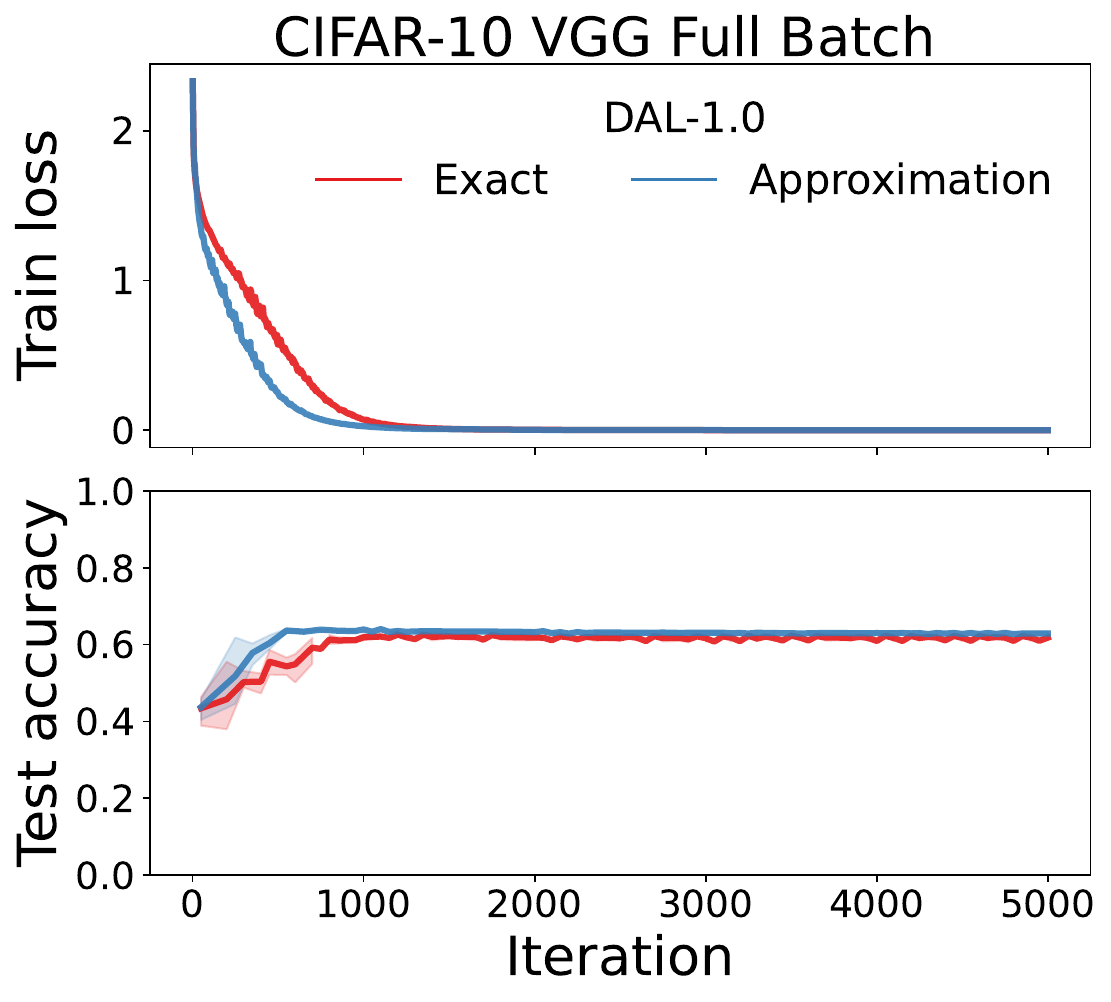}%
    } \end{subfloat}%
\begin{subfloat}[DAL-0.5]{
 \includegraphics[width=0.333\columnwidth]{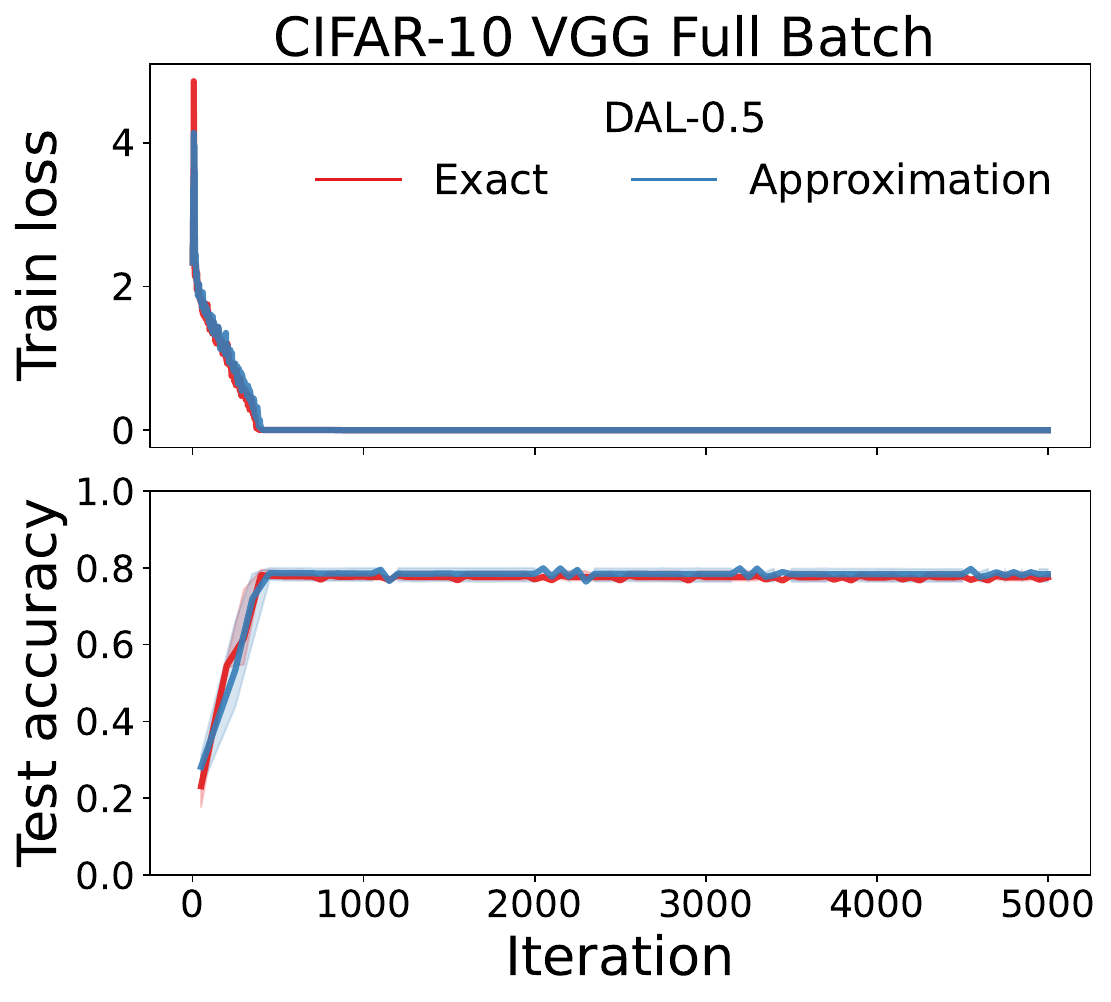}%
    } \end{subfloat}%
\begin{subfloat}[DAL-0.25]{
 \includegraphics[width=0.333\columnwidth]{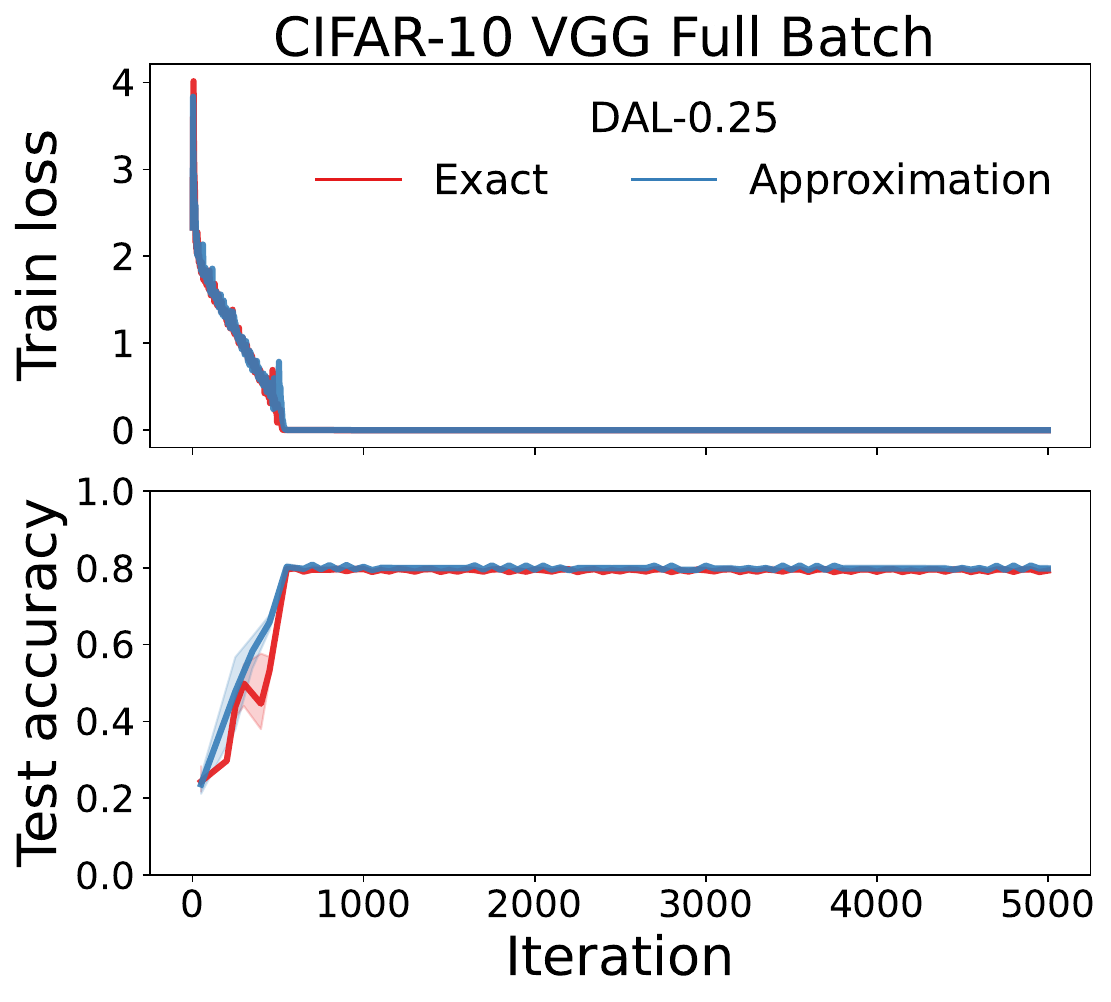}%
    } \end{subfloat}%
\caption[The effects of the approximation of $\nabla_{\vtheta}^2 E \nabla_{\vtheta} E$ on DAL; results on full-batch training on CIFAR-10.]{CIFAR-10 DAL results with a Hessian vector product computation of $\nabla_{\vtheta}^2 E \nabla_{\vtheta} E$ compared to an approximation. In this case, we observe no difference in performance between the two approaches.}
\label{fig:dal_cifar_full_batch}
\end{figure}

\begin{figure}[th!]
\begin{subfloat}[DAL-1]{
 \includegraphics[width=0.33\columnwidth]{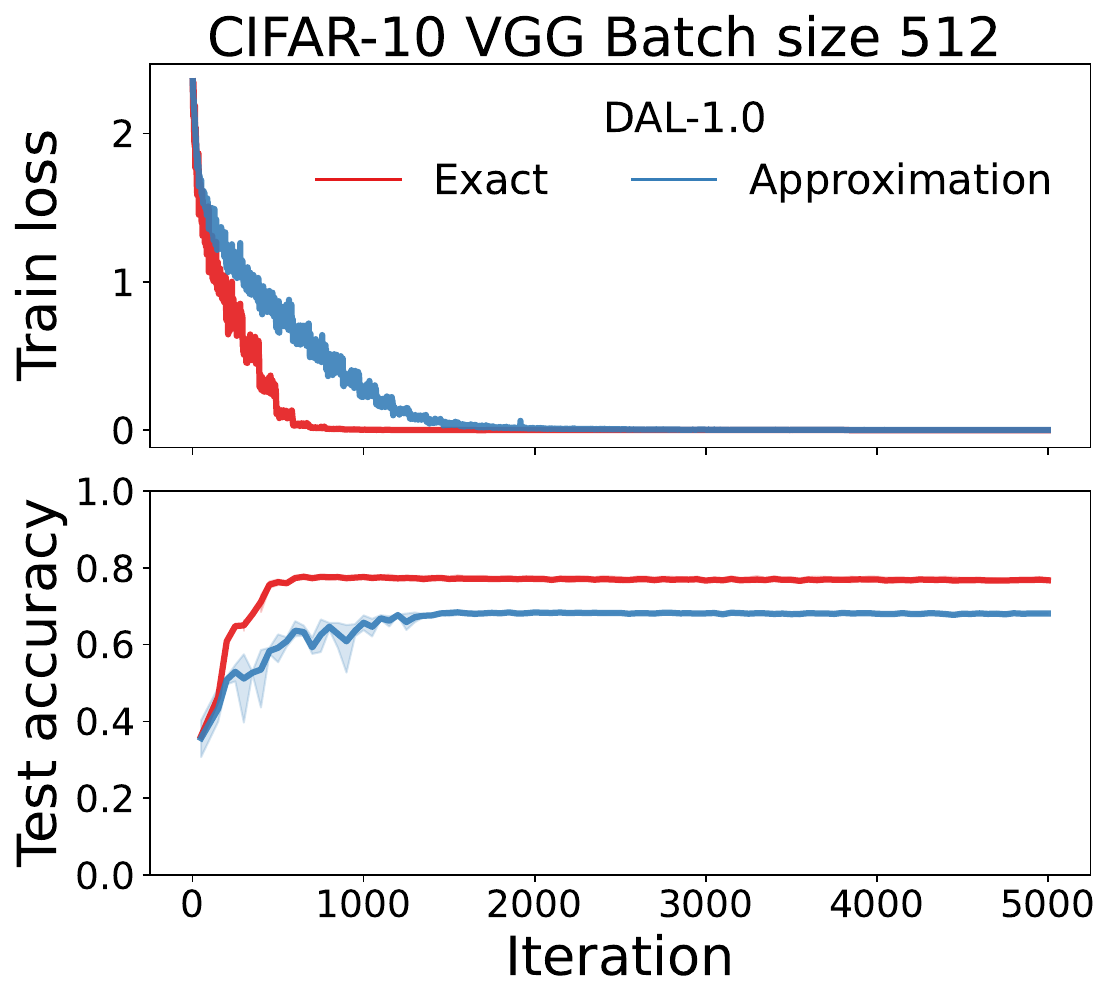}%
    } \end{subfloat}%
\begin{subfloat}[DAL-0.5]{
 \includegraphics[width=0.33\columnwidth]{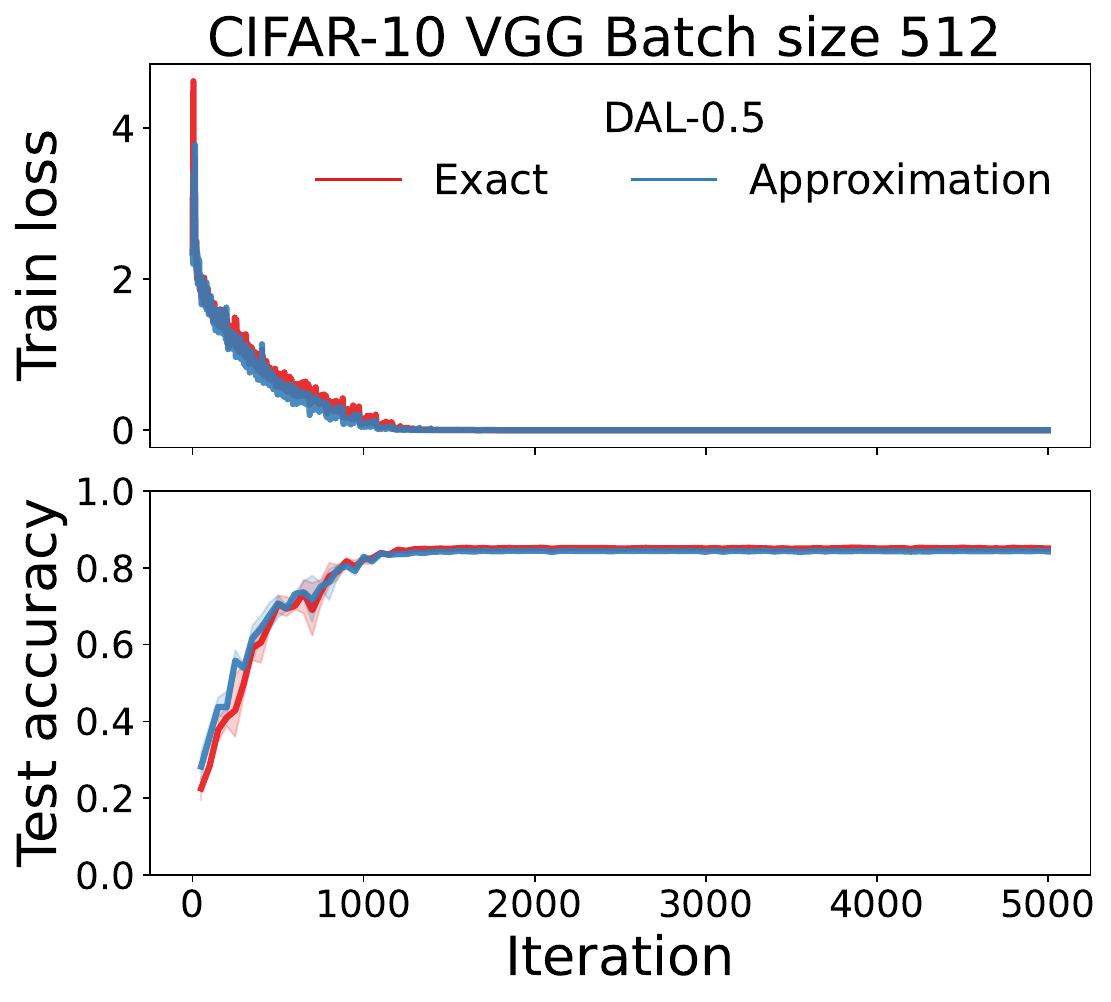}%
    } \end{subfloat}%
\begin{subfloat}[DAL-0.25]{
 \includegraphics[width=0.33\columnwidth]{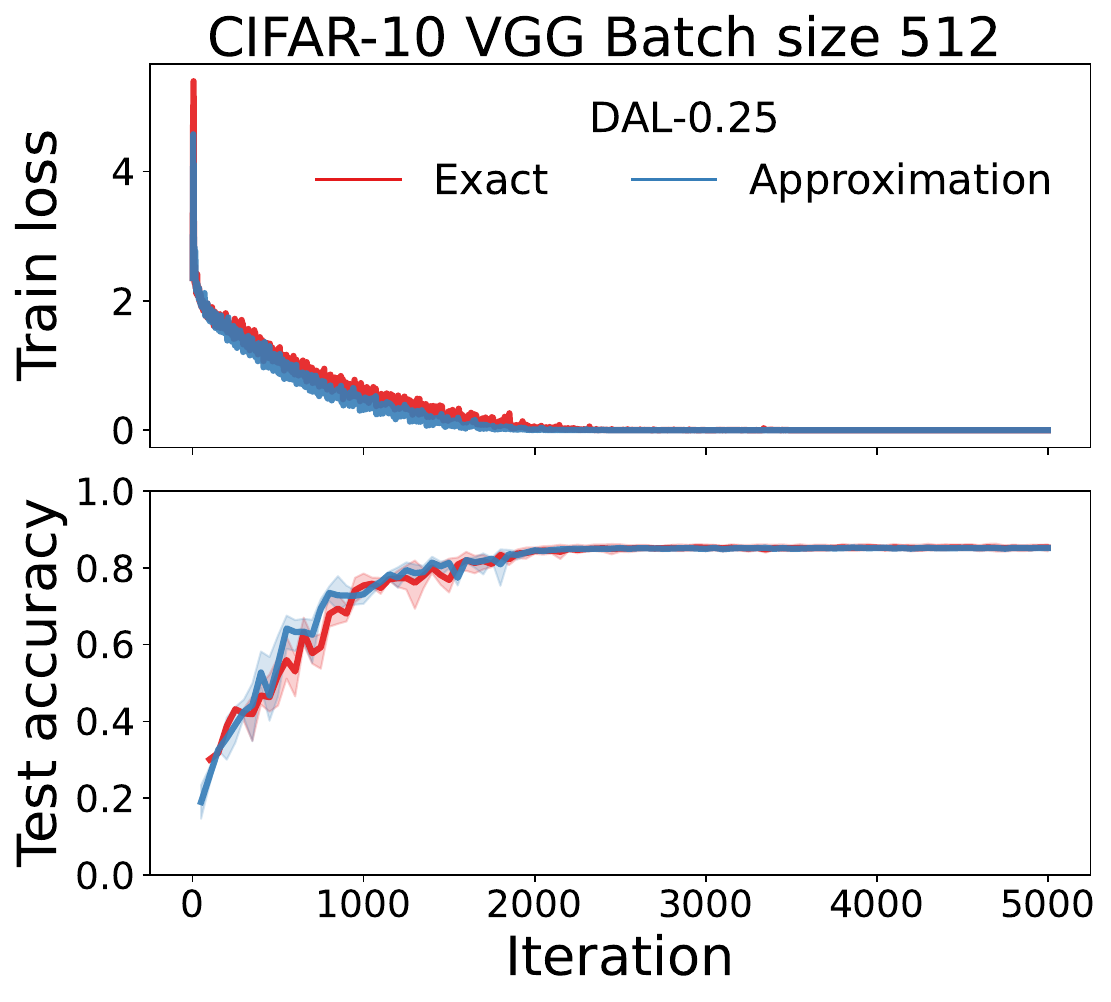}%
    } \end{subfloat}%
\caption[The effects of the approximation of $\nabla_{\vtheta}^2 E \nabla_{\vtheta} E$ on DAL; results with batch size 512 on CIFAR-10.]{CIFAR-10 DAL results with a Hessian vector product computation of $\nabla_{\vtheta}^2 E \nabla_{\vtheta} E$ compared to an approximation. For DAL-1, using the approximation leads to a decrease in test accuracy, but the same is not observed for DAL-$0.5$ and DAL-$0.25$.}
\label{fig:dal_cifar_sgd}
\end{figure}

\begin{figure}[th!]
\begin{subfloat}[Batch size 1024.]{
 \includegraphics[width=0.333\columnwidth]{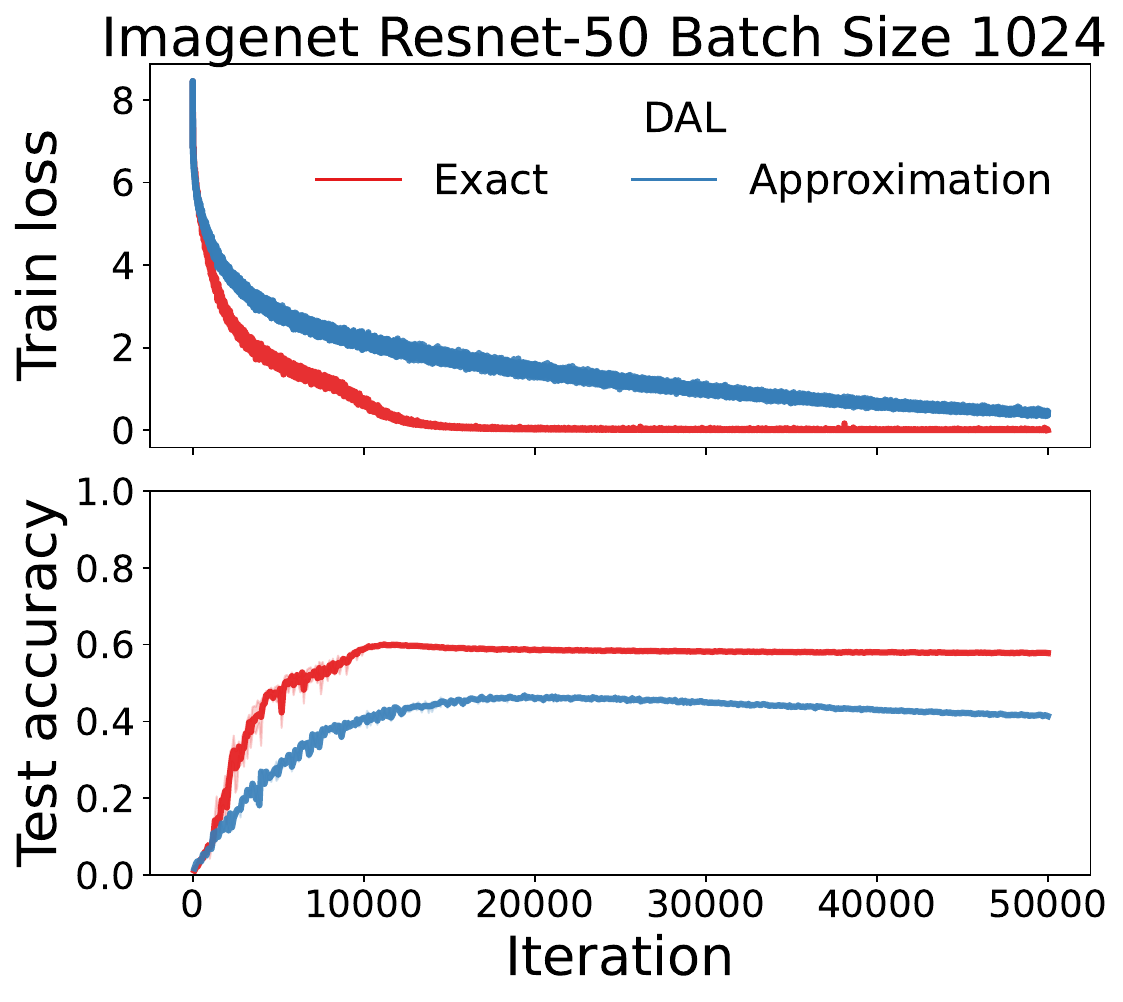}%
    } \end{subfloat}%
\begin{subfloat}[Batch size 2048.]{
 \includegraphics[width=0.333\columnwidth]{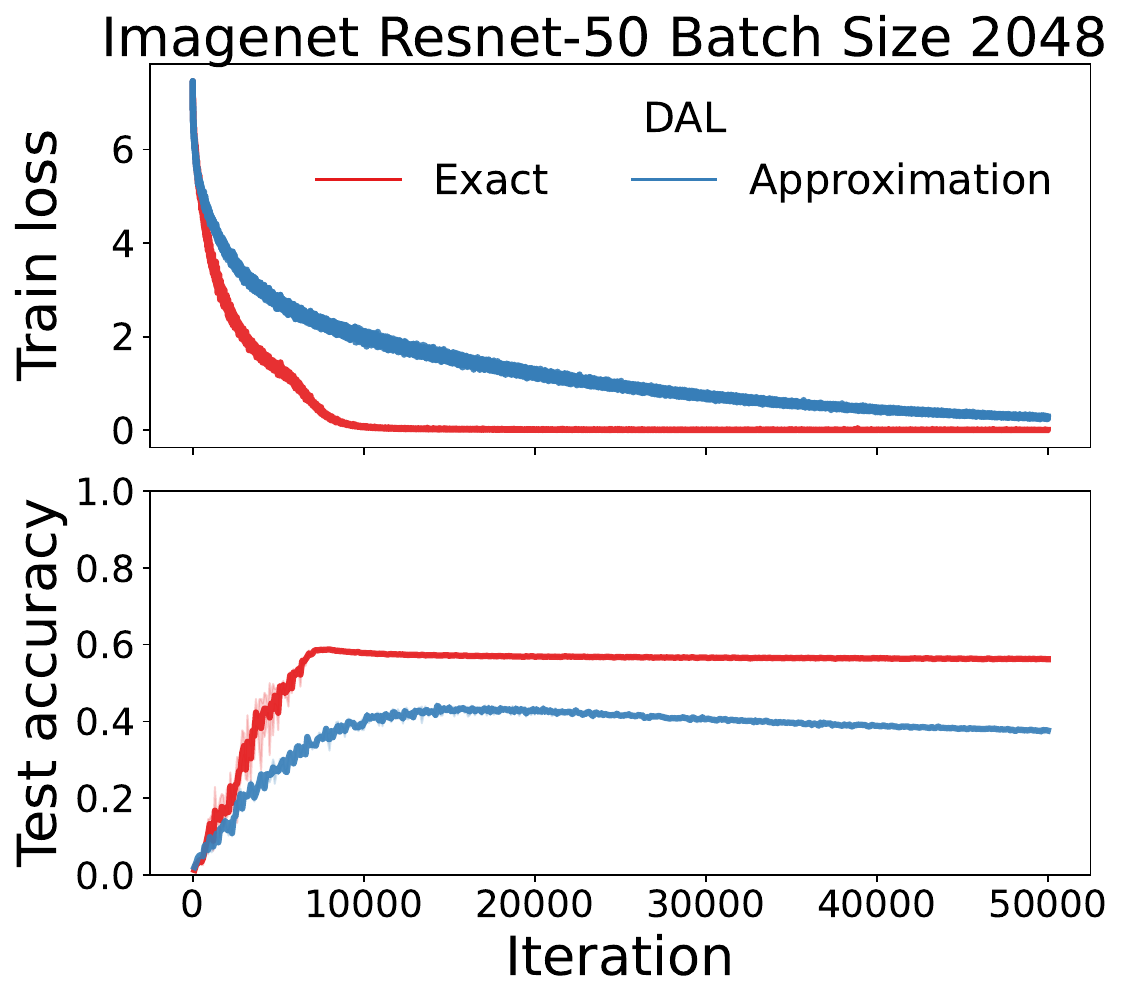}%
    } \end{subfloat}%
\begin{subfloat}[Batch size 4096.]{
 \includegraphics[width=0.333\columnwidth]{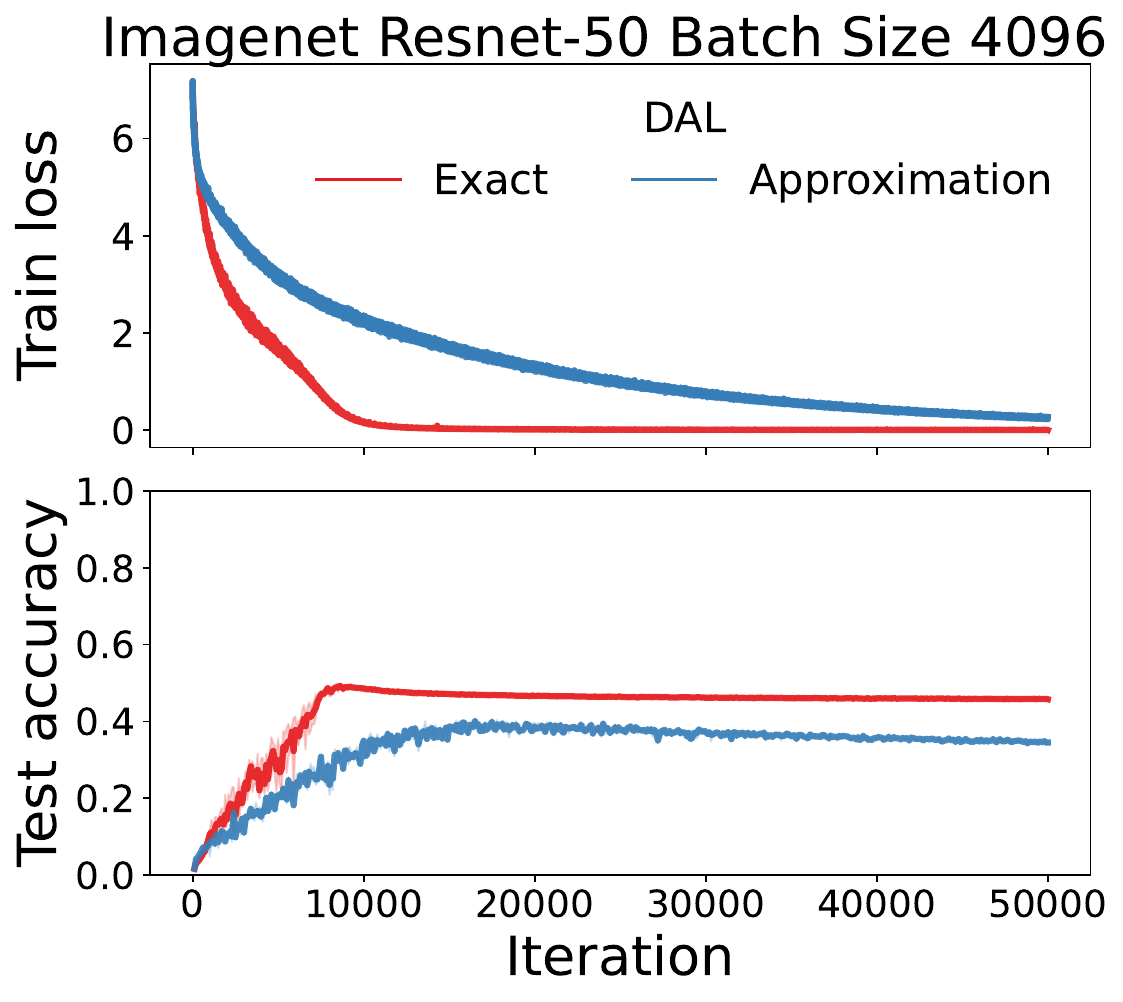}%
    } \end{subfloat}%
\caption[The effects of the approximation of $\nabla_{\vtheta}^2 E \nabla_{\vtheta} E$ on DAL; Imagenet results. ]{Imagenet DAL results with a Hessian vector product computation of $\nabla_{\vtheta}^2 E \nabla_{\vtheta} E$ compared to an approximation. Using the approximation results into a significant decrease in performance; we suspect that this can be resolved by using another value for $\epsilon$.}
\label{fig:dal_approx_comp}
\end{figure}

\begin{figure}[th!]
\begin{subfloat}[Batch size 1024.]{
 \includegraphics[width=0.333\columnwidth]{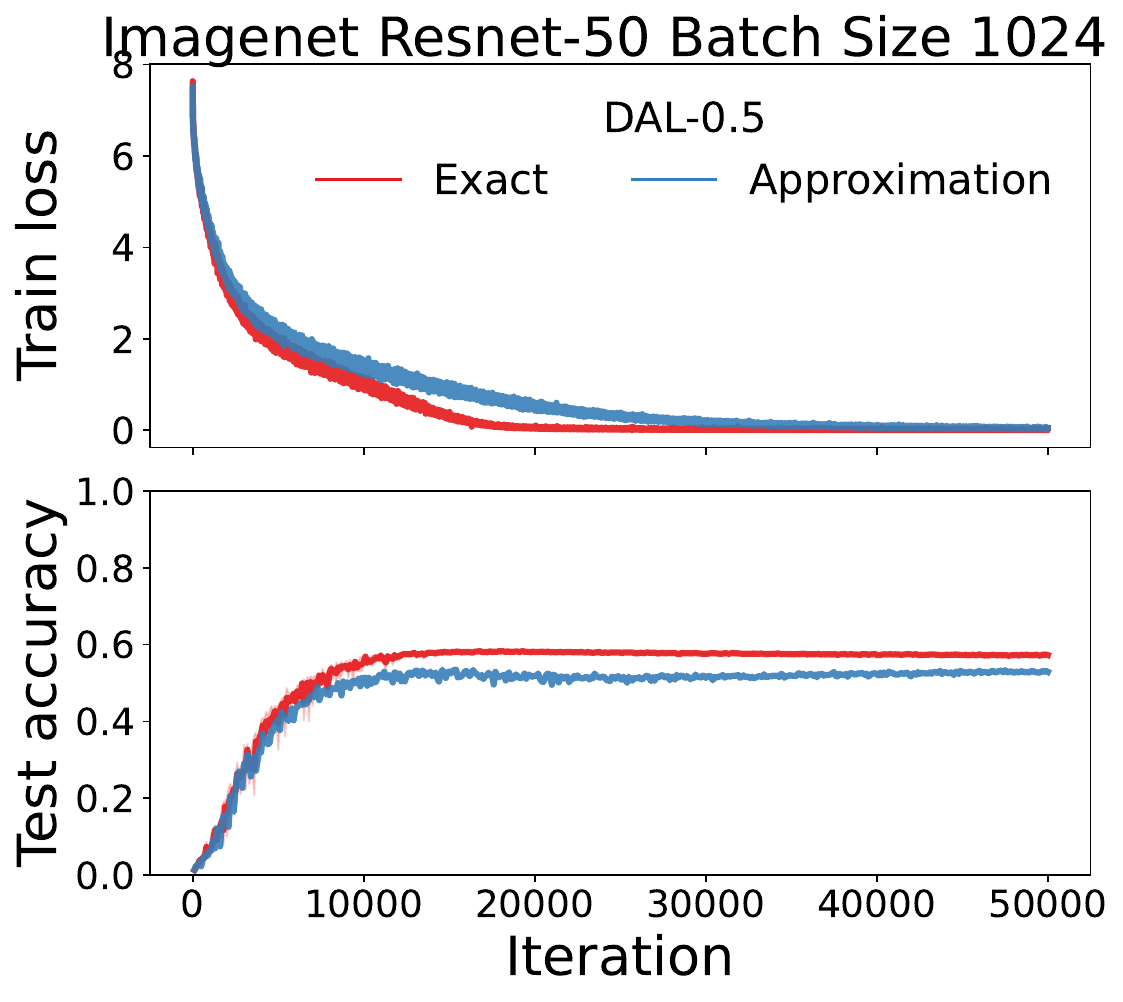}%
    }\end{subfloat}%
\begin{subfloat}[Batch size 2048.]{
 \includegraphics[width=0.333\columnwidth]{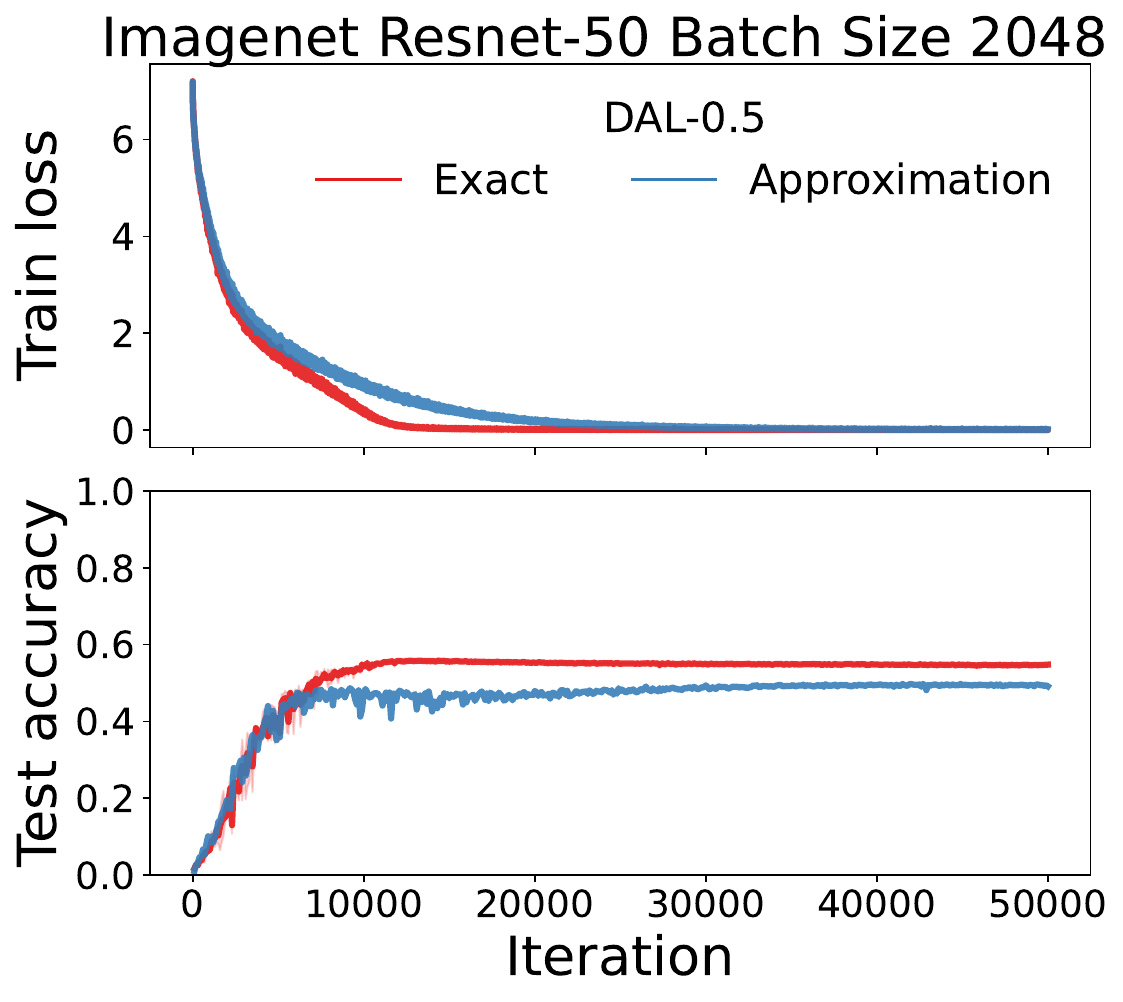}%
    } \end{subfloat}%
\begin{subfloat}[Batch size 4096.]{
 \includegraphics[width=0.333\columnwidth]{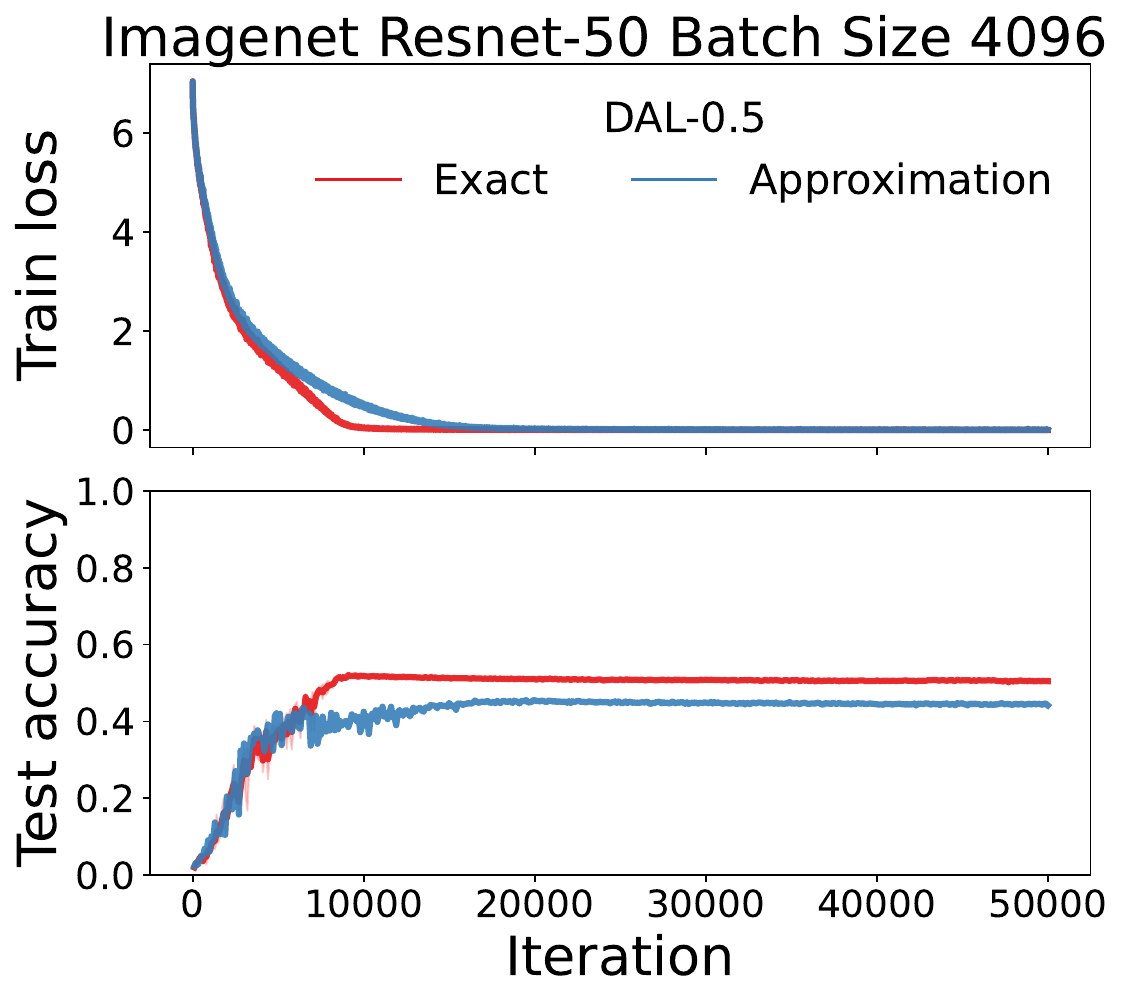}%
    } \end{subfloat}%
\caption[The effects of the approximation of $\nabla_{\vtheta}^2 E \nabla_{\vtheta} E$ on DAL-$0.5$; Imagenet results.]{Imagenet DAL$-0.5$ results with a Hessian vector product computation of $\nabla_{\vtheta}^2 E \nabla_{\vtheta} E$ compared to an approximation. Using the approximation results into a significant decrease in performance; we suspect that this can be resolved by using another value for $\epsilon$.}
\label{fig:dal_0_5_approx_comp}
\end{figure}

\newpage
\section{Additional experimental results}

\textbf{Oscillations for non linear functions}. 
We show that in the case of non-linear functions in Figure~\ref{fig:1d_cosine}, using $\cos$. Here too, the PF is better at describing the behaviour of GD than the NGF and IGR flow, which stop at local minima.

\begin{figure}[h]
\centering
\begin{subfloat}[Medium sized learning rate $h = 0.5$.]{
 \includegraphics[width=0.48\columnwidth]{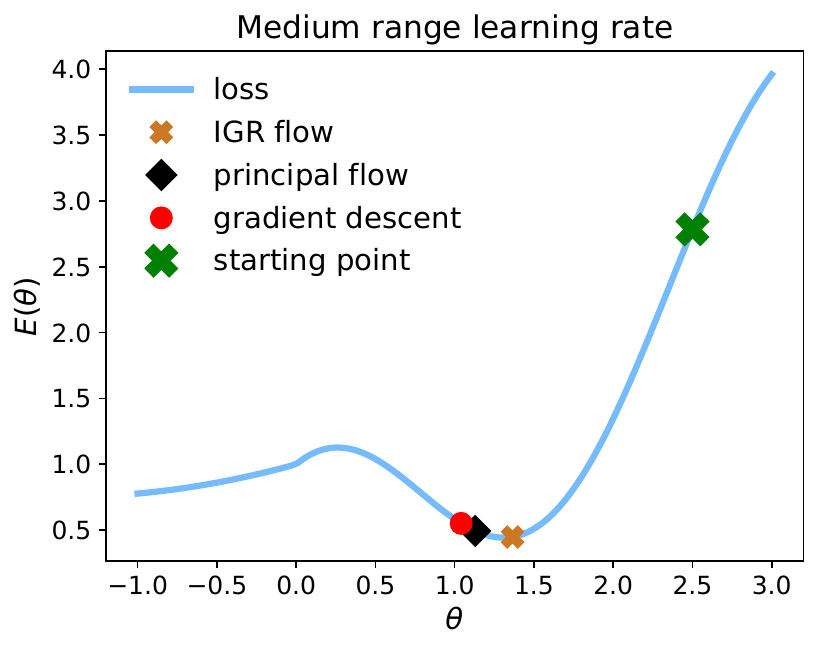}%
}\end{subfloat}%
\begin{subfloat}[Large learning rate $h = 0.85$.]{
 \includegraphics[width=0.48\columnwidth]{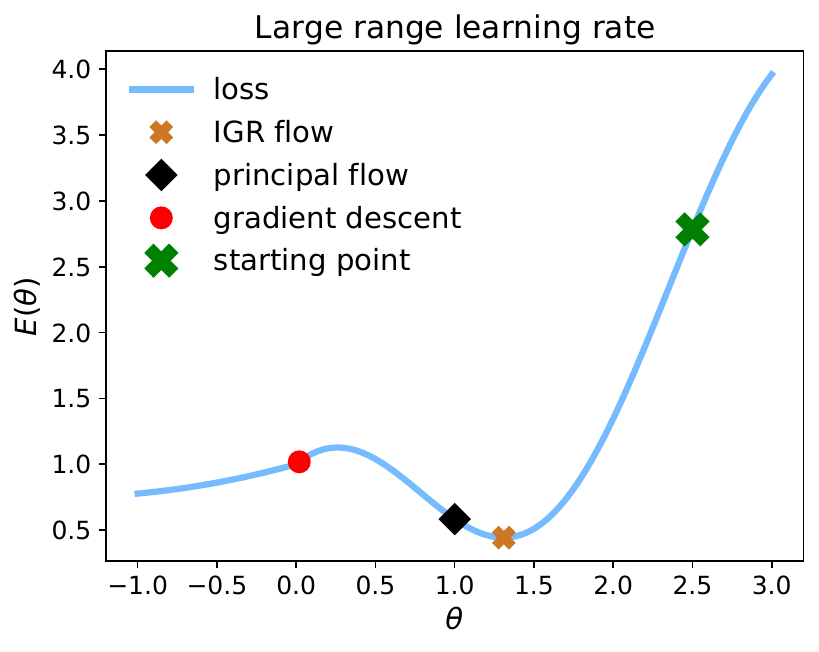}%
}\end{subfloat}%
\caption[Continuous time flows approximating gradient descent for a 1-D non quadratic function.]{Results with a non quadratic function, including $\cos$ and $\sin$. The function used is: $E(\theta) = \cos(\theta) + \theta, \text{if } \theta < 0; (\theta/3)^22 + 1 + \theta/3)$ otherwise. The principal flow and the IGR flow stay on the cosine branch, but the PF is able to exit the local minimum. The main reason for using this function was to show the escape of gradient descent into a flatter valley and to assess what the corresponding flows do.}
\label{fig:1d_cosine}
\end{figure}

\noindent \textbf{Global error for the UCI breast cancer dataset.}
We show in Figure~\ref{fig:breast_cancer_principal_flow} that the PF is better at tracking the behaviour of GD in the case of NNs both in the stability and edge of stability areas.

\begin{figure}
\centering
\begin{subfloat}[Stability.]{
 \includegraphics[width=0.45\columnwidth]{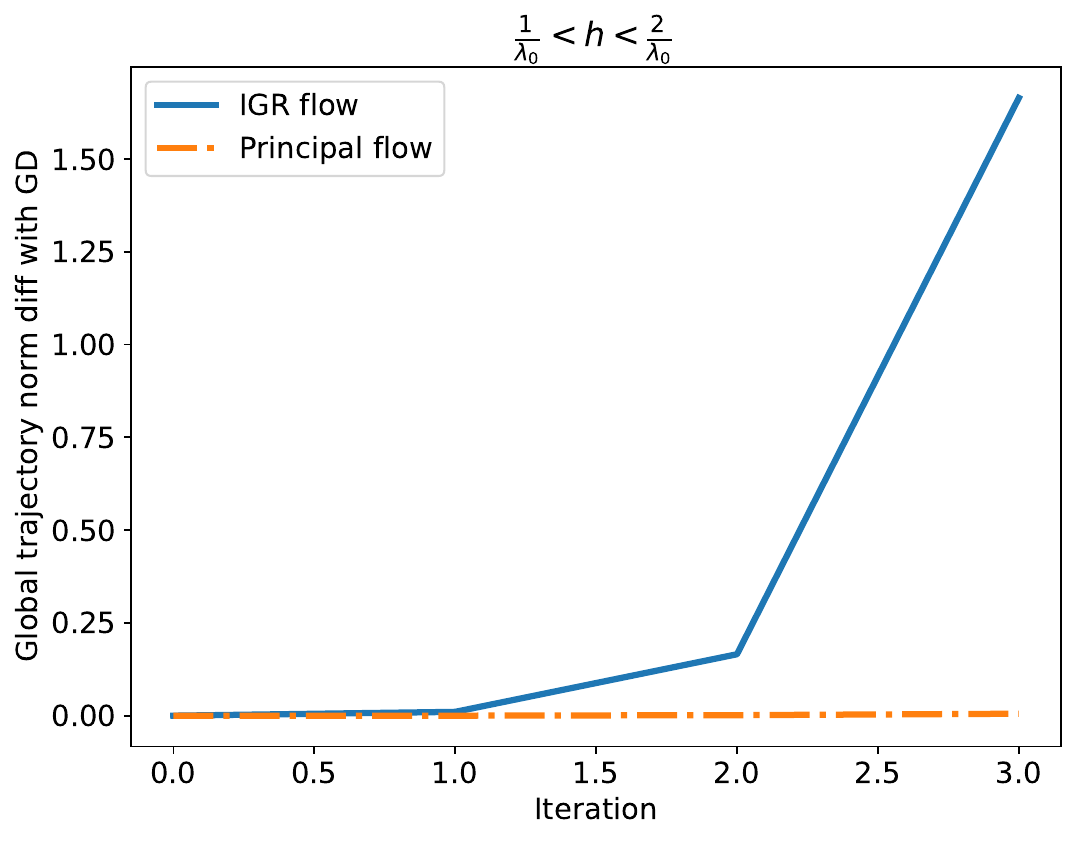}
}\end{subfloat}
\begin{subfloat}[Edge of stability.]{
 \includegraphics[width=0.45\columnwidth]{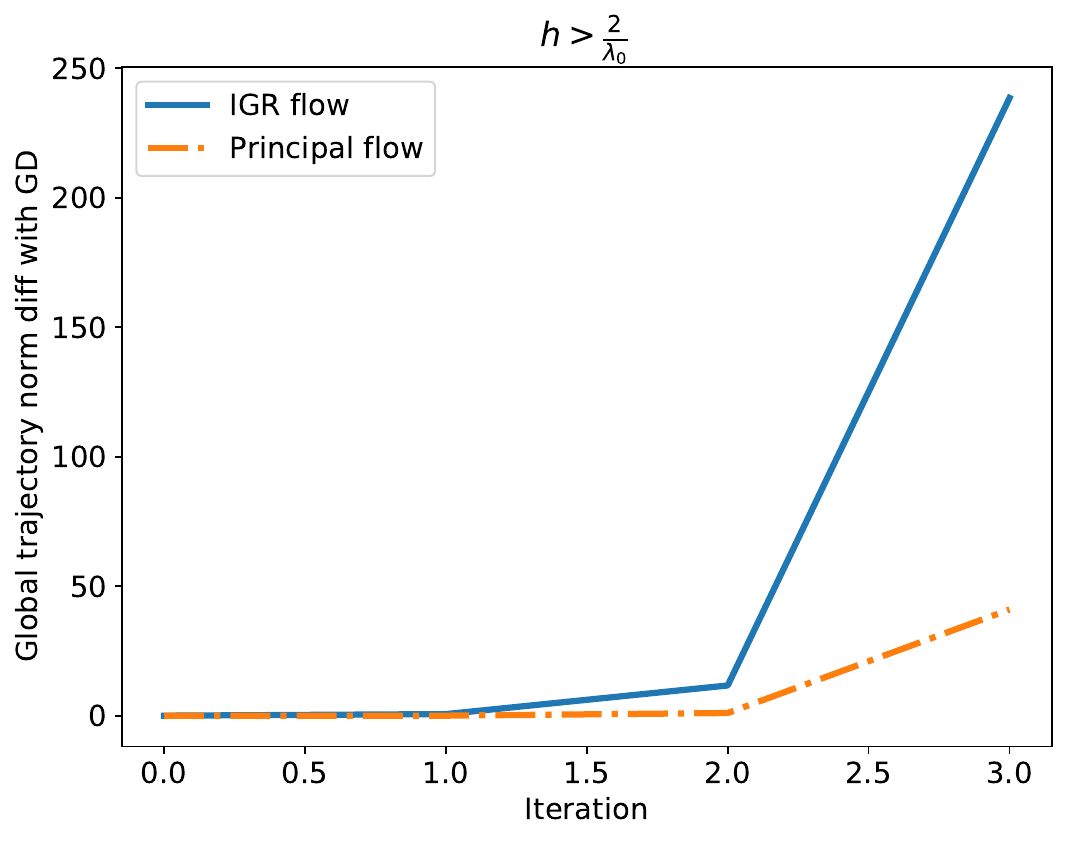}
}
\end{subfloat}
\caption[Global parameter error between gradient descent and continuous-time flows  on an MLP trained on the UCI breast cancer dataset.]{Global parameter error $\norm{\vtheta(nh) - \vtheta_n}$ for IGR ad PF flows  on an MLP trained on the UCI breast cancer dataset. We initialise both gradient descent and the IGR and PF flows at a set of initial parameters, and run them for time step $2h$ and compare the error after each time step $h$. The PF is better at tracking the trajectory of GD than the IGR flow by a significant margin.}
\label{fig:breast_cancer_principal_flow}
\end{figure}

\noindent \textbf{Additional edge of stability experiments}.
We show that one dimension is sufficient to cause instability in training on MNIST in Figure \ref{fig:mnist_continuous_time_swap} (as we have done in Figure \ref{fig:instabilities_change_learning_rate_loss} for CIFAR-10). 
We show the connection between the stability coefficient $sc_0$ and areas where the loss increases in Figure \ref{fig:cifar_peak_commons}: this shows that when the loss increases, it is in an area where $sc_0$ is also increasing. We note that to get a full picture of the behaviour of the PF we would need to use continuous-time computation, but since that is computationally prohibitive we use the incomplete but easily available discrete data. Figure~\ref{fig:edge_of_stability_results_lambda} zooms in the behaviour of $\lambda_0$ as dependent of the behaviour of the corresponding flows on MNIST and CIFAR-10 and Figure~\ref{fig:edge_of_stability_results_cifar_resnet} shows additional Resnet-18 results.

\begin{figure}
\centering
\begin{subfloat}[Iteration 200.]{
 \includegraphics[width=0.33\columnwidth]{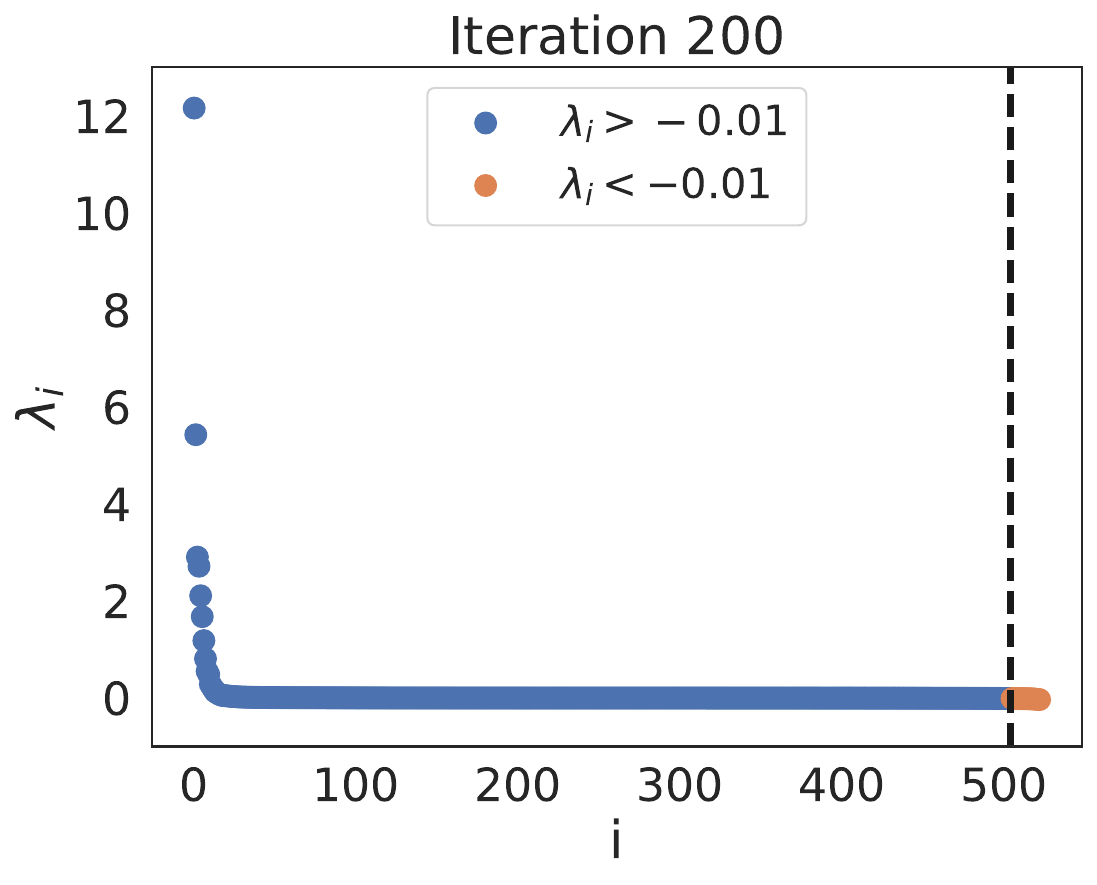}%
}\end{subfloat}%
\begin{subfloat}[Iteration 225.]{
 \includegraphics[width=0.33\columnwidth]{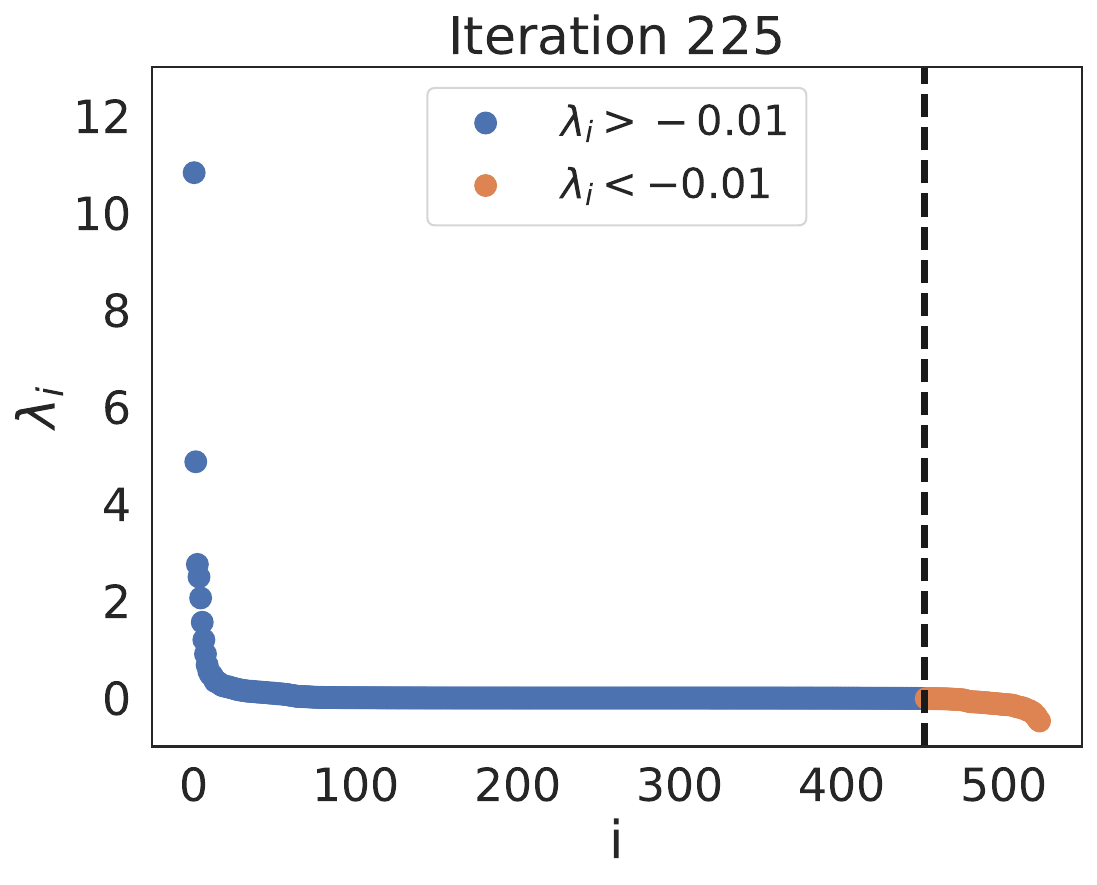}%
}\end{subfloat}%
\begin{subfloat}[Iteration 250.]{
 \includegraphics[width=0.33\columnwidth]{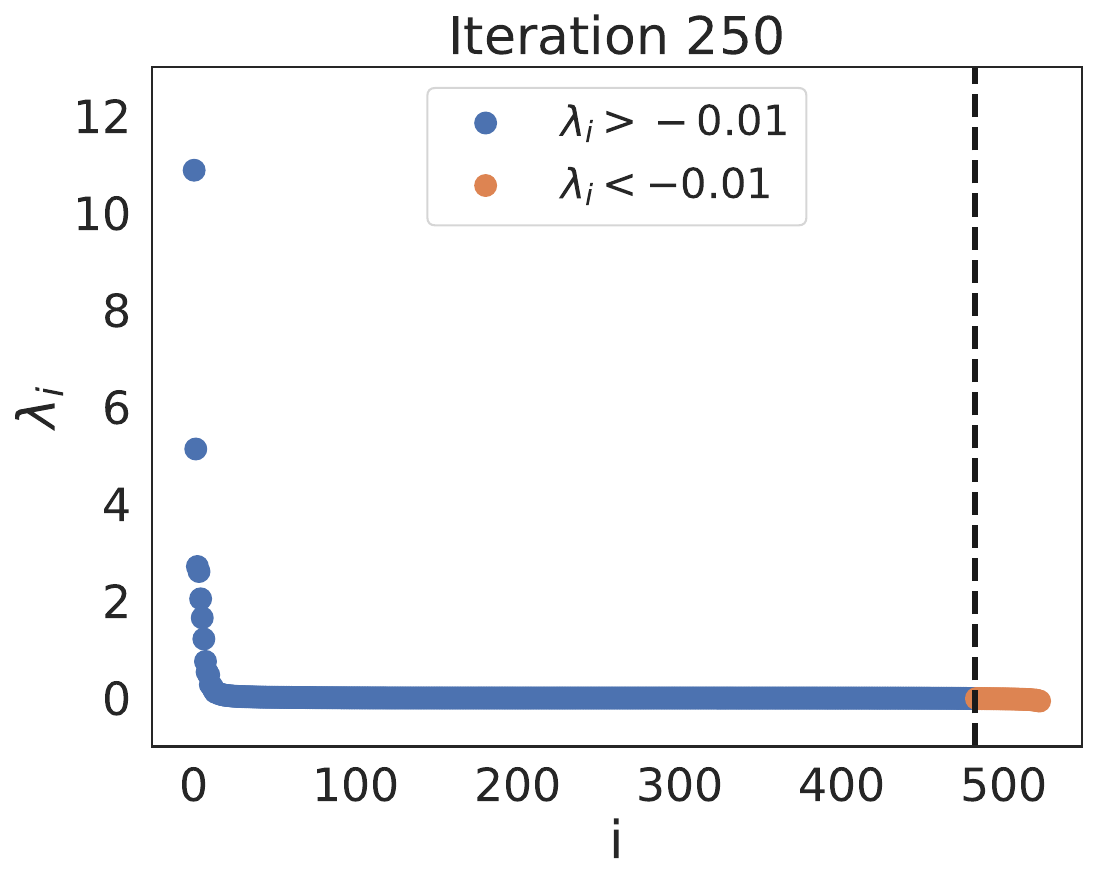}%
}\end{subfloat}%
\caption[The eigenspectrum obtained along the gradient descent path of a 5 layer small MLP trained on the UCI Iris dataset.]{The eigenspectrum obtained along the gradient descent path of a 5 layer small MLP trained on the UCI Iris dataset. The Hessian eigenvalues are largely positive. The result accompanies Figure~\ref{fig:model_of_gd_at_edge_of_stability} in the main thesis.}
\label{fig:eigspectrum_small_mlp}
\end{figure}

\begin{figure}[tbh]
\centering
\begin{subfloat}[Training loss.]{
  \includegraphics[width=0.45\columnwidth]{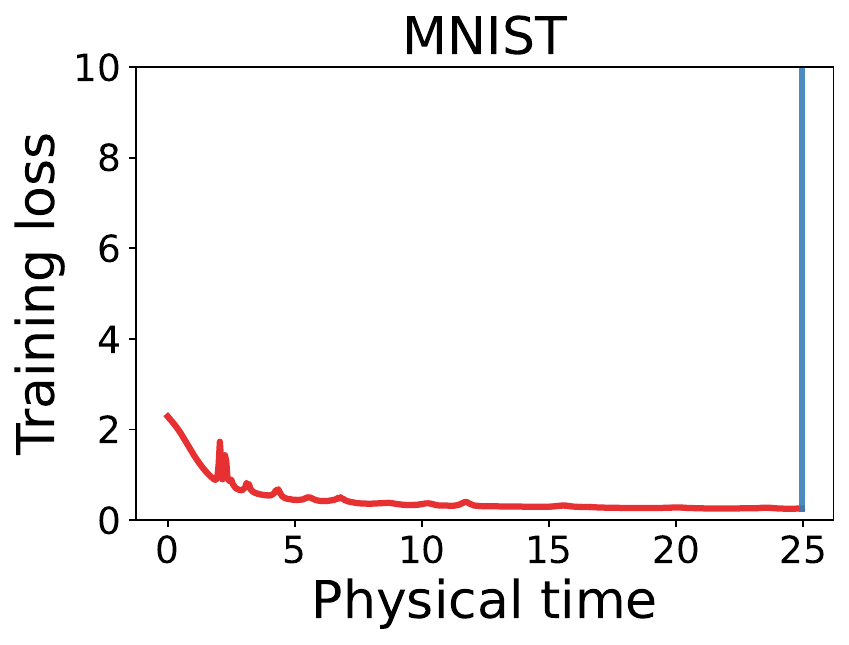}%
  \label{fig:one_dim_sign_swap_loss_mnist}
}\end{subfloat}%
\begin{subfloat}[$\lambda_0$.]{
  \includegraphics[width=0.45\columnwidth]{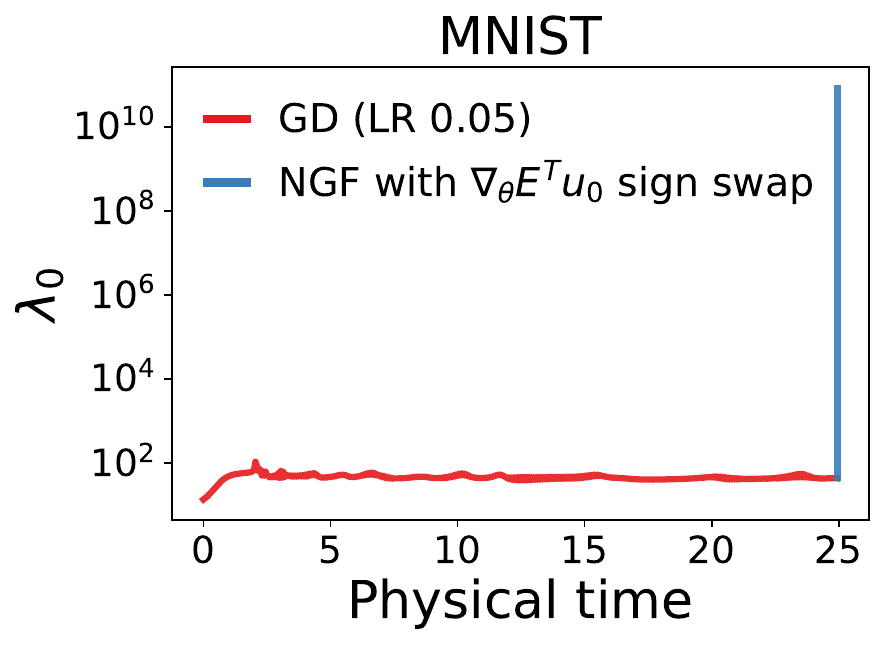}
  \label{fig:one_dim_sign_swap_lambda_mnist}
}\end{subfloat}%
\caption[One eigendirection is sufficient to cause instability in continuous-time; MNIST results.]{One eigendirection is sufficient to lead to instabilities. To create a situation similar to that of the PF in neural network training, we construct a flow given by the NGF in all eigendirections but $\vu_0$; in the direction of $\vu_0$, we change the sign of the flow's vector field. This leads to the flow ${\dot{\vtheta} = \left(\nabla_{\vtheta}E^T \vu_0\right) \vu_0 + \sum_{i=1}^{D-1} -\left(\nabla_{\vtheta}E^T \vu_i\right) \vu_i}$. We show this flow can be very unstable when initialised in an edge of stability area, with increases in loss \subref{fig:one_dim_sign_swap_loss_mnist} and $\lambda_0$ \subref{fig:one_dim_sign_swap_lambda_mnist}. MNIST results when changing one direction in continuous-time: one eigendirection is sufficient to cause instability. The model was trained with learning rate $0.05$ until the edge of stability area is reached after which the flow ${\dot{\vtheta} = \left(\nabla_{\vtheta}E^T \vu_0\right) \vu_0 + \sum_{i=1}^{D-1} -\left(\nabla_{\vtheta}E^T \vu_i\right) \vu_i}$ is used.}
\label{fig:mnist_continuous_time_swap}
\end{figure}

\begin{figure}[tbh]
\centering
\begin{subfloat}[Proportion of loss function local peaks \\ that are also stability coefficient local peaks.]{
  \includegraphics[width=0.495\columnwidth]{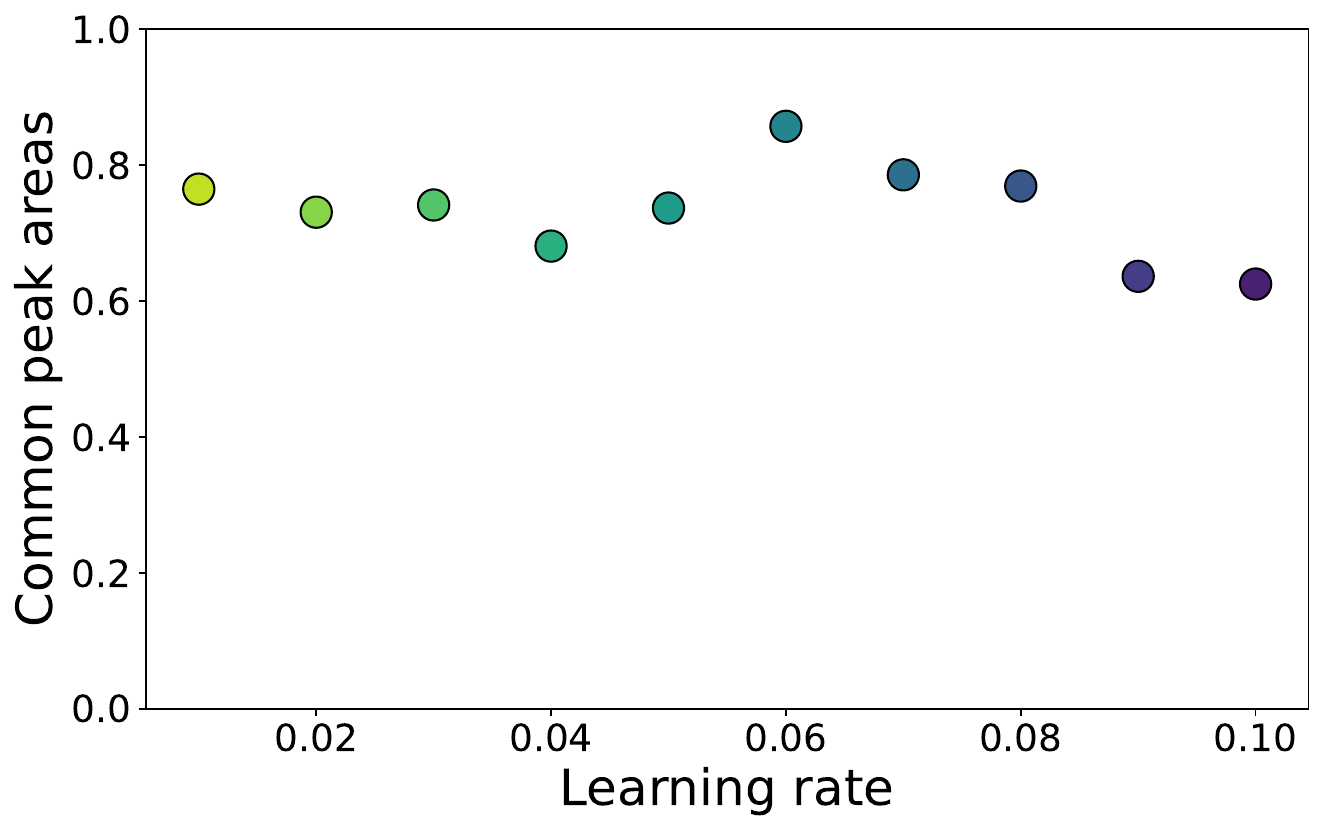}%
  \label{fig:common_peaks_prop}
} \end{subfloat}%
\begin{subfloat}[Delta in peak areas of the loss and the stability coefficient counted in iterations.]{
  \includegraphics[width=0.495\columnwidth]{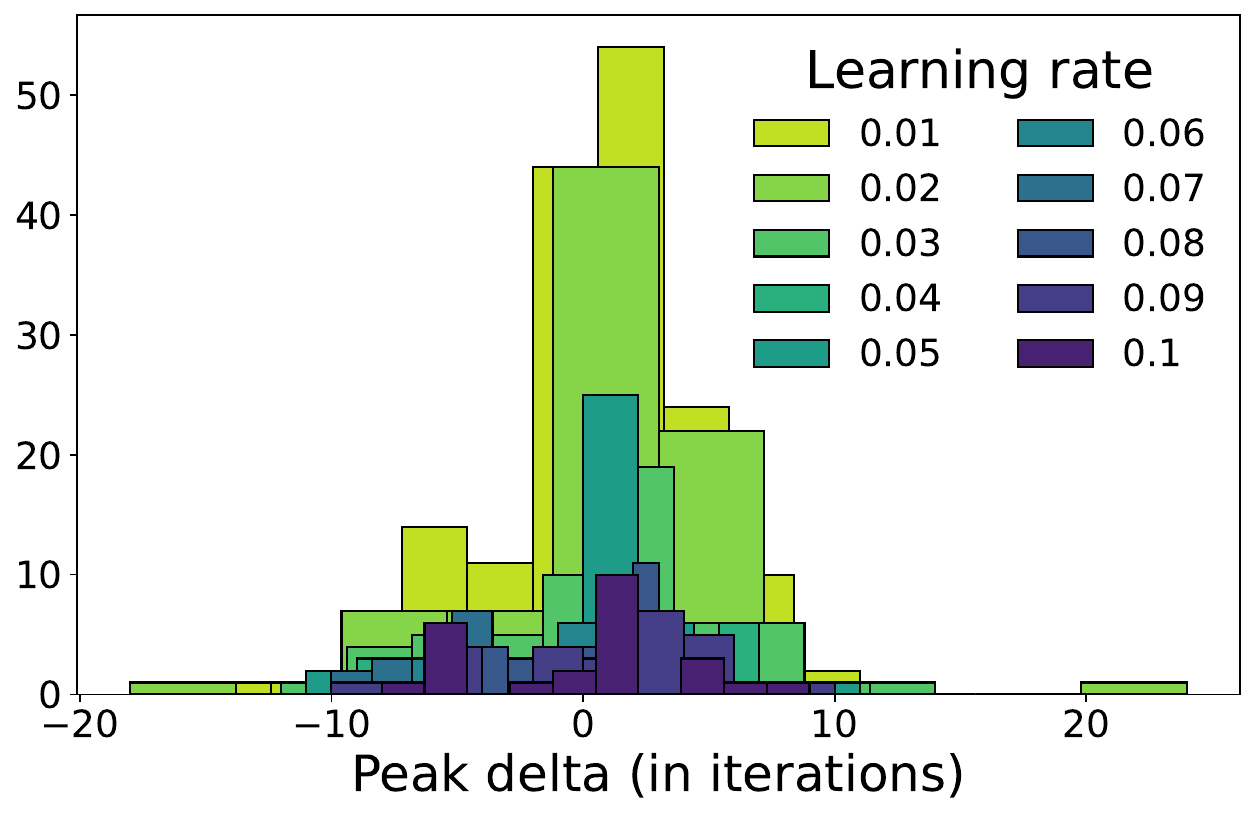}%
  \label{fig:common_peaks_delta}
 }\end{subfloat}%
\caption[Peak areas of loss increase and the stability coefficient for the largest eigendirection of the PF.]{Peak areas of loss increase and the stability coefficient for the largest eigendirection of the PF. We find peaks in our training signals via \lstinline{scipy.signal.find_peaks}. Here, we normalise the coefficient by the gradient norm to remove the gradient norm as a potential confounder. Results on CIFAR-10. \subref{fig:common_peaks_prop} shows that most loss function peaks are 5 iterations away from a local stability coefficient peak. \subref{fig:common_peaks_delta} shows the delta in iterations between a loss function peak and the closes stability coefficient peak. }
\label{fig:cifar_peak_commons}
\end{figure}

\begin{figure}[tbh]
\centering
\begin{subfloat}[MNIST MLP.]{
 \includegraphics[width=0.48\columnwidth]{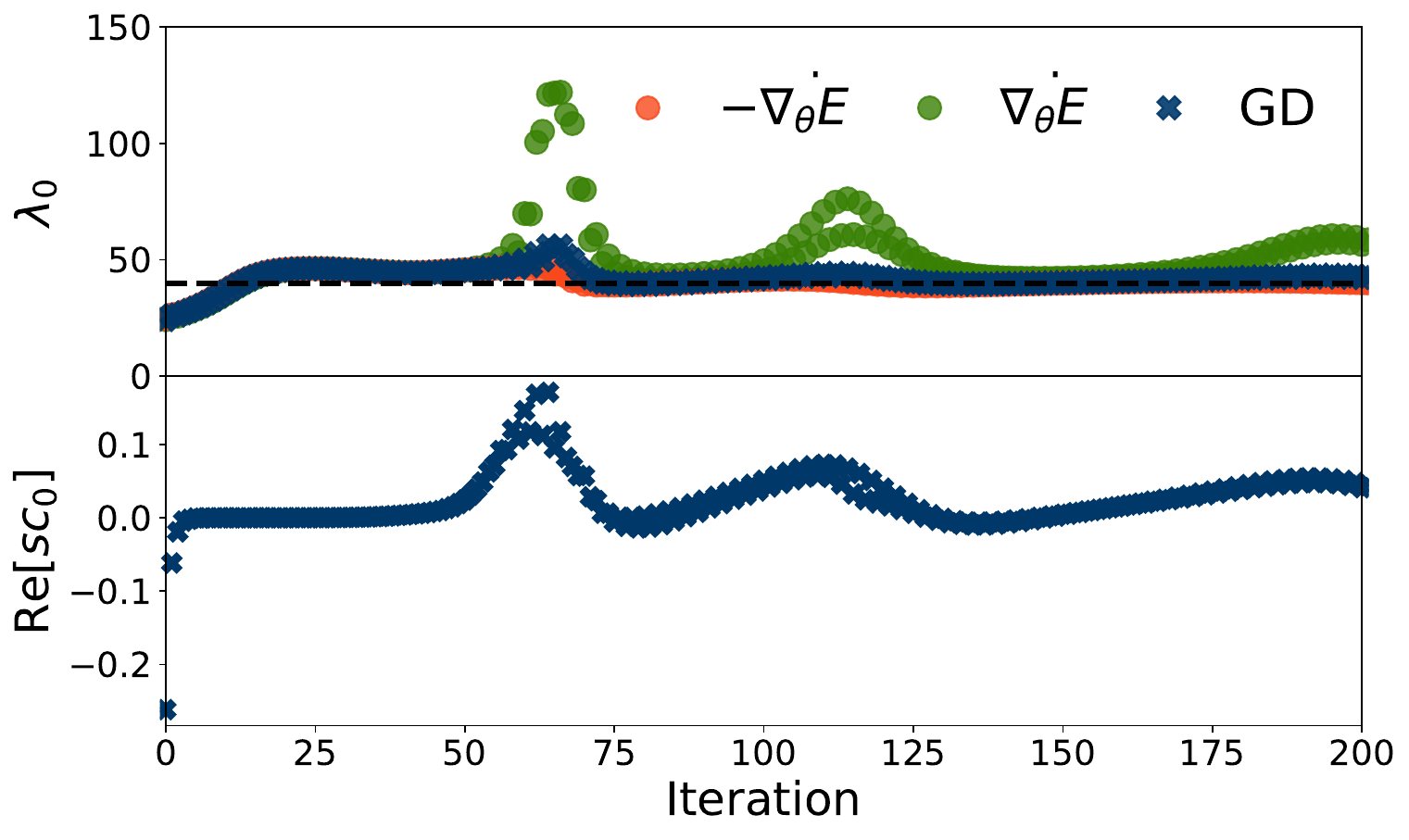}%
 }
 \end{subfloat}
\begin{subfloat}[CIFAR-10 VGG.]{
 \includegraphics[width=0.48\columnwidth]{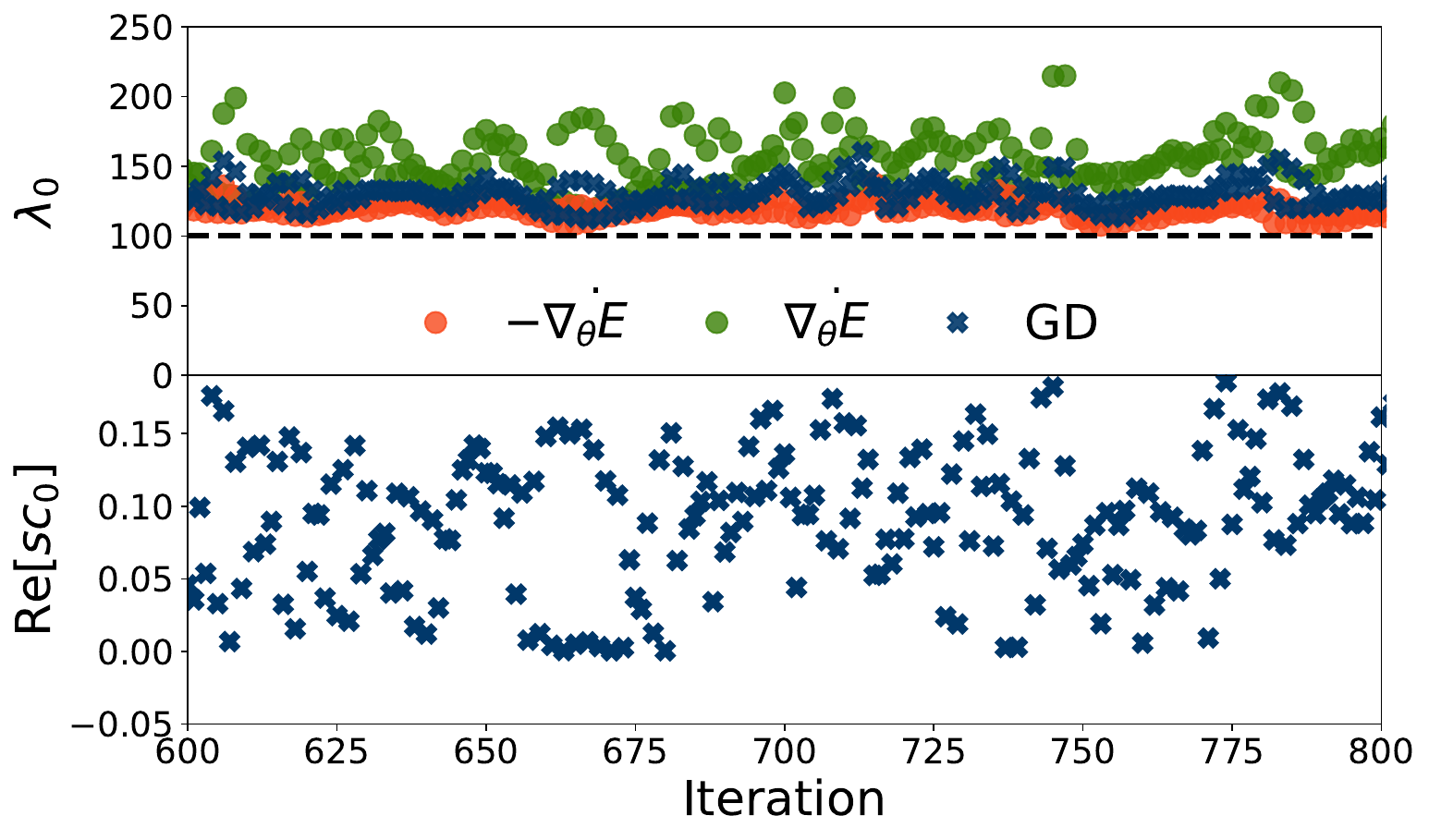}%
 }
 \end{subfloat}
\caption[Understanding changes to $\lambda_0$ using the PF.]{Understanding changes to $\lambda_0$ using the PF: increases in $\lambda_0$ above $2/h$ correspond to increases in the flow $\dot{\vtheta} = \nabla_{\vtheta} E$. The stability coefficient reflects $sc_0$ the changes in $\lambda_0$.}
\label{fig:edge_of_stability_results_lambda}
\end{figure}

\begin{figure}[tbh]
\centering
  \includegraphics[width=0.9\columnwidth]{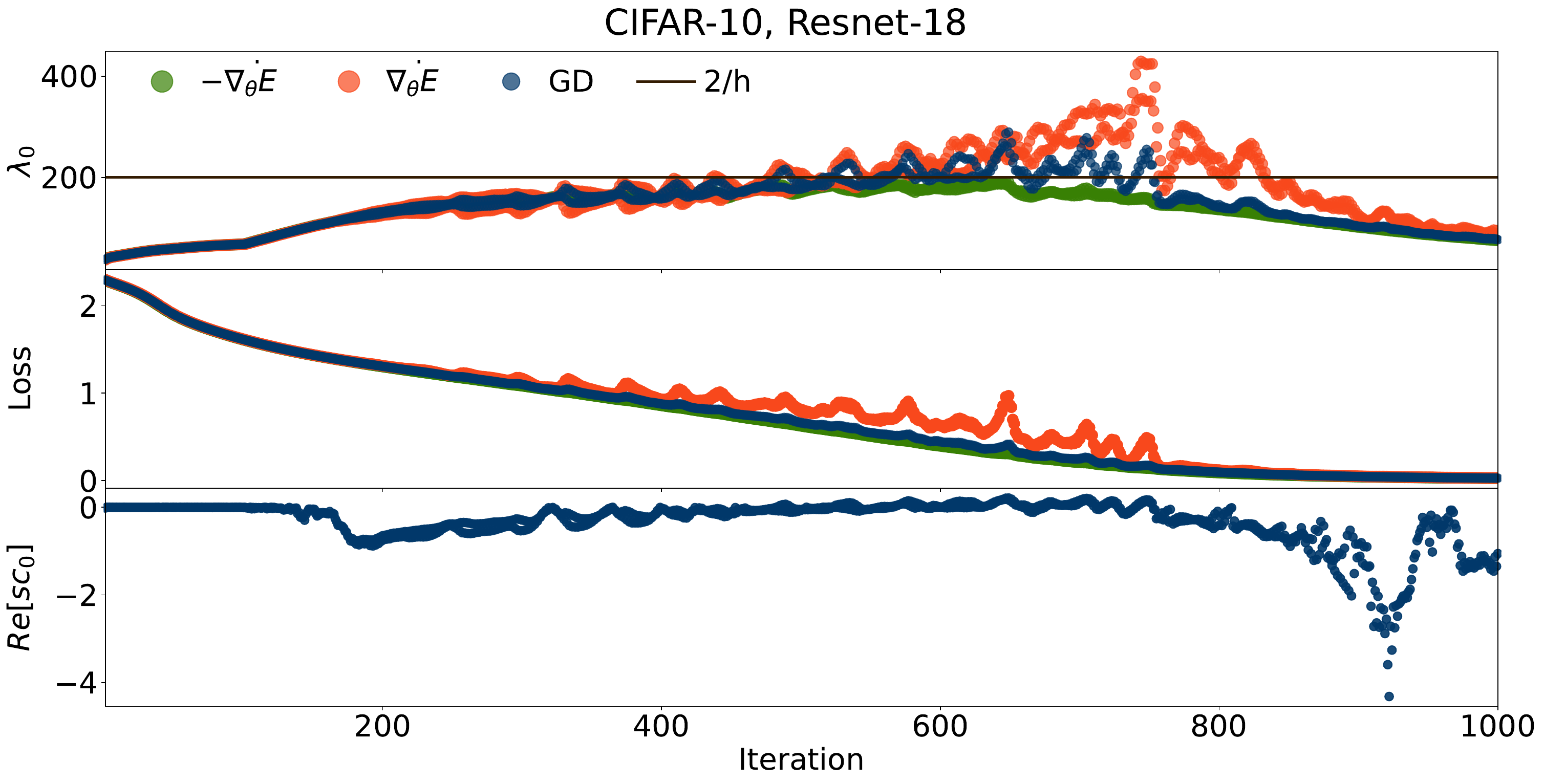}
\caption[The connection between the PF, $\lambda_0$ and loss instabilities with Resnet-18 trained on CIFAR-10.]{Learning rate: $0.01$. Edge of stability results: the connection between the PF, stability coefficients and loss instabilities with Resnet-18.  Together with the behaviour of gradient descent, we plot the behaviour of the NGF and positive gradient flow initialized at $\vtheta_t$ and simulated for time $h$ for each iteration $t$. \rebuttalrone{The analysis we performed based on the PF suggests that when $\Re[sc_0] > 0$ and large we should expected gradient descent to exhibit behaviours close to those of the positive gradient flow. What we observe empirically is that increases in loss value of gradient descent are proportional to the increase of the positive gradient flow in that area (can be seen best between iterations 600 and 800); the same behaviour can be seen in relation to the eigenvalue $\lambda_0$. Results for a larger learning rate are shown in Figure~\ref{fig:instabilities_resnet}. }}
\label{fig:edge_of_stability_results_cifar_resnet} 
\end{figure}

\noindent \textbf{DAL}.
We use gradient descent training with a fixed learning rate to measure the quantities we would like to use as a learning rate in DAL, to see if they have a reasonable range and show results in Figure~\ref{fig:connections_quantities}. We show sweeps across batch sizes Figures \ref{fig:power_sweeps_batch_sizes_vgg}, \ref{fig:imagenet_lr_scaling_across_batch_sizes}, \ref{fig:imagenet_lr_sqrt_scaling_across_batch_sizes}. We show DAL-$p$ sweeps with effective learning rates, training losses and test accuracies in Figure~\ref{fig:power_sweeps_all}. We show results with DAL on a mean square loss in Figure \ref{fig:least_square_loss_dal}.

\begin{figure}[tbh!]
\begin{subfloat}[VGG full batch.]{
  \includegraphics[width=0.33\columnwidth]{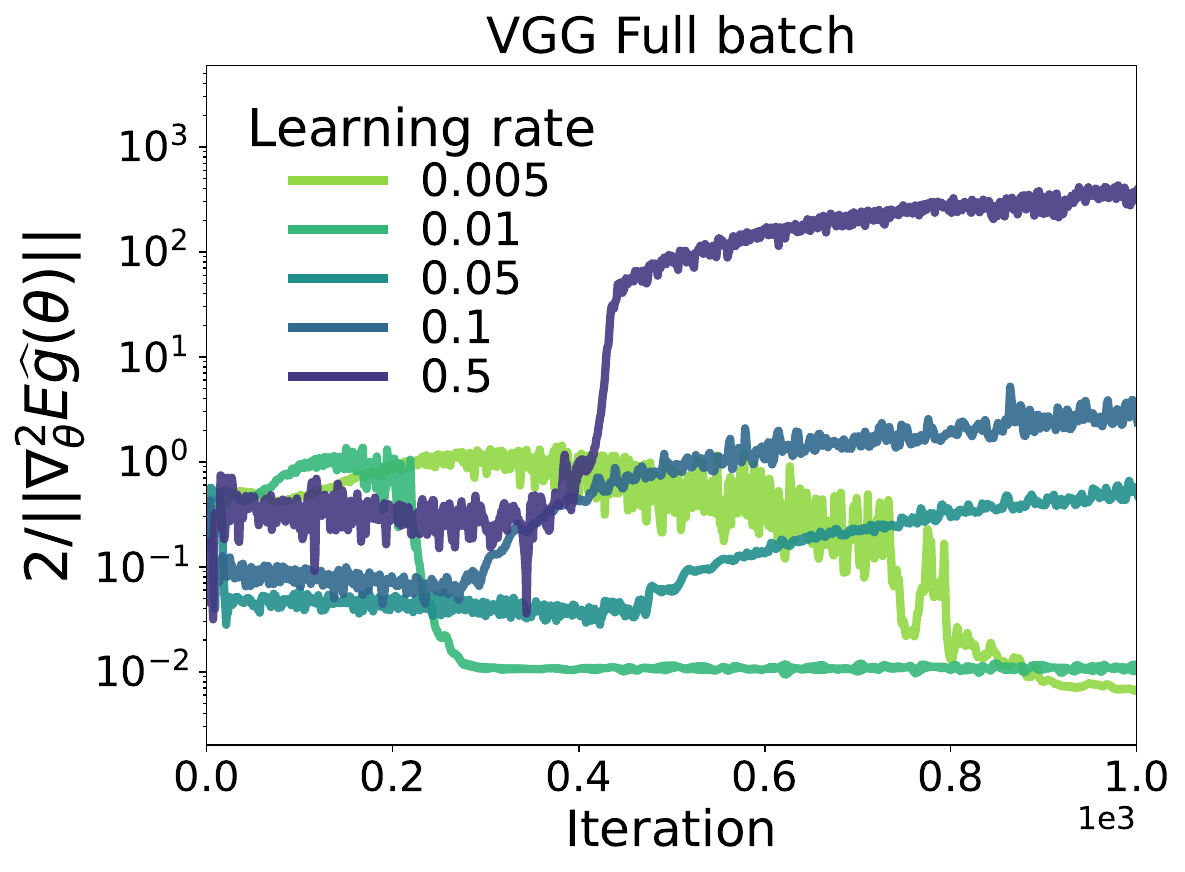}%
}
\end{subfloat}%
\begin{subfloat}[Resnet-18 full batch.]{
 \includegraphics[width=0.33\columnwidth]{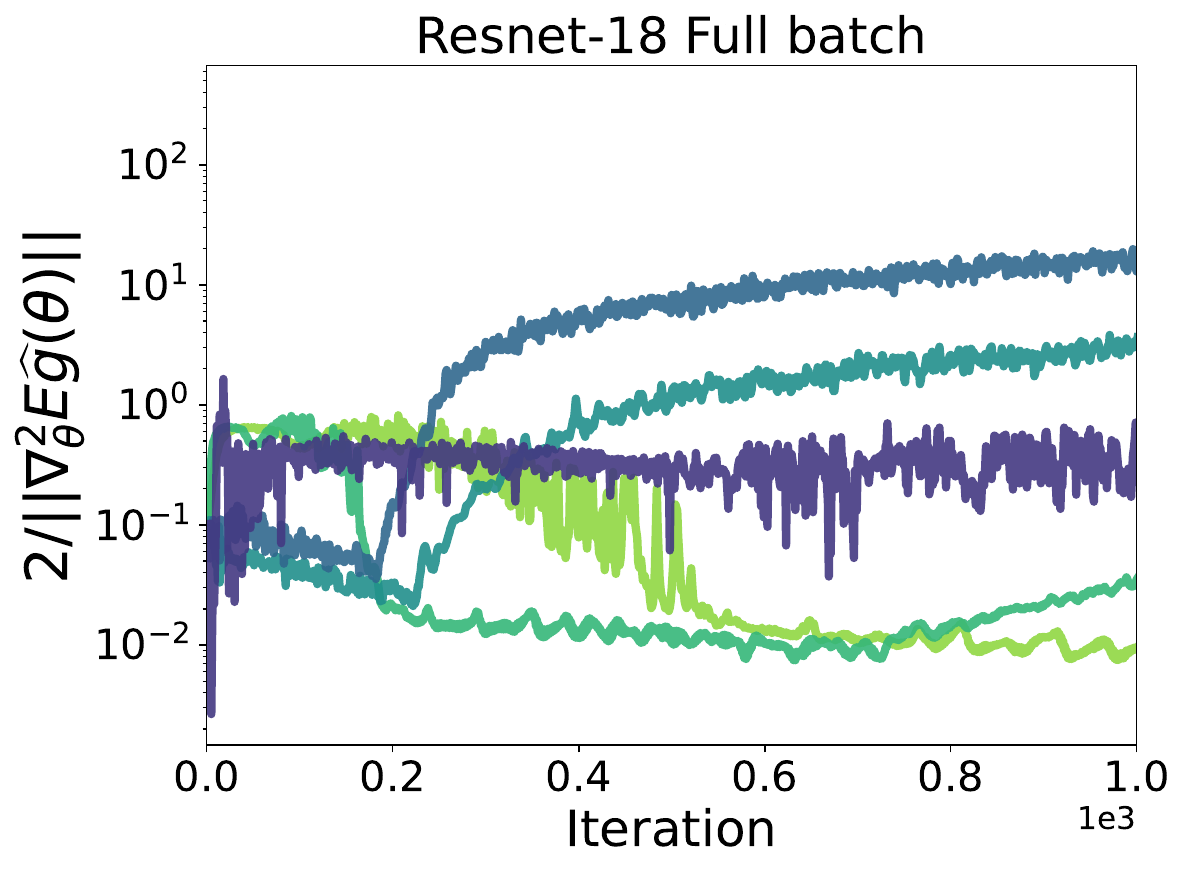}%
}
\end{subfloat}%
\begin{subfloat}[VGG batch size 64.]{
  \includegraphics[width=0.33\columnwidth]{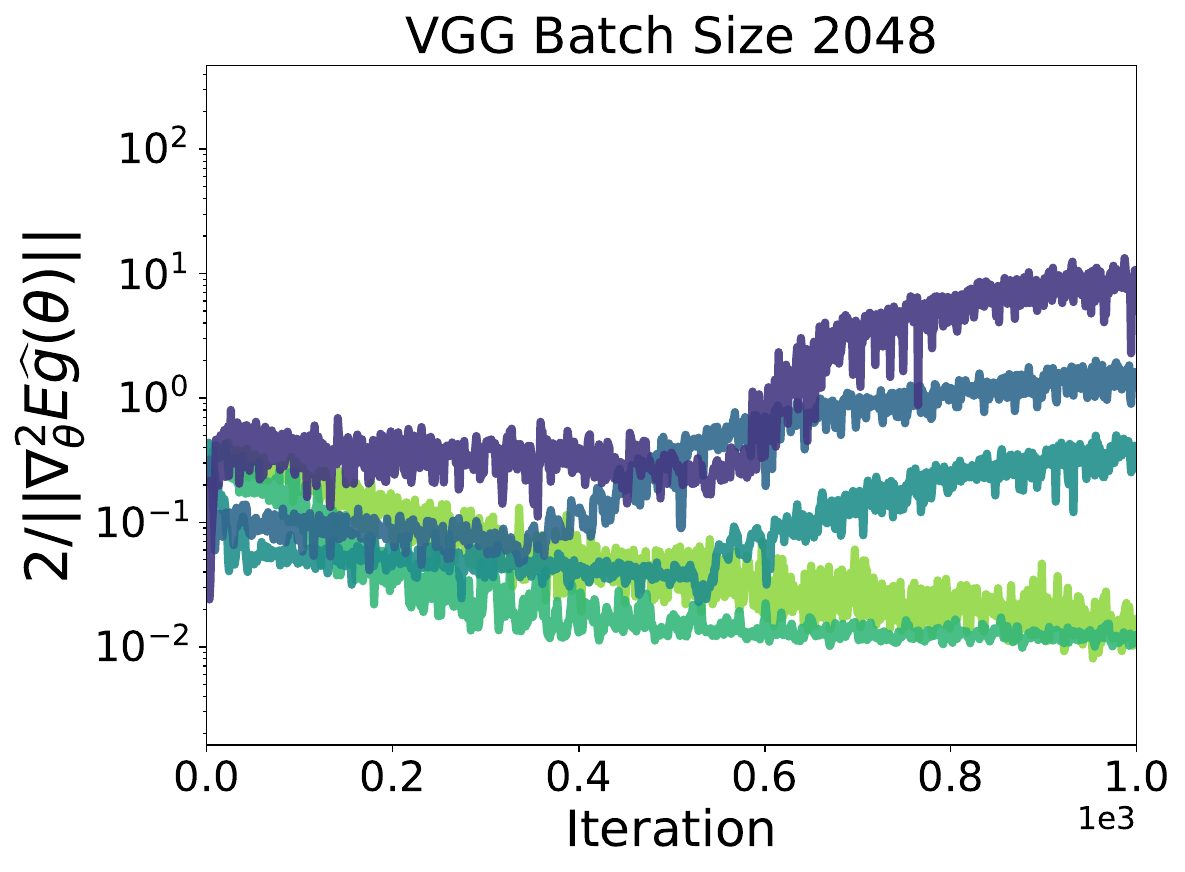}%
}
\end{subfloat}%
 \caption[Assessing whether the value of the DAL learning rate for a fixed learning rate sweep has magnitudes which can be used as a learning rate in training.]{Evaluating the value of the DAL learning rate for a fixed learning rate sweep to assess whether it would have reasonable magnitudes. We observe that the quantity is mainly the ranges $10^{-2}$ to $10$, which can be used for model training. We note that this is a preliminary check only, as using DAL in training changes the optimisation trajectory and thus can affect these ranges. As we have seen however, DAL can be suitably used for model training.}
\label{fig:connections_quantities}
\end{figure}

\begin{figure}
\centering
\begin{subfloat}[Batch size 64.]{
  \includegraphics[width=0.45\columnwidth]{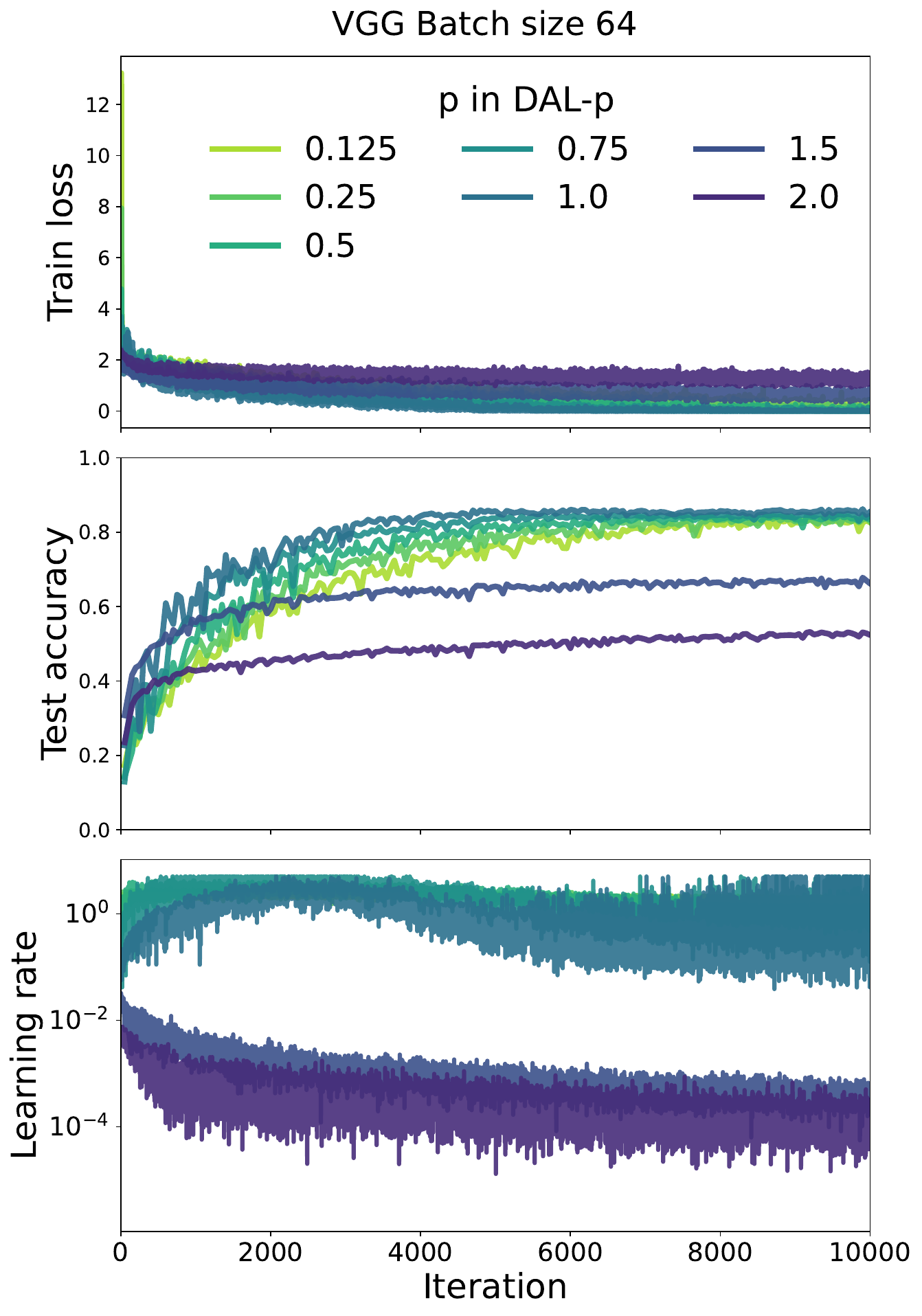}%
}\end{subfloat}%
\begin{subfloat}[Batch size 256.]{
 \includegraphics[width=0.45\columnwidth]{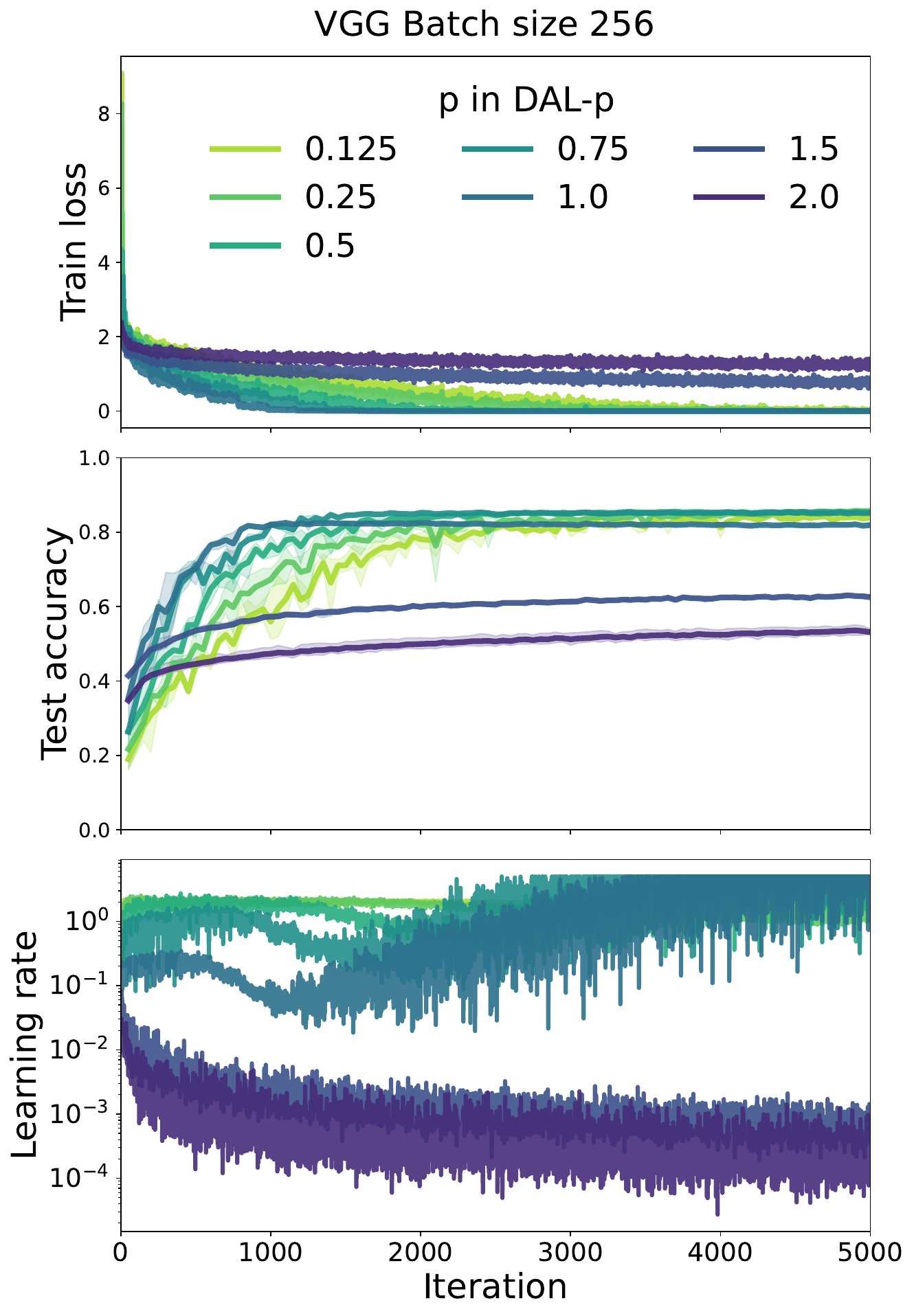}
}\end{subfloat}\\
\begin{subfloat}[Batch size 1024.]{
 \includegraphics[width=0.45\columnwidth]{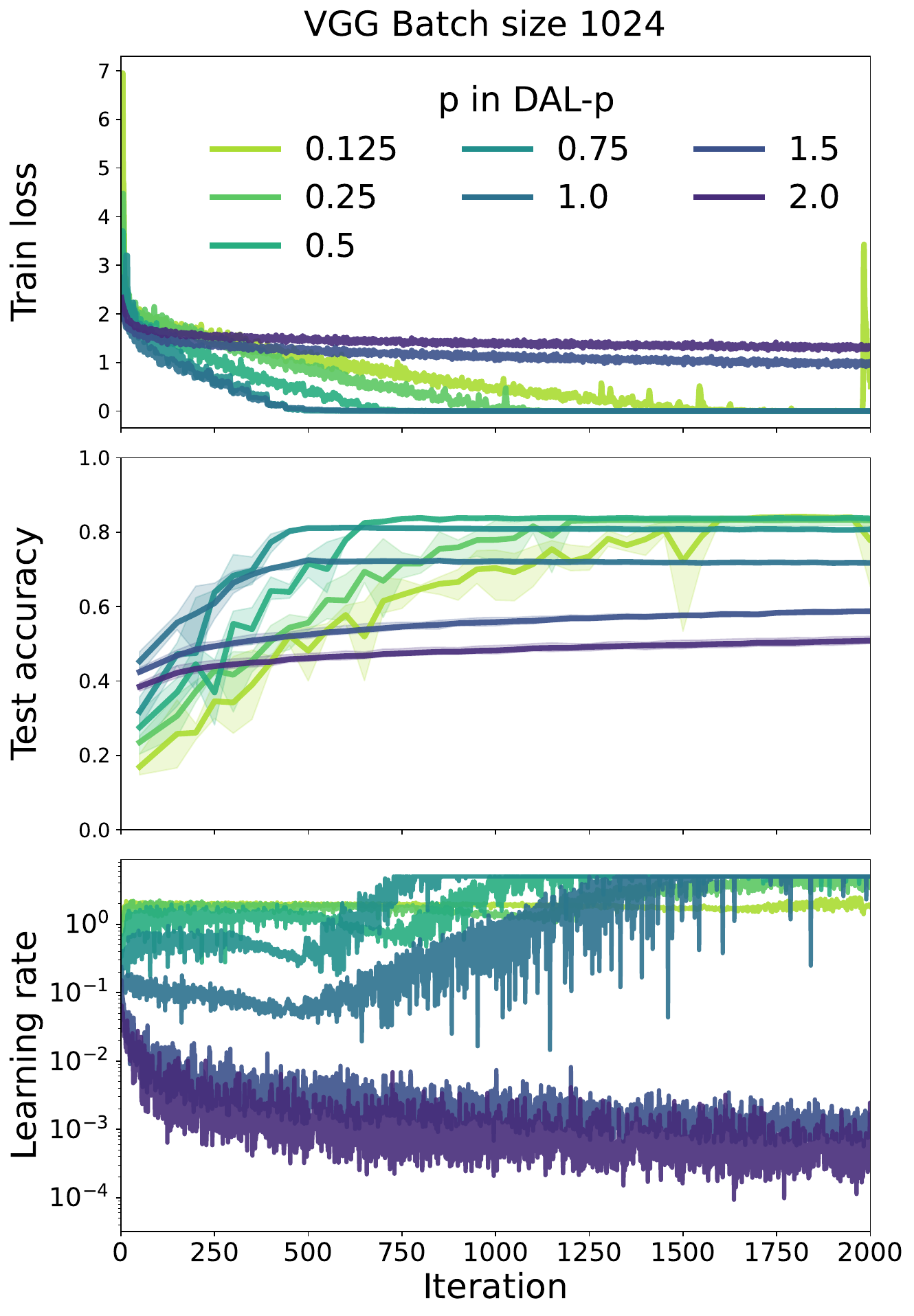}%
}\end{subfloat}%
\begin{subfloat}[Full batch.]{
  \includegraphics[width=0.45\columnwidth]{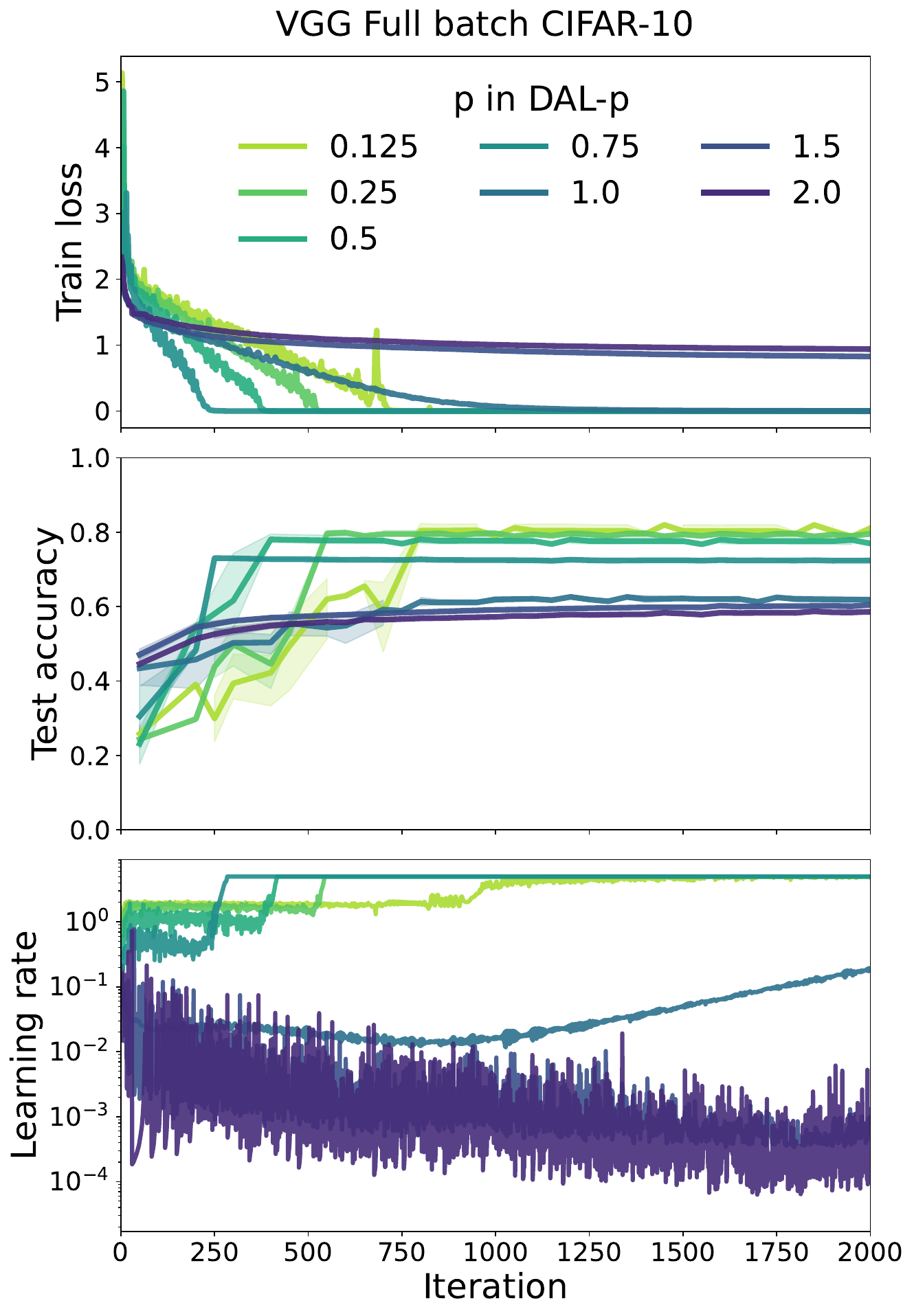}%
}\end{subfloat}%
\caption[DAL-$p$ sweep with a VGG model trained on CIFAR-10 for small and large batch sizes.]{DAL-$p$ training a VGG model on CIFAR-10. Sweep across batch sizes: discretisation drift helps test performance, but at the cost of stability. We also show the effective learning rate and train losses and test accuracies.}
\label{fig:power_sweeps_batch_sizes_vgg}
\end{figure}

\begin{figure}[tb]
\centering
\begin{subfloat}[Training loss.]{
 \includegraphics[width=0.52\columnwidth]{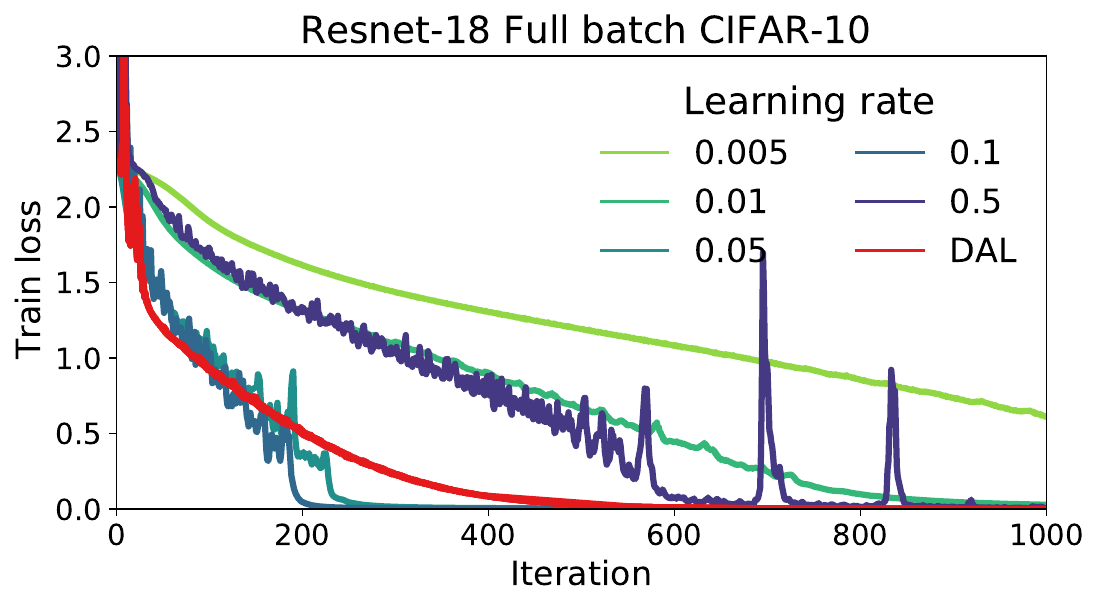}
}%
\end{subfloat}%
\begin{subfloat}[Test accuracy.]{
   \includegraphics[width=0.45\columnwidth]{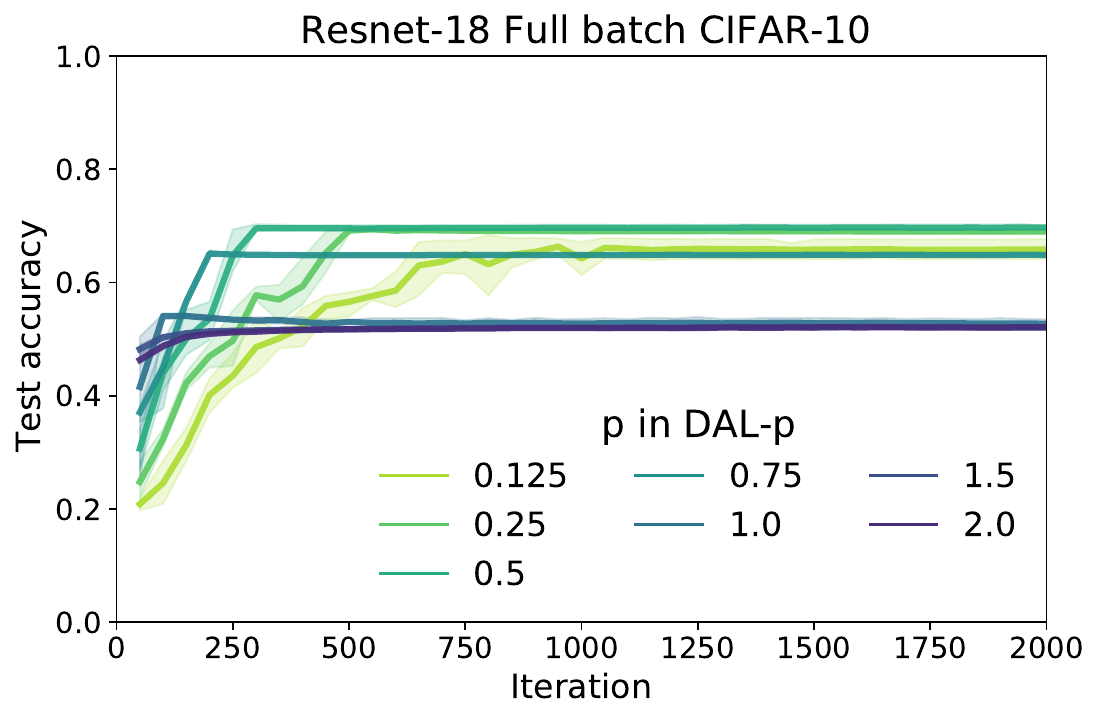}
}%
\end{subfloat}%
\caption[Models trained using a learning rate sweep or DAL on CIFAR-10 using a Resnet-18 model.]{DAL and DAL$-p$ results with Resnet-18 on CIFAR-10: DAL results in improved stability without requiring a hyperparameter sweep, while DAL-$p$ achieves high generalisation.}
\label{fig:dal_resnet}
\end{figure}

\begin{figure}[tbh]
\begin{subfloat}[Batch size 1024.]{
 \includegraphics[width=0.45\columnwidth]{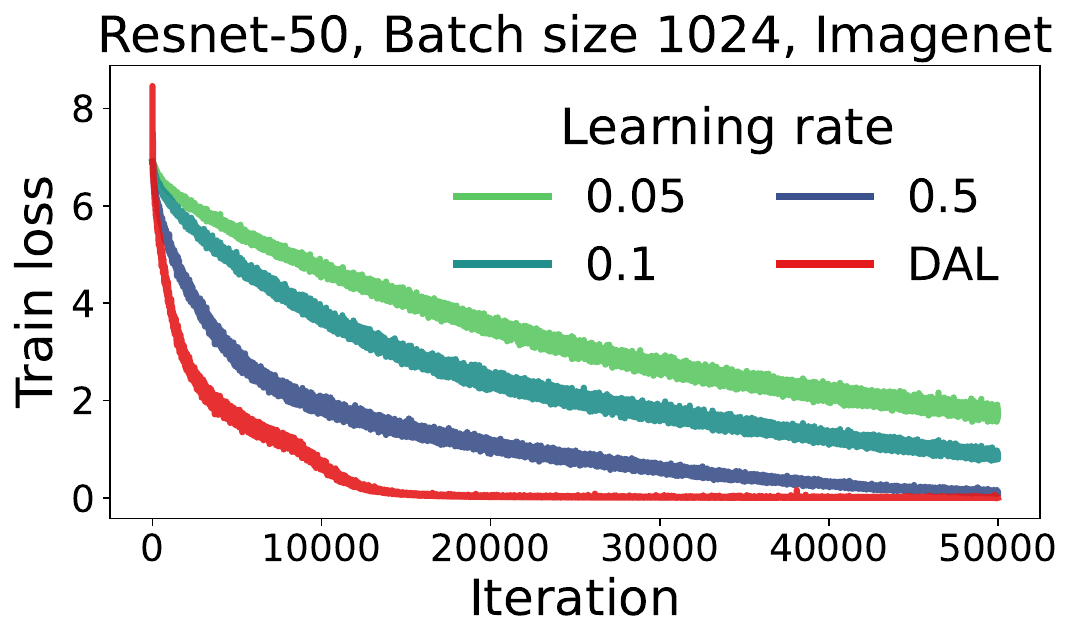}
}\end{subfloat}%
\begin{subfloat}[Batch size 2048.]{
  \includegraphics[width=0.45\columnwidth]{imagenet_train_loss_h_g_scaling_2048}
}\end{subfloat}\\
\begin{subfloat}[Batch size 4096.]{
  \includegraphics[width=0.45\columnwidth]{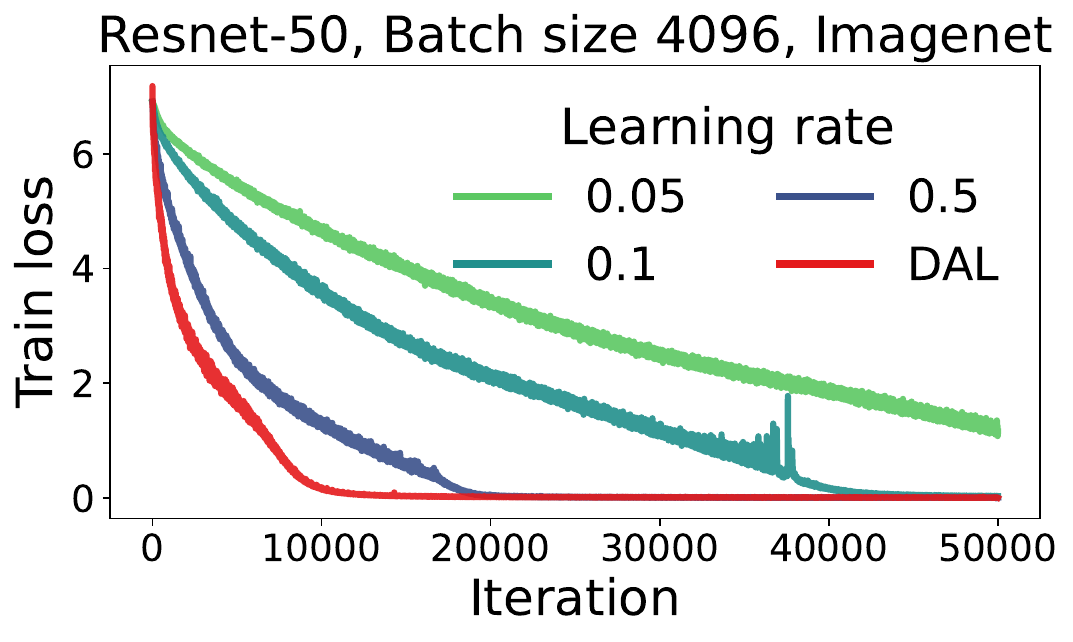}
}\end{subfloat}%
\begin{subfloat}[Batch size 8192.]{
  \includegraphics[width=0.45\columnwidth]{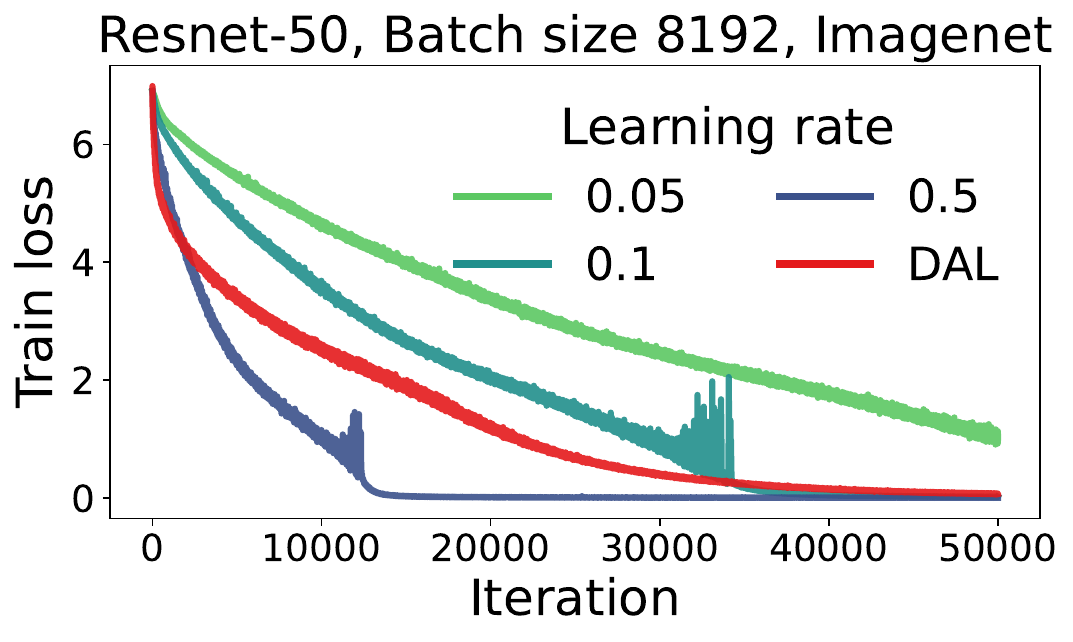}
}\end{subfloat}%
\caption[DAL on Imagenet, a learning rate sweep across batch sizes.]{DAL: Imagenet results across batch sizes. We observe quick convergence and stable training.}
\label{fig:imagenet_lr_scaling_across_batch_sizes}
\end{figure}

\begin{figure}[tbh]
\begin{subfloat}[Batch size 1024.]{
 \includegraphics[width=0.45\columnwidth]{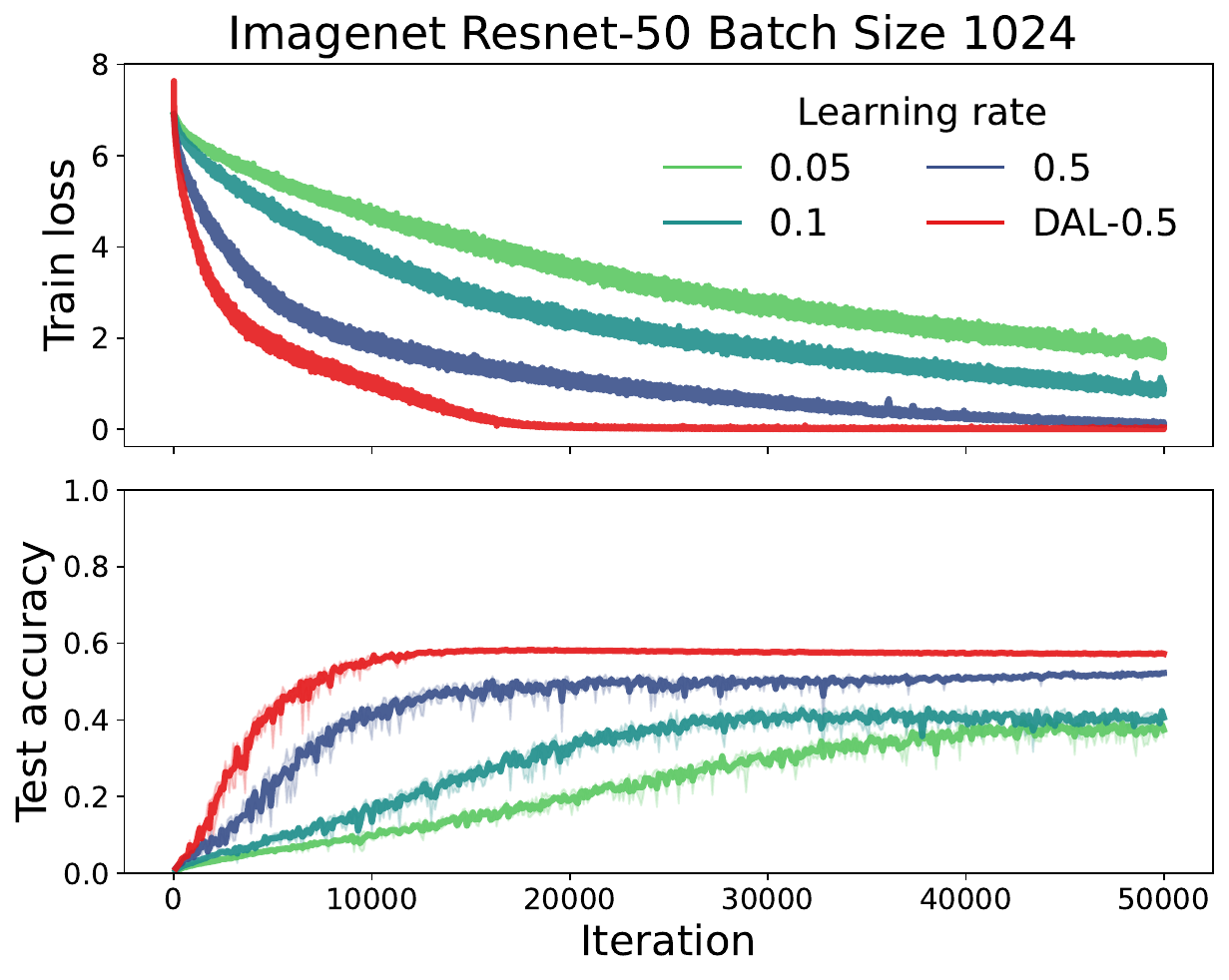}
}\end{subfloat}%
\begin{subfloat}[Batch size 2048.]{
  \includegraphics[width=0.45\columnwidth]{imagenet_2048_train_test_h_g_scaling_sqrt}
}\end{subfloat}\\
\begin{subfloat}[Batch size 4096.]{
  \includegraphics[width=0.45\columnwidth]{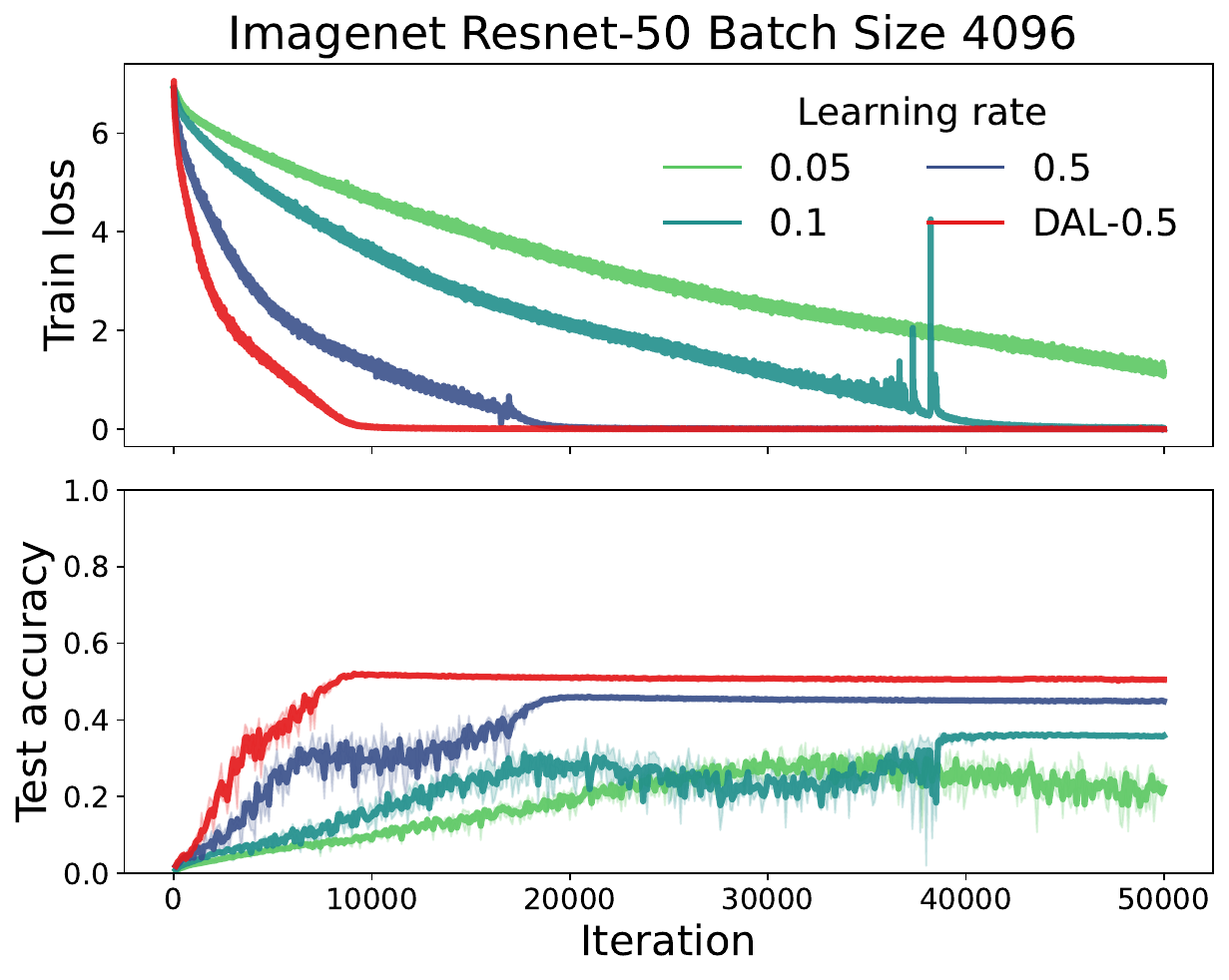}
}\end{subfloat}%
\begin{subfloat}[Batch size 8192.]{
  \includegraphics[width=0.45\columnwidth]{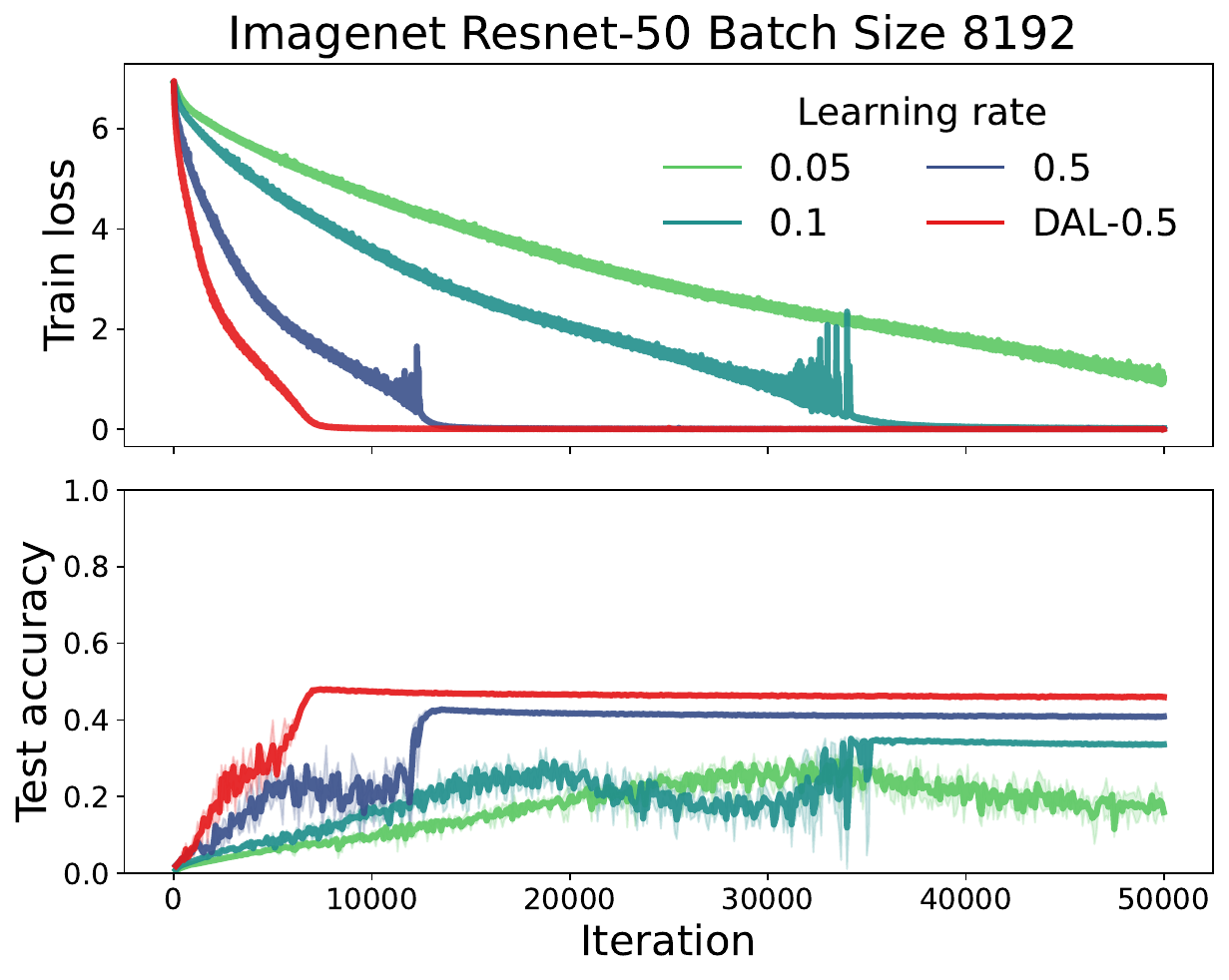}
}\end{subfloat}%
\caption[DAL$-0.5$: on Imagenet, a learning rate sweep across batch sizes.]{DAL$-0.5$: Imagenet results across batch sizes. We observe quick training and good generalisation performance.}
\label{fig:imagenet_lr_sqrt_scaling_across_batch_sizes}
\end{figure}

\begin{figure}[tbh]
\centering
\begin{subfloat}[Resnet-18 on CIFAR-10.]{
 \includegraphics[width=0.45\columnwidth]{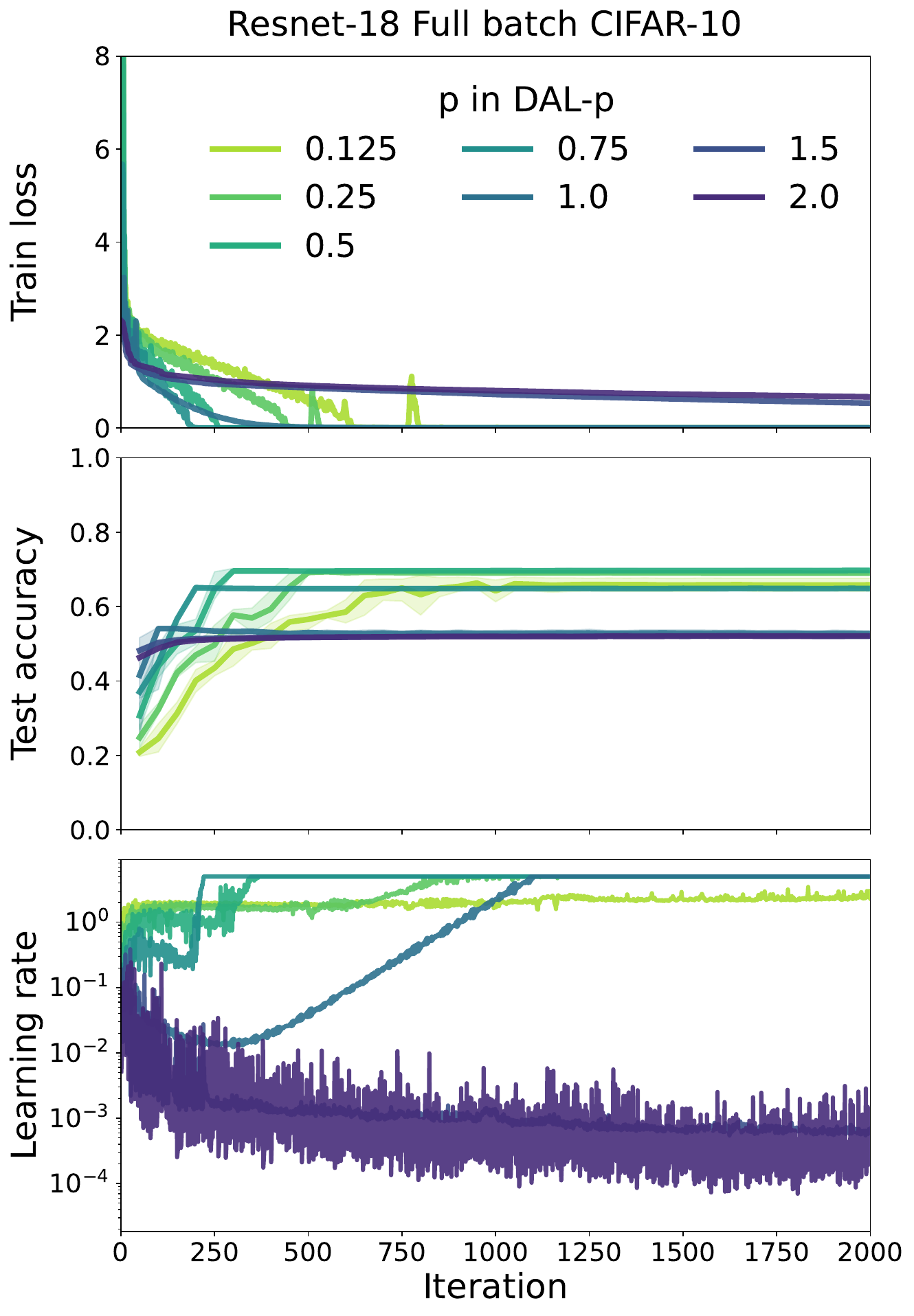}%
}\end{subfloat}%
\begin{subfloat}[Resnet-50 on Imagenet.]{
 \includegraphics[width=0.45\columnwidth]{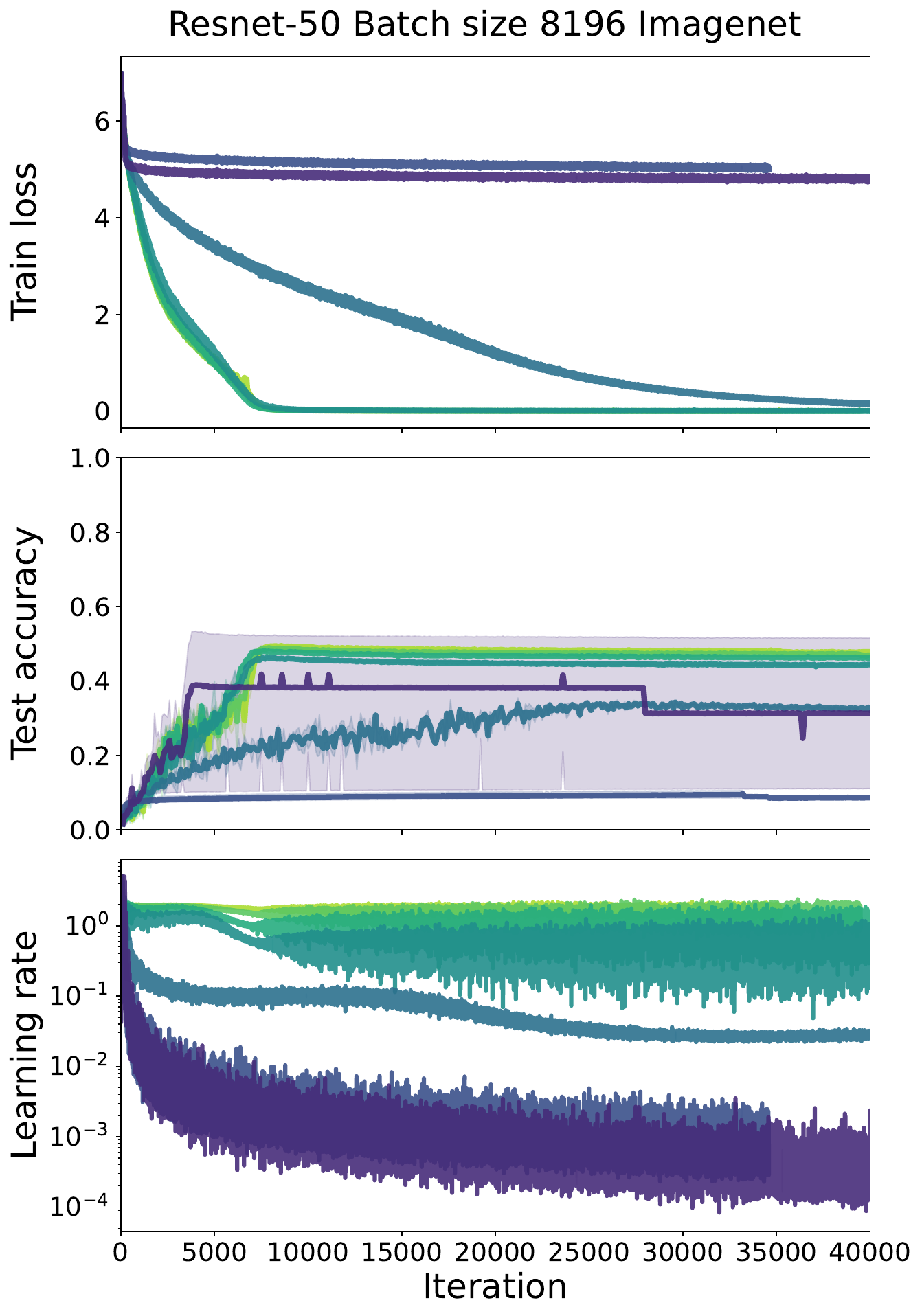}%
}\end{subfloat}%
\caption[DAL-$p$ sweep on VGG, Resnet-18 and Imagenet.]{DAL-$p$ sweep: discretisation drift helps test performance, but at the cost of stability. We also show the effective learning rate and train losses and test accuracies.}
\label{fig:power_sweeps_all}
\end{figure}

\begin{figure}
\centering
\begin{subfloat}[Training loss.]{
 \includegraphics[width=0.6\columnwidth]{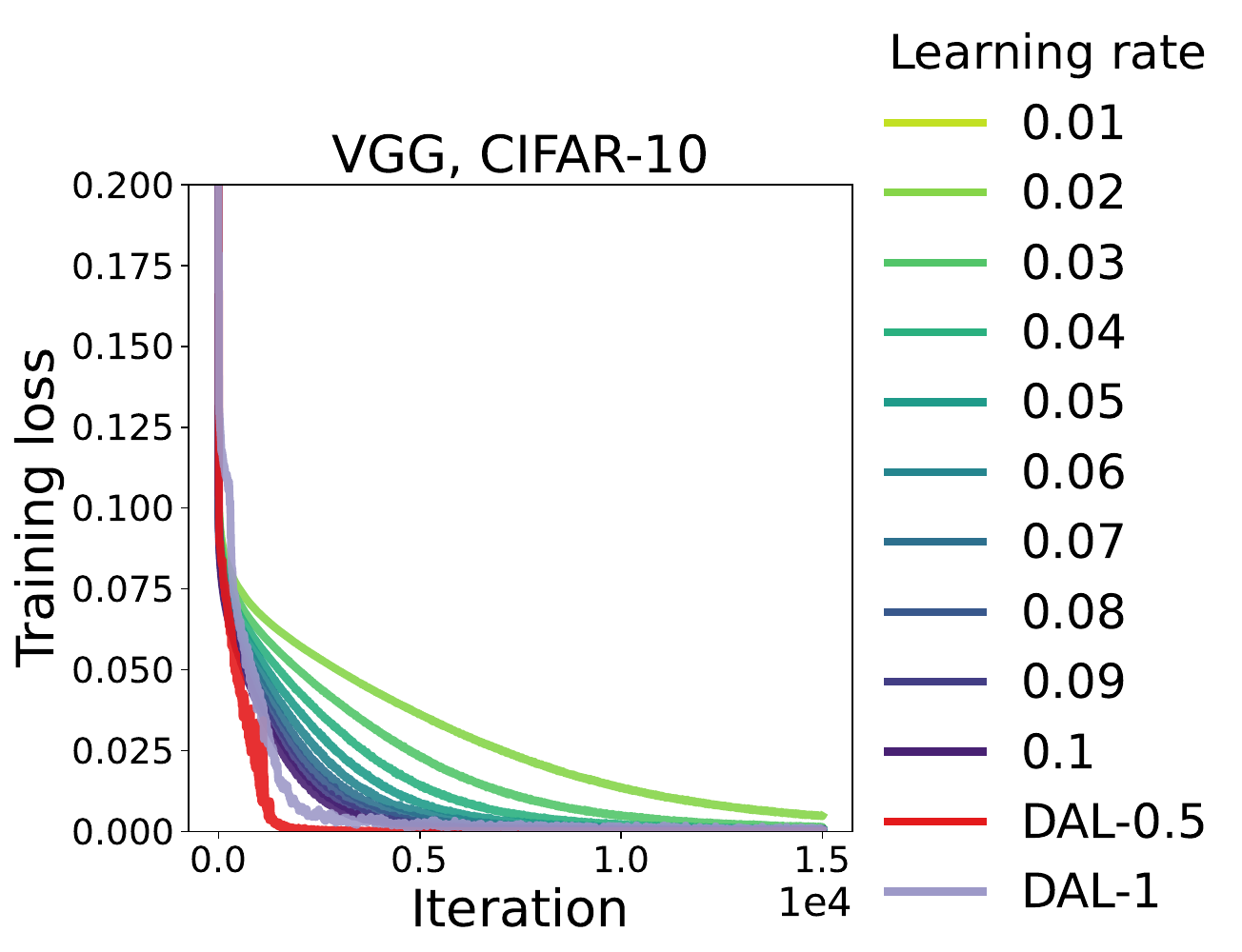}
}\end{subfloat}%
\begin{subfloat}[$\lambda_0$.]{
 \includegraphics[width=0.4\columnwidth]{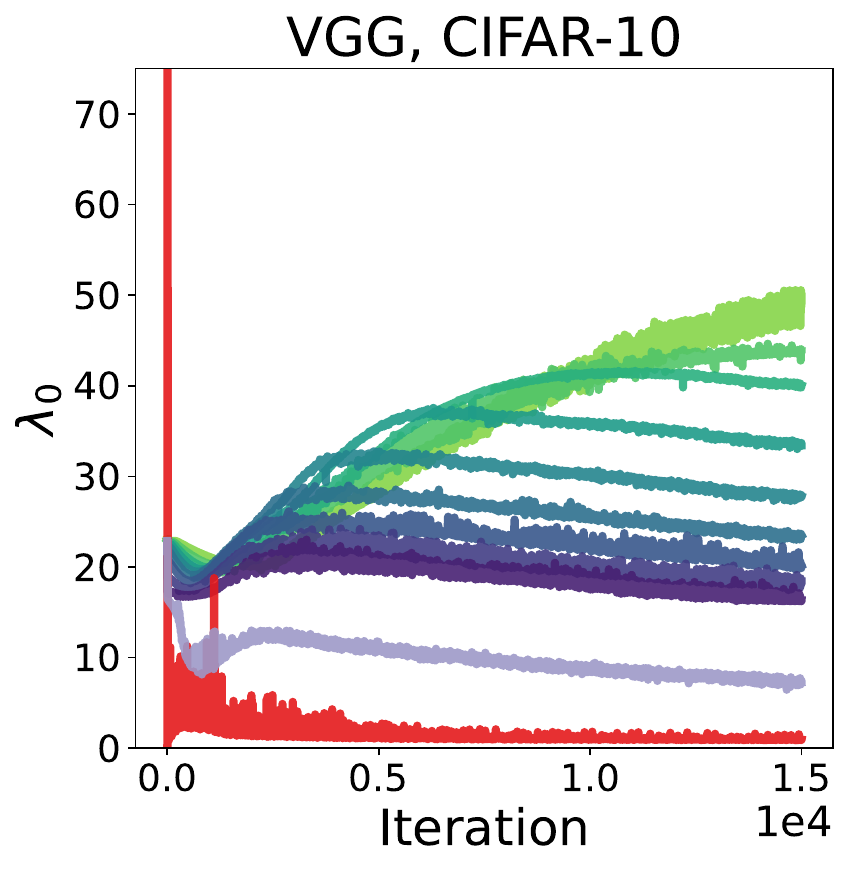}
}\end{subfloat}%
\caption[DAL results with a least square loss. ]{Results with a least square loss. DAL and DAL-$0.5$ lead to quicker training and lower $\lambda_0$ in this setting as well.}
\label{fig:least_square_loss_dal}
\end{figure}

\noindent \textbf{DAL learned landscapes.}
To investigate the landscapes learned by DAL-$p$, we use the method of~\citet{li2018visualizing} and compare against the landscapes learned using SGD. We have shown results with batch size 64 on CIFAR-10 in Figure~\ref{fig:DAL_p_64_landscape} in the main text. Here present additional results in Figures~\ref{fig:DAL_p_full_batch_landscape},~\ref{fig:DAL_p_imagenet_landscape}. Across datasets and batch sizes, we consistently observe that DAL-$p$ learns flatter landscapes. We also consistently observe that during training, $\lambda_0$ is smaller with DAL-$p$ than with SGD, as shown in Figure~\ref{fig:DAL_p_full_batch_eigen}.

\begin{figure}[t]
\centering
\begin{subfloat}[DAL]{
\includegraphics[width=0.49\columnwidth]{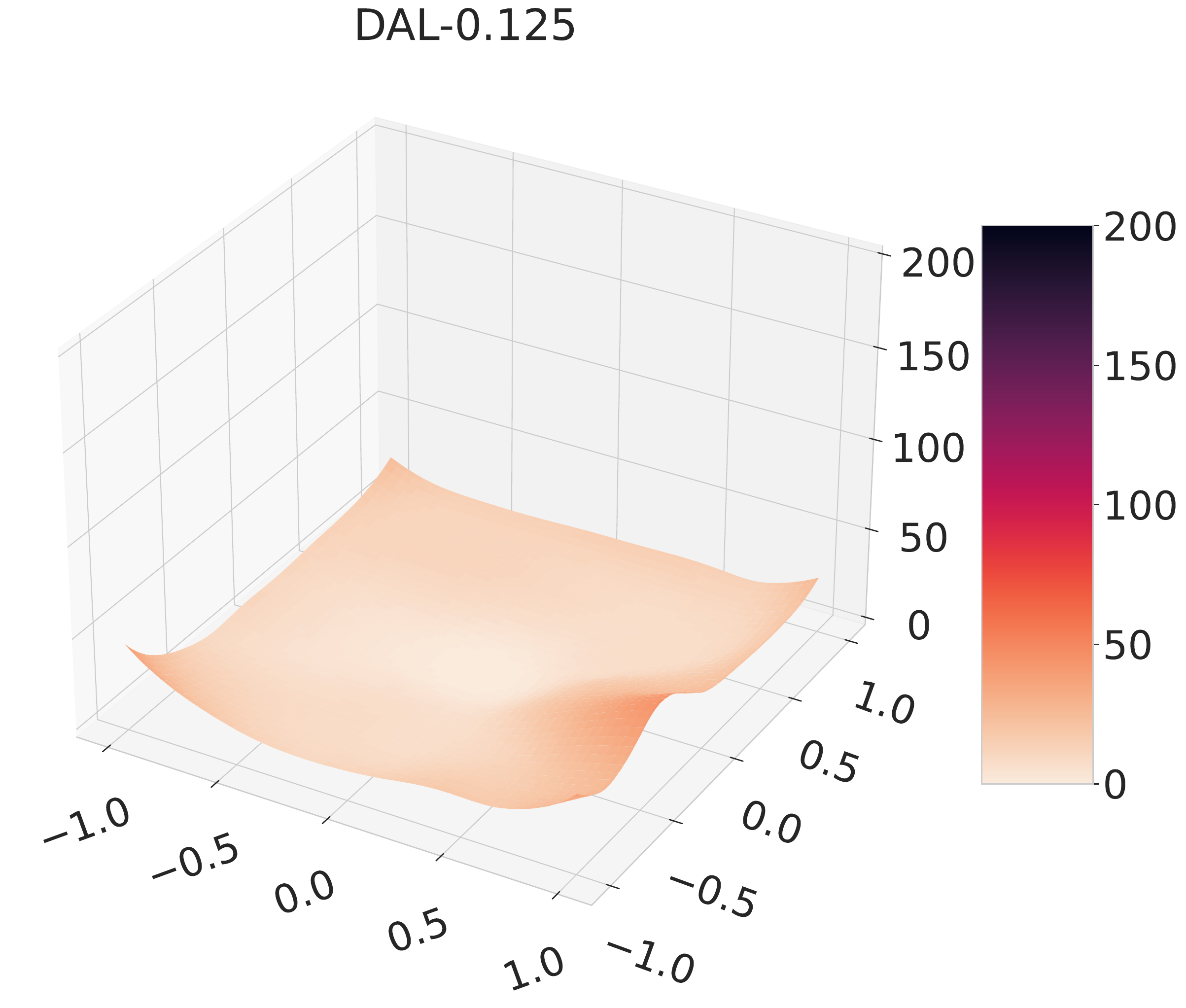}%
}
\end{subfloat}%
\begin{subfloat}[SGD]{
\includegraphics[width=0.49\columnwidth]{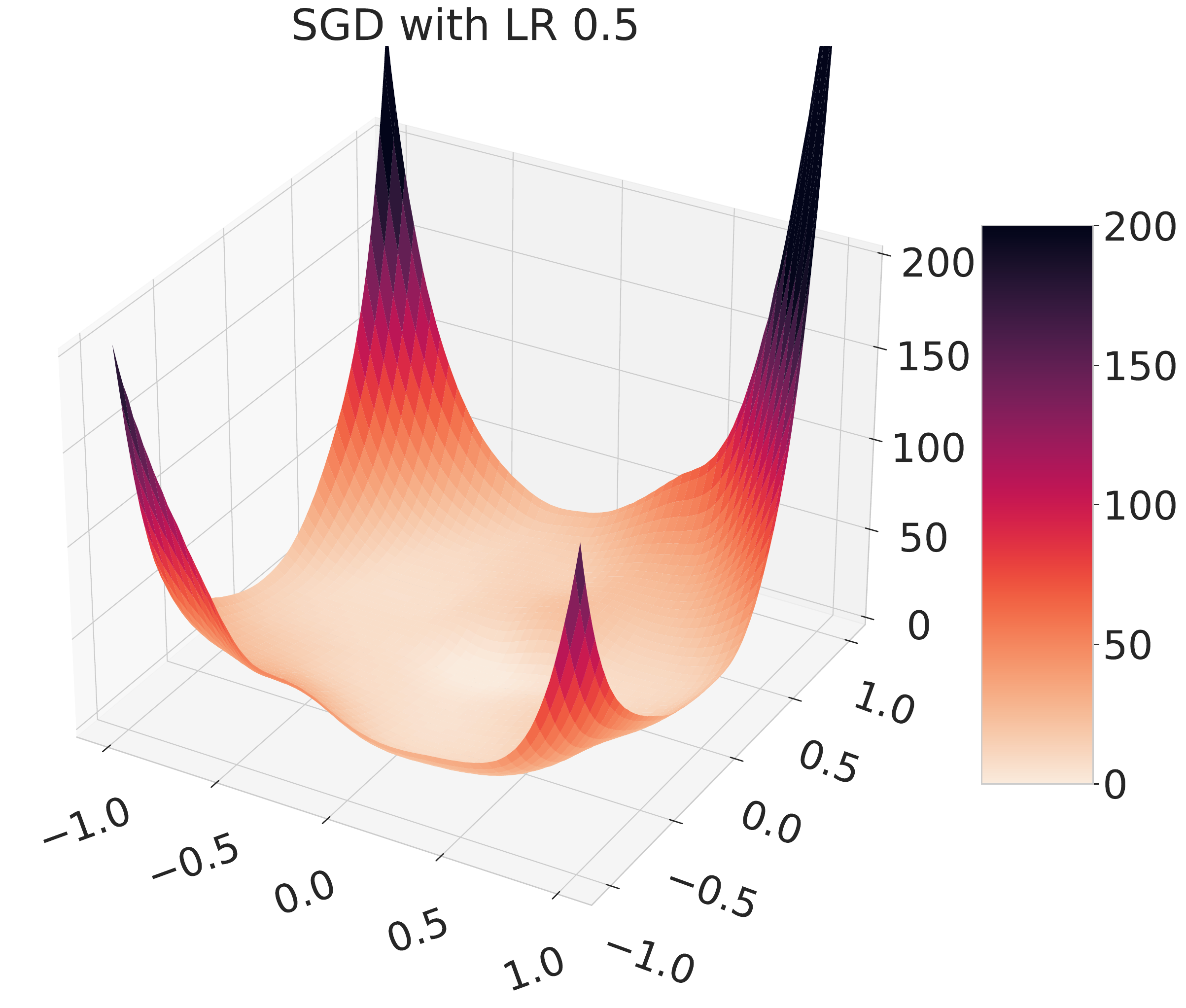}
}
\end{subfloat}

\begin{subfloat}[DAL]{
\includegraphics[width=0.49\columnwidth]{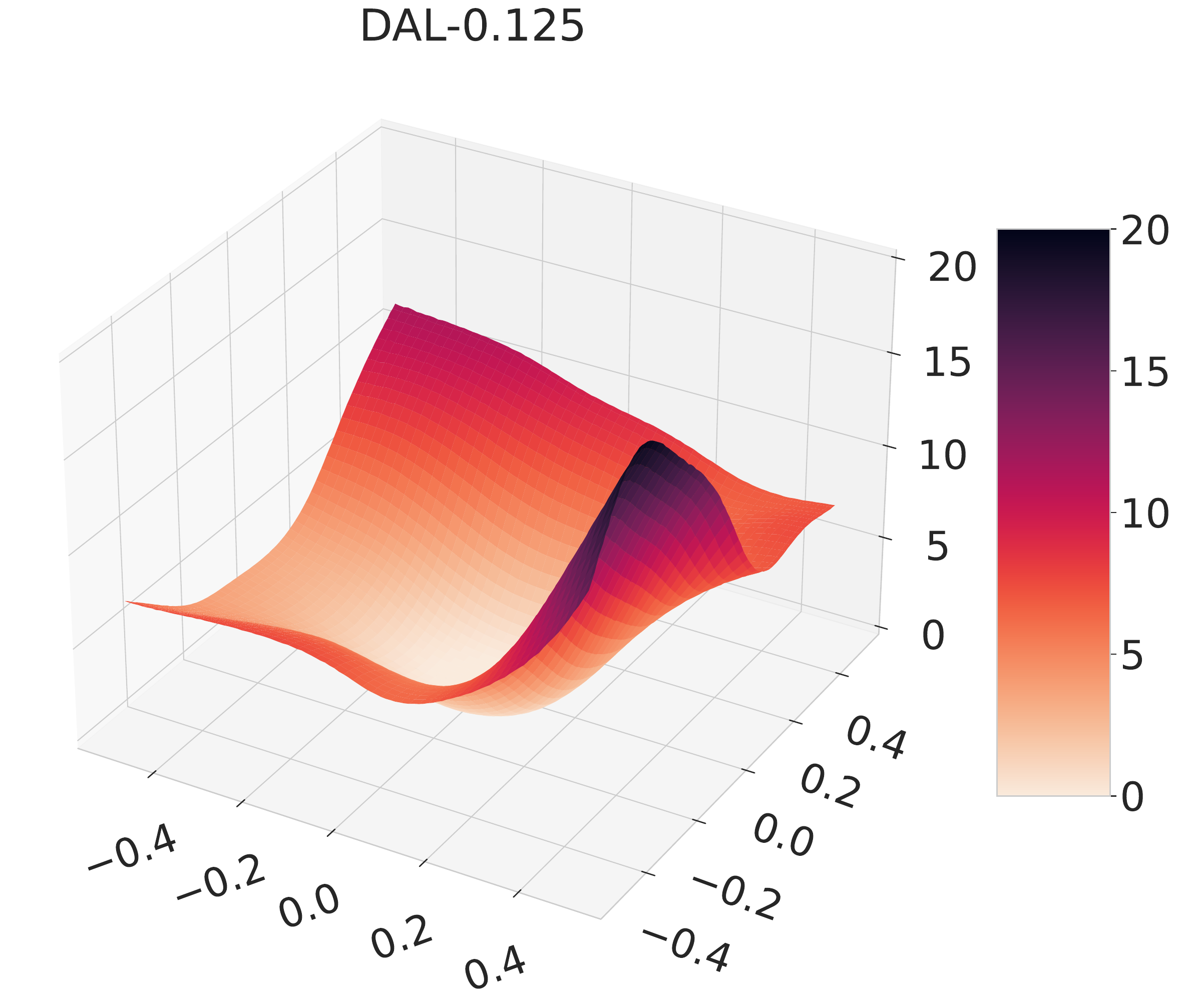}
}%
\end{subfloat}%
\begin{subfloat}[SGD]{
\includegraphics[width=0.49\columnwidth]{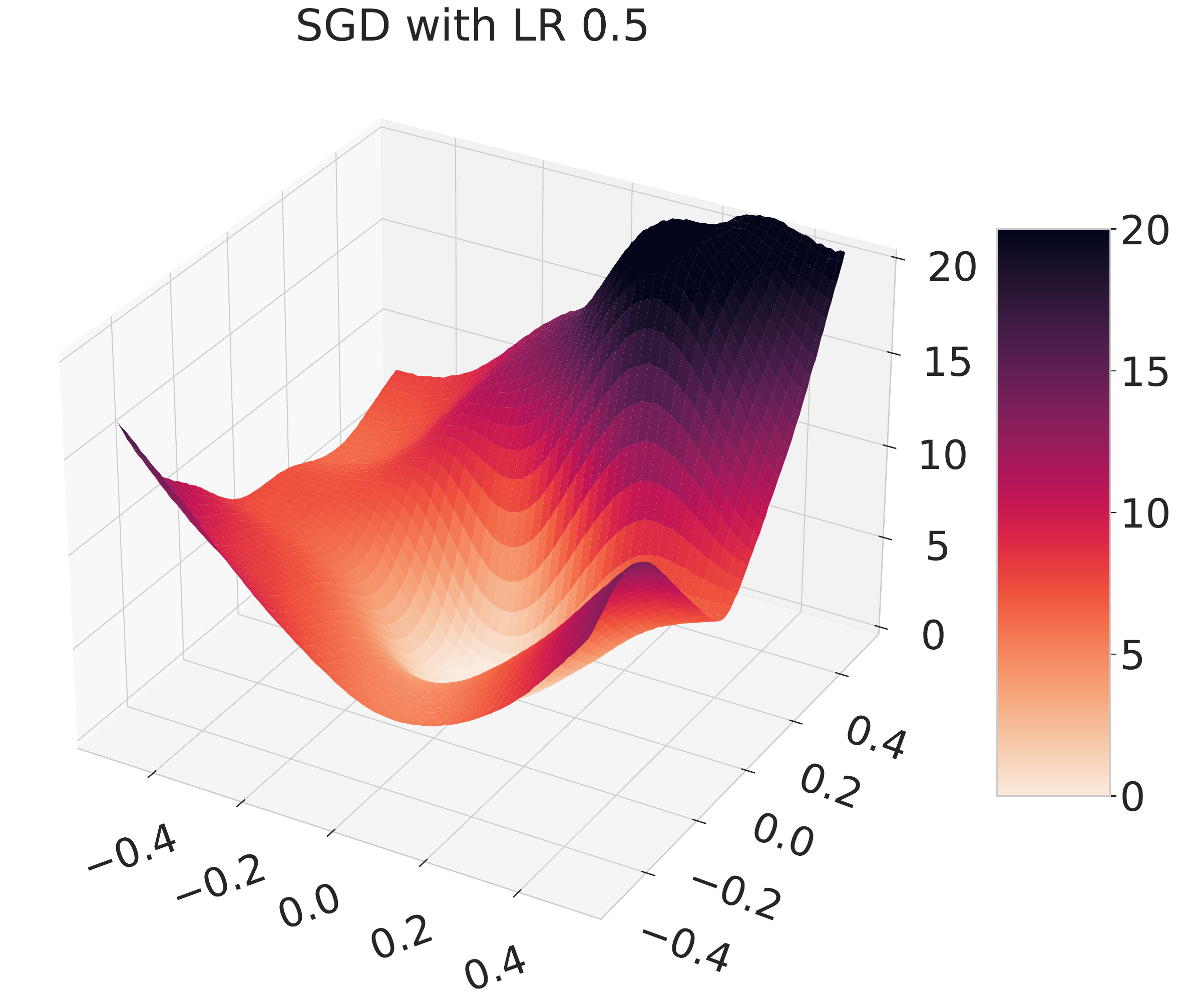}%
}
\end{subfloat}%
\caption[DAL and gradient descent learned landscapes; full batch on CIFAR-10.]{\textbf{CIFAR-10, full batch, at various levels of zoom into around the convergence points.}  The 2D projection of the DAL-$p$ and SGD learned landscapes on CIFAR-10, using a VGG model. The visualisation is made using the method of~\citet{li2018visualizing}. The DAL-$p$ model achieves an accuracy of 82\% which the SGD model achieves 77\% accuracy. We also show the trajectory of $\lambda_0$ for both models in Figure~\ref{fig:DAL_p_full_batch_eigen}.}
\label{fig:DAL_p_full_batch_landscape}
\end{figure}

\begin{figure}[t]
\begin{subfloat}[DAL]{
\includegraphics[width=0.4\columnwidth]{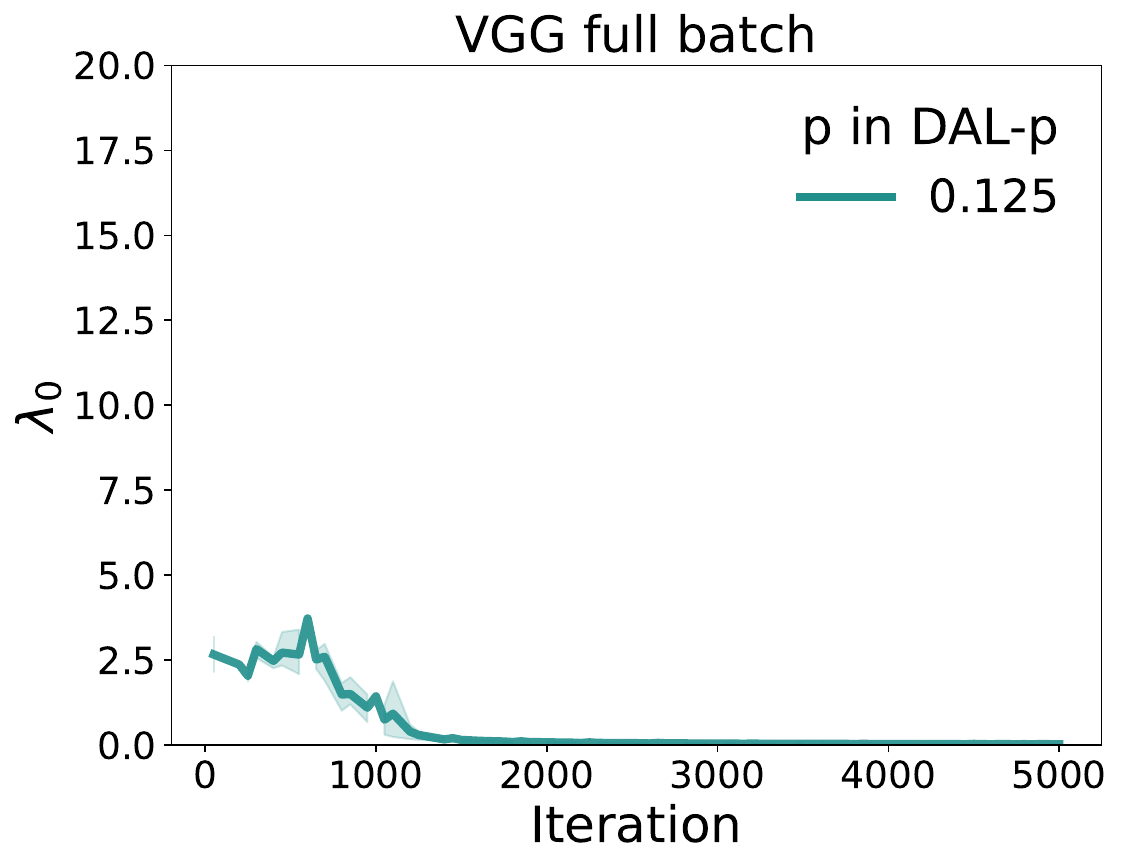}
}
\end{subfloat}
\begin{subfloat}[SGD]{
\includegraphics[width=0.4\columnwidth]{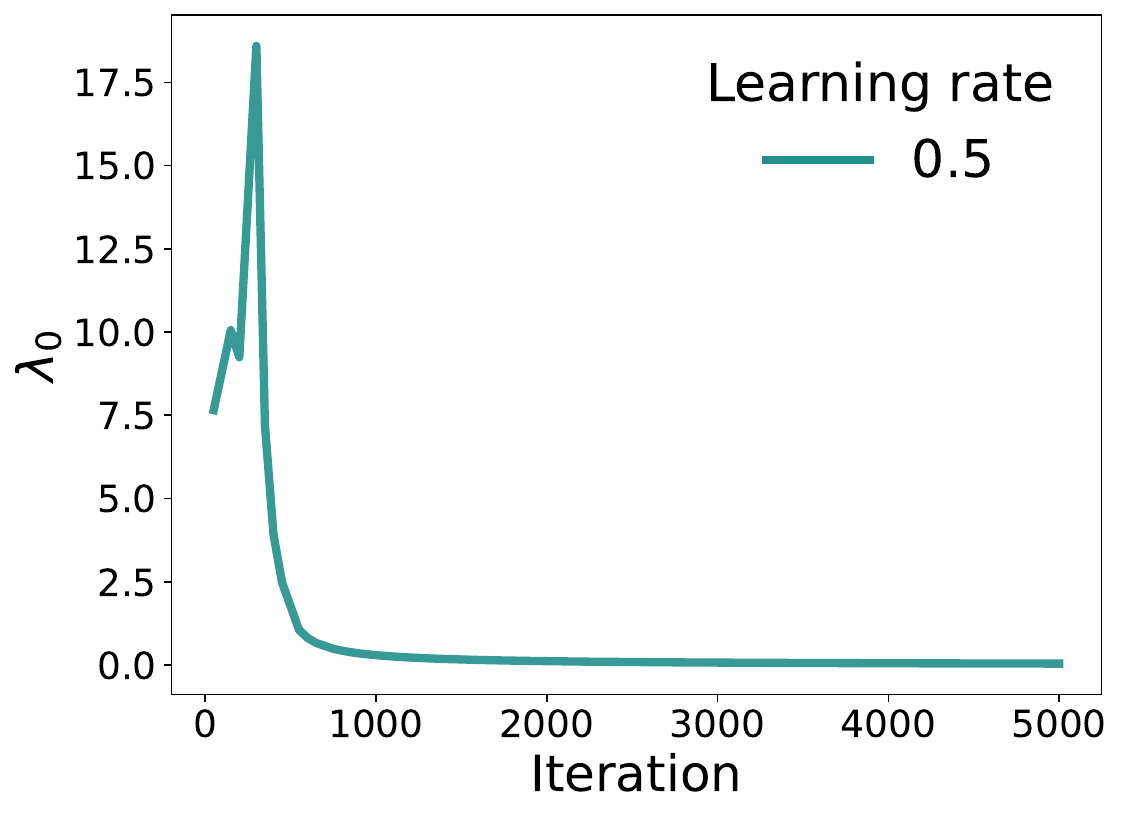}
}
\end{subfloat}
\caption[$\lambda_0$ in DAL and gradient descent; full batch on CIFAR-10.]{$\lambda_0$ for learned models using full batch gradient descent and DAL-$p$ on CIFAR-10: DAL traverses areas of space with smaller $\lambda_0$.}
\label{fig:DAL_p_full_batch_eigen}
\end{figure}

\begin{figure}[t]
\centering
\begin{subfloat}[DAL]{
\includegraphics[width=0.49\columnwidth]{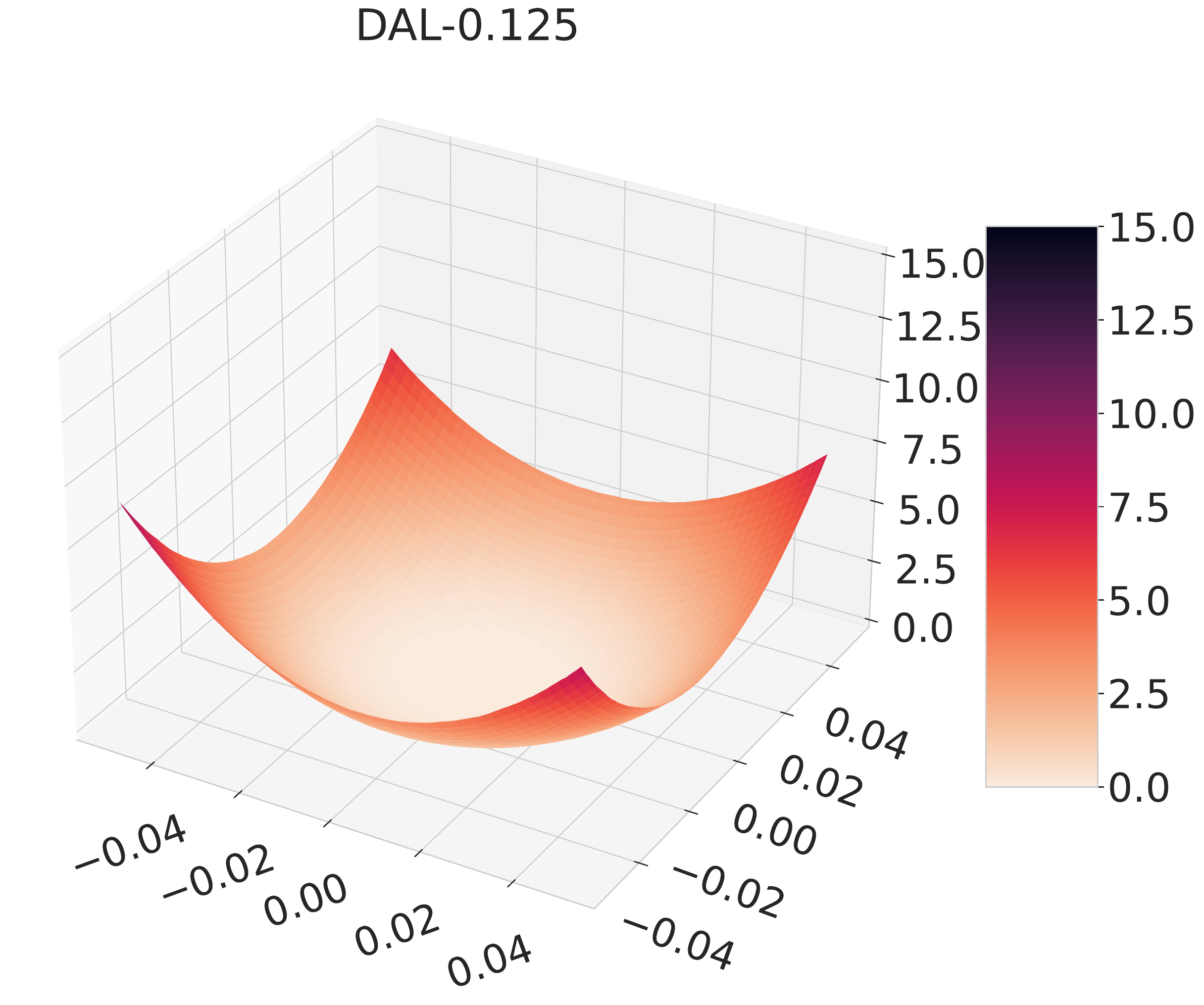}
}%
\end{subfloat}%
\begin{subfloat}[SGD]{
\includegraphics[width=0.49\columnwidth]{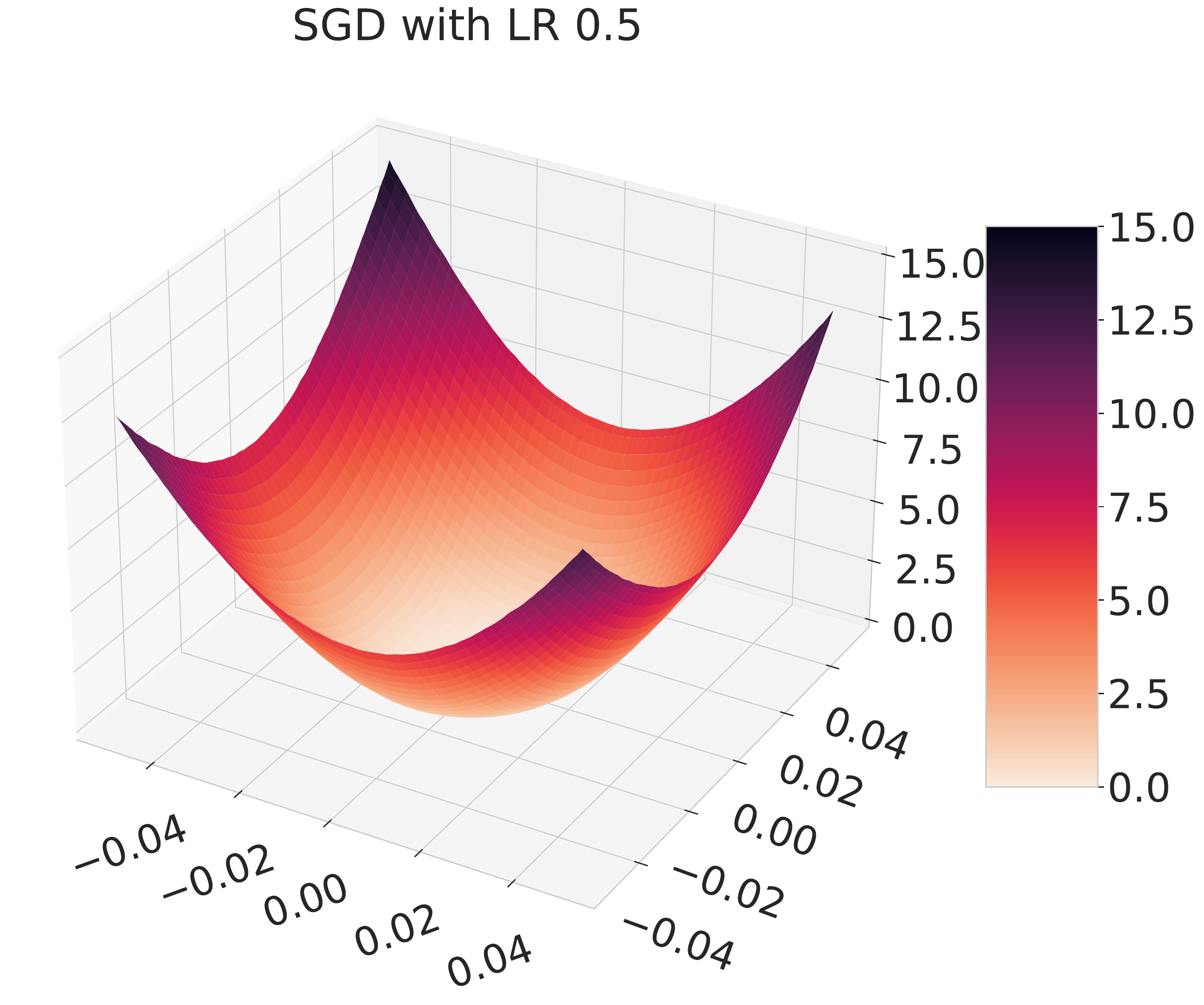}%
}
\end{subfloat}%
\caption[DAL and gradient descent learned landscapes; batch size 8192 on ImageNet.]{\textbf{Imagenet, batch size 8192}. The 2D projection of the DAL-$p$ and SGD learned landscapes, using the method of~\citet{li2018visualizing}. The DAL-$p$ model achieves an accuracy of 50\% which the SGD model achieves 42\% accuracy.}
\label{fig:DAL_p_imagenet_landscape}
\end{figure}

\begin{figure}
\begin{subfloat}[Fixed learning rate sweep.]{
  \includegraphics[width=0.45\columnwidth]{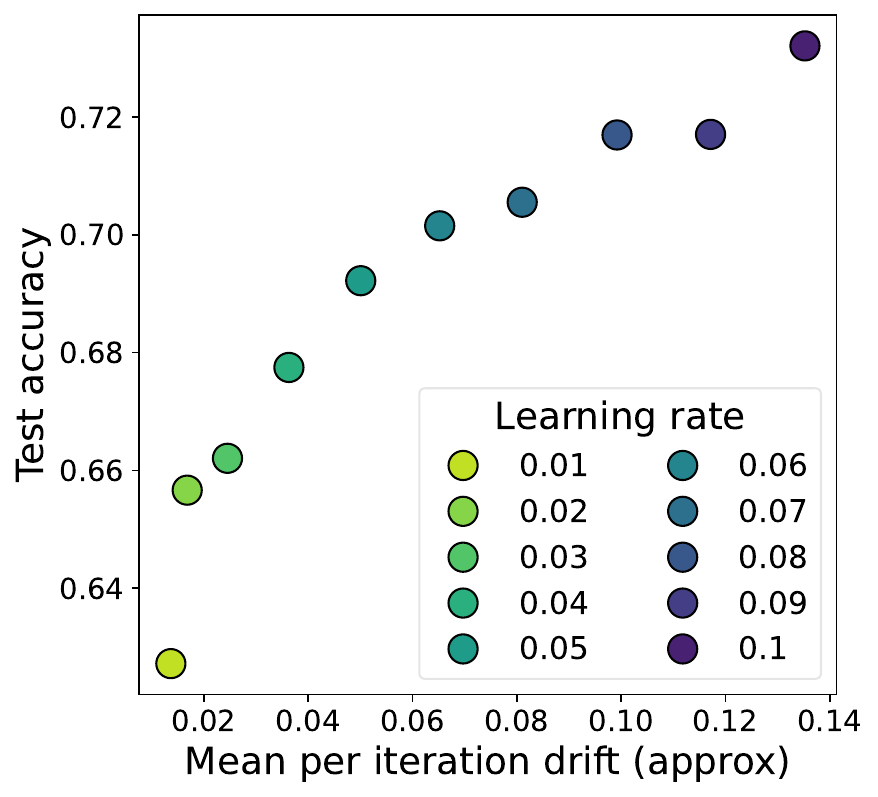}
  \includegraphics[width=0.45\columnwidth]{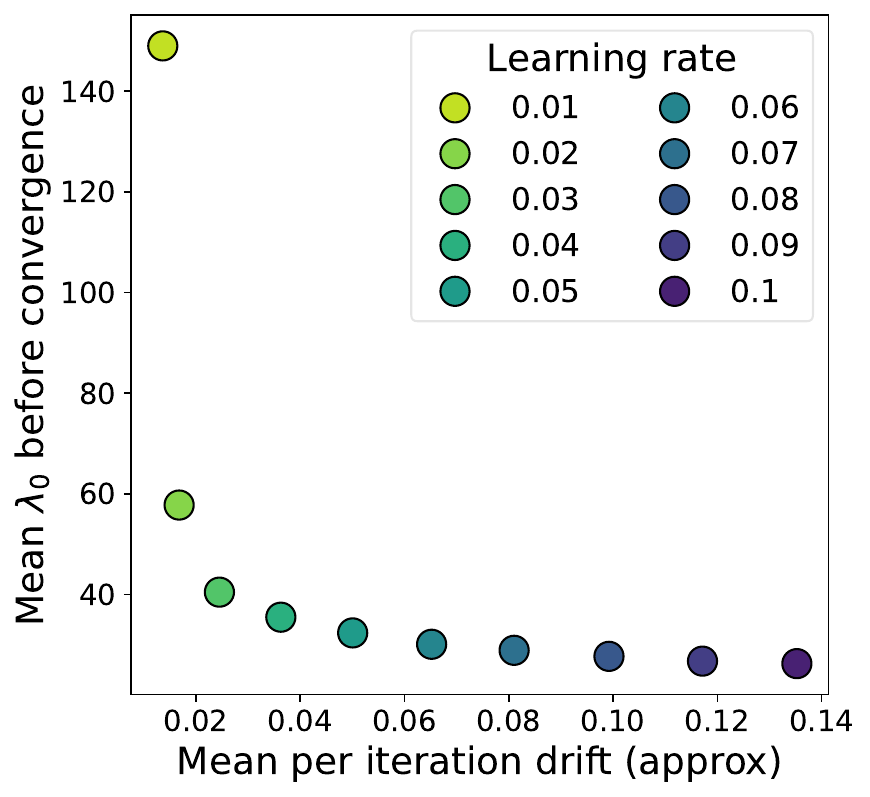}
}\end{subfloat}
\begin{subfloat}[DAL-$p$ sweep.]{
  \includegraphics[width=0.45\columnwidth]{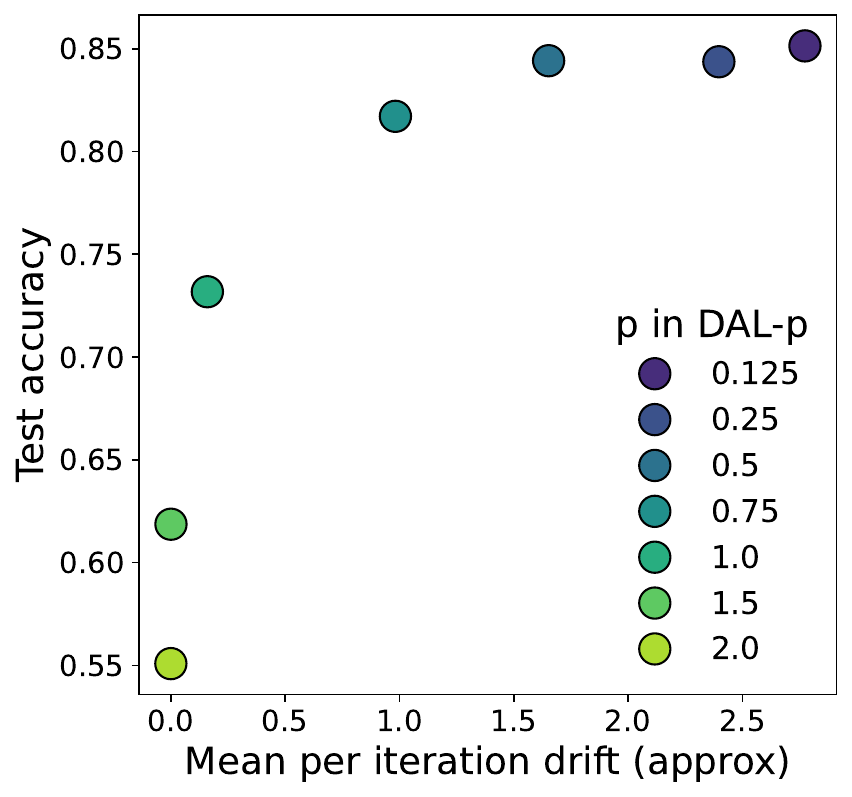}
  \includegraphics[width=0.45\columnwidth]{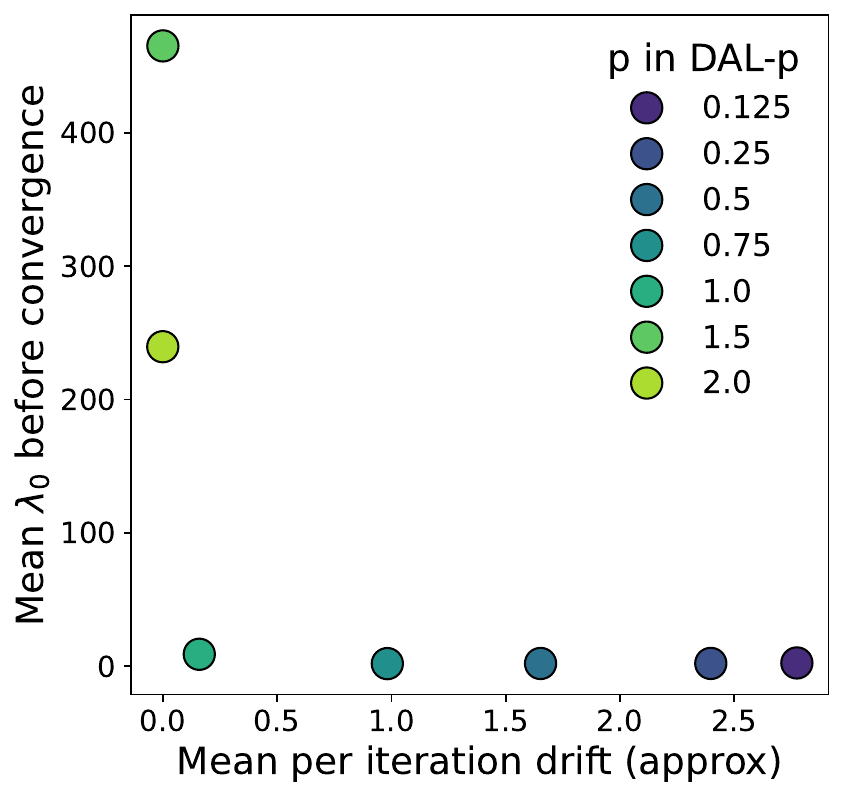}
}\end{subfloat}
\caption[VGG trained on CIFAR-10 with batch size 1024: connection between drift, test error and $\lambda_0$.]{VGG CIFAR-10 with batch size 1024: connection between drift, test error and $\lambda_0$. We observe the same patterns for other batch sizes, namely that the higher the drift the more generalisation and the lower $\lambda_0$, both with fixed learning rates and DAL.}
\label{fig:drift_lambda_sgd_connection}
\end{figure}

\noindent \textbf{Momentum}. We show additional results with learning rate adaptation and momentum in Figure \ref{fig:momentum_imagenet_1024}.

\begin{figure}[htb!]
\begin{subfloat}[Training loss.]{
 \includegraphics[width=0.45\columnwidth]{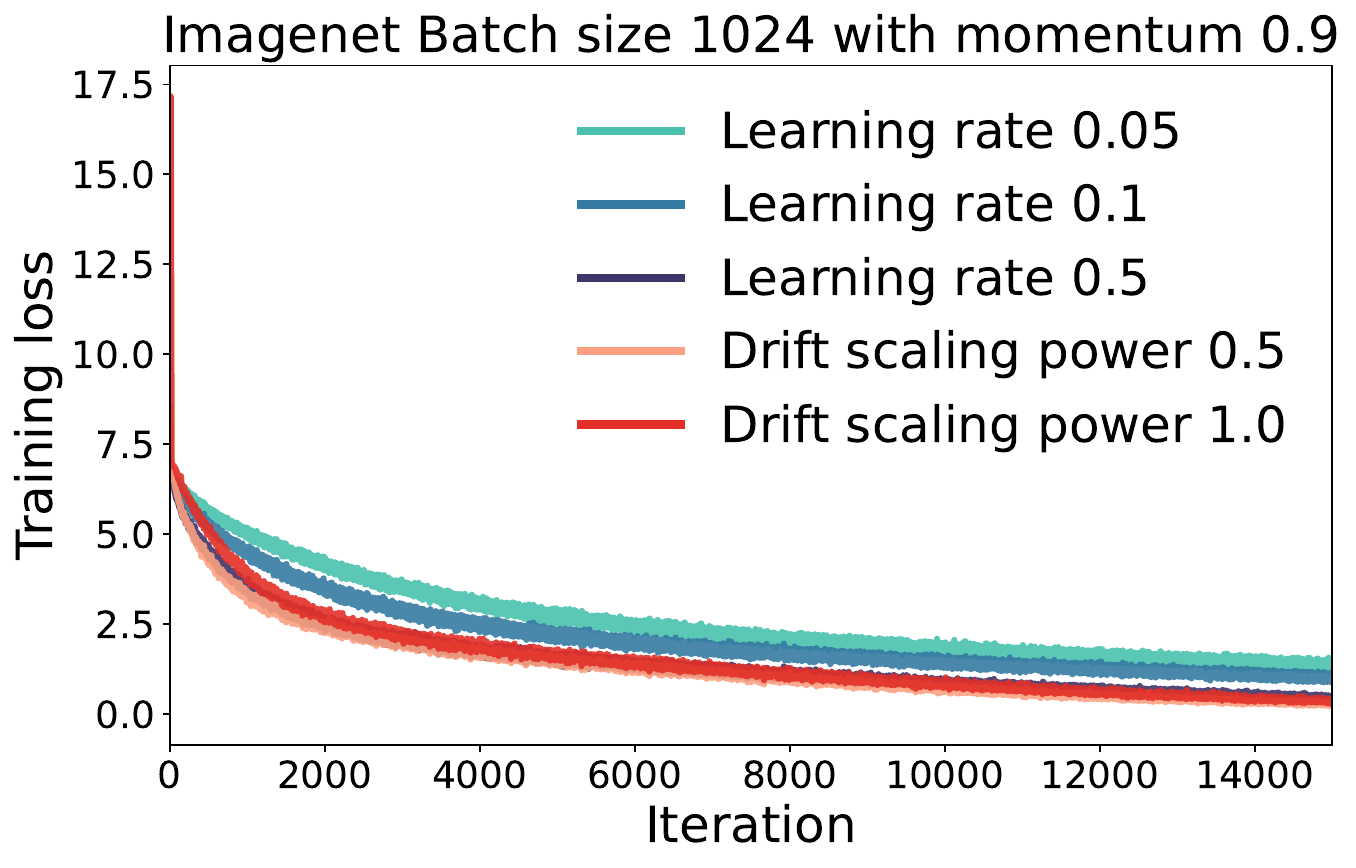}%
}
\end{subfloat}
\begin{subfloat}[Test accuracy.]{
  \includegraphics[width=0.45\columnwidth]{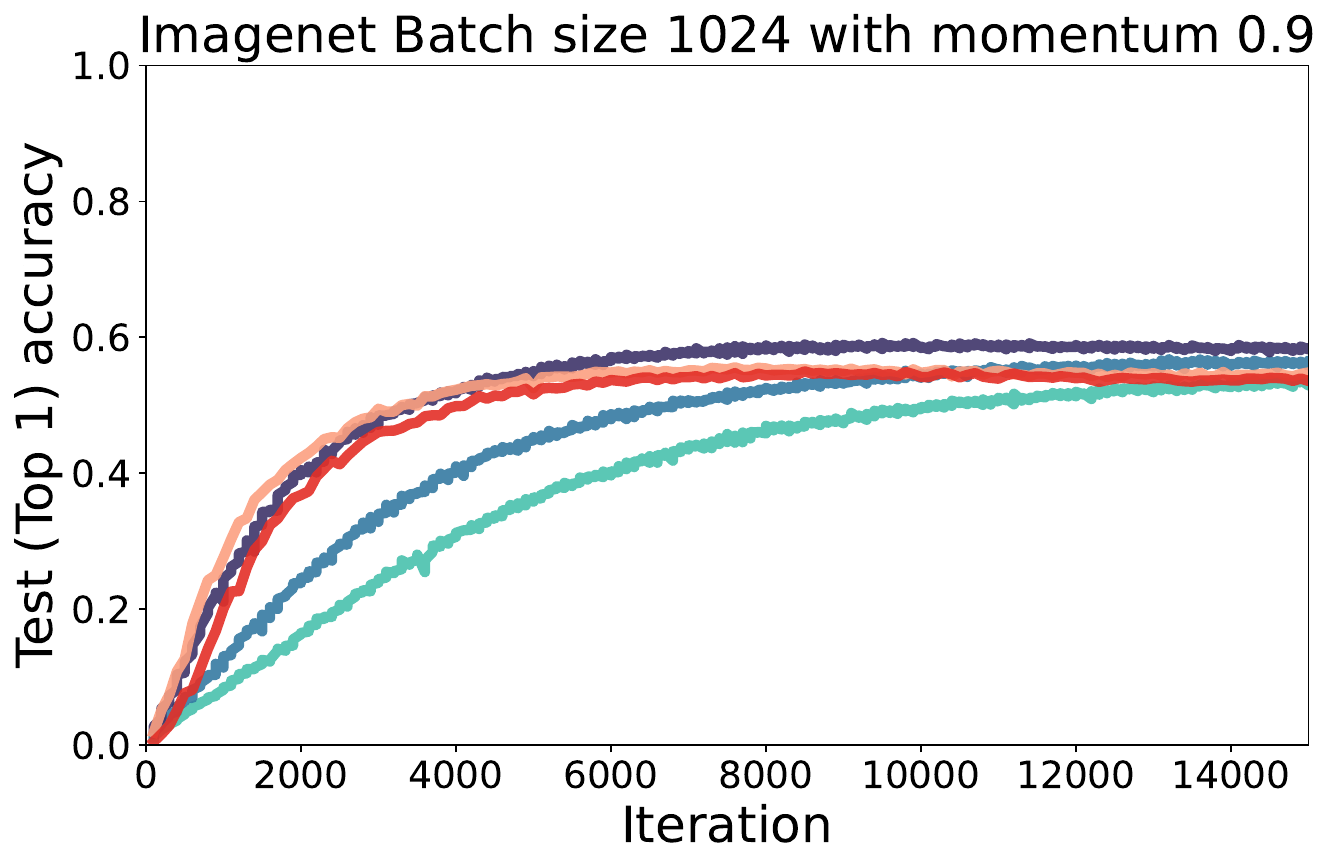}
}\end{subfloat}
\caption[DAL-$p$ with momentum $0.9$ on Imagenet. The model is Resnet-50 trained with batch size 1024.]{Integrating information on drift with momentum: with smaller batch sizes, we observe a smaller empirical gain. Since these are preliminary results, more investigation is needed into why that is. These results are consistent with the observation obtained with vanilla gradient descent, where we say larger gains with larger batch sizes on Imagenet.}
\label{fig:momentum_imagenet_1024}
\end{figure}

\noindent \textbf{Global discretisation error}. We show the global error in trajectory between the NGF and gradient descent in Figure \ref{fig:mnist_gradient_flow}. As previously observed \citep{cohen2021gradient}, gradient descent follows the NGF early in training, and the eigenvalues of the trajectory following the NGF keeps growing.

\begin{figure}[htb!]
\begin{subfloat}[$\norm{\vtheta(nh) - \vtheta_n}$]{
 \includegraphics[width=0.48\columnwidth]{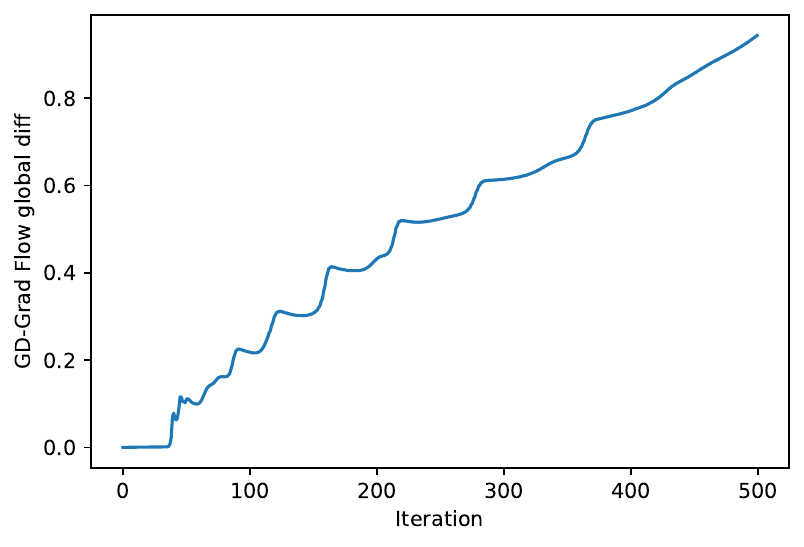}
 \label{fig:ngf_total_diff}
 }
\end{subfloat}
\begin{subfloat}[Leading eigenvalues $\lambda_i$ of the Hessian.]{
 \includegraphics[width=0.45\columnwidth]{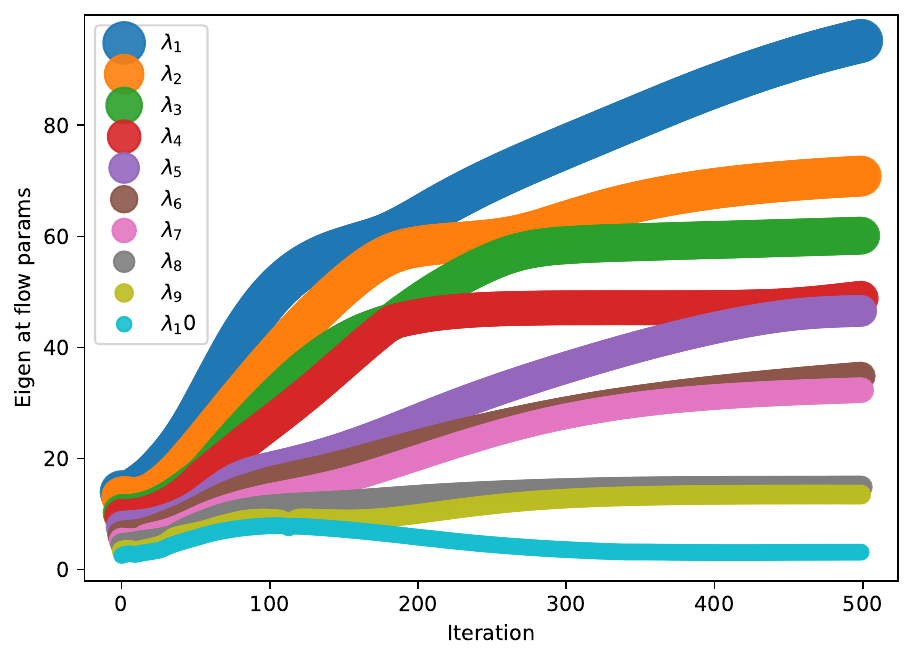}
 \label{fig:ngf_total_diff_lambda}
 }
\end{subfloat}
\caption[The global error in parameter space between the NGF and gradient descent; MNIST results.]{The global error in parameter space $\norm{\vtheta(nh) - \vtheta_n}$ between the NGF and gradient descent \subref{fig:ngf_total_diff}, as well as the behaviour of $\lambda_0$ of the NGF \subref{fig:ngf_total_diff_lambda}. We observe that early in training gradient descent follows the NGF, and then when following the NGF the leading eigenvalues grow.}
\label{fig:mnist_gradient_flow}
\end{figure}

\clearpage

\section{Figure reproduction details}
{\parindent0pt %
\begin{itemize}\item Figure \ref{fig:motivation_small_examples}: 2D convex plot. The loss function is $E = \vtheta^T A \vtheta$ with  $A = ((1, 0.0), (0.0, 0.01))$. The learning rate is 0.9.
\item Figure \ref{fig:intuition_convex1d_complex_part_needed}: 1D example. $E = \frac 1 2 (\vtheta - 0.6)^2$ and $h=1.2$
\item Figure \ref{fig:z_square}: $E(z) = \frac 1 2 z^2$. Learning rates are $0.8$, $1.5$ and $2.1$.
\item Figure \ref{fig:intuition_quadratic_2d}: Quadratic losses in 2 dimensions. $E= \vtheta^T A \vtheta$ with  $A = ((1, 0.0), (0.0, 0.01))$, with learning rates $0.5$, $0.9$ and $1.05$.
\item Figure \ref{fig:intuition_banana}: Banana function. Learning rates $0.0006$, $0.0017$ $0.005$.
\item Figure \ref{fig:intuition_nn_function_delta}: Error between gradient descent parameters and parameters obtained following continuous-time flows for multiple iterations on a small MLP: $\norm{\vtheta_{t+n} - \vtheta(nh)}$. Input size 2, MLP of with output units $10, 1$. Learning rates $0.1$, $0.2$ and $0.25$. We use a dataset with 5 examples where the input is generated from a Gaussian distribution with 2 dimensions and the targets are samples from a Gaussian distribution with 1 dimension. A mean square error loss is used.
\item Figure \ref{fig:prediction_grad_u}: Predictions of $\nabla_{\vtheta}E^T\vu_0$ using the PF on neural networks. Full-batch training with a VGG network on CIFAR-10. Learning rates $0.01$, $0.03$  and $0.05$.
\item Figure \ref{fig:reproduce_edge_of_stability}: Edge of stability on VGG networks on CIFAR-10. Full-batch training.
\item Figure \ref{fig:model_of_gd_at_edge_of_stability}: Edge of stability on a small 5 layer MLP with 10 units per layer on the UCI Iris dataset. Full-batch training. The learning rate used is 0.18.
\item Figure \ref{fig:edge_of_stability_results_local}: Edge of stability results. MNIST MLP with 4 layers. Learning rate $0.05$. For each model we approximate the NGF and the positive gradient flow initialised at the current gradient descent parameters and measure the value of the loss function and $\lambda_0$.
\item Figure \ref{fig:instabilities_short}: Using the PF and stability coefficient to understand instabilities in deep learning. MNIST results use a 4 layer MLP and a $0.05$ learning rate. CIFAR-10 MLP use a VGG network and a $0.02$ learning rate.
\item Figure \ref{fig:instabilities_resnet}: Understanding the behaviour of the loss through the PF and the behaviour of $\lambda_0$. Using a Resnet-18 trained on CIFAR-10 using a learning rate of $0.03$. For each model we approximate the NGF and the positive gradient flow initialised at the current gradient descent parameters and measure the value of the loss function and $\lambda_0$.
\item Figure \ref{fig:instabilities_change_learning_rate_loss}: Assessing whether 1 dimension is enough to cause instability in a continuous-time flow. We train a model with gradient descent until it reaches the edge of stability ($\lambda_0 \approx 2/h$), after which we use a approximate the continuous flow $\dot{\vtheta} = \nabla_{\vtheta}E^T \vu_0 \vu_0 + \sum_{i=1}^{D-1} -\nabla_{\vtheta}E^T \vu_i \vu_i$. The model is a VGG model trained on CIFAR-10 with learning rate $0.05$ and after the edge of stability is reached the flow is approximated using Euler steps of size $5 \times 10^{-5}$.
\item Figure \ref{fig:instability_largest_dot_prod}: The unstable dynamics of $\nabla_{\vtheta}E^T u$ in the edge of stability area. The model results are obtained from a VGG model trained on CIFAR-10 with a learning rate of $0.01$.
\item Figure \ref{fig:non_principal}: Learning rate sweep showing the value of the non-principal third-order term in training, on a Resnet-18 model trained on CIFAR-10. The coefficients used for the non-principal term are those in Eq~\eqref{eq:third_order_modified_vector_field}.
\item Figure \ref{fig:h_g_drift}: $||\nabla_{\vtheta}^2 E \nabla_{\vtheta} E||$ and the per-iteration drift as measured during training. The plots are obtained on an MNIST MLP with 4 layers and 100 units per layer, and a VGG model trained on CIFAR-10. The learning rates used are $0.05$, $0.01$ and $0.01$ respectively.
\item Figure \ref{fig:h_g_drift_spearman}: Correlation between $||\nabla_{\vtheta}^2 E \nabla_{\vtheta} E||$ and the iteration drift. The plots are obtained on an MNIST MLP with 4 layers and 100 units per layer, and a VGG model trained on CIFAR-10. Results are obtained over a learning rate sweep of 10 learning rates $0.01, 0.02, \dots, 0.1$.
\item Figure \ref{fig:dal}: Models trained using a learning rate sweep or DAL on CIFAR-10 and Imagenet. Models are VGG, Resnet-18 and Resnet-50 respectively.
\item Figure \ref{fig:l_r_scaling_quantities}: Key quantities in DAL versus fixed learning rate training: learning rate, and update norms. Results obtained on a Resnet-18 model trained on CIFAR-10.
\item Figure \ref{fig:l_r_scaling_sqrt}: DAL-0.5 results on a CIFAR-10 and Imagenet using  VGG, Resnet-18 and Resnet-50 model respectively.
\item Figure \ref{fig:power_sweeps}: DAL-$p$ sweep on full-batch training on CIFAR-10 with VGG and Resnet-18 models and a Resnet-50 model trained on Imagenet.
\item Figure \ref{fig:stability_lambda_performance}: VGG model trained on CIFAR-10 using full-batch gradient descent, either with a fixed learning rate in a sweep or a DAL-$p$ sweep. The drift per-iteration is approximated using $h^2/2 \nabla_{\vtheta}E^2 \nabla_{\vtheta}E$ or $h(\vtheta)^2/2 \nabla_{\vtheta}E^2 \nabla_{\vtheta}E$ (the adaptive learning rate) in case of DAL-p.
\item Figure \ref{fig:momentum_imagenet}: Resnet-50 results with $0.9$ momentum and a learning rate sweep or DAL. Models trained on Imagenet.
\item Figure \ref{fig:dal_cifar_full_batch}: CIFAR-10 DAL results with a Hessian vector product computation of $\nabla_{\vtheta}^2 E \nabla_{\vtheta} E$ compared to an approximation. We used $\epsilon = 0.01$ in the approximation. Full-batch training. 
\item Figure \ref{fig:dal_cifar_sgd}: CIFAR-10 DAL results with a Hessian vector product computation of $\nabla_{\vtheta}^2 E \nabla_{\vtheta} E$ compared to an approximation. We used $\epsilon = 0.01$ in the approximation. Batch size 512. 
\item Figure \ref{fig:dal_approx_comp}: Imagenet DAL results with a Hessian vector product computation of $\nabla_{\vtheta}^2 E \nabla_{\vtheta} E$ compared to an approximation. We used $\epsilon = 0.01$ in the approximation. 
\item Figure \ref{fig:dal_0_5_approx_comp}: Imagenet DAL-$0.5$ results with a Hessian vector product computation of $\nabla_{\vtheta}^2 E \nabla_{\vtheta} E$ compared to an approximation. We used $\epsilon = 0.01$ in the approximation. 
\item Figure \ref{fig:1d_cosine}: Results with a non quadratic function, including $\cos$ and $\sin$. The function used is: $E(\theta) = \cos(\theta) + \theta, \text{if } \theta < 0; (\theta/3)^22 + 1 + \theta/3)$ otherwise. 
\item Figure \ref{fig:validating_bea}: Per-iteration error between the NGF, IGR and PF and gradient descent. MNIST learning rate $0.05$. CIFAR learning rate $0.02$. We obtain the results by training a model with gradient descent $\vtheta_t = \vtheta_{t-1} - h \nabla_{\vtheta} E(\vtheta_{t-1})$ and at each iteration $t$ of gradient descent approximating the respective flows for time $h$ and computing the difference in norms between the resulting flow parameters and the gradient descent parameters at the next iteration: $\norm{\vtheta_t - \vtheta(h)}$ with $\vtheta(h) = \vtheta_{t-1}$. Results obtained on full-batch training.
\item Figure \ref{fig:breast_cancer_principal_flow}: Global parameter error for IGR ad PF flows compared  on an MLP trained on the UCI breast cancer dataset. MLP with 10 hidden units and 1 output unit. Learning rates $1.5$ and $5.$ respectively.
\item Figure \ref{fig:stability_analysis_nn}: We train a VGG model on CIFAR-10 with learning rate $0.1$, and as it reaches converges increase the learning rate slightly so that $\lambda_0 > 2/h$. We use a square loss here to avoid training the decrease in eigenvalues later in training associated with cross entropy losses. 
\item Figure \ref{fig:eigspectrum_small_mlp}: The eigenspectrum obtained along the gradient descent path of a 5 layer small MLP trained on the UCI Iris dataset. The Hessian eigenvalues are largely positive. The result accompanies Figure~\ref{fig:model_of_gd_at_edge_of_stability} in the main thesis.
\item Figure \ref{fig:instabilities_change_learning_rate_discrete}: Training a VGG model on CIFAR-10 and training with learning rate $0.05$ after which switching to $0.01$.
\item Figure \ref{fig:mnist_continuous_time_swap}: MNIST results when changing one direction in continuous-time. The model is a 4 layer MLP with 100 hidden units per layer. The model was trained with learning rate $0.05$ until the edge of stability area is reached after which the flow $\dot{\vtheta} = \nabla_{\vtheta}E^T \vu_0 \vu_0 + \sum_{i=1}^{D-1} -\nabla_{\vtheta}E^T \vu_i \vu_i$ is used.
\item Figure \ref{fig:cifar_peak_commons}: Peak areas of loss increase and the stability coefficient for the largest eigendirection of the PF. Results using a VGG model trained with full-batch gradient descent on CIFAR-10 across a learning rate sweep.
\item Figure \ref{fig:edge_of_stability_results_lambda}: Understanding changes to $\lambda_0$ using the PF. MLP with 4 layers with 100 units each trained on MNIST, and VGG model trained on CIFAR-10. Learning rate $0.05$ and $0.01$ respectively.
\item Figure \ref{fig:edge_of_stability_results_cifar_resnet}: The connection between the PF, $\lambda_0$ and loss instabilities with Resnet-18 trained on CIFAR-10 with a learning rate $0.01$.
\item Figure \ref{fig:connections_quantities}: Evaluating the value of the DAL learning rate for a vanilla (fixed) learning rate sweep to assess whether it would have reasonable magnitudes. Models trained on CIFAR-10 using a learning rate sweep.
\item Figure \ref{fig:power_sweeps_batch_sizes_vgg}: DAL-$p$ sweep with a VGG model trained on CIFAR-10 for small and large batch sizes.
\item Figure \ref{fig:imagenet_lr_scaling_across_batch_sizes}: DAL on Imagenet, a learning rate sweep across batch sizes.
\item Figure \ref{fig:imagenet_lr_sqrt_scaling_across_batch_sizes}: DAL$-0.5$: on Imagenet, a learning rate sweep across batch sizes.
\item Figure \ref{fig:power_sweeps_all}: DAL-$p$ sweep on VGG, Resnet-18 and Imagenet.
\item Figure \ref{fig:least_square_loss_dal}: DAL results with a least square loss. Full-batch training of a VGG model on CIFAR-10.

\item Figure \ref{fig:imagenet_lr_scaling_per_parameter}: Resnet-50 Imagenet training with a learning rate per parameter. Batch size 8192.
\item Figure \ref{fig:drift_lambda_sgd_connection}: VGG trained on CIFAR-10 with batch size 1024: connection between drift, test error and $\lambda_0$.
\item Figure \ref{fig:momentum_imagenet_1024}: DAL-$p$ with momentum $0.9$ on Imagenet. The model is Resnet-50 trained with batch size 1024.
\item Figure \ref{fig:mnist_gradient_flow}: MNIST results with a 4 layer MLP with 100 hidden units. The learning rate used is $0.05$.
\end{itemize}
}%

\chapter{Continuous time models of gradient descent in two-player games}

\section{Proof of the main theorems}
\label{sec:dd_gan_proofs}

\textbf{Notation}: We use $\vphi \in \mathbb{R}^m$ to denote the parameters of the first player
and $\vtheta \in \mathbb{R}^n$ for the second player. We assume that the players parameters are updated simultaneously
with learning rates $\lrp h$ and $\lrt h$ respectively.
We consider the vector fields $f(\vphi, \vtheta): \mathbb{R}^m \times \mathbb{R}^n \rightarrow \mathbb{R}^m$ and
$g(\vphi, \vtheta): \mathbb{R}^m \times \mathbb{R}^n \rightarrow \mathbb{R}^n$. 
The Jacobian $\jacthetaf(\vphi, \vtheta) \in \mathbb{R}^{n, m}$ denotes the matrix with entries $\big[\jacthetaf(\vphi, \vtheta)\big]_{i, j} = \frac{d f_j}{d \vtheta_{i}}$ with $i \in \{1, ..., n\}, j \in\{1, ..., m\}$. 

We now prove Theorem~\ref{thm:sim} and Theorem~\ref{thm:alt} in the main thesis, corresponding to the \textit{simultaneous} Euler updates and to the \textit{alternating} Euler updates, respectively.
In both cases, our goal is to find corrections $f_1$ and $g_1$ to the original system
\begin{align}
 \dot{\vphi} &=  f( \vphi, \vtheta), \label{eq:ode-1}\\
 \dot{\vtheta}  &= g( \vphi, \vtheta), \label{eq:ode-2}
\end{align}
such that the modified continuous system
\begin{align}
 \dot{\vphi} &=  f( \vphi, \vtheta)  + h f_1( \vphi, \vtheta), \\
 \dot{\vtheta}  &= g( \vphi, \vtheta) + h g_1( \vphi, \vtheta),
\end{align}
follows the discrete steps of the method with a local error of order $\mathcal O(h^3)$.
More precisely, if $(\vphi_{t}, \vtheta_{t})$ denotes the discrete step of the method at time $t$, and $(\vphi(h), \vtheta(h))$ corresponds to the continuous solution of the modified system above starting at $(\vphi_{t-1}, \vtheta_{t-1})$, we want that the local errors for both players, i.e.,
$$
\| \vphi_{t} - \vphi(\lrp h) \|
\qquad \textrm{ and } \qquad
\| \vtheta_{t} - \vtheta(\lrt h) \|
$$
to be of order $\mathcal O(h^3)$. In the expression above, $\lrp h$ and $\lrt h$ are the effective learning rates (or step-sizes) for the first and the second player respectively for both the simultaneous and alternating Euler updates.

Our proofs from BEA follow the same steps (see Section~\ref{remark:bea_proofs} for an overview):
\begin{enumerate}
  \item Expand the discrete updates to find a relationship between $\vphi_t$ and $\vphi_{t-1}$ and $\vtheta_t$ and $\vtheta_{t-1}$ up to order $\mathcal{O}(h^2)$.
  \item Expand the changes in continuous-time of the modified flow given by BEA.
  \item Find the first order Discretisation Drift (DD) by matching the discrete and continuous updates up to second order in learning rates.
\end{enumerate}

\textbf{Notation}: To avoid cluttered notations, we use $f_{(t)}$ to denote the $f(\vphi_t, \vtheta_t)$ and $g_{(t)}$ to denote $g(\vphi_t, \vtheta_t)$ for all $t$. If no index is specified, we use $f$ to denote $f(\vphi, \vtheta)$, where $\vphi$ and $\vtheta$ are variables in a continuous system.

\subsection{Simultaneous updates (Theorem~\ref{thm:sim})}
\label{sec:sim_updates}

Here we prove Theorem~\ref{thm:sim} in the main thesis, which we reproduce here:

The simultaneous Euler updates with learning rates $\lrp h$ and $\lrt h$ respectively are given by:
\begin{align}
\vphi_t &= \vphi_{t-1} + \lrp h f(\vphi_{t-1}, \vtheta_{t-1} ) \label{eq:supp_simup1}\\
\vtheta_{t} &= \vtheta_{t-1} + \lrt h g(\vphi_{t-1},\vtheta_{t-1}) \label{eq:supp_simup2}
\end{align}
\begin{customthm}{(DD for simultaneous Euler updates)} \label{thm:supp_2_players_sim} (Theorem~\ref{thm:sim}) The discrete \emph{simultaneous} Euler updates
in \eqref{eq:supp_simup1} and \eqref{eq:supp_simup2} follow the continuous system
\label{thm:supp_sim}
\begin{align}
\dot{\vphi} &= f - \frac{\lrp h}{2} \left(\jacphif f + \jacthetaf g \right) \\
\dot{\vtheta} &= g - \frac{\lrt h}{2} \left(\jacphig f + \jacthetag g \right)
\end{align}
with an error of size $\mathcal O(h^3)$  after one update step.
\end{customthm}

\textbf{Step 1: Expand the updates per player via a Taylor expansion.} \\
We expand the numerical scheme to find a relationship between $\vphi_t$ and $\vphi_{t-1}$ and $\vtheta_t$ and $\vtheta_{t-1}$ up to order $\mathcal{O}(h^2)$.
In the case of the simultaneous Euler updates, this does not require any change to Eqs~\eqref{eq:supp_simup1} and~\eqref{eq:supp_simup2}.

\textbf{Step 2: Expand the continuous-time changes for the modified flow given by BEA.} 

\begin{lemma}
If:
\begin{align}
\dot{\vphi} = \tilde{f}(\vphi, \vtheta) \\
\dot{\vtheta} = \tilde{g}(\vphi, \vtheta)
\end{align}
where
\begin{align}
\tilde{f}(\vphi, \vtheta) = f(\vphi, \vtheta) + \sum_{i=1} {\tau_\phi}^i f_i (\vphi, \vtheta) \\
\tilde{g}(\vphi, \vtheta) = g(\vphi, \vtheta) + \sum_{i=1} {\tau_\theta}^i g_i (\vphi, \vtheta)
\end{align}
and $\tau_{\theta}$ and $\tau_{\phi}$ are scalars. 
Then---for ease of notation, we drop the argument $\tph$ on the evaluations of $\vphi$ and $\theta$ on the RHS:
\begin{align}
\vphi(\tph + \tau_\phi) =  \vphi(\tph) + \tau_\phi f + \tau_\phi^2 f_1 + \tau_\phi^2\frac{1}{2} \jacphif f + \tau_\phi^2\frac{1}{2} \jacthetaf g + \mathcal{O}(\tau_\phi^3)
\end{align}
\label{thm:cont}
\end{lemma}

\begin{proof}
We perform a Taylor expansion:
\begin{align}
\vphi(&\tph + \tau_\phi) \\& = \vphi + \tau_\phi\dot{\vphi} + \tau_\phi^2\frac{1}{2}\ddot{\vphi} + \mathcal{O}(\tau_\phi^3)\\
               &= \vphi + \tau_\phi \tilde{f} + \tau_\phi^2\frac{1}{2} \dot{\tilde{f}} +  \mathcal{O}(\tau_\phi^3) \\
               &= \vphi + \tau_\phi \tilde{f} + \tau_\phi^2\frac{1}{2} \left(\jacparam{\vphi}{\tilde{f}} \tilde{f}  + \jacparam{\vtheta}{\tilde{f}} \tilde{g} \right)  + \mathcal{O}(\tau_\phi^3)\\
              &= \vphi + \tau_\phi (f + \tau_\phi f_1 + \mathcal{O}(\tau_\phi^2))+ \tau_\phi^2\frac{1}{2} \left(\jacparam{\vphi}{\tilde{f}} \tilde{f}  + \jacparam{\vtheta}{\tilde{f}} \tilde{g} \right)  + \mathcal{O}(\tau_\phi^3)\\
              &= \vphi + \tau_\phi f + \tau_\phi^2 f_1 + \tau_\phi^2\frac{1}{2} \left(\jacparam{\vphi}{\tilde{f}} \tilde{f}  + \jacparam{\vtheta}{\tilde{f}} \tilde{g} \right)  + \mathcal{O}(\tau_\phi^3)\\
              &= \vphi + \tau_\phi f + \tau_\phi^2 f_1  \\ & \quad+\tau_\phi^2\frac{1}{2} \left(\jacparam{\vphi}{\tilde{f}} \left(f + \tau_\phi f_1 + \mathcal{O}(\tau_\phi^2)\right)  + \jacparam{\vtheta}{\tilde{f}}\left(g + \tau_\theta g_1 + \mathcal{O}(\tau_\theta^2)\right)\right)  + \mathcal{O}(\tau_\phi^3)\\
              &= \vphi + \tau_\phi f + \tau_\phi^2 f_1 + \tau_\phi^2\frac{1}{2} \jacparam{\vphi}{\tilde{f}} f + \tau_\phi^2\frac{1}{2} \jacparam{\vtheta}{\tilde{f}} g + \mathcal{O}(\tau_\phi^3)\\
              &= \vphi + \tau_\phi f + \tau_\phi^2 f_1 + \tau_\phi^2\frac{1}{2} \jacphif f + \tau_\phi^2\frac{1}{2} \jacthetaf g + \mathcal{O}(\tau_\phi^3)
\end{align}
where we assumed that $\tau_\phi$ and $\tau_\theta$ are in the same order of magnitude.
\end{proof}

\textbf{Step 3: Matching the discrete and modified continuous updates.} \\
We substitute the values given by the discrete updates, namely $\vphi_{t-1}$ and $\vtheta_{t-1}$, in Lemma~\ref{thm:cont} in order to calculate the displacement according to the continuous updates:
\begingroup
\begin{align}
\vphi(\tph + \tau_\phi) &= \vphi_{t-1} + \tau_\phi f_{(t-1)} + \tau_\phi^2 f_1(\vphi_{t-1}, \vtheta_{t-1})  \\
    & \hspace{2.2em}+ \frac{1}{2} \tau_\phi^2   \evaljacgames{\vphi}{f}{{(t-1)}} f_{(t-1)} + \frac{1}{2}\tau_\phi^2  \evaljacgames{\vtheta}{f}{{(t-1)}} g_{(t-1)} + \mathcal{O}(\tau_\phi^3) \\
\label{eq:sim_ode_phi} 
\vtheta(\tph + \tau_\theta) &= \vtheta_{t-1} + \tau_\theta g_{(t-1)} + \tau_\theta^2 g_1(\vphi_{t-1}, \vtheta_{t-1}) \\
  & \hspace{2.2em} + \frac{1}{2} \tau_\theta^2  \evaljacgames{\vphi}{g}{{(t-1)}} f_{(t-1)} + \frac{1}{2}\tau_\theta^2  \evaljacgames{\vtheta}{g}{{(t-1)}} g_{(t-1)} + \mathcal{O}(\tau_\theta^3)
\label{eq:sim_ode_theta}
\end{align}
\endgroup
In order to find $f_1$  and $g_1$ such that the continuous dynamics of the modified updates $f + \lrp h f_1$
and $g + \lrt h g_1$ match the discrete updates in Eqs~\eqref{eq:supp_simup1} and~\eqref{eq:supp_simup2}, we look for the corresponding continuous increments of the discrete updates in the modified continuous system, such that $\norm{\vphi(\tph + \tau_\phi)   - \vphi_t }$ and $\norm{\vtheta(\tph + \tau_\theta)   - \vtheta_t }$ are $\mathcal{O}(h^3)$.
 The first order terms in Eqs~\eqref{eq:supp_simup1} and~\eqref{eq:supp_simup2} and  Eqs~\eqref{eq:sim_ode_phi} and~\eqref{eq:sim_ode_theta} suggest that:
\begin{align}
\lrp h = \tau_{\phi} \\
\lrt h = \tau_{\theta}
\end{align}
We can now proceed to find $f_1$ and $g_1$ from the second order terms
\begin{align}
  & 0  = \lrp^2 h^2 f_1(\vphi_{t-1}, \vtheta_{t-1}) + \frac{1}{2}  \lrp^2 h^2 \evaljacgames{\vphi}{f}{{(t-1)}} f_{(t-1)}  + \frac{1}{2} \lrp^2 h^2  \evaljacgames{\vtheta}{f}{{(t-1)}} g_{(t-1)} \\
  & f_1(\vphi_{t-1}, \vtheta_{t-1}) = - \frac{1}{2}  \evaljacgames{\vphi}{f}{{(t-1)}} f_{(t-1)} - \frac{1}{2}  \evaljacgames{\vtheta}{f}{{(t-1)}} g_{(t-1)}.
\end{align}
Similarly, for $g_1$ we obtain
\begin{align}
g_1(\vphi_{t-1}, \vtheta_{t-1}) &= - \frac{1}{2} \evaljacgames{\vphi}{g}{{(t-1)}} f_{(t-1)} - \frac{1}{2}  \evaljacgames{\vtheta}{g}{{(t-1)}} g_{(t-1)}.
\end{align}
Leading to the first order corrections
\begin{align}
f_1(\vphi_{t-1}, \vtheta_{t-1}) &= - \frac{1}{2}  \evaljacgames{\vphi}{f}{{(t-1)}} f_{(t-1)} - \frac{1}{2}  \evaljacgames{\vtheta}{f}{{(t-1)}} g_{(t-1)} \\
g_1(\vphi_{t-1}, \vtheta_{t-1}) &= - \frac{1}{2} \evaljacgames{\vphi}{g}{{(t-1)}} f_{(t-1)} - \frac{1}{2}  \evaljacgames{\vtheta}{g}{{(t-1)}} g_{(t-1)}.
\label{eq:sim_geneal_eq}
\end{align}
We have found the functions $f_1$ and $g_1$ such that after one discrete optimisation step the flows $\dot{\vphi} = f + \lrp h f_1$ and $\dot{\vtheta} = g + \lrt h g_1$ follow the discrete updates up to order $\mathcal{O}(h^3)$, finishing the proof.

\subsection{Alternating updates (Theorem~\ref{thm:alt})}
\label{sec:multiple_updates}

Here we prove Theorem~\ref{thm:alt} in the main thesis, which we reproduce here.
For the \textit{alternating Euler updates}, the players take turns to update their parameters, and can perform multiple updates each. We denote the number of alternating updates of the first  player (resp. second player) by $m$ (resp. $k$).
We scale the learning rates by the number of updates, leading to the following updates $ \vphi_{t} := \vphi_{m,t} $ and $\vtheta_{t} := \vtheta_{k, t}$ where
\begin{align}
 \vphi_{i, t} &=  \vphi_{i-1, t} + \frac{\lrp h}{m}   f(\vphi_{i-1,t}, \vtheta_{t-1}) , \hspace{1em} i = 1 \dots m, \label{eq:supp_altup1} \\
 \vtheta_{j, t} &=  \vtheta_{j-1, t} + \frac{\lrt h}{k} g(\vphi_{m, t}, \vtheta_{j-1, t}), \hspace{1em} j = 1 \dots k. \label{eq:supp_altup2}
\end{align}
\begin{customthm}{DD for alternating Euler updates} \label{thm:supp_2_players_alt} (Theorem~\ref{thm:alt})
The discrete \emph{alternating} Euler updates in \eqref{eq:supp_altup1} and \eqref{eq:supp_altup2} follow the continuous system
\begin{align}
\dot{\vphi} &= f - \frac{\lrp h}{2} \left(\frac{1}{m}\jacphif f + \jacthetaf g \right) \\
\dot{\vtheta} &= g - \frac{\lrt h}{2} \left((1- \frac {2\lrp} {\lrt})\jacphig f + \frac{1}{k} \jacthetag g \right)
\end{align}
with an error of size $\mathcal O(h^3)$ after one update step.
\end{customthm}

\noindent \textbf{Step 1: Discrete updates} \\
In the case of alternating Euler discrete updates, we have
\begingroup
\allowdisplaybreaks
\begin{align}
 \vphi_{1,t} &=  \vphi_{t-1} + \frac{\lrp}{m} h f(\vphi_{t-1}, \vtheta_{t-1}) \\
 \dots\\
 \vphi_{m,t} &=  \vphi_{m-1,t} + \frac{\lrp}{m} h f(\vphi_{m-1, t}, \vtheta_{t-1}) \\
 \vtheta_{1,t} &=  \vtheta_{t-1} + \frac{\lrt}{k} h g(\textcolor{black}{\vphi_{m,t}}, \vtheta_{t-1}) \\
 \dots \\
 \vtheta_{k,t} &=  \vtheta_{k-1, t-1} + \frac{\lrt}{k} h g(\vphi_{m,t}, \vtheta_{k-1, t}) \\
 \vphi_t &= \vphi_{m,t} \\
 \vtheta_t &= \vtheta_{k,t}.
\end{align}
\endgroup

\begin{lemma}
Given updates $\vphi_{m,t} = \vphi_{m-1, t} + h f(\vphi_{m-1, t}, \vtheta_{t-1})$, we can write:
\begin{align}
 \vphi_{m,t} &= \vphi_{t-1} + mh f_{(t-1)} + \frac{m (m-1)}{2}h^2 \evaljacgames{\vphi}{f}{{(t-1)}} f_{(t-1)}  + \mathcal{O}(h^3)
\end{align}

\begin{proof}
Proof by induction.
\label{thm:discrete}
Base step.
\begin{align}
 \vphi_{2,t} &=  \vphi_{1,t} + h f(\vphi_{1,t}, \vtheta_{t-1}) \\
            &=  \vphi_{t-1} +  h f(\vphi_{t-1}, \vtheta_{t-1}) + h f(\vphi_{1,t}, \vtheta_{t-1}) \\
             &= \vphi_{t-1} + h f_{(t-1)} + h f\big(\vphi_{t-1} + h f_{(t-1)}, \vtheta_{t-1} \big) \\
             &= \vphi_{t-1} + h f_{(t-1)} + h \left(f_{(t-1)} + h  \evaljacgames{\vphi}{f}{{(t-1)}} f_{(t-1)} + \mathcal{O}(h^2) \right) \\
             &= \vphi_{t-1} + 2h f_{(t-1)} + h^2  \evaljacgames{\vphi}{f}{{(t-1)}} f_{(t-1)} + \mathcal{O}(h^3) 
\end{align}
Inductive step:
\vspace{-1em}
\begin{align}
\vphi_{m+1, t} &= \vphi_{m, t} + h f(\vphi_{m, t}, \vtheta_{t-1}) \\
                 &= \vphi_{t-1} + mh f_{(t-1)} + \frac{m (m-1)}{2}h^2  \evaljacgames{\vphi}{f}{{(t-1)}} f_{(t-1)} \\
                 &\quad + h f\Big(\vphi_{t-1} + mh f_{(t-1)} + \frac{m (m-1)}{2}h^2  \evaljacgames{\vphi}{f}{{(t-1)}} f_{(t-1)} + \mathcal{O}(h^3), \vtheta_{t-1}\Big) \nonumber\\&\quad+ \mathcal{O}(h^3) \\
                &= \vphi_{t-1} + mh f_{(t-1)} + \frac{m (m-1)}{2}h^2  \evaljacgames{\vphi}{f}{{(t-1)}} f_{(t-1)}\\
                &\quad + h \left(f_{(t-1)} + \evaljacgames{\vphi}{f}{{(t-1)}} \left(mh f_{(t-1)} + \frac{m (m-1)}{2}h^2  \evaljacgames{\vphi}{f}{{(t-1)}} f_{(t-1)}\right) \right) \nonumber\\&\quad+ \mathcal{O}(h^3) \\
 &= \vphi_{t-1} + (m +1)h f_{(t-1)} + \frac{m (m-1)}{2}h^2  \evaljacgames{\vphi}{f}{{(t-1)}} f_{(t-1)} \\
                &\quad + h \evaljacgames{\vphi}{f}{{(t-1)}} \left(mh f_{(t-1)} + \frac{m (m-1)}{2}h^2  \evaljacgames{\vphi}{f}{{(t-1)}} f_{(t-1)})\right)+ \mathcal{O}(h^3) \\
                &= \vphi_{t-1} + (m +1)h f_{(t-1)} + \frac{m (m-1)}{2}h^2  \evaljacgames{\vphi}{f}{{(t-1)}} f_{(t-1)} \\
                &\quad + h^2 m \evaljacgames{\vphi}{f}{{(t-1)}} f_{(t-1)}+ \mathcal{O}(h^3) \\
                 &= \vphi_{t-1} + (m +1)h f_{(t-1)} + \frac{m (m+1)}{2}h^2  \evaljacgames{\vphi}{f}{{(t-1)}} f_{(t-1)} + \mathcal{O}(h^3)
\end{align}
\end{proof}
\end{lemma}

From Lemma~\ref{thm:discrete} with  $h = \lrp h/m$ we have that
\begin{align}
 \vphi_{m,t} &= \phi_{t-1} + \lrp h f_{(t-1)} + \frac{m-1}{2m} \lrp^2 h^2  \evaljacgames{\vphi}{f}{{(t-1)}} f_{(t-1)}  + \mathcal{O}(h^3).
\end{align}
We now turn our attention to the second player.
We define $g_{\miditerationindex} = g(\vphi_{m, t}, \vtheta_{t-1} )$.
This is captures the difference between simultaneous and alternating updates.
From Lemma~\ref{thm:discrete} with $h = \lrt h/k$ we have that
\begin{align}
 \vtheta_{k, t} &= \vtheta_{t-1} + \lrt h 
 g_{\miditerationindex} + \frac{(k - 1)}{2k}\lrt^2 h^2 \evaljacgames{\vtheta}{g}{\miditerationindex} g_{\miditerationindex} + \mathcal{O}(h^3).
 \label{eq:sup_alt_theta_exp}
\end{align}
We now expand $g_{\miditerationindex}$ by Taylor expansion:
\begin{align}
g_{\miditerationindex} &= g(\vphi_{m, t}, \vtheta_{t-1}) \\
     &= g(\vphi_{t-1} +  \lrp h f_{(t-1)} +  \lrp^2 \frac{m-1}{2m}h^2  \evaljacgames{\vphi}{f}{{(t-1)}} f_{(t-1)} + \mathcal{O}(h^3), \vtheta_{t-1}) \\
     &= g_{(t-1)} + \evaljacgames{\vphi}{g}{{(t-1)}}\left( \lrp h f_{(t-1)} + \frac{(m -1)}{2m} \lrp^2 h^2  \evaljacgames{\vphi}{f}{{(t-1)}} f_{(t-1)} + \mathcal{O}(h^3)\right) 
\label{eq:mid_g}
\end{align}
If we replace the above in Eq~\eqref{eq:sup_alt_theta_exp}:
\begingroup
\begin{align}
 &\vtheta_{k, t} = \vtheta_{t-1} + \lrt h g_{\miditerationindex} + \frac{k-1}{2k}\lrt^2 h^2  \evaljacgames{\vtheta}{g}{\miditerationindex} g_{\miditerationindex} + \mathcal{O}(h^3) \\
                 &= \vtheta_{t-1} + \lrt h \left(g_{(t-1)} + \evaljacgames{\vphi}{g}{{(t-1)}} \left(\lrp h f_{(t-1)} + \frac{m -1}{2m}\lrp^2 h^2  \evaljacgames{\vphi}{f}{{(t-1)}} f_{(t-1)}\right)  \right)
                 \\ & \hspace{1.2em} +\frac{k-1}{2k}\lrt^2 h^2 \evaljacgames{\vtheta}{g}{\miditerationindex} g_{\miditerationindex} + \mathcal{O}(h^3) \\
                 &= \vtheta_{t-1} + \lrt h g_{(t-1)} + \lrt  h \evaljacgames{\vphi}{g}{{(t-1)}} \left(\lrp h f_{(t-1)} + \frac{m -1}{2m}\lrp^2 h^2  \evaljacgames{\vphi}{f}{{(t-1)}} f_{(t-1)} \right) 
                 \\ & \hspace{1.2em} + \frac{k-1}{2k}\lrt^2 h^2  \evaljacgames{\vtheta}{g}{\miditerationindex} g_{\miditerationindex} + \mathcal{O}(h^3) \\
                &= \vtheta_{t-1} + \lrt h g_{(t-1)} + \lrt \lrp h^2  \evaljacgames{\vphi}{g}{{(t-1)}} f_{(t-1)} + \frac{k-1}{2k}\lrt^2 h^2  \evaljacgames{\vtheta}{g}{\miditerationindex} g_{\miditerationindex} + \mathcal{O}(h^3) 
\end{align}
\begin{align}
&\vtheta_{k, t}
                = \vtheta_{t-1} +  \lrt h g_{(t-1)} + \lrt \lrp h^2  \evaljacgames{\vphi}{g}{{(t-1)}} f_{(t-1)} \\
                & \hspace{1.2em} + \frac{k-1}{2k} h^2 \evaljacgames{\vtheta}{g}{\miditerationindex} \left(g_{t-1} + \evaljacgames{\vphi}{g}{{(t-1)}}\left(\lrp h f_{(t-1)} + \frac{(m -1)}{2m} \lrp^2 h^2  \evaljacgames{\vphi}{f}{{(t-1)}} f_{(t-1)} \right) \right)  \\
                 & \hspace{1.2em}+  \mathcal{O}(h^3) \\
                 &= \vtheta_{t-1} + \lrt h g_{(t-1)} + \lrt \lrp h^2  \evaljacgames{\vphi}{g}{{(t-1)}}f_{(t-1)} + \frac{k-1}{2k}\lrt^2 h^2 \evaljacgames{\vtheta}{g}{\miditerationindex} g_{(t-1)} + \mathcal{O}(h^3) \\
                &= \vtheta_{t-1} + \lrt h g_{(t-1)} + \lrt \lrp h^2  \evaljacgames{\vphi}{g}{{(t-1)}}f_{(t-1)} \\
                & \hspace{1em} + \frac{k-1}{2k}\lrt^2 h^2
                   \evaljacgames{\vtheta}{\left(g + \jacphig \left(\lrp h f + \frac{(m -1)}{2m}\lrp^2 h^2  \jacphif f \right) \right)}{t-1} g_{(t-1)}+ \mathcal{O}(h^3) \owntag{using Eq \eqref{eq:mid_g}} \\
                &= \vtheta_{t-1} +\lrt h g_{(t-1)} +\lrt \lrp h^2  \evaljacgames{\vphi}{g}{{(t-1)}}f_{(t-1)} + \frac{k-1}{2k}\lrt^2 h^2  \evaljacgames{\vtheta}{g}{{(t-1)}} g_{(t-1)}  + \mathcal{O}(h^3)
\end{align}
\endgroup
We then have:
\begin{align}
 \vphi_{m,t} &= \vphi_{t-1} + \lrp h f_{(t-1)} + \frac{m-1}{2m}\lrp^2 h^2  \evaljacgames{\vphi}{f}{{(t-1)}} f_{(t-1)}  + \mathcal{O}(h^3)
 \label{eq:alt_general_learning_rates_discrete_phi} \\
 \vtheta_{k,t} &= \vtheta_{t-1} + \lrt h g_{(t-1)} + \lrt \lrp h^2 \evaljacgames{\vphi}{g}{{(t-1)}} f_{(t-1)} + \frac{k-1}{2k} \lrt^2 h^2  \evaljacgames{\vtheta}{g}{{(t-1)}} g_{(t-1)}  + \mathcal{O}(h^3)
 \label{eq:alt_general_learning_rates_discrete_theta}
\end{align}

\textbf{Step 2: Expand the continuous-time changes for the modified flow given by BEA}
(Identical to the simultaneous update case.)

\textbf{Step 3: Matching the discrete and modified continuous updates.} \\
The linear terms are identical to those in the simultaneous updates:
\begin{align}
\tau_\phi = \lrp h  \\
\tau_\theta = \lrt h
\end{align}
We then obtain $f_1$ from matching the quadratic terms in Eqs~\eqref{eq:sim_ode_phi} and Eqs~\eqref{eq:alt_general_learning_rates_discrete_phi}---below we denote $f_1(\vphi_{t-1}, \vtheta_{t-1})$ by $f_1$ and $g_1(\vphi_{t-1}, \vtheta_{t-1})$ by $g_1$, for brevity:
\begin{align}
&\lrp^2 h^2 f_1 + \frac{1}{2} \lrp^2 h^2 \evaljacgames{\vphi}{f}{{(t-1)}} f_{(t-1)}  + \frac{1}{2} \lrp^2 h^2   \evaljacgames{\vtheta}{f}{{(t-1)}} g_{(t-1)}\nonumber \\ &\hspace{10em}=  \frac{m-1}{2m} \lrp^2 h^2  \evaljacgames{\vphi}{f}{{(t-1)}} f_{(t-1)} \\
& f_1 + \frac{1}{2}  \evaljacgames{\vphi}{f}{{(t-1)}} f_{(t-1)}  + \frac{1}{2}    \evaljacgames{\vtheta}{f}{{(t-1)}} g_{(t-1)} =  \frac{m-1}{2m}   \evaljacgames{\vphi}{f}{{(t-1)}} f_{(t-1)} \\
&  f_1 =  \left(\frac{m-1}{2m} -\frac{1}{2}\right)  \evaljacgames{\vphi}{f}{{(t-1)}} f_{(t-1)}  - \frac{1}{2} \evaljacgames{\vtheta}{f}{{(t-1)}} g_{(t-1)} \\
&f_1 =  - \frac{1}{2m} \evaljacgames{\vphi}{f}{{(t-1)}} f_{(t-1)} - \frac{1}{2} \evaljacgames{\vtheta}{f}{{(t-1)}} g_{(t-1)}
\end{align}
For $g_1$, from Eqs~\eqref{eq:sim_ode_theta} and \eqref{eq:alt_general_learning_rates_discrete_theta}:
\begin{align}
&  \lrt^2 h^2 \Big(g_1 + \frac{1}{2} \evaljacgames{\vphi}{g}{{(t-1)}} f_{(t-1)} +  \frac{1}{2}  \evaljacgames{\vtheta}{g}{{(t-1)}} g_{(t-1)}\Big) \nonumber \\
& \hspace{10em}= \lrt \lrp h^2 \evaljacgames{\vphi}{g}{{(t-1)}} f_{(t-1)} + \frac{k-1}{2k} \lrt^2 h^2  \evaljacgames{\vtheta}{g}{{(t-1)}} g_{(t-1)} \\
& g_1 +  \frac{1}{2} \evaljacgames{\vphi}{g}{{(t-1)}} f_{(t-1)} +  \frac{1}{2}  \evaljacgames{\vtheta}{g}{{(t-1)}} g_{(t-1)} \nonumber\\ &\hspace{10em}= \frac{\lrp}{\lrt} \evaljacgames{\vphi}{g}{{(t-1)}} f_{(t-1)} + \frac{k-1}{2k}   \evaljacgames{\vtheta}{g}{{(t-1)}} g_{(t-1)} \\
&  g_1 +  \frac{1}{2} \evaljacgames{\vphi}{g}{{(t-1)}} f_{(t-1)} +  \frac{1}{2}  \evaljacgames{\vtheta}{g}{{(t-1)}} g_{(t-1)} \nonumber\\ &\hspace{10em}= \frac{\lrp}{\lrt} \evaljacgames{\vphi}{g}{{(t-1)}} f_{(t-1)} + \frac{k-1}{2k}   \evaljacgames{\vtheta}{g}{{(t-1)}} g_{(t-1)} \\
&   g_1 = \left(\frac{\lrp}{\lrt} -\frac{1}{2}\right) \evaljacgames{\vphi}{g}{{(t-1)}} f_{(t-1)} - \frac{1}{2k}   \evaljacgames{\vtheta}{g}{{(t-1)}} g_{(t-1)}
\end{align}
We thus have that
\begin{align}
f_1(\vphi_{t-1}, \vtheta_{t-1}) &=  - \frac{1}{2m} \evaljacgames{\vphi}{f}{{(t-1)}} f_{(t-1)} - \frac{1}{2} \evaljacgames{\vtheta}{f}{{(t-1)}} g_{(t-1)}
\label{eq:alternating_f1}\\
g_1(\vphi_{t-1}, \vtheta_{t-1}) &= \left(\frac{\lrp}{\lrt} -\frac{1}{2}\right) \evaljacgames{\vphi}{g}{{(t-1)}} f_{(t-1)} - \frac{1}{2k}   \evaljacgames{\vtheta}{g}{{(t-1)}} g_{(t-1)}.
\label{eq:alternating_g1}
\end{align}
We found $f_1$ and $g_1$ such that after one optimisation step the flows $\dot{\vphi} = f + \lrp h f_1$ and $\dot{\vtheta} = g + \lrt h g_1$ follow the discrete updates up to order $\mathcal{O}(h^3)$, finishing the proof.

\section{Proof of the main corollaries}

In this section, we will write the modified equations in the case of using gradient descent common-payoff games and zero-sum games.
To do so, we will replace $f$ and $g$ with the values given by gradient descent in the first order corrections we have derived in the previous sections. In the common-payoff case $f = - \nabla_{\vphi} E$ and $g = -\nabla_{\vtheta} E$, while in the zero-sum case  $f = \nabla_{\vphi} E$ and $g = -\nabla_{\vtheta} E$, where $E(\vphi, \vtheta)$ is a function of the player parameters. 
We will use the following identities, which are obtained from the chain rule:
\begin{align}
\jacparam{\vphi}{\nabla_{\vphi} E}{\nabla_{\vphi} E} &=  \nabla_{\vphi} \Big(\frac{\|\nabla_{\vphi} E\|^2}2\Big) \label{eq:normssamephi}\\
\jacparam{\vtheta}{\nabla_{\vtheta} E}{\nabla_{\vtheta} E} &=  \nabla_{\vtheta} \Big(\frac{\|\nabla_{\vtheta} E\|^2}2\Big) \label{eq:normssametheta} \\
 \jacparam{\vtheta}{\nabla_{\vphi} E}{\nabla_{\vtheta} E} &= {\left(\jacparam{\vphi}{\nabla_{\vtheta} E}\right)}^T{\nabla_{\vtheta} E} =  \nabla_{\vphi} \Big(\frac{\|\nabla_{\vtheta} E\|^2}2\Big) \label{eq:normscrosstheta}\\
 \jacparam{\vphi}{\nabla_{\vtheta} E}{\nabla_{\vphi} E} &= {\left(\jacparam{\vtheta}{\nabla_{\vphi} E}\right)}^T{\nabla_{\vphi} E} =   \nabla_{\vtheta} \Big(\frac{\|\nabla_{\vphi} E\|^2}2\Big)\label{eq:normscrossphi}.
\end{align}

\subsection{Common-payoff alternating two-player games}

\begin{customcor}{common-payoff alternating two-player games} (Corollary~\ref{cor:common_payoff_alt}) In a two-player common-payoff game with common loss $E$, \textit{alternating} gradient descent -- as described in Eqs~\eqref{eq:supp_altup1} and~\eqref{eq:supp_altup2}---with one update per player follows a gradient flow given by the modified losses
\begin{align}
\tilde E_{\vphi}&= E + \frac{\lrp h}{4} \left(\norm{\nabla_{\vphi} E}^2  + \norm{\nabla_{\vtheta} E}^2\right) \\
\tilde E_{\vtheta}&=  E + \frac{\lrt h}{4} \left((1 - \frac {2 \lrp} {\lrt}) \norm{\nabla_{\vphi} E}^2  + \norm{\nabla_{\vtheta} E}^2\right)
\end{align}
with an error of size $\mathcal O(h^3)$  after one update step.
\end{customcor}

In the common-payoff case, both players minimise the loss function $E$.
Substituting $f=-\nabla_{\vphi} E$ and $g=-\nabla_{\vtheta} E$ into the corrections $f_1$ and $g_1$ for the alternating Euler updates in Theorem~\ref{thm:alt}
and using the gradient identities above yields
\begin{align}
f_1 & = - \frac{1}{2} \left(\jacphif f + \jacthetaf g \right) \\
    &=  -\frac{1}2\left(
                                \frac 1m \jacparam{\vphi}{\nabla_{\vphi} E} \nabla_{\vphi} E
                                + \jacparam{\vtheta}{\nabla_{\vphi} E} \nabla_{\vtheta} E
                      \right) \owntag{\text{def} f, g}\\
        &= -\frac{1}2\left(
                                \frac 1m \nabla_{\vphi} \frac{\|\nabla_{\vphi} E\|^2}2\
                                +\nabla_{\vphi} \frac{\|\nabla_{\vtheta} E\|^2}2
                      \right) \owntag{via \eqref{eq:normssamephi},\eqref{eq:normscrosstheta}}\\
    &= -\nabla_{\vphi} \left(\frac{1}{4m} \|\nabla_{\vphi} E\|^2  + \frac{1}{4} \|\nabla_{\vtheta} E\|^2 \right) \\ \nonumber\\
g_1 &= - \frac{1}{2} \left((1- \frac {2\lrp} {\lrt})\jacphig f + \frac{1}{k} \jacthetag g \right) \\
 &= -\frac{1}2\left(
                                (1 - \frac{2\lrp}\lrt)  \jacparam{\vphi}{\nabla_{\vtheta} E} \nabla_{\vphi} E
                                + \frac 1k \jacparam{\vtheta}{\nabla_{\vtheta} E} \nabla_{\vtheta} E
                       \right) \owntag{\text{def} f, g}\\
    &= -\frac{1}2\left(
                                (1 - \frac{2\lrp}\lrt) \nabla_{\vtheta} \frac{\|\nabla_{\vphi} E\|^2}2
                                + \frac 1k\nabla_{\vtheta}\frac{\|\nabla_{\vtheta} E\|^2}2
                       \right) \owntag{via \eqref{eq:normssametheta},\eqref{eq:normscrossphi}}\\
    &= -\nabla_{\vtheta} \left(
                                \frac{1}4 (1 - \frac{2\lrp}\lrt)  \|\nabla_{\vphi} E\|^2
                                + \frac {1}k\|\nabla_{\vtheta} E\|^2
                         \right)
\end{align}
Now, replacing the gradient expressions for $f_1$ and $g_1$ calculated above into the modified equations $\dot \vphi = -\nabla_{\vphi} E + \lrp hf_1$ and $\dot \vtheta = -\nabla_{\vtheta} E + \lrt hg_1$ and factoring out the gradients, we obtain the modified equations in the form of the flows:
\begin{align}
\dot \vphi   &= -\nabla_{\vphi} \widetilde E_{\vphi} \\
\dot \vtheta &= -\nabla_{\vtheta} \widetilde E_{\vtheta},
\end{align}
with the following modified losses for each players:
\begin{align}
\widetilde E_{\vphi} & = E + \frac{\lrp h}4  \left(\frac 1m\|\nabla_{\vphi} E\|^2 + \|\nabla_{\vtheta} E\|^2\right) \\
\widetilde E_{\vtheta} & = E + \frac{\lrt h}4 \left( (1 - \frac{2\lrp}{\lrt} )\|\nabla_{\vphi} E\|^2 + \frac 1k \|\nabla_{\vtheta} E\|^2 \right).
\end{align}
We obtain Corollary 5.1 by setting the number of player updates to one: $m=k=1$.

\subsection{Zero-sum simultaneous two-player games}
\label{sec:zero-sum-sim}

\begin{customcor}{Zero-sum simultaneous two-player games} (Corollary~\ref{cor:zs-sim}) In a zero-sum two-player differentiable game, \textit{simultaneous} gradient descent updates - as described in Eqs~\eqref{eq:supp_simup1} and~\eqref{eq:supp_simup2} - follows a gradient flow given by the modified losses
\begin{align}
\tilde E_{\vphi}&= - E + \frac{\lrp h}{4} \norm{\nabla_{\vphi} E}^2  -  \frac{\lrp h}{4} \norm{\nabla_{\vtheta} E}^2
 \\
\tilde E_{\vtheta}&= E -  \frac{\lrt h}{4} \norm{\nabla_{\vphi} E}^2  +  \frac{\lrt h}{4} \norm{\nabla_{\vtheta} E}^2
\end{align}
with an error of size $\mathcal O(h^3)$ after one update step.
\end{customcor}
Substituting $f = \nabla_{\vphi} E$ and $g=-\nabla_{\vtheta} E$ into the corrections $f_1$ and $g_1$ for the simultaneous Euler updates and using the gradient identities above yields
\begin{align}
f_1 & = - \frac{1}{2} \left(\jacphif f + \jacthetaf g \right) \\
    &=  -\frac{1}2\left(
                               \jacparam{\vphi}{\nabla_{\vphi} E}  \nabla_{\vphi} E
                                -  \jacparam{\vtheta}{\nabla_{\vphi} E} \nabla_{\vtheta} E
                      \right) \owntag{\text{def} f, g}\\
    &= -\frac{1}2\left(
                                  \nabla_{\vphi} \frac{\|\nabla_{\vphi} E\|^2}2
                                - \nabla_{\vphi} \frac{\|\nabla_{\vtheta} E\|^2}2
                      \right) \owntag{via \eqref{eq:normssamephi},\eqref{eq:normscrosstheta}}\\
    &= -\nabla_{\vphi} \left(\frac{1}{4} \|\nabla_{\vphi} E\|^2  - \frac{1}{4} \|\nabla_{\vtheta} E\|^2 \right) \\ 
g_1 &=  - \frac{1}{2} \left(\jacphig f + \jacthetag g \right) \\
    &=  - \frac{1}{2} \left(-  \jacparam{\vphi}{\nabla_{\vtheta} E} \nabla_{\vphi} E + \jacparam{\vtheta}{\nabla_{\vtheta} E} \nabla_{\vtheta} E \right) \owntag{\text{def} f, g}\\
    &= -\frac{1}2\left(
                                - \nabla_{\vtheta} \frac{\|\nabla_{\vphi} E\|^2}2
                                + \nabla_{\vtheta}\frac{\|\nabla_{\vtheta} E\|^2}2
                       \right) \owntag{via \eqref{eq:normssametheta},\eqref{eq:normscrossphi}} \\
    &= -\nabla_{\vtheta} \left(
                                - \frac{1}4 \|\nabla_{\vphi} E\|^2
                                + \frac{1}4 \|\nabla_{\vtheta} E\|^2
                         \right)
\end{align}
Now, replacing the gradient expressions for $f_1$ and $g_1$ calculated above into the modified equations $\dot \vphi = -\nabla_{\vphi}(-E) + \lrp hf_1$ and $\dot \vtheta = -\nabla_{\vtheta} E + \lrt hg_1$ and factoring out the gradients, we obtain the modified equations in the form of the flows:
\begin{align}
\dot \vphi   &= -\nabla_{\vphi} \widetilde E_{\vphi} \\
\dot \vtheta &= -\nabla_{\vtheta} \widetilde E_{\vtheta},
\end{align}
with the following modified losses for each players:
 \begin{align}
\widetilde E_{\vphi} & = -E + \frac{\lrp h}{4} \|\nabla_{\vphi} E\|^2  - \frac{\lrp h}{4} \|\nabla_{\vtheta} E\|^2 \\
\widetilde E_{\vtheta} & = E  - \frac{\lrt h}4 \|\nabla_{\vphi} E\|^2 + \frac{\lrt h}4 \|\nabla_{\vtheta} E\|^2 .
 \end{align}

\subsection{Zero-sum alternating two-player games}

\begin{customcor}{Zero-sum alternating two-player games} (Corollary~\ref{cor:zs-alt}) In a zero-sum two-player differentiable game, \textit{alternating} gradient descent - as described in Eqs~\eqref{eq:supp_altup1} and~\eqref{eq:supp_altup2} - follows a gradient flow given by the modified losses
\begin{align}
\tilde E_{\vphi}&= -E + \frac{\lrp h}{4m} \norm{\nabla_{\vphi} E}^2  - \frac{\lrp h}{4} \norm{\nabla_{\vtheta} E}^2
\\
\tilde E_{\vtheta}&= E - \frac{\lrt h}{4} (1 - \frac{2 \lrp}{\lrt}) \norm{\nabla_{\vphi} E}^2  + \frac{\lrt h}{4k}\norm{\nabla_{\vtheta} E}^2
\end{align}
with an error of size $\mathcal O(h^3)$ after one update step.
\end{customcor}

In this last case, substituting $f=\nabla_{\vphi} E$ and $g=-\nabla_{\vtheta} E$ into the corrections $f_1$ and $g_1$ for the alternating Euler updates and using the gradient identities above yields the modified system as well as the modified losses exactly in the same way as for the two previous cases above. (This amounts to a single sign change in the proof of Corollary 5.1)

\subsection{Self and interaction terms in zero-sum games}

\begin{remark}
Throughout the Appendix, we will refer to self terms and interaction terms, as originally defined in our thesis (Definition~\ref{def:self_interaction_terms}), and we will also use this terminology to refer to terms in our derivations that originate from the self terms and interaction terms. While a slight abuse of language, we find it useful to emphasise the provenance of these terms in our discussion.\end{remark}

For the case of zero-sum games trained with simultaneous gradient descent, the self terms encourage the minimisation of the player's own gradient norm, while the interaction terms encourage the maximisation of the other player's gradient norm:
\begin{align}
\widetilde E_{\vphi} & = -E + \underbrace{\frac{\lrp h}{4} \|\nabla_{\vphi} E\|^2}_{self} \underbrace{- \frac{\lrp h}{4} \|\nabla_{\vtheta} E\|^2}_{interaction}, \\
\widetilde E_{\vtheta} & = E  \underbrace{- \frac{\lrt h}4 \|\nabla_{\vphi} E\|^2}_{interaction} \underbrace{+ \frac{\lrt h}4 \|\nabla_{\vtheta} E\|^2 }_{self}.
\end{align}

Similar terms are obtained for alternating gradient descent, with the only difference that the sign of the second player's interaction term can be positive, depending on the learning rate ratio of the two players.

\section{Discretisation drift in Runge--Kutta 4}
\label{sec:dd_rk}

\subsection{Runge--Kutta 4 for one player}

 Runge--Kutta 4 (RK4) is a Runge--Kutta a method of order 4. That is, the discrete steps of RK4 coincide with the dyanamics of the flow they discretise up to $\mathcal O(h^5)$ (i.e., the local error after one step is of order $\mathcal O(h^5)$). The modified equation for a method of order $n$ starts with corrections at order $h^{n+1}$ (i.e., all the lower corrections vanish; see \cite{backward_lifespan} and \cite{GNI} for further details). This means that RK4 has no DD up to order $\mathcal O(h^5)$, and why for small learning rates RK4 can be used as a proxy for the exact flow.

\subsection{Runge--Kutta 4 for two players}

When we use equal learning rates and simultaneous updates, the two-player game is always equivalent to the one player case, so Runge--Kutta 4 will have a local error of $\mathcal O(h^5)$. However, in the case of two-players games, we have the additional freedom of having different learning rates for each of the players. We now show that when the learning rates of the two players are different, \textit{RK4 also also has a drift effect of order $2$ and the DD term comes exclusively from the interaction terms}. To do so, we use BEA and apply the same steps we performed for simultaneous and alternating Euler updates.

\vspace{2em}
\textbf{Step 1: Expand the updates per player via a Taylor expansion.} \\
The simultaneous Runge--Kutta 4 updates for two players require defining:
\begin{align}
 k_{1, \vphi} &= f(\vphi_{t-1}, \vtheta_{t-1}) \\
 k_{1, \vtheta} &= g(\vphi_{t-1}, \vtheta_{t-1}) \\
 k_{2,\vphi} &= f(\vphi_{t-1} + \frac{\lrp h}{2} k_{1, \vphi}, \vtheta_{t-1} + \frac{\lrt h}{2} k_{1, \vtheta}) \\
 k_{2,\vtheta} &= g(\vphi_{t-1} + \frac{\lrp h}{2} k_{1, \vphi}, \vtheta_{t-1} + \frac{\lrt h}{2} k_{1, \vtheta}) \\
 k_{3,\vphi} &= f(\vphi_{t-1} + \frac{\lrp h}{2} k_{2, \vphi}, \vtheta_{t-1} + \frac{\lrt h}{2} k_{2, \vtheta}) \\
 k_{3,\vtheta} &=  g(\vphi_{t-1} + \frac{\lrp h}{2} k_{2, \vphi}, \vtheta_{t-1} + \frac{\lrt h}{2} k_{2, \vtheta}) \\
 k_{4,\vphi} &= f(\vphi_{t-1} + \frac{\lrp h}{2} k_{3, \vphi}, \vtheta_{t-1} + \frac{\lrt h}{2} k_{3, \vtheta}) \\
 k_{4,\vtheta} &=  g(\vphi_{t-1} + \frac{\lrp h}{2} k_{3, \vphi}, \vtheta_{t-1} + \frac{\lrt h}{2} k_{3, \vtheta})
 \end{align}
 From here we can define the update vectors:
 \begin{align}
 k_{\vphi} &= \frac{1}{6} \left(k_{1,\vphi} + 2 k_{2,\vphi} + 2 k_{3,\vphi} + k_{4,\vphi}\right) \\
 k_{\vtheta} &= \frac{1}{6} \left(k_{1,\vtheta} + 2 k_{2,\vtheta} + 2 k_{3,\vtheta} + k_{4,\vtheta}\right) 
 \end{align}
And the parameter updates
\begin{align}
 \vphi_{t}  &= \vphi_{t-1} + \lrp h  k_{\vphi} \\
 \vtheta_{t}  &= \vtheta_{t-1} + \lrt h k_{\vtheta}
\end{align}
To find the discretisation drift we expand each intermediate step:
\begingroup
\begin{align}
 k_{1, \vphi} &= f_{(t-1)} \\
 k_{1, \vtheta} &= g_{(t-1)} \\
 k_{2,\vphi} &= f\left(\vphi_{t-1} + \frac{\lrp h}{2} k_{1, \vphi}, \vtheta_{t-1} + \frac{\lrt h}{2} k_{1, \vtheta}\right) \\
    &= f_{(t-1)} + \frac{\lrt h}{2}  \evaljacgames{\vtheta}{f}{{(t-1)}} k_{1, \vtheta}+ \frac{\lrp h}{2}  \evaljacgames{\vphi}{f}{{(t-1)}} k_{1,\vphi} + \mathcal{O}(h^2) \\
    &= f_{(t-1)} + \frac{\lrt h}{2}  \evaljacgames{\vtheta}{f}{{(t-1)}} g_{(t-1)} + \frac{\lrp h}{2} \evaljacgames{\vphi}{f}{{(t-1)}} f_{(t-1)}  + \mathcal{O}(h^2) \\
 k_{2,\vtheta} &= g\left(\vphi_{t-1} + \frac{\lrp h}{2} k_{1, \vphi}, \vtheta + \frac{\lrt h}{2} k_{1, \vtheta}\right) \\
    &=g_{(t-1)} + \frac{\lrt h}{2}  \evaljacgames{\vtheta}{g}{{(t-1)}} k_{1, \vtheta} + \frac{\lrp h}{2}  \evaljacgames{\vphi}{g}{{(t-1)}} k_{1,\vphi}  + \mathcal{O}(h^2) \\
    &=g_{(t-1)} + \frac{\lrt h}{2}  \evaljacgames{\vtheta}{g}{{(t-1)}} g_{(t-1)} + \frac{\lrp h}{2} \evaljacgames{\vphi}{g}{{(t-1)}} f_{(t-1)}  + \mathcal{O}(h^2) \\
  k_{3,\vphi} &= f_{(t-1)} + \frac{\lrt h}{2}  \evaljacgames{\vtheta}{f}{{(t-1)}} k_{2, \vtheta} + \frac{\lrp h}{2} \evaljacgames{\vphi}{f}{{(t-1)}} k_{2,\vphi}  + \mathcal{O}(h^2) \\
    &= f_{(t-1)} + \frac{\lrt h}{2}  \evaljacgames{\vtheta}{f}{{(t-1)}} g_{(t-1)} + \frac{\lrp h}{2} \evaljacgames{\vphi}{f}{{(t-1)}} f_{(t-1)}  + \mathcal{O}(h^2) \\
k_{3,\vtheta} &=g_{(t-1)} + \frac{\lrt h}{2} \evaljacgames{\vtheta}{g}{{(t-1)}} g_{(t-1)} + \frac{\lrp h}{2} \evaljacgames{\vphi}{g}{{(t-1)}} f_{(t-1)} + \mathcal{O}(h^2) \\
k_{4,\vphi} &= f_{(t-1)} + \frac{\lrt h}{2}  \evaljacgames{\vtheta}{f}{{(t-1)}} g_{(t-1)} + \frac{\lrp h}{2} \evaljacgames{\vphi}{f}{{(t-1)}} f_{(t-1)}  + \mathcal{O}(h^2) \\
k_{4,\vtheta} &=g_{(t-1)} + \frac{\lrt h}{2} \evaljacgames{\vtheta}{g}{{(t-1)}} g_{(t-1)} + \frac{\lrp h}{2} \evaljacgames{\vphi}{g}{{(t-1)}}  f_{(t-1)} + \mathcal{O}(h^2) 
\end{align}
\endgroup
with the update direction:
\begin{align}
k_{\vphi} &= \frac{1}{6} \left(k_{1,\vphi} + 2 k_{2,\vphi} + 2 k_{3,\vphi} + k_{4,\vphi}\right) \\
         &= f_{(t-1)} + \frac{ \lrt h}{2}  \evaljacgames{\vtheta}{f}{{(t-1)}} g_{(t-1)} + \frac{\lrp h}{2}  \evaljacgames{\vphi}{f}{{(t-1)}} f_{(t-1)}  + \mathcal{O}(h^2) \\
k_{\vtheta} &= \frac{1}{6} \left(k_{1,\vtheta} + 2 k_{2,\vtheta} + 2 k_{3,\vtheta} + k_{4,\vtheta}\right) \\
        &= g_{(t-1)} + \frac{\lrt h}{2}  \evaljacgames{\vtheta}{g}{{(t-1)}} g_{(t-1)} + \frac{\lrp h}{2} \evaljacgames{\vphi}{g}{{(t-1)}} f_{(t-1)} + \mathcal{O}(h^2) 
\end{align}
and thus the discrete update of the Runge--Kutta 4 for two players are:
\begin{align}
\vphi_{t}  &= \vphi_{t-1} + \lrp h f_{(t-1)} + \frac{1}{2} \lrp \lrt h^2  \evaljacgames{\vtheta}{f}{{(t-1)}} g_{(t-1)} + \frac{1}{2} \lrp^2 h^2  \evaljacgames{\vphi}{f}{{(t-1)}} f_{(t-1)} + \mathcal{O}(h^3) \label{eq:rk_1}\\
\vtheta_{t}  &= \vtheta_{t-1} + \lrt h g_{(t-1)} + \frac{1}{2} \lrt^2 h^2  \evaljacgames{\vtheta}{g}{{(t-1)}} g_{(t-1)} + \frac{1}{2} \lrp\lrt h^2 \evaljacgames{\vphi}{g}{{(t-1)}} f_{(t-1)} + \mathcal{O}(h^3) \label{eq:rk_2}
\end{align}

\textbf{Step 2: Expand the continuous-time changes for the modified flow }\\
(Identical to the simultaneous Euler updates.)

\textbf{Step 3: Matching the discrete and modified continuous updates.} \\
As in the always in Step 3,  we substitute  $\vphi_{t-1}$ and $\vtheta_{t-1}$ in Lemma~\ref{thm:cont}:
\begin{align}
\vphi(\tph + \tau_\phi) &= \vphi_{t-1} + \tau_\phi f_{(t-1)} + \tau_\phi^2 f_1(\vphi_{t-1}, \vtheta_{t-1}) + \frac{1}{2} \tau_\phi^2  \evaljacgames{\vphi}{f}{{(t-1)}}  f_{(t-1)} \nonumber \\
   & \hspace{2em} + \frac{1}{2}\tau_\phi^2 \evaljacgames{\vtheta}{f}{{(t-1)}} g_{(t-1)}  + \mathcal{O}(\tau_\phi^3) \\
\vtheta(\tph + \tau_\theta) &= \vtheta_{t-1} + \tau_\theta g_{(t-1)} + \tau_\theta^2 g_1(\vphi_{t-1}, \vtheta_{t-1}) + \frac{1}{2} \tau_\theta^2 \evaljacgames{\vphi}{g}{{(t-1)}} f_{(t-1)} \nonumber \\
  & \hspace{2em} + \frac{1}{2}\tau_\theta^2  \evaljacgames{\vtheta}{g}{{(t-1)}} g_{(t-1)} + \mathcal{O}(\tau_\theta^3)
\end{align}

For the first order terms we obtain $\tau_{\phi} = \lrp h$ and $\tau_{\theta} = \lrt h$. We match the $\mathcal{O}(h^2)$ terms in the equations above with the discrete Runge--Kutta 4 updates shown in Eq~\eqref{eq:rk_1} and~\eqref{eq:rk_2}
and notice that:
\begin{align}
f_1(\vphi_{t-1}, \vtheta_{t-1}) = \frac 1 2 (\frac{\lrt}{\lrp}- 1)  \evaljacgames{\vtheta}{f}{(t-1)}  g_{(t-1)}\\
g_1(\vphi_{t-1}, \vtheta_{t-1}) = \frac 1 2(\frac{\lrp}{\lrt}- 1) \evaljacgames{\vphi}{g}{(t-1)} f_{(t-1)}
\end{align}
Thus, if $\lrp \ne \lrt$ RK4 has second order drift. This is why, in all our experiments comparing with RK4, we use the same learning rates for the two players $\lrp = \lrt$, to ensure that we use a method which has no DD up to order $\mathcal{O}(h^5)$.

\section{Stability analysis}
\label{sec:dd_gan_stability_analysis}

In this section, we give all the details of the stability analysis results, to showcase how the modified flows we have derived can be used as a tool for stability analysis. We  provide the full computation for the Jacobian of the modified vector fields for simultaneous and alternating Euler updates, as well as the calculation of their trace, and show how this can be used to determine the stability of the modified vector fields. While analysing the modified vector fields is not equivalent to analysing the discrete dynamics due to the higher order errors of the drift which we ignore, it provides a better approximation than what is often used in practice, namely the original flows, which completely ignore the drift.

\subsection{Simultaneous Euler updates}

Consider a two-player game with dynamics given by $\vphi_t = \vphi_{t-1}  +  \lrp h f_{(t-1)}$ and  $\vtheta_t = \vtheta_{t-1}  +  \lrt h g_{(t-1)} $ (Eqs \eqref{eq:supp_simup1} and \eqref{eq:supp_simup2}). The modified dynamics for this game are given by $\dot{\vphi} = \widetilde{f} $, $\dot{\vtheta} = \widetilde{g}$, where $\widetilde{f} = f - \frac{ \lrp h}{2} (\jacphif f + \jacthetaf g)$ and $\widetilde{g} = g - \frac{ \lrt h}{2} (\jacphig f + \jacthetag g )$ (Theorem \ref{thm:supp_2_players_sim}).

The stability of this system can be characterised by the modified Jacobian matrix evaluated at the equilibria of the two-player game. The equilibria that we are interested in for our stability analysis are the steady-state solutions of Eqs \eqref{eq:supp_simup1} and \eqref{eq:supp_simup2}, given by $f = \myvec{0}, g = \myvec{0}$. These are also equilibrium solutions for the steady-state modified equations\footnote{There are additional steady-state solutions for the modified equations. However, we can ignore these since they are spurious solutions that do not correspond to steady states of the two-player game, arising instead as an artifact of the $\mathcal{O}(h^3)$ error in BEA.} given by $\tilde{f} = \mathbf{0}, \tilde{g} = \mathbf{0}$.

The modified Jacobian can be written, using block matrix notation as:
\begin{gather}
 \widetilde{\vJ}
 =
  \begin{bmatrix}
    \jacparam{\vphi}{\widetilde{f}} & \jacparam{\vtheta}{\widetilde{f}}\\
    \jacparam{\vphi}{\widetilde{g}} &
   \jacparam{\vtheta}{\widetilde{g}}
   \end{bmatrix}
\end{gather}
Next, we calculate each term in this block matrix. (In the following analysis, each term is evaluated at an equilibrium solution given by $f = \mathbf{0}, g = \mathbf{0}$). We find:
\begin{align}
\jacparam{\vphi}{\tilde{f}} &= \jacphif - \frac{ \lrp h}{2} \left(( \jacphif)^2 +  \jactwoparam{\vphi}{\vphi}{f} f + \jacthetaf \jacphig   + \jactwoparam{\vphi}{\vtheta}{f} g \right)\\
&=\jacphif - \frac{ \lrp h}{2} \left(( \jacphif)^2 + \jacthetaf \jacphig   \right) \\
\jacparam{\vtheta}{\tilde{f}} &= \jacthetaf - \frac{ \lrp h}{2} \left(\jacphif \jacthetaf +  \jactwoparam{\vtheta}{ \vphi}{f} f + \jacthetaf \jacthetag  +  \jactwoparam{\vtheta}{\vtheta}{f} g \right)\\
&= \jacthetaf - \frac{ \lrp h}{2} \left(\jacphif \jacthetaf  + \jacthetaf \jacthetag \right) \\
\jacparam{\vphi}{\tilde{g}} &= \jacphig - \frac{ \lrt h}{2} (\jacthetag \jacphig +  \jactwoparam{\vphi}{\vtheta}{g} g + \jacphig \jacphif  +\jactwoparam{\vphi}{\vphi}{g} ) f \\
&= \jacphig - \frac{ \lrt h}{2} ( \jacthetag \jacphig + \jacphig \jacphif  )\\
\jacparam{\vtheta}{\tilde{g}} &= \jacthetag - \frac{ \lrt h}{2} \left( (\jacthetag)^2+  \jactwoparam{\vtheta}{\vtheta}{g} g +  \jacphig \jacthetaf +   \jactwoparam{\vtheta}{\vphi}{g} f \right)\\
&= \jacthetag - \frac{ \lrt h}{2} \left( (\jacthetag)^2+  \jacphig \jacthetaf \right)
\end{align}
where $\jactwoparam{\vphi}{\vtheta}{f} g$ is the matrix in $\mathbb{R}^{m, m}$ with $\left(\jactwoparam{\vphi}{\vtheta}{f} g\right)_{i,j} = \left(\jacparam{\vtheta}{\frac{\partial f_i}{\partial \vphi_j}}\right)^T g$.

Given these calculations, we can now write
\begin{gather}
 \widetilde{\vJ}
 =
  \begin{bmatrix}
    \jacparam{\vphi}{\widetilde{f}} & \jacparam{\vtheta}{\widetilde{f}}\\
    \jacparam{\vphi}{\widetilde{g}} &
   \jacparam{\vtheta}{\widetilde{g}}
   \end{bmatrix}
   = \vJ - \frac{h}{2}\mathbf{K}_{\text{sim}}
\end{gather}
where $\vJ$ is the Jacobian of the unmodified flow
\begin{gather}
 \vJ
 =
  \begin{bmatrix}
    \jacphif & \jacthetaf\\
    \jacphig & \jacthetag
   \end{bmatrix}
\end{gather}
and
\begin{gather}
\mathbf{K}_{\text{sim}}
 =
 \begin{bmatrix}
    \lrp (\jacphif)^2 +  \lrp \jacthetaf \jacphig  & \lrp \jacphif \jacthetaf +  \lrp \jacthetaf \jacthetag \\
   \lrt \jacthetag  \jacphig +  \lrt \jacphig \jacphif &
   \lrt(\jacthetag)^2 +  \lrt \jacphig \jacthetaf 
   \end{bmatrix}.
\end{gather}

Using the modified Jacobian, we can use Remark~\ref{rem:stability_analysis} to determine the stability of fixed points.
A necessary condition for stability is that the trace of the modified Jacobian is less than or equal to zero (i.e. $\Tr(\widetilde{\vJ})\leq0$), since the trace is the sum of eigenvalues. Using the property of trace additivity and the trace cyclic property we see that:\begin{equation}\label{eqn:modified_trace}
\Tr(\widetilde{\vJ}) = \Tr(\vJ) - \frac{ h}{2}\left(\lrp \Tr((\jacphif)^2)+ \lrt  \Tr((\jacthetag)^2)    \right) - \frac{ h}{2}(\lrp + \lrt )\Tr( \jacthetaf \jacphig ).
\end{equation}

\textbf{Instability caused by discretisation drift}.
We now use the above analysis to show that the equilibria of a two-player game following the modified flow obtained simultaneous Euler updates as defined by Eqs \eqref{eq:supp_simup1} and \eqref{eq:supp_simup2} can become asymptotically unstable for some games.

There are choices of $f$ and $g$ that have stable equilibrium without DD, but are unstable under DD. For example, consider the zero-sum two-player game with $f=\nabla_{\vphi} E $ and $g=- \nabla_{\vtheta} E $. 
In this case $\jacphif = \nabla_{\vphi}^2 E$, $\jacthetag = - \nabla_{\vtheta}^2 E$ and $\jacthetaf = {\jacphig}^T = \nabla_{\vphi, \vtheta} E$. We can thus use $\Tr(A^T A) = \|A\|_2$ where $\|.\|_2$ denotes the Frobenius norm, we have \begin{equation}\label{eqn:modified_trace_zero_sum}
\Tr(\widetilde{\vJ}) = \Tr(\vJ) - \frac{ h}{2}\left(\lrp {\| \nabla_{\vphi}^2 E\|}^2_2  + \lrt {\| \nabla_{\vtheta}^2 E\|}^2_2    \right) + \frac{ h}{2}(\lrp + \lrt ) {\| \nabla_{\vphi, \vtheta} E\|}^2_2.
\end{equation}
 The Dirac-GAN is an example of a zero-sum two-player game that is stable without DD, but becomes unstable under DD with  $\Tr(\widetilde{\vJ})  = h(\lrp + \lrt ) {\| \nabla_{\vphi, \vtheta} E\|}^2_2/2 > 0$ (see Section \ref{sec:dirac-gan-stability}).

\subsection{Alternating Euler updates}
\label{sec:dd_gan_stability_analysis_alternating}

Consider a two-player game with dynamics given by Eqs \eqref{eq:supp_altup1} and \eqref{eq:supp_altup2}. The modified dynamics for this game are given by $\dot{\vphi} = f - \frac{\lrp h}{2} \left(\frac{1}{m}\jacphif f + \jacthetaf g \right)$,
$ \dot{\vtheta} = g - \frac{\lrt h}{2} \left((1- \frac {2\lrp} {\lrt})\jacphig f + \frac{1}{k} \jacthetag g \right)$ (Theorem \ref{thm:supp_2_players_alt}).

The stability of this system can be characterised by the modified Jacobian matrix evaluated at the equilibria of the two-player game. The equilibria that we are interested in for our stability analysis are the steady-state solutions of Eqs \eqref{eq:supp_simup1} and \eqref{eq:supp_simup2}, given by $f = \myvec{0}, g = \myvec{0}$. These are also equilibrium solutions for the steady-state modified equations\footnote{There are additional steady-state solutions for the modified equations. However, we can ignore these since they are spurious solutions that do not correspond to steady states of the two-player game, arising instead as an artifact of the $\mathcal{O}(h^3)$ error in BEA.} given by $\tilde{f} = \mathbf{0}, \tilde{g} = \mathbf{0}$.

The modified Jacobian can be written, using block matrix notation as:
\begin{gather}
 \widetilde{\vJ}
 =
  \begin{bmatrix}
    \jacparam{\vphi}{\widetilde{f}} & \jacparam{\vtheta}{\widetilde{f}}\\
    \jacparam{\vphi}{\widetilde{g}} &
   \jacparam{\vtheta}{\widetilde{g}}
   \end{bmatrix}
\end{gather}
Next, we calculate each term in this block matrix. (In the following analysis, each term is evaluated at an equilibrium solution given by $f = \mathbf{0}, g = \mathbf{0}$). We find:
\begin{align}
\jacparam{\vphi}{\tilde{f}} &= \jacphif - \frac{ \lrp h}{2} \left(\frac 1 m ( \jacphif)^2 +  \frac 1 m  \jactwoparam{\vphi}{\vphi} f f + \jacthetaf \jacphig   +  \jactwoparam{\vphi}{\vtheta} f  g \right)\\
&=\jacphif - \frac{ \lrp h}{2} \left(\frac 1 m ( \jacphif)^2 + \jacthetaf \jacphig   \right) \\
\jacparam{\vtheta}{\tilde{f}} &= \jacthetaf - \frac{ \lrp h}{2} \left(\frac 1 m \jacthetaf \jacphif  + \frac 1 m  \jactwoparam{\vtheta}{\vphi}{f} f + \jacthetaf \jacthetag  +  \jactwoparam{\vtheta}{\vtheta}{f} g \right)\\
&= \jacthetaf - \frac{ \lrp h}{2} \left(\frac 1 m \jacphif \jacthetaf  + \jacthetaf \jacthetag \right) \\
\jacparam{\vphi}{\tilde{g}} &= \jacphig - \frac{ \lrt h}{2} (\frac 1 k \jacthetag \jacphig + \frac 1 k  \jactwoparam{\vphi}{\vtheta}{g} g + (1 - \frac{2 \lrp}{\lrt})\jacphig \jacphif  + (1 - \frac{2 \lrp}{\lrt}) \jactwoparam{\vphi}{\vphi}{g} f) \\
&= \jacphig - \frac{ \lrt h}{2} (\frac 1 k \jacthetag \jacphig + (1 - \frac{2 \lrp}{\lrt})\jacphig \jacphif  )\\
\jacparam{\vtheta}{\tilde{g}} &= \jacthetag - \frac{ \lrt h}{2} \left(\frac 1 k (\jacthetag)^2 + \frac 1 k  \jactwoparam{\vtheta}{\vtheta}{g} g + (1 - \frac{2 \lrp}{\lrt}) \jacphig \jacthetaf + (1 - \frac{2 \lrp}{\lrt})   \jactwoparam{\vtheta}{\vphi}{g} f \right)\\
&= \jacthetag - \frac{ \lrt h}{2} \left(\frac 1 k (\jacthetag)^2+  (1 - \frac{2 \lrp}{\lrt}) \jacphig \jacthetaf \right)
\end{align}
where $\jactwoparam{\vphi}{\vtheta}{f} g$ is the matrix in $\mathbb{R}^{m, m}$ with $\left(\jactwoparam{\vphi}{\vtheta}{f} g\right)_{i,j} = \left(\jacparam{\vtheta}{\frac{\partial f_i}{\partial \vphi_j}}\right)^T g$.

Given these calculations, we can now write:
\begin{gather}
 \widetilde{\vJ}
 =
  \begin{bmatrix}
    \jacparam{\vphi}{\widetilde{f}} & \jacparam{\vtheta}{\widetilde{f}}\\
    \jacparam{\vphi}{\widetilde{g}} &
   \jacparam{\vtheta}{\widetilde{g}}
   \end{bmatrix}
   = \vJ - \frac{h}{2}\mathbf{K}_{\text{alt}}
\end{gather}
where $\vJ$ is the Jacobian of the unmodified flow:
\begin{gather}
 \vJ
 =
  \begin{bmatrix}
    \jacphif & \jacthetaf\\
    \jacphig & \jacthetag
   \end{bmatrix}
   \label{eq:jacobian}
\end{gather}
and
\begin{gather}
\mathbf{K}_{\text{alt}}
 =
  \begin{bmatrix}
    \frac {\lrp} {m} ( \jacphif)^2 + \lrp \jacthetaf \jacphig   & \frac{ \lrp} {m} \jacphif \jacthetaf +  \lrp \jacthetaf \jacthetag\\
   \frac{\lrt}{k} \jacthetag \jacphig+  \lrt (1 - \frac{2 \lrp}{\lrt}) \jacphig \jacphif&
   \frac {\lrt}{k}( \jacthetag)^2 +  \lrt (1 - \frac{2 \lrp}{\lrt}) \jacphig \jacthetaf
   \end{bmatrix}
\end{gather}

Using the modified Jacobian, we can use Remark~\ref{rem:stability_analysis} to determine the stability of fixed points.
A necessary condition for stability is that the trace of the modified Jacobian is less than or equal to zero (i.e. $\Tr(\widetilde{\vJ})\leq0$), since the trace is the sum of eigenvalues. Using the property of trace additivity and the trace cyclic property we see that:\begin{equation}\label{eqn:modified_trace_alternating}
\Tr(\widetilde{\vJ}) = \Tr(\vJ) - \frac{ h}{2}\left(\frac{\lrp}{m} \Tr((\jacphif)^2)+ \frac {\lrt}{k}  \Tr((\jacthetag)^2)    \right) - \frac{ h}{2}(\lrt - \lrp)\Tr( \jacthetaf \jacphig ).
\end{equation}
We note that unlike for simultaneous updates, even if $\Tr( \jacthetaf \jacphig )$ is negative, if $\lrt < \lrp$, the trace of the modified system will stay negative, so the necessary condition for the system to remain stable is still satisfied. However, since this is not a sufficient condition, the modified system could still be unstable.

\section{SGA in zero-sum games}
\label{app:sga}

For clarity, this section reproduces Symplectic Gradient Adjustment (SGA) from~\citet{balduzzi2018mechanics} for zero-sum games, to showcase the appearance of interaction terms. We have two players $\vphi$ and $\vtheta$ minimizing loss functions $E$ and $-E$, respectivelly, then SGA defines the vector
\begin{gather}
 \vxi
 =
  \begin{bmatrix}
    \nabla_{\vphi} E \\
   - \nabla_{\vtheta} E
   \end{bmatrix}.
   \label{eq:sga_neg_vector_field}
\end{gather}
They then define $\mathbf{H}_{\vxi}$ the game Hessian (their Section 2.2)
\begin{gather}
 \mathbf{H}_{\vxi} = 
  \begin{bmatrix}
    \jactwoparam{\vphi}{\vphi} E &  \jactwoparam{\vphi}{\vtheta} E \\
   -  \jactwoparam{\vtheta}{\vphi} E &
   - \jactwoparam{\vtheta}{\vtheta} E
   \end{bmatrix}.
   \label{eq:j}
\end{gather}
Thus
$\mathbf{H}_{\vxi}$ has the \emph{anti-symmetric} component
\begin{gather}
 \vA
 = \frac{1}{2}(\mathbf{H}_{\vxi} - \mathbf{H}_{\vxi}^T) =
  \frac{1}{2} \begin{bmatrix}
    0 & \jactwoparam{\vphi}{\vtheta} E +  \jactwoparam{\vtheta}{\vphi} E^T \\
    -  \jactwoparam{\vtheta}{\vphi} E - \jactwoparam{\vphi}{\vtheta} E^T &
   0
   \end{bmatrix} = \begin{bmatrix}
    0 & \jactwoparam{\vphi}{\vtheta} E  \\
    -  \jactwoparam{\vtheta}{\vphi} E &
   0
   \end{bmatrix}.
   \label{eq:j_anti_symmetric}
\end{gather}
The vector field $\vxi$ is modified according to SGA as
\begin{align}
 \hat{\vxi}
 = \vxi + \zeta \vA^T \, \vxi &= \begin{bmatrix}
    \nabla_{\vphi} E \\
   - \nabla_{\vtheta} E
   \end{bmatrix} + \zeta
  \begin{bmatrix}
    0 & \jactwoparam{\vphi}{\vtheta} E  \\
    -  \jactwoparam{\vtheta}{\vphi} E &
   0
   \end{bmatrix}  \begin{bmatrix}
    \nabla_{\vphi} E \\
   - \nabla_{\vtheta} E
   \end{bmatrix} \\
   &=  \begin{bmatrix}
    \nabla_{\vphi} E \\
   - \nabla_{\vtheta} E
   \end{bmatrix} + \zeta
  \begin{bmatrix}
     - \jactwoparam{\vphi}{\vtheta} E \nabla_{\vtheta} E \\
     - \jactwoparam{\vtheta}{\vphi} E \nabla_{\vphi} E 
   \end{bmatrix},
   \label{eq:j_sga}
\end{align}
where $\zeta$ is either a hyperparameter or adjusted dynamically in training~\citep{balduzzi2018mechanics}.
Thus, via Eqs~\eqref{eq:normscrossphi} and~\eqref{eq:normscrosstheta} the modified vector field can be simplified to
\begin{gather}
 \hat{\vxi}
   = \begin{bmatrix} \nabla_{\vphi} (E - \zeta \frac{1}{2} \norm{\nabla_{\vtheta}E}^2)\\
     \nabla_{\vtheta} (- E - \zeta \frac{1}{2} \norm{\nabla_{\vphi}E}^2)
   \end{bmatrix}.
   \label{eq:j_zero_su}
\end{gather}
Therefore, since $\hat{\vxi}$ defines the negative of the vector field followed by the system, the modified losses for the two players can be written respectively as
\begin{align}
\widetilde{L}_1 &= E - \frac{\zeta}{2}\norm{\nabla_{\vtheta}E}^2 \\
\widetilde{L}_2 &= -E - \frac{\zeta}{2}\norm{\nabla_{\vphi}E}^2.
\end{align}
Thus if $\zeta< 0$, which is what we use in our experiments, the functional form of the modified losses given by SGA is the same used to cancel the interaction terms of DD in the case of simultaneous gradient descent updates in zero sum games. We do however highlight a few differences in our approach compared to SGA: our approach extends to alternating updates and provides the optimal regularisation coefficients; cancelling the interaction terms of the drift is different compared to SGA for general games.

\section{DiracGAN - an illustrative example}
\label{app:dirac_gan}
In their work assessing the convergence of GAN training~\citet{mescheder2018training}
introduce the example of the DiracGAN, where the GAN is trying to learn a delta distribution with mass at zero.
More specifically, the generator $G(z; \theta) = \theta$ with parameter $\theta$ parametrises constant functions whose images $\{\theta\}$ correspond to the support of the delta distribution $\delta_\theta$. The discriminator is a linear model $D(x;\phi) = \phi \cdot x$ with
parameter $\phi$.
The loss function is given by
\begin{align}
E(\theta, \phi) = l(\theta \phi) + l(0),
\label{eq:supp_dirac_gan}
\end{align}
where $l$ depends on the GAN used; for the standard GAN it is $l = - \log (1 + e^{-t})$. As in~\citet{mescheder2018training}, we assume $l$ is continuously differentiable with $l'(x) \ne 0$ for all $x \in \mathbb{R}$. The partial derivatives of the loss function
\begin{align}
\frac{\partial E}{\partial\phi} = l'(\theta \phi) \theta, \hspace{3em}  \frac{\partial E}{\partial\theta} = l'(\theta \phi) \phi,
\end{align}
lead to the underlying continuous dynamics
\begin{align}
\dot{\phi} =  f(\theta, \phi) =  l'(\theta \phi) \theta, \hspace{3em} \dot{\theta} =  g(\theta, \phi) =  -l'(\theta \phi) \phi.
\label{eq:supp_dirac_gan_vector_field}
\end{align}

Thus the only equilibrium of the game is $\theta = 0$ and $\phi = 0$.

\subsection{Reconciling discrete and continuous updates in Dirac-GAN}
\label{sec:dirac_gan_reconciling}
\citet{mescheder2018training} observed a discrepancy between the continuous and discrete dynamics of DiracGAN.
They show that, for the problem in Eq~\eqref{eq:supp_dirac_gan}, the continuous dynamics preserve $\theta^2 + \phi^2$, and thus cannot converge (Lemma 2.3 in~\citet{mescheder2018training}), since
\begin{align}
\frac{d \left(\theta^2 + \phi^2\right)}{dt} = 2 \theta \frac{d \theta}{dt} + 2 \phi \frac{d \phi}{dt} = - 2 \theta l'(\theta \phi) \phi + 2 \phi l'(\theta \phi) \theta = 0.
\end{align}
They also observe that when following the discrete dynamics of simultaneous gradient descent that $\theta^2 + \phi^2$ increases in time (Lemma 2.4 in~\citet{mescheder2018training}). We resolve this discrepancy between conclusions reached with continuous-time and discrete-time analysis in DiracGAN, by showing that the modified continuous dynamics given by DD result in behaviour consistent with that of the discrete updates.

\begin{proposition} Following the continuous-time modified flow we obtained using BEA describing simultaneous gradient descent increases $\theta^2 + \phi^2$  in DiracGAN.
\end{proposition}
\begin{proof}

We assume both the generator and the discriminator use learning rates $h$, as in~\citet{mescheder2018training}. We first compute terms used by the zero-sum Colloraries~\ref{cor:zs-sim} and~\ref{cor:zs-alt} in the main thesis.
\begin{align}
\norm{\nabla_{\theta} E}^2 = l'(\theta \phi)^2 \phi^2 \\
\norm{\nabla_{\phi} E}^2 = l'(\theta \phi)^2 \theta^2
\end{align}
and
\begin{align}
\nabla_{\theta} \norm{\nabla_{\theta} E}^2 &= 2 \phi^3 l'(\theta \phi) l''(\theta \phi)\\
\nabla_{\theta} \norm{\nabla_{\phi} E}^2 &=  2 \theta l'(\theta \phi)^2 + 2 \theta^2 \phi l'(\theta \phi) l''(\theta \phi) \\
\nabla_{\phi} \norm{\nabla_{\theta} E}^2 &= 2 \phi l'(\theta \phi)^2 + 2 \phi^2 \theta l'(\theta \phi) l''(\theta \phi) \\
\nabla_{\phi} \norm{\nabla_{\phi} E}^2 &= 2 \theta^3  l'(\theta \phi) l''(\theta \phi).
\end{align}
Thus, the modified flows are given by:
\begin{align}
\dot{\phi} &=   l'(\theta \phi) \theta + \frac{h}{2} \left[- \theta^3  l'(\theta \phi) l''(\theta \phi) + \phi l'(\theta \phi)^2 +  \phi^2 \theta l'(\theta \phi) l''(\theta \phi)\right] \\
\dot{\theta} &= - l'(\theta \phi) \phi - \frac{h}{2} \left[ - \theta l'(\theta \phi)^2 - \theta^2 \phi l'(\theta \phi) l''(\theta \phi)  +  \phi^3 l'(\theta \phi) l''(\theta \phi) \right].
\end{align}
By denoting $l'(\theta \phi)$ by $l'$ and $l''(\theta \phi)$ by $l''$, then we have:
\begin{align}
&\frac{d \left(\theta^2 + \phi^2\right)}{dt}
= 2 \theta \frac{d \theta}{dt} + 2 \phi \frac{d \phi}{dt} \\
&= 2 \theta \left(- l' \phi - \frac{h}{2} \left[ - \theta l'^2 - \theta^2 \phi l'l''  +  \phi^3 l'l'' \right]\right) \nonumber \\
  & \hspace{1em}+  2\phi \left(l' \theta + \frac{h}{2} \left[- \theta^3  l'l'' +  \phi l'^2 +  \phi^2 \theta l'l''\right]\right) \\
  &= 2 \theta \left( - \frac{h}{2} \left[ - \theta l'^2 - \theta^2 \phi l'l''  +  \phi^3 l'l'' \right]\right)
    +  2 \phi \left(\frac{h}{2} \left[- \theta^3  l'l'' +  \phi l'^2 + \phi^2 \theta l'l''\right]\right) \\
  & = h \theta^2 l'^2 + h \theta^3 \phi l'l'' - h  \phi^3 \theta l'l'' - h \phi \theta^3 l'l'' + h \phi^2 l'^2 + h \phi^3 \theta l'l'' \\
  & =  h \theta^2 l'^2 + h \phi^2 l'^2 > 0
\end{align}
for all $\phi,\theta\neq 0$, which shows that $\theta^2 + \phi^2$ is not preserved and it will strictly increase for all values away from the equilibrium (we have used the assumption that $l'(x) \ne 0, \forall x \in \mathbb{R}$).
\end{proof}

We have thus identified a continuous system which exhibits the same behaviour as described by Lemma 2.4 in~\citet{mescheder2018training}, where a discrete system is analysed.

\subsection{DD changes the convergence behaviour of Dirac-GAN}
\label{sec:dirac-gan-stability}
The Jacobian of the unmodified Dirac-GAN evaluated at the equilibrium solution given by $\phi = 0$, $\theta = 0$. Since each parameter is one dimensional, we can write
\begin{gather}
 \vJ
 =
  \begin{bmatrix}
    \nabla_{\phi,\phi}E & \nabla_{\theta,\phi}E\\
   - \nabla_{\phi,\theta}E & - \nabla_{\theta,\theta}E
   \end{bmatrix}
   =
  \begin{bmatrix}
    0&  l'(0) \\
    -l'(0) & 0
   \end{bmatrix}
\end{gather}
We see that $\Tr(\vJ)=0$ and the determinant $|\vJ| = l'(0)^2 $. Therefore, the eigenvalues of this Jacobian are ${\lrt}_{\pm} = \Tr(\vJ)/2 \pm \sqrt{\Tr(\vJ)^2 - 4|J|}/2 = \pm  il'(0) $ (Reproduced from \citet{mescheder2018training}). This is an example of a stable \emph{center equilibrium}.

Next we calculate the Jacobian of the modified flows given by DiracGAN, evaluated at an equilibrium solution and find $\widetilde{\vJ} = \vJ - h\Delta/2$, where
\begin{align}
 \Delta
 =&
  \begin{bmatrix}
    \lrp ( \nabla_{\phi, \phi} E)^2 -  \lrp \nabla_{\phi,\theta}E\nabla_{\theta, \phi}E   & \lrp \nabla_{\theta, \phi}E\nabla_{\phi, \phi}E -  \lrp \nabla_{\theta,\theta}E\nabla_{\theta,\phi}E\\
   \lrt \nabla_{\phi,\theta}E\nabla_{\theta,\theta}E -  \lrt \nabla_{\phi,\phi}E\nabla_{\phi,\theta}E&
   \lrt( \nabla_{\theta,\theta} E)^2 -  \lrt \nabla_{\theta,\phi}E\nabla_{\phi,\theta}E
   \end{bmatrix}
   \\
   =&
     \begin{bmatrix}
     -  \lrp \nabla_{\phi,\theta}E\nabla_{\theta, \phi}E   & 0\\
   0& -  \lrt \nabla_{\theta,\phi}E\nabla_{\phi,\theta}E
   \end{bmatrix}
   \\
   =&
  \begin{bmatrix}
    -\lrp l'(0)^2 &  0 \\
    0 & -\lrt l'(0)^2
   \end{bmatrix}
\end{align}
so
\begin{gather}
 \widetilde{\vJ}
 =
  \begin{bmatrix}
    h\lrp l'(0)^2/2&  l'(0) \\
    -l'(0) & h\lrt l'(0)^2/2
   \end{bmatrix}.
\end{gather}
Now, we see that the trace of the modified Jacobian for the Dirac-GAN is $\Tr(\widetilde{\vJ})  = (h/2)(\lrp + \lrt ) l'(0)^2 > 0$, so the modified flows induced by gradient descent in DiracGAN are unstable.

\subsection{Explicit regularisation stabilises Dirac-GAN}
\label{sec:dirac-gan-exp-reg}

Here, we show that we can use our stability analysis to identify forms of explicit regularisation that can counteract the destabilizing impact of DD. Consider the Dirac-GAN with explicit regularisation of the following form:  $E_{\vphi}  = - E + \gamma{\| \nabla_{\theta} E\|}^2$ and $E_{\vtheta}  = E + \zeta {\| \nabla_{\phi} E\|}^2$ where $\phi_t = \phi_{t-1}  -  \lrp h \nabla_{\phi} E_{\vphi} $ and  $\theta_t = \theta_{t-1}  -  \lrt h \nabla_{\theta} E_{\vtheta}  $ and with $\gamma, \zeta \sim \mathcal{O}(h)$. The modified Jacobian for this system is given by
\begin{align}
 \widetilde{\vJ}
   =&
  \begin{bmatrix}
    h\lrp/2 \nabla_{\phi,\theta}E\nabla_{\theta, \phi}E - \gamma \nabla_{\phi, \phi} \norm{ \nabla_{\theta}E}^2  & \nabla_{\theta,\phi}E\\
   -\nabla_{\phi,\theta}E & h\lrt/2 \nabla_{\theta, \phi}E\nabla_{\phi, \theta}E -\zeta \nabla_{\theta, \theta} \norm{ \nabla_{\phi}E}^2
   \end{bmatrix}
   \\
   =&
  \begin{bmatrix}
    (h\lrp/2 -2 \gamma)l'(0)^2  &  l'(0) \\
    -l'(0) &  (h\lrt/2 -2\zeta)l'(0)^2
   \end{bmatrix}
\end{align}
The determinant of the modified Jacobian is $|\widetilde{J}| =  (h\lrp/2 -2\gamma)(h\lrt/2 -2\zeta)l'(0)^4 + l'(0)^2$ and the trace is $\Tr(\widetilde{\vJ})  = (h\lrp/2 -2 \gamma)l'(0)^2 + (h\lrt/2 -2\zeta)l'(0)^2$. A necessary and sufficient condition for asymptotic stability is  $|\widetilde{\vJ}|>0 $ and $\Tr(\widetilde{\vJ}) <0$ (since this guarantees that the eigenvalues of the modified Jacobian have negative real part). Therefore, if $\gamma > h\lrp /4$ and $\zeta > h\lrt /4$, the system is asymptotically stable. We note however that in practice, when using discrete updates, the exact threshold for stability will have a $\mathcal{O}(h^3)$ correction, arising from the $\mathcal{O}(h^3)$ error in our BEA. Also, we see that when  $\gamma = h\lrp /4$ and $\zeta = h\lrt /4$, the contribution of the cross-terms is cancelled out (up to an $\mathcal{O}(h^3)$ correction). In Figure~\ref{fig:dirac_gan_explicit}, we see an example where this explicit regularisation stabilises the DiracGAN, so that it converges toward the equilibrium solution.

\section{The PF for games: linearisation results around critical points}
\label{sec:pf_linearisation_games}

Assume we are in the general setting described in Section~\ref{sec:bea_pf_games}. To simplify our notation we denote $[\vphi, \vtheta] = \vpsi$ and 
the vector  $[\nabla_{\vphi} E_{\vphi}, \nabla_{\vtheta} E_{\vtheta}]$ as $u(\vpsi)$. Thus our discrete simultaneous gradient descent updates become
\begin{align}
\vpsi_t = \vpsi_{t-1} - h u(\vpsi_{t-1}).
\label{eq:euler_gen_pf}
\end{align}
We also note that from here on we no longer use the structure of the vector field, so the below proof applies to any Euler update.

Assume we are around a critical point $\vpsi^*$, i.e. $u(\vpsi^*) = 0$. We can write
\begin{align}
 u(\vpsi) \approx  u(\vpsi^*) + \nablavpsiu(\vpsi) (\vpsi - \vpsi^*) = \nablavpsiu(\vpsi^*) (\vpsi - \vpsi^*) .
\label{eq:critical_point_approx_games}
\end{align}
Replacing this in the discrete updates in Eq~\eqref{eq:euler_gen_pf}
\begin{align}
 \vpsi_t - \vpsi^* \approx  (\vI- h \nablavpsiu(\vpsi^*)) (\vpsi_{t-1} - \vpsi^*)  \approx   (\vI- h \nablavpsiu(\vpsi^*))^t (\vpsi_{0} - \vpsi^*) .
\end{align}
If we now take the Jordan normal form of $\nablavpsiu(\vpsi^*) =  \vP^{-1} \vJ \vP$,
we can write
\begin{align}
 \vpsi_t - \vpsi^* &\approx  (\vI- h \nablavpsiu(\vpsi^*))^t (\vpsi_{0} - \vpsi^*) \\
                   &=   (\vI- h \vP^{-1} \vJ \vP)^t (\vpsi_{0} - \vpsi^*) \\
                   &=  \vP^{-1} (\vI- h \vJ)^t  \vP(\vpsi_{0} - \vpsi^*). 
\end{align}
Under the above approximation
\begin{align}
\vpsi_t =\vpsi^* + \vP^{-1} (\vI- h \vJ)^t  \vP(\vpsi_{0} - \vpsi^*),
\label{eq:linearisation_expansion_vpsi_t}
\end{align}
which we can write in continuous form as (we use that iteration $n$ is at time $t =nh$):
\begin{align}
\vpsi(t) =\vpsi^* + \vP^{-1} (\vI- h \vJ)^{t/h}  \vP(\vpsi_{0} - \vpsi^*) \\
         = \vpsi^* + \vP^{-1} e^{\log (\vI- h \vJ)^{t/h}}  \vP(\vpsi_{0} - \vpsi^*).
\end{align}
Taking the derivative w.r.t. $t$
\begin{align}
\dot{\vpsi} &= \frac{1}{h} \vP^{-1}  \log (\vI- h \vJ)e^{\log (\vI- h \vJ)^{t/h}} \vP(\vpsi_0 - \vpsi^*) \\
            &=  \frac{1}{h} \vP^{-1}  \log (\vI- h \vJ)e^{\log (\vI- h \vJ)^{t/h}} \vP \vP^{-1} (e^{\log (\vI- h \vJ)^{t/h}})^{-1}\vP (\vpsi - \vpsi^*) \\
            &=  \frac{1}{h} \vP^{-1}  \log (\vI- h \vJ)\vP (\vpsi - \vpsi^*).
\end{align}
From here, if we assume $u(\vpsi^*)$ is invertible we can write from the approximation above (Eq. \eqref{eq:critical_point_approx_games}) that $\vpsi - \vpsi^* = \nablavpsiu(\vpsi^*)^{-1} u(\vpsi) = \vP^{-1} \vJ^{-1} \vP u(\vpsi) $
and replacing this in the above we obtain
\begin{align}
\dot{\vpsi} &= \frac{1}{h} \vP^{-1}  \log (\vI- h \vJ) \vP \vP^{-1} \vJ^{-1} \vP u(\vpsi) \\
            &=  \frac{1}{h} \vP^{-1}  \log (\vI- h \vJ) \vJ^{-1} \vP u(\vpsi),
\end{align}
which is the generalisation of the PF presented in Eq~\eqref{eq:gen_pf_games}.

\section{GAN Experimental details}

Unless otherwise specified, all GAN experiments use the CIFAR-10 dataset and the CIFAR-10 convolutional SN-GAN architectures---see Table 3 in Section B.4 in~\citet{miyato2018spectral}.

\noindent \textbf{Libraries}: We use JAX~\citep{jax2018github} to implement our models, with Haiku~\citep{haiku2020github} as the neural network library, and Optax~\citep{optax2020github} for optimisation.

\noindent \textbf{Computer Architectures}: All models are trained on NVIDIA V100 GPUs. Each model is trained on 4 devices.

\subsection{Implementing explicit regularisation}
\label{app:unbiased_est}

The loss functions we are interested are of the form
\begin{align}
  L = \mathbb{E}_{p(\vx)} f_{\vtheta}(\vx)
\end{align}
We then have
\begin{align}
  \nabla_{\vtheta_i} L &= \mathbb{E}_{p(\vx)}  \nabla_{\vtheta_i} f_{\vtheta}(\vx) \\
\norm{\nabla_{\vtheta} L}^2 &= \sum_{i=1}^{i = |\vtheta|} \left(\nabla_{\vtheta_i} L\right)^2 =  \sum_{i=1}^{i = |\vtheta|} \left(\mathbb{E}_{p(\vx)}  \nabla_{\vtheta_i} f_{\vtheta}(\vx)\right)^2
\end{align}
to obtain an unbiased estimate of the above using samples, we have that:
\begin{align}
\norm{\nabla_{\vtheta} L}^2 &=  \sum_{i=1}^{i = |\vtheta|} \left(\mathbb{E}_{p(\vx)}  \nabla_{\vtheta_i} f_{\vtheta}(\vx)\right)^2 \\
& = \sum_{i=1}^{i = |\vtheta|} (\mathbb{E}_{p(\vx)}  \nabla_{\vtheta_i} f_{\vtheta}(\vx)) (\mathbb{E}_{p(\vx)}  \nabla_{\vtheta_i} f_{\vtheta}(\vx)) \\
& \approx \sum_{i=1}^{i = |\vtheta|} \left(\frac{1}{N} \sum_{k=1}^{N} \nabla_{\vtheta_i} f_{\vtheta}(\widehat{\vx_{1,k}})\right) \left(\frac{1}{N} \sum_{j=1}^{N}\nabla_{\vtheta_i} f_{\vtheta}(\widehat{\vx_{2,j}})\right)
\end{align}
so we have to use two sets of samples $\vx_{1,k} \sim p(\vx)$ and $\vx_{2,j} \sim p(\vx)$ from the true distribution (by splitting the batch into two or using a separate batch) to obtain the correct norm. To compute an estimator for $\nabla_{\vphi} \norm{\nabla_{\vtheta} L}^2$, we can compute the gradient of the above unbiased estimator of $\norm{\nabla_{\vtheta} L}^2$. However, to avoid computing gradients for two sets of samples, we derive another unbiased gradient estimator, which we use in all our experiments:
\begin{align}
\frac{2}{N} \sum_{i=1}^{i = |\vtheta|} \sum_{k=1}^N \nabla_{\vphi} \nabla_{\vtheta_i} f_{\vtheta}(\widehat{\vx_{1, j}}) \nabla_{\vtheta_i} f_{\vtheta}(\widehat{\vx_{2, j}})
\label{eq:unbiased_grads_one_backprop_N_terms}
\end{align}

\section{Additional experimental results}
\label{sec:dd_gan_exp_res_app}

\subsection{Additional results using zero-sum GANs}

We show additional results showcasing the effect of DD on zero-sum games in Figure~\ref{fig:supp_idd_zero_sum_games}. We see that not only do simultaneous updates perform worse than alternating updates when using the best hyperparameters, but that simultaneous updating is much more sensitive to hyperparameter choice. We also see that multiple updates can improve the stability of a GAN trained using zero-sum losses, but this strongly depends on the choice of learning rate.

 \begin{figure}[t]
 \centering
 \begin{subfloat}[Hyperparameter sensitivity.]{
  \includegraphics[width=0.45\columnwidth]{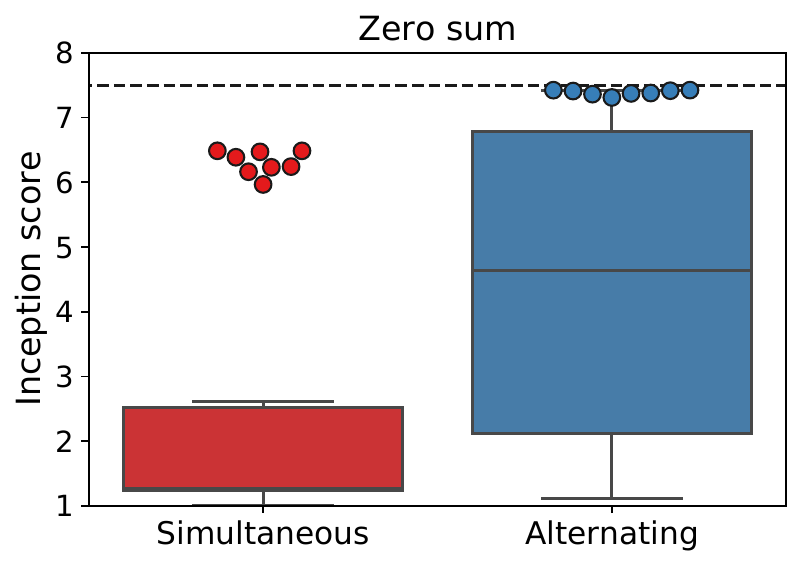}%
 }
 \end{subfloat}%
  \begin{subfloat}[Learning curves.]{
  \includegraphics[width=0.45\columnwidth]{best_sim_vs_alt_zero_sum_is}
 }
 \end{subfloat}\\
  \begin{subfloat}[The effect of multiple alternating updates.]{
  \includegraphics[width=0.45\columnwidth]{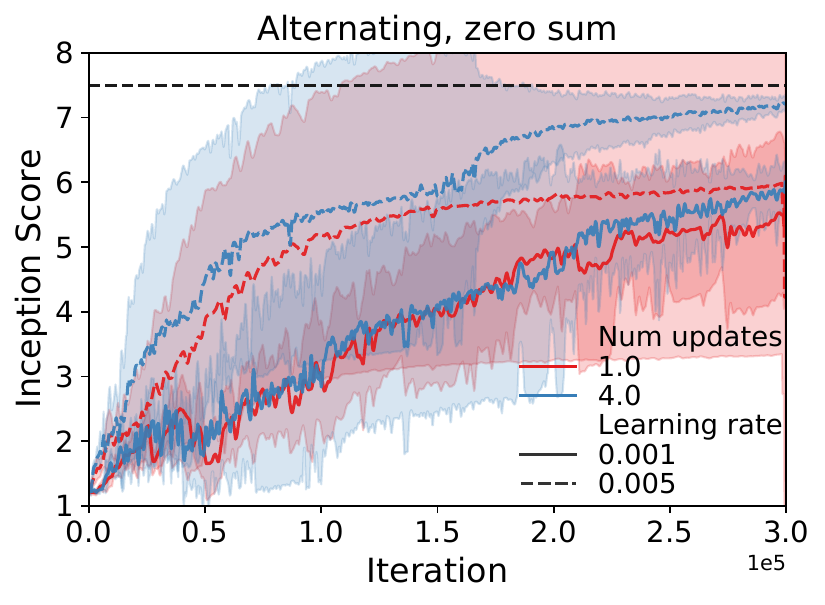}
 }
 \end{subfloat}%
   \begin{subfloat}[Equal learning rates.]{
    \includegraphics[width=0.45\columnwidth]{rk_euler_comparison_saturating_no_regularization}
 }
 \end{subfloat}%
    \caption{The effect of discretisation drift on zero-sum games; using the saturating generator loss.}
   \label{fig:supp_idd_zero_sum_games}
\end{figure}

\textbf{Least squares GANs}.
In order to assess the robustness of our results independent of the GAN loss used, we perform additional experimental results using GANs trained with a least square loss (LS-GAN~\citep{ls_gan}). We show results in Figure~\ref{fig:idd_zero_sum_games_ls_gan}, where we see that for the least square loss too, the learning rate ratios for which the generator drift does not maximise the discriminator norm (learning rate ratios above or equal to 0.5) perform best and exhibit less variance.

 \begin{figure}[t]
 \centering
  \begin{subfloat}[Hyperparameter sensitivity.]{
  \includegraphics[width=0.49\columnwidth]{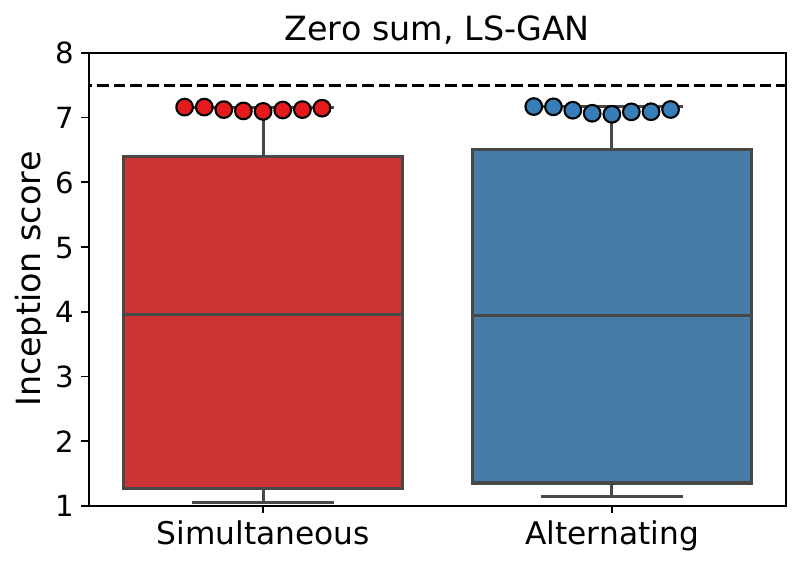}%
  \label{fig:square_loss_gan_sensitivity}
 }
 \end{subfloat}%
  \begin{subfloat}[The effect of learning rate ratios for alternating updates.]{
  \includegraphics[width=0.49\columnwidth]{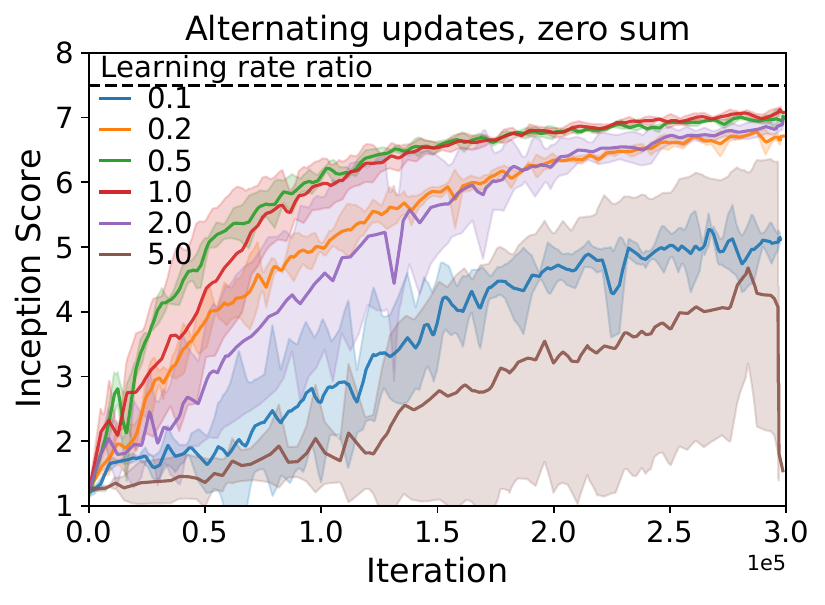}%
  \label{fig:square_loss_gan_ratios}
 }
 \end{subfloat}%
  \caption{The effect of discretisation drift on zero-sum games; least square losses.}
   \label{fig:idd_zero_sum_games_ls_gan}
\end{figure}

\subsection{GANs using the non-saturating loss}

Next, we explore how the strength of the DD depends on the game dynamics. We do this by comparing the \textit{relative effect} that numerical integration schemes have across different games. To this end, we consider the non-saturating loss introduced in the original GAN paper ($- \log D(G(\vz; \vtheta); \vphi)$). This loss has been extensively used since it helps to avoid problematic gradients early in training. When using this loss, we see that there is little difference between simultaneous and alternating updates (Figure~\ref{fig:supp_non_saturating}), unlike the saturating loss case. These results demonstrate that since DD depends on the underlying dynamics, it is difficult to make general game-independent predictions about which numerical integrator will perform best.

\begin{figure*}[t]
  \centering
  \begin{subfloat}[Hyperparameter sensitivity.]{
    \includegraphics[width=0.45\columnwidth]{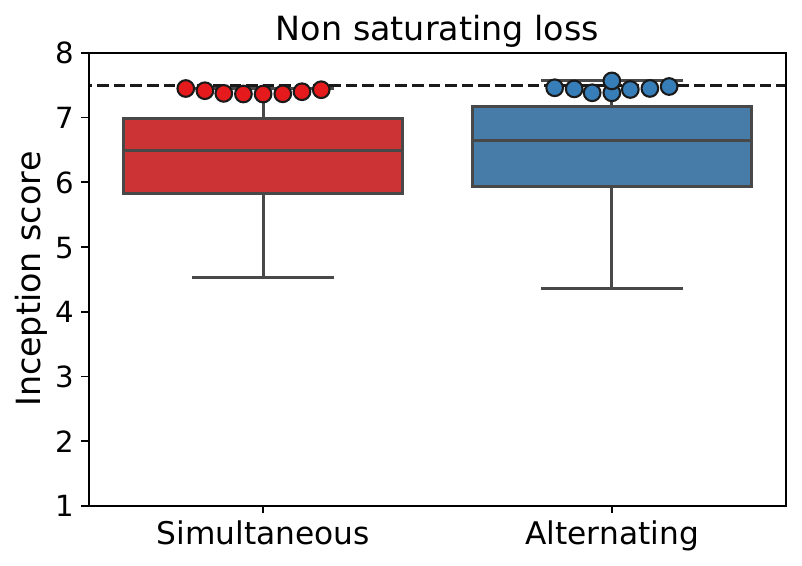}%
  }
  \end{subfloat}
  \begin{subfloat}[Learning curves.]{
    \includegraphics[width=0.45\columnwidth]{best_sim_vs_alt_zero_non_satuarting_is}
  }
  \end{subfloat}\\
  \begin{subfloat}[The effect of multiple alternating updates.]{
    \includegraphics[width=0.45\columnwidth]{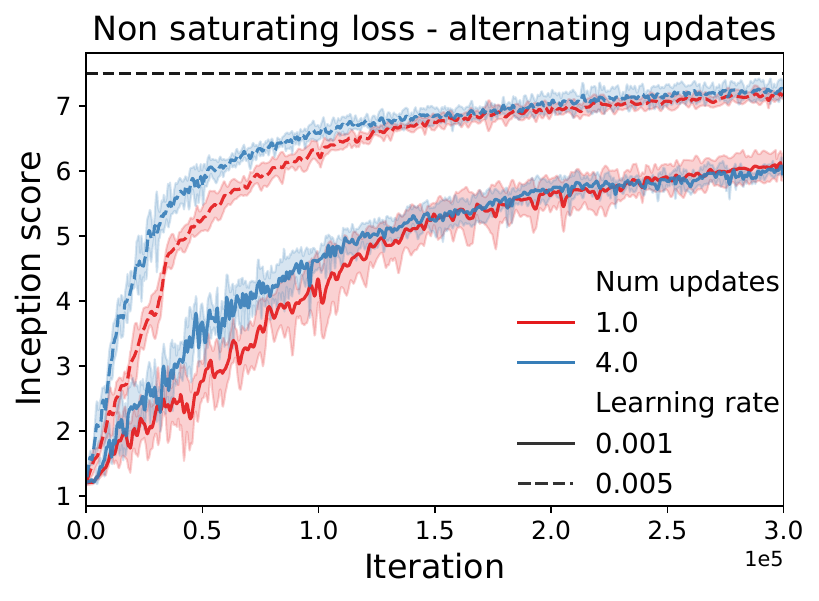}%
  }
  \end{subfloat}
   \begin{subfloat}[Equal learning rates.]{
    \includegraphics[width=0.45\columnwidth]{rk_euler_comparison_non_saturating_no_regularization}%
  }
  \end{subfloat}
  \caption[The effect of discretisation drift depends on the game.]{The effect of discretisation drift depends on the game: with the non saturating loss, the relative performance of different numerical estimators is different compared to the saturating loss, and the effect of explicit regularisation is also vastly different.}
  \label{fig:supp_non_saturating}
\end{figure*}

\subsection{Explicit regularisation in zero-sum games trained using simultaneous gradient descent}

We now show additional experimental results and visualisations obtained using explicit regularisation obtained using the original zero sum GAN objective, as presented in~\citet{goodfellow2014generative}.

We show the improvement that can be obtained compared to gradient descent with simultaneous updates by cancelling the interaction terms in Figure~\ref{fig:sgd_vs_cancel_drift_interaction}.
We additionally show results obtained from strengthening the self terms in Figure~\ref{fig:sgd_vs_cancel_drift_interaction_all_types_reg}.

In Figure~\ref{fig:adam_comparison}, we show that by cancelling interaction terms, SGD becomes a competitive optimisation algorithm when training the original GAN. We also note that in the case of Adam, while convergence is substantially faster than with SGD, we notice a degrade in performance later in training. This is something that has been observed in other works as well (e.g.~\citep{odegan}).

\begin{figure}[t]
 \centering
  \begin{subfloat}[Inception Score ($\uparrow$).]{
  \includegraphics[width=0.45\columnwidth]{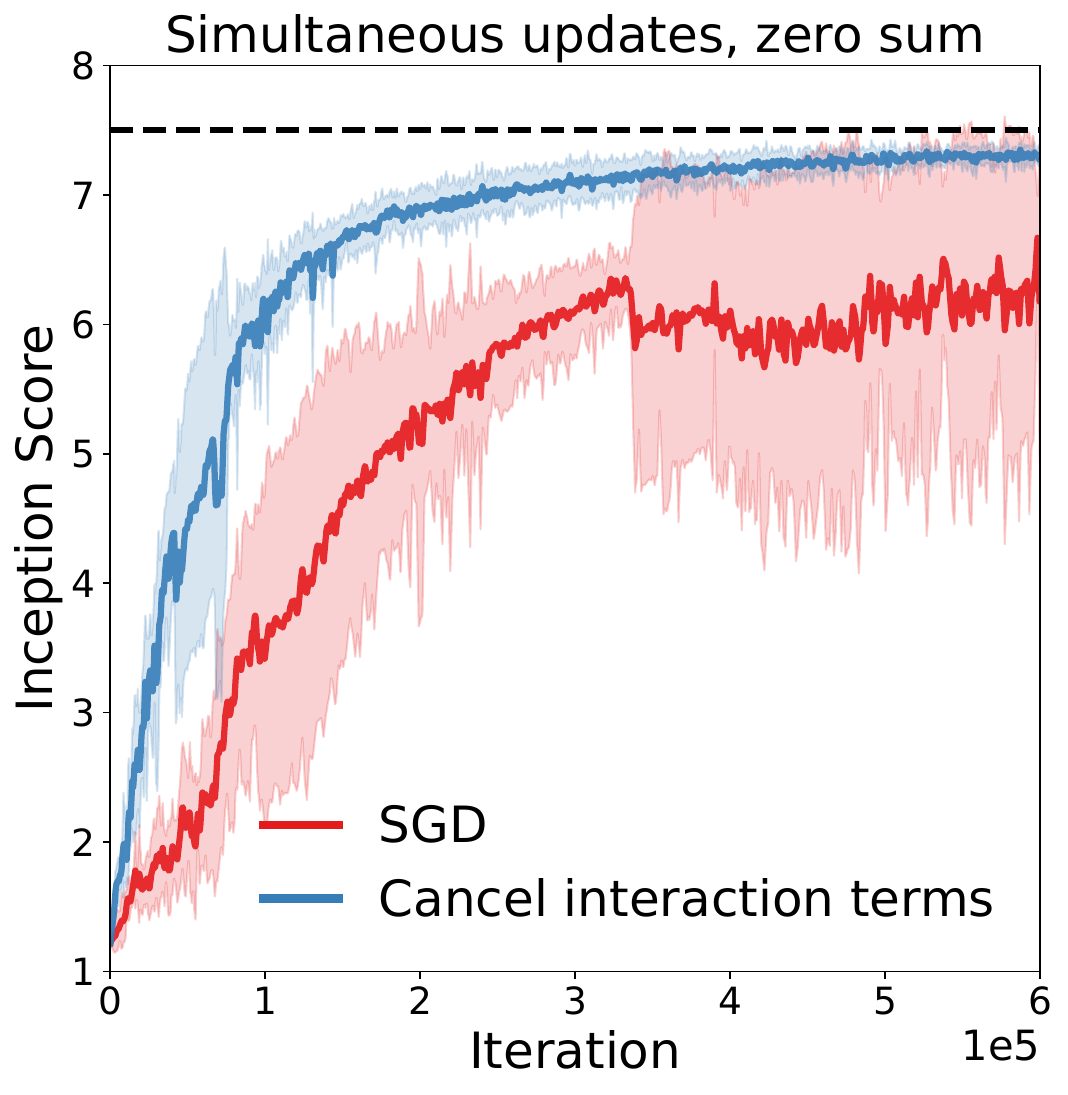}%
} \end{subfloat}
\hspace{2em}
 \begin{subfloat}[Fréchet Inception Distance ($\downarrow$).]{
  \includegraphics[width=0.45\columnwidth]{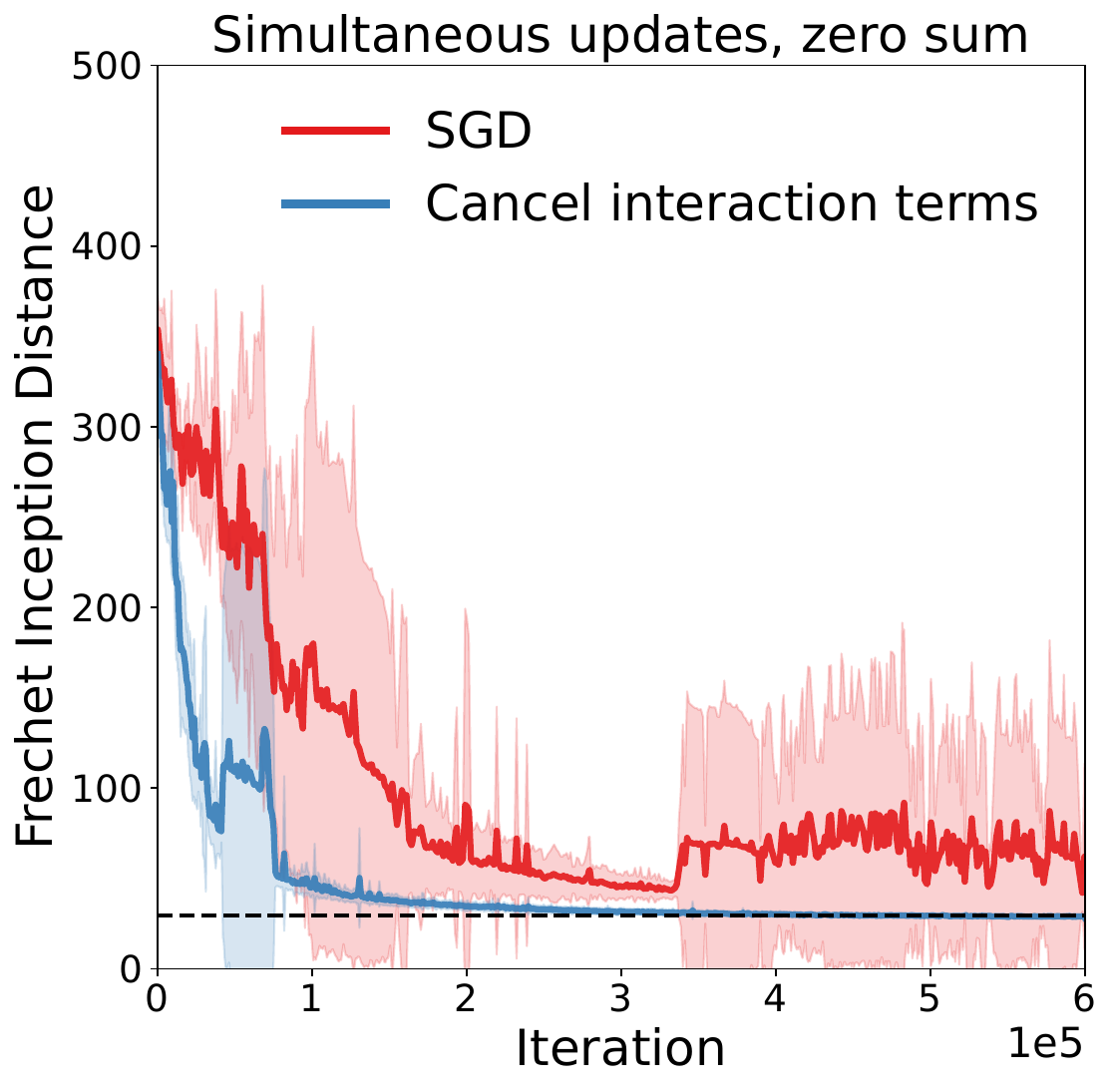}%
} \end{subfloat}%
  \caption[Using explicit regularisation to cancel the interaction terms of discretisation drift leads to a substantial performance improvement.]{Using explicit regularisation to cancel the effect of the interaction components of drift leads to a substantial improvement compared to SGD without explicit regularisation.}
  \label{fig:sgd_vs_cancel_drift_interaction}
\end{figure}

\begin{figure}[t]
 \centering
  \begin{subfloat}[Inception Score ($\uparrow$).]{
  \includegraphics[width=0.49\columnwidth]{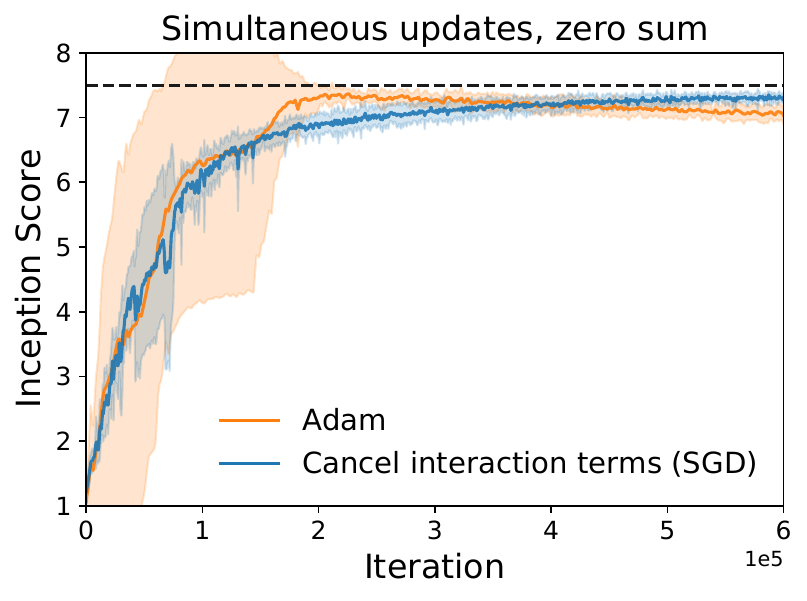}%
} \end{subfloat}%
 \begin{subfloat}[Fréchet Inception Distance ($\downarrow$).]{
  \includegraphics[width=0.49\columnwidth]{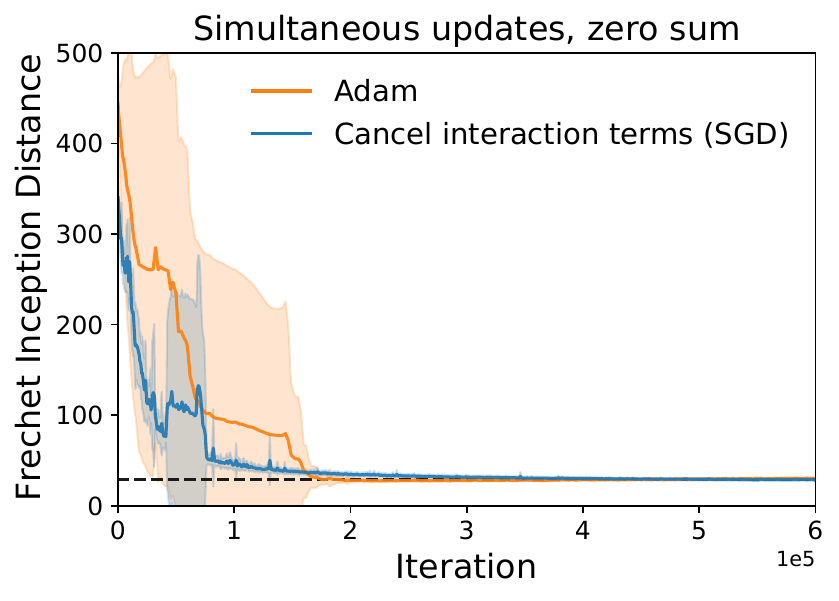}%
} \end{subfloat}%
  \caption[Using explicit regularisation to cancel the interaction terms of discretisation drift obtains the same peak performance as Adam.]{Using explicit regularisation to cancel the effect of the interaction components of drift allows us to obtain the same peak performance as Adam using SGD without momentum.}
  \label{fig:adam_comparison}
\end{figure}

\begin{figure}[t]
 \centering
  \begin{subfloat}[Inception Score ($\uparrow$).]{
  \includegraphics[width=0.49\columnwidth]{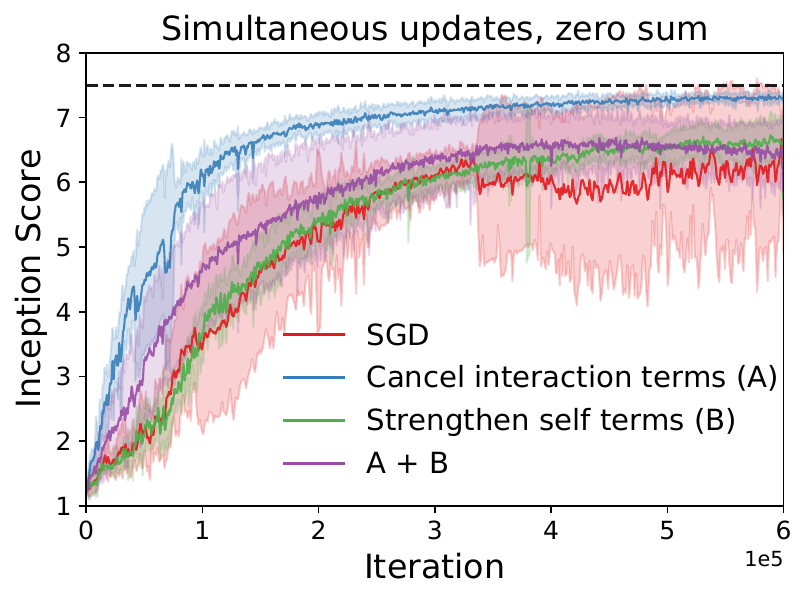}%
} \end{subfloat}%
 \begin{subfloat}[Fréchet Inception Distance ($\downarrow$).]{
  \includegraphics[width=0.49\columnwidth]{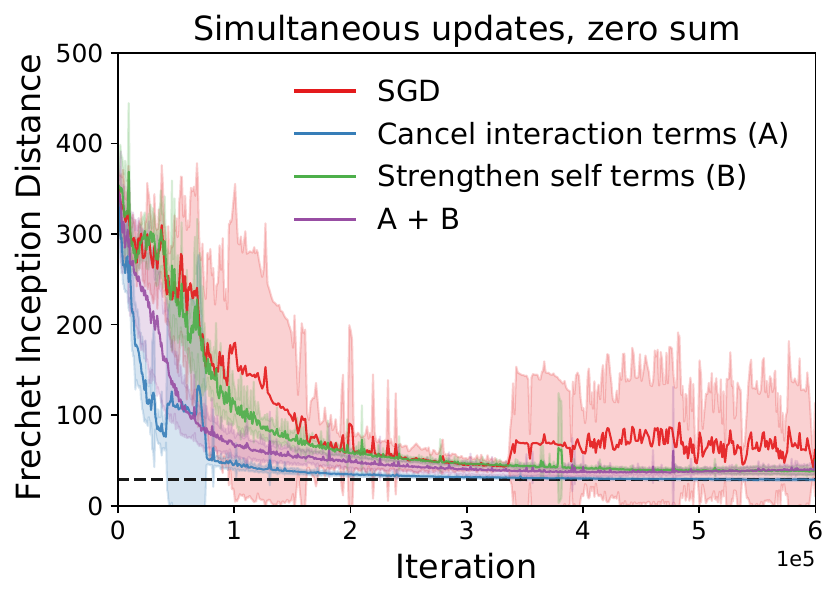}%
} \end{subfloat}%
  \caption[The form of discretisation drift allows us to construct efficient regularisers without the need of a hyperparameter sweep.]{Using explicit regularisation to cancel the effect of the drift leads to a substantial improvement compared to SGD without explicit regularisation. Strengthening the self terms---the terms which minimise the player's own norm---does not lead to a substantial improvement; this is somewhat expected since while the modified flows give us the exact coefficients required to \textit{cancel} the drift, they do not tell us how to strengthen it, and our choice of exact coefficients from the drift might not be optimal.}
  \label{fig:sgd_vs_cancel_drift_interaction_all_types_reg}
\end{figure}

\textbf{More percentiles}.
Throughout the main thesis, we displayed the best $10\%$ performing models for each optimisation algorithm used. We now expand that to show performance results across the best $20\%$ and $30\%$ of models in Figure~\ref{fig:multiple_percentages_sgd_cancel_interaction_terms}. We observe a consistent increase in performance obtained by cancelling the interaction terms.

\begin{figure}[t]
 \centering
  \begin{subfloat}[Top 10\% models.]{
  \includegraphics[width=0.33\columnwidth]{cancel_interaction_sgd_comparison_simultaneous_is_10}%
} \end{subfloat}%
 \begin{subfloat}[Top 20\% models.]{
  \includegraphics[width=0.33\columnwidth]{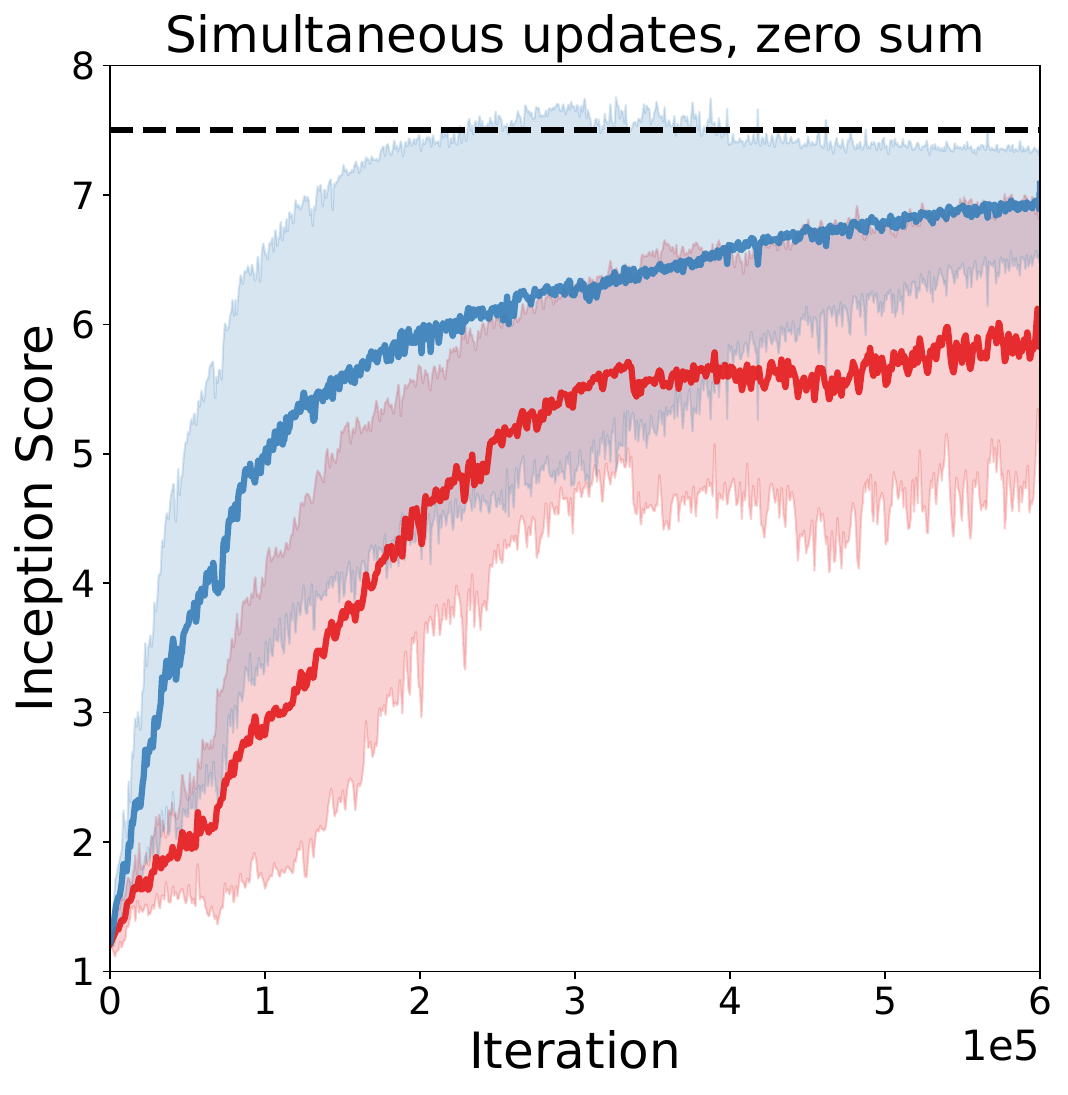}%
} \end{subfloat}%
\begin{subfloat}[Top 30\% models.]{
  \includegraphics[width=0.33\columnwidth]{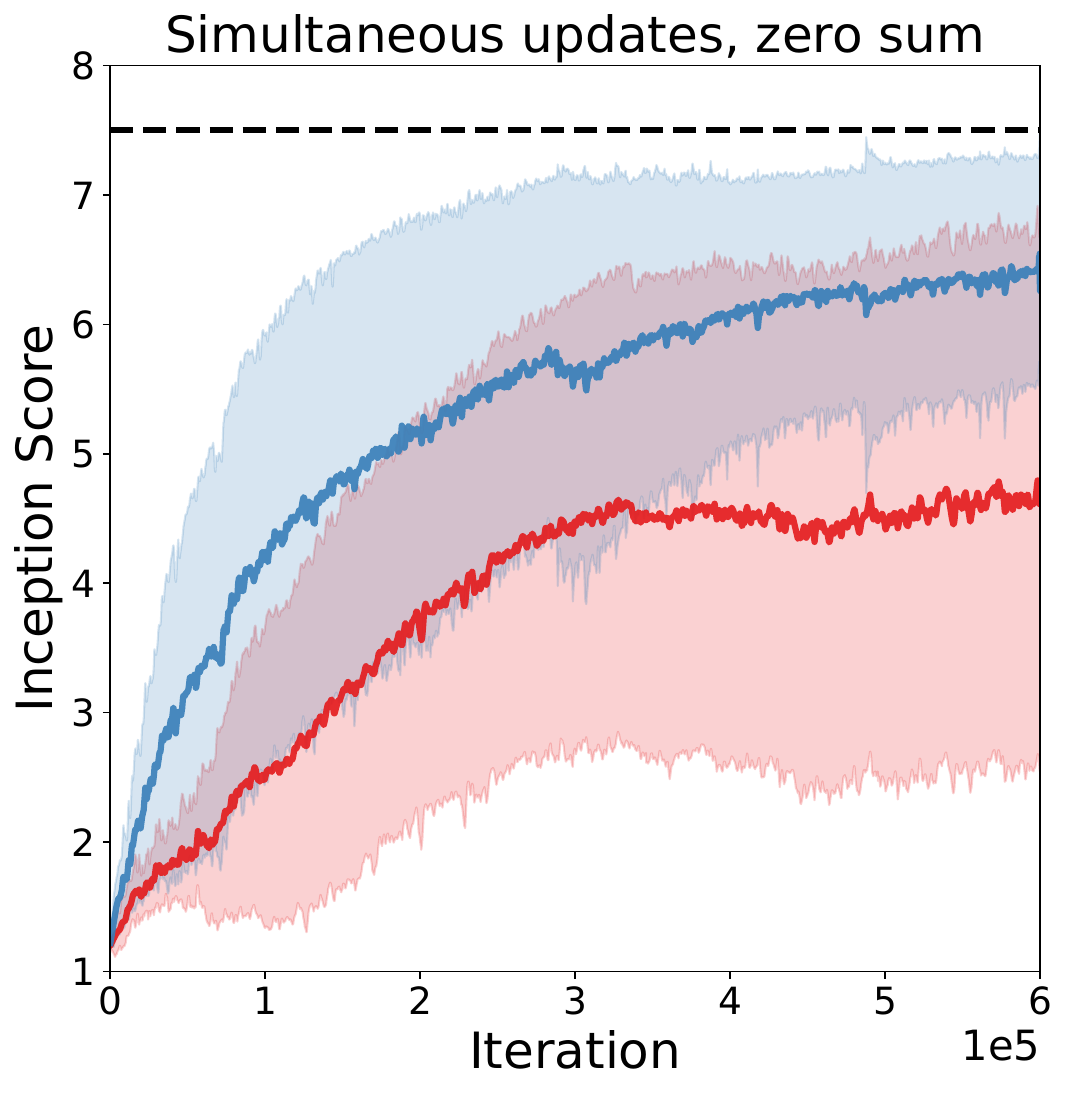}%
} \end{subfloat}%
  \caption[Cancelling interaction terms increases performance across multiple percentages used to select top models.]{Performance across the top performing models for vanilla SGD and with cancelling the interaction terms. Across all percentages, cancelling the interaction terms improves performance.}
  \label{fig:multiple_percentages_sgd_cancel_interaction_terms}
\end{figure}

\textbf{Batch size comparison}.
We now show that the results which show the efficacy of cancelling the interaction terms are resilient to changes in batch size in Figure~\ref{fig:batch_size_comparison}.

\begin{figure}[t]
 \centering
  \begin{subfloat}[Batch size 64.]{
  \includegraphics[width=0.33\columnwidth]{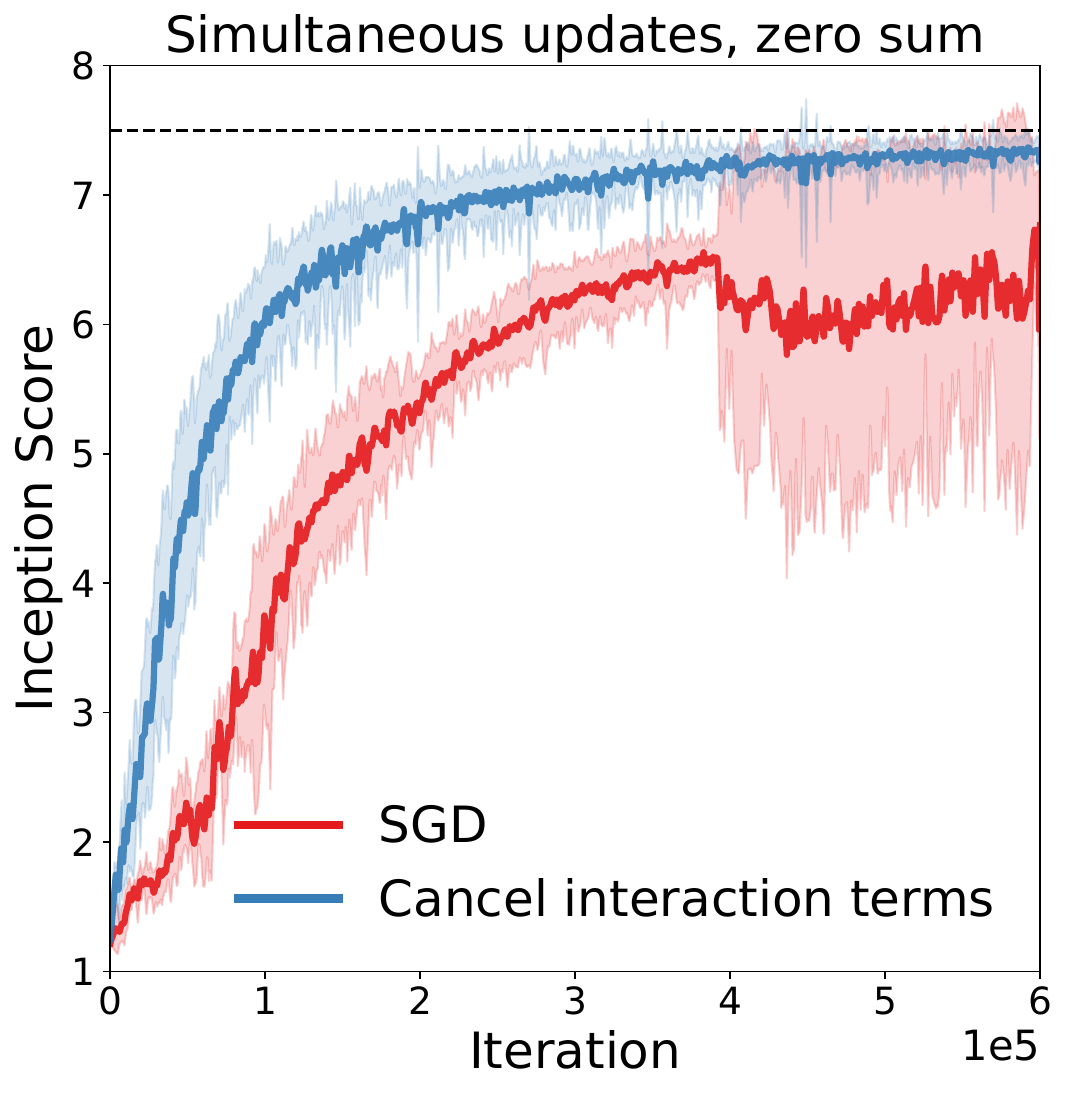}%
} \end{subfloat}%
 \begin{subfloat}[Batch size 128.]{
  \includegraphics[width=0.33\columnwidth]{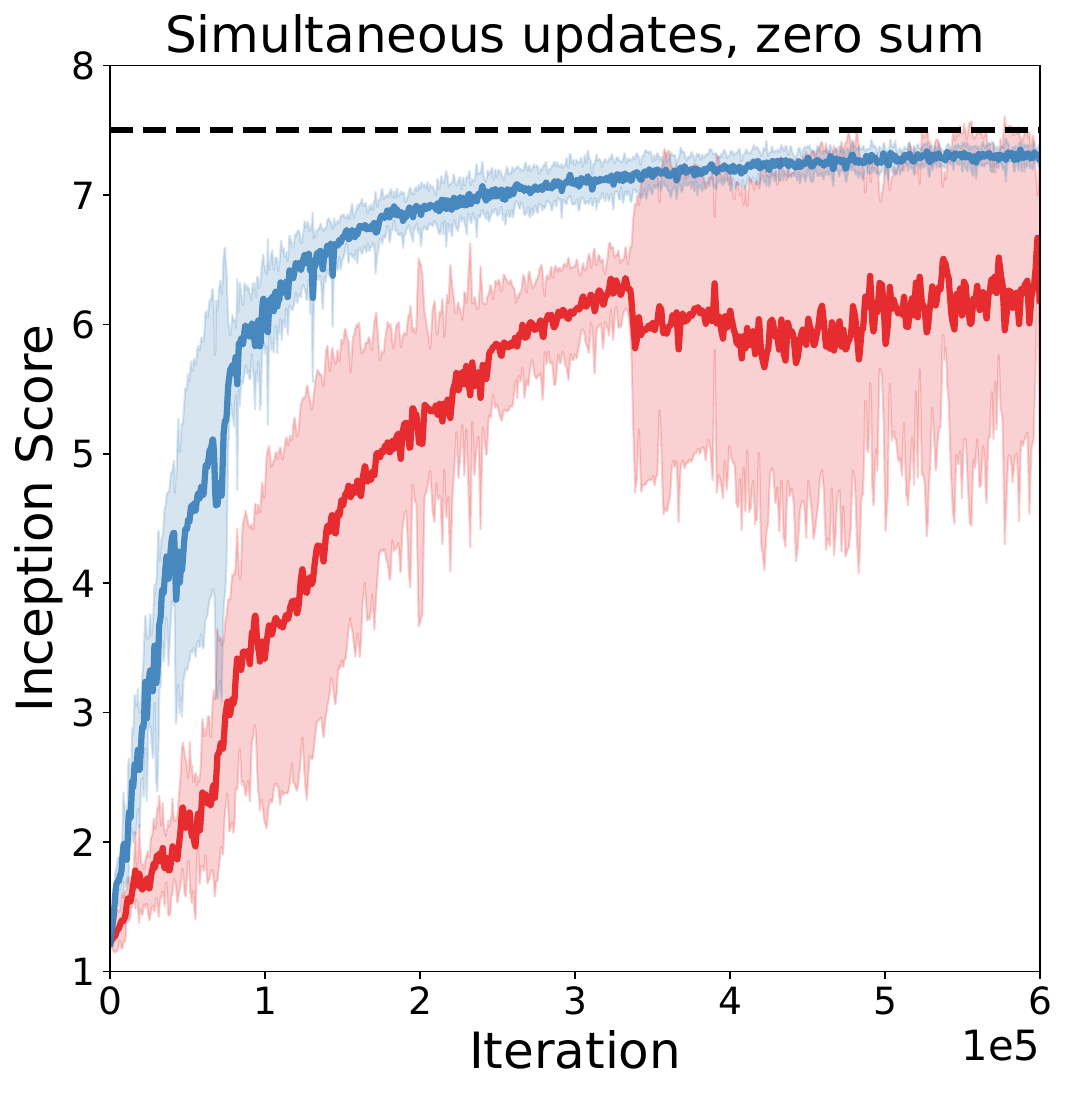}%
} \end{subfloat}%
\begin{subfloat}[Batch size 256.]{
  \includegraphics[width=0.33\columnwidth]{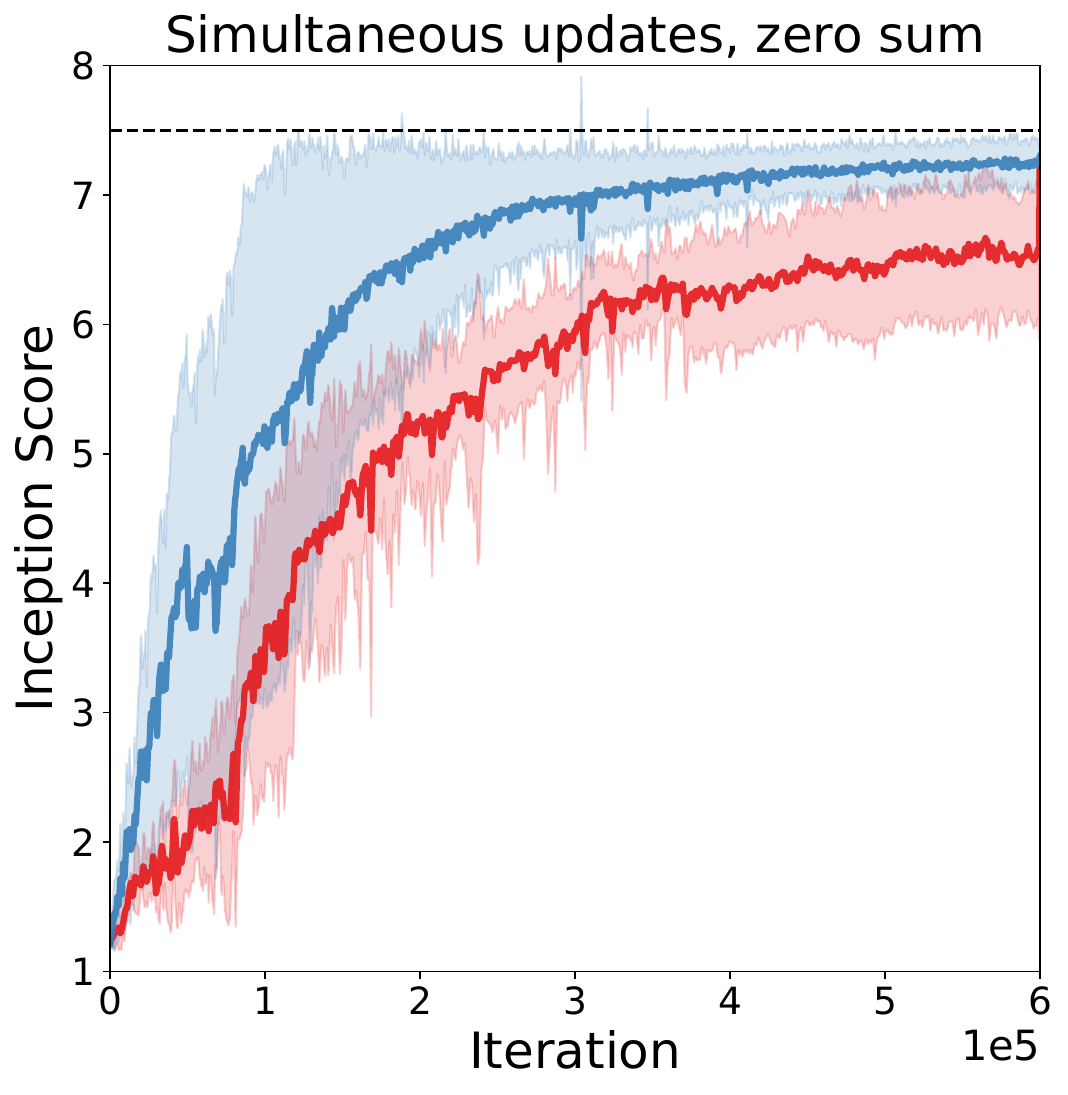}%
} \end{subfloat}%
  \caption[Cancelling interaction terms increases performance across batch sizes.]{Performance when changing the batch size. We consistently see that cancelling te interaction terms improves performance.}
  \label{fig:batch_size_comparison}
\end{figure}

\textbf{Comparison with Symplectic Gradient Adjustment}.
We show results comparing with Symplectic Gradient Adjustment (SGA)~\citep{balduzzi2018mechanics} in Figure~\ref{fig:supp_sga_comp} (best performing models) and Figure~\ref{fig:sga_box_plots} (quantiles showing performance across all hyperparameters and seeds). We observe that despite having the same functional form as SGA, cancelling the interaction terms of DD performs better; this is due to the choice of regularisation coefficients, which in the case of cancelling the interaction terms is provided by Corollary~\ref{cor:zs-sim}.

\begin{figure}[t]
 \centering
  \begin{subfloat}[Inception Score ($\uparrow$).]{
  \includegraphics[width=0.49\columnwidth]{explicit_regularization_sim_updates_comparison_with_sga_is}%
} \end{subfloat}%
 \begin{subfloat}[Fréchet Inception Distance ($\downarrow$).]{
  \includegraphics[width=0.49\columnwidth]{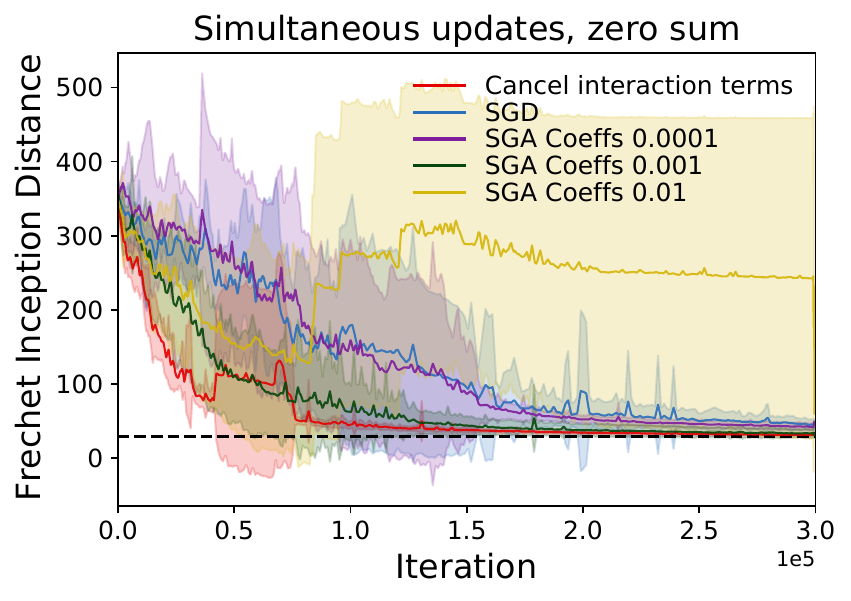}%
} \end{subfloat}%
  \caption[Comparing cancelling interactions terms with Symplectic Gradient Adjustment.] {Comparison with Symplectic Gradient Adjustment (SGA). Cancelling the interaction terms results in similar performance, but with less variance. The performance of SGA heavily depends on the strength of regularisation, adding another hyperparmeter to the hyperparameter sweep, while cancelling the interaction terms of the drift requires no other hyperparameters, since the explicit regularisation coefficients strictly depend on learning rates.}
  \label{fig:supp_sga_comp}
\end{figure}

\begin{figure}[t]
 \centering
  \begin{subfloat}{
  \includegraphics[width=\columnwidth]{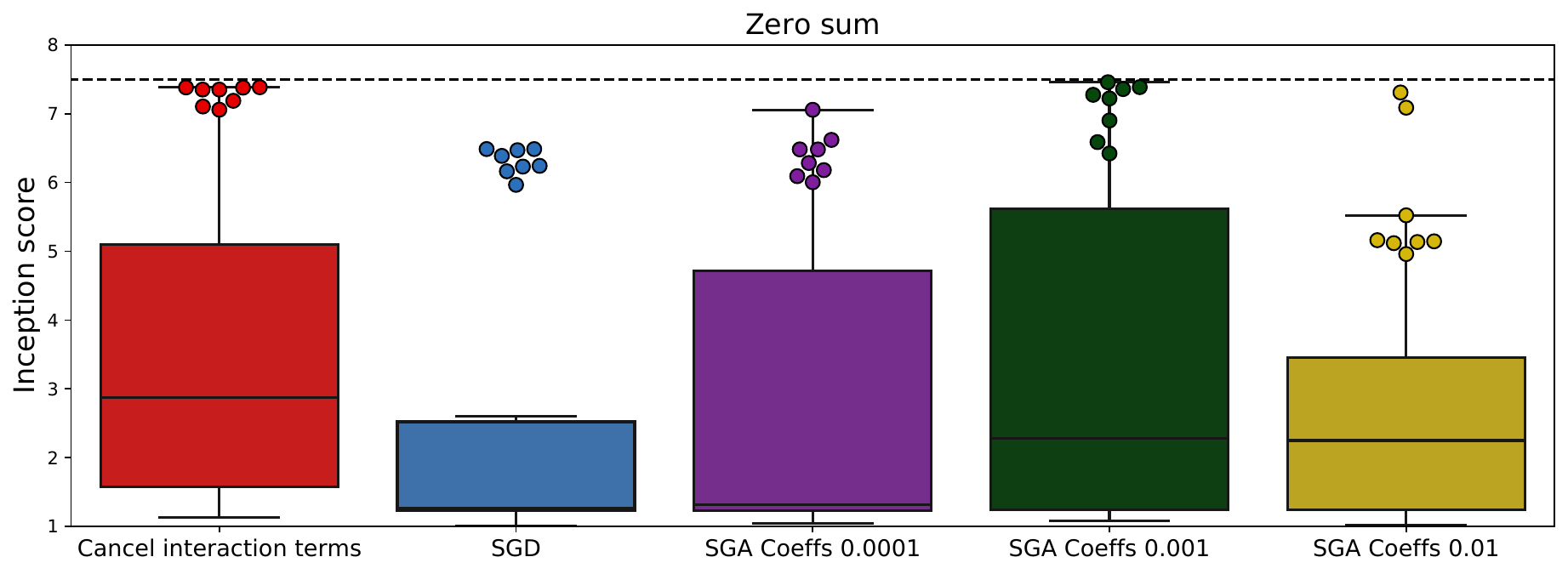}%
} \end{subfloat}%
  \caption[Comparing hyperparameter sensitivity of cancelling interactions terms and Symplectic Gradient Adjustment.]{Comparison with Symplectic Gradient Adjustment (SGA),  obtained from \textit{all models in the sweep}. Without requiring an additional sweep over the regularisation coefficient, cancelling the interaction terms results in better performance across the learning rate sweep and less sensitivity to hyperparameters.}
  \label{fig:sga_box_plots}
\end{figure}

\textbf{Comparison with Consensus Optimisation}.
We show results comparing with Consensus Optimisation (CO)~\citep{mescheder2017numerics} in Figure~\ref{fig:consensus_opt_com} (best performing models) and Figure~\ref{fig:consensus_opt_box_plots} (quantiles showing performance across all hyperparameters and seeds). We observe that cancelling the interaction terms performs best, and that additionally strengthening the self terms does not provide a performance benefit.

\begin{figure}[t]
 \centering
  \begin{subfloat}[Inception Score ($\uparrow$).]{
  \includegraphics[width=0.49\columnwidth]{consensus_opt_comp_is}%
} \end{subfloat}%
 \begin{subfloat}[Fréchet Inception Distance ($\downarrow$).]{
  \includegraphics[width=0.49\columnwidth]{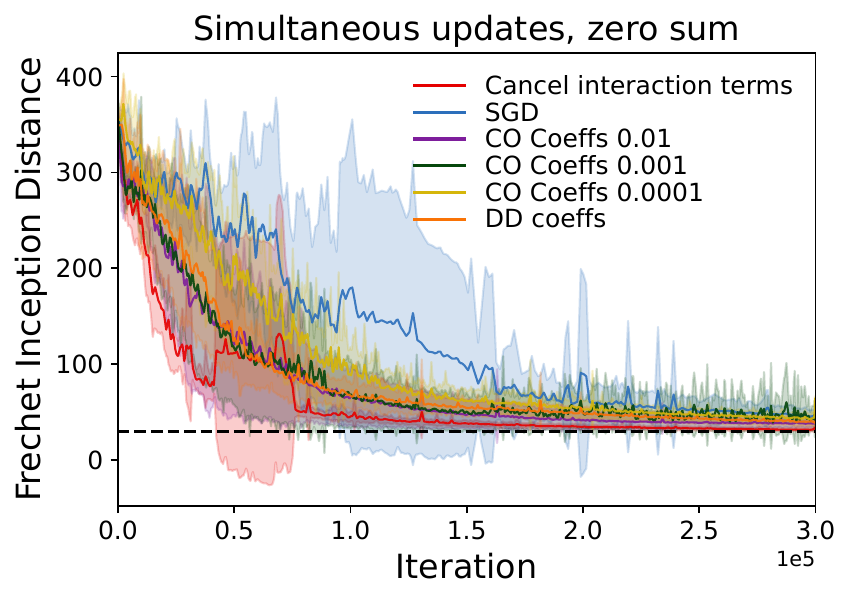}%
} \end{subfloat}%
  \caption[Comparing cancelling interactions terms with Consensus Optimisation.]{Comparison with Consensus Optimisation (CO). Despite not requiring additional hyperparameters compared to the standard SGD learning rate sweep, cancelling the interaction terms of the drift performs better than consensus optimisation. Using consensus optimisation with a fixed coefficient can perform better than using the drift coefficients when we use them to the strengthen the norms -- this is somewhat expected since while the modified flows give us the exact coefficients required to \textit{cancel} the drift, they do not tell us how to strengthen it, and our choice of exact coefficients from the drift might not be optimal.}
  \label{fig:consensus_opt_com}
\end{figure}

\begin{figure}[t]
 \centering
  \begin{subfloat}{
  \includegraphics[width=\columnwidth]{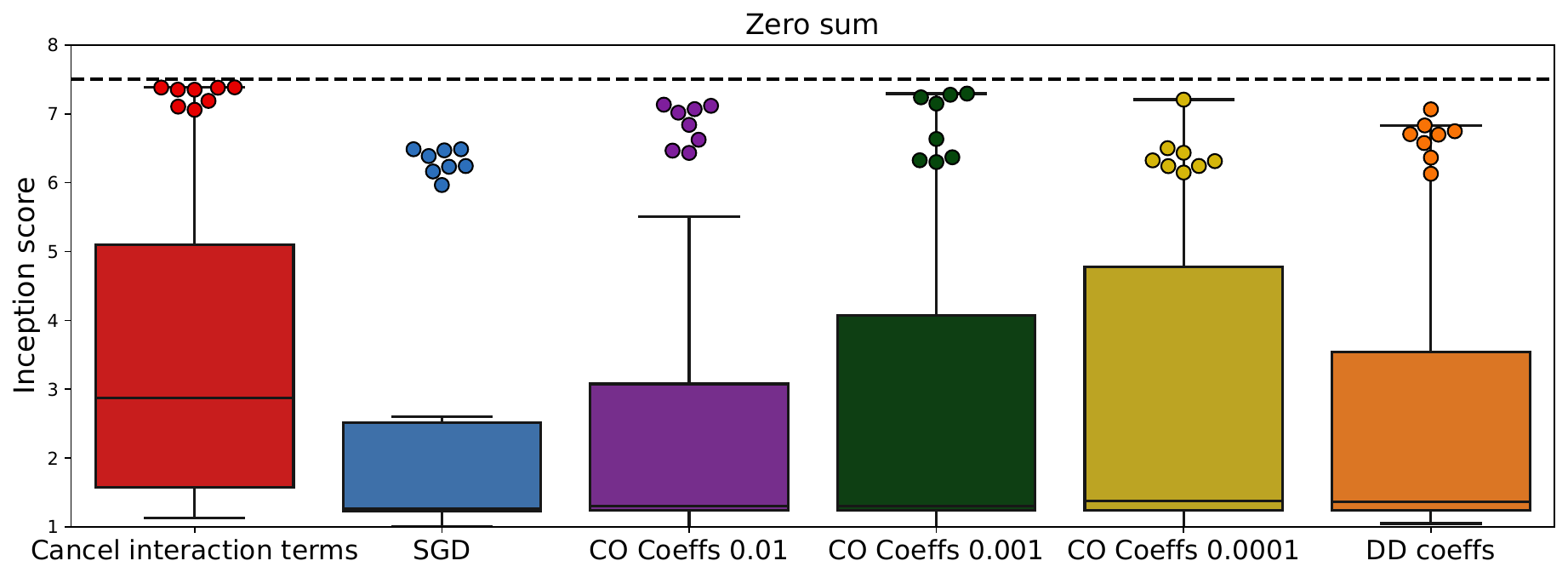}%
} \end{subfloat}%
  \caption[Comparing hyperparameter sensitivity of cancelling interactions terms and Consensus Optimisation.]{Comparison with Consensus Optimisation (CO), obtained from \textit{all models in the sweep}. Without requiring an additional sweep over the regularisation coefficient, cancelling the interaction terms results in better performance across the learning rate sweep and less sensitivity to hyperparameters.}
  \label{fig:consensus_opt_box_plots}
\end{figure}

\textbf{Variance across seeds}.
We have mentioned in the main text the challenge with variance across seeds observed when training GANs with SGD, especially in the case of simultaneous updates in zero sum games. We first notice that performance of simultaneous updates depends strongly on learning rates, with most models not learning. We also notice variance across seeds, both in vanilla SGD and when using explicit regularisation to cancel interaction terms. In order to investigate this effect, we ran a sweep of 50 seeds for the best learning rates we obtain when cancelling interaction terms in simultaneous updates, namely a discriminator learning rate of $0.01$ and a generator learning rate of $0.005$, we obtain Inception Score results with mean 5.61, but a very large standard deviation of 2.34. Indeed, as shown in Figure~\ref{fig:seed_performance_cancel_interaction_terms}, more than 50\% of the seeds converge to an IS grater than 7.
To investigate the reason for the variability, we repeat the same experiment, but clip the gradient value for each parameter to be in $[-0.1, 0.1]$, and show results in Figure~\ref{fig:seed_performance_cancel_interaction_terms_clip}. We notice that another $10\%$ of jobs converge to an IS score grater than 7, and $10\%$ drop in the number of jobs that do not manage to learn. This makes us postulate that the reason for the variability is due to large gradients, perhaps early in training. We contrast this variability across seeds with the consistent performance we obtain by looking at the best performing \textit{models} across learning rates, where as we have shown in the main thesis and throughout the Appendix, we obtain consistent performance which consistently leads to a substantial improvement compared to SGD without explicit regularisation, and obtains performance comparable with Adam.

\begin{figure}[t]
 \centering
  \begin{subfloat}[Learning curves.]{
  \includegraphics[width=0.49\columnwidth]{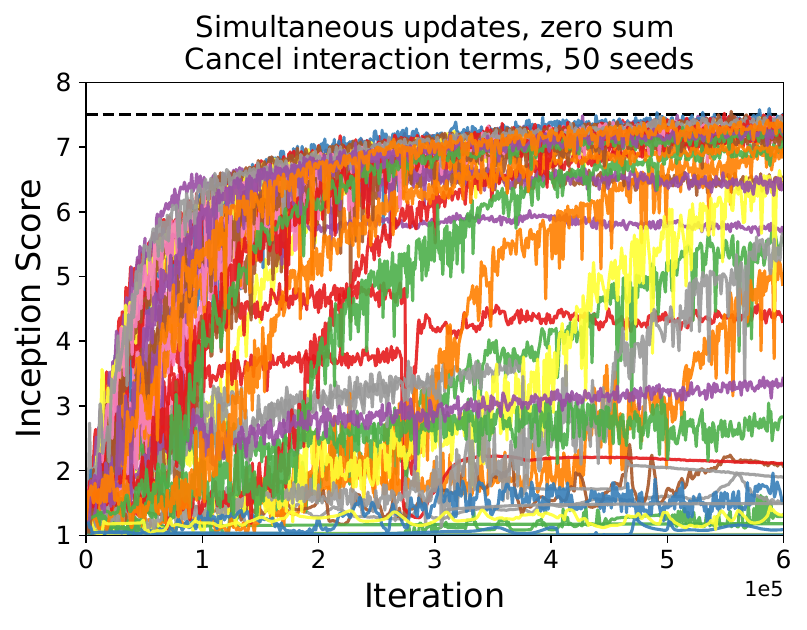}%
} \end{subfloat}%
 \begin{subfloat}[Performance Histogram.]{
  \includegraphics[width=0.49\columnwidth]{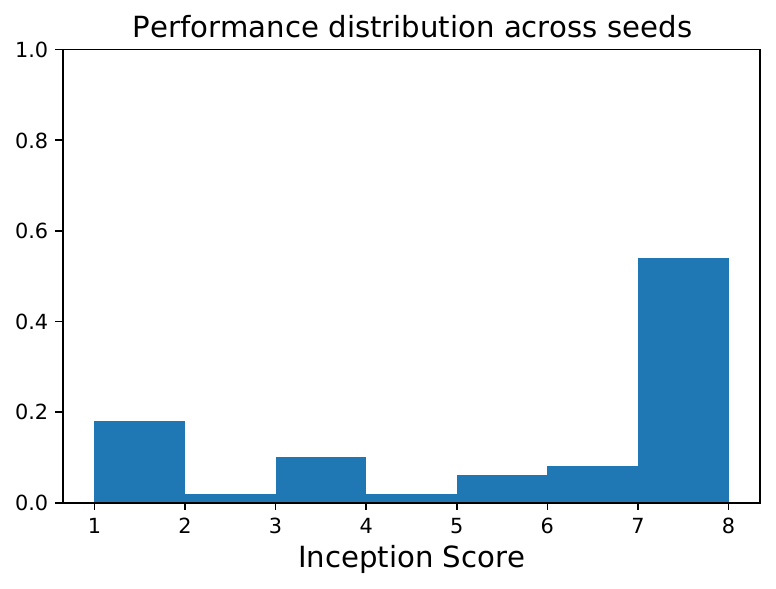}%
} \end{subfloat}%
  \caption[Variability across seeds when cancelling interaction terms.]{Variability across seeds for the best performing hyperparameters, when cancelling interaction terms in simultaneous updates for the original GAN, with a zero sum loss.}
  \label{fig:seed_performance_cancel_interaction_terms}
\end{figure}

\begin{figure}[t]
 \centering
  \begin{subfloat}[Learning curves.]{
  \includegraphics[width=0.49\columnwidth]{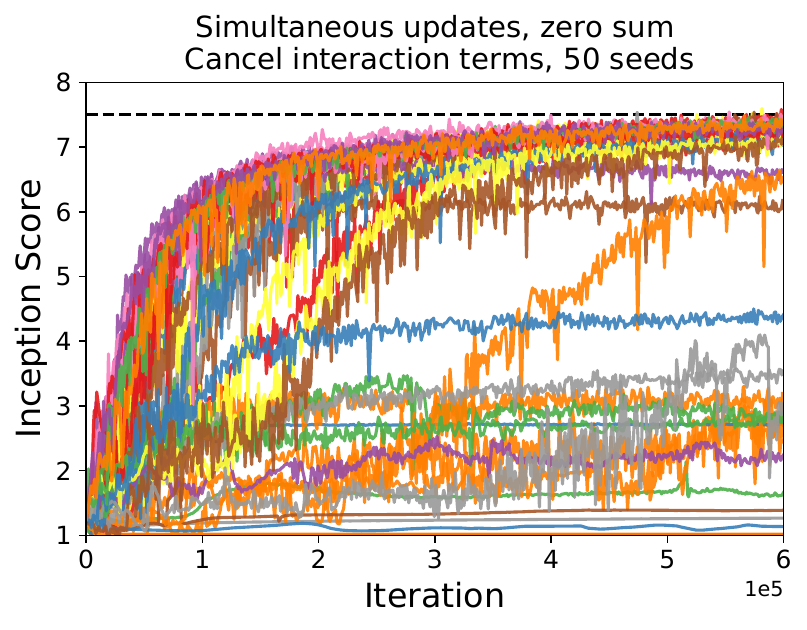}%
} \end{subfloat}%
 \begin{subfloat}[Performance Histogram.]{
  \includegraphics[width=0.49\columnwidth]{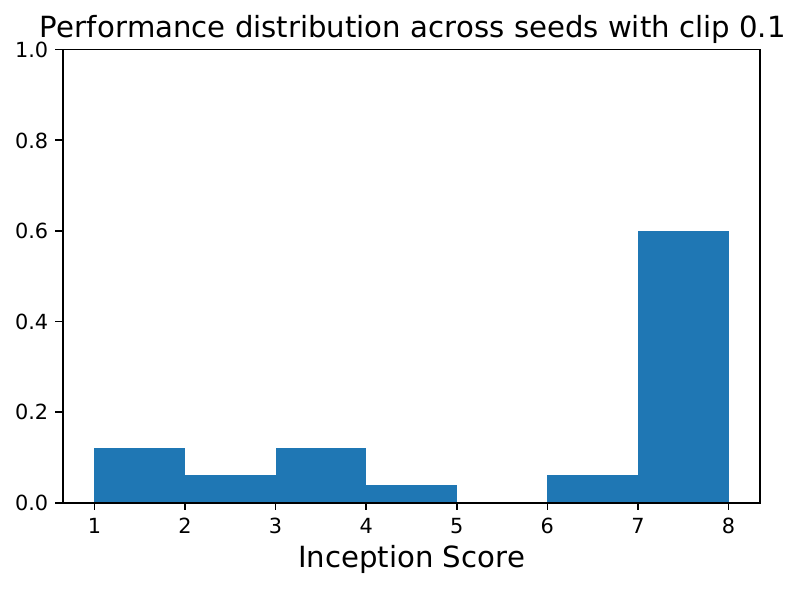}%
} \end{subfloat}%
  \caption[Gradient clipping can reduce variability across seeds when cancelling interaction terms.]{\textit{With gradient clipping}. Gradient clipping can reduce variability. This suggests that the instabilities observed in gradient descent are caused by large gradient updates.}
  \label{fig:seed_performance_cancel_interaction_terms_clip}
\end{figure}

\subsection{Explicit regularisation in zero-sum games trained using alternating gradient descent}

We perform the same experiments as done for simultaneous updates also with alternating updates. To do so, we cancel the effect of DD using the same explicit regularisation functional form, but updating the coefficients to be those of alternating updates. We show results in Figure~\ref{fig:sgd_vs_cancel_drift_interaction_alternating}, where we see very little difference in the results compared to vanilla SGD, perhaps apart from less instability early on in training. We postulate that this could be due the effect of DD in alternating updates can can be beneficial, especially for learning rate ratios for which the second player also minimises the gradient norm of the first player. We additionally show results obtained from strengthening the self terms in Figure~\ref{fig:sgd_vs_cancel_drift_interaction_all_types_reg_alt}.

\textbf{More percentiles}.
Throughout the main thesis, we displayed the best $10\%$ performing models for each optimisation algorithm used. We now expand that to show performance results across the $20\%$ and $30\%$ jobs in Figure~\ref{fig:multiple_percentages_sgd_cancel_interaction_terms_alternating}. We observe a consistent increase in performance obtained by cancelling the interaction terms.

\begin{figure}[t]
 \centering
  \begin{subfloat}[Inception Score ($\uparrow$).]{
  \includegraphics[width=0.49\columnwidth]{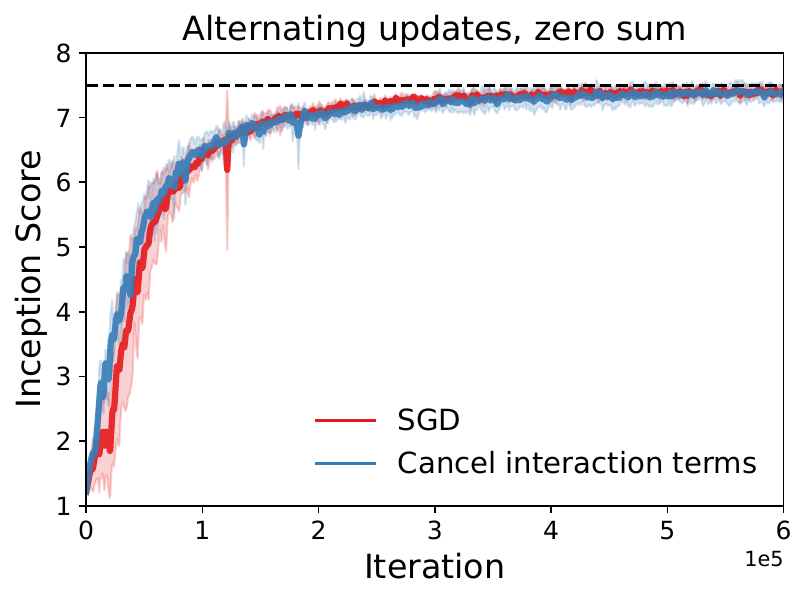}%
} \end{subfloat}%
 \begin{subfloat}[Fréchet Inception Distance ($\downarrow$).]{
  \includegraphics[width=0.49\columnwidth]{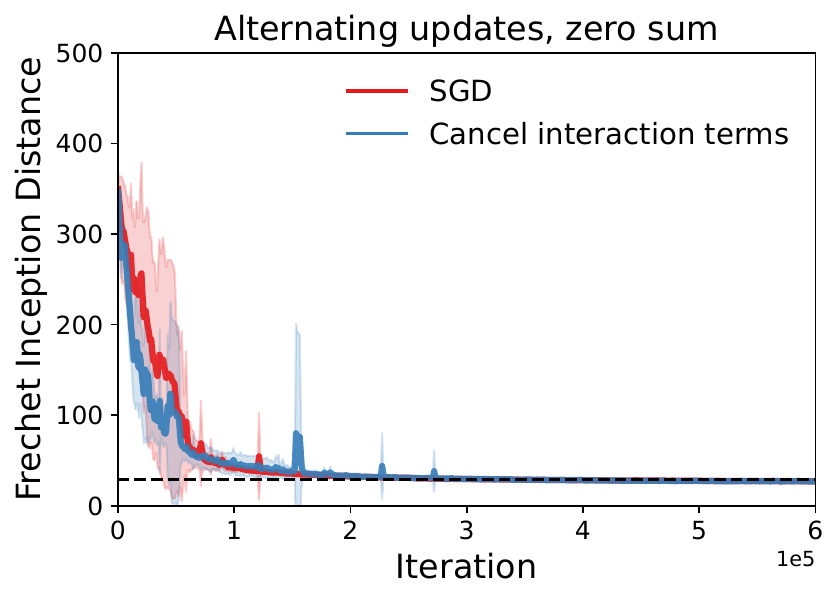}%
} \end{subfloat}%
  \caption[Cancelling interaction terms of discretisation drift in alternating gradient descent updates.]{In alternating updates, using explicit regularisation to cancel the effect of the interaction components of drift does not substantially improve performance compared to SGD, but can reduce variance. This is expected, given that the interaction terms for the second player in the case of alternating updates can have a beneficial regularisation effect.}
  \label{fig:sgd_vs_cancel_drift_interaction_alternating}
\end{figure}

\begin{figure}[t]
 \centering
  \begin{subfloat}[Inception Score ($\uparrow$).]{
  \includegraphics[width=0.49\columnwidth]{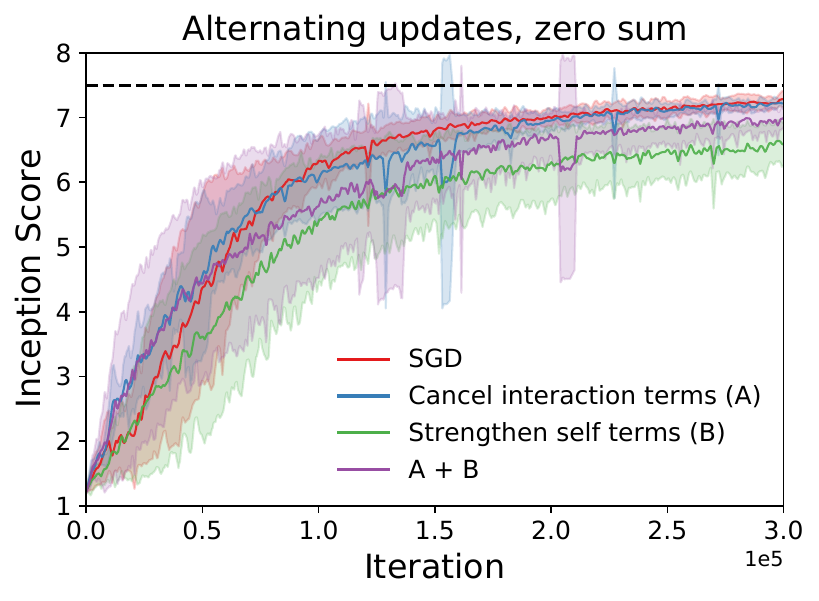}%
} \end{subfloat}%
 \begin{subfloat}[Fréchet Inception Distance ($\downarrow$).]{
  \includegraphics[width=0.49\columnwidth]{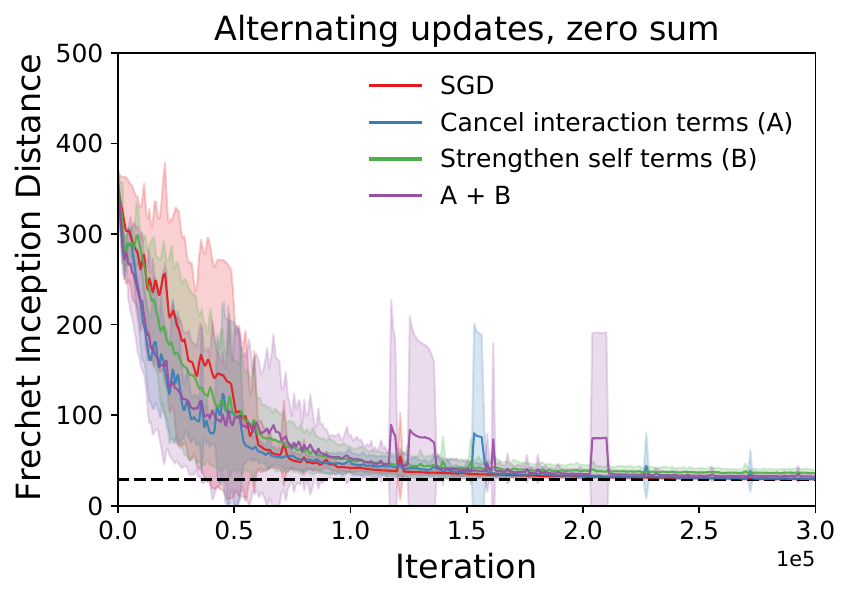}%
} \end{subfloat}%
  \caption[Variability when cancelling interaction terms of discretisation drift in alternating gradient descent updates.]{In alternating updates, using explicit regularisation to cancel the effect of the interaction components of drift does not substantially improve performance compared to SGD, but can reduce variance. This is expected, given that the interaction terms for the second player in the case of alternating updates can have a beneficial regularisation effect. Strengthening the self terms---the terms which minimise the player's own norm---does not lead to a substantial improvement; this is somewhat expected since while the modified flows give us the exact coefficients required to \textit{cancel} the drift, they do not tell us how to strengthen it, and our choice of exact coefficients from the drift might not be optimal.}
  \label{fig:sgd_vs_cancel_drift_interaction_all_types_reg_alt}
\end{figure}

\begin{figure}[t]
 \centering
  \begin{subfloat}[Top 10\% models.]{
  \includegraphics[width=0.33\columnwidth]{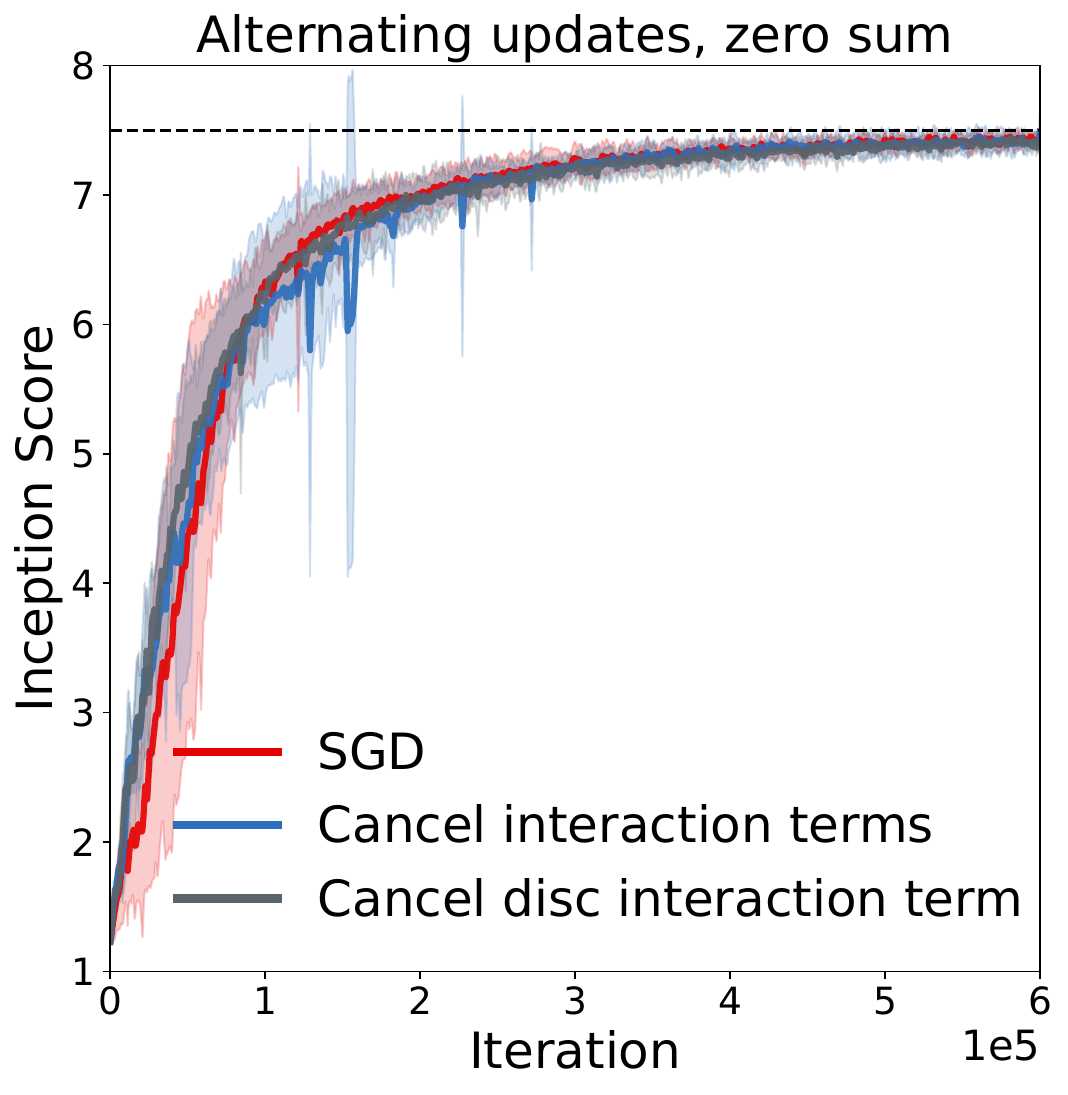}%
} \end{subfloat}%
 \begin{subfloat}[Top 20\% models.]{
  \includegraphics[width=0.33\columnwidth]{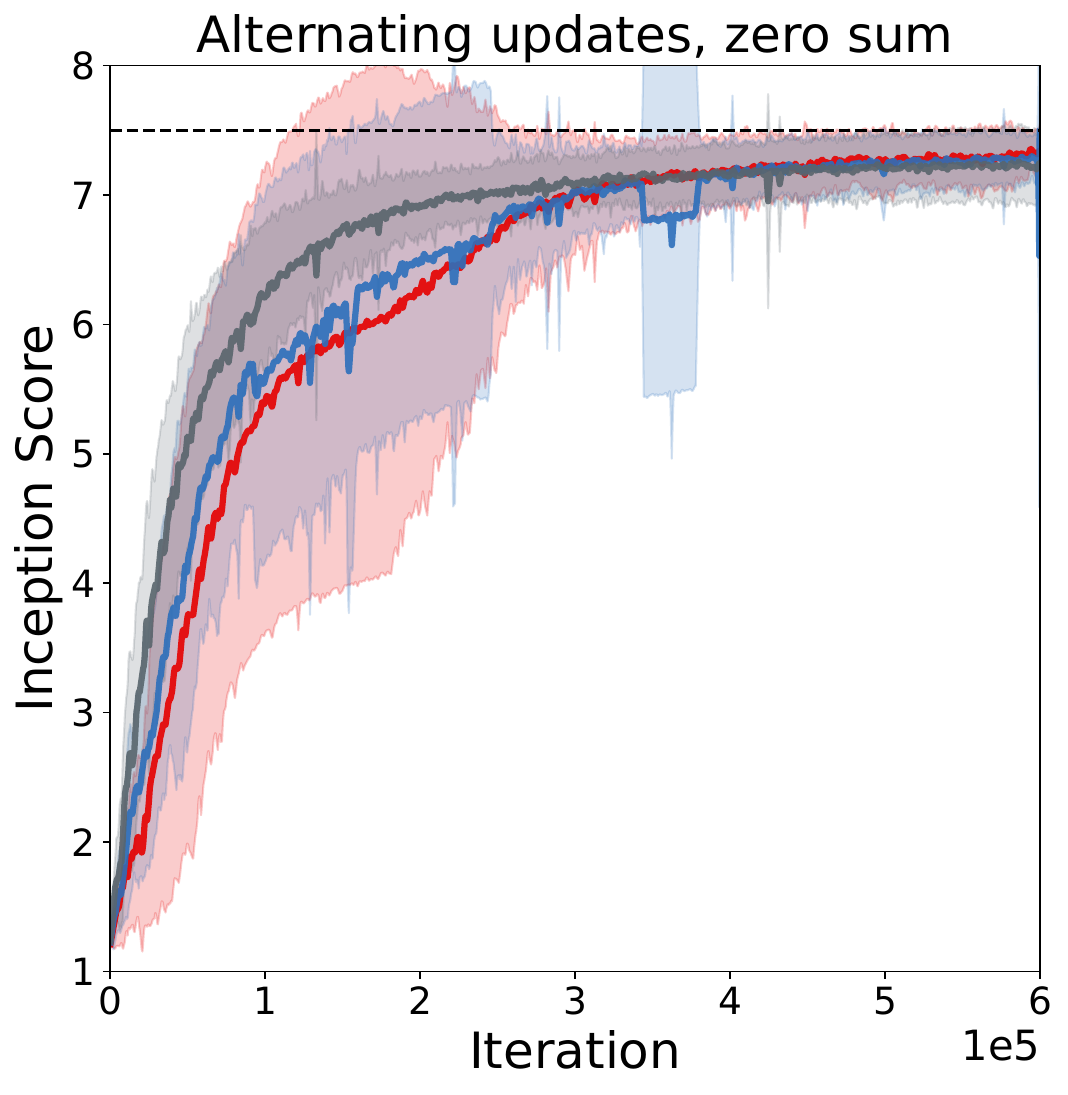}%
} \end{subfloat}%
\begin{subfloat}[Top 30\% models.]{
  \includegraphics[width=0.33\columnwidth]{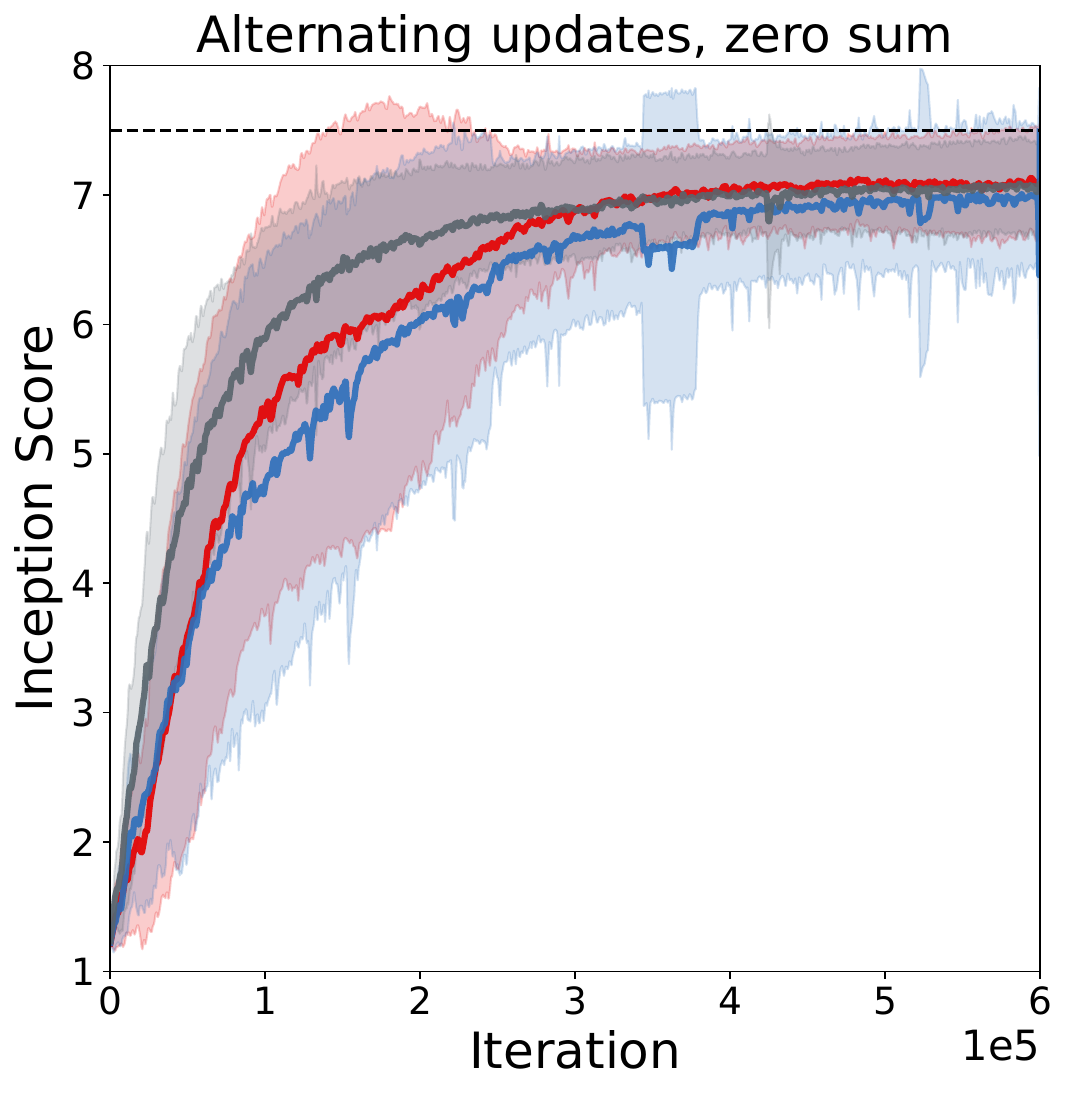}%
} \end{subfloat}%
  \caption[Cancelling interaction terms in alternating updates can increase performance across multiple percentages used to select top models.]{Performance across the top performing models for vanilla SGD and with cancelling the interaction terms and only cancelling the discriminator interaction terms. We notice that cancelling only the discriminator interaction terms can result in higher performance across more models, and that the interaction term of the generator can play a positive role, likely due to the smaller strength of the generator interaction term compared to simultaneous updates.}
  \label{fig:multiple_percentages_sgd_cancel_interaction_terms_alternating}
\end{figure}

\clearpage

\section{Individual figure reproduction details}

\begin{itemize}
\item Figure~\ref{fig:alternating_updates_supervised_learning} performs MNIST classification results with alternating updates. We use an MLP with layers of size $[100, 100, 100, 10]$ and a learning rate of $0.08$.
The batch size used is $50$ and models are trained for 1000 iterations.
 Error bars are obtained from 5 different seeds.
 \item Figure~\ref{fig:dirac_gan} uses a learning rate of $0.01$ for both the discriminator and the generator. The continuous-time simulation is obtained using Runge--Kutta4.
  \item Figure~\ref{fig:idd_zero_sum_games} is obtained using sweep over $\{0.01, 0.005, 0.001, 0.0005\}$ for the discriminator and for the generator learning rates. We restrict the ratio between the two learning rates to be in the interval $[0.1, 10]$ to ensure the validity of our approximations. The architecture is Spectral Normalised GAN. Batch size of 128. When comparing with Runge--Kutta 4, we no longer perform a cross product over discriminator and generator learning rates, but instead restrict ourselves to equal learning rates.
 \item Figure~\ref{fig:sim_vs_alt_zero_sum_learning_rates} controls for the number of experiments which have the same learning rate ratio. To do so, we obtain 5 learning rates uniformly sampled from the interval $[0.001, 0.01]$ which we use for the discriminator, we fix the learning rate ratios to be in $\{0.1, 0.2, 0.5, 1., 2., 5.\}$ and we obtain the generator learning rate from the discriminator learning rate and the learning rate ratio. Batch size of 128.
 \item Figure~\ref{fig:sgd_adam_sim_comparison},~\ref{fig:sgd_vs_cancel_drift_interaction},~\ref{fig:adam_comparison},~\ref{fig:sgd_vs_cancel_drift_interaction_all_types_reg},~\ref{fig:multiple_percentages_sgd_cancel_interaction_terms},
 ~\ref{fig:seed_performance_cancel_interaction_terms},
  and~\ref{fig:seed_performance_cancel_interaction_terms_clip} use the unbiased estimator described in Section~\ref{app:unbiased_est} for explicit regularisation. SGD results are obtained from a sweep of $\{0.01, 0.005, 0.001, 0.0005\}$ for the discriminator and for the generator learning rates. We restrict the ratio between the two learning rates to be in the interval $[0.1, 10]$ to ensure the validity of our approximations. For Adam, we use the learning rate sweep $\{10^{-4}, 2 \times 10^{-4}, 3 \times 10^{-4}, 4 \times 10^{-4}\}$ for the discriminator and the same for the generator, with $\beta_1 = 0.5$ and $\beta_2 = 0.99$. All models use simultaneous updates and batch size of 128.
 \item Figure~\ref{fig:sga_consensus_opt_main_paper},~\ref{fig:consensus_opt_com},~\ref{fig:consensus_opt_box_plots},~\ref{fig:supp_sga_comp},~\ref{fig:sga_box_plots} use the unbiased estimator described in Section~\ref{app:unbiased_est} for explicit regularisation. Results are obtained from a sweep of $\{0.01, 0.005, 0.001, 0.0005\}$ for the discriminator and for the generator learning rates. We restrict the ratio between the two learning rates to be in the interval $[0.1, 10]$ to ensure the validity of our approximations. The SGA and CO explicit regularisation coefficients are taken from a sweep over $\{0.01, 0.001, 0.0001\}$. All models use simultaneous updates and batch size of 128.
 \item Figures~\ref{fig:sgd_adam_sim_alt_updates},~\ref{fig:sgd_vs_cancel_drift_interaction_alternating}~\ref{fig:sgd_vs_cancel_drift_interaction_all_types_reg_alt},~\ref{fig:multiple_percentages_sgd_cancel_interaction_terms_alternating} use the same experimental setup as Figure~\ref{fig:sgd_adam_sim_comparison}, with the exception of the update type: here we use alternating updates.
 \item Figure~\ref{fig:non_saturating} uses the same experimental setting as Figure~\ref{fig:idd_zero_sum_games}, but the non-saturating GAN loss is used.
 \item Figure~\ref{fig:supp_idd_zero_sum_games} uses the same experimental setting as Figure~\ref{fig:idd_zero_sum_games}.
 \item Figure~\ref{fig:idd_zero_sum_games_ls_gan} uses a least square loss. Figure~\ref{fig:square_loss_gan_sensitivity} uses the same experimental setting as Figure~\ref{fig:idd_zero_sum_games}, while Figure~\ref{fig:square_loss_gan_ratios} uses the same experimental setting as Figure~\ref{fig:sim_vs_alt_zero_sum_learning_rates}.
 \item Figure~\ref{fig:supp_non_saturating} uses the same experimental setting as Figure~\ref{fig:supp_idd_zero_sum_games}, but uses the non-saturating generator loss.
 \item Figure~\ref{fig:multiple_percentages_sgd_cancel_interaction_terms} shows results from the same experiment Figure~\ref{fig:sgd_adam_sim_comparison}.
 \item Figure~\ref{fig:batch_size_comparison} uses the same experimental setting as Figure~\ref{fig:sgd_adam_sim_comparison}, but instead of using a fixed batch size of 128, performs a sweep over batch sizes. 
\end{itemize}

\chapter{Finding new implicit regularisers by revisiting backward error analysis}
\label{ch:bea_stochasticity_supp}

We now provide the proofs required to write modified losses for stochastic gradient descent in supervised learning discussed in Chapter~\ref{ch:bea_stochasticity}. We denote by $\vtheta$ the parameters of the model and by $f$ the update function. Since we are now interested in the stochastic setting, we write $f$ as an argument both of data and parameters: $f(\vtheta;\vx)$ denotes the application of $f$ at input $\vx$ using parameters $\vtheta$. We often consider $f = - \nabla_{\vtheta} E(\vtheta; \vx)$, where $E$ is the loss function. We will denote as
\begin{align}
 E(\vtheta; \vX^{t}) = \frac 1 B \sum_{i=1}^{B}E(\vtheta; \vx_i^t) 
\end{align}
and 
\begin{align}
 f(\vtheta; \vX^{t}) &= - \frac 1 B \sum_{i=1}^{B} \nabla_{\vtheta} E(\vtheta; \vx_i^t) 
\label{eq:f_hat_def} \\
 f(\vtheta; \{\vX^{t}, ...\vX^{t+n-1}\}) &= \frac 1 n \sum_{i=0}^{n-1} f(\vtheta; \{\vX^{t+i}\}) \label{eq:multiple_batches_def}
\end{align}
as averages over batches for convenience and clarity.

Our goal is to find $\dot{\vtheta}$ such that the distance between $n$ steps of stochastic gradient descent (SGD) with learning rate $h$ and $\dot{\vtheta}$ is of order $\mathcal{O}(h^3)$. 
We take the same approach as in the other BEA proofs:
\begin{itemize}
  \item We expand the $n$ discrete SGD updates up to $\mathcal{O}(h^3)$.
  \item We expand change of the modified continuous updates of the form $\dot{\vtheta} = f + h f_1$ in time $nh$, where $f$ is the gradient function obtained using the concatenation of all batches used in the first step.
  \item We match the terms between the discrete and continuous updates of $\mathcal{O}(h^2)$  to find $f_1$.
\end{itemize}

\section{Two consecutive steps of SGD}

\textbf{Step 1: Expand the discrete updates.} \\
From the definition of SGD:
\begin{align}
 \vtheta_{t} &= \vtheta_{t-1} + h  f(\vtheta_{t-1}; \vX^{t}) \label{eq:sgd1_app} \\
 \vtheta_{t +1 } &= \vtheta_{t} + h   f(\vtheta_{t}; \vX^{t+1}) \label{eq:sgd2_app}
\end{align}
We expand the gradient descent steps and obtain:
\begingroup
\allowdisplaybreaks
\begin{align}
 \vtheta_{t +1 } &= \vtheta_{t} + h   f(\vtheta_{t}; \vX^{t+1}) \\
                &= \vtheta_{t-1} + h  f(\vtheta_{t-1}; \vX^{t}) + h   f(\vtheta_{t}; \vX^{t+1}) \owntag{From \eqref{eq:sgd1_app}} \\
                 &= \vtheta_{t-1} + h  f(\vtheta_{t-1}; \vX^{t}) + h   f(\vtheta_{t-1} + h  f(\vtheta_{t-1}; \vX^{t}); \vX^{t+1}) \owntag{From \eqref{eq:sgd2_app}} \\
                  &= \vtheta_{t-1} + h  f(\vtheta_{t-1}; \vX^{t}) + h   f(\vtheta_{t-1}; \vX^{t+1}) \\ &\hspace{2em}+  h^2 \evaljacdata{\vtheta}{f}{\vtheta_{t-1}}{\vX^{t+1}}  f(\vtheta_{t-1}; \vX^{t}) +  \mathcal{O}(h^3) \owntag{Taylor expansion}\\
                  &= \vtheta_{t-1} + 2h \left(\frac{1}{2} f(\vtheta_{t-1}; \vX^{t}) +  \frac{1}{2} f(\vtheta_{t-1}; \vX^{t+1})\right) \owntag{Grouping $f$ terms} \\ &\hspace{2em}+  h^2 \evaljacdata{\vtheta}{f}{\vtheta_{t-1}}{\vX^{t+1}}  f(\vtheta_{t-1}; \vX^{t}) +  \mathcal{O}(h^3)\\
                    &= \vtheta_{t-1} + 2h \!\!\!\underbrace{ f(\vtheta_{t-1}; \{\vX^{t}, \vX^{t+1} \})}_{\substack{\text{update with a batch obtained by}\\{\text{concatenating the two batches}}}} \owntag{Eq~\eqref{eq:multiple_batches_def}}
                    \\ &\hspace{2em}+ \, h^2 \evaljacdata{\vtheta}{f}{\vtheta_{t-1}}{\vX^{t+1}}  f(\vtheta_{t-1}; \vX^{t}) +  \mathcal{O}(h^3)
\label{eq:two_steps_sgd}
\end{align}
\endgroup

\textbf{Step 2: Taylor expansion of modified flow} \\
We now expand what happens in a time of $2h$ in continuous-time, by using the form of $\dot{\vtheta}$ given by BEA
\begin{align}
\dot{\vtheta} =  f(\vtheta; \{\vX^{t}, \vX^{t+1} \}) + 2 h f_1(\vtheta; \{\vX^{t},\vX^{t+1}\} )
\end{align}
thus for any $\tau$
\begin{align}
{\vtheta}(\tau + 2h) &= {\vtheta}(\tau) + 2h \dot{\vtheta}(\tau) + 4h^2 \ddot{{\vtheta}}(\tau) + \mathcal{O}(h^3) \\
&= {\vtheta} + 2 h  f(\vtheta; \{\vX^{t}, \vX^{t+1} \}) \\ &\hspace{2em}+ 4 h^2 \left[f_1 + \frac{1}{2}  \jacparam{\vtheta}{f(\vtheta; \{\vX^{t}, \vX^{t+1} \})}  f(\vtheta; \{\vX^{t}, \vX^{t+1} \})\right] + \mathcal{O}(h^3).
\label{eq:disp_sgd_flow}
\end{align}

\textbf{Step 3: Matching terms of the second order} \\
From the above expansion of the discrete updates (Step 1) and of the continuous updates (Step 2) we can find the value of $f_1$ at the last initial parameters, $\vtheta_{t-1}$:
\begin{align}
4 h^2 \left[f_1 + \frac{1}{2}  \jacparam{\vtheta}{f(\vtheta; \{\vX^{t}, \vX^{t+1} \})}  f(\vtheta; \{\vX^{t}, \vX^{t+1} \})\right] = h^2 \evaljacdata{\vtheta}{f}{\vtheta_{t-1}}{\vX^{t+1}}  f(\vtheta_{t-1}; \vX^{t})
\end{align}
This leads to:
\begin{align}
f_1(\vtheta_{t-1}) &= \underbrace{- \frac{1}{2}  \evaljacdata{\vtheta}{f}{\vtheta_{t-1}}{\{\vX^{t}, \vX^{t+1} \}}  f(\vtheta_{t-1}; \{\vX^{t}, \vX^{t+1} \})}_{\text{the IGR full-batch drift term}} \owntag{From flow: \eqref{eq:disp_sgd_flow}} \\ &+ \frac 1 4 \evaljacdata{\vtheta}{f}{\vtheta_{t-1}}{\vX^{t+1}}  f(\vtheta_{t-1}; \vX^{t}) \owntag{From SGD: \eqref{eq:two_steps_sgd}}
\label{eq:sgd_f1_vheta_t}
\end{align}
From here we choose a function $f_1$ which satisfies the above constraint
\begin{align}
f_1(\vtheta) &= \underbrace{- \frac{1}{2}  \jacparam{\vtheta}{f(\vtheta; \{\vX^{t}, \vX^{t+1} \})}  f(\vtheta; \{\vX^{t}, \vX^{t+1} \})}_{\text{the IGR full-batch drift term}} \\ &\quad + \frac 1 4 \jacparam{\vtheta}{f(\vtheta; \vX^{t+1})}  f({\color{red}\vtheta_{t-1}}; \vX^{t}).
\end{align}
We have now found the modified flow $\dot{\vtheta}$, which by construction follows two Euler updates with an error of $\mathcal{O}(h^3)$. Note the use of the initial parameters (highlighted in red) in the flow's vector field; this choice is required to ensure we can write a modified loss function in the single objective optimisation setting by using $f = - \nabla_{\vtheta} E$:
\begin{align}
f_1(\vtheta) &=  - \frac{1}{2}  \jacparam{\vtheta}{\nabla_{\vtheta}E(\vtheta; \{\vX^{t}, \vX^{t+1} \})}  \nabla_{\vtheta}E(\vtheta; \{\vX^{t}, \vX^{t+1} \}) \\ &\quad+ \frac 1 4 \jacparam{\vtheta}{\nabla_{\vtheta}E(\vtheta; \vX^{t+1})}  \nabla_{\vtheta}E({\color{red}\vtheta_{t-1}}; \vX^{t}) \\
&= - \nabla_{\vtheta} \left(\frac{1}{4} \|\nabla_{\vtheta} E(\vtheta;\{\vX^{t}, \vX^{t+1} \}) \|^2  - \frac 1 4 \nabla_{\vtheta} E(\vtheta; \vX^{t+1})^T \nabla_{\vtheta} E({\color{red}{\vtheta_{t  -1}}}; \vX^t) \right) \owntag{using \eqref{eq:normssametheta}},
\end{align}
where $ E(\vtheta;\{\vX^{t}, \vX^{t+1} \}) = \frac {1} {2} \left( E(\vtheta;\vX^{t}) + E(\vtheta;\vX^{t+1})\right)$.
We can now write
\begin{align}
\dot{\vtheta} = - \nabla_{\vtheta} \Bigg(&E(\vtheta;\{\vX^{t}, \vX^{t+1} \}) \\&+ 2h \left(\frac{1}{4} \|\nabla_{\vtheta} E(\vtheta;\{\vX^{t}, \vX^{t+1} \}) \|^2  - \frac 1 4 \nabla_{\vtheta} E(\vtheta; \vX^{t+1})^T \nabla_{\vtheta} E({\color{red}{\vtheta_{t  -1}}}; \vX^t) \right) \Bigg),
\end{align}
and since the vector field is a negative gradient, this leads to the modified loss
\begin{align}
\tilde{E} =& E(\vtheta;\{\vX^{t}, \vX^{t+1} \}) \\&+ 2h \left(\frac{1}{4} \|\nabla_{\vtheta} E(\vtheta;\{\vX^{t}, \vX^{t+1} \}) \|^2  - \frac 1 4 \nabla_{\vtheta} E(\vtheta; \vX^{t+1})^T \nabla_{\vtheta} E({\color{red}{\vtheta_{t  -1}}}; \vX^t) \right).
\end{align}

\section{Multiple steps of SGD}

We will now derive a similar result, for $n$ SGD steps. 

\noindent\textbf{Step 1: Expand discrete updates}
\begin{align}
 \vtheta_{t} &= \vtheta_{t-1} + h  f(\vtheta_{t-1}; \vX^{t}) \\
\dots \nonumber\\
 \vtheta_{t + n -1 } &= \vtheta_{t + n -2} + h   f(\vtheta_{t + n -2}; \vX^{t+n -1})
\end{align}
From Eq~\eqref{eq:two_steps_sgd}, we know that if we expand two steps we obtain
\begin{align}
 \vtheta_{t +1 } &= \vtheta_{t-1} + 2h  f(\vtheta_{t-1}; \{\vX^{t}, \vX^{t+1} \}) + h^2 \evaljacdata{\vtheta}{f}{\vtheta_{t-1}}{\vX^{t+1}}  f(\vtheta_{t-1}; \vX^{t})  +  \mathcal{O}(h^3) \label{eq:ih_multiple_steps}.
\end{align}
Then, by expanding the third step
\begingroup
\allowdisplaybreaks
\begin{align}
 \vtheta_{t +2 } &= \vtheta_{t +1} + h  f(\vtheta_{t+1}; \vX^{t+2}) \\
                 &=  \vtheta_{t-1} + 2h  f(\vtheta_{t-1}; \{\vX^{t}, \vX^{t+1} \}) + h^2 \evaljacdata{\vtheta}{f}{\vtheta_{t-1}}{\vX^{t+1}}  f(\vtheta_{t-1}; \vX^{t}) \owntag{Eq~\eqref{eq:ih_multiple_steps}} \\ &\hspace{1em}+    h  f(\vtheta_{t+1}; \vX^{t+2}) +  \mathcal{O}(h^3) \\
                 &=  \vtheta_{t-1} + 2h  f(\vtheta_{t-1}; \{\vX^{t}, \vX^{t+1} \}) + h^2 \evaljacdata{\vtheta}{f}{\vtheta_{t-1}}{\vX^{t+1}}  f(\vtheta_{t-1}; \vX^{t}) \\ &\hspace{1em}+   h  f\left({\vtheta_{t-1} + 2h  f(\vtheta_{t-1}; \{\vX^{t}, \vX^{t+1} \}) + h^2 \evaljacdata{\vtheta}{f}{\vtheta_{t-1}}{\vX^{t+1}}  f(\vtheta_{t-1}; \vX^{t})}; \vX^{t+2}\right) \owntag{Eq~\eqref{eq:ih_multiple_steps}} \\&\hspace{1em} +  \mathcal{O}(h^3)  \\
                 &=  \vtheta_{t-1} + 2h  f(\vtheta_{t-1}; \{\vX^{t}, \vX^{t+1} \}) + h^2 \evaljacdata{\vtheta}{f}{\vtheta_{t-1}}{\vX^{t+1}}  f(\vtheta_{t-1}; \vX^{t}) \\ 
                 &\hspace{1em}+   h  f(\vtheta_{t-1};\vX^{t+2})  \\
                 &\hspace{1em}+ h \evaljacdata{\vtheta}{f}{\vtheta_{t-1}}{\vX^{t+2}} \left(2h  f(\vtheta_{t-1}; \{\vX^{t}, \vX^{t+1} \}) + h^2 \evaljacdata{\vtheta}{f}{\vtheta_{t-1}}{\vX^{t+1}}  f(\vtheta_{t-1}; \vX^{t})\right) \owntag{Taylor expansion} \\&\hspace{1em} +  \mathcal{O}(h^3) \\
                      &=  \vtheta_{t-1} + 2h  f(\vtheta_{t-1}; \{\vX^{t}, \vX^{t+1} \}) + h^2 \evaljacdata{\vtheta}{f}{\vtheta_{t-1}}{\vX^{t+1}}  f(\vtheta_{t-1}; \vX^{t}) \\ 
                 &\hspace{1em}+   h  f(\vtheta_{t-1};\vX^{t+2})  \\
                 &\hspace{1em}+ 2h^2  \evaljacdata{\vtheta}{f}{\vtheta_{t-1}}{\vX^{t+2}}  f(\vtheta_{t-1}; \{\vX^{t}, \vX^{t+1} \})  \owntag{Simplifying $h^3$ term} \\&\hspace{1em} +  \mathcal{O}(h^3)
\end{align}

\begin{align}
 \vtheta_{t +2 } &=  \vtheta_{t-1} + 3h  f(\vtheta_{t-1}; \{\vX^{t}, \vX^{t+1}, \vX^{t+2} \}) \owntag{Grouping $f$ terms} \\ &\hspace{1em} + h^2 \evaljacdata{\vtheta}{f}{\vtheta_{t-1}}{\vX^{t+1}}  f(\vtheta_{t-1}; \vX^{t}) \\ 
                 &\hspace{1em}+ 2h^2  \evaljacdata{\vtheta}{f}{\vtheta_{t-1}}{\vX^{t+2}}  f(\vtheta_{t-1}; \{\vX^{t}, \vX^{t+1} \})  +  \mathcal{O}(h^3) \\
                  &=  \vtheta_{t-1} + 3h  f(\vtheta_{t-1}; \{\vX^{t}, \vX^{t+1}, \vX^{t+2} \}) + h^2 \evaljacdata{\vtheta}{f}{\vtheta_{t-1}}{\vX^{t+1}}  f(\vtheta_{t-1}; \vX^{t}) \\ 
                 &\hspace{1em}+ h^2  \evaljacdata{\vtheta}{f}{\vtheta_{t-1}}{\vX^{t+2}}  f(\vtheta_{t-1}; \vX^{t}) + h^2 \evaljacdata{\vtheta}{f}{\vtheta_{t-1}}{\vX^{t+2}}  f(\vtheta_{t-1}; \vX^{t+1}) \owntag{Eq~\eqref{eq:multiple_batches_def}} \\&\hspace{1em} +  \mathcal{O}(h^3) .
\end{align}
\endgroup
Thus by induction we get
\begin{align}
 \vtheta_{t +n-1 } &= \vtheta_{t-1} +  n h  f(\vtheta_{t-1}; \{\vX^{t}, \dots, \vX^{t+n-1} \})
                  \\ &+  h^2 \sum_{\tau = 0}^{n-1} \sum_{\mu =\tau+1 }^{n-1}  \evaljacdata{\vtheta}{f}{\vtheta_{t-1}}{\vX^{t+\mu}}  f(\vtheta_{t  -1}; \vX^{t+\tau}) + \mathcal{O}(h^3). \label{eq:sgd_n_exp}
\end{align}

\noindent\textbf{Step 2: Taylor expansion of modified flow}
\begin{align}
{\vtheta}(\tau + nh) &= {\vtheta} + n h  f(\vtheta; \{\vX^{t}, \dots, \vX^{t+n-1} \}) \\ &\quad+ n^2 h^2 \left[f_1 + \frac{1}{2} \jacparam{\vtheta}{f(\vtheta; \{\vX^{t}, \dots, \vX^{t+n-1} \})}  f(\vtheta; \{\vX^{t}, \dots, \vX^{t+n-1} \}) \right] \\& \quad+ \mathcal{O}(h^3)
\label{eq:taylor_exp_sgd}
\end{align}

\noindent\textbf{Step 3: Matching terms} \\
From the above expansion of the discrete updates (Step 1) and of the continuous updates (Step 2) we can find the value of $f_1$ at $\vtheta_{t-1}$ by matching the terms of order $\mathcal{O}(h^2)$:
\begin{align}
\sum_{\tau = 0}^{n-1} &\sum_{\mu =\tau+1 }^{n-1}  \evaljacdata{\vtheta}{f}{\vtheta_{t-1}}{\vX^{t+\mu}}  f(\vtheta_{t  -1}; \vX^{t+\tau}) \owntag{SGD: Eq~\eqref{eq:sgd_n_exp}} \\ &= n^2 \left[f_1 + \frac{1}{2} \jacparam{\vtheta}{f(\vtheta; \{\vX^{t}, \dots, \vX^{t+n-1} \})}  f(\vtheta; \{\vX^{t}, \dots, \vX^{t+n-1} \}) \right] \owntag{Flow: Eq~\eqref{eq:taylor_exp_sgd}}
\end{align}
Leading to
\begin{align}
f_1(\vtheta_{t-1}) &= \underbrace{- \frac{1}{2} \evaljacdata{\vtheta}{f}{\vtheta_{t-1}}{\{\vX^{t}, \dots, \vX^{t+n-1} \}}  f(\vtheta_{t-1}; \{\vX^{t}, \dots, \vX^{t+n-1} \})}_{\text{the usual drift term}} \\ &\quad + \frac{1}{n^2}\sum_{\tau = 0}^{n-1} \sum_{\mu =\tau+1 }^{n-1}  \evaljacdata{\vtheta}{f}{\vtheta_{t  -1}}{\vX^{t+\mu}}  f(\vtheta_{t  -1}; \vX^{t+\tau}) .
\label{eq:f_1_sgd_ours}
\end{align}
From here we choose a function $f_1$ which satisfies the above constraint.
\begin{align}
f_1(\vtheta) &= \underbrace{- \frac{1}{2} \jacparam{\vtheta}{f(\vtheta; \{\vX^{t}, \dots, \vX^{t+n-1} \})}  f(\vtheta; \{\vX^{t}, \dots, \vX^{t+n-1} \})}_{\text{the usual drift term}} \\ &\quad+ \frac{1}{n^2} \sum_{\tau = 0}^{n-1} \sum_{\mu =\tau+1 }^{n-1}  \jacparam{\vtheta}{f(\vtheta; \vX^{t+\mu})}  f({\color{red}{\vtheta_{t  -1}}}; \vX^{t+\tau}) 
\end{align}
We now replace $f = - \nabla_{\vtheta} E$, using that in this case $\jacthetaf = \nabla_{\vtheta}^2 E$,
\begin{align}
f_1(\vtheta_{t-1}) &= - \frac{1}{2} \jacparam{\vtheta}{\nabla_{\vtheta}E(\vtheta; \{\vX^{t}, \dots, \vX^{t+n-1} \})}  \nabla_{\vtheta}E(\vtheta; \{\vX^{t}, \dots, \vX^{t+n-1} \}) \\ &\quad + \frac{1}{n^2} \sum_{\tau = 0}^{n-1} \sum_{\mu =\tau+1 }^{n-1}  \jacparam{\vtheta}{\nabla_{\vtheta}E(\vtheta; \vX^{t+\mu})}  \nabla_{\vtheta}E({\color{red}{\vtheta_{t  -1}}}; \vX^{t+\tau}) \\
  &= - \nabla_{\vtheta}\Big(\frac{1}{4} \norm{ \nabla_{\vtheta}E(\vtheta; \{\vX^{t}, \dots, \vX^{t+n-1} \})}^2 \owntag{using \eqref{eq:normssametheta}} \\&\quad \quad \quad \quad - \frac{1}{n^2} \sum_{\tau = 0}^{n-1} \sum_{\mu =\tau+1 }^{n-1}  \nabla_{\vtheta} E(\vtheta; \vX^{t+\mu})^T \nabla_{\vtheta}E({\color{red}{\vtheta_{t  -1}}}; \vX^{t+\tau}) \Big)  .
\end{align}
This leads to the modified loss
\begin{align}
\tilde{E}(\vtheta) &= E(\vtheta; \{\vX^{t}, \dots, \vX^{t+n-1} \})\\ 
                   & \quad+  \frac{n h}{4} \norm{ \nabla_{\vtheta}E(\vtheta; \{\vX^{t}, \dots, \vX^{t+n-1} \})}^2 \\
                   & \quad- \frac{h}{n} \sum_{\tau = 0}^{n-1} \sum_{\mu =\tau+1 }^{n-1}  \nabla_{\vtheta} E(\vtheta; \vX^{t+\mu})^T \nabla_{\vtheta}E({\color{red}{\vtheta_{t  -1}}}; \vX^{t+\tau}).
\end{align}
Through algebraic manipulation we can write the above in the form
\begin{align}
\tilde{E}(\vtheta) &= E(\vtheta; \{\vX^{t}, \dots, \vX^{t+n-1} \})\\ 
                   & \quad + \frac{n h}{4} \norm{ \nabla_{\vtheta}E(\vtheta; \{\vX^{t}, \dots, \vX^{t+n-1} \})}^2 \\
                   & \quad - \frac{h}{n}\sum_{\mu =1 }^{n-1} \left[ \nabla_{\vtheta} E(\vtheta; \vX^{t+\mu})^T \left(\sum_{\tau = 0}^{\mu-1}    \nabla_{\vtheta}E({\color{red}{\vtheta_{t  -1}}}; \vX^{t+\tau})\right)\right].
\end{align}

We can now compare this with the modified loss we obtained by ignoring stochasticity and assuming both updates have been done with a full batch; this entails using the IGR loss:
\begin{align}
\tilde{E} = E(\vtheta;\{\vX^{t}, \dots, \vX^{t+n-1} \}) + \frac{h}{4} \|\nabla_{\vtheta} E(\vtheta; \{\vX^{t}, \dots, \vX^{t+n-1} \}) \|^2 
\end{align}

Thus, when using multiple batches, there is the additional pressure to maximise the dot product of the gradients obtained using the last $k$ batches that came before the current batch, evaluated at the parameters at which we started the $n$ iterations.

\section{Multiple steps of full-batch gradient descent}
\label{sec:igr_multiple_steps}

To contrast our results with multiple steps of gradient descent \textit{with the same batch} (full-batch case), we show that the IGR flow follows gradient descent with error of order $\mathcal{O}(h^3)$ after $n$ gradient descent steps.
The proof steps follow the same approach as above, with two main differences: first, we assume all batches are the same, and second, we no longer fix the starting parameters when choosing $f_1$. With the same steps as before we obtain Eq~\eqref{eq:f_1_sgd_ours}:
\begin{align}
f_1(\vtheta_{t-1}) &= - \frac{1}{2} \evaljacdata{\vtheta}{f}{\vtheta_{t-1}}{\{\vX^{t}, \dots, \vX^{t+n-1} \}}  f(\vtheta_{t-1}; \{\vX^{t}, \dots, \vX^{t+n-1} \}) \\ &\quad + \frac{1}{n^2}\sum_{\tau = 0}^{n-1} \sum_{\mu =\tau+1 }^{n-1}  \evaljacdata{\vtheta}{f}{\vtheta_{t  -1}}{\vX^{t+\mu}}  f(\vtheta_{t  -1}; \vX^{t+\tau}) .
\end{align}
Assuming identical batches and using that $f$ is an empirical average, we obtain $f(\cdot; \{\vX^{t}, \dots, \vX^{t+n-1} \}) = f(\cdot; \vX^{t}) = f(\cdot; \vX^{t + i})$, for any choice of $i$.
We then have 
\begin{align}
f_1(\vtheta) &= - \frac{1}{2} \jacparam{\vtheta}{f(\vtheta, \vX^{t})}  f(\vtheta; \vX^{t}) + \frac{1}{n^2}\sum_{\tau = 0}^{n-1} \sum_{\mu =\tau+1 }^{n-1}  \jacparam{\vtheta}{f(\vtheta, \vX^{t})}  f(\vtheta; \vX^{t}) = \\
   &= (- \frac{1}{2} + \frac{n(n-1)}{2 n^2})\jacparam{\vtheta}{f(\vtheta, \vX^{t})}  f(\vtheta; \vX^{t}) \\
   &= - \frac{1}{2n} \jacparam{\vtheta}{f(\vtheta, \vX^{t})}  f(\vtheta; \vX^{t})
\end{align}
Replacing $f = - \nabla_{\vtheta}E(\cdot; \vX^{t})$ into the form of the modified flow in Eq~\ref{eq:taylor_exp_sgd}, we obtain:
\begin{align}
\dot{\vtheta} &= - \nabla_{\vtheta}E(\vtheta; \vX^{t}) - \frac{nh}{2n} \nabla_{\vtheta}^2{E(\vtheta, \vX^{t})}  \nabla_{\vtheta}E(\vtheta; \vX^{t}) \\
&= - \nabla_{\vtheta}E(\vtheta; \vX^{t}) - \frac{h}{2} \nabla_{\vtheta}^2{E(\vtheta, \vX^{t})}  \nabla_{\vtheta}E(\vtheta; \vX^{t}),
\end{align}
which recovers the IGR flow. Importantly, the flow  in the full-batch case \textit{does not depend on the number of iterations $n$}.

\subsection{Expectation over all shufflings}
\label{sec:sgd_igr_comp}
We contrast our results with those of~\citet{igr_sgd}, who construct a modified loss over an epoch of stochastic gradient descent in expectation over all possible shufflings of the batches in the epoch. That is, \citet{igr_sgd} describe expected value of the modified loss $\mathbb{E}_{\sigma} \left[\tilde{E}(\vtheta; \{\vX^{\sigma(t)}, \dots, \vX^{\sigma(t+n-1)} \}) \right]$, where $n$ is the number of batches in an epoch, and $t$ is the iteration at which the new epoch start, while $\sigma$ denotes the set of all possible permutations of batches $\{1, ...n\}$. While our results do not require an expectation and account for the exact batches used, to constrast our results with \citet{igr_sgd}, we also take an expectation over all possible batch shufflings (but not the elements in the batch) in an epoch in our results shown in Eq~\eqref{eq:modified_sup_learning}, and obtain
\begin{align}
\mathbb{E_\sigma} \left[E_{sgd}(\vtheta) \right]&= E(\vtheta; \{\vX^{t}, \dots, \vX^{t+n-1} \}) + \frac{nh}{4}  \norm{\nabla_{\vtheta} E(\vtheta;\{\vX^{t}, \dots, \vX^{t+n-1} \})}^2 
 \\& \hspace{1em}- \frac{h}{n} \mathbb{E_\sigma} \left[ \sum_{k=0}^{n-1} \nabla_{\vtheta} E(\vtheta;\vX^{t+k}) ^T\left(\sum_{i=0}^{k-1} \nabla_{\vtheta} E(\vtheta_{t-1}; \vX^{t+i})\right) \right] \\
                   &= E(\vtheta; \{\vX^{t}, \dots, \vX^{t+n-1} \}) + \frac{nh}{4}  \norm{\nabla_{\vtheta} E(\vtheta;\{\vX^{t}, \dots, \vX^{t+n-1} \})}^2 
                   \\& \hspace{1em}- \frac{h}{n} \frac{1}{2} \mathbb{E_\sigma} \left[\sum_{k=0}^{n-1} \sum_{i=0, i \ne k}^{n-1} \nabla_{\vtheta} E(\vtheta;\vX^{t+k}) ^T \nabla_{\vtheta} E(\vtheta_{t-1}; \vX^{t+i})\right] ,
\end{align}
We used the symmetry of the permutation structure since for each permutation where $\sigma(i) < \sigma(j)$ there is also a permutation where $\sigma(i) > \sigma(j)$ by swapping the values of $\sigma(i)$ and $\sigma(j)$.
We expand the last term
\begin{align}
&\mathbb{E}_{\sigma} \left[E_{sgd} \right] 
                   =  E(\vtheta; \{\vX^{t}, \dots, \vX^{t+n-1} \}) + \frac{nh}{4}  \norm{\nabla_{\vtheta} E(\vtheta;\{\vX^{t}, \dots, \vX^{t+n-1} \})}^2 \\
                   &\hspace{1em} - \frac{h}{n} \frac{1}{2} \mathbb{E_\sigma} \left[\sum_{k=0}^{n-1} \sum_{i=0}^{n-1} \nabla_{\vtheta} E(\vtheta;\vX^{t+k}) ^T \nabla_{\vtheta} E(\vtheta_{t-1}; \vX^{t+i})\right]\\&\hspace{1em}  -  \frac{h}{n} \frac{1}{2} \mathbb{E_\sigma} \left[\sum_{k=0}^{n-1}  \nabla_{\vtheta} E(\vtheta;\vX^{t+k}) ^T \nabla_{\vtheta} E(\vtheta_{t-1}; \vX^{t+k})\right]
\end{align}
From here
\begin{align}
 \mathbb{E}_{\sigma} \left[E_{sgd} \right]                   &=  E(\vtheta; \{\vX^{t}, \dots, \vX^{t+n-1} \}) + \frac{nh}{4}  \norm{\nabla_{\vtheta} E(\vtheta;\{\vX^{t}, \dots, \vX^{t+n-1} \})}^2 
                    \\&\hspace{1em}- \frac{h}{2n} \nabla_{\vtheta} E(\vtheta; \{\vX^{t}, \dots, \vX^{t+n-1} \})^T \nabla_{\vtheta} E(\vtheta_{t-1}; \{\vX^{t}, \dots, \vX^{t+n-1} \} 
                   \\&\hspace{1em}  -  \frac{h}{2n}\left[\sum_{k=0}^{n-1}  \nabla_{\vtheta} E(\vtheta;\vX^{t+k}) ^T \nabla_{\vtheta} E(\vtheta_{t-1}; \vX^{t+k})\right].
\end{align}
We obtain that the pressure to minimise individual batch gradient norms is translated into a pressure to maximise the dot product between the gradients at the end of the epoch with those at the beginning of the epoch, both for each batch and for the entire dataset.

\section{Two-player games}

\subsection{Effects on the non-saturating GAN}
\label{app:non_sat_gan}

We now expand of implicit regularisation terms found in Section~\ref{sec:dd_general_modified_losses} on GANs, specifically, the effect of the interaction terms, which takes the form of a dot product. We will denote the first player, the discriminator, as $D$, parametetrised by $\vphi$, and the generator as $G$, parametrised by $\vtheta$. We denote the data distribution as $p^*(\vx)$ and the latent distribution $p(\vz)$.
Consider the non-saturating GAN loss described by \citet{goodfellow2014generative}:
\begin{align}
    E_{\vphi}(\vphi, \vtheta) &= \mathbb{E}_{p^*(\vx)} \log D(\vx; \vphi) + \mathbb{E}_{p(\vz)} \log (1 - D(G(\vz; \vtheta); \vphi)) \\
     E_{\vtheta}(\vphi, \vtheta) &=  \mathbb{E}_{p(\vz)} - \log D(G(\vz; \vtheta); \vphi) 
\end{align}

Consider the interaction term for the discriminator (see Eq \eqref{eq:disc_int_app}), and replace the definitions of the above loss functions:
\begin{align}
\nabla_{\vtheta} & E_{\vphi}^T \nabla_{\vtheta} E_{\vtheta}(\vphi_{t-1}, \vtheta_{t-1}) \\ &= \nabla_{\vtheta} \left(\mathbb{E}_{p^*(\vx)} \log D(\vx; \vphi) + \mathbb{E}_{p(\vz)} \log (1 - D(G(\vz; \vtheta); \vphi)\right)^T \nabla_{\vtheta} E_{\vtheta}(\vphi_{t-1}, \vtheta_{t-1}) \\
 &= \nabla_{\vtheta} \left(\mathbb{E}_{p(\vz)} \log (1 - D(G(\vz; \vtheta); \vphi)\right)^T \nabla_{\vtheta} E_{\vtheta}(\vphi_{t-1}, \vtheta_{t-1}) \\
  &=\left(\mathbb{E}_{p(\vz)}  \nabla_{\vtheta}  \log (1 - D(G(\vz; \vtheta); \vphi)\right)^T \nabla_{\vtheta} E_{\vtheta}(\vphi_{t-1}, \vtheta_{t-1}) \\
    &=\left(- \mathbb{E}_{p(\vz)} \frac{1}{1 - D(G(\vz; \vtheta); \vphi)} \nabla_{\vtheta}  D(G(\vz; \vtheta); \vphi)\right)^T \nabla_{\vtheta} E_{\vtheta}(\vphi_{t-1}, \vtheta_{t-1}) \\
    &=\left(- \mathbb{E}_{p(\vz)} \frac{1}{1 - D(G(\vz; \vtheta); \vphi)} \nabla_{\vtheta}  D(G(\vz; \vtheta); \vphi)\right)^T  \\ & \hspace{5em} \left( - \mathbb{E}_{p(\vz)} \frac{1}{D(G(\vz; \vtheta_{t-1}); \vphi_{t-1})} \nabla_{\vtheta}  D(G(\vz; \vtheta_{t-1}); \vphi_{t-1})\right) \\
    &=\left(\mathbb{E}_{p(\vz)} \frac{1}{1 - D(G(\vz; \vtheta); \vphi)} \nabla_{\vtheta}  D(G(\vz; \vtheta); \vphi)\right)^T \\ & \hspace{5em} \left( \mathbb{E}_{p(\vz)} \frac{1}{D(G(\vz; \vtheta_{t-1}); \vphi_{t-1})} \nabla_{\vtheta}  D(G(\vz; \vtheta_{t-1}); \vphi_{t-1})\right)
\end{align}
where the expectations can be evaluated at the respective mini-batches used in the updates for iterations $t$ and $t-1$, respectively. Consider $\vz^i_t$ the latent variable with index $i$ in the batch at time $t$. Then the above is approximated as
\begin{align}
 \frac{1}{B^2} \sum_{i,j=1}^B &c^{non-sat}_{i,j}  \nabla_{\vtheta}  D(G(\vz^i_t; \vtheta); \vphi)^T   \nabla_{\vtheta}  D(G(\vz^j_{t-1}; \vtheta_{t-1}); \vphi_{t-1}), \hspace{3em} \text{with} \\
    &c^{non-sat}_{i,j} = \frac{1}{1 - D(G(\vz^i_t; \vtheta); \vphi)} \frac{1}{D(G(\vz^j_{t-1}; \vtheta_{t-1}); \vphi_{t-1})}.
\end{align}

\chapter{\smoothnesstitle}

\section{Individual figure reproduction details}
\label{app:smoothness_exp}

\begin{itemize}
	\item Figure~\ref{fig:smoothness_double_descent}: the U-shape curved is obtained by fitting polynomials of increased degree on a simple one dimensional regression problem, with the true underlying function $f(x) = \sin(5 x) - \sin( 10x)$. The double descent curve is obtained by training Resnet-18 models on CIFAR-10.
	\item Figure~\ref{fig:smooth_wass_mlp} uses samples from two Beta distributions, one with parameters $3$ and $5$, and one with parameters $2$ and $3$. The MLP approximation to the density ratio is obtained with an MLP with units $[10, 10, 40, 1]$. The Wasserstein optimal critic is approximated using a linear programming approach~\citep{cuturi2014ground}.
	\item Figures~\ref{fig:two_moons_decision_surface},~\ref{fig:two_moons_layers_constant},~\ref{fig:two_moons_layers_local_constant} use the an MLP with 4 layers and 100, 100, 100 and 1 output unit, respectively. The shallow MLP has 2 layers of 100 and 1 unit. All methods were trained for 100 iterations on 50 data points. We use the Adam optimiser, with learning rate $0.01$, $\beta_1=0.9$, $\beta_2=0.999$. To compute the local Lipschitz function of the decision surface learned on two moons in Figure~\ref{fig:two_moons_layers_local_constant}, we split the space into small neighbourhoods (2500 equally sized grids); for each grid, we sample 2500 random pairs of points in the grid and report $\max \norm {f(\vx) - f(\vy)} / \norm{\vx - \vy}$.
	\item Figure~\ref{fig:mnist_classification} is obtained using an MLP with 4 layers of 1000, 1000, 1000 and 10 units each
and are trained for 500 iterations at batch size 100, reaching an accuracy of 95\% on the entire test set. The models are trained with the Adam optimiser, with $\beta_1=0.9$ $\beta_2 = 0.999$, and learning rates $0.001$ and $0.0005$.
	\item Figure~\ref{fig:sn_mom} For the GAN CIFAR-10 experiments, we use the architectures specified in the Spectral normalization paper~\citep{miyato2018spectral}. We use the Adam optimiser~\citep{kingma2014adam} with default $\beta_2$, $\beta_1$ as specified in the respective subfigure and a learning rate sweep $\{10^{-4}, 2 \times 10^{-4}, 3 \times 10^{-4}, 4 \times 10^{-4}\}$ for the discriminator and the same for the generator.
\end{itemize}

\chapter{Spectral Normalisation in Reinforcement learning}

\section{Additional experimental results}

\subsection{The weak correlation between smoothness and performance}
\label{sub:rl_smoothness_weak_correlations}

In Section~\ref{sec:rl_spectral_schedulers} we have made the argument that smoothness and performance are not strongly correlated. We visulise the individual results used to obtain the correlation values shown in the main thesis in Figure~\ref{fig:smoothness_correlate_performance_rl}.

\begin{figure}[t]
    \centering
 \begin{subfloat}[Average over seeds.]{
    \includegraphics[width=0.475\columnwidth]{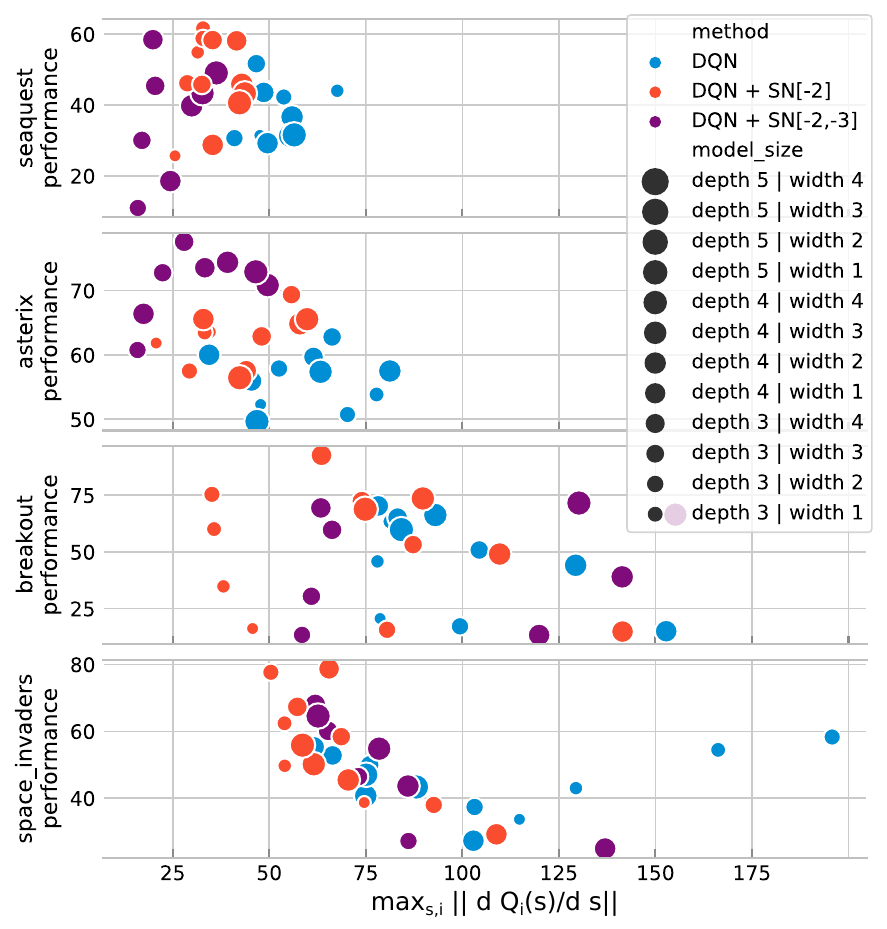}%
    \label{fig:smoothness_correlate_performance_rl_avg_seeds}%
 }
\end{subfloat}%
 \begin{subfloat}[Individual seeds.]{
 \includegraphics[width=0.525\columnwidth]{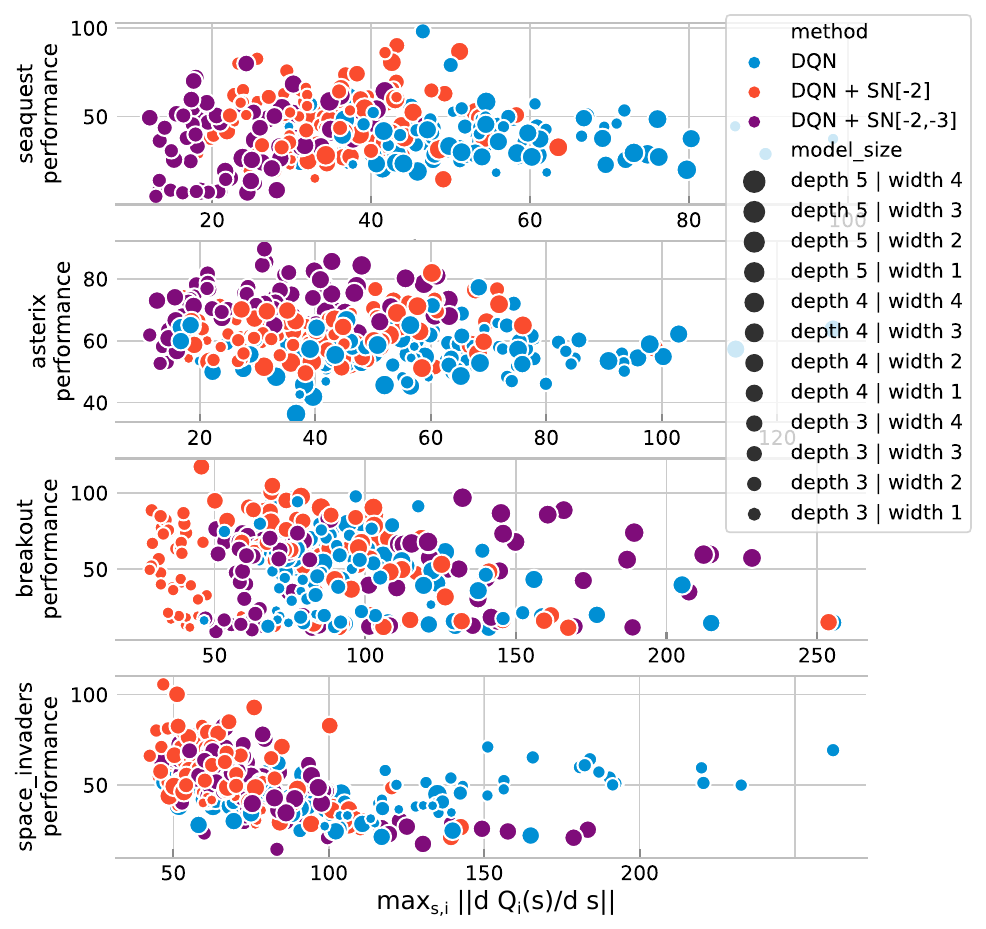}%
    \label{fig:smoothness_correlate_performance_rl_seeds}%
 }
\end{subfloat}%
    \caption[Applying Spectral Normalisation on a subset of layers does not always result in smoother networks.]{Applying normalisation does not always result in smoother networks. Even if normalising a subset of the network's layers makes the network less smooth than the baseline while performance still improves. Results averaged over 10 seeds in \subref{fig:smoothness_correlate_performance_rl_avg_seeds}; individual seeds shown in~\subref{fig:smoothness_correlate_performance_rl_seeds}.  Results are obtained from a sweep over architectures detailed in Table~\ref{tab:rl_architectures_sweep}.}
    \label{fig:smoothness_correlate_performance_rl}
\end{figure}

\subsection{Other regularisation methods}

\label{sub:rl_other_regularisation}
\begin{figure}[ht]
    \centering
    \includegraphics[width=\columnwidth]{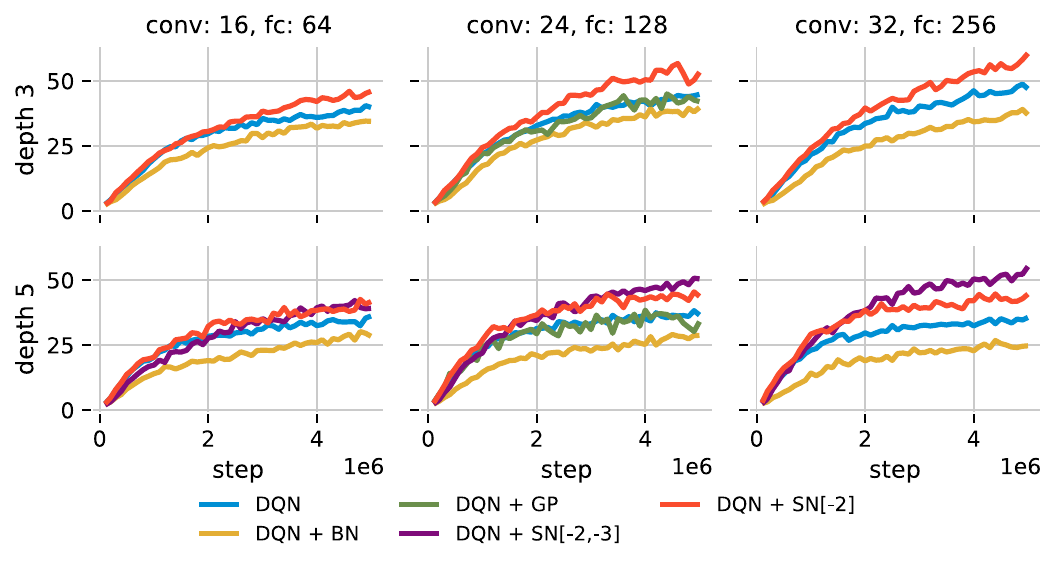}
    \caption[Other normalisation and regularisation methods do not recover Spectral Normalisation performance when applied to the DQN critic.]{\textbf{Regularisation does not recover Spectral Normalisation performance.} Performance on MinAtar games of Spectral Normalisation, gradient penalties and Batch Normalisation. Each line is an average over normalized scores of each game. Ten seeds for each configuration.}
    \label{fig:rl_minatar_regularisation}
\end{figure}

We investigated whether other regularisation methods imposing smoothness constraints can have similar effects on the agent's performance. To this end we ran experiments with both gradient penalties and Batch Normalisation on several architectures (Table~\ref{tab:rl_architectures_sweep}). We consistently observe that these approaches do not recover the performance of Spectral Normalisation.

\paragraph{Batch Normalisation.} For each of the architectures we employed Batch Normalisation after the ReLU activation of every convolutional or linear layer except the output.

\paragraph{Gradient Penalty.} We did extensive experimentation with gradient penalty regulairsation. We tried multiple forms of regularisation, including  penalising the norm of the sum of the gradients of all actions $\left\vert\left\vert \sum_{i=1}^{|\mathcal{A}|}\frac{d Q_i}{d {\vs}} \right\vert\right\vert_2$, regularising the expected norm of each Q-value with respect to the state $\sum_{i=1}^{|\mathcal{A}|} \left\vert\left\vert \frac{d Q_i}{d {\vs}} \right\vert\right\vert_2$, and also regularising the norm of the gradient of the Q-value associated with the optimal action $\left\vert\left\vert \frac{d \max Q_i}{d {\vs}} \right\vert\right\vert_2$. In all cases we swept through a wide range of penalty coefficients $\zeta$. In Figure~\ref{fig:rl_minatar_regularisation} we report the results of the best setting we could identify.

\paragraph{Relaxations to 1-Lipschitz normalisation.} 

We now investigate whether relaxing the 1-Lipschitz constraint imposed by Spectral Normalisation to a $K$-Lipschitz constraint allows us to normalise more layers without a decrese in performance. We have seen in Chapter \ref{ch:smoothness} that this can occur on simple classification tasks; the need for relaxing the Lipschitz constraint has also been observed by \citet{gouk2020regularisation}.
Figure~\ref{fig:minatar_lipschitz_k} shows that $K > 1$ values Spectral Normalisation can be applied to multiple layers, and it further leads to an increase in performance. As expected, we also observe that for rather large $K$ (such as $K=10$), Spectral Normalisation no longer provides a benefit.
The increased computation required when approximating the spectral norm for all the layers and the addition of one hyperparameter per layer determined us to not pursue this setup further.

\begin{figure}[ht!]
    \centering
    \includegraphics[width=\columnwidth]{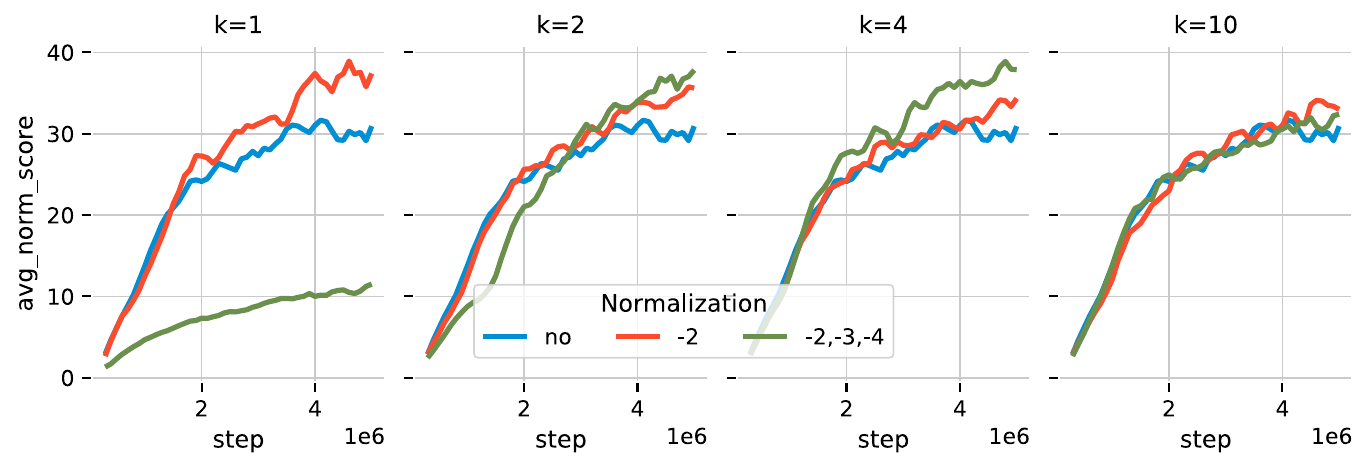}
    \caption[Assessing the effect of the Lipschitz constant $K$ when applying Spectral Normalisation to a DQN agent.]{
        A sweep over Lipschitz constants $K$ when normalising all but the last layer of the critic. We see that while for $K=1$, normalising 3 layers drastically reduces performance, this is not the case for $K>1$. For large $K$, however, we also do not observe any performance improvement.
        Each line is an average over normalised scores of 4 games $\times$ 10 seeds.}
    \label{fig:minatar_lipschitz_k}
\end{figure}

\subsection{\textsc{divOut}, \textsc{divGrad} and \textsc{mulEps}}

\begin{figure*}[ht]
    \centering
    \includegraphics[width=\textwidth]{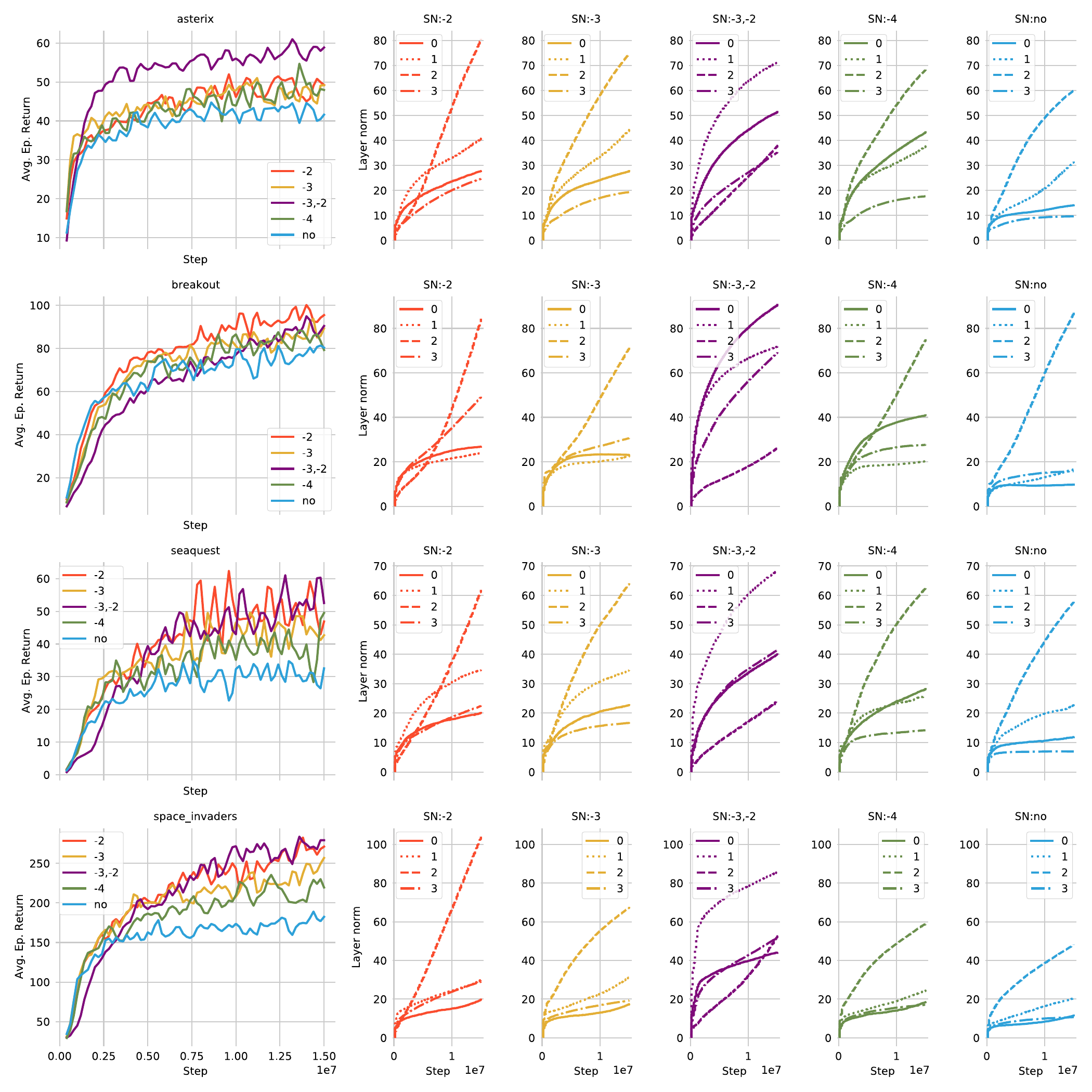}
    \caption[Spectral radii in a long training run  on MinAtar using a 4-layer critic.]{All spectral radii for the 15M experiment on MinAtar using a 4-layer architecture (conv=24-24,fc=128). Colors code the subsets of layers that are normalised (consistent with the rest of the document), while line styles code the four layers. Note how the penultimate layer has the largest spectral norm across all normalisation variants. 10 seeds.}
    \label{fig:dqn-minatar-all-radii-2}
\end{figure*}

 We present results comparing \textsc{divOut}, \textsc{mulEps} and \textsc{divGrad} across multiple normalised layers and architectures in Figure~\ref{fig:schedulers_detailed}.
\textsc{divOut} has a close behaviour to that of Spectral Normalisation. In contrast, the \textsc{mulEps} and \textsc{divGrad} optimisers outperform Spectral Normalisation when all hidden layers are normalised.
\begin{figure*}[ht]
    \centering
    \includegraphics[width=\textwidth]{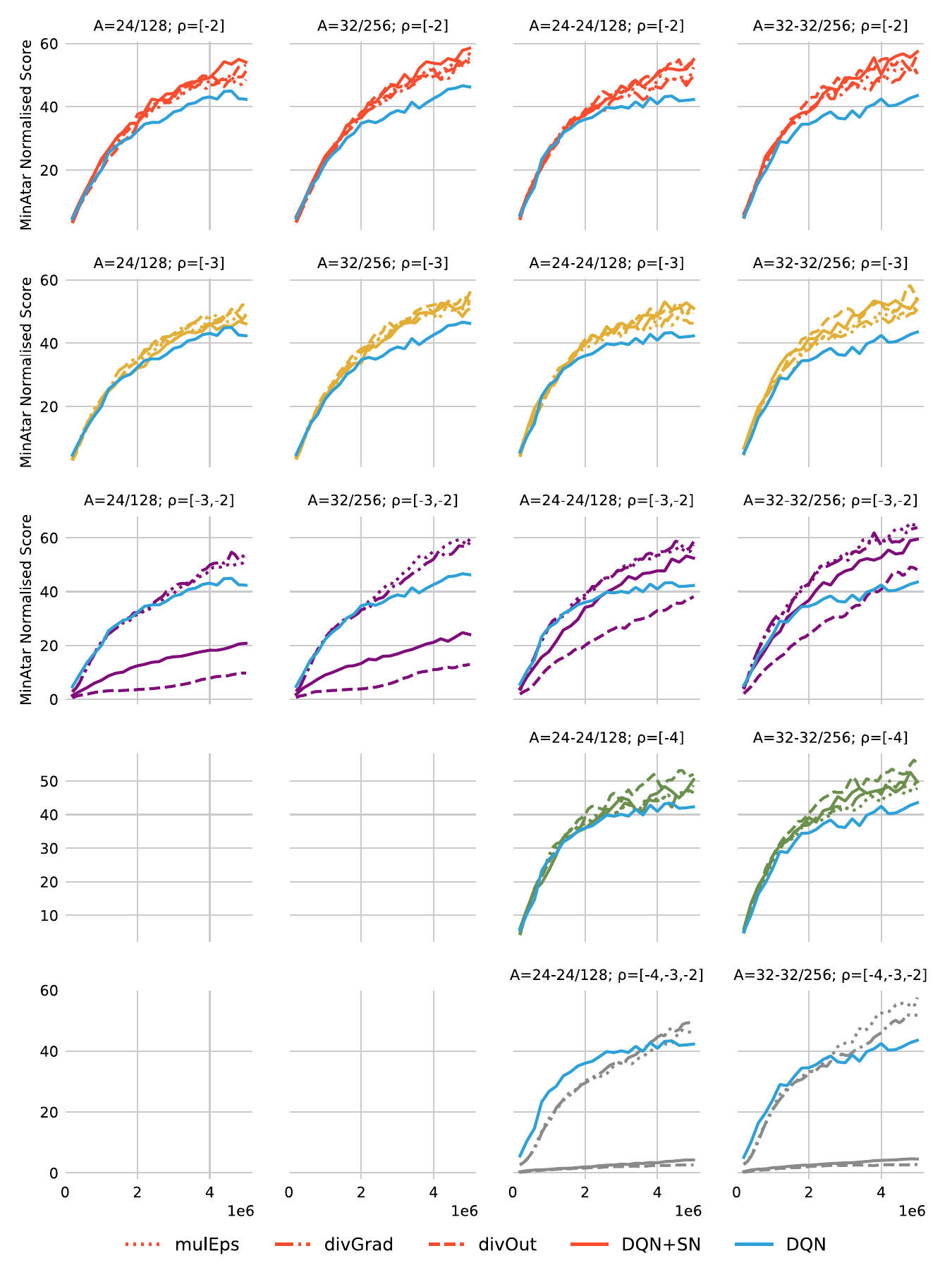}
    \caption[MinAtar Normalised Scores for the four architectures in Table~\ref{tab:rl_architectures_sweep}.]{MinAtar Normalised Scores for the four architectures in Table~\ref{tab:rl_architectures_sweep} and various subsets of layers whose spectral radii are used for Spectral Normalisation or optimisation methods we derived based on Spectral Normalisation. Notice that \textsc{divOut} behaves similarly to Spectral Normalisation (even when they fail to train), while \textsc{mulEps} and \textsc{divGrad} converge even when all hidden layers are normalised.}
    \label{fig:schedulers_detailed}
\end{figure*}

\clearpage

\section{Experimental details}

\subsection{Application of Spectral Normalisation}
\label{sec:app-rl-sn}

We use power iteration to approximate the spectral norm of weight matrices, as suggested by ~\citet{miyato2018spectral}.
For convolutional layers we adapt the procedure in \citet{gouk2020regularisation} and the two matrix-vector multiplications are replaced by convolutional and transposed convolutional operations.

\paragraph{Backpropagating through the norm.} Since the parameters tuned during optimisation are the unnormalised weights $\weight{i}$, we investigated whether there are any advantages in backpropagating through the power iteration step.
In a set of experiments performed on a subset of games of Atari we noticed no loss in performance when treating the spectral norm as a constant from the backpropagation perspective, and thus for computational efficiency we adopt this approach.

\subsection{Measuring smoothness: the norm of the Jacobians}

Experiments in Section~\ref{sub:rl_smoothness_weak_correlations} used the maximum norm of the Jacobians w.r.t. the inputs as an indirect metric for network function's smoothness. This is to avoid using potential loose bounds discussed in Chapter~\ref{ch:smoothness}, since the exact computation of the Lipschitz constant of a network is NP-hard \citep{virmaux2018lipschitz}. For each network we collected thousands of states (\emph{on-policy}), computing the maximum Euclidean norm of the Jacobian w.r.t. the inputs: $\max_{i, \vx} \left\vert\left\vert\frac{d Q_i}{d {\vs}} \right\vert\right\vert_2$, where $\left\vert\left\vert \vs \right\vert\right\vert_2$ denotes the Frobenius norm and $Q_i$ denotes the $Q$ value corresponding to action $i$.

We note that what we compute is a lower bound of the Lipschitz constant. To see why consider $Q(\cdot;\vtheta): \mathbb{R}^{|\mathcal{S}|} \rightarrow \mathbb{R}^{|\mathcal{A}|}$ with Lipschitz constant w.r.t. the Frobenius norm $K$. We then we have that:
\begin{align}
\max_{i} \left\vert\left\vert \frac{d Q_i}{d {\vs}} \right\vert\right\vert_2^2 
\le \left\vert\left\vert \frac{d Q}{d {\vs}} \right\vert\right\vert_2^2  \le \min(|\mathcal{S}|, |\mathcal{A}|) \left\vert\left\vert \frac{d Q}{d {\vs}} \right\vert\right\vert_{op}^2 \le \min(|\mathcal{S}|, |\mathcal{A}|) K^2.
\end{align}
Since we chose the value of $\vs$ with maximum norm, this ensures that the bound is as tight as possible with the available data (though this can be made tighter by computing the $\max \left\vert\left\vert \frac{d Q}{d \vs} \right\vert\right\vert_2^2$).

\subsection{Evaluation}

The mean and median Human Normalised Score have been critiqued in the past \citep{machado2018revisiting,toromanoff2019deep} because they can be dominated by some large normalised scores.
However we notice that with very few exceptions ($3/54$ for DQN and $11/54$ games for C51) normalising at most one of the network's layers will not degrade the performance compared to the baseline but instead it will improve upon it, often substantially.

\subsection{Minatar}
\label{sec:app-minatar}

\paragraph{Game selection.} MinAtar \cite{young19minatar} benchmark is a collection of five games that reproduce the dynamics of ALE counterparts, albeit in a smaller observational space. Out of \emph{Asterix, Breakout, Seaquest, Space Invaders} and \emph{Freeway} we excluded the latter from all our experiments since all agents performed similarly on this game.

\paragraph{Network architecture.} All experiments on MinAtar are using a convolutional network with $L_C$ convolutional layers with the same number of channels, a hidden linear layer, and the output layer. The number of input channels is game-dependent in MinAtar. All convolutional layers have a kernel size of $3$ and a stride of $1$. All hidden layers have rectified linear units. Whenever we vary depth we change the number of convolutional layers $L_C$, keeping the two linear layers. When the width is varied we change the width of both convolutional layers  and the penultimate linear layer. All convolutional layers are always identically scaled.
We list all the architectures used in various experiments described in this section in Table~\ref{tab:rl_architectures_sweep}.

 \paragraph{General hyperparameter settings.} In all our MinAtar experiments we used the same set of hyperparameters returned by a small grid search around the initial values published by \citet{young19minatar}. We list the values we settled on in Table~\ref{tab:app-minatar_hyperparameters}. For the rest of this section we only mention how we deviate from this set of hyperparameters and settings for each of the experiments that follow.

\begin{table}[t]
    \vskip 0.15in
    \begin{center}
        \begin{small}
            \begin{tabular}{lcc}
                \toprule
                \textsc{Hyperparameter}        & \textsc{Value} \\
                \midrule
                \rule{0pt}{2ex}%
                discount $\gamma$               & \textsc{0.99} \\
                update frequency                & \textsc{4} \\
                target update frequency         & \textsc{4,000} \\
                \midrule
                \rule{0pt}{1ex}%
                starting $\epsilon$             & \textsc{1.0} \\
                final $\epsilon$                & \textsc{0.01} \\
                $\epsilon$ steps                & \textsc{250,000} \\
                $\epsilon$ schedule             & linear \\
                warmup steps                    & \textsc{5,000} \\
                \midrule
                \rule{0pt}{1ex}%
                replay size                     & \textsc{100,000} \\
                history length                  & \textsc{1} \\
                \midrule
                \rule{0pt}{1ex}%
                cost function                   & \textsc{MSE Loss} \\
                optimiser                       & \textsc{Adam} \\
                learning rate $h$            & \textsc{0.00025} \\
                damping term $\epsilon$         & \textsc{0.0003125} \\
                $\beta_1, \beta_2$              & \textsc{(0.9, 0.999)} \\
                \midrule
                \rule{0pt}{1ex}%
                validation steps                & \textsc{125000} \\
                validation $\epsilon$           & \textsc{0.001} \\
                \bottomrule
            \end{tabular}
        \end{small}
    \end{center}
    \caption{
        MinAtar default hyperparameter settings, used for all MinAtar experiments unless otherwise specified.
    }
    \label{tab:app-minatar_hyperparameters}
\end{table}

\paragraph{MinAtar Normalised Score.} In our work we present many MinAtar experiments as averages over the four games we tested on. Since in MinAtar the range of the expected returns is game dependent we normalise the score. Inspired by the Human Normalised Score in \citet{mnih2015human} we take the largest score ever recorded by a baseline agent in our experiments and use it to compute $\text{MNS} = 100 \times (\text{score}_\text{agent} - \text{score}_\text{random}) / (\text{score}_\text{max} - \text{score}_\text{random})$. We use the resulting MinAtar Normalised Score whenever we report performance aggregates over the games.

\begin{table}[t]
    \vskip 0.15in
    \begin{center}
        \begin{small}
            \begin{tabular}{lcc}
                \toprule
                \textsc{Game}   & \textsc{Max}    & \textsc{Random} \\
                \midrule
                \rule{0pt}{2ex}%
                Asterix         & \textsc{78.90}     & \textsc{0.49} \\
                Breakout        & \textsc{122.88}    & \textsc{0.52} \\
                Seaquest        & \textsc{93.91}     & \textsc{0.09} \\
                Space Invaders  & \textsc{360.92}    & \textsc{2.86} \\
                \bottomrule
            \end{tabular}
        \end{small}
    \end{center}
    \caption{
        MinAtar maximum and random scores used for computing the MinAtar Normalised Score.
    }
    \label{tab:app-minatar_normalization_scores}
\end{table}

\noindent \textbf{Netwok architecture}. The default network architecture used for MinAtar experiments is shown in Table~\ref{tab:detault_atari_network}.

\begin{table}[t]
    \vskip 0.15in
    \begin{center}
        \begin{small}
            \begin{tabular}{ll}
                \toprule
                Network hyperparameter           & Value \\
                \midrule
                Number of convolutional channels & 16\\
                Filter size for convolutional channels &  3 $\times$ 3 \\
                Stride for convolutional channels &  1\\
                Hidden units for fully connected layers &  128 \\
                Number of fully connected layers &  2 \\
                Number of convolutional layers &  2 \\
                Number of outputs &  Number of actions \\
                \bottomrule
            \end{tabular}
        \end{small}
    \end{center}
    \caption{
        Network used for MinAtar experiments. The last fully connected layer has an output size given by the number of actions.
    }
    \label{tab:detault_minatar_network}
\end{table}

\subsection{Atari experiments}
\label{sec:app-atari}

\textbf{Evaluation protocols on Atari.}
Since we mostly compare our ALE results with the \rainbow~agent, we adopt the evaluation protocol from  \citet{hessel2018RainbowCI}.
Every 250K training steps in the environment we suspend the learning and evaluate the agent on 125K steps (or 500K frames). All the agents we train on ALE follow this validation protocol, the only difference being the validation epsilon value: $\epsilon=0.001$ for C51 and DQN-Adam that we directly compare to \rainbow, which uses the same value we use.

\noindent\textbf{DQN-Adam}.
We list the full details in Table~\ref{tab:app-dqn_adam_hyperparameters} especially since these hyperparameters differ considerably from the the original DQN agent; since we also wanted to be able to compare our results with those of \rainbow we use similar hyperparameters to those in \citet{hessel2018RainbowCI}.

\begin{table}[t]
    \vskip 0.15in
    \begin{center}
        \begin{small}
            \begin{tabular}{lcc}
                \toprule
                \textsc{Hyper-parameter}        & \textsc{Value} \\
                \midrule
                \rule{0pt}{2ex}%
                discount $\gamma$               & \textsc{0.99} \\
                update frequency                & \textsc{4} \\
                target update frequency         & \textsc{8000} \\
                \midrule
                \rule{0pt}{1ex}%
                starting $\epsilon$             & \textsc{1.0} \\
                final $\epsilon$                & \textsc{0.01} \\
                $\epsilon$ steps                & \textsc{250000} \\
                $\epsilon$ schedule             & linear \\
                warmup steps                    & \textsc{20000} \\
                \midrule
                \rule{0pt}{1ex}%
                replay size                     & \textsc{1M} \\
                batch size                      & \textsc{32} \\
                history length                  & \textsc{4} \\
                \midrule
                \rule{0pt}{1ex}%
                cost function                   & \textsc{SmoothL1Loss} \\
                optimiser                       & \textsc{Adam} \\
                learning rate $h$            & \textsc{0.00025} \\
                damping term $\epsilon$         & \textsc{0.0003125} \\
                $\beta_1, \beta_2$              & \textsc{(0.9, 0.999)} \\
                \midrule
                \rule{0pt}{1ex}%
                validation steps                & \textsc{125000} \\
                validation $\epsilon$           & \textsc{0.001} \\
                \bottomrule
            \end{tabular}
        \end{small}
    \end{center}
    \caption{
        DQN-Adam hyperparameters.
    }
    \label{tab:app-dqn_adam_hyperparameters}
\end{table}

\noindent \textbf{Netwok architecture}. The network architecture used for Atari experiments is shown in Table~\ref{tab:detault_atari_network}.
\begin{table}[t]
    \vskip 0.15in
    \begin{center}
        \begin{small}
            \begin{tabular}{ll}
                \toprule
                Network hyperparameter           & Value \\
                \midrule
                Number of convolutional channels & 32, 64, 64 \\
                Filter size for convolutional channels &  8 $\times$ 8, 4 $\times$ 4, 3 $\times$ 3 \\
                Stride for convolutional channels &  4, 2, 1\\
                Hidden units for fully connected layers &  512 \\
                Number of fully connected layers &  2 \\
                Number of outputs &  Number of actions \\
                \bottomrule
            \end{tabular}
        \end{small}
    \end{center}
    \caption[Network used for Atari experiments. ]{
        Network used for Atari experiments, taken from~\citet{hessel2018rainbow}. The last fully connected layer has an output size given by the number of actions.
    }
    \label{tab:detault_atari_network}
\end{table}

\section{Additional hyperparameter sweeps}

Network architecture results from other sweeps are shown in Table~\ref{tab:rl_architectures_sweep}.

\begin{table}[t]
    \vskip 0.15in
    \begin{center}
        \begin{small}
            \begin{tabular}{lccc}
                \toprule
                Experiment & \thead{Nr. of \\ conv layers}   & \thead{Conv \\ width}    & \thead{FC \\ width} \\
                \midrule
                \rule{0pt}{2ex}%
                \makecell[l]{Sweep over optimisation hyperparameters (Figure~\ref{fig:hypersensitivity_rl}.)} &
                \makecell{
                   \textsc{1} \\ \textsc{2} \\ \textsc{3} \\
                   \textsc{4} \\ \textsc{2} \\ \textsc{3} \\
                } &
                \makecell{
                   \textsc{24} \\ \textsc{24} \\ \textsc{24} \\
                   \textsc{24} \\ \textsc{32} \\ \textsc{32} \\
                } &
                \makecell{
                   \textsc{128} \\ \textsc{128} \\ \textsc{128} \\
                   \textsc{128} \\ \textsc{256} \\ \textsc{256} \\
                }\\
                \midrule
                \rule{0pt}{2ex}%
                \makecell[l]{Regularisation (Figure~\ref{fig:rl_minatar_regularisation})} &
                \makecell{
                   \textsc{1} \\ \textsc{3} \\ \textsc{1} \\
                   \textsc{3} \\ \textsc{1} \\ \textsc{3} \\
                } &
                \makecell{
                   \textsc{16} \\ \textsc{16} \\ \textsc{24} \\
                   \textsc{24} \\ \textsc{32} \\ \textsc{32} \\
                } &
                \makecell{
                   \textsc{64} \\ \textsc{64} \\ \textsc{128} \\ 
                   \textsc{128} \\ \textsc{256} \\ \textsc{256} \\ 
                }\\
                \midrule
                \rule{0pt}{2ex}%
                \makecell[l]{
                    Optimisation methods (Figure~\ref{fig:schedulers_detailed})\\
                } &
                \makecell{
                   \textsc{1} \\ \textsc{2} \\ \textsc{1} \\ \textsc{2} \\
                } &
                \makecell{
                   \textsc{24} \\ \textsc{24} \\ \textsc{32} \\ \textsc{32} \\
                } &
                \makecell{
                   \textsc{128} \\ \textsc{128} \\ \textsc{256} \\ \textsc{256} \\ 
                }\\
                \bottomrule
                   \makecell[l]{
                   Smoothness analysis (Figures~\ref{fig:minatar_SN_depth_vs_width}~\ref{fig:smoothness_correlate_performance_rl} and~\ref{fig:smoothness_correlate_performance_rl_seeds},Tables~\ref{tab:smoothness_correlate_performance_rl} and \ref{tab:smoothness_sn_applied_performance_rl})
                } &
                 \makecell{
                     1\\ 2\\ 1\\ 3\\ 2\\ 3\\ 3\\ 2\\ 2\\ 1\\ 3\\ 1 \\
                } &
                 \makecell{
                    8\\ 8\\ 32\\ 8\\ 32\\ 32\\ 24\\ 16\\ 24\\ 24\\ 16\\ 16 \\
                } &
                \makecell{
                 32\\ 32\\ 256\\ 32\\ 256\\ 256\\ 128\\ 64\\ 128\\ 128\\ 64\\ 64 \\
                } 
            \end{tabular}
        \end{small}
    \end{center}
    \caption[Details of neural architecture sweeps.]{Details of neural architecture sweeps. The total number of critic layers is the number of convolutional layers specified here, followed by two linear layers.}
    \label{tab:rl_architectures_sweep}
\end{table}

\clearpage

\section{Individual figure reproduction details}

\begin{itemize}
\item Figure~\ref{fig:rl_main_atari_results} uses the default Atari network shown in Table~\ref{tab:detault_atari_network}. The DQN-Adam hyperparameters as shown in Table~\ref{tab:app-dqn_adam_hyperparameters}. The evaluation protocol is described in Section~\ref{sec:app-atari}. The C51 hyperparameters are the same as those in \citet{hessel2018RainbowCI}.
\item Figure~\ref{fig:minatar_SN_depth_vs_width} performs a sweep over depth and width of networks on MinAtar. The optimisation and reinforcement learning hyperparameters as shown in Table~\ref{tab:app-minatar_hyperparameters}. The model sweep is described in Table~\ref{tab:rl_architectures_sweep}.
\item Figure~\ref{fig:atari_dqn_hns} presents both Atari and MinAtar results. The Atari results in Figure~\ref{fig:atari_dqn_ale_g54_hns} use the default Atari network shown in Table~\ref{tab:detault_atari_network}. The DQN-Adam hyperparameters as shown in Table~\ref{tab:app-dqn_adam_hyperparameters}. The Figure~\ref{fig:min_atar_dqn_ale_g54_hns} uses the hyperparameters as shown in Table~\ref{tab:app-minatar_hyperparameters} and the network in Table~\ref{tab:detault_minatar_network}.
\item Tables~\ref{tab:smoothness_sn_applied_performance_rl} and~\ref{tab:smoothness_correlate_performance_rl} use data from the same experiment as Figure~\ref{fig:minatar_SN_depth_vs_width}.
\item Figure~\ref{fig:hypersensitivity_rl} uses a MinAtar sweep over architectures detailed in Table~\ref{tab:rl_architectures_sweep}, with a sweep over learning rates $h \in \{0.00001, ..., 0.00215\}$ and $\epsilon \in \{0.00001, ..., 0.01\}$. The remaining hyperparameters are the ones in Table~\ref{tab:app-minatar_hyperparameters}.
\item Figure~\ref{fig:spectral_schedulers_rl} used the default MinAtar optimisation and reinforcement learning hyperparameters as shown in Table~\ref{tab:app-minatar_hyperparameters}, and the network in Table~\ref{tab:detault_minatar_network}.
\item Figures~\ref{fig:mul_eps},~\ref{fig:div_out_gans_d_and_g_sat_main},~\ref{fig:div_out_gans} use the architectures in Spectral Normalised GAN~\citep{miyato2018spectral}, with the Adam default hyperparameters with the exception of $\beta_1 = 0.5$, and learning rates sweeps as a cross product of $\{0.0001, 0.0002, 0.0003, 0.0004\}$ for the discriminator and generator. Simultaneous updates are used.
\item Figure~\ref{fig:spectral_norm_entropy} uses the architectures in Spectral Normalised GAN~\citep{miyato2018spectral}, with the Adam default hyperparameters with the exception of $\beta_1 = 0.5$, and learning rates sweeps as a cross product of $\{0.0001, 0.0002, 0.0003, 0.0004\}$ for the discriminator and generator. The generator uses the non-saturating loss. Alternating updates are used.
\item Figures~\ref{fig:spectral_norm_histogram_examples_early} and~\ref{fig:spectral_norm_histogram_examples} use the architectures in Spectral Normalised GAN~\citep{miyato2018spectral}, with the Adam default hyperparameters with the exception of $\beta_1 = 0.5$, and learning rates $0.0001$ for both the discriminator and generator. The generator is trained using the non-saturating loss.  Alternating updates are used.
\item Figure~\ref{fig:adam_curvature} uses the architectures in Spectral Normalised GAN~\citep{miyato2018spectral}, with the Adam default hyperparameters with the exception of $\beta_1 = 0.5$, and learning rates $0.0004$ for both the discriminator and generator. The $\epsilon$ values are provided in each subfigure. The generator is trained using the non-saturating loss. Simultaneous updates are used.
\item Figures~\ref{fig:smoothness_correlate_performance_rl} and~\ref{fig:smoothness_correlate_performance_rl_seeds} use a sweep over architectures detailed in Table~\ref{tab:rl_architectures_sweep}. The optimisation and reinforcement learning hyperparameters as shown in Table~\ref{tab:app-minatar_hyperparameters}.
\item Figure~\ref{fig:rl_minatar_regularisation} used a sweep over architectures detailed in Table~\ref{tab:rl_architectures_sweep}. The optimisation and reinforcement learning hyperparameters as shown in Table~\ref{tab:app-minatar_hyperparameters}.
\item Figure~\ref{fig:minatar_lipschitz_k} uses the hyperparameters as shown in Table~\ref{tab:app-minatar_hyperparameters} and the network in Table~\ref{tab:detault_minatar_network}.
\item Figure~\ref{fig:schedulers_detailed} is a different visualisation of the same experiment shown in Figure~\ref{fig:spectral_schedulers_rl}. The architectures are those in Table~\ref{tab:rl_architectures_sweep}.
\end{itemize}

\chapter{Geometric Complexity: a smoothness complexity measure implicitly regularised in deep learning}

\label{ch:gc_appendix}

\section{Figure and experiments reproduction details}

\begin{itemize}
\item Figure~\ref{fig:gc_initialisation}: the MLP shown have 500 units per layer.  GC is computed using 100 data points.
\item Figure~\ref{fig:training_sequence}: the MLP shown have 300 units per layer and 3 layers. The learning rate used is $0.02$.
\item Figure~\ref{fig:gc_explicit_regularisation}: The network used is a Resnet-18, with a batch size of 512 and a learning rate of $0.02$. Each experiment uses 3 seeds. The activation function is ReLu.
\item Figure~\ref{fig:explicit_gc_regularisation_cifar}: The network used is a Resnet-18, with a batch size of 128 and a learning rate of $0.02$. Each experiment uses 3 seeds. The activation function is ReLu.
\item Figure~\ref{fig:gc_learning_rate}: The network used is a Resnet-18, trained with batch size 512. Each experiment uses 3 seeds. The activation function is ReLu.
\item Figure~\ref{fig:gc_batch_sizes}: The network used is a Resnet-18, trained with learning rate $0.02$. Each experiment uses 3 seeds. The activation function is ReLu.
\item Figure~\ref{fig:double_descent}:  The network used is a Resnet-18, trained with learning rate $0.02$ and batch size 512. The activation function is ReLu.
\end{itemize}

\section{Additional experimental results}
\label{sec:gc_exp_app}

We present additional the implicit and explicit regularisation experiments using gradient descent \emph{with momentum}, which is widely used in practice. We observe the same phenomena as with gradient descent without momentum, namely that more implicit regularisation through decreased batch size and increased learning rate (Figure~\ref{fig:gc_mom_implicit_regularisation}) and explicit regularisation using gradient norm, spectral and L2 regularisation (shown in Figures~\ref{fig:gradient_norm_gc_cifar_app},~\ref{fig:spectral_gc_cifar_app} and~\ref{fig:l2_gc_cifar_app} respectively) produces solutions with higher test accuracy and lower Geometric Complexity (GC).

 \begin{figure}[t]
  \centering
  \begin{subfloat}[Learning rate.]{
  \includegraphics[width=0.45\linewidth]{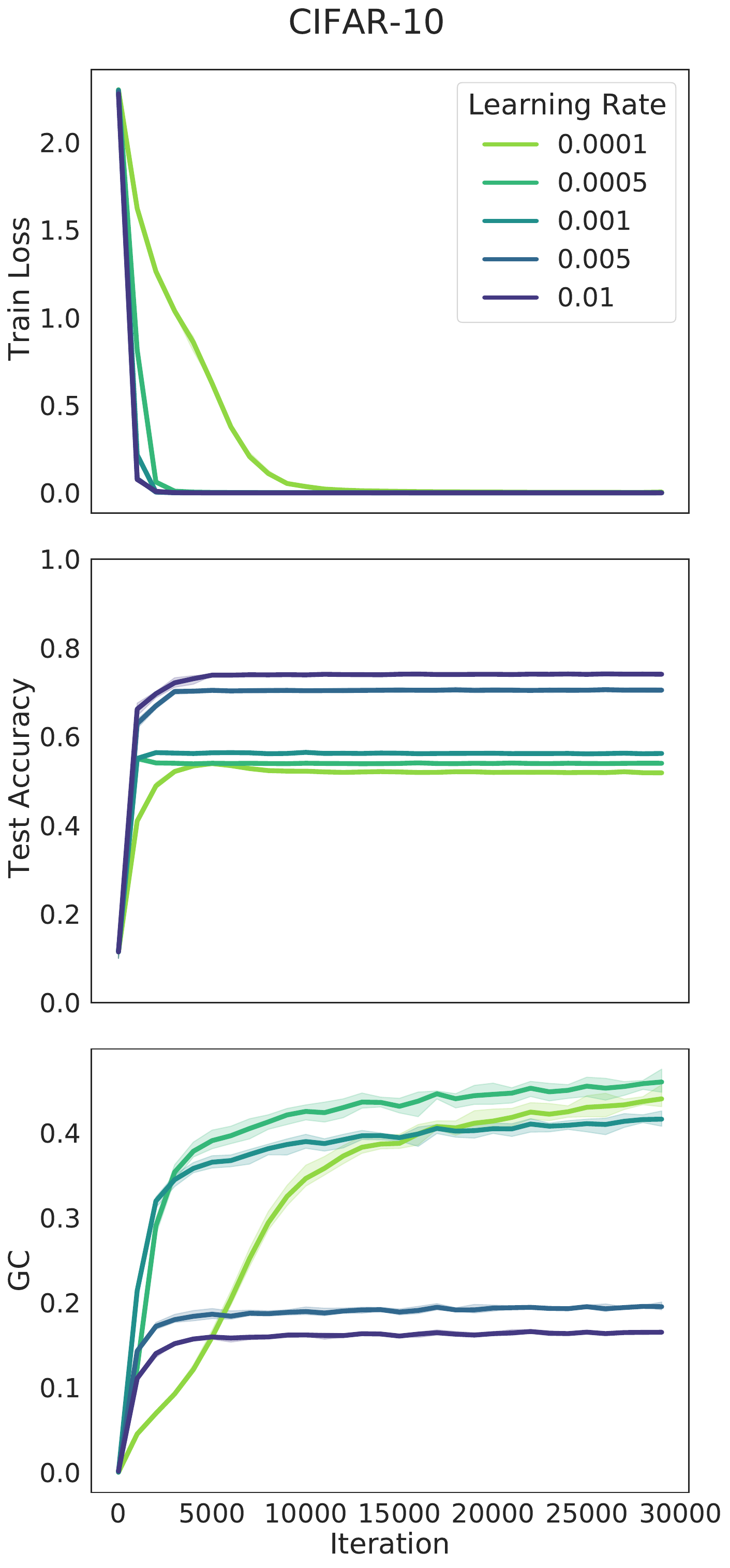}
  \label{fig:lr_gc_mom}
  }
  \end{subfloat}
  \begin{subfloat}[Batch size.]{
    \includegraphics[width=0.45\linewidth]{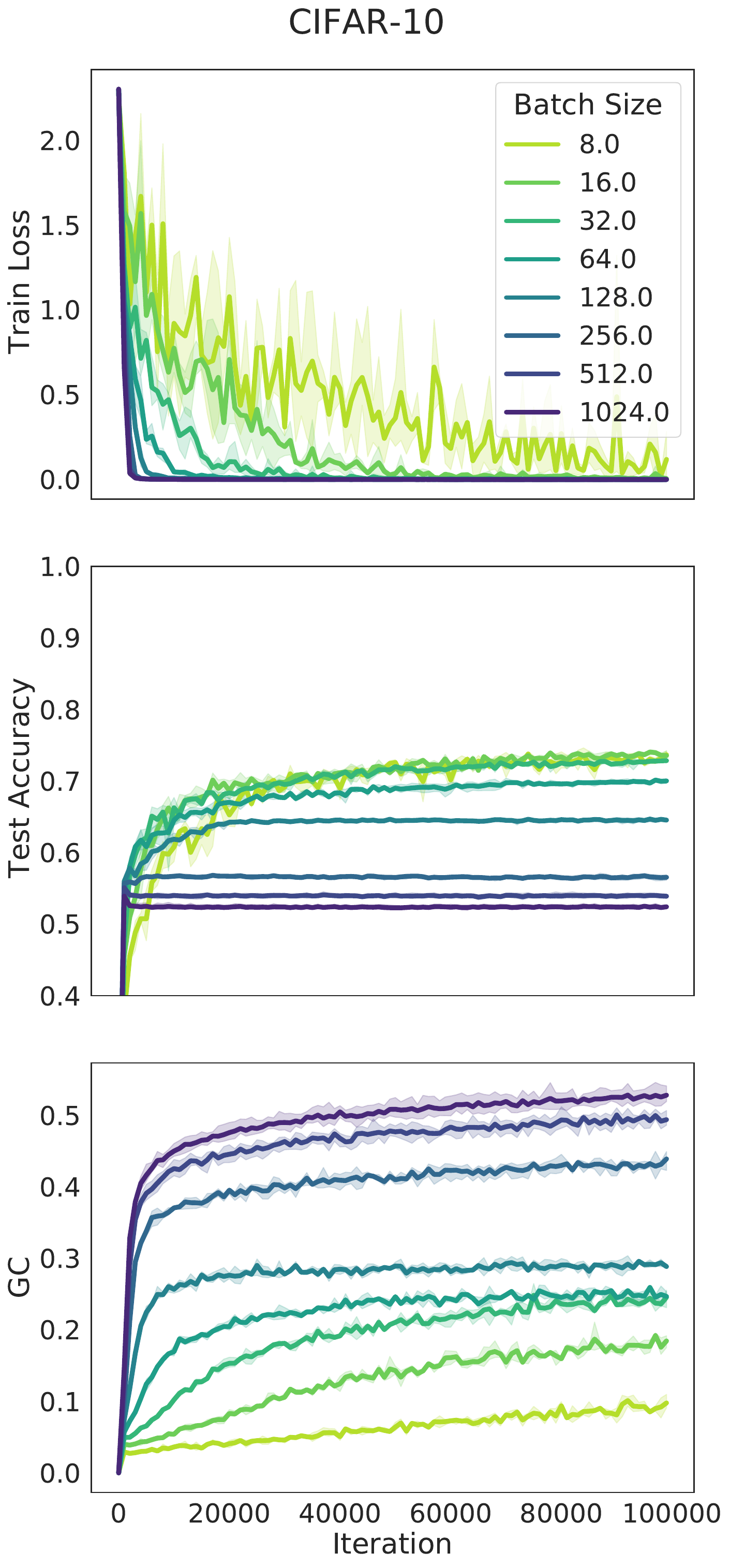}
  \label{fig:bs_gc_mom}
  }
  \end{subfloat}
  \caption[The implicit regularisation effect of increased learning rates and decreased batch sizes leads to decreased Geometric Complexity (GC) when momentum is used.]{\textbf{The implicit regularisation effect of increased learning rates \subref{fig:lr_gc_mom} and decreased batch sizes \subref{fig:bs_gc_mom} leads to decreased GC when momentum is used.} The network used is a Resnet-18, the momentum rate is $0.9$. For the learning rate experiments, the batch size used is 512. For the batch size experiments, the learning rate used is $0.005$. Each experiment uses 3 seeds. }
  \label{fig:gc_mom_implicit_regularisation}
\end{figure}

 \begin{figure}[t]
  \centering
  \begin{subfloat}[Vanilla SGD.]{
  \includegraphics[width=0.49\linewidth]{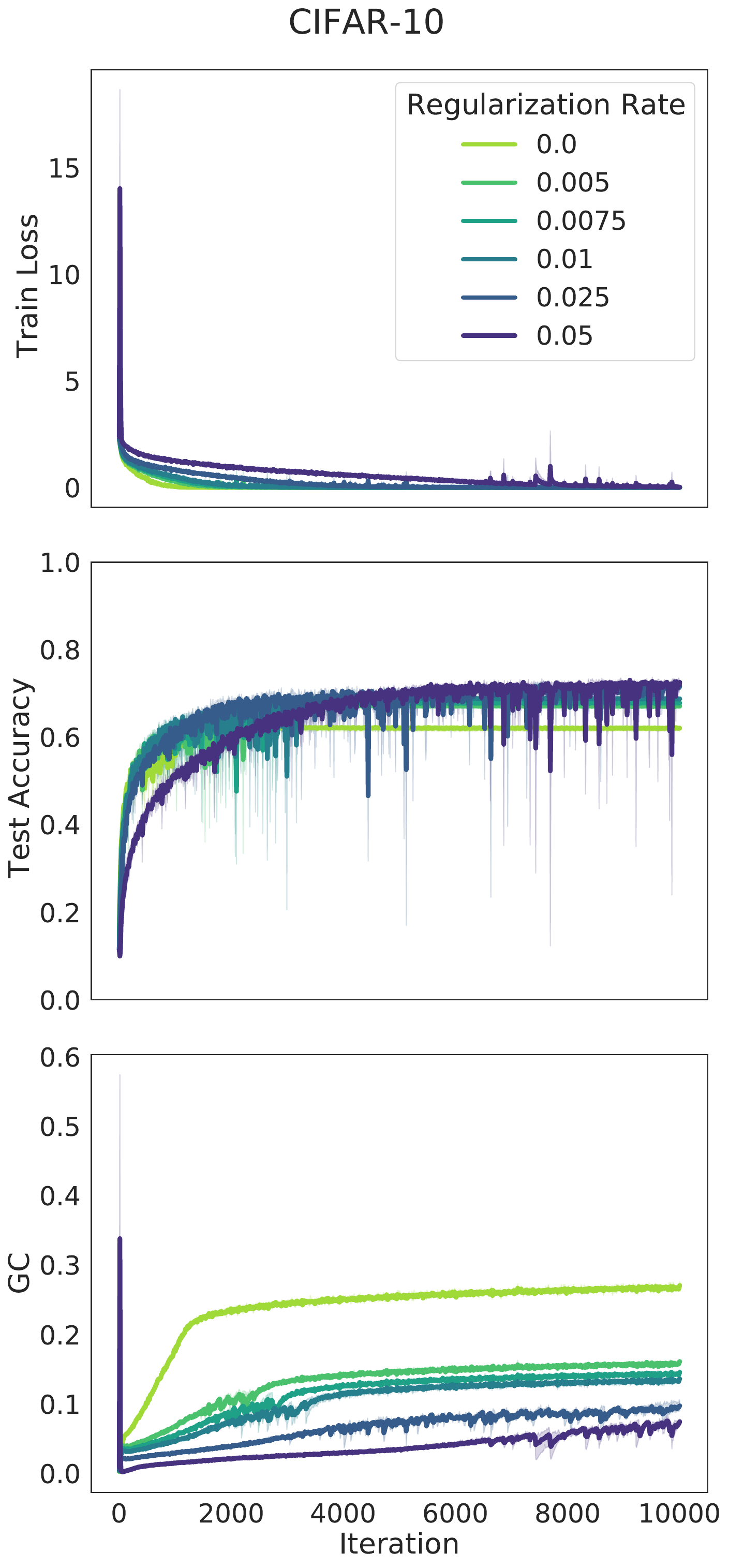}%
  \label{fig:gc_curves_gradient_reg}
  }
  \end{subfloat}%
 \begin{subfloat}[Momentum $0.9$.]{
  \includegraphics[width=0.49\linewidth]{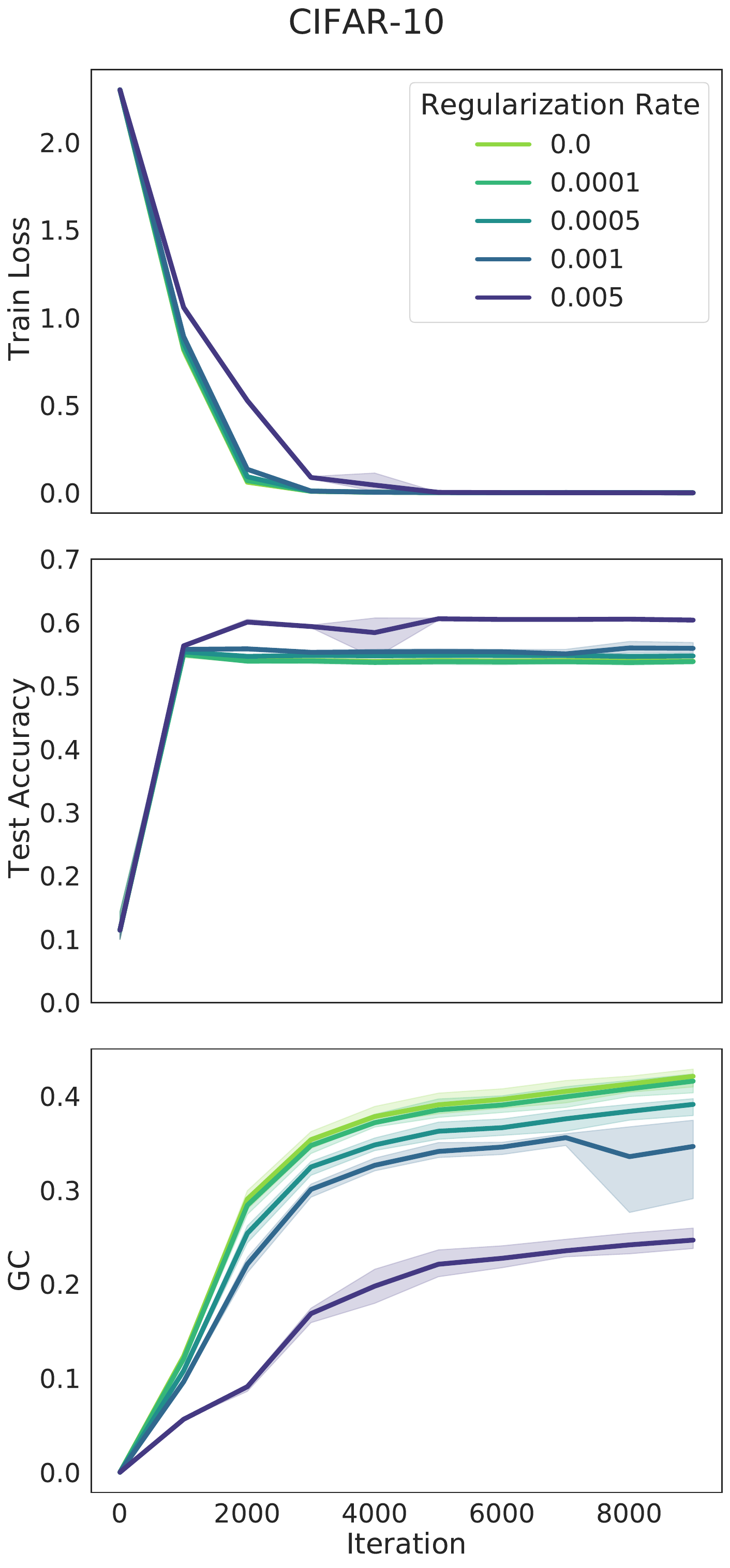}%
  \label{fig:gc_curves_gradient_mom}
  }
  \end{subfloat}%
  \caption[Gradient norm regularisation leads to decreased GC when using stochastic gradient descent.]{\textbf{Gradient norm regularisation leads to decreased Geometric Complexity when using stochastic gradient descent, with and without momentum.}. The network used is a Resnet-18, with a batch size of 512 and a learning rate of $0.005$ for momentum, and $0.02$ for vanilla gradient descent. Each experiment uses 3 seeds. }
  \label{fig:gradient_norm_gc_cifar_app}
\end{figure}

 \begin{figure}[t]
  \centering
  \begin{subfloat}[Vanilla SGD.]{
  \includegraphics[width=0.49\linewidth]{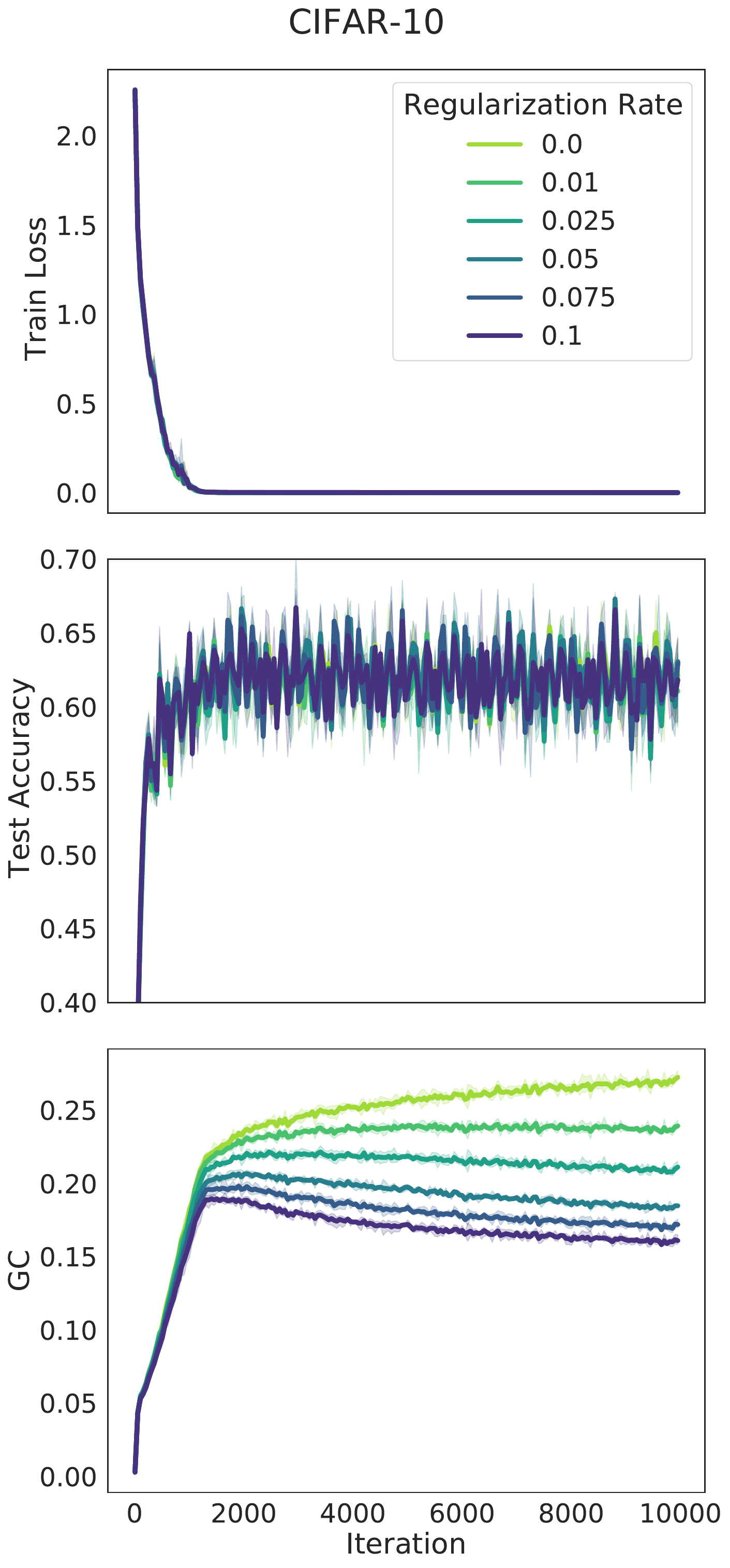}%
  \label{fig:gc_curves_spectral_reg}
  }
  \end{subfloat}%
 \begin{subfloat}[Momentum $0.9$.]{
  \includegraphics[width=0.49\linewidth]{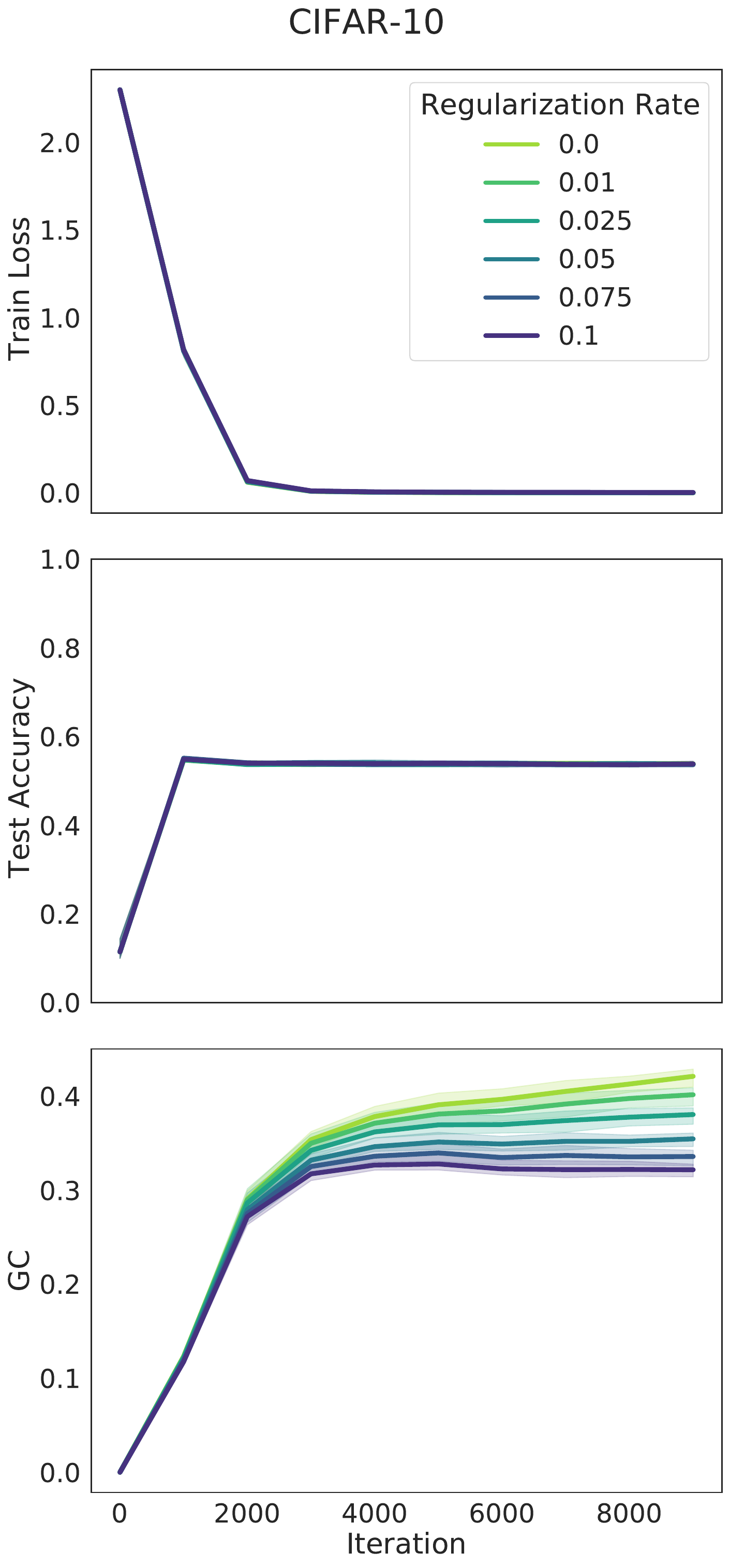}%
  \label{fig:gc_curves_spectral_mom}
  }
  \end{subfloat}%
  \caption[Spectral regularisation leads to decreased Geometric Complexity when using stochastic gradient descent.]{\textbf{Spectral regularisation leads to decreased GC when using stochastic gradient descent, with \subref{fig:gc_curves_spectral_mom} and without momentum \subref{fig:gc_curves_spectral_reg}.} The network used is a Resnet-18, with a batch size of 512 and  a learning rate of $0.005$ for momentum, and $0.02$ for vanilla gradient descent. Each experiment uses 3 seeds.}
  \label{fig:spectral_gc_cifar_app}
\end{figure}

\begin{figure}[t]
  \centering
  \begin{subfloat}[Vanilla SGD.]{
  \includegraphics[width=0.5\linewidth]{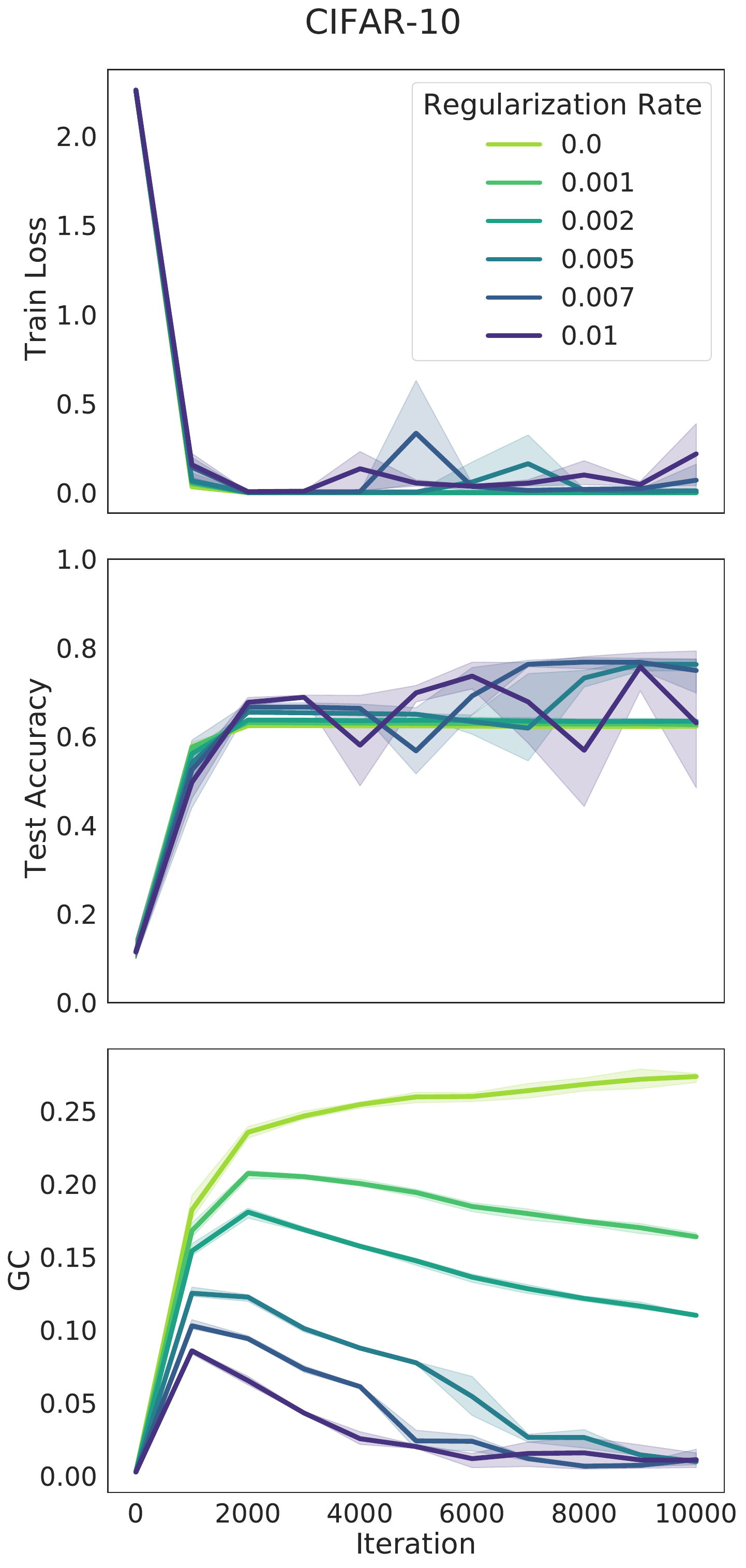}%
  \label{fig:gc_curves_l2_reg}
  }
  \end{subfloat}%
 \begin{subfloat}[Momentum $0.9$.]{
  \includegraphics[width=0.49\linewidth]{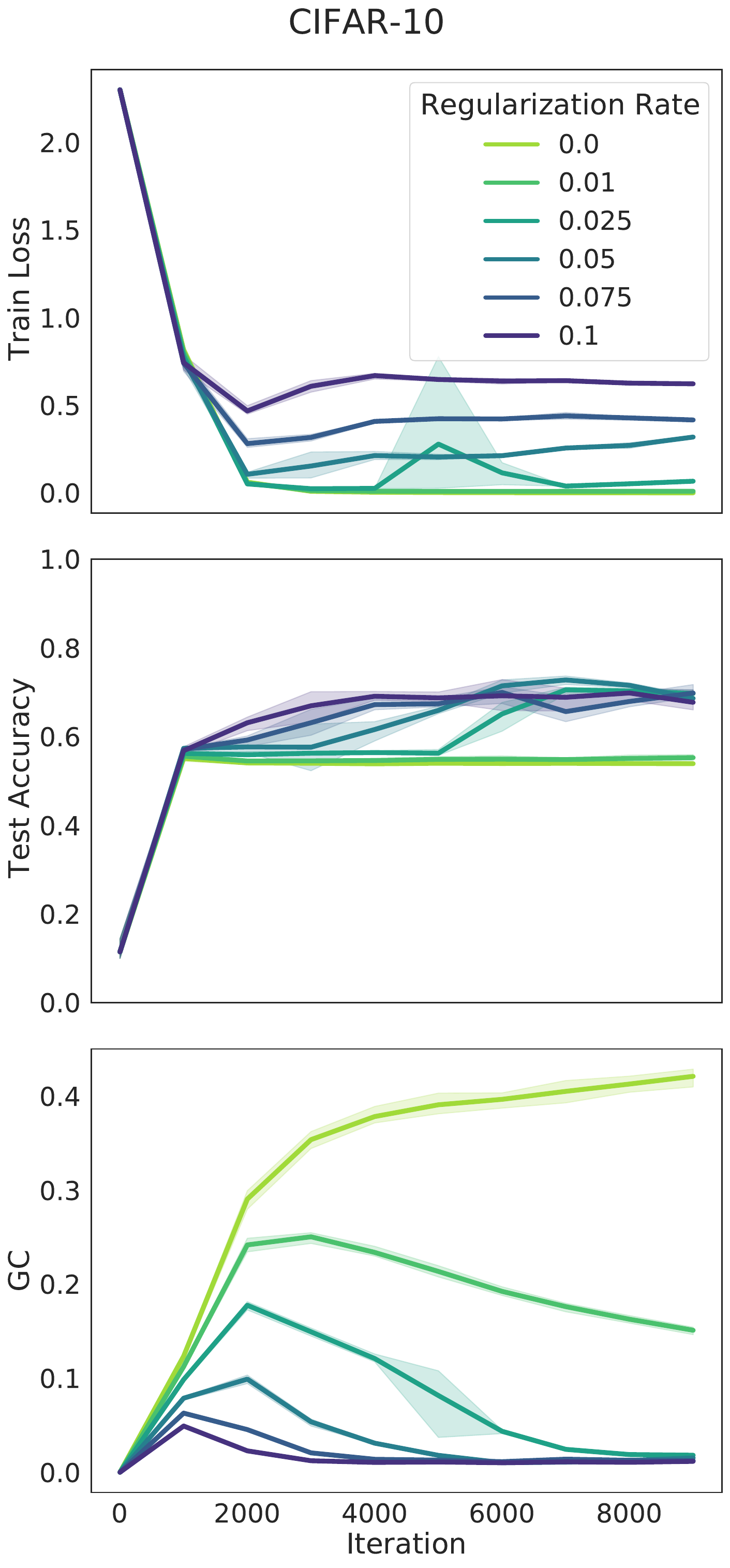}%
  \label{fig:gc_curves_l2_reg_mom}
  }
  \end{subfloat}%
  \caption[$L_2$ regularisation leads to decreased GC when using stochastic gradient descent.]{\textbf{$L_2$ regularisation leads to decreased Geometric Complexity when using stochastic gradient descent, with \subref{fig:gc_curves_l2_reg_mom} and without momentum \subref{fig:gc_curves_l2_reg}.} The network used is a Resnet-18, with a batch size of 512 and  a learning rate of $0.005$ for momentum, and $0.02$ for vanilla gradient descent. Each experiment uses 3 seeds.}
  \label{fig:l2_gc_cifar_app}
\end{figure}

\end{document}